%% file: main.tex
\documentclass[phd,ianc,oneside,logo,doublespacing,sansheadings,romanprepages,parskip]{infthesis}
\pdfoutput=1

\usepackage[utf8]{inputenc} 
\usepackage[T1]{fontenc}  
\usepackage{hyperref}    
\usepackage{url}      
\usepackage{booktabs}    
\usepackage{amsfonts}    
\usepackage{amsmath}
\usepackage{bm}
\newcommand\numberthis{\addtocounter{equation}{1}\tag{\theequation}}
\newcommand{\triangleq}{=\mathrel{\mathop:}}
\usepackage{nicefrac}    
\usepackage{microtype}   
\usepackage{lipsum}
\usepackage{graphicx}
\usepackage{float}
\usepackage[natbibapa]{apacite}
\usepackage{pdfpages}
\usepackage{epstopdf}
\usepackage[ruled,vlined]{algorithm2e}
\usepackage{graphicx}
\usepackage{subcaption}
\usepackage{float}

\newcommand{\KL}{D_\mathrm{KL}}
\newcommand{\E}{\mathbb{E}}

\title{Applications of the Free Energy Principle to Machine Learning and Neuroscience}
\author{Beren Millidge}

\abstract{%

In this thesis, we explore and apply methods inspired by the free energy principle to two important areas in machine learning and neuroscience. The free energy principle is a general mathematical theory of the necessary information-theoretic behaviours of systems which maintain a separation from their environment. A core postulate of the theory is that complex systems can be seen as performing variational Bayesian inference and minimizing an information-theoretic quantity called the variational free energy. The free energy principle originated in, and has been extremely influential in theoretical neuroscience, having spawned a number of neurophysiologically realistic process theories, and maintaining close links with Bayesian Brain viewpoints. 

The thesis is split into three main parts where we apply methods and insights from the free energy principle to understand questions first in \emph{perception}, then \emph{action}, and finally \emph{learning}. Specifically, in the first section, we focus on the theory of predictive coding, a neurobiologically plausible process theory derived from the free energy principle under certain assumptions, which argues that the primary function of the brain is to minimize prediction errors. We focus on scaling up predictive coding architectures and simulate large-scale predictive coding networks for perception on machine learning benchmarks; we investigate predictive coding's relationship to other classical filtering algorithms, and we demonstrate that many biologically implausible aspects of current models of predictive coding can be relaxed without unduly harming the performance of predictive coding models which allows for a potentially more literal translation of predictive coding theory into cortical microcircuits.
 
In the second part of the thesis, we focus on the application of methods deriving from the free energy principle to \emph{action}. We study the extension of methods of `active inference', a neurobiologically grounded account of action through variational message passing, to utilize deep artificial neural networks, allowing these methods to `scale up' to be competitive with state of the art deep reinforcement learning methods. Additionally, we show that these active inference inspired methods can bring conceptual clarity and novel perspectives to deep reinforcement learning. We show how active inference reveals the importance of deep generative models and model-based planning for adaptive action, as well as information-seeking exploration which arises under a unified mathematical framework from active inference. Finally, we provide a unified mathematically principled framework for understanding and deriving many information-seeking exploration objectives through the lens of a dichotomy between `evidence' and `divergence' objectives. We show that this distinction is crucial for understanding and relating the many exploratory objectives in both the reinforcement learning, active inference, and cognitive science communities and that this provides a general mathematical framework for specifying the objectives underlying intelligent, adaptive behaviour.

Finally, we focus on applications of the free energy principle to questions of \emph{learning} where we attempt to understand how credit assignment can take place in the brain. First, we demonstrate that, under certain conditions, the predictive coding algorithm can closely approximate the backpropagation of error algorithm along arbitrary computation graphs, which underlies the training of essentially all contemporary machine learning architectures, thus indicating a potential path to the direct implementation of machine learning algorithms in neural circuitry. Finally, we explore other algorithms for biologically plausible credit assignment in the brain, and present Activation Relaxation, a novel algorithm which can approximate backprop using only local learning rules which are substantially simpler than those necessary for predictive coding. We additionally show that the some relaxations that apply to predictive coding, also work for the activation relaxation algorithm, thus producing an extremely elegant and effective algorithm for local approximations to backprop in the brain.

In sum, we believe we have demonstrated the theoretical utility of the free energy principle, by demonstrating how methods inspired by it can interface productively with other fields, specifically neuroscience and machine learning, to develop and improve existing methods, as well as inspire novel advances, in all three areas of perception, action, and learning. Moreover, throughout this thesis, we demonstrate implicitly, the theoretical benefit brought about by the FEPs unified treatment of these seemingly disparate processes, under the rubric of free energy minimization.
}

\begin{document}

\begin{preliminary}


\maketitle

\begin{acknowledgements}
I would like to thank my supervisor, Richard Shillcock, for all his help and advice over the years; for his giving me freedom to work on the topics in this thesis even though they did not align with his planned research trajectory, and for his perseverance in handling my endless drafts. Secondly, a huge thanks to Christopher L Buckley, Alec Tschantz, and Anil Seth for hosting me for a year at the Sackler Centre and the Evolutionary and Adaptive Systems Group (EASY) at the University of Sussex. You have all taught me so much about the craft of research, and without your mentorship and guidance, I would not be half the researcher I am today. I would also like to thank Conor Heins for many stimulating conversations around the free energy principle and related topics. Finally, and above all, I would like to thank my wife, Mycah Banks, for her unending love and support throughout this entire process; her aid with proofreading and figure preparation for many papers, and her sacrifice in putting up with a husband who always has some more research to do. Without you, none of this would be possible. Thank you, Mycah.
\end{acknowledgements}

\standarddeclaration


\tableofcontents

 \listoffigures

\end{preliminary}


\include{chap1}

\include{chap2}
\include{chap3}
\include{chap4}
\include{chap5}
\include{chap6}
\include{chap7}

\appendix
\include{appendix_A}
\include{appendix_B}

\include{Appendix_C}



\bibliography{thesis}

\end{document}

%% file: chap1.tex
\chapter{Introduction}
The Free Energy Principle (FEP) \citep{friston2006free,friston2012free,friston2019particularphysics,parr2020Markov} is an emerging theory in theoretical neuroscience which aims to tackle an extremely deep and fundamental question -- can one characterise necessary behaviour of any system that maintains a statistical separation from its environment \citep{parr2020Markov,friston2019particularphysics,bruineberg2020emperor}? Specifically, it argues that any such system can be seen as performing an elemental kind of Bayesian inference where the dynamics of the internal states of such a system can be interpreted as minimizing a variational free energy functional \citep{beal2003variational} \footnote{Hence the name the ` free energy principle'}, and thus performing approximate (variational) Bayesian inference \citep{friston2019particularphysics}. The FEP is thus effectively a formalization and generalization of the Ashbyan good regulator principle \citep{conant1970every}, where an intrinsic property of these kinds of systems is that they in some sense come to embody a Bayesian model of their surroundings, and the perform a inference using this model \citep{baltieri2020predictions}.

The  free energy principle therefore provides a close link between the notions of self-organisation and dissipative structures in thermodynamics \citep{prigogine1973theory,seifert2008stochastic}, with cybernetic notions of feedback, regulation, and control \citep{wiener2019cybernetics,kalman1960contributions,johnson2005pid}, to more `cognitive' ideas of inference and learning \citep{schmidhuber1991possibility,dayan2008decision,rao1999predictive}. Specifically, we see that, in some sense, all of these notions can be construed as necessary properties and consequences of systems that self-organize to, and maintain themselves at, a non-equilibrium steady state. While having developed over time into a very general theory of self-organising systems, the free energy principle has emerged from theoretical neuroscience as a way to understand the properties of biological and cognitive systems, especially the brain \citep{friston2003learning,friston2006free}. As such, the most developed process theories, which are explicitly inspired by the  free energy principle -- predictive coding \citep{mumford1992computational,rao1999predictive,friston2005theory}, and active inference \citep{friston2012active,friston2015active,friston2017active}, each purport to be theories of inference, action, and learning in the brain. A core tenet of the  free energy principle is that perception, action, and learning can all be unified under a single inference objective which minimizes a single objective -- the variational free energy. As we shall see, these different `process theories' arise simply from the choice of a specific parametrization of the generative model and the variational density, where certain choices have been found to be useful and also give potentially biologically plausible inference and learning rules. In this thesis, we focus on the high level applications of the free energy principle to neuroscience and to machine learning, and make multiple contributions to the literature through the application of the free energy principle to all of perception, action, and learning. Specifically, in this thesis, we are primarily concerned with \emph{scaling up} methods which have emerged in the theoretical neuroscience literature, to the extremely challenging and complex tasks which can be solved with modern machine learning methods. The value of scaling up such methods is twofold. Firstly, FEP inspired process theories often possess significant biological plausibility, in that they provide a potential account of what the brain is doing -- thus, if they can scale them up to handle the sort of tasks that the brain must solve, then we can be more confident that these theories could, in theory, be actually implemented in the brain. Secondly, the free energy principle and its process theories contain many insights which can potentially be used to improve and extend current state of the art methods in machine learning. In this thesis, we aim to present both kinds of contributions -- firstly by demonstrating that FEP inspired models and process theories can scale, and secondly, by showcasing how ideas from the FEP can be used to advance the field of machine learning or neuroscience on its own terms.

\section{Thesis Overview}

This thesis is organized into three main parts. We begin with a detailed introduction and mathematical walkthrough of the Free Energy Principle (FEP) as presented in its most recent incarnation \citep{friston2019free,parr2020Markov} (Chapter 2) and then, in the first section of original work (Chapter 3), we consider applications of the free energy principle to \emph{perception} -- and make contributions to the FEP process theory of \emph{predictive coding}. In the second section (Chapters 4 and 5), we consider applications of the free energy principle to \emph{action}, and work with the process theory of \emph{active inference}. Finally, in the third section (Chapter 6), we consider applications of the free energy principle to \emph{learning}, and especially focus on what FEP-inspired models and process theories can tell us about the nature of credit assignment in the brain. Below is a chapter by chapter breakdown of the work in the thesis and what I see are the main contributions to both machine learning and neuroscience in each. Throughout the thesis, since each chapter covers a fairly distinct topic, I have tried to make each chapter modular and mostly independent of the others. Each chapter contains an introduction and mini literature review on the field it discusses, as well as presenting my original work.

In Chapter 2, I will give a detailed overview of the free energy principle, starting from first principles, and include a discussion of the mathematical assumptions and provide some of my opinions on the philosophical nature of the theory and its potential utility. I will then give a brief walk-through of discrete state space active inference as presented in \citep{friston2015active,friston2017process,da2020active} which will be the focus of the `scaling up' work in Chapter 4.

In Chapter 3, we deal principally with models of perception and predictive coding. We begin by giving a brief overview and mathematical walkthrough of predictive coding theory as it is presented in \citep{friston2005theory,friston2008hierarchical,buckley2017free}. We then cover in depth two contributions to the theory of predictive coding. 
First, we present work where we scale up and empirically test the performance of large scale predictive coding networks on machine learning datasets, which had not been tested before in the literature. We also clarify the relationship between predictive coding and other known algorithms such as Kalman filtering. Secondly, we discuss relaxing various relatively un-biologically plausible aspects of the predictive coding equations, such as the need for symmetric forward and backwards weights, the necessity of using nonlinear derivatives in the update rules, and the one-to-one error to value neuron connectivity required by the standard algorithms. All of these conditions put serious constraints on the biological plausibility of the algorithm, and here we show that to some extent they can each be relaxed without harming performance. 

Then, in Chapters 4 and 5, I present my work on the applications of the free energy principle to questions of action selection and control. Chapter 4 focuses predominantly on scaling up active inference methods to achieve results comparable to those achieved in the deep reinforcement learning literature, while Chapter 5 takes a more abstract and mathematical approach and investigates in depths the mathematical origin of objective functionals which combine both exploitatory and exploratory behaviour -- an approach which has the potential to finesse the exploration-exploitation dilemna \citep{friston2015active}.

Specifically, in Chapter 4, I will first review the rudiments of reinforcement learning (RL), and its current incarnation in deep reinforcement learning, including the two paradigms of model-free and model-based approaches. I will then present two of my contributions of merging active inference and deep reinforcement learning under the new paradigm of deep active inference. I will first discuss deep active inference in the model-free paradigm, and show how in this case the active inference equations can naturally be understood as specifying an actor-critic architecture with a bootstrapped value function, except one where the value-function becomes the expected-free energy functional, which provides an intrinsic source of exploratory drive to the algorithm which can improve performance. I then discuss the similarities and differences to standard deep reinforcement learning algorithms and empirically compare the performance of the algorithms on a number of challenging continuous control tasks from OpenAI gym \citep{brockman2016openai}. 

Then I will present a second piece of work which applies active inference instead to the model-based reinforcement learning paradigm. We will show that in this case, we can use the powerful generative `world models' \citep{ha_recurrent_2018} of active inference to work as transition models of the learnt dynamics, and then the use of action selection in planning. The use of the expected free energy functional again furnishes an intrinsic exploratory for active inference agents, which we again show is crucial to effective, goal-directed exploration and that it empirically improves performance on a suite of continuous control tasks. We then conceptualize reinforcement learning through the lens of inference, and understand the distinction between model-free and model-based reinforcement learning through the lens of iterative and amortised inference. We then demonstrate how these two types of inference can be \emph{combined}, leading to a novel hybrid inference algorithm which we show attains both the sample efficiency of model-based reinforcement learning with the higher asymptotic performance and fast computation time of model-free RL.

Then, in Chapter 5, we move into a more abstract, mathematical domain. Here we grapple with deep questions underlying the objective functions of reinforcement learning. Specifically, we wish to understand the mathematical origin and nature of the expected free energy term which grants deep active inference agents their superior exploratory capacities. Having tackled this, we turn to the deeper question of the mathematical origin of information-seeking exploratory terms within the inference objective optimized in reinforcement learning methods, and thus the mathematical origin of exploratory drives. We present a new dichotomy between \emph{evidence} and \emph{divergence} objectives, and demonstrate how only \emph{divergence} objectives, which intuitively can be seen as minimizing the divergence between the predicted and desired futures, rather than simply maximizing the \emph{likelihood} of the desired future, are required to obtain such terms. We then relate this fundamental dichotomy to a number of objectives prominent in both the cognitive science, neuroscience, and machine learning literatures. Finally, we further seek to explore the general possible space of variational objective functionals for control, and provide a wide-ranging categorisation of the potential of such functionals within our framework.


Finally, in Chapter 6, we turn to the application of insights and ideas from the free energy principle to \emph{learning}. Specifically, we focus on the vexing question of how to achieve credit assignment in the brain. This is necessary since, we assume, that most of the statistical `parameters' in the brain -- such as synaptic weights -- start out initialized fairly randomly during development and thus need to be trained or learned through interactions with the environment \footnote{It is possible that some pathways, especially low-level subcortical pathways may be, to some extent, hardwired by evolution. However, it is generally considered infeasible for the immense number and complexity of the neocortical circuitry to be hardwired in this way}. Understanding how this learning can take place is a fundamental question within neuroscience. One approach, which has recently been immensely successful in machine learning with large artificial neural networks, is the idea of learning through gradient descent using the backpropagation of error algorithm. Since backpropagation of error is such a successful algorithm in artificial neural network -- which, although simplified are nevertheless generally quite a close substrate to biological neural networks -- it is very likely that it would also work to successfully train biological neural networks, if it could be implemented in a biologically plausible manner in such networks. The question, then, becomes whether and how backprop can be implemented in biologically plausible neural networks. While this is an extremely broad question which cannot likely be answered in a single thesis, we present two novel contributions to this question here.

Specifically, in Chapter 6, we first provide a brief review of the credit assignment problem in the brain, as well as the backpropagation algorithm (and automatic differentiation in general), for context, and then present our two contributions to this field. First, we demonstrate how under certain conditions, predictive coding itself can be utilized as a biologically plausible method of credit assignment in the brain, can apply to any arbitrary computation graph, and can be used to train modern machine learning architectures such as CNNs and LSTMs with performance comparable to backprop. Secondly, we introduce a novel, simpler algorithm for credit assignment in the brain, which we call Activation Relaxation and then discuss their similarities and differences. We end with a discussion of the current state of credit assignment algorithms and backpropagation in the brain, and the importance of this field of research.

Finally, in Chapter 7, we provide a discussion and overview of the work in our thesis. We will briefly survey what has been achieved, and where the limitations and directions for future work lie, as well as the implications of the work of this thesis. 

\section{Statement of Contributions}

This statement provides a detailed overview of the work undertaken in this PhD which has resulted in research papers, both the papers included in this thesis and also those not included. I provide a brief summary of the key results and narrative of each paper, as well as a detailed breakdown my contributions. * denotes equal contribution.

\subsection{Included in Thesis}






\subsubsection{Chapter 3}

\begin{itemize}

\item \emph{Predictive Coding -- a Theoretical and Experimental review} (2021). \textbf{Beren Millidge}, Alexander Tschantz, Anil Seth, Christopher L Buckley. \emph{In Preparation}

This paper provides a full review of  recent advances in predictive coding, as well as the mathematical basis of the theory. It covers all of the mathematical, implementational, and neuronal aspects of predictive coding theory. As a first author paper, I conceptualised the idea, collated the necessary materials for the review, and wrote the paper. Alexander Tschantz, Christopher L Buckley, and Anil Seth, contributed edits and other editorial suggestions.

\item  \emph{Neural Kalman Filtering} (2021). \textbf{Beren Millidge}, Alexander Tscahtnz, Anil Seth, Christopher L Buckley. \emph{Arxiv}

This paper reviews the close connection between Kalman filtering and linear predictive coding, demonstrates that predictive coding can closely approximate the performance of Kalman filtering on filtering tasks, and proposes a low-level neural implementation of predictive coding which could be implemented in the brain. As a first author paper, I conceptualised the idea, worked out the mathematical derivations, implemented the model and experiments, and wrote a first draft of the paper. Alexander Tschantz, Christopher L Buckley, and Anil Seth contributed editorial and narrative suggestions.

\item  \emph{Relaxed Predictive Coding} (2020). \textbf{Beren Millidge}, Alexander Tschantz, Anil Seth, Christopher L Buckley. \emph{Arxiv}

This paper shows how several biologically implausible aspects of the predictive coding algorithm -- backwards weight symmetry, nonlinear derivatives, and one-to-one error unit connectivity -- can be relaxed without unduly harming performance on challenging object recognition tasks. As a first author paper, I conceptualised the idea, implemented the code and experiments, and wrote up a first draft of the paper. Alexander Tschantz, Anil Seth, and Christopher L Buckley contributed editorial suggestions.



\item \emph{Implementing Predictive Processing and Active Inference: Preliminary Steps and Results} (2019). \textbf{Beren Millidge}, Richard Shillcock. 

This paper provides reference implementations of multi-layer predictive coding networks trained for object recognition within a machine learning paradigm. As a first author paper, I conceptualised the idea, wrote the code and experiments, and wrote up the initial draft of the paper. Richard Shillcock contributed edits.

\end{itemize}

\subsubsection{Chapter 4}
\begin{itemize}
\item \emph{Deep Active Inference as Variational Policy Gradients} (2019). \textbf{Beren Millidge}. Published in the \emph{Journal of Mathematical Psychology}. 
\newline

This paper merges active inference and model-free deep reinforcement learning to create a deep active inference agent, very similar to actor-critic methods in deep reinforcement learning. The performance of deep active inference and deep reinforcement learning is compared on a suite of OpenAI Gym continuous control tasks. As a sole author paper, I conceptualised the idea, executed it mathematically and in code, designed and implemented the experiments and wrote up the paper. 

\item \emph{Reinforcement Learning through Active Inference} (2020). Alexander Tschantz*, \textbf{Beren Millidge}* , Anil Seth, Christopher L Buckley. Published in ICLR workshop on \emph{Bridging AI and Cognitive Science}. \newline

This paper applies active inference to model-based reinforcement learning methods for continuous control. We use an ensemble transition model parametrised by deep neural networks for model-based planning. The expected free energy and  free energy-of-the-expected future objective functionals provide additional exploratory bonuses which allow considerably greater and faster performance of the method compared to standard deep reinforcement learning baselines. I was joint first author on this paper. While the initial idea was primarily conceptualized by Alexander Tschantz and Christopher L Buckley, I contributed equally to the design and implementation of the algorithm, the experiments, and the writing of the paper.

\item \emph{Reinforcement Learning as Iterative and Amortised inference} (2020). \textbf{Beren Millidge}*, Alexander Tschantz*, Anil Seth, Christopher L Buckley. \emph{Arxiv}. \newline

This short workshop paper derives the key mathematical result of understanding model-free and model-based reinforcement learning in terms of iterative and amortised inference. It then partitions known reinforcement learning algorithms into a quadrant based on two orthogonal axes -- firstly whether it uses iterative or amortised reinforcement learning, and secondly whether we optimize over plans or over policies. As a joint first author, I contributed equally (with Alexander Tschantz) in the idea, formulation of the dichotomy, and the mathematical derivations. I also was the primary author of the text of the paper. Alexander Tschantz, Christopher L Buckley, and Anil Seth also contributed edits to the paper draft.

\item \emph{Control as Hybrid Inference} (2020). Alexander Tschantz, \textbf{Beren Millidge}, Anil Seth, Christopher L Buckley. Published in ICML workshop on the \emph{Theoretical Foundations of Reinforcement Learning}. \newline 

This paper combines both iterative (model-based) and amortised (model-free) reinforcement learning methods to obtain a hybrid method which combines both the sample-efficiency of model-based RL, with the asymptotic performance and fast computation of model-free methods. As second author, I contributed equally to the idea, the mathematical derivation, and architectural formulation of the hybrid agent. Alexander Tschantz took the lead with implementing the agent in code, designing and running, the experiments, and writing up the initial draft of the paper. I then contributed paper edits along with Anil Seth and Christopher L Buckley.
\end{itemize}

\subsubsection{Chapter 5}
\begin{itemize}

\item \emph{Whence the Expected Free Energy} (2020). \textbf{Beren Millidge}, Alexander Tschantz, Anil Seth, Christopher L Buckley. Published in \emph{Neural Computation}.

This paper investigates the mathematical origin of the expected free energy functional in active inference, demonstrates its relationship to other algorithm, and proposes a novel, more principled objective, the  free energy of the expected future (FEEF). As a first author, I primarily conceptualised the idea and worked out the mathematical results. I also wrote up the initial draft and was instrumental in handling later edits. Christopher L Buckley also contributed significantly to some of the mathematical results. Alexander Tschantz, Christopher L Buckley, and Anil Seth, also contributed through edits to the main text of the paper.

\item  \emph{On the relationship of Active Inference and Control as Inference} (2020). \textbf{Beren Millidge}, Alexander Tschantz, Anil Seth, Christopher L Buckley. Published in the \emph{IEEE International Workshop on Active Inference}.

This paper derives the relationship between active inference methods, and the control as inference paradigm which is popular within the reinforcement learning community. As first author, I conceptualised the idea, derived the primary mathematical results, and wrote the initial draft of the paper. Alexander Tschantz, Anil Seth,  and Christopher L Buckley contributed paper edits.

\item \emph{Understanding the Origin of Information-Seeking Exploration in Probabilistic Objectives for Control} (2021).\textbf{Beren Millidge}, Alexander Tschantz, Anil Seth, Christopher L Buckley. Submitted to \emph{Arxiv}. 

This paper introduces the dichotomy between evidence and divergence objectives, demonstrates how divergence objectives are necessary for the emergence of information maximizing exploration, and unifies many disparate objectives proposed in the machine learning and cognitive science communities under this formalism. As a first author paper, I conceptualised and derived the mathematical results, and wrote up a first draft of the paper. Alexander Tschantz, Christopher L Buckley, and Anil Seth contributed paper edits and narrative suggestions. Alexander Tschantz contributed heavily to the cognitive science and psychophysics sections.
\end{itemize}

\subsubsection{Chapter 6}

\begin{itemize}



\item \emph{Predictive Coding Approximates Backprop along Arbitrary Computation Graphs} (2020). \textbf{Beren Millidge}, Alexander Tschantz, Anil Seth, Christopher L Buckley. \emph{Arxiv}

This paper demonstrates that predictive coding can approximate the backpropagation of error algorithm along arbitrary computation graphs. Predictive coding is used to train state of the art machine learning architectures, and obtains identical performance to backprop even for deep and complex architectures. As a first author paper, I conceptualised the idea, implemented the models and experiments in code, and wrote up the initial draft of the paper. Alexander Tschantz, Christopher L Buckley, and Anil Seth contributed paper edits and narrative suggestions.

\item \emph{Activation Relaxation: A Local, Dynamical Approximation to Backpropagation in the Brain} (2020). \textbf{Beren Millidge}, Alexander Tschantz, Anil Seth, Christopher L Buckley. \emph{Arxiv}

This paper introduces a novel algorithm for approximating backpropagation in a local, biologically plausible way, which we call the activation relaxation algorithm. Crucially, this approach is significantly simpler than predictive coding, in that it does not require a special population of error neurons. Additionally, we show that several of the remaining biologically implausible aspects of the algorithm -- specifically the symmetric backwards weights -- can also be relaxed, leading to an extremely simple and biologically plausible algorithm for credit assignment. As a first author paper, I invented the algorithm and performend the mathematical derivations. I implemented the model in code, and ran the experiments. I wrote up the initial draft of the paper. Alexander Tschantz, Christopher L Buckley, and Anil Seth contributed editorial suggestions.

\item \emph{Investigating the Scalability and Biological Plausibility of the Activation Relaxation Algorithm} (2020). \textbf{Beren Millidge}, Alexander Tschantz, Anil Seth, Christopher L Buckley. Published at \emph{NeurIPS 2020 Workshop: Beyond Backprop}.

This paper extends and empirically tests the Activation Relaxation algorithm on more challenging tasks including large-scale CNN models. Moreover, it investigates the degree to which the loosening the assumptions of the activation relaxation algorithm hinder performance. As a first author paper, I conceptualized the core ideas to test, implemented the experiment and analyzed the results. I wrote up the initial draft of the paper. Alexander Tschantz, Christopher L Buckley, and Anil Seth contributed editorial suggestions.
\end{itemize}
\subsection{Not Included in the Thesis}
\begin{itemize}

\item \emph{Combining active inference and hierarchical predictive coding -- a tutorial review and case study} (2019). \textbf{Beren Millidge}, Richard Shillcock.

This paper uses hierarchical predictive coding networks as a dynamics model for a simple active inference approach which is then applied to discrete action reinforcement learning tasks such as the cart-pole. As a first author paper, I conceptualised the idea, implemented the code and experiments, and wrote up the initial draft. Richard Shillcock contributed paper edits and suggestions.

\item \emph{A predictive processing account of visual saliency using cross-predicting autoencoders} (2018). \textbf{Beren Millidge}, Richard Shillcock. \emph{Psyarxiv}.

This paper demonstrates how applying a cross-modal prediction objective in predictive coding, allows for the development of error representations which provide a good empirical match to estimates of visual saliency in natural image scenes. As a first author paper, I contributed equally in conceptualising the idea with my supervisor, Richard Shillcock. I implemented the models and experiments, and wrote up the initial draft of the paper. Richard Shillcock then contributed with editorial suggestions.

\item  \emph{Exploring infant vocal imitation in Tadarida brasiliensis mexicana} (2019). Richard Shillcock, \textbf{Beren Millidge}, Andrea Ravignani. Published in \emph{Neurobiology of Speech and Language}.

This paper introduces a multi-agent model of vocal imitation in bat infants, which provides a gradient soundscape which can guide mothers to pups in a crowded bat colony. As a second author paper, I contributed substantially to the development of the model. I implemented the model in code and ran the experiments. I contributed substantially to the writing of the paper.



\item \emph{Curious inferences: reply to Sun and Firestone on the Dark Room Problem} (2020). Anil Seth, \textbf{Beren Millidge}, Christopher L Buckley, Alexander Tschantz. Published in \emph{Trends in Cognitive Science}. 

This short response argues against the \emph{Dark Room Problem} in predictive coding by suggesting that the intrinsic exploratory drives of the expected free energy and other objectives such as the  free energy of the expected future suffice to drive the agent away from dark-room environments. I was involved in the conceptualisation and writing of the piece, although the main impetus behind this response lay with Anil Seth.

\item \emph{The Acquisition of Culturally Patterned Attention Styles under Active Inference} (2020). Axel Constant, Alexander Tschantz, \textbf{Beren Millidge}, Filipo Criado-Boado, Andy Clark. \emph{Arxiv}.

This paper presents an active inference model of culturally patterned saccade behaviour trained on archaeological vase patterns. It demonstrates that more complex patterns result in more vertically oriented saccade behaviour, thus corroborating experimental studies. I jointly designed and implemented the active inference model with Alexander Tschantz and ran the experiments. I also wrote the first draft of the methods section of the paper. Andy Clark, Felipe Criado-Boado, and Axel Constant conceptualised the idea and experiments.





\end{itemize}

%% file: chap2.tex
\chapter{The Free Energy Principle}

The free energy principle is a grand theory, arising out of theoretical neuroscience, with deep ambitions to provide a unified understanding of the nature of self organisation under the rubric of Bayesian inference \citep{friston2006free,friston_free_2019,friston2010free,friston2012free}. Perhaps the central postulate of this theory is the `Free Energy Lemma' which states that one can interpret any self organizing system, of any type and on any scale, as performing a kind of elemental Bayesian inference upon the external environment that surrounds it \citep{friston2013life,friston2012ao,friston2019particularphysics}. More generally than this, it claims to provide a recipe, in terms of a set of statistical independencies -- which we call the `Markov Blanket', following \citep{pearl2011Bayesian} -- which define precisely and mathematically what it means to be a system at all \citep{friston2019particularphysics}. Understanding self-organization through the lens of inference provides an exceptionally powerful perspective for understanding the nature of self-organizing systems, as it allows one to immediately grasp the nature of the dynamics which undergird self-organization, as well as apply the extremely large and powerful literature on Bayesian inference methods and algorithms to the dynamics of self-organizing systems \citep{parr2020modules,parr2020Markov,yedidia2011message}. 
Moreover, by framing everything in statistical terms -- in terms of conditional independencies, generative models, and approximate variational distributions -- the free energy principle provides a novel and powerful vocabulary to talk about such systems, as well as to ask questions such as `what kind of generative model does this system embody?' \citep{baltieri2020predictions, maturana2012autopoiesis} which would be impossible to ask and answer without it. Ultimately, this new statistical and inferential perspective upon dynamics may lead to important advances or novel insights.

This perspective also has exceptionally close relationships with early cybernetic views of control and regulation \citep{wiener2019cybernetics,conant1970every,kalman1960new}, and philosophically the FEP can be seen as a mathematical generalization of Ashby's notion that every good regulator of a system must become, in effect, a model of the system \citep{conant1970every}. The FEP nuances this notion slightly by instead stating that every system that regulates itself against the external environment, must in some sense embody a generative model of the environment, and also that the flow of the internal states of the system necessarily perform approximate variational inference upon an approximate posterior distribution over the external states of the environment, such that, broadly, they track the fluctuations in the external environment.

The free energy principle originated in theoretical neuroscience, as an attempt to understand the mathematical properties that a self-organising living, biotic system, \emph{must} possess in order to sustain itself against thermodynamic equilibrium. It was first and especially applied to understanding the function of the brain \citep{friston2006free,friston2010action,friston2012history}, and has been developed into two main process theories -- predictive coding \citep{rao1999predictive,friston2003learning,friston2005theory,friston2008hierarchical} and active inference \citep{friston2009reinforcement,friston2012active,friston2015active,friston2017process,friston2018deep,da2020active} which have been investigated in a wide variety of paradigms, where it has been used to investigate a wide variety of phenomena from \citep{friston2014anatomy,friston2015knowing,friston2015active}, information foraging and saccades \citep{parr2017uncertainty,parr2018active,parr2019computational} exploratory behaviour \citep{schwartenbeck2013exploration,friston2015active,friston2017curiosity,friston2020sophisticated}, concept learning \citep{schwartenbeck_computational_2019}, and a variety of neuropsychiatric disorders \citep{lawson2014aberrant,adams2012smooth,mirza2019impulsivity,cullen2018active}. These process theories translate the abstract formulation of the FEP into concrete and practical algorithms by specifying certain generative models, variational distributions, and inference procedures, and have been shown to be extremely useful both in providing powerful and biologically plausible theories of learning and inference in the brain, and also in developing highly effective inference algorithms which have advanced the state of the art in machine learning \citep{parr2019neuronal,millidge_deep_2019,tschantz2020reinforcement,millidge2020relationship}. 

In this chapter, we will provide a relatively self-contained step through of the key mathematical results of the most recent incarnation of the free energy principle as presented in \citep{friston2019particularphysics, parr2020Markov}, as well as the details of the discrete-state-space active inference process theory \citep{friston2015active,da2020active}. While none of the material in this chapter is original, it is necessary (especially the material on active inference), to understand what is to come in later chapters. Since this thesis covers a fairly wide range of topics, each individual thesis chapter also comes with its own literature review covering the necessary background for the original material in that chapter.

It is important to note that the material in the first section of this chapter (walkthrough of the Free Energy principle as described in \citet{friston2019particularphysics}) is not original to me and draws heavily from an unpublished monograph \citep{friston2019particularphysics}, albeit a monograph which has widely been viewed as the canonical reference point for the theory. Additionally, many core elements of the theory presented here have been published elsewhere \citep{friston2013life, friston2020some,parr2020Markov,friston2020sentience}. Notably, a fair amount of the material covered here is also controversial within the community and the validity of many required assumptions still remains to be assessed. Where relevant, in this section, we provide additional disclaimers highlighting core assumptions and potentially problematic elements of the mathematical exposition -- and additionally in the discussion section we include an itemized list of all assumptions as well as critical discussions on each. While some of this material is somewhat extraneous to the original work covered in later chapters, we believe that this presentation of the free energy principle gives the reader valuable context into the broader paradigm of the FEP which has inspired much of the original work in this thesis. Additionally, by condensing the logical flow of the FEP, and providing a detailed critical discussion of the logic and assumptions required, we aim to provide a broader service to the community by helping to make clear the current state as well as current controversies and debates at the cutting-edge of the free energy community.

Finally, it is important to note that the process theories derived from the FEP -- which we will primarily focus on in the rest of the thesis -- do not strictly require the full mathematical structure of the FEP to hold for their validity. As scientific theories about the real world, they rise or fall on empirical considerations independently of the overall mathematical construct of the FEP, and thus while the material in this chapter is useful contextually, it is not necessary to understand the work in the chapters that follow. Throughout we have aimed to make sure that each chapter, by and by large, is `modular', so that they can be read and understood in isolation. As such, we have striven to ensure that each chapter contains sufficient background information within it to let it be understood and evaluated independently of the others.

\section{History and Logical Structure}
\begin{figure}
  \centering
  \includegraphics[scale=0.2]{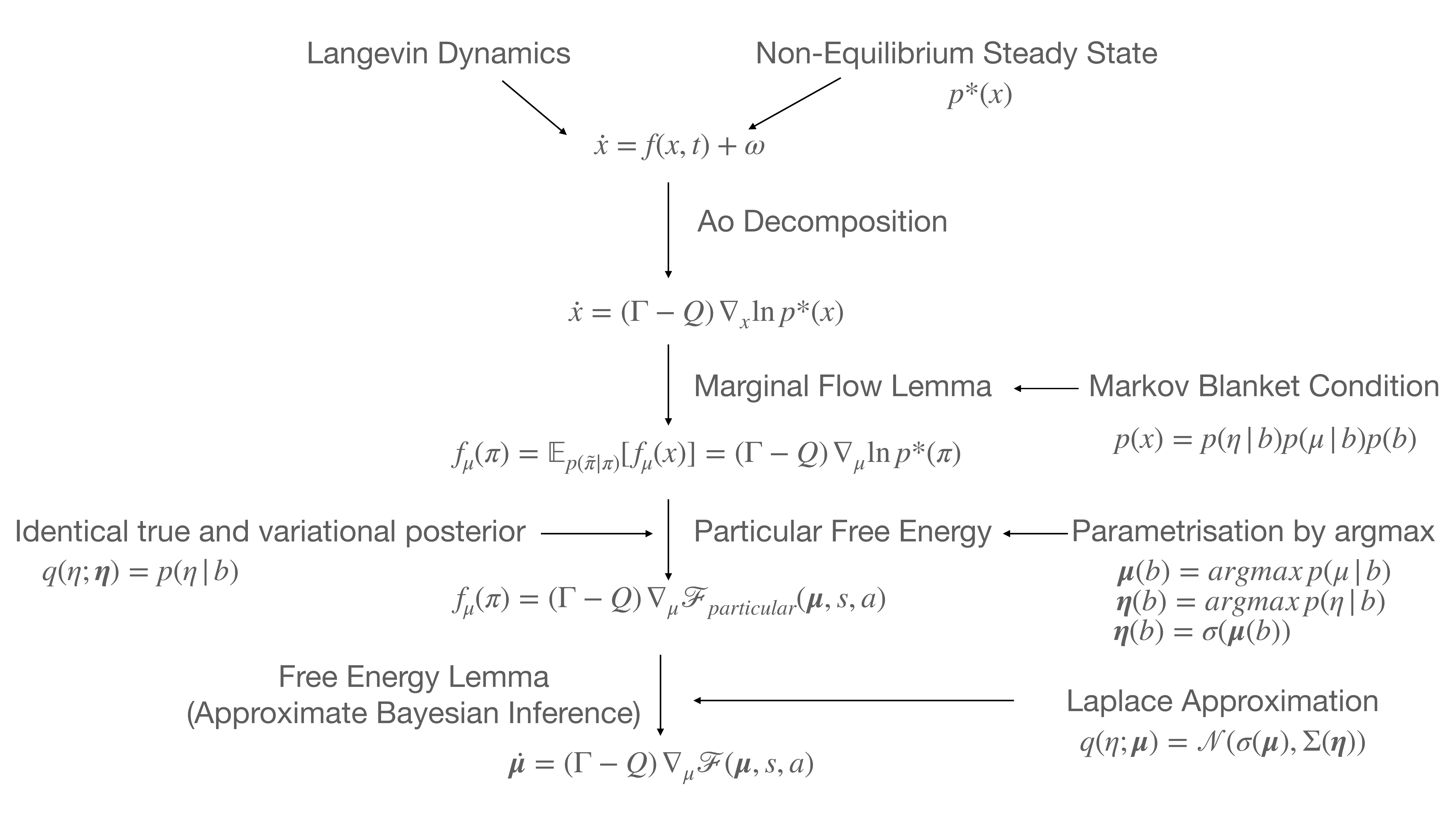}
  \caption{The logical flow of the argument of the FEP from the initial formulation to the crucial approximate Bayesian inference lemma. We begin with a setting of random Langevin stochastic dynamical systems, which possess a non-equilibrium-steady state. By applying the Ao decomposition, we can understand the dynamics in terms of a gradient descent upon the surprisal. Upon the addition of a Markov Blanket partition, we can express subsets in terms of their own marginal flows via the marginal flow lemma. If we then identify the internal states as parametrizing a variational distribution over the external states, we can interpret the marginal flow on the surprisal as a flow on the variational free energy, under the Laplace approximation.}
\end{figure}
Historically, the free energy principle has evolved over the course of about fifteen years. Its intellectual development can best be seen in two phases. In the first phase, an intuitive and heuristic treatment emerged with \citet{friston2006free} which stated that the imperative to minimize variational free energy emerged from a necessary imperative of minimizing the system's entropy, or log model evidence, which is upper bounded by variational free energy. This imperative emerges due to the self-sustaining nature of biological systems such as brains, in that they maintain a set distribution against the inexorably increasing entropic nature of thermodynamic reality \citep{friston2009free}. In order to do so, systems must constantly seek to reduce and maintain their entropy across their state space. Since the VFE is computationally tractable while the entropy itself is not, it was postulated that neural systems maintain themselves by implicitly minimizing this proxy rather than the actual entropy itself \citep{friston2010free}.

Later, in the second phase \citep{friston2013life}, this heuristic argument and intuition was related more formally to concepts in stochastic thermodynamics \citep{friston2012ao,friston2012free}. Specifically, the framework developed mathematically into a description of stochastic dynamics (as stochastic differential equations) separated into `external, internal, and blanket' states by a statistical construct called a \emph{Markov Blanket}. This blanket makes precise the statistical independence conditions required to make sense of talking about a `system' as distinct from its `environment'. Moreover, by separating the `blanket' into `sensory' and `active' states, one can obtain a statistical description of the core elements of a perception-action-loop, a central concept in cybernetics, control theory, and reinforcement learning. Secondly, the theory developed a precise notion of what it means to maintain a stable `phenotype' which is interpreted mathematically as a non-equilibrium steady-state density (NESS) over the state-space. This steady state is non-equilibrium due to the presence of `solenoidal flows' which are flows orthogonal to the gradient of the NESS density. Mathematically, such flows do not increase or decrease the entropy of the steady-state-density, but do, however, in contrast to an equilibrium steady state (ESS), provide a clear arrow of time. Given this, it is claimed, that under certain conditions, one can draw a relationship between the flow dynamics and the process of variational Bayesian inference through the minimization of the variational free energy (VFE)-- which measures the discrepancy between an approximate posterior and generative model -- and that the dynamics that result from this specific kind of flow under a Markov blanket at the NESS density can be seen as approximating a gradient descent upon the VFE, thus licensing the interpretation of the system as performing a basic kind of Bayesian inference or, `self-evidencing' T \citep{hohwy2008predictive,clark2015surfing}

While the intuitions and basic logical structure of the theory has remained roughly constant since \citet{friston2013life,friston2012free}, the mathematical formulation and some of the arguments have been refined in the most recent \citet{friston2019particularphysics} monograph and related papers \citep{friston2020some,parr2020Markov}. These papers have drawn close connections between the formulation of free energy principle, and many aspects of physics including the principle of least action in classical mechanics, and notions of information length and the arrow of time in stochastic thermodynamics. Additionally, the Particular Physics monograph \citep{friston2019particularphysics} contains a novel information-geometric gloss on the nature of the Bayesian inference occurring in the system. Specifically, it argues that the internal states of the system can be seen as points on an \emph{statistical manifold} that parametrize distributions over the external states, and that thus the internal states can be described using a `dual-aspect information geometry.' According to this perspective, internal states evolve in both the `intrinsic' state space of the system's physical dynamics, while simultaneously parameterising a manifold of statistical beliefs about external states - the so-called `extrinsic' information geometry. 

While the mathematical depths of the FEP often appears formidably complex to the uninitiated, the actual logical structure of the theory is relatively straightforward. First, we want to define what it means to be `a system' that keeps itself apart from the outside `environment' over a period of time. The FEP answers this question precisely its own way. We define a `system' as a dynamical system which has a non-equilibrium steady state (NESS) which it maintains over an appreciable length of time, and that the dynamics are structured in such a way that they obey the `Markov Blanket Condition'. Specifically, having a NESS can be intuitively thought of as defining dynamics which produce something like a system -- i.e. a recognizable pattern of states which persists relatively unchanged for some period of time. For instance, we can think of the biological systems in such a manner. Biological organisms maintain relatively steady states, against constant entropic dissipation, for relatively long (by thermodynamic standards) periods of time. Of course, from a purely thermodynamical perspective, in resisting entropy themselves, biological organisms are not countering the law of thermodynamics. To achieve their steady state requires a constant influx of energy -- hence it is a non-equilibrium steady state (NESS). From this perspective, we can understand biological organisation to be the process of creating `dissipative structures' \citep{prigogine1973theory, kondepudi2014modern} which only manage to maintain themselves at steady state and reduce their own entropy at the expense of consuming energy and increasing the entropy production rate of their environment \citep{prigogine2017non}. Illustrative physical examples of similar NESS states are Benard convection cells , and the Belousov-Zhabotinsky reaction \citep{zwanzig2001nonequilibrium}. In practical terms, we can consider the NESS density to be the `phenotype' of the system. From the perspective of the FEP, we are not usually concerned with whether a set of dynamics possesses a NESS density, or how convergence \emph{to} the NESS density works, instead we take it as an axiom that we possess a system with a NESS density, and are instead concerned with the dynamical behaviour of the system \emph{at} the NESS density. While this is clearly a special case, nevertheless dynamical systems at NESS already exhibit rich behaviours to effectively maintain themselves there, and it is these properties which necessarily any system which maintains itself at NESS, which are the fundamental object of study of the FEP. 

Secondly, now that we have a system which has a NESS density, and thus exhibits some stability through time, we also require a statistical way to separate the `system' from the `environment'. The FEP handles this by stipulating that any system it considers must fulfil a set of criteria which we call the Markov Blanket conditions. These conditions, deriving from the idea of Markov blankets in Bayesian networks \citep{pearl2011Bayesian,pearl2014probabilistic}, set forth a set of conditional independence requirements that allow a system to be statistically separated from its environment \footnote{Whenever we say Markov Blanket, following standard use in the literature, we mean the \emph{minimal} Markov blanket -- i.e. the Markov Blanket which requires the fewest number of blanket states to achieve the required conditional independencies.}. Specifically, we require that the dynamics of the system can be partitioned into three sets of states -- `internal' states which belong to the system of study, `external' states which correspond to the environment, and `blanket states' which correspond to the boundary between the system and its environment. Specifically, we require the internal states to be conditionally independent of the external states given the blanket states, and vice versa. Thus all `influence' of the environment must travel through the blanket, and cannot directly interact with the internal states of the system which are `shielded' behind the blanket \footnote{Interestingly, mathematically, the MB condition and all of the FEP is completely symmetrical between `internal' and `external' states. Thus from the perspective of the system, the `external states' are its environment, but from the perspective of the environment, the `external states' are the system. This means that the environment models and performs inference about the system just as the system models and performs inference on the environment. We can thus think of the environment-system interaction as a duality of inference, where each tries to model and infer the other in a loop.}

Now that we have a system with a NESS density which obeys the Markov Blanket conditions, so that we can partition it into external, internal, and blanket states, we then wish to understand the dynamics of the system \emph{at} the NESS density, so we can understand the necessary behaviours of the system to allow the NESS to be maintained. The derivation of the FEP then uses the Helmholtz (Ao) decomposition \citep{yuan2017sde,yuan2011potential,yuan2012beyond} to represent the dynamics as a gradient flow on the log of the NESS density (which is called the surprisal) with both dissipative (in the direction of the gradient) and solenoidal (orthogonal to the gradient) components. Now that we can express the flows of the system in terms of gradients of the log NESS density, we then invoke the \emph{Marginal Flow Lemma} to write out the dynamics of each component of the partitioned dynamics (i.e. external, internal, and blanket states) solely in terms of a gradient flow on its own marginal NESS density. This means that we can express, for instance, the dynamics of the internal states solely in terms of gradient flows on the marginal NESS density over the internal and blanket states.

Given this marginal partition, we can analyze and understand each of the flows in each partition of the system independently. Specifically, to understand the Ashbyan notion that `every good regulator of a system is a model of the system' , we wish to understand the relationship between the flows of the internal and external states, which are statistically separated from the blanket. Despite this separation, it is possible to define a mapping between the most likely internal state, given a specific configuration of the blanket states, and the distribution over the most likely external state of the system. We can use this mapping to interpret internal states as parametrizing a variational or approximate distribution \emph{over} the external states. This interpretation sets up the `dual-aspect' information geometry of the internal states, since internal state changes simultaneously represent changes in parameters of the distribution over internal states (which can potentially be non-parametric), and changes to the parameters of the variational distribution over external states. This latter interpretation means that the internal states parameterise a statistical manifold equipped with a Fisher information metric (if the variational distribution is in the exponential family), and in general becomes amenable to the techniques of information geometry \citep{amari1995information,ollivier2017information} Finally, given that we can interpret the internal states as paramterising a \emph{distribution} over external states, we can reconsider the gradient flow upon the log NESS density with a new light. Specifically, we can understand the marginal NESS density to represent the implicit \emph{generative model} of the system, and the gradient flow dynamics as a descent upon the free energy, with a perfect Bayes-optimal posterior. Alternatively, if we invoke an approximate posterior distribution over external states which is parametrized by the internal states, we can represent the gradient flow of the internal states as performing an approximate minimization of the variational free energy (VFE), and thus the internal states of the system can be interpreted as performing approximate variational Bayes. This is the \emph{key result} of the FEP. It states, simply, that the necessary dynamics of any system that maintains itself at a non-equilibrium steady state, and possesses a Markov Blanket, can be interpreted as modelling, and performing approximate variational inference upon the external states beyond its own Markov Blanket. It thus generalizes and makes precise Ashby's notion that every good regulator must in some sense be a model of the system \citep{conant1970every}. Here we see that in order to maintain a non-equilibrium steady state, to counteract the dissipative forces inherent in thermodynamics, it is necessary to perform some kind of inference about the environment beyond the system itself. 

\section{Formulation}

Here we begin the precise mathematical description of the FEP. We aim to provide a consistent notation, and more detailed derivations of key results than are often presented. The presentation in this chapter mostly follows the order of presentation in \citet{friston2019particularphysics}, although many circumstantial topics are omitted to focus on the main flow of the argument. We begin with the basic mathematical setting and formulation of the theory. We assume that the dynamics we wish to describe can be expressed in terms of a Langevin stochastic differential equation \citep{jaswinskistochastic},
\begin{align*}
\label{FEP_dynamics}
\frac{dx}{dt} = f(x) + \omega \numberthis
\end{align*}
where $x = [x_0 \dots x_N]$ is a vector of states of some dimensionality, and f(x) is an arbitrary nonlinear but differentiable function of the states. Expressing the dynamics in terms of a Langevin stochastic differential equation is a very flexible parametrization of the dynamics, and is the standard form studied in the field of stochastic differential equations, thus allowing the immediate use of results from that field. Specifically, here we assume already that this process is not history dependent. The dynamics only depend on the instantaneous values of the states. In practice, history dependent systems can be represented in this fashion, albeit somewhat unintuitively by adding sufficient statistics of the history to the state itself. $\omega$ is assumed to be white (zero autocorrelation) Gaussian noise with zero mean such that $\omega$ = $\mathcal{N}(x; 0, 2\Gamma)$ where $\Gamma$ is the half the variance of the noise. Zero autocorrelation means that the covariance between the noise at any two time instants, even infintesimally close together, is 0 -- $\E[\omega_t \omega_{t+\delta}^T] = 0$. We assume that this noise is added additively to the dynamics.

This stochastic differential equation can also be represented not in terms of dynamically changing states, but in terms of a dynamically changing \emph{probability distribution} over states. This transformation is achieved through the Fokker-Planck equation, by which we can derive that the change in the distribution over states can be written as,
\begin{align*}
\label{FP_equation}
\frac{d p(x,t)}{dt} = - \nabla_x f(x,t)p(x,t) + \nabla_x \Gamma \nabla_x p(x,t) \numberthis
\end{align*}

Where $p(x,t)$ is the instantaneous distribution over the states at a given time $t$. 
Here $\nabla_x f(x,t)$ is the gradient function and simply denotes the vector of partial derivatives of the function $f$ with respect to each element of the vector $x$. $\nabla_x f(x) = [ \frac{\partial f(x,t)}{\partial x_0}, \frac{\partial f(x,t)}{\partial x_N}, \dots , \frac{\partial f(x,t)}{\partial x_N}]$. $\nabla^2_x f(x)$ represents the matrix of second partial derivatives of the function.

Next, we presuppose that the dynamics expressed in Equation \ref{FEP_dynamics} tend towards a non-equilibrium steady state $\lim_{t \to \infty} p(x,t) \to p^*(x)$ where we represent the steady state distribution as $p^*(x)$. Note that this distribution no longer depends on time, since it is by definition at a steady state. We use $p^*$ to make clear that this distribution is at steady state. By definition a steady state distribution does not change with time, so that $\frac{dp^*(x)}{dt} = 0$.

The distinction between an equilibrium steady state and a non-equilibrium steady state (NESS) distribution is subtle and important. An equilibrium steady state, mathematically, is one where the property of detailed balance holds. This means that any transition between states at equilibrium is just as likely to go in the `forwards' direction as it is to go in the `backwards' direction. In effect, the dynamics are completely symmetric to time, and thus there is no notion of an arrow of time in such systems. Conversely, a non-equilibrium steady state is one where detailed balance does not hold, so there is a directionality to the dynamics, and thus an arrow of time, even though the actual distribution over states remains constant. From a thermodynamic perspective, the equilibrium-steady-state is the inexorable endpoint of the second law of thermodynamics, since it is the maximum entropy state. Conversely, a NESS is not a maximum entropy solution, since the directionality of the dynamics means that there is a degree of predictability in the system which could in theory be exploited to produce work. Non-equilibrium steady states can arise in thermodynamic systems but require an external source of driving energy as a constant input to the system, which is then dissipated to the external surroundings and gives the NESS a positive entropy production rate. To take an intuitive example, we can think about the thermodynamic equilibrium of a cup of coffee with cream added. The equilibrium steady state (ESS) is when the coffee and cream have completely diffused into one another, so that the cream maintains a constant proportion throughout the entire coffee cup. This will be the inevitable result (by the second law of thermodynamics) of adding an initially low entropy highly concentrated cream scoop into the coffee. On the other hand, we can think of the non-equilibrium steady state (NESS) as to be when the cream and coffee are equally diffused throughout, but somebody \footnote{Of course the analogy fails here since this represents a system with external driving (the person stirring) whereas the true NESS has no external driving and as such is just `intrinsically being stirred' with no stirrer.} is constantly stirring the coffee in a specific direction. Here, we are at steady state because the concentrations of cream and coffee don't change over time, but nevertheless there is a directionality to the dynamics in the direction of the stirring. This directionality is only maintained due to a constant input of energy \footnote{It's important to note that here we are using physical intuition and concepts like `energy' in a purely metaphorical sense. All results here apply to arbitrary SDEs which do not necessarily follow the same constraints as physical systems -- i.e. respect conservation of energy} to the system (the stirring) \footnote{Interestingly, physical experience with this analogy would suggest that the solenoidal dynamics leading to NESS would lead to faster convergence to the NESS density compared to the strictly dissipative dynamics leading to ESS -- effectively, stirring helps the cream diffuse faster. This insight has been applied to the design of highly efficient Markov-Chain-Monte-Carlo samplers in machine learning \citep{metropolis1953equation,neal2011MCMC,betancourt2013generalizing}}. The flow caused by the stirring is referred to as the `solenoidal flow' and mathematically is necessarily orthogonal to the gradient of the steady state distribution. This is necessary so that the solenoidal flow does not ascend or descent the gradient of the density, and thus change the steady state distribution which, as a steady state, by definition cannot change \footnote{Importantly, in this coffee-cream example, we are not claiming that if the stirring is removed then it will settle into a different steady state distribution, merely that the steady state with the stirring is non-equilibrium steady state (NESS) while the steady-state without stirring is an equilibrium steady state (ESS) due to the lack of solenoidal flow. Adding solenoidal flow to an ESS always can generate a NESS while the converse is not true. There are NESSs which can exist solely in virtue of their solenoidal dynamics without a corresponding ESS. An example of this would be a spinning top, which remains spinning solely due to its solenoidal motion.}. Biological self organizing systems are often considered to be `dissipative structures', or non-equilibrium steady states from the perspective of thermodynamics \citep{prigogine1973theory,kondepudi2014modern}, since they maintain a relatively steady state over time which requires a constant influx of energy to maintain. 

Given that we presuppose a system with a NESS density, we wish to understand the dynamics \emph{at} the NESS density -- specifically, how does the solenoidal flow help prevent the system from relaxing into an equilibrium-steady-state (ESS)? To understand this, we utilize the Helmholtz decomposition \citep{yuan2017sde,yuan2012beyond,friston2012ao} to rewrite the dynamics at the NESS into a form of a dissipative and solenoidal ascent upon the gradient of the log NESS density,
\begin{align*}
\frac{dx}{dt} = (\Gamma(x) - Q(x))\nabla_x \ln p^*(x) \numberthis
\end{align*}
Where $\Gamma(x)$ is a dissipative component of the flow which tries to ascend the log density. It is the amplitude of the random fluctuations in the original SDE formulation \citep{jordan1998variational,yuan2010constructive,yuan2011potential}, which in effect are constantly trying to `smooth out' the NESS density and increase its entropy. Conversely, the $Q(x)$ represents the solenoidal portion of the flow which, although orthogonal to the gradient of the log potential, successfully counteracts the dissipative effects of the $\Gamma(x)$ terms to maintain the dynamics at a steady state. While $\Gamma(x)$ and $Q(x)$ can in theory be state-dependent, from here on out we typically assume that they are not -- $\Gamma(x) = \Gamma$; $Q(x) = Q$, and additionally assume that $\Gamma$ is a diagonal matrix \footnote{Technically, we only need to assume a \emph{block-diagonal} matrix, but we also typically also assume that the noise in each state dimension is independent}, so there is no cross-correlation between states in the noise added to the system.

It is straightforward to verify that the Helmholtz decomposition of the dynamics satisfies the steady state condition $\frac{dp^*(X)}{dt} = 0$ by plugging this form into the Fokker-Planck equation (Equation \ref{FP_equation}),
\begin{align*}
\frac{dp^*(x)}{dt} &= - \nabla_x \big[(\Gamma - Q)\nabla_x \ln p^*(x) \big] p^*(x) + \Gamma \nabla^2_x p^*(x) \\
 &= - \nabla_x \big[(\Gamma - Q)\frac{\nabla_x p^*(x)}{p^*(x)} \big] p^*(x) + \Gamma \nabla^2_x p^*(x) \\
&= - \nabla_x \big[(\Gamma - Q) \nabla_x p^*(x) \big] + \Gamma \nabla^2_x p^*(x) \\
&= -\Gamma \nabla^2_x p^*(x) + \nabla_x Q \nabla_x p^*(x) + \Gamma \nabla^2_x p^*(x) \\
&= \nabla_x Q \nabla_x p^*(x) = 0 \numberthis
\end{align*}

Where the last line follows because, by definition, the gradient of the solenoidal flow with respect to the gradient of the log density is 0, since the solenoidal flow must be orthogonal to the gradient of the density, which is represented by the solenoidal $Q$ matrix being antisymmetric $Q = -Q^T$.

\section{Markov Blankets}

From these preliminaries, we have a set of dynamics of states $x$, which possess a NESS density, and we can express the dynamics at the NESS density in terms of dissipative $\Gamma$ and a solenoidal $Q$ flows on the gradient of the log density. Now, we begin to explore the statistical structure of these dynamics in terms of a Markov Blanket. Specifically, we next require that we can partition the states $x$ of the dynamics into three separate units. External states $\eta$, internal states $\mu$, and blanket states $b$ such that $x = [\eta,\mu,b]$. Intuitively, the external states represent the `environment'; the internal states represent the `system' we wish to describe, and the blanket states represent the statistical barrier between the system and its environment. For instance, we might wish to describe the dynamical evolution of a simple biological system such as a bacterium in such a manner. Here, the internal states would describe the internal cellular environment of the bacterium -- the cytoplasm, the nucleus, the ribosomes etc. The external states would be the environment outside the bacterium, while the blanket states would represent the cell membrane, sensory epithelia, and potentially active instruments such as the flagella which interact physically with the external environment. The key intuition behind the FEP is that although all influence between external and internal states is mediated by the blanket states, simply maintaining the non-equilibrium steady state against environmental perturbations requires that the internal states in some sense model and perform (variational) Bayesian inference on the external states. The Markov Blanket condition is straightforward. It simply states that the internal and external states must be independent given the blanket states,
\begin{align*}
\label{MB_condition}
p^*(x) = p^*(\eta,\mu,b) = p^*(\eta | b)p^*(\mu|b)p^*(b) \numberthis
\end{align*}

\begin{figure}
  \centering
  \includegraphics[scale=1]{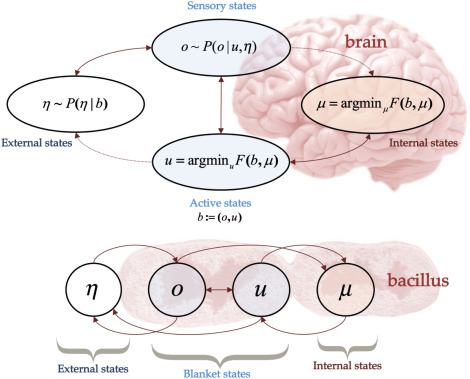}
  \caption{The intuition behind the Markov Blanket partition. The brain (or bacillus) consists of internal states $\mu$ which are separated from the outside world (external states $\eta$ by the blanket states $b$, which can themselves be partitioned into sensory states $s$, representing the sensory epithelia, and which are directly influenced by external states, and active states $a$ representing the organisms effectors and which are directly influenced by internal states, and act on external states. We see that perception concerns the minimization of free energy of the \emph{internal states}, while action concerns the minimization of the expected free energy of the \emph{active states}. Figure originally appeared in \citet{friston2019particularphysics}}
\end{figure}

While in probabilistic terms this factorisation is straightforward, it has more complex consequences for the dynamical flow of the system. Firstly, we additionally decompose the blanket states into sensory $s$ and active $a$ states such that $b = [s,a]$ and thus, ultimately $x = [\eta,\mu,s,a]$. Sensory states are blanket states that are causal children of the external states -- i.e. the states that the environment acts on directly. Active states are those blanket states that are not causal children of the external states. Essentially, external states influence sensory states, which influence internal states, which influence active states, which influence external states. The circular causality implicit in this loop is what allows the Markov Blanket condition to represent the perception-action loop. For notational purposes, we also define \emph{autonomous} states $\alpha = [a,\mu]$ which consist of active and internal states, and \emph{particular} states $\pi = [\mu,s,a]$ which consist of sensory, active, and internal states.

The next step is to understand what the conditional independence requirements put forth in Equation \ref{MB_condition} imply for the dynamics of the flow. Specifically, we obtain the \emph{marginal flow lemma} (see \citet{friston2019physics} for a full derivation), which states that the marginal flow of a partition, averaged under its complement, can be expressed as an Ao-decomposed flow on the gradients of the log of the marginal distribution. For instance, the flow of the internal states $\mu$, averaged under the complement $\tilde{\pi}$ of the particular states $\pi$ can be expressed as,

\begin{align*}
  f_\mu (\pi) \triangleq \E_{p(\tilde{\pi} | \pi)}[\frac{d \mu (x)}{dt}] = (\Gamma_{\mu \mu} - Q_{\mu \mu)} \nabla_\mu \ln p^*(\pi) + Q_{\mu \tilde{\mu}} \nabla_{\tilde{\mu}} \ln p^*(\pi) \numberthis
\end{align*}

Importantly, we see that the marginal flow lemma allows us to express the flow of a subset of states, averaged under their complements, in terms of independent Helmholtz decompositions on their marginal NESS densities, if we ignore solenoidal coupling terms (such as $Q_{\mu \tilde{\mu}}$). This allows us to investigate in detail the information-theoretic interactions of one set of states with another, and allows us to gain intuition and understanding of the core information-theoretic properties of the perception-action loop. For instance, using the marginal flow lemma we can express the flow of autonomous (active and internal) $\alpha = (a,i)$ as,
\begin{align*}
f_\alpha(x) = (\Gamma_{\alpha \alpha} - Q_{\alpha \alpha}) \nabla_\alpha \ln p^*(\pi) \numberthis
\end{align*}
where we see that autonomous states follow a gradient descent on the marginal NESS density of the internal, sensory, and active states, and attempt to suppress their surprisal or, on average, their entropy. We can use a series of mathematical `inflationary devices' to express this surprisal in terms of its interaction with the external states beyond the blanket.
\begin{align*}
  -\ln p^*(\pi) &= \E_{p^*(\eta | \pi)}\big[ -\ln p^*(\pi) \big] \\
  &= \E_{p^*(\eta | \pi)}\big[ \ln p^*(\eta | \pi) -\ln p^*(\eta,\pi) \big] \\
  &= \E_{p^*(\eta | \pi)}\big[ \ln p^*( \eta| \pi) -\ln p^*(\pi | \eta) - \ln p^*(\eta) \big] \\
  &=\underbrace{\E_{p^*(\eta | \pi)}\big[ -\ln p^*(\pi | \eta) \big]}_{\text{Inaccuracy}} + \underbrace{\KL \big[p^*(\eta | \pi) || p(\eta)\big]}_{\text{Complexity}} \numberthis
\end{align*}
Thus we can see that the flow of autonomous states acts to minimize the inaccuracy (maximize accuracy) and minimize the complexity of the external states with respect to the particular states of the system in question. Parsed into more intuitive terms, we can thus see that the flow of `system' states ($\pi$) aim to maximize the \emph{likelihood} of the internal states given the external states -- i.e. perform maximum likelihood inference on themselves (c.f. `self evidencing' \citep{hohwy2016self}) -- while simultaneously minimizing the complexity -- or the divergence between the external states given the internal states, and the `prior' distribution over the external states. In short, by re-expressing the flow in information-theoretic terms, we can obtain a decomposition of the entropy term into intuitive and interpretable sub-components which can help us reason about the kinds of behaviours these systems must exhibit.

\section{Variational Inference}

Variational inference is a technique and method for approximating intractable integrals in Bayesian statistics \citep{feynman1998statistical,jordan1998introduction,ghahramani2001propagation,jordan1999introduction,fox2012tutorial,neal1998view}. Typically, a direct application of Bayes-rule to compute posteriors in complicated systems fails due to the intractability of the log model evidence, which appears in the denominator of Bayes' rule. While there exist numerical or sampling-based methods to precisely compute this integral, they typically scale poorly with the dimension of the problem -- a phenomenon which is known as the curse of dimensionality \citep{goodfellow2016deep}. Variational techniques originated from methods in statistical physics in the 1970s and 1980s \citep{feynman1998statistical}, and were then taken up in mainstream statistics and machine learning in the 1990s \citep{ghahramani2001propagation,beal2003variational,jordan1998introduction} where they have become an influential, often dominant approach for approximating posteriors and fitting complex high-dimensional Bayesian models to data \citep{feynman1998statistical,jordan1999introduction,ghahramani2000graphical,beal2003variational,blei2017variational,kingma_auto-encoding_2013,dayan1995helmholtz}.

The core idea of variational inference is to approximate an intractable inference problem with a tractable optimization problem. Thus, instead of directly computing a posterior distribution $p(H | D)$ where $H$ is some set of hypotheses and $D$ is the data, we instead postulate an approximate or variational distribution $q(H | D; \theta)$ which is often, although not always, parametrized with some fixed number of parameters $\theta$. We then seek to optimize the parameters $\theta$ to minimize the divergence between the approximate and true posterior,
\begin{align*}
\theta^* = \underset{\theta}{argmin} \, \KL[q(H | D; \theta) || p(H | D)] \numberthis
\end{align*}
Unfortunately, this optimization problem is itself intractable since it contains the intractable posterior as an element. Instead, we minimize a tractable bound on this quantity called the variational free energy (VFE) $\mathcal{F}(D,\theta)$,
\begin{align*}
\mathcal{F}(D, \theta) &= \KL[q(H | D; \theta) || p(H, D)] \\
&= \KL[q(H | D; \theta) || p(H | D)] - \ln p(D) \\
&\geq \KL[q(H | D; \theta) || p(H | D)] \numberthis
\end{align*}

Since the VFE is based on a divergence between the variational distribution and the generative model $p(D,H)$, it is tractable as we assume we know the generative model that gave rise to the data. By minimizing the VFE, therefore, we reduce the divergence between the true and approximate posteriors, and thus improve our estimate of the posterior. 

Secondly, the variational free energy is simultaneously a bound upon the log model evidence $\ln p(D)$, a quantity of great important for model-selection \citep{geweke2007Bayesian,friston2018Bayesian}, and which is usually intractable to compute due to the implicit integration over all possible hypotheses (or parameters) $p(D) = \int dH p(D | H)p(H)$.
\begin{align*}
\ln p(D) &= \KL[q(H | D; \theta) || p(H, D)] - \mathcal{F}(D,\theta) \\
&\geq -\mathcal{F}(D,\theta) \numberthis
\end{align*}
The second line follows due to the non-negativity of the KL divergence. The VFE is the foundation of the free energy principle as, we shall show, we can interpret self-organizing systems which maintain themselves at a non-equilibrium-steady state to be implicitly minimizing the VFE, and thus performing variational Bayesian inference.

It is important to note here that while the variational free energy $\mathcal{F}$ is not technically a KL divergence, since the two distributions it involves do not share the same support (one being a posterior and the other a joint), for notational convenience in this thesis we slightly abuse the KL notation to represent free energies of one form or another. Formally, we will use $\mathcal{F}[q(x), p(x,y)] = \E_{q(x)}[\ln p(y | x)] + \KL[q(x)||p(x)] = \ln p(y) + \KL[q(x)||p(x|y)] := \KL[q(x)||[(x,y)]$.

We can gain some intuition for the effects of minimizing the VFE by decomposing into various constituent terms. Here we showcase two different decompositions which each give light to certain facets of the objective function,
\begin{align*}
\label{VFE_decomp}
  \mathcal{F}(D, \theta) &= \underbrace{\E_{q(H | D; \theta})[\ln p(H,D)]}_{\text{Energy}} - \underbrace{\mathbf{H}[q(H | D; \theta)]}_{\text{Entropy}} \numberthis \\ 
  &= -\underbrace{\E_{q(H | D;\theta)}[\ln p(D | H)]}_{\text{Accuracy}} + \underbrace{\KL[q(H | D;\theta)||p(H)]}_{\text{Complexity}} \numberthis
\end{align*}
Here we see that we can decompose the variational free energy into two separate decompositions, each consisting of two terms. The first decomposition splits the VFE into an `energy' term, which effectively scores the likelihood of the generative model averaged under the variational distribution, whilst the entropy term encourages the variational distribution to become maximally entropic. Essentially, this decomposition can be interpreted as requiring that the variational distribution maximize the joint probability of the generative model (energy), while simultaneously remaining as uncertain as possible (entropy) \footnote{Interestingly, this energy, entropy decomposition is precisely why this information-theoretic quantity is named the variational \emph{free energy}. The thermodynamic free energy, a central quantity in statistical physics, has an identical decomposition into the energy and the entropy.}. The second decomposition -- into an `accuracy' and a `complexity' term -- speaks more to the role of the VFE in inference. Here the accuracy term can be interpreted as driving the variational density to produce a maximum likelihood fit of the data, by maximizing their likelihood under the variational density. The complexity term can be seen as a regularizer, which tries to keep the variational distribution close to the prior distribution, and thus restrains variational inference from pure maximum-likelihood fitting.
\section{Intrinsic and Extrinsic information geometries}

Now, we wish to understand the relationship between the internal states and the external states, which are separated by the blanket states. Importantly, the existence of the blanket means that we can define a mapping between the most likely internal state, given a specific blanket state, and a distribution over external states \footnote{This function is defined if we assume injectivity between the most likely internal and blanket states \citep{parr2020Markov}.}. We define the most likely internal and external states given a blanket state as,
\begin{align*}
\bm{\eta}(b) &= \underset{\eta}{argmax} \, p(\eta | b) \\
\bm{\mu}(b) &= \underset{\mu}{argmax} \, p(i | b) \numberthis
\end{align*}

Next, we assume that there is a smooth and differentiable function $\sigma$ which maps between the most likely internal and external states given a blanket state,
\begin{align*}
\bm{\eta}(b) = \sigma(\bm{\mu}(b)) \numberthis
\end{align*}
Importantly, we interpret the output of this function -- the most likely external states given the blanket states -- as parametrizing the mean over a full distribution over the external states, as a function of the internal states $q(\eta; \bm{\eta}(b)) = q(\eta; \sigma(\bm{\mu}(b)))$. This allows us to interpret the flow of internal states as parametrising distributions over the external states. 

Crucially, we can say that if any given set of internal states parametrizes a distribution over external states, then the space of internal states effectively represent a \emph{space of distributions} over external states, parametrized by internal states. This space of distributions may be, and usually is, curved and non-euclidean in nature. The field of information geometry has emerged to allow us to describe and mathematically characterise such spaces correctly \citep{amari1995information,caticha2015basics}. A key result in information geometry is that the space of parameters of families of exponential distributions is a non-euclidean space with the Fisher Information as its metric. A metric is simply a notion of distance for a given space. For instance, in Euclidean space, the metric is $\sqrt{\sum_\mu^N x_\mu^2}$ where $N$ is the dimensionality of the space. We can represent general coordinate transformers on spaces with any metric through the use of a metric tensor $\bm{G}$. Essentially, we measure differences between distributions in terms of the KL divergence, and thus if we want to see how an infinitesimal change in the parameters of a distribution results in changes to the distribution itself, we can measure an infinitesimal change in their KL divergence as a function of the infinitesimal change in the parameters. i.e. 
\begin{align*}
\frac{\partial p(x; \theta)}{\partial \theta} = \lim_{\delta \theta \to 0} \KL[p(x; \theta) || p(x; \theta + \delta \theta)] \numberthis
\end{align*}
In the case of the space of parameters of exponential distributions, the metric tensor is the Fisher information, which arises as from the Taylor expansion of the infinitesimal KL divergence between the two distributions. We define $\theta' = \theta + \delta \theta$. Specifically, since there is only an infintesimal change, we can Taylor-expand around $\theta' = \theta$ to obtain,
\begin{align*}
  \KL[p(x;\theta)||p(x;\theta')] &\approx \underbrace{\KL[p(x;\theta)||p(x;\theta)]}_{=0} + \underbrace{\frac{\partial \KL[p(x;\theta)||p(x;\theta')]}{\partial \theta}|_{\theta = \theta'}(\theta - \theta')}_{=0} \\ &+ \frac{\partial^2 \KL[p(x;\theta)||p(x;\theta')]}{\partial \theta^2}|_{\theta = \theta'}(\theta - \theta')^2 \numberthis
\end{align*}`
Where the first two terms vanish, so we need only handle the second term,
\begin{align*}
  \KL[p(x;\theta)||p(x;\theta')] &\approx \frac{\partial^2 \KL[p(x;\theta)||p(x;\theta')]}{\partial \theta^2}|_{\theta = \theta'}(\theta - \theta')^2 \\
  &= \int p(x; \theta) \frac{\partial \ln p(x;\theta)}{\partial \theta}\frac{\partial \ln p(x;\theta)}{\partial \theta} \\ 
  &= \mathbb{F} \numberthis
\end{align*}
where $\mathcal{\mu}$ is the Fisher information. Since the internal states can be interpreted as parametrizing distributions over external states, as parameters, they lie on an information-geometric manifold with a Fisher information metric. This is the extrinsic information geometry. Simultaneously, the internal states also parametrize (implicitly) a second (empirical) distribution over the internal states. This parametrization gives rise to a second information geometry -- the intrinsic geometry, since it represents the relationship the internal states have to the distribution over themselves. Specifically, suppose $\bm{\mu}$ define the sufficient statistics of a variational density over internal states $q(\mu; \bm{\mu})$, and $\bm{\eta} = \sigma(\bm{\mu})$ define the sufficient statistics of the variational density over external states $q(\eta;\bm{\eta})$, then we can see that the internal states in fact parametrize \emph{two} densities and thus partake in two simultaneous information geometries. First, there is a metric defined over the space of internal densities,
\begin{align*}
  \mathcal{\mu}(\bm{\mu}) = \frac{\partial ^2 \KL[q(\mu;\bm{\mu})||q(\mu;\bm{\mu} + \delta \bm{\mu})]}{\partial \bm{\mu}^2}|_{\bm{\mu} + \delta \bm{\mu} = \bm{\mu}}\numberthis
\end{align*}
which is called the \emph{intrinsic} information geometry. And secondly, a metric defined over the space of external densities, parametrized by internal states,
\begin{align*}
  \mathcal{\mu}(\bm{\eta}) =\frac{\partial^2 \KL[q(\eta; \bm{\eta})||q(\eta;\bm{\eta} + \delta \bm{\eta})]}{\partial \bm{\eta}^2}|_{\bm{\eta} + \delta \bm{\eta} = \bm{\eta}} \numberthis
\end{align*}
which is called the \emph{extrinsic} information geometry. These well-defined intrinsic and extrinsic information geometries, allow us to interpret the motion of the internal as also movement on the intrinsic and extrinsic statistical manifolds. Crucially, enabling us to make mathematically precise the link between two conceptually distinct ideas -- dynamical motion in space, and variational inference (i.e. Bayesian belief updating) on parameters of distributions. Using this underlying information-geometric framework, in the next section we shall go on to see how we can interpret the dynamics of a non-equilibrium system at NESS as performing approximate variational Bayesian inference on its external environment.

\section{Self-Organization and Variational Inference}
Here we present the key results of the free energy principle via the free energy lemma. This says, firstly, that the dynamics of the autonomous states can be interpreted as minimizing a free energy functional over the external states, and thus can be construed as performing a kind of elemental Bayesian (variational) inference. Specifically, we will first consider the general case in terms of the `particular' free energy, which stipulatively assumes that the system obtains the correct posterior at every time-point, rendering the traditional variational bound superfluous, and thus demonstrating that in a way self-organizing systems maintaining themselves at NESS can be construed as performing \emph{exact} Bayesian inference on the generative model they embody through their NESS density. We thus reach the key statement of the FEP -- that the dynamics of self-organizing systems that maintain themselves at NESS can be interpreted as performing exact Bayesian inference on the external states beyond their blanket or, alternatively, they can be interpreted as approximating approximate (variational) Bayesian inference. 

We then introduce the general case of the \emph{variational free energy}, which is in general a bound upon the log of the NESS density, and we show in the special case of assuming that the variational distribution over external states which is parametrized by the internal states can be approximated by the Laplace approximation, that we can interpret the flow of autonomous states as directly performing a descent upon the variational free energy and thus directly performing variational Bayesian inference. Since we, as the modeller, can specify the variational distribution in any desired way, then this means that this interpretation is tenable for an extremely wide range of systems. The Laplace approximation approximates the variational distribution as a Gaussian where the variance is a function of the curvature at the mean. Intuitively, this assumption is that the Gaussian is tightly peaked around the mean value. This approximation is theoretically well-justified, due to the underlying Gaussianity of the stochastic noise in the system, and the likely concentration of the probability mass around the mean. Moreover, the Gaussian distribution arises regularly in nature whenever averages over large numbers of independent events are taken (c.f. the Central Limit Theorem (CLT)), and can thus be considered a natural modelling choice for distribution of the mode of the external states given the blanket, which likely is composed of contributions from a large number of specific external states. 

To recall, we can write the flow of autonomous states $\alpha = (i,a)$ in terms of a gradient descent on the log NESS density of the particular states $\ln p(\pi)$ with both dissipative and solenoidal components via the Helmholtz decomposition.
\begin{align*}
f_\alpha(x)= -(\Gamma - Q)\nabla_\alpha \ln p^*(s, \alpha) \numberthis
\end{align*}

Then we can define the particular free energy as the variational free energy, where the variational distribution over external states, is stipulatively defined to be equal to the `true' posterior distribution over external states given the particular states $q(\eta | \pi) = p^*(\eta | \pi)$ \footnote{We implicitly assume here that the variational distribution can be stipulated to be of the same family of the true posterior, so that they can match one another}. With this assumption, we can define the particular free energy using the standard form for the variational free energy
\begin{align*}
\mathcal{F}_{particular} &= \KL[q(\eta | \pi) || p^*(\eta,\pi)] \\
&= -\underbrace{\E_{q(\eta | \pi}[\ln p^*(\eta | \pi)]}_{\text{Accuracy}} + \underbrace{\KL[(q\eta | \pi) || p^*(\pi)]}_{\text{Complexity}} \\
&= \underbrace{\ln p^*(\pi)}_{\text{Evidence}} + \underbrace{\KL[q(\eta | \pi) || p^*(\eta | \pi)]}_{\text{Bound}} \\
&= \ln p^*(\pi) \numberthis
\end{align*}
where the last line follows because the bound is always 0 since we have defined the variational and true posteriors to be the same. Importantly, we see that the particular free energy is then equal to the log of the NESS density over the sensory, internal, and active states. As such, we can rewrite the dynamics of the autonomous states directly in terms of the particular free energy,
\begin{align*}
f_\alpha(x)= -(Q - \Gamma) \nabla_\alpha \mathcal{F}_{particular}(s, \alpha) \numberthis
\end{align*}

While this may seem like just a mathematical sleight of hand, it demonstrates how systems which maintain the statistical structure of a Markov Blanket at equilibrium can in fact be interpreted as performing variational Bayesian inference with a correct posterior distribution. If, conversely, we relax this assumption somewhat, so that, as is typical for variational inference when the class of distributions represented under the variational density does not include the true posterior, then we retain an approximate relationship. That is, when $q(\eta | \pi;\theta) \approx p(\eta | \pi)$, we obtain,
\begin{align*}
\mathcal{F} &= \KL[q(\eta |\pi) || p^*(\eta,\pi)] \\
&= \ln p^*(\pi) + \KL[q(\eta | \pi) || p^*(\eta | \pi)] \\
&\approx \ln p^*(\pi) \\
&\implies f_\alpha(x) \approx -(Q - \Gamma) \nabla_\alpha \mathcal{F}(s, \alpha)
\end{align*}

So we can see that in this case, we can interpret the dynamics of the autonomous states as \emph{approximating} approximate Bayesian inference. This is perhaps the most general statement of the FEP -- that the dynamics of a system which maintains the statistical structure of a Markov Blanket at NESS against external dissipative perturbations, can be interpreted as performing approximate variational Bayesian inference to optimize a distribution over the external states of the environment, parametrized by its own internal states. The distinction between variational and particular free energy, with the particular free energy always using the stipulatively correct posterior, while being somewhat a mathematical trick, is also a useful philosophical distinction to draw. In effect, we can think of the system as always performing correct Bayesian inference, simply because the inference is over the system itself, where the generative model of the system is simply its NESS density. Conversely, we can see the approximation arising from the approximate variational distribution as being related to the imperfection of our own understanding of the system as an exogenous modeller. The system is perfectly happy using its Bayes-optimal posterior at all times. A variational distribution distinct from this posterior must be, in some sense, the creature and creation of the modeller, not of the system, and as such the approximations to the dynamics that arise from this approximation is due to the approximations implicit in modelling rather than in the dynamics of the system per-se. It is also important to note that while we have used an approximation sign, in reality the variational free energy is an \emph{upper bound} upon the log model evidence or the particular free energy -- i.e. $\mathcal{F} \geq \mathcal{F}_{particular}$ and the approximate dynamics can be interpreted as driving the system towards the minimization of this bound, and thus increasing the accuracy of the approximation in a manner analogous to the similar process inherent in variational inference.

While in the general case above, the relationship between the dynamics of the system and variational inference is only approximate, if we are only interested in the\emph{mode} of the external states -- i.e. the most likely external state configuration -- instead of the full distribution, then the approximation becomes exact and we can directly see that the dynamics of the system do perform variational inference upon the mode of the external states. Here we can see that, in a sense, the maximum-a-posteriori (MAP) modes for the internal states precisely track the MAP modes for the external states and thus, under the Laplace approximation, can be seen as directly performing a minimization of the variational free energy.

Firstly, recall from previously that we had defined the smooth mapping between the modes of the external and internal states given the blanket state, $\bm{\eta}(b) = \sigma(\bm{\mu}(b))$. By applying the chain rule to this function, it is straightforward to derive the flow of the external mode with respect to the internal mode,
\begin{align*}
\label{2_2_4}
f_{\bm{\eta}}(b) = \frac{\partial \sigma(\bm{\mu}(b))}{\partial \bm{\mu}(b)} f_{\bm{\mu}}(b)  \numberthis
\end{align*}
Then, assuming that the mapping is invertible (requiring that the internal states and external states have the same dimensionality), or rather in the general case that it has a Moore-Penrose pseudoinverse, we can express the dynamics of the internal mode in terms of the dynamics of the external mode,
\begin{align*}
f_{\bm{\mu}}(b)  = \frac{\partial \sigma(\bm{\mu}(b))}{\partial \bm{\mu}(b)}^{-1} f_{\bm{\eta}}(b)  \numberthis
\end{align*}
Similarly, we can derive the expression of the NESS density over the external mode in terms of the mode of the internal states, which provides a precise mapping, called the \emph{synchronization manifold}, between the two densities, even though they are in fact separated by the Markov Blanket,
\begin{align*}
\label{2_2_6}
\frac{\partial \ln p(\bm{\eta}(b) | b)}{\partial \mu} = \frac{\partial \ln p(\bm{\eta}(b) | b)}{\partial \bm{\eta}(b)}\frac{\partial \sigma(\bm{\mu}(b))}{\partial \mu} \numberthis
\end{align*}
Combining Equation \ref{2_2_4} and Equation \ref{2_2_6} and using the fact that the flow of the external mode, by the marginal flow lemma is, $f_{\bm{\eta}}(b)  = (\Gamma_\eta - Q_\eta) \nabla_\eta \ln p(\bm{\eta}(b) | b)$, we can express the flow of the internal mode in terms of the marginal NESS density over the external states, thus understanding how the internal states probabilistically track changes in their environment,
\begin{align*}
f_{\bm{\mu}}(b)  &= \frac{\partial \sigma(\bm{\mu}(b))}{\partial \bm{\mu}(b)}^{-1} \frac{d \bm{\eta}(b)}{dt} \\
&= \frac{\partial \sigma(\bm{\mu}(b))}{\partial \bm{\mu}(b)}^{-1} (\Gamma_\eta - Q_\eta) \nabla_\eta \ln p(\bm{\eta}(b) | b) \\
&= \frac{\partial \sigma(\bm{\mu}(b))}{\partial \bm{\mu}(b)}^{-1} (\Gamma_\eta - Q_\eta) \frac{\partial \sigma(\bm{\mu}(b))}{\partial \bm{\mu}(b)}^{-1} \frac{\partial \sigma(\bm{\mu}(b))}{\partial \bm{\mu}(b)} \nabla_\eta \ln p(\bm{\eta}(b) | b) \\
 &= (\Gamma_\sigma - Q_\sigma) \nabla_\mu \ln p(\sigma(\bm{\mu}(b))) \numberthis
\end{align*}
where $(\Gamma_\sigma - Q_\sigma) = \frac{\partial \sigma(\bm{\mu}(b))}{\partial \bm{\mu}(b)}^{-1} (\Gamma_\eta - Q_\eta) \frac{\partial \sigma(\bm{\mu}(b))}{\partial \bm{\mu}(b)}^{-1}$. Crucially, this expression allows us to write the flow of the internal mode as a gradient descent on the NESS density of the external mode as a function of the internal mode, given the blanket, with respect to the internal states. Fascinatingly, this relationship takes the same general form of the Helmholtz decomposition with separate dissipative $\Gamma_\sigma$ and solenoidal $Q_\sigma$ components which are simply the original dissipative and solenoidal components with respect to the internal states modulated by the inverse of the mapping function $\sigma$. In effect, this implements a coordinate transform between the coordinates of the flow of the external states to the coordinates of the flow of the mode of the external states, as a function of internal states.

Now we demonstrate how we can interpret this gradient descent on the NESS density of the mode over external states in terms of a direct descent on the variational free energy, and thus as directly and exactly performing variational inference. First, we must define our variational distribution $q(\bm{\eta} | b; \bm{\mu})$ which is a distribution over the modes of external states, given the blanket states, parametrized by the mode of the internal states. Since we are only interested now in distributions over the \emph{mode} of the external states, a reasonable assumption is that it is approximately Gaussian distributed due to the central limit theorem. This means that a Laplace approximation, which is a Gaussian approximation where the covariance is simply a function of the mean, derived via a second order Taylor-expansion of the density at the mode, is a good approximation to use here. We thus define the variational density as,
\begin{align*}
& q(\bm{\eta} | b; \bm{\mu}) = \mathcal{N}(\bm{\eta}; \bm{\mu}, \Sigma(\bm{\mu})) \\
& \text{where} \, \, \Sigma(\bm{\mu}) = \frac{\partial^2 \sigma(\bm{\mu})}{\partial \sigma^2}^{-1} \numberthis
\end{align*}
Importantly, if we substitute this definition of $q$ into the variational free energy and drop constants unrelated to the variational parameters $\bm{\mu}$, we obtain,
\begin{align*}
\mathcal{F} &= \ln p(\bm{\mu}, b) + \frac{1}{2}tr(\Sigma(\bm{\mu})) \frac{\partial^2 \sigma(\bm{\mu})}{\partial \sigma^2}^{-1} + \ln | \Sigma(\bm{\mu}) | \\
&\implies \frac{\partial \mathcal{F}}{\partial \mu} = \frac{\partial \ln p(\bm{\mu}, b)}{\partial \mu}
\end{align*}
The second line follows since this is the only term where $i$ is directly utilized. Then, from this definition, we can see that the variational free energy is actually precisely the gradient term we see in the expression for the flow of the internal state mode, thus allowing us to rewrite it as,
\begin{align*}
\label{Free_energy_descent}
f_{\bm{\mu}}(b)  = (\Gamma_\sigma - Q_\sigma) \nabla_\mu \mathcal{F} \numberthis
\end{align*}
After this thicket of mathematics, we thus see a crucial result for the FEP. That, with a Laplace-encoded variational density, we can see that the mode of the internal states precisely tracks the mode of the external states, and the dynamics that allows it to do so are precisely those of a gradient descent on the variational free energy, thus enabling an exact interpretation of the flow of the internal states as performing Bayesian inference on the external states. This proof demonstrates the fundamentally Ashbyan nature of self-organization at non-equilibrium steady state, where systems, in order to maintain their steady state, and thus existence as distinct systems, are necessarily forced to engage in some degree of modelling or tracking external states of the environment, in order to counter their dissipative perturbations. Interestingly, this exact relationship to variational inference only emerges when considering the \emph{modes} of the system, not the full distribution over environmental and internal states as was done previously, where we only obtained an approximation to variational inference. Perhaps this is because, in some sense, the system need not perform inference on full distributions, but only on modes. This perhaps makes more intuitive sense within the cybernetic Ashbyan paradigm where, in general, the system is seen as significantly smaller than the environment, and thus simply cannot be expected to encode a fully accurate model of the entire environment which, in the extreme case, includes the entire rest of the universe. Instead, the system simply models and tracks coarse-grained environmental variables such as the mode.

\section{The Expected Free Energy and Active Inference}
So far, we have only considered the relationship between internal and external states, and observed that the flow of the internal state can be considered to be performing a variational gradient descent on the parameters of the variational density over external states. The internal state dynamics exactly follow a variational gradient descent if we assume that the internal states parametrize a Laplacian approximate posterior, or they approximately follow a variational gradient descent if we assume a broader class of variational posteriors. From this, we can interpret the flow of the internal states as performing some kind of `perceptual' inference about the causes of fluctuations in the blanket states -- namely, the external states. But what about the active states? How do they fit into this picture? 

First, we recall from the approximate Bayesian inference lemma that we can express the flow of the autonomous states (active and internal) in terms of an approximate gradient descent on the variational free energy (Equation \ref{Free_energy_descent}). By the marginal flow lemma, if we ignore solenoidal coupling between internal and active states, we can partition this descent into separate (marginal) descents on the internal and the active states, allowing us to write the flow of the active states as
\begin{align*}
   f_a(x) \approx (\Gamma_{aa} - Q_{aa}) \nabla_a \mathcal{F}(s, \alpha) \numberthis
\end{align*}
where $\Gamma_{aa}$ and $Q_{aa}$ are the block matrices corresponding solely to the interactions between active states in the larger $\Gamma$ and $Q$ matrices. Crucially, if we recall the definition of the variational free energy,
\begin{align*}
  \mathcal{F}(\pi) &= \KL[q(\eta | \pi; \bm{\mu})||p^*(\eta, \pi)] \\
  &= -\underbrace{\E_{q(\pi | \bm{\mu})}[ -\ln p^*(\pi | \eta)]}_{\text{Inaccuracy}} + \underbrace{\KL[q(\eta | \pi ; \bm{\mu})||p^*(\eta)]}_{\text{Complexity}} \numberthis
\end{align*}
Crucially, the only term in this decomposition that depends on the active states $a$ is the first \emph{inaccuracy} term. Thus, we can straightforwardly write down the flow of the active states as,
\begin{align}
  f_a(x) \approx (\Gamma_{aa} - Q_{aa}) \nabla_a \E_{q(\eta | \bm{\mu})}[ -\ln p^*(\pi | \eta)] \numberthis
\end{align}
Where we can intuitively see that the flow of the active states effectively \emph{minimize} inaccuracy (or maximize accuracy). In effect, we can interpret the flow of the active states at the NESS density to try to ensure that the variational `beliefs' encoded by the blanket and internal states of the system are as accurate as possible. Since active states can only influence external states and \emph{not} internal states, the way this is achieved is by acting upon the external states to bring them into alignment with the beliefs represented by the internal states -- hence \emph{active inference}.

While this provides a good characterisation of the flow of the system at equilibrium, we are often also interested in the properties of dynamical systems as the self-organize \emph{towards} equilibrium. Specifically, we wish to characterise the nature of the active states during this process of self-organization, so that we can understand the necessary kinds of active behaviour any self-organizing system must evince. To begin to understand the nature of this self-organization we first define another information theoretic quantity, the \emph{Expected Free Energy} (EFE) which serves as an upper-bound on surprisal throughout the entire process of self-organization, with equality only at the equilibrium itself. Since we have this upper-bound, we can interpret self-organizing systems away from equilibrium, by following their surprisal dynamics as approximating expected free energy minimization, using logic directly analogous to the approximate Bayesian inference lemma. Conversely, turning this logic around lets us \emph{construct} self-organizing systems by defining some desired NESS density, and then prescribing dynamics which simply minimize the EFE.

To handle systems away from equilibrium, we define some new terminology. We define $p(\eta_t, \mu_t, s_t, a_t | \eta_0, \mu_0, s_0, a_0)$ to be the probability density over the variables of the system at some time $t$, which depends on some set of initial conditions $e_0, \mu_0, s_0, a_0$. To simplify, we average over the external initial condition and only represent the particular initial condition $\pi_0 = (\mu_0, s_0, a_0)$. Next we define the expected free energy $\mathcal{G}(\pi)$ similarly to the variational free energy, but with the current-time predictive density taking the place of the approximate variational posterior, and the NESS density taking the place of the generative model.
\begin{align*}
  \mathcal{G}(\pi) &= \E_{p(\eta_t, \pi_t) | \pi_t)}[\ln p(\eta_t | \pi_t, \pi_0) - \ln p^*(\eta, \pi)] \\
  &= \underbrace{\E_{p(\eta_t, \pi_t) | \pi_t)}[-\ln p^*(\pi | \eta)]}_{\text{Ambiguity}}+ \underbrace{\KL[p(\eta_t | \pi_t, \pi_0)||p^*(\eta)]}_{\text{Risk}} \numberthis
\end{align*}
We see that the EFE mandates the minimization of both ambiguity (i.e. avoiding situations which are heavily uncertain) and risk (avoiding large divergences between the current state density and the equilibrium state. It is straightforward to see that the EFE is an upper bound on the expected predictive surprisal at any time-point, by using the fact that the KL-divergence is always greater than or equal to 0,
\begin{align*}
  &\KL[p(\eta_t, \pi_t | \pi_0)||p^*(\eta, \pi)] \geq 0 \\
  &\implies \mathcal{G}(\pi_t) + \E_{p(\eta_t, \pi_t) | \pi_t)}\ln p(\pi_t | \pi_0)] \geq 0 \\
  &\implies \mathcal{G}(\pi_t) \geq - \E_{p(\eta_t, \pi_t) | \pi_t)}[\ln p(\pi_t | \pi_0)]
\end{align*}
Similarly, it is straightforward to see that at equilibrium, the EFE simply becomes the surprisal.
\begin{align*}
  \KL[p(\eta_t, \pi_t | \pi_0)||p^*(\eta, \pi)] &=  \mathcal{G}(\pi_t) + \E_{p(\eta_t, \pi_t) | \pi_t)}[\ln p(\pi_t | \pi_0)] =0 \\
  &\implies \mathcal{G}(\pi_t) = - \E_{p(\eta_t, \pi_t) | \pi_t)}[\ln p(\pi_t | \pi_0)] \numberthis
\end{align*}
Since this is the case, we can understand the EFE as effectively quantifying the discrepancy between the current predictive density and the equilibrium. Because of this, we can see that the EFE is necessarily a Lyapunov function of self-organizing dynamics, and it makes sense to interpret self-organizing dynamics under a Markov blanket as minimizing the EFE. Conversely, if one wants to define a set of dynamics that self-organize to some given attractor $p^*(\eta,\pi)$ then one simply needs to define dynamics that minimize the EFE to achieve convergence to the equilibrium (in the case where there are no local minima). 

Taking this converse approach allows us to move from simply providing an interpretative characterisation of given dynamics in terms of inference, and move instead to constructing or defining systems, or agents, which can achieve specific goals. This approach is taken in the literature on active inference process theories \citep{friston2012active,friston2015active,friston2017active,da2020active} where instead of simply describing a given stochastic differential equation, we instead consider the NESS density to be the \emph{preferences} or \emph{desires} of the agent often represented as a Boltzmann distribution over environmental rewards $p^*(\eta,\pi) = exp(-r(\eta))$ and the active states (the agent's actions) being computed through a minimization of the EFE, with this minimization either taking place directly as a gradient descent in continuous time and space \citep{friston2009reinforcement} or else as an explicit model-based planning algorithm as in the discrete-time and discrete-space formulation \citep{friston2017process,tschantz2020reinforcement,millidge_deep_2019,millidge2020relationship}.

\section{Philosophical Status of the FEP}
It is worth stepping back from the mathematical morass, at this point, to try to define at a high level what kind of theory, philosophically speaking the FEP is, and what kind of claims it makes. There have been numerous debates in the literature about whether the FEP is `falsifiable', or whether it is `correct', and whether or not it makes any specific, empirical claims \citep{williams2020brain,andrews2020math}. However often debates on this matter are obscured or confused by the challenging and deep mathematical background required for a full understanding of the specifics of the FEP. It is clear from the mathematics that the FEP offers only an `interpretation' of already extant dynamics. In short, FEP presupposes the existence of the kinds of dynamics it wishes to make sense of -- dynamical systems which organize themselves into a non-equilibrium steady state, and which maintain the requisite statistical independency structure of the Markov Blanket condition. Once these conditions are satisfied, the FEP gives an interpretation of the dynamical evolution of such a system as performing a kind of variational Bayesian inference whereby the internal states of the system (defined by the Markov Blanket partition) can be seen as inferring or representing external states which are otherwise statistically isolated behind the Markov Blanket. Crucially, the FEP, in its most general formulation does not make any specific predictions about the flow of the system. It offers an interpretation only. While systems that implement the FEP can be derived, and several process theories have been explicitly derived from within the FEP framework \citep{friston2005theory,friston2015active}, all such theories necessitate making specific and ultimately arbitrary modelling choices, such as of the generative model and variational density. Such choices sit below the level of abstraction that the mathematical theory of the FEP exists at. The FEP offers a mathematical interpretation only of certain dynamical structures.

The FEP is often compared and analogised to the principle of least action in physics \citep{lanczos2012variational} which allows one to describe many physical processes (although not all) as minimizing the path integral of a functional called the `action' over a trajectory of motion \citep{sussman2015structure}. This argument is often used to claim, in my opinion correctly, that the FEP is a mathematical `principle' or interpretation and therefore cannot be falsified or empirically tested. In my opinion, however, the principle of least action is, in its philosophical status, not directly analogous to the FEP. While the relationship between the path integral of the action and the dynamics prescribed by the Euler-Lagrange equations is simply a mathematical truth, the principle of least action itself, as applied to physics contains a fundamentally empirical and falsifiable claim -- that physical systems in the real world can be well described through its own mathematical apparatus -- that is of dynamics derived from minimizing an action. This claim is in principle falsifiable. Not all dynamical systems can be derived from least action principles. If physical systems predominantly came from the class that cannot be so derived, the principle of least action in physics would be effectively falsified, and the mathematical apparatus underlying it would have become nothing more than an arcane mathematical curiosity. So far as we know, there is no a-priori reason why much of physics can be so well understood through action principles, and indeed there are areas of physics -- such as statistical mechanics and thermodynamics, and dissipative non-conservative systems in general -- which \emph{cannot} (so far) be described straightforwardly in these terms.

It appears a closer physics analogy to the FEP might be one direction of Noether's theorem. Noether's theorem proves a direct correspondences between symmetries or invariances in a given system, and conservation laws. For instance, in physical systems, time-translation symmetry implies the conservation of energy, and rotational symmetry (of the underlying euclidean space, not any given object within it) implies the conservation of angular momentum. The FEP, similarly, can be thought to show a correspondence between the dynamics of a certain kind of system (NESS density, Markov Blanket conditions) and the dynamics of variational Bayesian inference. Interestingly, while the `forward' direction from the NESS density and Markov Blanket conditions is treated in the FEP, any reverse conditions -- i.e. whether the presence of Bayesian inference dynamics implies any kind of statistical structure upon the dynamics of the system remains unclear, and this is likely a fruitful direction for further theoretical work. Noether's theorem, unlike the principle of least action, matches more closely than the principle of least action since it only specifies correspondences between certain kinds of mathematical objects (symmetries and conservation laws) just as the FEP only specifies a correspondence between dynamical flows at NESS of a system with a Markov Blanket, and the gradient flows on the variational free energy.

While its status as a mathematical principle and interpretation only can shield the FEP from the possibility of an empirical `falsification', this does not mean that the theory is not subject to some kind of implicit intellectual review. Much of the core motivation behind the FEP has been to try to derive universal properties of the kind of biological self-organizing systems which give rise to structured behaviour including relatively `high level' processes such as the perception-action-loop, explicit perception and inference about the causes of the external world and, ultimately, prospective inference and planning. For instance, much of the FEP literature has been focused on and applied to understanding brain function \citep{friston2008hierarchical,friston2015active,friston2017process}. This ambition renders the FEP open to questions about its `applicability', if not its falsifiability. The FEP imposes relatively stringent conditions that dynamical systems must satisfy for the logical steps in the FEP to hold. In the next section, we present a detailed itemized list and critical discussion of all the assumptions required. Some especially key assumptions, which substantially restrict the potential class of systems the FEP can apply to are:
\begin{itemize}
\item That the system in question can be adequately represented as a Langevin equation (i.e. the system is Markov and does not depend on history) with additive white Gaussian noise.
\item that the dynamical system as a whole have a well-defined NESS density (including over the external states).
\item that the system obey the Markov Blanket conditions, which are, in general, relatively restrictive about the kinds of flows that are possible, and appear to have become more restrictive in \citet{friston2020some}, which precludes any solenoidal coupling between active and sensory states (indeed the didactic treatments of the free energy lemma typically require a block-diagonal Q matrix, meaning \emph{no} solenoidal coupling between subsets of states). If this assumption is relaxed, then there are additional solenoidal coupling terms in the flow of the internal states, so at best one can say that gradient ascent upon the surprisal is a \emph{component of the flow}.
\item That there be an injective mapping between the most-likely internal state given the blanket and the mode of the distribution of external states given the blanket states, which is additionally smooth and differentiable (this is required for the dual-aspect information geometry, and thus the identification with Bayesian inference).
\end{itemize}
These conditions are quite strict about the class of systems that the FEP can apply to, and it is unclear if `real systems' of the kind that FEP desires to explain -- such as biological self-organization, and especially brains, can fulfil them. If it turns out that such systems flagrantly violate the conditions for the FEP, then the FEP cannot be said to apply to them and thus cannot be of use in understanding them, even as an interpretatory device. In this case, the FEP would fail the applicability criterion, and would cease to be particularly useful for its original goals of neuroscience, even if it remains not technically falsified and does, in fact, apply to some obscure mathematical class of dynamical systems. Importantly, many of the assumptions of the FEP, when interpreted strictly, do not appear to hold in general for complex biological systems such as brains. For instance, to take extreme but illustrative examples, it is clear that no biological system is ever in a true non-equilibrium steady state, since eventually all such organisms will age and die, and indeed eventually the entire universe will likely decay to a thermodynamic equilibrium state. Additionally, the Markov Blanket assumption is directly violated by things such as x-rays (and indeed gravity) which can directly interact with `internal states' of the brain, such as neurons, without first passing through the Markov Blanket of the physical boundaries of the brain and the sensory epithelium. As such, for a real physical system, we must take the assumptions of the FEP to be only approximations, which hold locally, or approximately, but not for all time and with complete perfection. It remains to be seen, and empirically investigated if possible, the extent to which the mathematical interpretations and logical statements of the FEP remain robust to such slight relaxations of its core assumptions.

While the FEP provides a mathematical interpretation of certain kinds of dynamics in terms of inference, it also, largely, remains to be seen whether such an interpretation is useful for spurring new ideas, questions, and developments within the fields the FEP hopes to influence -- such as neuroscience, cognitive science, and dynamical systems theory. Returning to our anaologies of the least action principle and Noether's theorem, while both of these mathematical results only provide interpretations of known dynamics, by operating at a high level of abstraction they provide powerful capabilities for generalization. For instance the principle of least action allows for dynamics to be derived, via the Euler-Lagrange equations, directly from the high level specification of the action. For instance, the potentially new or counterfactual laws of physics can be derived simply by postulating a given Lagrangian or Hamiltonian and then working through the mathematical machinery of the principle of least action to derive the ensuing dynamics. Additionally, by investigating invariances in the action, one can often understand the kinds of invariances and degrees of freedom that exist in the actually realized dynamics. Similarly, Noether's theorem allows one to play with setting up certain conserved quantities or symmetries a-priori, and then work out precisely the consequences that these entail for the dynamics. 

It is currently unclear to what extent the FEP offers such powerful advantages of abstraction and generalization. This is largely due to the FEP being immature as a field compared to the cornerstones of classical physics, and the majority of the research effort so far has gone into making the theory precise rather than deriving consequences and generalizations from it, but there are some promising initial signs which have just begun to emerge in the literature of the power the FEP perspective offers. From a practical perspective, the FEP appears to offer a number of novel techniques. Firstly, given a desired NESS density, the free energy lemma provides a straightforward way of deriving dynamics which will necessarily reach that density, due to the fact that the variational free energy becomes a Lyapunov function of the system as a whole. This approach has strong potential links to Markov-Chain-Monte-Carlo methods in machine learning and statistics, which aim to approximate an intractable posterior distribution by the time evolution of a Markov process \citep{metropolis1953equation,neal2011MCMC,betancourt2017conceptual,chen2014stochastic,brooks2011handbook}. The FEP provides a new perspective on such systems as fundamentally performing variational Bayesian inference, and may in future be used to develop improved algorithms in this domain, akin to the developments of Hamiltonian \citep{betancourt2013generalizing} and Riemannian MCMC \citep{girolami2011riemann} methods. For instance, there is much potential in the idea of solenoidal flow speeding up convergence to the desired equilibrium density \citep{ma2015complete}. Conversely, the FEP, through the Helmholtz decomposition, may additionally provide tools for \emph{inferring the eventual NESS density} given a specific set of dynamics \citep{ma2015complete,friston2019particularphysics}. This would allow, again, for an analytical or empirical characterisation of the ultimate fate of a system, and allow for characterising different kinds of systems purely by their dynamics far from equilibrium. 

A second strand of potentially directly useful research which has begun to arise from the FEP is empirical and statistical methodologies for defining, computing, and approximating Markov Blankets. This implies the ability to infer the statistical independency structure of the dynamics purely either from analytical knowledge of the dynamics or, alternatively, from purely observed trajectories. There are already two approaches to achieve this in the literature. One which utilizes graph theory in the form of the graph Laplacian to infer nodes of the Markov blanket based on the parents, and children of parents of the largest eigenstates of the Jacobian \citep{palacios2017biological,friston2013life, friston2020parcels}. A second approach directly uses the Hessian of the dynamics to attempt to read off the conditional independency requirements it implies \citep{friston2020parcels}. These approaches may have substantial merit and utility in understanding the effective statistical independency structure of complex dynamical processes, especially questions regarding functional independence in the brain. This strand of research heavily relates to the question of \emph{abstraction} in dynamical systems -- namely, whether complex systems can or cannot be straightforwardly partitioned into independent subsystems which can then be abstracted over. For instance, the ideal would be the ability to, given a complex high-dimensional dynamical system, parse this system into individual `entities' (separated by Markov blankets) which interact with each other according to another set of (hopefully simpler) dynamical rules. This would allow for an automatic procedure to transform a high dimensional complex system into a simpler, low-dimensional approximate system more amenable for analysis and, ultimately understanding \citep{friston2013life,parr2020modules,friston2007parcels}.

\section{Discussion of Assumptions required for the FEP}
Here we provide a general overview and short discussion of every assumption required at each stage of the FEP. Ultimately, the overall picture that emerges is that the FEP requires many assumptions to work, and it is unlikely that all of them can be fulfilled by the kinds of complex self-organizing systems that the FEP ultimately `wants' to be about -- such as biological self organization and, ultimately, brains. However, this is not necessarily overly problematic for the FEP as many of its assumptions may be approximately, or locally true over small enough time periods. This is not necessarily a bad thing -- almost all of the sciences ultimately use simplified models to try to understand their ultimate objects of study in a more tractable way. The FEP is simply continuing that tradition, but if we do this, we need to make explicit the key distinction between the model and the reality or, more memorably, the map and the territory.

The first set of key assumptions that the FEP makes comes through the definition of the kinds of stochastic dynamical systems that it works with. Specifically, we make the following assumptions about the form of the dynamics we deal with,
\begin{itemize}
  \item The system as a whole can be modelled as a Langevin SDE of the form $\frac{dx}{dt} = f(x) + \omega$
  \item The noise $\omega$ is Gaussian with 0 mean and a covariance matrix $2\Gamma$.
  \item The noise is \emph{additive} to the dynamics
  \item $\Gamma$ does not change with time
  \item $\Gamma$ has no state dependence (no heteroscedastic noise)
  \item $\Gamma$ is a diagonal matrix (each state dimension has independent noise)
  \item The dynamics $f(x)$ do not themselves change with time.
\end{itemize}
We also must make the following assumptions about the system as a whole,
\begin{itemize}
  \item The system is \emph{ergodic}, which means that state and time averages coincide or, alternatively, that there must be some probability of ultimately reaching every part of the system from every other part.
  \item The system possesses a well characterized nonequilibrium-steady-state density (NESS), which does not change over time
  \item Once the system reaches this NESS density it cannot escape it -- there is no metastability or multiple competing attractors.
\end{itemize}
These assumptions setup the basic formalism we wish to consider. From here, we then apply the Ao decomposition to rewrite the dynamics in the form of a gradient descent on the log of the potential function with dissipative and solenoidal components $f(x)= (Q - \Gamma)\nabla_x \ln p^*(x)$. To be able to implement this decomposition requires,
\begin{itemize}
  \item The dynamics function $f$ be smooth and differentiable
\end{itemize}

Now, we apply the Markov Blanket conditions at the NESS density,
\begin{itemize}
  \item The state space $x$ can be partitioned into a set of four states -- internal $i$, external $e$, active $a$ and sensory $s$ which, at the NESS density fulfill the following conditional independence relationships:
  $p^*(x) = p^*(\eta | s,a)p^*(\mu | s,a)p^*(s,a)$.
  \item We thus require \emph{all partitions} to be at NESS, including the \emph{external states}. This means that the environment also has to be at steady state, not just the system.
  \item We often assume no solenoidal coupling between internal and sensory states (internal states do not directly act on sensory states -- only the external states do), nor between active and external states (active states drive the external state but are not driven by it). Mathematically this corresponds to $Q_{s, \mu,} = 0$,$Q_{\eta,a} = 0$. 
  \item We may even require that $Q$ be block diagonal -- thus allowing for \emph{no} solenoidal coupling between subsets of the Markov Blanket at all.
\end{itemize}
Given the Markov Blanket conditions hold, we can then begin to move towards the free energy lemma. To begin with, we must first assume,
\begin{itemize}
  \item There is a unique argmax $\bm{\eta}, \bm{\mu}$ exists for both internal and external states for every blanket state $b$.
  \item That there exists a function $\sigma$ which maps from $\bm{\mu}$ to $\bm{\eta}$
  \item That $\sigma$ is invertible
  \item That $\sigma$ is differentiable  
  \item For the particular free energy, we assume that the variational posterior $q(\eta ; \bm{\mu})$ is equal to the true posterior, and thus that the true posterior can be represented by a vector of sufficient statistics ($\bm{\mu}$).
\end{itemize}
These assumptions on $\sigma$ are quite restrictive. A more detailed discussion of what these assumptions require can be found in the next section of this chapter, where every restriction is listed and discussed in some depth.

Finally, to reach the free energy lemma, we must make the following assumptions,
\begin{itemize}
  \item The flow of the sufficient statistic of external states $\bm{\eta}$ follows the same (Ao-decomposition) dynamics as the external states themselves
  \item The variational distribution $q(\eta;\bm{\mu})$ is a Laplace distribution (Gaussian) with a fixed covariance $\Sigma$ as a function of $\bm{\mu}$.
\end{itemize}
This first assumption has come under heavy controversy and is discussed in more detail later. These additional assumptions pertain to the Laplace approximation, but the final assumption here appears to go beyond what is typically required by variational Laplace where, since the conditional distribution is a function of the blanket, one would expect the conditional covariance to be one too.

\subsection{Assumptions on the Form of the Langevin Dynamics}

The FEP formulation makes reasonably strong assumptions about the nature of the dynamics that it models -- restricting them to the form of stochastic dynamics which can be written as a Langevin equation with additive Gaussian noise. While the assumptions on the dynamics function are not that strong, only requiring differentiability and time-independence, the restrictions on the noise in the system are quite severe.

Firstly, it is important to note that using additive white noise, while a common modelling assumption due to its mathematical simplicity, nevertheless imposes some restrictions on the kind of systems that can be modelled -- especially as complex self organizing systems typically evince some kind of colored smooth noise, as well as often power-law noise distributions which are associated with self-organized criticality \citep{ovchinnikov2016introduction}.

However, the further assumptions on the $\Gamma$ covariance matrix -- that it is diagonal, state-independent, and time-independent -- are also strong additional restrictions. Specifically, this means that the noise to every dimension in the system is completely independent of any other dimension, and that the noise is constant at every point throughout the state space and throughout time.
\subsubsection{Ergodicity and the Ao Decomposition}

The Ao decomposition requires both that the dynamics possess a consistent non-equilibrium steady state density (which forms the potential function) and also that the dynamics are ergodic. Additionally, this ergodicity assumption is implicitly used in the Bayesian mechanics, which allows expectations of the surprisal to be taken and interpreted as entropies, and thus to ultimately derive an interpretation of the dynamics in terms of accuracy and complexity. In general, for many biological and self-organizing systems, ergodicity does not hold and such systems typically exhibit substantial amounts of path dependence and irreversibility. This means that on a strict reading, for most systems the FEP desires to model, the ergodicity assumption does not hold. However, it may still be possible to describe ergodicity as holding locally in the small region of the state space around the NESS density and this may be sufficient for an approximate version of the FEP to hold, although the resistance of the FEP to slight perturbations of its assumptions remains unclear.

\subsection{The Markov Blanket Condition}
\subsubsection{Is Information Retained Behind the Blanket? -- The Time Synchronous Markov Blanket Condition}

A potentially substantial problem, which has been raised by Martin Biehl and Nathaniel Virgo, for the FEP is that the Markov Blanket conditions would appear to very strongly imply that the internal states cannot store any more information about the external states than the blanket states. This fact can be derived from a straightforward application of the data processing inequality. Translated into the terminology of biological systems like brains, this would mean that the state of the brain could contain no more information about the environment than the state of the sensory epithelia and actuators at the current time. In effect, this would rule out systems obeying the FEP from exhibiting any sort of long term memory or learning -- clearly a very undesirable side-effect. 

In discussions within the community, there have been many attempts to finesse this apparent difficulty with appeals to notion of nested temporal scales and the local applicability of the FEP. The intuitive argument is that if the Markov Blanket conditions rule out information storage on the macroscale where they apply locally, they may nevertheless allow for the slow accumulation of information over a longer timescale. Effectively, if we can imagine that there are two kinds of variables -- `fast' variables which can change over the a given timescale and `slow' variables which do not. Then, if we can consider the slow variables fixed over some timescale, then we can consider the fast variables to reach a NESS density over that timescale, however over longer timescales, the values of the fast variables can influence the slow variables leading to them changing over time, and thus inducing a different NESS density over among the fast variables. The change in the slow variables can be considered to be learning, and could allow for the accumulation of information over time. This process of timescale separation is directly analogous to the classical distinction between inference (fast) and learning (slow) in machine learning and control theory, and can also be expressed physically in terms of an adiabatic reduction \citep{friston2019particularphysics} which explicitly separates out the dynamics of the system into fast and slow eigenmodes. This construction, however, does require a notion of `approximate' NESS for a timescale which is long enough for the `fast' variables but also short enough for the `slow' variables to appear fixed.

\subsubsection{The Real Constraints on Solenoidal Coupling?}

While the Markov blanket conditions only explicitly disallow solenoidal coupling directly between the internal and external states -- $Q_{\mu,\eta} = 0$, the free energy lemma appears to require a significantly greater reduction of solenoidal coupling. Specifically, the free energy lemma requires that, for a straightforward identification of the surprisal with the free energy, that the form of the dynamics for each marginal subset of states in the partition take the same form as the dynamics of the full set of states $x$. Specifically, this means that \emph{all} solenoidal coupling between the subsets must be suppressed, since if they were not then, by the marginal flow lemma, there would be additional solenoidal coupling terms in Equation \ref{Free_energy_descent}, which would complicate the relation to free energy minimization with additional solenoidal terms. As such, for the free energy lemma, as currently presented, we appear to have the extremely strong condition of the diagonality of $Q$, where each subset in the Markov Blanket is only allowed solenoidal interactions with itself. 

It is important to note that this restriction is significantly stronger than those required just by the Markov Blanket condition, and indeed is stronger even than the flow constraints proposed in \citet{friston2020some}. While this does not entirely rule out any interactions between different subsets of the Markov blanket, it does mean that all interactions have to be mediated through the gradient term, since both the $\Gamma$ and $Q$ matrices are assumed to be diagonal. However, it may be that the additional solenoidal terms in the free energy lemma as a result of non-diagonal Q are not that deleterious to the theory since as these are purely solenoidal terms, they are orthogonal to the flow and do not affect the ultimate minima of the system.

\subsection{Assumptions of the free energy Lemma}
\subsubsection{The $\sigma$ function}

The existence and general properties of the $\sigma$ function have also recently elicited much discussion and debate within the community. Specifically, it is not at all clear that this function exists in the general case, for arbitrary dynamics functions $f$ and conditional NESS distributions $p^*(\eta | b)$ and $p^*(\mu | b)$. In later papers it is assumed to exist under the condition of injectivity between $\bm{\eta}$ and $\bm{\mu}$. In effect, this means that there must be a unique mapping between $\bm{\eta}$ and $\bm{\mu}$ for all blanket states -- i.e. that for every blanket state, if the argmax of the internal states is $\bm{\mu}$, then the argmax of the external states must be $\bm{\eta}$. Additionally, there must be a corresponding (and separate) external argmax for every internal argmax. There may, however, be some external argmaxes with no corresponding internal argmaxes (although the converse condition does not hold). This requires that the dimensionality of the external states be greater than or equal to the dimensionality of the internal states -- which should generally hold for most reasonable systems where we can safely assume that the environment is larger than the system itself. This injectivity condition also guarantees invertibility in the case that the internal and external state spaces are of the same dimension. It is also possible to use the Moore-Penrose pseudoinverse for the case where the external state space is larger, at the cost of the free energy lemma becoming approximate instead of exact.

The differentiability of the $\sigma$ function is a more stringent condition. In many cases this is unlikely to be met, since the argmax functions which the $\sigma$ function maps between are generally nondifferentiable. It remains unclear to what extent differentiable $\sigma$ functions can exist in systems of interest.
\subsubsection{The flow of the Sufficient Statistics $\eta$}

An additional important assumption necessary for the free energy lemma, is that the flow of the sufficient statistics of the external mode follow the same flow as the external states generally. This assumption turns out to be crucial to the free energy lemma which relies heavily in the fact that the flow of the sufficient statistic $\bm{\eta}$ can be written as a gradient descent on the log surprisal -- which can then be expressed in terms of a free energy under the Laplace approximation.

This assumption is also problematic and has been the source of much discussion within the community. The extent to which this assumption is justified remains unclear. Specifically, it appears to rule out the use of arbitrary functions $\xi$ (to be discussed in the next section) to parametrize the external sufficient statistic (although not the internal sufficient statistic). The assumption effectively holds to the extent to which one can describe the sufficient statistic as equal to some external state $\bm{\eta}(b) \approx \eta$, which may occur often for the argmax but not necessarily always. It remains to be seen whether the argmax is in fact the optimal such function -- which is dependent on the blanket, but which can identify a consistent $\eta$ to identify with and thus partake in the same dynamics.


\subsubsection{Potential and Optimal $\xi$ Functions}


While the didactic treatment of the FEP in \citep{friston2019particularphysics, parr2020Markov}, it is assumed that the $\sigma$ function relates the \emph{argmax} of $\eta$ and of $\mu$, this is a simple assumption and is not particularly required by the theory. It only requires that there be \emph{some function} not that it necessarily be an argmax. This means that we could, in theory use an arbitrary function $\bm{\mu}(b) = \xi(b)$ instead of the argmax. Indeed, we might desire to make this function contain \emph{as much information as possible} about the true conditional distribution of the internal states given the external states, so that when the $\sigma$ function maps this to the sufficient statistic of the external density it can be seen as performing inference with the most information possible between the external and internal states. An additional benefit of defining an arbitrary function for $\xi$ instead of using $\xi(b) = argmax \, p(\mu| b)$ is that we can make $\xi$ differentiable, which alleviates much of the difficulty of making $\sigma$ differentiable as well.

While this approach brings many benefits, it also has the drawback of the necessity to choose a suitable function $\xi$ which introduces another degree of freedom into the modelling process. One possible condition is that we could chose the optimal $\xi$ to be the one that contains the most information about the internal state or, alternatively minimizes the KL between the approximate conditional distribution over the internal states parametrized vy $\xi$ and the true conditional over the blanket states. That is, we could define,
\begin{align*}
  \xi^* = \underset{\xi}{argmin} \, \KL[q(\mu; \xi(b)) || p(\mu | b)]
\end{align*}

This would reduce the number of degrees of freedom of $\xi$ and provide a valid modelling target, although the actual computability of this minimization process is potentially a problem, as is whether this objective is actually optimal. Nevertheless, the use of an arbitrary $\xi$ function for the sufficient statistics of the internal states may yet resolve or ameliorate some of the difficulties with the free energy lemma, and is an interesting inroad to begin understanding various relaxations or extensions to the current incarnation of the free energy principle.

\section{Active Inference}

In the previous section, we have covered the very general and abstract form of the FEP, here we elucidate the central process theory that has emerged from the FEP literature in theoretical neuroscience -- Active Inference \citep{friston2012active,friston2009reinforcement}. Active inference is a normative theory of perception, decision-making, and learning which ties these three core cognitive processes together under the general paradigm of variational inference via the minimization of the variational free energy \citep{friston2015active,friston2017process}. Specifically, it views all of these processes as emerging out of a central imperative of the system to minimize its free energy over time, and thus perform inference. Perception can be quite clearly stated as an inference problem of inferring the hidden states and causes of the world from sensory observations. Learning too can be interpreted as inference over the parameters of the generative model, which takes place on a slower timescale than perceptual inference. Finally, action selection, decision-making, and planning can be described as inference on policies over trajectories into the future. While there are numerous methods to perform this inference, active inference chooses to minimize an expected free energy functional which encodes goals in terms of a prior over future states \citep{da2020active}.

Active inference has been applied widely and productively in theoretical neuroscience, and active inference models have been proposed for planning and navigation \citep{kaplan2018planning}, saccadic eye movements \citep{parr2017uncertainty,parr2018active,parr2018anatomy} and visual foraging \citep{parr2019computational,heins2020deep}, and general planning problems in reinforcement learning and machine learning \citep{ueltzhoffer_deep_2018,tschantz2020reinforcement,tschantz2020control,millidge_deep_2019}. Additionally, by simulating various lesions or incorrect update rules in the active inference scheme, one may obtain interesting behavioural anomalies or shortfalls which one can then analogize to known psychiatric disorders -- an approach known as computational psychiatry \citep{parr2019computational,cullen2018active}. By drawing correspondences between disorders of behaviour in tractable artificial systems and the psychiatric symptoms of disorders in humans or other animals, one may be able to shed new light upon the actual mechanistic underpinnings of such disorders, which may lead to novel hypotheses, experimental protocols and, ultimately, treatments. Computational psychiatric approaches using active inference have pioneered statistical, mechanistic models of impulsivity \citep{mirza2019impulsivity}, visual neglect \citep{parr2018computational}, autism \citep{lawson2014aberrant}, schizophrenia \citep{adams2012smooth} substance use disorder and addiction \citep{schwartenbeck2015optimal}, and rumination \citep{hesp2020sophisticated}. Finally, the epistemic imperatives that arise from the minimization of the expected free energy functional have given rise to a number of simulation studies applying the approach to exploration tasks \citep{schwartenbeck2013exploration,friston2015active,friston2017curiosity}, visual foraging and other information-seeking saccade behaviour \citep{heins2020deep,parr2017active}, and exploration in complex sparse-reward environments from reinforcement learning \citep{tschantz2020reinforcement}, which will be significantly expanded upon in later chapters of this thesis.

Broadly, there are two main classes of active inference models in the literature -- continuous-time, continuous-state models, which are an extension of predictive coding models of brain function \citep{friston2009reinforcement,friston2010action,pio2016active,baltieri2017active,baltieri2019pid,millidge2019combining}, and discrete-time discrete-state-space active inference models which have been heavily developed in the literature in the past decade, and perhaps now form the main theory of active inference applied to the brain \citep{da2020active,friston2017process}. All of these models, however, ultimately are derived from the same mathematical apparatus. As such, an advantage of active inference for modelling behaviour is that due to its developed and shared mathematical apparatus, different models are specified simply through the generative model and variational distribution, and can be directly compared through Bayesian model comparison techniques. This can be used to fit active inference models to empirical behavioural data in a straightforward fashion. Continuous time, continuous state based models are explained in depth in Chapter 3, where I discuss my work with these models in the context of predictive coding. Here, we present an introduction to discrete-state-space active inference which shall form the basis of my work in Chapters 4 and 5 of the thesis. 

Here we introduce a more standard notation which shall be used for the rest of the thesis. We consider our agent to be situated in a Partially Observed Markov Decision Process (POMDP) \citep{sutton1990integrated,kaelbling1996reinforcement}. The agent is given observations $o$, and must infer the hidden states of the world $x$ that gave rise to the observations. The agent may additionally possess models with parameters $\theta$, which can also be optimized. Finally, action selection consists of inferring actions $a$, or policies $\pi = [a_0, a_1, \dots a_N]$ which are simply sequences of actions in order to achieve some desired goal. Active inference is based around the fundamental imperative of minimizing the variational free energy (VFE). We may recall from Equation \ref{VFE_decomp} that the VFE consists of the KL divergence between a variational distribution $q(x | o)$ and a generative model $p(o,x)$. 

\begin{align*}
\mathcal{F}(o) = \KL[q(x | o) || p(o,x)] \numberthis
\end{align*}

If we additionally want to infer the parameters $\theta$, we can extend the generative model and variational density to include a distribution over the parameters -- this provides a fully Bayesian treatment of parameters in contrast to many machine learning schemes which treat them effectively as point distributions,
\begin{align*}
\mathcal{F}(o) = \KL[q(x, \theta | o) || p(o,x,\theta)] \numberthis
\end{align*}

In order to implement a specific active inference scheme, the key thing to specify is the nature of the generative model, and the nature of the variational distribution. These two distributions suffice to completely specify the model, and with these distributions set, the processes of learning, inference, and action selection can be handled by the standard mathematical apparatus of the theory. Specifically, given a specific variational distribution and a generative model, we can implement perception as a minimization of the VFE with respect to the variational distribution with respect to the hidden states, and we can implement learning as the minimization of the VFE with respect to the parameters
\begin{align*}
\text{Perception: } &\underset{q(x | o)}{argmin} \, \mathcal{F}(o) \\
\text{Learning: } &\underset{q(\theta | x,o)}{argmin} \, \mathcal{F}(o) \numberthis
\end{align*}
There are then two separate ways to implement action. The most straightforward approach, which is utilized in the continuous time version of active inference is to similarly implement action as a gradient descent on the VFE with respect to action,
\begin{align*}
\text{Action: } &\underset{a}{argmin} \, \mathcal{F}(o(a)) \numberthis
\end{align*}
Where we have made the implicit dependence of the observations, and hence the VFE on action explicit, which makes such a minimization non-trivial. A second approach, which is typically used in the discrete-state-space paradigm is to assume a specific functional form for the variational posterior over policies -- that of a softmax distribution over the Expected Free Energy (EFE) $\mathcal{G}(o,x)$ \citep{friston2015active},
\begin{align*}
\label{EFE_softmax}
\text{Action (discrete-state-space): } Q(\pi) = \sigma(-\mathcal{G}(o,x)) \numberthis
\end{align*}
where $\sigma$ is a softmax function. Effectively, what this states is that the optimal policy is a softmax distributions over the path-integrals of the EFE into the future. Effectively, the optimal distribution over policies is simply one that selects policies in proportion with the exponentiated EFE resulting from executing that policy in the future. This means that the policy with the greatest EFE is most likely, while policies with lesser EFE are exponentially less likely to be selected based on the difference between their EFE and that of the best policy. 

Discrete state-space active inference therefore optimizes two complementary objective functions. Optimizing the variational free energy, which is used for perception and learning, ensures that the agent learns an accurate world model, and is able to accurately infer the hidden states of the world from current observations. The second objective, the expected free energy, is used to score potential plans or action policies, to allow the agent to make decisions which are adaptive relative to its goals. To successfully predict and infer with trajectories in the future requires a highly developed and accurate world-model, able to make accurate multi-step predictions of the consequences of action. Such a world model is provided by the minimization of the VFE in the inference and learning steps. This separation of inference and action selection into two separate objectives -- the VFE and the EFE -- introduces a measure of complexity into the theory which may or may not be unavoidable. In Chapter 5, we focus especially on this question and investigate the nature of the EFE, and whether all facets of inference, learning, and action selection can be subsumed under a \emph{single} unified objective.

\subsection{Discrete State-space models and Perception}

The core component of the discrete-state-space model is the discrete generative models and variational densities it is based upon. Specifically, we split the generative model into a likelihood and prior distribution $p(o,x) = p(o|x)p(x)$, and then represent each of these distributions as a categorical distribution,
\begin{align*}
p(o,x ) &= p(o|x)p(x) \\
p(o|x) &= Cat(o; \hat{o}) = \bm{A} \\
p(x) &= Cat(x; \hat{x}) = \bm{B} \numberthis
\end{align*}


A categorical distribution is one that simply directly assigns some probability value to every possible discrete contingency. The parameters $\hat{o}$ and $\hat{x}$ of these distributions are simply these probability values, which can be represented straightforwardly in terms of matrices. $\bm{A} \in \mathcal{R}^{O} \times \mathcal{R}^{X}$ is simply a normalized matrix of probabilities representing the likelihood contingencies -- that is, for every hidden state $x$, what is the probability of each potential outcome. Similarly, the transition matrix $\bm{B} \in \mathcal{R}^{X} \times \mathcal{R}^{X}$ is a matrix of probabilities representing the probability of transitioning from any one hidden state to any other hidden state. Similarly, we define our variational distribution $q(x | o = o_\mu;\hat{x}_q) \in \mathcal{R}^{X}$ to be a categorical distribution of the probability of each discrete hidden state as a function of the observed observation $o_\mu$. With our variational and generative model set, we can explicitly write out and evaluate the VFE,
\begin{align*}
\mathcal{F} &= \KL[q(x | o) || p(o,x)] \\
&= \E{q(x | o; \hat{x}_q)}[\ln q(x | o; \hat{x}_q)] - \E_{q(x | o; \hat{x}_q)}[\ln p(o|x; \hat{o})] - \E_{q(x | o; \hat{x}_q)}[\ln p(x; \hat{x})] \\
&= \hat{x}_q \ln \hat{x}_q - \hat{x}_q \ln \bm{A} - \hat{x}_q \ln \bm{B} \numberthis
\end{align*}
Where we have simply explicitly written out the variational free energy in terms of the parameters of the categorical distributions. The expectation operator $\E[]$ can be computed as a simple dot product instead of an integral due to the discrete state space. Importantly, because both our variational density and generative models are categorical distributions, we can derive an analytical expression for the minimum of $\mathcal{F}$ with respect to the variational parameters $\hat{x}_q$, allowing for an exact Bayes-optimal single-step update for perception,
\begin{align*}
\frac{\partial \mathcal{F}}{\partial \hat{x}_q} &= \frac{\partial}{\partial \hat{x}_q}[\hat{x}_q \ln \hat{x}_q - \hat{x}_q \ln \bm{A} - \hat{x}_q \ln \bm{B}]\\
&= \ln \hat{x}_q + \bm{1} - \ln \bm{A} - \ln \bm{B} \\
& \frac{\partial \mathcal{F}}{\partial \hat{x}_q} = 0 \implies \hat{x}_q^* = \sigma(-\ln \bm{A} - \ln \bm{B}) \numberthis
\end{align*}

Learning can be approached similarly, by placing suitable hyperpriors (typically dirichlet \citep{schwartenbeck_computational_2019}) upon the parameters of the $\bm{A}$ and $\bm{B}$ matrices and then minimizing the VFE with respect to these parameters. For more information on how learning is implemented see \citet{da2020active,friston2017process}.

\subsection{Action Selection and the Expected Free Energy}


Action selection is then handled via the variational posterior being equal to the softmaxed path integral of the EFE through time (Equation \ref{EFE_softmax}). Typically, in small discrete state spaces, this path integral can be computed exactly, by simply computing the EFE for every single policy and every single possible trajectory through the state-space. Unfortunately, this approach scales exponentially in the time horizon, and is thus not suitable for long, open-ended tasks, although it remains a highly effective method for simulating short tasks, such as single trials in a psychophysical, or simple decision-making, paradigm \citep{friston_active_2015,schwartenbeck2015optimal,friston2020sophisticated}. Various methods have been proposed to handle this exponential complexity. One commonly proposed method is to simply prune potential trajectories which become too unlikely (i.e. have too low an EFE) to be worth considering further. Typically, such methods, however, do not reduce the algorithm to a smaller (polynomial) complexity class, but instead simply reduce the exponential coefficient which allows the method to scale to slightly larger tasks but does not remove the fundamental exponential complexity of the algorithm. Other approaches involve approximating the path integral with either bootstrapping value-function methods \citep{millidge_deep_2019}, which take advantage of the recursive temporal decomposition of the EFE, or alternatively Monte-Carlo techniques which approximate the EFE through a random sampling of trajectories, which corresponds to classical model-predictive control algorithms \citep{kappen2012optimal}. An additional consideration in the EFE is the need to specify a desired or goal state for the action selection mechanism to achieve. This can be considered to be a probabilistic description of rewards in reinforcement learning and psychology, or of utility in economics. Mathematically, this specification is achieved by defining a \emph{biased} generative model $\tilde{p}(o,x)$ which contains a desired distribution $\tilde{p}$ which encodes the rewards or utility as effectively priors in the inference procedure.

Since action selection is dependent entirely on the path-integral of the EFE, the properties of the EFE functional essentially determines the kind of behaviour that finding trajectories that minimize the EFE will induce. Here, we showcase two decompositions of the EFE, and discuss its intrinsic exploratory drive.

\begin{align*}
\mathcal{G}(o,x) &= \E_{q(o,x)}[\ln q(x) - \ln \tilde{p}(o,x)] \\
&= \underbrace{\E_{q(o,x)}[\ln p(o|x)]}_{\text{Ambiguity}} + \underbrace{\KL[q(x) || \tilde{p}(x)]}_{\text{Risk}} \\
&= \underbrace{\E_{q(o,x)}[\ln \tilde{p}(o)]}_{\text{Extrinsic Value}} -\underbrace{\E_{q(o)}\KL[q(x | o) || q(x)]}_{\text{Information Gain}} + \underbrace{\E_{q(o)}\KL[q(x |o) || p(x|o)]}_{\text{Posterior Divergence}} \numberthis
\end{align*}
The first decomposition into \emph{risk} and \emph{ambiguity} obtains when the goal distribution is specified in terms of the hidden states of the world $\tilde{p}(x)$. In this case, EFE minimization can be thought of as directly trying to match the expected states of the world to the desired states, and thus achieving one's goals, while simultaneously trying to minimize the \emph{ambiguity} of the observations one receives. The second decomposition into \emph{extrinsic} and \emph{intrinsic} value (information gain) occurs when the goal distribution is specified in terms of the observations $\tilde{p}(o)$. The extrinsic value term can be thought of as the expected reward or expected utility, since it is the average amount of reward expected under the predicted observation distribution. Most interestingly in this decomposition, however, is the information gain term which encourages the agent to maximize the divergence between the variational prior and the variational posterior. Effectively this term encourages the agent to seek out information in the environment which will maximally update its beliefs about the world. In effect, this term encourages a specific kind of information-seeking exploration, where agents that minimize the EFE are effectively driven to seek out and integrate resolvable uncertainty about the world into their world model.

This intriguing property of active inference agents which minimize the EFE has been extensively investigated in the literature -- from simple tasks such as the T-maze which require deciding whether to gain information by seeking out a cue or not \citep{friston2015active}, to planning visual saccades in a way which maximizes the information about the scene gained \citep{parr2017uncertainty,heins2020deep} and to using directed exploration to develop highly sample efficient and powerful general reinforcement learning algorithms for sparse-reward environments \citep{millidge2019deep}. Moreover, the fact that the EFE naturally gives rise to an information-seeking exploration term offers a promising and fascinating avenue for resolving the exploration-exploitation tradeoff and deriving optimal exploration strategies directly from variational Bayesian inference algorithms.

\section{Discussion}

In this chapter, we have already covered a substantial amount of ground. We have reviewed the core tenets of the free energy principle, provided a mathematically detailed walk-through of the core results, and discussed their philosophical implications and meaning. We have additionally provided a short review of a core process theory -- discrete state space active inference -- which we will fundamentally build upon in various ways in the rest of this thesis. Chapter 3 will focus on applying the free energy principle to perception -- and will focus on the implementation and extension of the process theory of predictive coding. Chapter 4 will focus on merging active inference as presented here with modern deep reinforcement learning methods to allow active inference approaches, which are currently bottlenecked due to their discrete explicit tabular representations and the exponential complexity of the action selection algorithm, to be extended to challenging machine learning problems. Chapter 5 will focus especially on the Expected Free Energy term and will seek to unravel the origin of its information seeking properties and in the process will reveal deep connections between active inference and other variational Bayesian approaches to action such as control as inference, as well as revealing a substantially richer landscape of potential variational functionals for control than has been previously realized.

%% file: chap3.tex
\chapter{Predictive Coding}

\section{Introduction}

In this chapter, we consider the application of the free energy principle to perception. Here, we focus entirely on visual perception and the process theory of predictive coding \citep{friston2003learning,friston2005theory,bastos2012canonical,buckley2017free,spratling2017review}. This chapter is organized into four relatively independent sections which each present a separate piece of work, which extends or contributes to the theory or practice of predictive coding. The general theme of our work aims to scale up predictive coding to reach the levels of performance achieved by machine learning, as well as to understand the potential biological plausibility of the theory. Both of these are important for understanding the potential predictive coding has as a general theory of cortical function since, if it is actually implemented in the brain, it must meet both bars of extremely high scalability (since the brain effortlessly handles perception and inference with extremely detailed and complex inputs, as well as constructing extremely powerful and general representations), as well as the biological plausibility necessary to allow the dynamics prescribed by predictive coding to be implemented by neural circuitry.

We begin with a mathematical introduction to predictive coding and its dynamics. This is followed with the presentation of our work where we experiment with implementing larger scale predictive coding networks than previously in the literature, and validate their performance and capabilities on benchmark machine learning datasets -- thus demonstrating that predictive coding as a theory can be scaled up to the standards of modern deep learning. We also experiment with dynamical predictive coding networks using generalized coordinates (as introduced \citep{friston2008DEM}) as well as combining both hierarchical and dynamical predictive coding networks, although these implementations are only tested on relatively small toy tasks and scaling these up to larger and much more challenging dynamical tasks, such as video prediction, remains an important avenue for future work.

In the second section, we focus on understanding how predictive coding can be written as a filtering algorithm -- and thus can be applied productively to fundamentally dynamical instead of static stimuli. We demonstrate precisely how predictive coding is related to Kalman filtering -- a ubiquitous and extremely successful Bayesian filtering algorithm \citep{kalman1960contributions,kalman1961new} -- and also show how predictive coding can extend this algorithm to allow for the online learning of the parameters of the generative model (as well as inference of the states) -- a capability which is not usually achieved with Kalman filtering alone. We validate the performance of this algorithm on simple filtering tasks.

Fourthly, we investigate the biological plausibility of predictive coding, show how the standard model possesses three key implausibilities -- weight transport, nonlinear derivatives, and one-to-one error unit connectivity, and show how each can be overcome with biologically plausible additions to the algorithm without causing much of a degradation in the classification performance of the algorithm.

\section{Predictive Coding}

Predictive coding is an influential theory in computational and cognitive neuroscience, which proposes a potential unifying theory of cortical function \citep{friston2003learning,friston2005theory,rao1999predictive,friston2010free,clark2013whatever,seth2014cybernetic} -- namely that the core function of the brain is simply to minimize prediction error, where the prediction errors denote mismatches between predicted input and the input actually received. This minimization can be achieved in multiple ways: through immediate inference about the hidden states of the world, which can explain perception \citep{beal2003variational}, through updating a global world-model to make better predictions, which could explain learning \citep{friston2003learning,neal1998view}, and finally through action to sample sensory data from the world that conforms to the predictions \citep{friston2009reinforcement}, which potentially provides an account of adaptive behaviour and control. Prediction error minimization can also be influenced by modulating the \emph{precision} (or inverse variance) of sensory signals, which may shed light on the neural implementation of attention mechanisms \citep{feldman2010attention,kanai2015cerebral}. Predictive coding boasts an extremely developed and principled mathematical framework, which formulates it as a variational inference algorithm \citep{blei2017variational,ghahramani2000graphical, jordan1998introduction}, alongside many empirically tested computational models with close links to machine learning \citep{beal2003variational,dayan1995helmholtz,hinton1994autoencoders}, which address how predictive coding can be used to solve challenging perceptual inference and learning tasks similar to those faced by the brain. Moreover, predictive coding also has been translated into neurobiologically plausible microcircuit process theories \citep{bastos2012canonical,shipp2016neural,shipp2013reflections} which are increasingly supported by neurobiological evidence \citep{walsh2020evaluating}. Predictive coding as a theory is also supported by a large amount of empirical evidence and offers a single mechanism that accounts for diverse perceptual and neurobiological phenomena such as end-stopping \citep{rao1999predictive}, bistable perception \citep{hohwy2008predictive,weilnhammer2017predictive}, repetition suppression \citep{auksztulewicz2016repetition}, illusory motions \citep{lotter2016deep,watanabe2018illusory}, and attentional modulation of neural activity \citep{feldman2010attention,kanai2015cerebral}. As such, and perhaps uniquely among neuroscientific theories, predictive coding encompasses all three layers of Marr's hierarchy by providing a well-characterised and empirically supported view of `what the brain is doing' at the computational, algorithmic, and implementational level \citep{marr1982vision}.

The core intuition behind predictive coding is that the brain is composed of a hierarchy of layers, which each make predictions about the activity of the layers below \citep{clark2015surfing,friston2008hierarchical}. These descending downward predictions at each level are compared with the activity and inputs of each layer to form prediction errors -- which is the information in each layer which could not be successfully predicted. These prediction errors are then fed upwards to serve as inputs to higher levels, which can can then be utilized to reduce their own prediction error. The idea is that, over time, the hierarchy of layers instantiates a range of predictions at multiple scales, from the fine details in local variations of sensory data at low levels, to global invariant properties of the causes of sensory data (e.g., objects, scenes) at higher or deeper levels.\footnote{This pattern is widely seen in the brain \citep{hubel1962receptive,grill2004human} and also in deep (convolutional) neural networks \citep{olah2017feature}, but it is unclear whether this pattern also holds for deep predictive coding networks, primarily due to the relatively few instances of deep convolutional predictive coding networks in the literature so far.}. Predictive coding theory claims that the goal of the brain as a whole, in some sense, is to minimize these prediction errors, and in the process of doing so performs both perceptual inference and learning. Both of these processes can be operationalized via the minimization of prediction error, first through the optimization of neuronal firing rates on a fast timescale, and then the optimization of synaptic weights on a slow timescale \citep{friston2008hierarchical}. Predictive coding proposes that using a simple unsupervised loss function, such as simply attempting to predict incoming sensory data, is sufficient to develop complex, general, and hierarchically rich representations of the world in the brain, an argument which has found recent support in the impressive successes of modern machine learning models trained on unsupervised predictive or autoregressive objectives \citep{radford2019language,kaplan2020scaling,brown2020language}. Moreover, the fact that, in these machine learning models, errors are computed at every layer means that each layer only has to focus on minimizing local errors rather than a global loss. This property potentially enables predictive coding to learn in a biologically plausible way using only local and Hebbian learning rules \citep{whittington2017approximation,millidge2020predictive,friston2003learning}.

While originating from many varied intellectual currents, including the speculations of Helmholtz \citep{helmholtz1866concerning}, ideas in information theory \citep{shannon1948mathematical} and Barlow's minimum redundancy principle \citep{barlow1961possible}, as well as ideas from cybernetics \citep{wiener2019cybernetics,seth2014cybernetic} and early work on machine learning \citep{jordan1998introduction,hinton1994autoencoders}, modern predictive coding can be best described as a variational inference algorithm \citep{beal2003variational} on the hidden causes of sensory sensations, under Gaussian and Laplace assumptions. Variational inference is a method of approximate Bayesian inference, arising in statistical physics \citep{feynman1998statistical}, which turns an intractable inference problem into a potentially tractable optimization problem. In brief, we postulate a variational density $q$, under the control of the modeller, and try to minimize the divergence between this variational density and the true posterior. Since this divergence is not tractable either (since it contains the true posterior), instead we optimize a tractable bound on this divergence known as the variational free energy $\mathcal{F}$ which measures the expected difference between the logs of the variational posterior and the generative model. To make this concrete, suppose we have observations (or data) $o$; we wish to infer hidden, or latent, states of the world $x$, with a generative model $p(o,x)$ and a variational density $q(x | o;\phi)$ with parameters $\phi$. Then, we can write the variational free energy $\mathcal{F}$ as,
\begin{flalign*}
 \KL[q(x | o; \phi) || p(x | o)] &= \KL[q(x | o; \phi) || \frac{p(o,x)}{p(o)}] \\ 
 &= \KL[q(x | o; \phi) || p(o,x)] + \mathbb{E}_{q(x | o ;\phi)}[\ln p(o)] \\
 &= \KL[q(x | o; \phi) || p(o,x)] + \ln p(o) \\
 &\leq \KL[q(x | o; \phi) || p(o,x)] = \mathcal{F} \numberthis
\end{flalign*}

To derive a specific variational inference algorithm -- such as predictive coding -- we must explicitly specify the forms of the variational posterior and the generative model. In the case of predictive coding, we assume a Gaussian form for the generative model $p(o,x ; \theta) = p(o | x;\theta)p(x;\theta) = \mathcal{N}(o; f(x;\theta_1), \Sigma_{2})\mathcal{N}(x; g(\bar{\mu};\theta_2), \Sigma_1)$ where we first partition the generative model into likelihood $p(o|x;\theta)$ and prior $p(x;\theta)$ terms. The mean of the likelihood Gaussian distribution is assumed to be some function $f$ of the hidden states $x$, which can be parametrized with parameters $\theta$, while the mean of the prior Gaussian distribution is set to some arbitrary function $g$ of the prior mean $\bar{\mu}$. We also assume that the variational posterior is a dirac-delta (or point mass) distribution $q(x | o;\phi) = \delta(x - \mu)$ with a center $\phi = \mu$\footnote{In previous works, predictive coding has typically been derived by assuming a Gaussian variational posterior under the Laplace approximation. This approximation effectively allows you to ignore the variance of the Gaussian and concentrate only on the mean. This procedure is effectively identical to the dirac-delta definition made here, and results in the same update scheme. However, the derivation using the Laplace approximation is much more involved so, for simplicity, here we use the Dirac delta definition. See Appendix C, or \citet{buckley2017free} for a detailed walkthrough of the Laplace derivation}. 

Given these definitions of the variational posterior and the generative model, we can write down the concrete form of the variational free energy to be optimized. We first decompose the variational free energy into an `\emph{Energy}' and an `\emph{Entropy}' term
\begin{flalign*}
 \mathcal{F} &= \KL[q(x | o;\phi) || p(o,x ; \theta)] \\
 &= \underbrace{\mathbb{E}_{q(x | o;\phi)}[\ln q(x | o;\phi)]}_{\text{Entropy}} - \underbrace{\mathbb{E}_{q(x | o;\phi)}[\ln p(o,x ; \theta)]}_{\text{Energy}} \numberthis
\end{flalign*}

where, since the entropy of the dirac-delta distribution is 0 (it is a point mass distribution), we can ignore the entropy term and focus solely on writing out the energy.
\begin{flalign*}
 \underbrace{\mathbb{E}_{q(x | o;\phi)}[\ln p(o,x ; \theta)]}_{\text{Energy}} &= \mathbb{E}_{ \delta ( x - \mu)}[\ln \big( \mathcal{N}(o; f(\theta_1 x), \Sigma_{1})\mathcal{N}(x; g(\theta_2 \bar{\mu}), \Sigma_2)\big)] \\
 &= \ln \mathcal{N}(o; f(\theta_1 \mu), \Sigma_{2}) + \ln \mathcal{N}(\mu; g(\theta_2 \bar{\mu}), \Sigma_1) \\
 &= \frac{(o-f(\theta_1 \mu)^2}{\Sigma_2} - \ln 2 \pi \Sigma_2 + \frac{(\mu - g(\theta_2 \bar{\mu}))^2}{\Sigma_1} - \ln 2 \pi \Sigma_1 \\
 &= \Sigma_2^{-1} \epsilon_o^2 + \Sigma_1^{-1}\epsilon_x^2 - \ln 4\pi \Sigma_1 \Sigma_2 \numberthis
\end{flalign*}

where we define the prediction errors $\epsilon_o = o - f(\theta_1 \mu)$ and $\epsilon_x = \mu - g(\theta_2 \bar{\mu})$. We thus see that the energy term, and thus the variational free energy, is simply the sum of two squared prediction error terms, weighted by their inverse variances, plus some additional log variance terms. 

Finally, to derive the predictive coding update rules, we must make one additional assumption -- that the variational free energy is optimized using the method of gradient descent such that,
\begin{flalign*}
 \frac{d \mu}{dt} = -\frac{\partial \mathcal{F}}{\partial \mu} \numberthis
\end{flalign*}
Given this, we can derive dynamics for all variables of interest ($\mu, \theta_1, \theta_2)$ by taking derivatives of the variational free energy $\mathcal{F}$. The update rules are as follows
\begin{flalign*}
 \label{PC_equations}
 \frac{d\mu}{dt} &= -\frac{\partial \mathcal{F}}{\partial \mu} = \Sigma_2^{-1} \epsilon_o \frac{\partial f}{\partial \mu} \theta^T - \Sigma^{-1}_1 \epsilon_x \\
 \frac{d\theta_1}{dt} &= \frac{\partial \mathcal{F}}{\partial \theta_1} = -\Sigma_2^{-1} \epsilon_o \frac{\partial f}{\partial \theta_1} \mu^T\\
 \frac{d\theta_2}{dt} &= \frac{\partial \mathcal{F}}{\partial \theta_2} = -\Sigma_1^{-1} \epsilon_x \frac{\partial g}{\partial \theta_2} \bar{\mu}^T \numberthis
\end{flalign*}
Furthermore while it is possible to run the dynamics for the $\mu$ and the $\theta$ simultaneously, it is often better to treat predictive coding as an EM algorithm \citep{dempster1977maximum} and alternate the updates. Empirically, it is typically best to run the optimization of the $\mu$s, with fixed $\theta$ until close to convergence, and then run the dynamics on the $\theta$ with fixed $\mu$ for a short while \citep{friston2005theory}. This implicitly enforces a separation of timescales upon the model where the $\mu$ are seen as dynamical variables which change quickly while the $\theta$ are slowly-changing parameters. For instance, the $\mu$s are typically interpreted as rapidly changing neural firing rates, while the $\theta$s are the slowly changing synaptic weight values \citep{rao1999predictive,friston2005theory}.

Finally, we can see how this derivation of predictive coding maps onto putative psychological processes of perception and learning. The updates of the $\mu$ can be interpreted as a process of perception, since the $\mu$ is meant to correspond to the true latent state of the environment generating the $o$ observations. By contrast, the dynamics of the $\theta$ can be thought of as corresponding to learning, since these $\theta$ effectively define the mapping between the latent state $\mu$ and the observations $o$. 

\section{Hierarchical predictive coding}
Thus far, we have only derived a predictive coding scheme with a single level of latent variables $\mu_1$. However, the expressivity of such a scheme is limited. The success of deep neural networks in machine learning have demonstrated that having hierarchical sets of latent variables is key to allowing methods to learn abstractions and to handle intrinsically hierarchical dependencies of the sort humans intuitively perceive \citep{krizhevsky2012imagenet,hinton2012neural}. Predictive coding can be straightforwardly extended to handle hierarchical dynamics of arbitrary depth. This is done through postulating multiple layers of latent variables $x1 \dots x_L$ and then defining the generative model as follows,
\begin{flalign*}
 p(x_0 \dots x_L) = p(x_L)\prod_{l=0}^{L-1} p(x_{l} | x_{l+1}) \numberthis
\end{flalign*}
where $p(x_{l} | x_{l+1}) = \mathcal{N}(x_l; f_l(\theta_{l+1} x_{l+1}, \Sigma_l)$ and the final layer $p(x_L) = \mathcal{N}(x_L | \bar{x_L}, \Sigma_L)$ has an arbitrary prior $\bar{x_L}$ and the latent variable at the bottom of the hierarchy is set to the observation actually received $x_0 = o$. Similarly, we define a separate variational posterior for each layer $q(x_{1:L} | o) = \prod_{l=1}^L \delta(x_l - \mu_l)$, then the variational free energy can be written as a sum of the prediction errors at each layer,
\begin{flalign*}
 \mathcal{F} = \sum_{l=1}^L \Sigma_l^{-1} \epsilon_l^2 + \ln 2\pi \Sigma_l \numberthis
\end{flalign*}
where $\epsilon_l$ = $\mu_l - f_l(\theta_{l+1} \mu_{l+1})$. Given that the free energy divides nicely into the sum of layer-wise prediction errors, it comes as no surprise that the dynamics of the $\mu$ and the $\theta$ are similarly separable across layers.
\begin{flalign*}
 \label{PC_hierarchical_mu}
 \frac{d\mu_l}{dt} &= -\frac{\partial \mathcal{F}}{\partial \mu_l} = \Sigma_{l-1}^{-1} \epsilon_{l-1} \frac{\partial f_{l-1}}{\partial \mu_l} \theta_{l}^T - \Sigma^{-1}_l \epsilon_l \numberthis
\end{flalign*}
\newline 
\begin{flalign*}
 \label{PC_hierarchical_theta}
 \frac{d\theta_l}{dt} &= -\frac{\partial \mathcal{F}}{\partial \theta_l} = \Sigma_l^{-1} \epsilon_{l-1} \frac{\partial f_{l-1}}{\partial \theta_l} \mu_l \numberthis
\end{flalign*}

We see that the dynamics for the variational means $\mu$ depend only on the prediction errors at their layer and the prediction errors on the level below. Intuitively, we can think of the $\mu$s as trying to find a compromise between causing error by deviating from the prediction from the layer above, and adjusting their own prediction to resolve error at the layer below. In a neurally-implemented hierarchical predictive coding network, \emph{prediction errors} would be the only information transmitted `upwards' from sensory data towards latent representations, while predictions would be transmitted `downwards'. Crucially for conceptual readings of predictive coding, this means that sensory data is \emph{not} transmitted directly up through the hierarchy, as is assumed in much of perceptual neuroscience. The dynamics for the $\mu$s are also fairly biologically plausible as they are effectively just the sum of the precision-weighted prediction errors from the $\mu$s own layer and the layer below, the prediction errors from below being transmitted back upwards through the synaptic weights $\theta^T$ and weighted with the gradient of the activation function $f_l$. 

Importantly, the dynamics for the synaptic weights is entirely local, needing only the prediction error from the layer below and the current $\mu$ at the given layer. The dynamics thus becomes a Hebbian rule between the presynaptic $\epsilon_{l-1}$ and postsynaptic $\mu_l$, weighted by the gradient of the activation function.

Based upon our previous work \citep{millidge2019implementing}, we present empirical evaluations and demonstrations of the expressive power of hierarchical predictive coding networks on standard machine learning benchmarks with learnable generative models. While some previous work has implemented hierarchical predictive coding models and tested them on `blind deconvolution' of simulated ERP data \citep{friston2008hierarchical,friston2005theory}, we present a key demonstration of predictive coding networks within a machine learning paradigm, and with a completely learnt generative model.

First, we tested the potential of predictive coding networks as autoencoders \citep{hinton1994autoencoders} from machine learning. Here, the goal of the network is simply to reconstruct its input data. In theory, this can be done trivially be learning the identity mapping, so to make it difficult we create an information bottleneck \citep{tishby2000information} in the hidden layers, such that the input is compressed to a much smaller latent code, which must then be decompressed to successfully reconstruct the image. Autoencoders of this type are widely used in machine learning and a probabilistic variant -- variational autoencoders \citep{kingma_auto-encoding_2013} are still state of the art at many image generation tasks \citep{child2020very}. Here we demonstrate that predictive coding networks can also function as powerful autoencoders. We first test the potential of predictive coding on the MNIST dataset -- a standard machine learning benchmark dataset of 60,000 28x28 grayscale handwritten digits. We utilized a three layer predictive coding network, with an input and output dimensionality of 784 (the size of a flattened vector of the MNIST digit), and a latent dimensionality of 20, meaning that the network had to learn to compress a 784 dimensional manifold into a 20 dimensional latent space. We trained the predictive coding network according to Equations \ref{PC_equations} with a batch size of 64 and a learning rate of 0.01. A sigmoid nonlinearity was used on the input and latent layers. We trained the networks for 100 epochs. Each epoch consisted of updating the $\mu$s using Equation \ref{PC_hierarchical_mu} for 100 steps, and then updating the weights $\theta$ using Equation \ref{PC_hierarchical_theta} once. The model was able to recreate MNIST digits successfully, as shown in the example reconstructions below:

\begin{figure}[H]
\centering
\begin{subfigure}{.3\linewidth}
 \centering
 \includegraphics[scale=0.4]{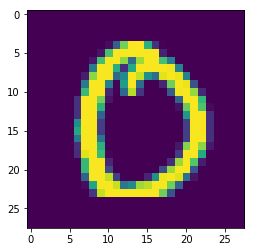}
\end{subfigure}
 \hfill
\begin{subfigure}{.3\linewidth}
 \centering
 \includegraphics[scale=0.4]{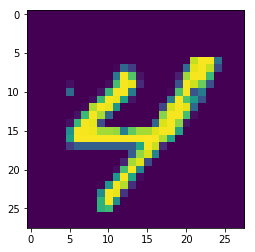}
\end{subfigure}
 \hfill
\begin{subfigure}{.3\linewidth}
 \centering
 \includegraphics[scale=0.4]{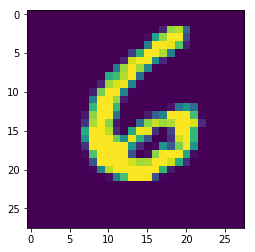}
\end{subfigure}

\bigskip
\begin{subfigure}{.3\linewidth}
 \centering
 \includegraphics[scale=0.4]{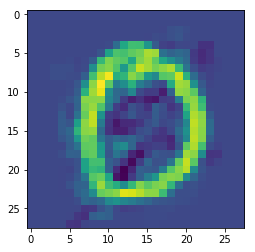}
\end{subfigure}
 \hfill
\begin{subfigure}{.3\linewidth}
 \centering
 \includegraphics[scale=0.4]{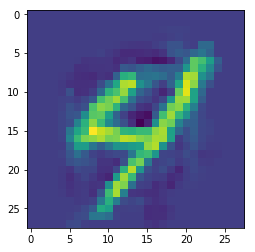}
\end{subfigure}
 \hfill
\begin{subfigure}{.3\linewidth}
 \centering
 \includegraphics[scale=0.4]{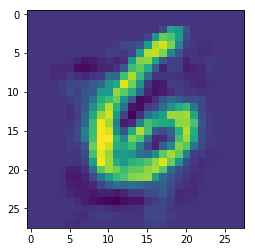}
\end{subfigure}
\caption{MNIST digits in the training set recreated by the network. Top row the actual digits, bottom row, the predictive reconstructions.}
\end{figure}
Additionally, the model was also able to recognize and reconstruct previously unseen MNIST digits, albeit with slightly lower fidelity. Nevertheless it is impressive how rapidly and well the network is able to generalize to completely unseen digits.
\newline
\bigskip

\begin{figure}[H]
\centering
\begin{subfigure}{.3\linewidth}
 \centering
 \includegraphics[scale=0.4]{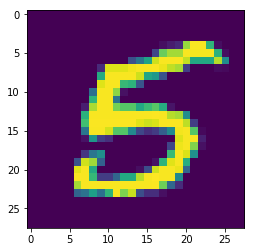}
\end{subfigure}
 \hfill
\begin{subfigure}{.3\linewidth}
 \centering
 \includegraphics[scale=0.4]{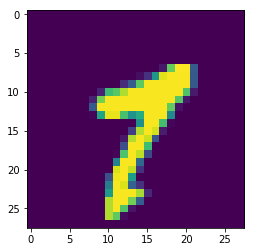}
\end{subfigure}
 \hfill
\begin{subfigure}{.3\linewidth}
 \centering
 \includegraphics[scale=0.4]{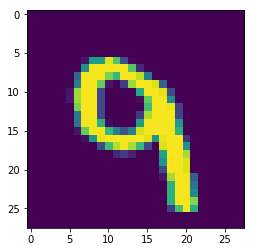}
\end{subfigure}

\bigskip
\begin{subfigure}{.3\linewidth}
 \centering
 \includegraphics[scale=0.4]{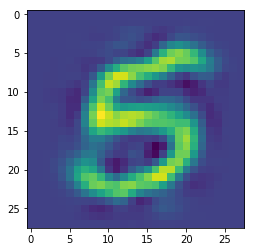}
\end{subfigure}
 \hfill
\begin{subfigure}{.3\linewidth}
 \centering
 \includegraphics[scale=0.4]{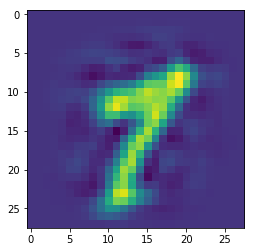}
\end{subfigure}
 \hfill
\begin{subfigure}{.3\linewidth}
 \centering
 \includegraphics[scale=0.4]{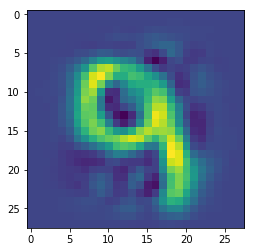}
\end{subfigure}
\caption{Unseen MNIST digits in the test set recreated by the network. Top row the actual digits, bottom row, the predictive reconstructions.}
\end{figure}

Since the model is a generative model, it is also able to generalize outside the training set to generate, or `dream', completely unseen digits by sampling from the latent space. Examples are shown below:

\begin{figure}[htb]
\centering
\begin{subfigure}{.3\linewidth}
 \centering
 \includegraphics[scale=0.4]{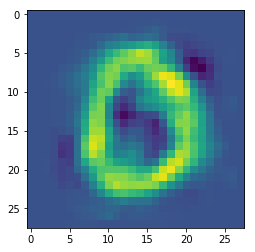}
\end{subfigure}
 \hfill
\begin{subfigure}{.3\linewidth}
 \centering
 \includegraphics[scale=0.4]{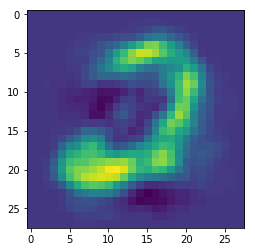}
\end{subfigure}
 \hfill
\begin{subfigure}{.3\linewidth}
 \centering
 \includegraphics[scale=0.4]{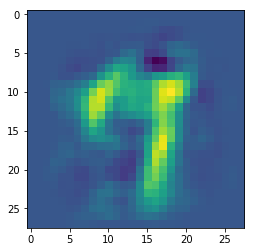}
\end{subfigure}
\caption{Images of hallucinated digits "dreamt" by the network. These were generated by sampling the latent space around the representations of some exemplar digits in the latent space, and then letting the predictive coding network generate its prediction from the chosen latent state.}
\end{figure}

\begin{figure}[H]
\centering
\includegraphics[scale=0.6]{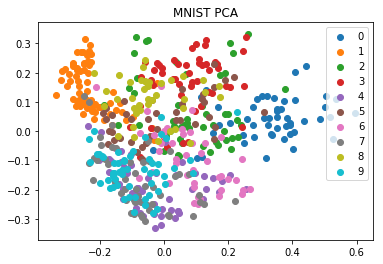}
\caption{A PCA clustering plot of the values of test MNIST digits in the latent space. Even though the 20 dimensional latent space has been reduced down to two, clusters are still visible. For instance, all the 1s are clustered in the top left corner. We thus see that predictive coding appears to be a powerful and fully unsupervised learning algorithm, capable of separating out distinct digits in the latent space, despite not being trained with any label information at all -- and purely on reconstruction.}
\label{PCA_figure}
\end{figure}

Finally, to visualize the latent space, PCA (principal components analysis) was applied to the learned representations in the 20 dimensional latent space to shrink it down to a two dimensional space. It is apparent upon inspection of Figure \ref{PCA_figure} that the latent space, even when shrunk down to two dimensions, does a good job of clustering the MNIST, putting all the 1s in the top left corner, or all the zeros in the middle right. This strongly indicates that the predictive coding model is able to learn the categories of digits despite being trained in an entirely unsupervised way without any knowledge of the true identities of the digits.

We additionally tested the network's capability to reconstruct CIFAR images, which are 32x32 colour images of natural scnes. Our network was the same in the MNIST case except that we used a latent dimension of 50. 

The hierarchical predictive coding CIFAR models are able to learn to reconstruct CIFAR images with impressive fidelity given that they were compressed from a 1024 dimensional image into a 50 dimensional latent state.

\begin{figure}[H]
\centering
\begin{subfigure}{.3\linewidth}
 \centering
 \includegraphics[scale=0.4]{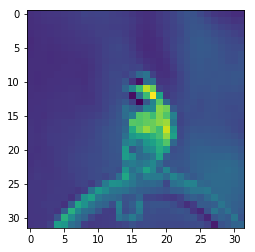}
\end{subfigure}
 \hfill
\begin{subfigure}{.3\linewidth}
 \centering
 \includegraphics[scale=0.4]{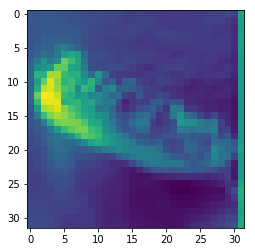}
\end{subfigure}
 \hfill
\begin{subfigure}{.3\linewidth}
 \centering
 \includegraphics[scale=0.4]{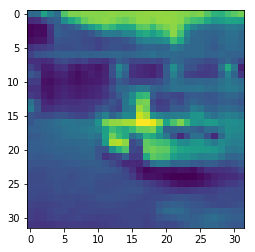}
\end{subfigure}

\bigskip
\begin{subfigure}{.3\linewidth}
 \centering
 \includegraphics[scale=0.4]{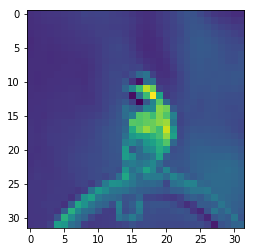}
\end{subfigure}
 \hfill
\begin{subfigure}{.3\linewidth}
 \centering
 \includegraphics[scale=0.4]{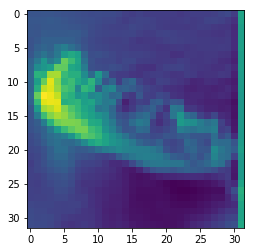}
\end{subfigure}
 \hfill
\begin{subfigure}{.3\linewidth}
 \centering
 \includegraphics[scale=0.4]{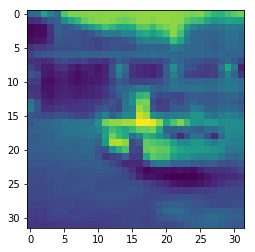}
\end{subfigure}
\begin{subfigure}{.3\linewidth}
 \centering
 \includegraphics[scale=0.4]{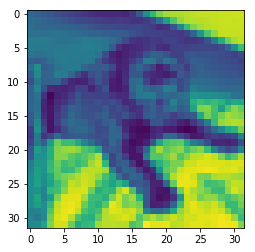}
\end{subfigure}
 \hfill
\begin{subfigure}{.3\linewidth}
 \centering
 \includegraphics[scale=0.4]{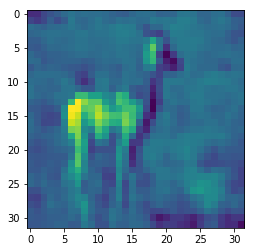}
\end{subfigure}
 \hfill
\begin{subfigure}{.3\linewidth}
 \centering
 \includegraphics[scale=0.4]{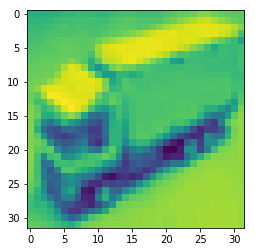}
\end{subfigure}

\bigskip
\begin{subfigure}{.3\linewidth}
 \centering
 \includegraphics[scale=0.4]{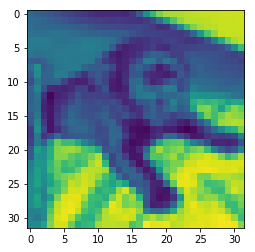}
\end{subfigure}
 \hfill
\begin{subfigure}{.3\linewidth}
 \centering
 \includegraphics[scale=0.4]{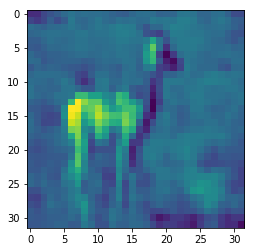}
\end{subfigure}
 \hfill
\begin{subfigure}{.3\linewidth}
 \centering
 \includegraphics[scale=0.4]{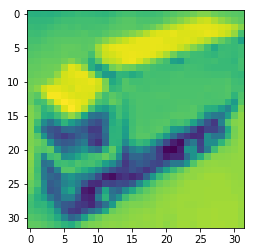}
\end{subfigure}
\caption{Test set CIFAR digits reconstructed by the network. The first of the two lines is the image and the second is the reconstruction. The network is extremely good at reconstructing CIFAR images.}
\end{figure}

Importantly, due to having learnt a latent space, we are able to \emph{interpolate} between images, and thus investigate the properties of the learnt latent space. To interpolate, we begin with two images $o_1$ and $o_2$ and take the difference vector in the latent space $\epsilon = f(o_2) - f(o_1)$ where $f$ is the encoder function. Then, we step in the latent space by a constant $\alpha$ and decode such that $\hat{o} = g(f(o_2) + \alpha \epsilon)$ where $g$ is the decoding function. We stepped in increments of $\alpha = 0.1$. Below we can see the network smoothly interpolating between an image of a horse and a cat. We see that the predictive coding network appears to naturally learn a smooth latent space, allowing for generalization to unseen images and smooth interpolation between classes in the latent space.
\begin{figure}[H]
\centering
\begin{subfigure}{.3\linewidth}
 \centering
 \includegraphics[scale=0.4]{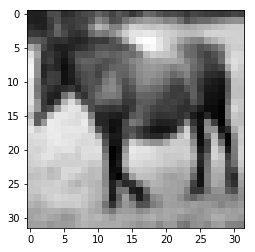}
\end{subfigure}
 \hfill
\begin{subfigure}{.3\linewidth}
 \centering
 \includegraphics[scale=0.4]{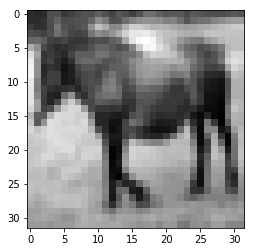}
\end{subfigure}
 \hfill
\begin{subfigure}{.3\linewidth}
 \centering
 \includegraphics[scale=0.4]{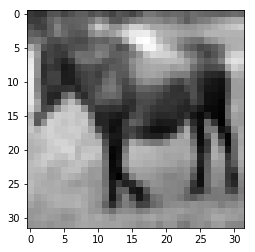}
\end{subfigure}

\bigskip
\begin{subfigure}{.3\linewidth}
 \centering
 \includegraphics[scale=0.4]{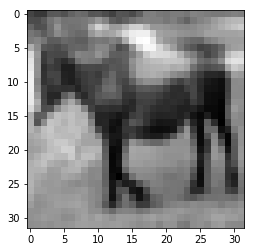}
\end{subfigure}
 \hfill
\begin{subfigure}{.3\linewidth}
 \centering
 \includegraphics[scale=0.4]{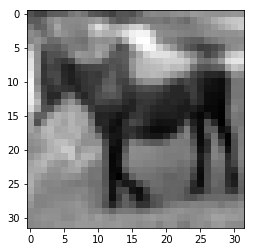}
\end{subfigure}
 \hfill
\begin{subfigure}{.3\linewidth}
 \centering
 \includegraphics[scale=0.4]{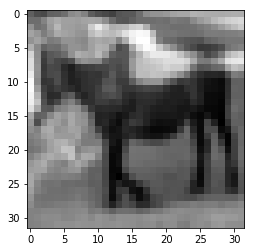}
\end{subfigure}
\begin{subfigure}{.3\linewidth}
 \centering
 \includegraphics[scale=0.4]{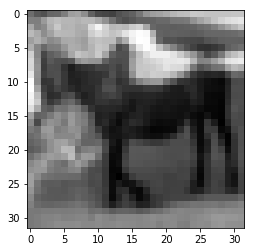}
\end{subfigure}
 \hfill
\begin{subfigure}{.3\linewidth}
 \centering
 \includegraphics[scale=0.4]{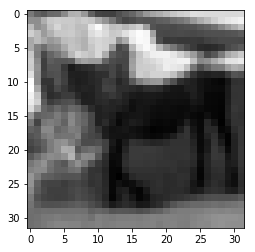}
\end{subfigure}
 \hfill
\begin{subfigure}{.3\linewidth}
 \centering
 \includegraphics[scale=0.4]{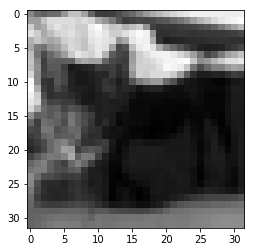}
\end{subfigure}
\caption{The CIFAR model interpolating between a horse and a cat. Read the images left to right top to bottom - like text. The interpolation is done by stepping in the latent space from the representation of the first image in the direction of the second until it is reached.}
\end{figure}

\subsection{Dynamical Predictive coding}

While so far, we have only considered modelling just a single static stimulus $o$. However, the data the brain receives comes in temporal sequences $\bar{o} = [o_1, o_2 \dots ] $. To model such temporal sequences, it is often useful to split the latent variables into \emph{states}, which can vary with time, and \emph{parameters} which cannot. In the case of sequences, instead of minimizing the variational free energy, we must instead minimize the \emph{free action} $\bar{\mathcal{F}}$, which is simply the path integral of the variational free energy through time:
\begin{align*}
 \mu^* &= \underset{\mu}{argmin} \, \, \bar{\mathcal{F}} \\
 \bar{\mathcal{F}} &= \int dt \mathcal{F}_t \\
 \mathcal{F}_t &= \KL \left[ q(x_t | o_t; \phi)|| p(o_t, x_t | x_{t-1}) \right] \numberthis
\end{align*}

While there are numerous methods to handle sequence data, one influential and elegant approach \citep{friston2008DEM,friston2008hierarchical,friston2010generalised} is to represent temporal data in terms of \emph{generalized coordinates of motion}. These coordinates represent not just the immediate observation state, but all the temporal derivatives of the observation. For instance, suppose that the brain represents beliefs about the position of an object. Under a generalized coordinate model, it would also represent beliefs about the velocity (first time derivative), acceleration (second time derivative), jerk (third time derivative) and so on. All these time derivative beliefs are concatenated to form a generalized state. The key insight into this dynamical formulation is, that when written in such a way, many of the mathematical difficulties in handling sequences disappear, leaving relatively straightforward and simple variational filtering algorithms which natively handle smoothly changing sequences. For instance, we maintain a coherent concept of a stationary state, since we can define it as one in which none of the time derivatives are changing. This allows the variational inference procedure to track a moving target by representing it as a steady state in a moving frame of reference.

Because the generalised coordinates are notationally awkward, we will be very explicit in the following. We denote the time derivatives of the generalized coordinate using a $'$, so $\mu'$ is the belief about the velocity of the $\mu$, just as $\mu$ is the belief about the `position' about the $\mu$. A key point of confusion is that there is also a `real' velocity of $\mu$, which we denote $\dot{\mu}$, which represents how the belief in $\mu$ actually changes over time. Importantly, this is not necessarily the same as the belief in the velocity: $\dot{\mu} \neq \mu'$, except at the equilibrium state, which can be understood as the path of least action. Intuitively, this makes sense as at equilibrium (mimimum of the free action, and thus perfect inference), our belief about the velocity of mu $\mu'$ and the `real' velocity perfectly match. Away from equilibrium, our inference is not perfect so they do not necessarily match. We denote the generalized coordinate representation of a state $\tilde{\mu}$ as simply a vector of each of the beliefs about the time derivatives $\tilde{\mu} = [\mu, \mu', \mu'', \mu''' \dots]$. We also define the operator $\mathcal{D}$ which maps each element of the generalised coordinate to its time derivative i.e. $\mathcal{D}\mu = \mu', \mathcal{D}\tilde{\mu} = [\mu', \mu'', \mu''',\mu'''' \dots]$. With this notation, we can define a dynamical generative model using generalized coordinates. Crucially, we assume that the noise $\omega$ in the generative model is not white noise, but is coloured, so it has non-zero autocorrelation and can be differentiated. Effectively, coloured noise allows one to model relatively slowly (not infinitely fast) exogenous forces on the system. For more information on coloured noise vs white noise see \citep{friston2008DEM,yuan2012beyond}. With this assumption we can obtain a generative model in generalized coordinates of motion by simply differentiating the original model.
\begin{align*}
 o &= f(x) + \omega_o && x = g(\bar{x}) + \omega_x \\
 o' &= f'(x)x' + \omega_o' && x' = g'(\bar{x})x' + \omega_x'\\
 o'' &= f'(x)x'' + \omega_o'' && x'' = g'(\bar{x})x'' + \omega_x'' \\
 &\dots && \dots \numberthis
\end{align*}
Where we have applied a local linearisation assumption \citep{friston2008DEM} which drops the cross terms in the derivatives. We can write these generative models more compactly in generalized coordinates.
\begin{align*}
 \tilde{o} = \tilde{f}(\tilde{x}) + \tilde{\omega}_o && \tilde{x} = \tilde{g}(\tilde{\bar{x}}) + \tilde{\omega}_x \numberthis
\end{align*}
which, written probabilistically is $p(\tilde{o},\tilde{x}) = p(\tilde{o} | \tilde{x})p(\tilde{x})$. It has been shown \citep{friston2008DEM} that the optimal (equilibrium) solution to this free action is the following stochastic differential equation,
\begin{align*}
 \label{dynamical_mu_integration}
 \dot{\tilde{\mu}} = \mathcal{D}\tilde{\mu} + \frac{\partial \mathbb{E}_{q(\tilde{x} | \tilde{o}; \tilde{\mu})}[\ln p(\tilde{o}, \tilde{x})]}{\partial \tilde{\mu}} + \tilde{\omega} \numberthis
\end{align*}
Where $\tilde{\omega}$ is the generalized noise at all orders of motion. Intuitively, this is because when $\frac{\partial \mathbb{E}_{q(x | o; \mu)}[\ln p(\tilde{o}, \tilde{x})]}{\partial \mu} = 0$ then $\dot{\tilde{\mu}} = \mathcal{D}\tilde{\mu}$, or that the `real' change in the variable is precisely equal to the expected change. This equilibrium is a dynamical equilibrium which moves over time, but precisely in line with the beliefs $\mu'$. This allows the system to track a dynamically moving optimal solution precisely, and the generalized coordinates let us capture this motion while retaining the static analytical approach of an equilibrium solution, which would otherwise necessarily preclude motion. There are multiple options to turn this result into a variational inference algorithm. Note, the above equation makes no assumptions about the form of variational density or the generative model, and thus allows multimodal or nonparametric distributions to be represented. For instance, the above equation (Equation \ref{dynamical_mu_integration}) could be integrated numerically by a number of particles in parallel, thus leading to a generalization of particle filtering \citep{friston2008variational}. Alternatively, a fixed Gaussian form for the variational density can be assumed, using the Laplace approximation. In this case, we obtain a very similar algorithm to predictive coding as before, but using generalized coordinates of motion. In the latter case, we can write out the free energy as,
\begin{align*}
 \mathcal{F}_t &= \ln p(\tilde{o} | \tilde{x})p(\tilde{x}) \\
 &\propto \tilde{\Sigma}^{-1}_o \tilde{\epsilon}_o^2 + \tilde{\Sigma}^{-1}_x \tilde{\epsilon}_x^2 \numberthis
\end{align*}
Where $\tilde{\epsilon}_o = \tilde{o} - \tilde{f}(\tilde{x})$ and $\tilde{\epsilon}_x = \tilde{o} - \tilde{g}(\tilde{\bar{x}})$. Moreover, the generalized precisions $\tilde{\Sigma}^{-1}$ not only encode the covariance between individual elements of the data or latent space at each order, but also the correlations between generalized orders themselves. Since we are using a unimodal (Gaussian) approximation, instead of integrating the stochastic differential equations of multiple particles, we instead only need to integrate the deterministic differential equation of the mode of the free energy,
\begin{align*}
 \dot{\tilde{\mu}} = \mathcal{D}\tilde{\mu} - \tilde{\Sigma}^{-1}_o \tilde{\epsilon}_o - \tilde{\Sigma}^{-1}_x \tilde{\epsilon}_x \numberthis
\end{align*}
which cashes out in a scheme very similar to standard predictive coding (compare to Equation \ref{PC_hierarchical_mu}), but in generalized coordinates of motion. The only difference is the $\mathcal{D}\tilde{\mu}$ term which links the orders of motion together. This term can be intuitively understood as providing the `prior motion' while the prediction errors provide `the force' terms. To make this clearer, let's take a concrete physical analogy where $\mu$ is the position of some object and $\mu'$ is the expected velocity. Moreover, the object is subject to forces $\tilde{\Sigma}^{-1}_o \tilde{\epsilon}_o + \tilde{\Sigma}^{-1}_x \tilde{\epsilon}_x$ which instantaneously affect its position. Now, the total change in position $\dot{\tilde{\mu}}$ can be thought of as first taking the change in position due to the intrinsic velocity of the object $\mathcal{D}\mu$ and adding that on to the extrinsic changes due to the various exogenous forces. 

Things get more complex when we consider a model which has both dynamical and hierarchical components where there are interactions between them. This we call a full-construct model following the lead of this tutorial \citep*{buckley2017free}. In a full construct model there is a dynamical hierarchy of levels where it is assumed that each dynamical order is only able to affect the level below:

\begin{flalign*}
o &= f(\mu ;\theta) \\
\mu &= f'(\mu' ; \theta')\\
\mu' &= f'(\mu'' ; \theta'') \\
... \numberthis
\end{flalign*}

Similarly, there is simultaneously a hierarchy of levels, where each level is assumed to be predicted by the level above it:
\begin{flalign*}
o &= g(\mu_0 ; \theta_0) \\
\mu_0 &= g'(\mu_1 ; \theta_1) \\
\mu_1 &= g'(\mu_2 ; \theta_2) \\
\mu_2 &= g'(\mu_3 ; \theta_3) \\
... \numberthis
\end{flalign*}
Therefore, each node in the lattice of hierarchical and dynamic hierarchies is influenced by two separate predictions - the dynamical prediction going from higher dynamical orders to lower, and the hierarchical prediction propagating from higher levels of the hierarchy to lower ones. Thus, a single state of a cause-unit $\mu_i^n$, where $i$ is the level of the hierarchy, and $n$ is the dynamical order, is defined to be:
\begin{align*}
\mu_i^n = f(\mu_i^{n+1} ; \theta_i^{n+1}) + g(\mu_{i+1}^n ; \theta_{i+1}^n) \numberthis
\end{align*}

This means that the variational free energy must sum over both dynamical and hierarchical prediction errors, such that:
\begin{align*}
\mathcal{F} = \sum_i \sum_n {\Sigma^n_i}^{-1} ({\epsilon_i^n})^2 \numberthis
\end{align*}

And that additionally the updates for the representation-units and the weights must take this into account. The revised update rules are presented below:
\begin{flalign*}
\frac{d\mu_i^n}{dt} &= \Sigma_{n+1}^{-1} \epsilon_i^{n+1} \frac{df}{d\mu_i^n}\theta_i^{n+1} + \Sigma_n^{-1} \epsilon_i^n + \Sigma_{i-1}^{-1} \epsilon_{i-1}^n \frac{dg}{d\mu_i^n} \theta_{i-1}^n + \Sigma_i^{-1} \epsilon_i^n \numberthis
\end{flalign*}

And for the weights the update rule thus becomes:
\begin{align*}
\frac{d\theta_i^n}{dt} = \Sigma_i^n \epsilon_i^n (\frac{df}{d\theta_i^n}{\mu_i^{n+1}}^T + \frac{dg}{d\theta_i^n} {\mu_{i+1}^n}^T) \numberthis
\end{align*}

These rules appear somewhat more complicated than the corresponding rules in the static case. Nevertheless they only incur a linear (in the order of generalized coordinates considered) additional computational cost.

Preliminary dynamical and full construct models were implemented and tested on simple stimuli. The first task the dynamical models were tested on was predicting a sine wave. This is the perfect toy-task since sine waves have analytic derivatives to any infinite degree. We used a dynamical model which represented three orders of generalized motion. The model was trained to predict a sine wave autoregressively and its' first two temporal derivatives. The model rapidly learned to predict the sine wave, as can be seen from the training graphs below. However, there was a consistent phase-error in the predictions it made, which could have been caused by the rapid rate of change of the sine-wave observations.

\begin{figure}[H]
\centering
\begin{subfigure}{.32\linewidth}
 \centering
 \includegraphics[width=0.8\linewidth]{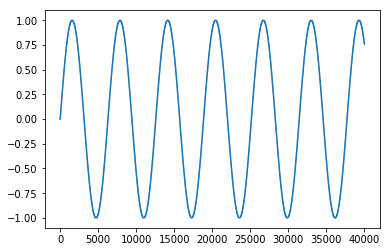}
 \caption{Incoming sense data - i.e. a sine wave}
\end{subfigure}
 \hfill
\begin{subfigure}{.32\linewidth}
 \centering
 \includegraphics[width=0.8\linewidth]{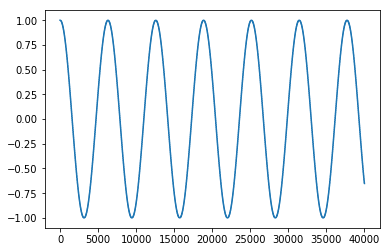}
 \caption{First derivative of the incoming sense data}
\end{subfigure}

\bigskip
\begin{subfigure}{.32\linewidth}
 \centering
 \includegraphics[width=0.8\linewidth]{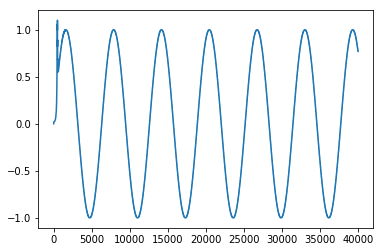}
 \caption{Predicted incoming sense data}
\end{subfigure}
 \hfill
\begin{subfigure}{.32\linewidth}
 \centering
 \includegraphics[width=0.8\linewidth]{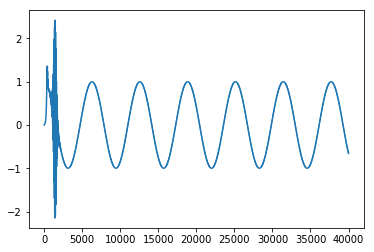}
 \caption{Prediction temporal derivative of the incoming sense data}
\end{subfigure}

\bigskip
 
\begin{subfigure}{.32\linewidth}
 \centering
 \includegraphics[width=0.8\linewidth]{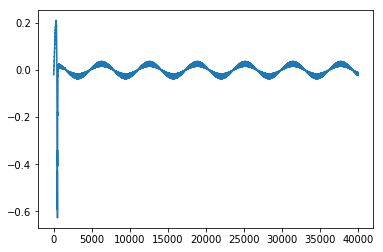}
 \caption{Prediction error at the first dynamical level}
\end{subfigure}
\hfill
\begin{subfigure}{.32\linewidth}
 \centering
 \includegraphics[width=0.8\linewidth]{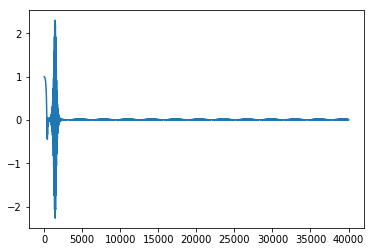}
 \caption{Prediction error at the second dynamical level}
\end{subfigure}
 
\bigskip
\begin{subfigure}{.32\linewidth}
 \centering
 \includegraphics[width=0.8\linewidth]{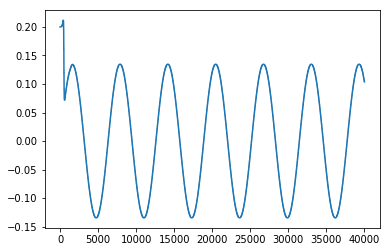}
 \caption{Acitvation of representation units at the first dynamical level}
\end{subfigure}
 \hfill
\begin{subfigure}{.32\linewidth}
 \centering
 \includegraphics[width=0.8\linewidth]{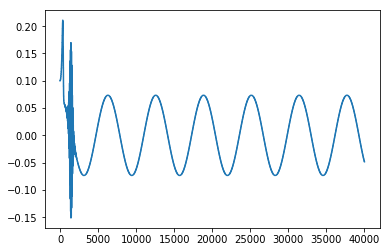}
 \caption{Activation of the representation units at the second dynamical level}\label{fig:image13}
\end{subfigure}
\caption{Prediction errors and prediction for simple toy dynamical models. The task of the dynamical predictive coding model is to learn to predict a sinewave using only the first two dynamical orders -- so including position, velocity, and acceleration. The model starts from randomly initialized parameters. We see that the model very quickly learns to match the incoming sine wave observations with only minimal error at the beginning.}
\end{figure}

The dynamical model does not only work with sine waves. The model was also tested on more jerky waveforms sawtooth waves. In this case a two layer linear dynamical model was used which learned to predict the sawtooth wave and its first temporal derivative. The model predicts the wave very successfully, including the temporal derivative, although there it is a little less successful. Once again there is a persistent patterned prediction error, likely caused by the lag time between the models predictions and the the observations it receives. 
\begin{figure}[H]
\centering
\begin{subfigure}{.32\linewidth}
 \centering
 \includegraphics[width=0.8\linewidth]{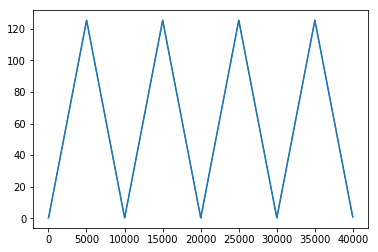}
 \caption{Incomding sense data - the sawtooth wave}
\end{subfigure}
 \hfill
\begin{subfigure}{.32\linewidth}
 \centering
 \includegraphics[width=0.8\linewidth]{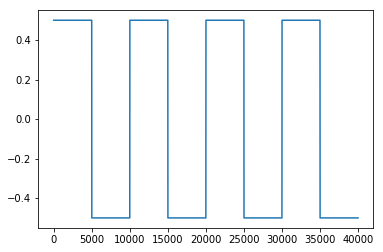}
 \caption{First derivative of the incoming sense data}
\end{subfigure}
\hfill
\begin{subfigure}{.32\linewidth}
 \centering
 \includegraphics[width=0.8\linewidth]{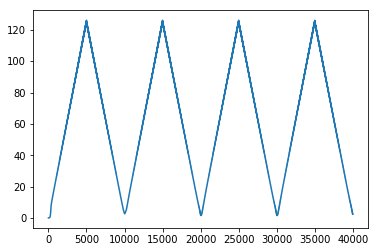}
 \caption{Predicted incoming sense data}
\end{subfigure}

\bigskip
\begin{subfigure}{.32\linewidth}
 \centering
 \includegraphics[width=0.8\linewidth]{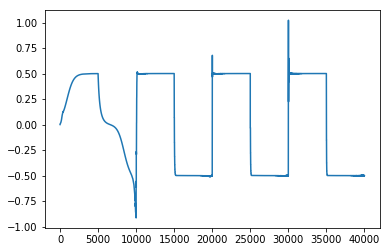}
 \caption{Prediction temporal derivative of the incoming sense data}
\end{subfigure}
\hfill
\begin{subfigure}{.32\linewidth}
 \centering
 \includegraphics[width=0.8\linewidth]{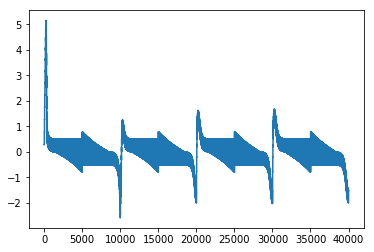}
 \caption{Prediction error at the first dynamical level}
\end{subfigure}
\hfill
\begin{subfigure}{.32\linewidth}
 \centering
 \includegraphics[width=0.8\linewidth]{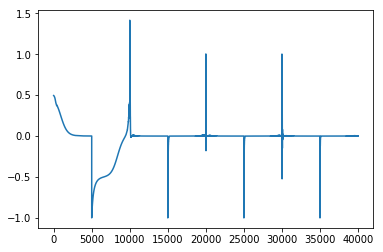}
 \caption{Prediction error at the second dynamical level}
\end{subfigure}
 
\bigskip
\begin{subfigure}{.32\linewidth}
 \centering
 \includegraphics[width=0.8\linewidth]{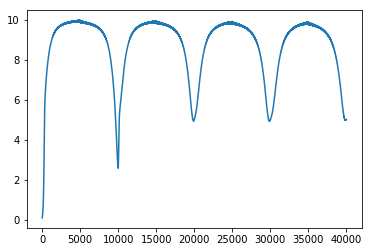}
 \caption{Activation of representation units at the first dynamical level}
\end{subfigure}
 \hfill
\begin{subfigure}{.32\linewidth}
 \centering
 \includegraphics[width=0.8\linewidth]{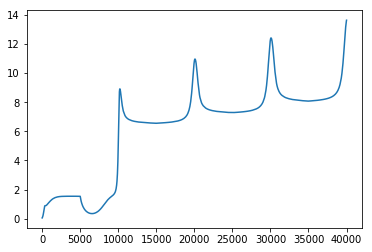}
 \caption{Activation of the representation units at the second dynamical level}
\end{subfigure}
\caption{Dynamical models tested on more challenging sawtooth and square wave inputs. The model was randomly initialized and only modelled the first two dynamical orders (so position, velocity, acceleration). Apart from a brief initial period of uncertainty, the model rapidly learned to predict these more challenging wave shapes.}
\end{figure}

We can also train `full-construct' models on dynamical stimuli. Here we used a model with two hierarchical layers and three dynamical layers and was trained autoregressively to predict a sine-wave. Training was `online' with a learning rate of $0.01$. The generative model parameters $\theta$ were updated initialized randomly and updated each epoch. For every `tick' of the sine wave, the variational parameters $\mu$ were updated for 100 steps. Training graphs are shown below:

\begin{figure}[H]
\centering
\begin{subfigure}{.3\linewidth}
 \centering
 \includegraphics[scale=0.35]{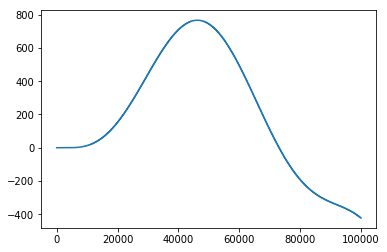}
 \caption{Incomding sense data - Sine wave}
\end{subfigure}
 \hfill
\begin{subfigure}{.3\linewidth}
 \centering
 \includegraphics[scale=0.35]{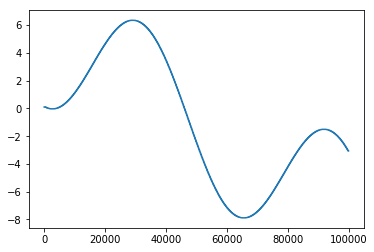}
 \caption{First derivative of the sense-data}
\end{subfigure}
\hfill
\begin{subfigure}{.3\linewidth}
 \centering
 \includegraphics[scale=0.35]{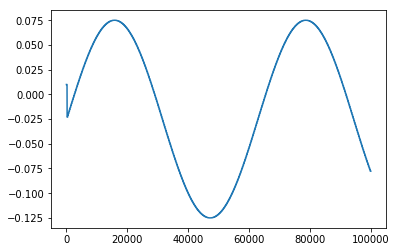}
 \caption{Predicted incoming sense data}
\end{subfigure}
\bigskip

\begin{subfigure}{.3\linewidth}
 \centering
 \includegraphics[scale=0.35]{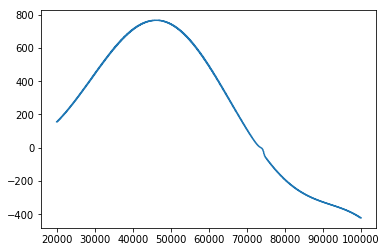}
 \caption{The models' prediction of the incomign sense data}
\end{subfigure}
\hfill
\begin{subfigure}{.3\linewidth}
 \centering
 \includegraphics[scale=0.35]{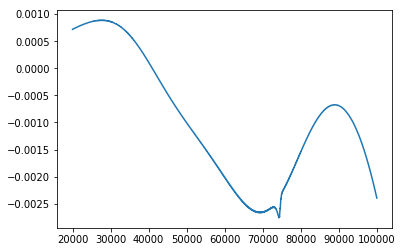}
 \caption{The models' prediction of the first derivative of the incoming sense-data}
\end{subfigure}
\hfill
\begin{subfigure}{.3\linewidth}
 \centering
 \includegraphics[scale=0.35]{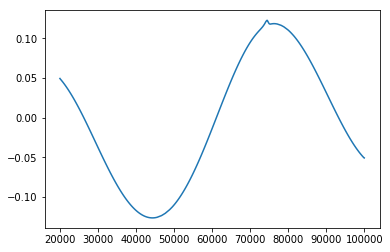}
 \caption{the models' prediction of the second derivative of the incoming sense-data}
\end{subfigure}
 
\bigskip
\begin{subfigure}{.3\linewidth}
 \centering
 \includegraphics[scale=0.35]{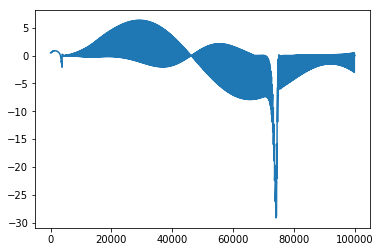}
 \caption{The prediction error at the first hierarcical layer}
\end{subfigure}
 \hfill
\begin{subfigure}{.3\linewidth}
 \centering
 \includegraphics[scale=0.35]{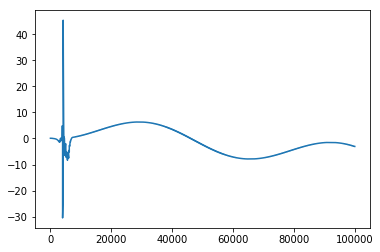}
 \caption{The prediction error at the second hierarcical layer}
\end{subfigure}
\hfill
\begin{subfigure}{.3\linewidth}
 \centering
 \includegraphics[scale=0.35]{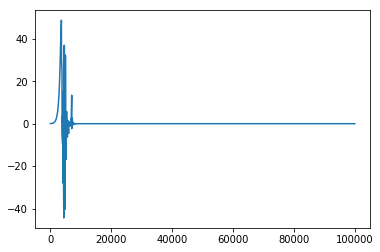}
 \caption{The prediction error at the first dynamical layer}
\end{subfigure}
 \hfill
\caption{The training graphs of the full construct model. It can successfully predict the first three temporal derivatives of a sine wave, and also minimise prediction error up to multiple hierarchical layers. The full construct model was randomly initialized, and learnt both parameters and inferred states purely online -- thus achieving a `double deconvolution' \citep{friston2008DEM}.}
\end{figure}

The top two rows of graphs show the incoming sense data and the first two temporal derivatives of the sense-data. The next two rows show the prediction errors over time for various levels of the hierarchy. The full construct model appears a bit less stable and successful than the simple dynamical model, likely because it is much more complex and has many more moving parts. Nevertheless it manages to learn the sine wave shapes relatively faithfully and does also manage to rapidly reduce the prediction error over time. Moreover these sorts of tasks do not really play well to the strength of the full-construct model since the input data (the sine wave) contains no suitable hierarchical structure for the higher levels to model. It seems likely that as these models are scaled up to more challenging tasks, the greater expressivity and power of the full-construct models will become more apparent. So far, we have only experimented with `full-construct' models on simple toy tasks such as sine waves. An interesting avenue for future would work be experimenting as to whether full construct models could be scaled up to handle challenging machine learning tasks dealing with sequential data such as video prediction. While predictive coding networks have been proposed for this task \citep{lotter2016deep}, none to our knowledge have explicitly utilized generalized coordinates in any capacity. There is, however, a literature using the predictive coding schemes with generalized coordinates in nonlinear (chaotic) systems with separation of temporal scales. This enables the recognition and prediction of things like birdsong and speech – and indeed their learning of particular songs. Crucially, these applications rest upon generalised coordinates of motion, usually up to 4th order motion \citep{friston2009predictive,friston2015duet,isomura2019Bayesian}
\section{Predictive Coding and Kalman Filtering}

A key intuition behind the utility of predictive coding is that it naturally handles \emph{filtering} tasks. Filtering tasks require constant updates of a moving state estimate given sequences of new data. Effectively, filtering is the task of learning and inferring movements in the hidden state of the world from changing input \emph{sequences} -- as opposed to the usual machine learning task of inferring hidden states (such as labels) from single static inputs. Importantly the core task faced by much of the brain is fundamentally one of filtering, since the inputs the brain receives are actually temporally extended sequences, rather than static flashes. For instance, in vision the task of the brain is not to categorize static images but rather to infer the state of, and ultimately interact with, a smoothly changing external world situated in continuous time. Moreover, it is known that the brain takes substantial advantage of the additional information given by integrating sequences over time such as optical flow \citep{gibson2002theory} and active motion to explore different angles on a given scene \citep{henderson2017gaze}.

If, as we generally assume throughout the thesis, that the brain is fundamentally a (Bayesian) inference machine, then the core task of the brain must be \emph{Bayesian Filtering} instead of static Bayesian inference \citep{sarkka2013Bayesian}. Bayesian filtering is mathematically somewhat more involved, due to the need to perform inference over sequences instead of single data-points, but there are a wide variety of algorithms in the literature which perform Bayesian filtering, often highly effectively \citep{kutschireiter2018nonlinear,kutschireiter2020hitchhiker}. Mathematically, we can formalize the filtering problem as follows \citep*{jaswinskistochastic,stengel1994optimal}. We have an estimated state $\hat{x}_t$, and some model of how the world evolves (the dynamics model): $\hat{x}_{t+1} = f(\hat{x}_t)$. We also receive observations $o$, and you have some model of how the observations depend on the estimated state (the observation model): $o = g(\hat{x}_t)$. The task, then, is to compute $p(\hat{x}_{t+1} | \hat{x}_t, o_{1...t})$. In the general nonlinear case, this calculation is analytically intractable and extremely expensive to compute exactly. Some form of approximate solution is required. Two forms of approximation are generally used. The first is to approximate the model - such as by assuming linearity of the dynamics and observation models. 

The second method is to approximate the posterior -- usually with a set of samples (or particles). This approach is taken by the class of particle filtering algorithms which track the changing posterior by propagating the particles through the dynamics and then resampling based upon updated measurement information \citep*{arulampalam2002tutorial,gordon1993novel}. This approach can handle general nonlinear filtering cases, but suffers strongly from the curse of dimensionality. If the state-space is high-dimensional the number of particles required for a good approximation grows rapidly \citep*{doucet2000sequential}.Moreover, there has been some fascinating work on implementing particle filtering methods in neural circuitry \citep*{kutschireiter2015neural}, as well as speculation about whether perhaps the brain may utilize particle or sampling methods for inference instead of variational ones \citep{sanborn2016Bayesian}.

Nevertheless, here we focus primarily on approximate variational approaches to inference. Specifically, we first demonstrate that predictive coding is naturally a filtering algorithm -- perhaps more naturally than one applied to static datasets. The only change to the algorithm is simply \emph{what} is predicted. If predictive coding is set up so as to predict the \emph{next} input \citep{mumford1992computational,clark_whatever_2013}, which is highly plausible in the brain, then it can perform variational Bayesian filtering. In this section, we explore this predictive coding filtering algorithm and show, crucially, that in the linear case it becomes a variant of Kalman filtering -- a fundamental and ubiquitous algorithm in classical control \citep{kalman1960contributions,kalman1960new}. Moreover, in the nonlinear case, predictive coding becomes a variant of extended Kalman filtering \citep{ollivier2019extended}. The Kalman Filter solves the general filtering problem by making two simplifying assumptions. The first is that both the dynamics model and the observation model are linear. The second assumption is that noise entering the system is white and Gaussian. This makes both the prior and likelihoods Gaussian. Since the Gaussian distribution is a conjugate prior to itself, this induces a Gaussian posterior, which can then serve as the prior in the next timestep. Since both prior and posterior are Gaussian, filtering can continue recursively for any number of time-steps without the posterior growing in complexity and becoming intractable. The Kalman filter is the Bayes-optimal solution provided that the assumptions of linear models and white Gaussian noise are met \citep*{kalman1960new}. The Kalman Filter, due to its simplicity and utility is widely used in engineering, time-series analysis, aeronautics, and economics \citep*{grewal2010applications,leondes1970theory,schneider1988analytical,harvey1990forecasting}.

Since predictive coding possesses several neurophysiologically realistic process theories \citep{bastos2012canonical}, this correspondence provides an avenue for a biologically plausible implementation of Kalman filtering in the brain. There is substantial evidence that the brain is capable of Bayes-optimal integration of noisy measurements, and is apparently in possession of robust forward models both in perception \citep*{zago2008internal,simoncelli2009optimal} and motor control \citep*{munuera2009optimal, gold2003influence,todorov2004optimality}. \citet*{de2013Kalman} have even shown that a Kalman filter successfully fits psychomotor data on visually guided saccades and smooth pursuit movement, although they remain agnostic on how it may be implemented in the brain. We demonstrate, however, a clear mathematical link of the relationship between Kalman filtering and predictive coding, allowing us first to use results from Kalman filtering to understand the performance of predictive coding algorithms, and second enabling us to utilize the process theories of predictive coding to understand how the brain may perform crucial filtering tasks.

First, we reveal the precise relationship between Kalman filtering and predictive coding -- namely that both optimize the same Bayesian objective, which is convex in the linear case. However, the Kalman filter solves the optimization problem analytically, thus giving rise to its algebraic complexities and especially the highly neurobiologically implausible Kalman gain matrix. Predictive coding, on the other hand, solves the objective through a process of gradient descent on the sufficient statistics of the variational distribution, thereby obtaining biologically plausible Hebbian update rules. Additionally, the fully Bayesian perspective granted by predictive coding also allows us to perform learning of the generative model -- i.e. learning the coefficients of the dynamics and likelihood matrices -- which allows us to handle cases where the model of the world is unknown, in contrast to traditional Kalman filtering which assumes accurate (and linear) dynamics and observation models of the world. While the close relationship between Kalman filtering and (linear) predictive coding has been hinted at before \citep{friston2005theory,friston2008hierarchical}, there it is claimed that predictive coding is `equivalent' to Kalman filtering -- which is not the case except insofar as the two algorithms optimize the same objective. \citet{baltieri2020Kalman} provide side by side comparisons of the update rules for predictive coding and Kalman filtering, but do not go beyond this superficial analysis to uncover the precise relationship between them. 

Secondly, we directly compare the performance of the Kalman filter and our predictive coding algorithm on a simplified location tracking task -- which the Kalman filter excels at. We show that despite the predictive coding algorithm performing a gradient descent instead of an analytical solution, it performs comparably with the Kalman filter and, due to the convexity of the underlying optimization problem, requires very few iterations to converge. This rapid convergence is important, since the brain is heavily time-constrained in its inferences -- choices often must be made fast. Secondly, we demonstrate that the learning rules for the likelihood and dynamics matrices allow us to perform online tracking even when the model is completely unknown. We only show that this is the case for the dynamics, however, as learning does not perform well with an unknown observation model. We hypothesize that this is due to the ill-posedness of the resulting optimization problem.

\subsection{The Kalman Filter}
The Kalman Filter is defined upon the following linear state-space \footnote{For simplicity, the model is presented in discrete time. The continuous time analogue of the Kalman filter is the Kalman-Bucy filter \citep*{kalman1961new}. Generalization of this scheme to continuous time is an avenue for future work.}
\begin{flalign*}
 x_{t+1} &= Ax_t + Bu_t + \omega & \\
 o_{t+1} &= Cx_{t+1} + z_t \numberthis
\end{flalign*}
Where $x_t$ represents the hidden or internal state at time t. $u_t$ is the control - or known inputs to the system - at time t. Matrices $A$,$B$, and $C$ parametrize the linear dynamics or observation models, and $\omega$ and $z$ are both zero-mean white noise Gaussian processes with covariance $\Sigma_\omega$ and $\Sigma_z$, respectively. Since the posterior $p(x_{t+1}|o_{1...t}, x_{t})$ is Gaussian, it can be represented by its two sufficient statistics -- the mean $\mu$ and covariance matrix $\Sigma_x$.

Kalman filtering proceeds by first `projecting' forward the current estimates according to the dynamics model. Then these estimates are `corrected' by new sensory data. The Kalman filtering equations are as follows:
\newline
\textbf{Projection}
\begin{flalign*}
 & \hat{\mu}_{t+1} = A\mu_t + Bu_t &\\
 & \hat{\Sigma}_x(t+1) = A\Sigma_x(t) A^T + \Sigma_\omega \numberthis
\end{flalign*}
\textbf{Correction}
\begin{flalign*}
 & \mu_{t+1} = \hat{\mu}_{t+1} + K(o_{t+1} - C\hat{\mu}_{t+1}) & \\
 & \Sigma_x(t+1) = (I - K)\hat{\Sigma}_x(t+1) \\
 & K = \hat{\Sigma}_x(t+1)C^T[C\hat{\Sigma}_x(t+1)C^T + \Sigma_z]^{-1} \numberthis
\end{flalign*}

Where $\mu_t$ and $\Sigma_x(t)$ are the mean and variance of the estimate of the state $x$ at time t, and $K$ is the Kalman gain matrix. Although these update rules provide an analytically exact solution to the filtering problem, the complicated linear algebra expressions, especially that for the Kalman gain matrix $K$, make it hard to see how such equations could be implemented directly in the brain. 

Importantly, these Kalman filtering equations can be derived directly from Bayes' rule. The mean of the posterior distribution is also the MAP (maximum-a-posteriori) point, since a Gaussian distribution is unimodal. Thus, to estimate the new mean, we simply have to estimate,
\begin{flalign*}
 \label{KF_MAP}
 &\underset{\hat{x}_{t+1}}{arg max} \, \, p(\hat{x}_{t+1} | o_{t+1}, \hat{x}_t) \propto \underset{\hat{x}_{t+1}}{argmax} \, p(o_{t+1} |\hat{x}_{t+1})p(\hat{x}_{t+1} | \hat{x}_t) \\
 &= \underset{\hat{x}_{t+1}}{arg max} \, N(o_{t+1};C\hat{x}_{t+1}, \Sigma_z)N(\hat{x}_{t+1}; A\hat{x}_t + Bu_t, \Sigma_\omega) \\
 &= \underset{\mu_{t+1}}{arg max} \, \frac{1}{Z}exp(-(y - C\mu_{t+1})^T\Sigma_Z(y - C\mu_{t+1}) \\ &+ (\mu_{t+1} - A\mu_t - Bu_t)^T\hat{\Sigma}_x(\mu_{t+1} - A\mu_t - Bu_t) \\
 &= \underset{\mu_{t+1}}{arg min} \, -(y - C\mu_{t+1})^T\Sigma_Z(y - C\mu_{t+1}) + (\mu_{t+1} - A\mu_t - Bu_t)^T\hat{\Sigma}_x (\mu_{t+1} - A\mu_t - Bu_t) \numberthis
\end{flalign*}
In the second line, the algebraic form of the Gaussian density is substituted and we have switched the maximization variable to $\mu_{t+1}$ due to the fact that the maximum of a Gaussian is also its mean. We also minimize the log probability instead of maximizing the probability, which gets rid of the exponential and the normalizing constant (which can be computed analytically since the posterior is Gaussian) \footnote{The log transformation is valid under maximization/minimization since the log function is monotonic.}. From this objective, one can simply solve analytically for the optimal $\mu$ and $\Sigma$. For a full derivation see Appendix A. 

\subsection{Predictive Coding as Kalman Filtering}

Here we demonstrate the relationship between predictive coding and Kalman filtering. First, we need to explicitly write out and adapt the mathematical apparatus of predictive coding to filtering problems. To do so, we need to perform variational inference over
full trajectories $o_{1:T}, x_{1:T}$ of observations and hidden states. If we then assume trajectories are Markov, and are thus licensed to apply a Markov factorization of the generative model $p(o_{1:T}, x_{1:T}) = p(o_1 | x_1)p(x_1) \prod_{t=2}^T p(o_t | x_t)p(x_t | x_{t-1})$ \footnote{Where, to make this expression not a function of $x_{t-1}$, we implicitly average over our estimate of $x_{t-1}$ from the previous timestep: $p(x_t | x_{t-1}) = \E_{q(x_{t-1})}[p(x_t | x_{t-1})]$} and a mean-field temporal factorization of the variational density, so that it is independent across timesteps $q(x_{1:T} ; \theta) = \prod_{t=1}^T q(x_t ; \theta)$, then the variational free energy of the trajectory factorizes into independently optimizable free-energies of a particular timestep,
\begin{align*}
 \mathcal{F}(o_{1:T}) &= \sum_{t=1}^T \mathcal{F}_t(o_t) \\
 \mathcal{F}_t(o_t) &= \KL[q(x_t ;\theta)||p(o_t, x_t | x_{t-1})] \numberthis
\end{align*}

This temporal factorization of the free energy means that the minimization at each timestep is independent of the others, and so we only need consider a single minimization of a single timestep to understand the solution, since all time-steps will be identical in terms of the solution method. Applying the linear Gaussian assumptions of the Kalman filter, we can specify our generative model in terms of Gaussian distributions,
\begin{align*}
 p(o_t, x_t | x_{t-1}) &= p(o_t | x_t)p(x_t | x_{t-1}) \\
 &= \mathcal{N}(o_t; Cx_t, \Sigma_z)\mathcal{N}(x_t | Ax_{t-1}, \Sigma_x) \numberthis
\end{align*}

Since we know the posterior is Gaussian, it makes sense to also use a Gaussian distribution for the variational approximate distribution. Importantly, for predictive coding we make an additional assumption -- the Laplace Approximation -- which characterises the variance of this Gaussian as an analytic function of the mean, thus defining,
\begin{align*}
 q(x_t; \theta) = \mathcal{N}(x_t; \mu_t, \sigma(\mu)) \numberthis
\end{align*}
where $\theta = [\mu_t,\sigma(\mu_t)]$ are the parameters of the variational distribution -- in this case a mean and variance since we have assumed a Gaussian variational distribution. With the variational distribution and generative model precisely specified, it is now possible to explicitly evaluate the variational free energy for a specific time-step,
\begin{align*}
 \mathcal{F}_t(o_t) = \mathcal{F}_t(o_t) &= \KL[q(x_t ;\theta)||p(o_t, x_t | x_{t-1})] \\
 &= -\E_{q(x_t;\theta)}[\ln p(o_t, x_t | x_{t-1})] - \mathbb{H}[q(x_t;\theta)] \numberthis
\end{align*}
Where the second term is the entropy of the variational distribution. Since we are only interested in minimizing with respect to the mean $\mu_t$ and the expression for the entropy of a Gaussian does not depend on the mean, we can ignore this entropy term in subsequent steps. The key quantity is the `energy' term $\E_{q(x_t;\theta}[\ln p(o_t, x_t | x_{t-1}]$. Since the Laplace approximation ensures that most of the probability distribution is near the mode $\mu_t$ of the variational distribution, we can well approximate the expectation using a Taylor expansion to second order around the mode,
\begin{align*}
 \E_{q(x_t;\theta}[\ln p(o_t, x_t | x_{t-1})] &\approx \ln p(o_t, \mu_t | \mu_{t-1}) + \E[\frac{\partial p(o_t, x_t | x_{t-1})}{\partial x_t}|_{x_t = \mu_t}[x_t - \mu_t] \\ &+ \E[\frac{\partial^2 p(o_t, x_t | x_{t-1})}{\partial x_t^2}|_{x_t = \mu_t}[x_t - \mu_t]^2 \\
 &= \ln p(o_t, \mu_t | \mu_{t-1}) + \frac{\partial p(o_t, x_t | x_{t-1})}{\partial x_t}|_{x_t = \mu_t}\underbrace{[\E[x_t] - \mu_t]}_{=0} \\ &+ \frac{\partial^2 p(o_t, x_t | x_{t-1})}{\partial x_t^2}|_{x_t = \mu_t}\underbrace{E[(x_t - \mu_t)^2]}_{=\sigma} \numberthis
\end{align*}

Since the first term vanishes as $\E[x_t] - \mu_t = \mu_t - \mu_t = 0$ and we can neglect the second term since it only depends on $\sigma$ and not $\mu$, then the only term that matters for the minimization is the first term $\ln p(o_t , \mu_t | \mu_{t-1})$. This means that we can write the overall optimization problem solved by predictive coding as,
\begin{align*}
 \underset{\mu_t}{arg min} \, \mathcal{F}_t(o_t) = \underset{\mu_t}{arg min} \, \ln p(o_t, \mu_t | \mu_{t-1}) \numberthis
\end{align*}
which is the same as the MAP optimization problem presented in Equation \ref{KF_MAP}. This means that ultimately the variational inference problem solved by predictive coding and the MAP estimation problem solved by the Kalman filter are the same although the interpretation of $\mu_t$ differs slightly -- from being a parameter of a Gaussian variational distribution versus simply a variable in the generative model -- the actual update rules involving $\mu_t$ are the same in both cases. Now we know that (linear) predictive coding and Kalman filtering share the same objective, we can precisely state their differences. While Kalman filtering analytically solves this objective directly, in predictive coding, we instead set the dynamics of the parameters to be a gradient descent on the variational free energy, which reduces to the MAP objective solved by the Kalman Filter.

For instance, we can derive the dynamics with respect to the variational parameters $\mu_{t+1}$ which, in neural process theories, are typically operationalized as the `activation' units as, 
\begin{flalign*}
\label{KF_mu}
 \frac{dL}{d\mu_{t+1}} &= 2C^T\Sigma_z y - (C^T \Sigma_z C + C^T\Sigma_z^T C)\mu_{t+1} + (\Sigma_x + \Sigma_x^T)\mu_{t+1} - 2\Sigma_x A\mu_t - 2\Sigma_x Bu_t & \\
 &= 2C^T\Sigma_z C\mu_{t+1} - 2C^T\Sigma_z C\mu_{t+1} + 2\Sigma_x \mu_{t+1} - 2\Sigma_x A\mu_t - 2\Sigma_x Bu_t \\
 &= -C^T \Sigma_z[y - C\mu_{t+1}] + \Sigma_x[\mu_{t+1} - A\mu_t - B\mu_t] \\
 &= -C^T \Sigma_z \epsilon_z + \Sigma_x \epsilon_x \numberthis
\end{flalign*}
Where $\epsilon_z = y - C\mu_{t+1}$ and $\epsilon_x = \mu_{t+1} - A\mu_t - Bu_t$.Thus, we can see that the gradient perfectly recapitulates the standard predictive coding scheme with precision weighted prediction errors. Similarly, by taking gradients with respect to the $A$, $B$, and $C$ matrices of the generative model, we obtain familiar looking update rules which consist of Hebbian update rules between the prediction errors and the presynaptic activations

\begin{flalign*}
\label{KF_A}
 \frac{dL}{dA} &= \frac{d}{dA}[-2\mu_{t+1}^T \Sigma_x A\mu_t + \mu_t^T A^T \Sigma_x A\mu_t + \mu_t^T A^T \Sigma_x Bu_t + u_t^TB^T \Sigma_x A \mu_t] & \\
 &= -2\Sigma_x^T \mu_{t+1} \mu_t^T + \Sigma_x^T A\mu_t \mu_t^T \Sigma_x A \mu_t\mu_t^T + \Sigma_x Bu_t\mu_t^T + \Sigma_x^T Bu_t\mu_t^T \\
 &= -\Sigma_x[\mu_{t+1} - A\mu_t - Bu_t]\mu_t^T \\
 &= -\Sigma_x \epsilon_x \mu_t^T \numberthis
\end{flalign*}

And similarly for the $B$ matrix.
\begin{flalign*}
\label{KF_B}
 \frac{dL}{dB} &= \frac{dL}{dB}[2u_t^TB^T\Sigma_x A\mu_t + u_t^TB^T\Sigma_x Bu_t - 2\mu_{t+1}^T \Sigma_x Bu_t] & \\
 &= (\Sigma_x + \Sigma_x^T)Bu_t u_t^T + 2\Sigma_x A \mu_t u_t^T - 2 \Sigma_x \mu_{t+1} u_t^T \\
 &= - \Sigma_x[\mu_{t+1} - A\mu_t - Bu_t]u_t^T \\
 &= - \Sigma_x \epsilon_x u_t^T \numberthis
\end{flalign*}
And the $C$ observation matrix.
\begin{flalign*}
\label{KF_C}
 \frac{dL}{dC} &= \frac{dL}{dC}[-2\mu_{t+1}^TC^TRy + \mu_{t+1}^T C^T R C \mu_{t+1}] &\\
 &= -2Ry\mu_{t+1}^T + 2RC\mu_{t+1}\mu_{t+1}^T \\
 &= -R[y - c\mu_{t+1}]\mu_{t+1}^T \\
 &= -R\epsilon_y \mu_{t+1}^T \numberthis
\end{flalign*}

\subsection{Results}

\begin{figure}[H]
 \begin{subfigure}{0.49\textwidth}
 \centering
 \includegraphics[width=.95\linewidth]{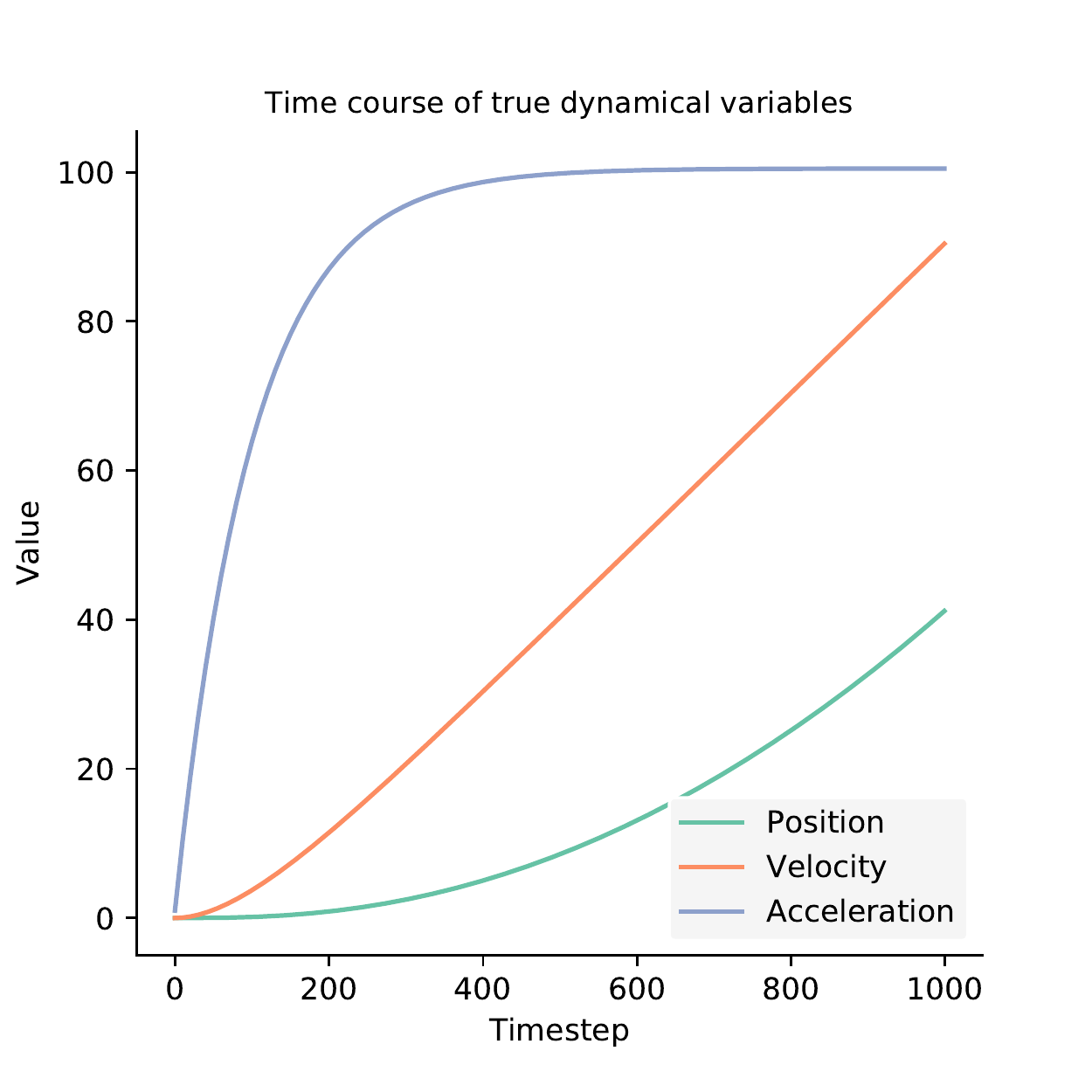}
 \caption{True Dynamics}
 \end{subfigure}%
 \begin{subfigure}{0.49\textwidth}
 \centering
 \includegraphics[width=.95\linewidth]{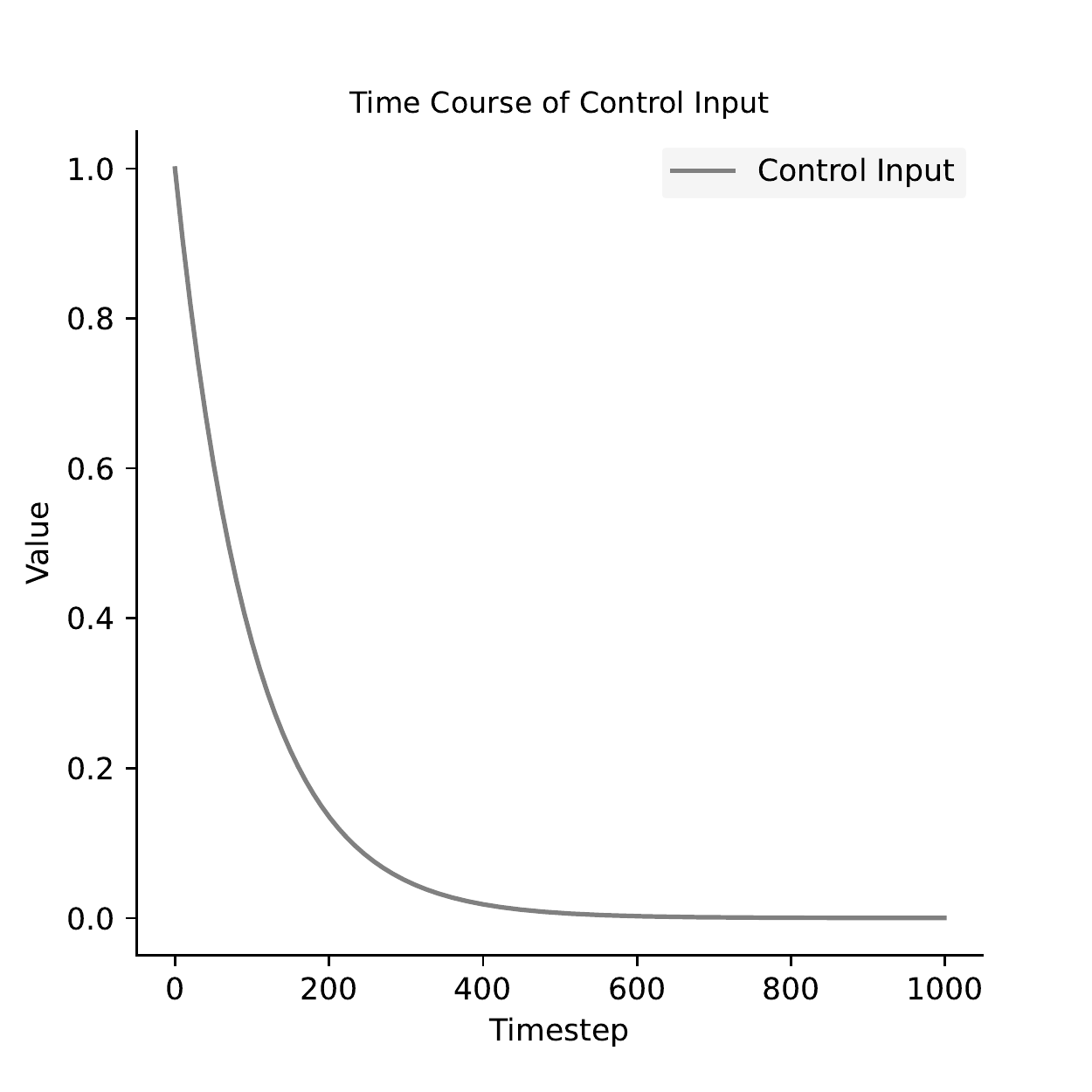}
 \caption{Control Input}
 \end{subfigure}
 \begin{subfigure}{0.49\textwidth}\quad
 \centering
 \includegraphics[width=.95\linewidth]{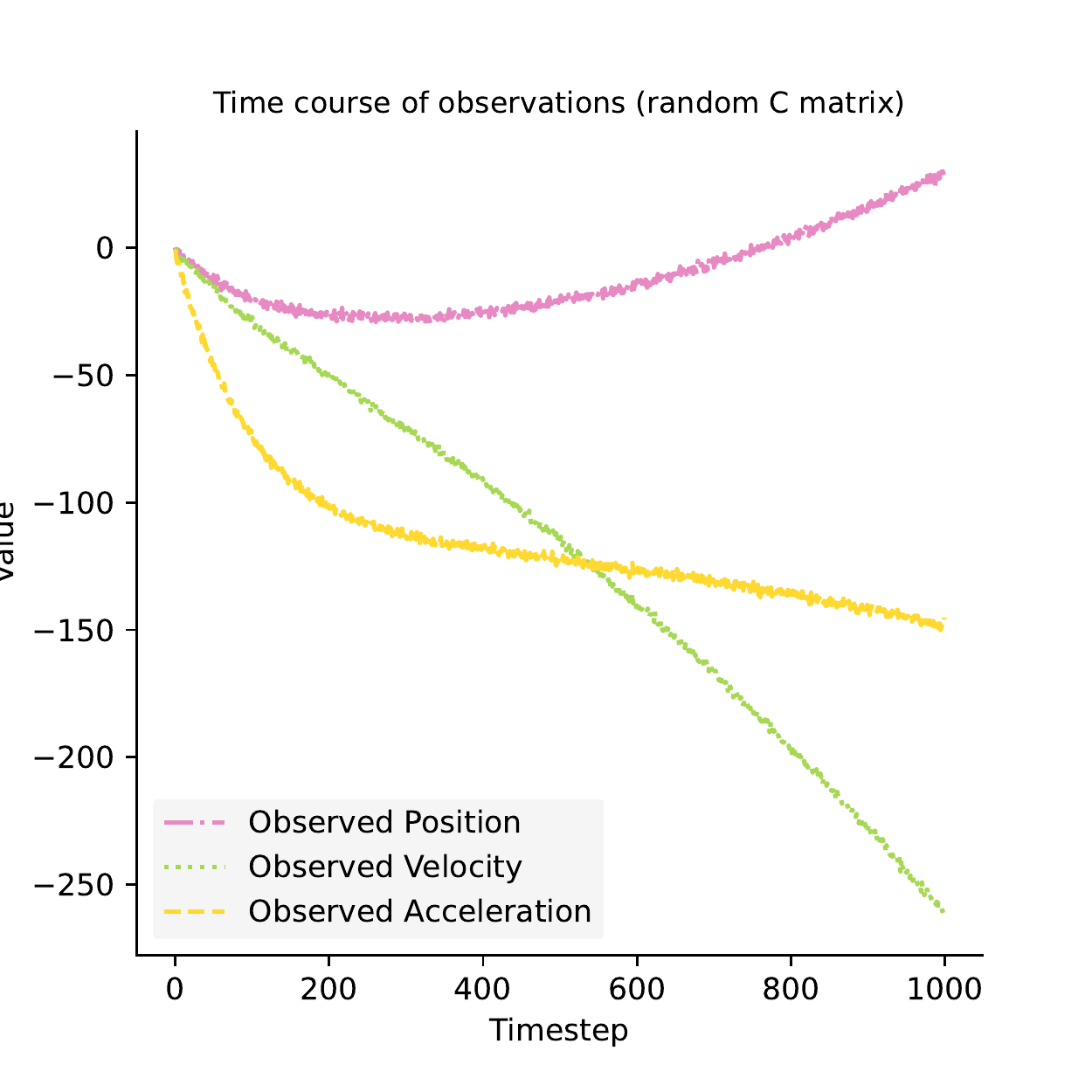}
 \caption{Observations}
 \end{subfigure}
\caption{The true dynamics, control input, and observations generated by a random C matrix. These are the source of truth that the predictive coding Kalman filter tries to approximate. The observations differ substantially from the true dynamics due to the random C matrix, which makes the inference problem faced by the predictive coding filter much more challenging, since it must de-scramble the observations to infer the true dynamics.}

\label{KF_true_dynamics_figure}
\end{figure}

We now compare the analytical Kalman filter with our predictive coding algorithm on a simple filtering application -- that of tracking the motion of an accelerating body given only noise sensor measurements. The body is accelerated with an initial high acceleration that rapidly decays according to an exponential schedule. The filtering algorithm must infer the position, velocity, and true acceleration of the body from only a kinematic dynamics model and noisy sensor measurements. The body is additionally perturbed by white Gaussian noise in all of the position, velocity and displacement. The control schedule and the true position, velocity and displacement of the body are shown in Figure \ref{KF_true_dynamics_figure} below.

The analytical Kalman filter was set up as follows. It was provided with the true kinematic dynamics matrix ($A$) and the true control matrix ($B$),
\begin{flalign*}
 A &= \begin{bmatrix}
 1 & dt & \frac{1}{2}dt^2 \\,
 0 & 1 & dt \\
 0 & 0 & 1
 \end{bmatrix} & \\
 B &= \begin{bmatrix}
 0 & 0 & 1
 \end{bmatrix} \numberthis
\end{flalign*}

The observation matrix C matrix was initialized randomly with coefficients drawn from a normal distribution with 0 mean and a variance of 1. This effectively random mapping of sensory states meant that the filter could not simply obtain the correct estimate directly but had to disentangle the measurements first. The Q and R matrices of the analytical Kalman filter were set to constant diagonal matrices, where the constant was the standard variance of the noise added to the system.

\begin{figure}[H]
 \begin{subfigure}{0.49\textwidth}
 \centering
 \includegraphics[width=.95\linewidth]{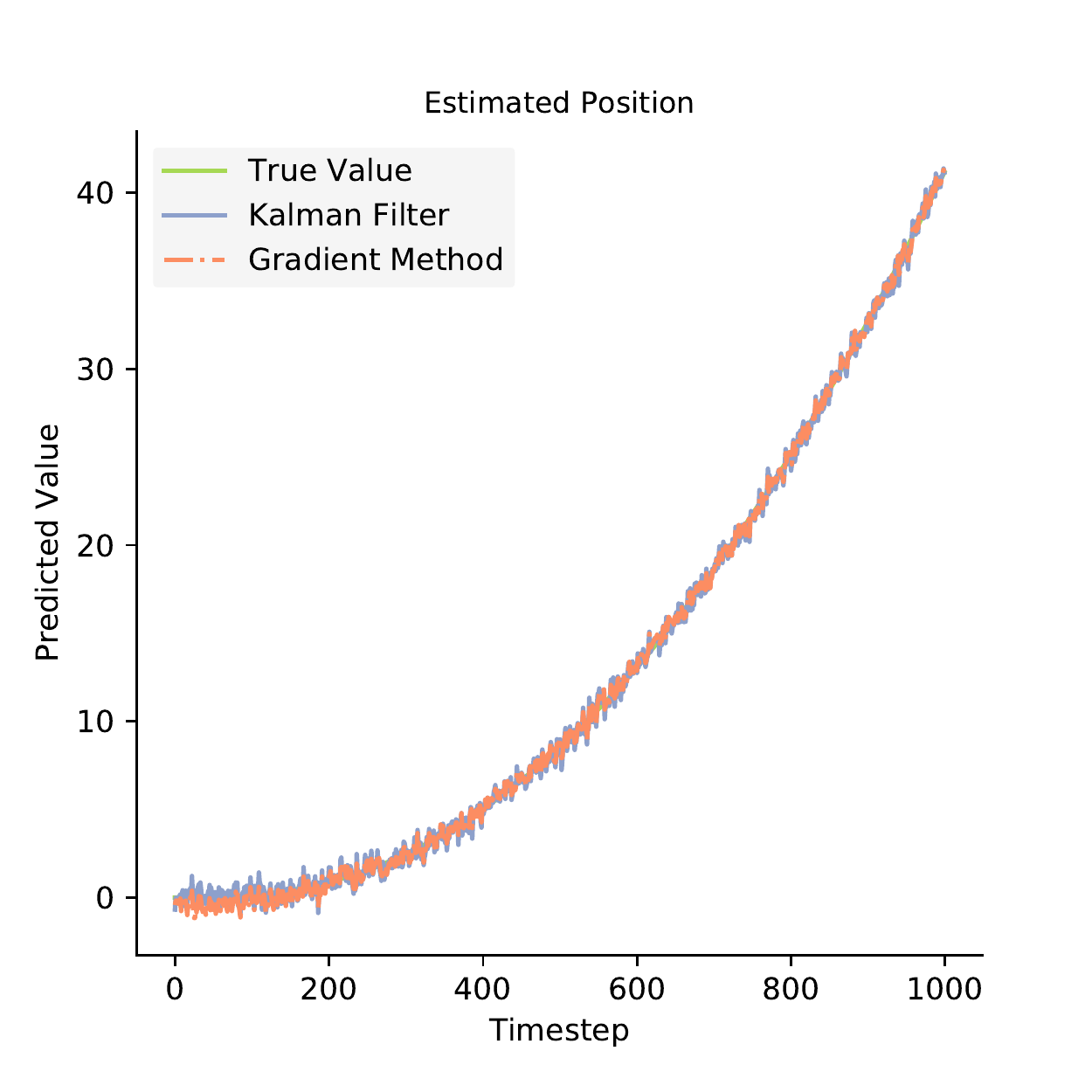}
 \caption{Position}
 \end{subfigure}%
 \begin{subfigure}{0.49\textwidth}
 \centering
 \includegraphics[width=.95\linewidth]{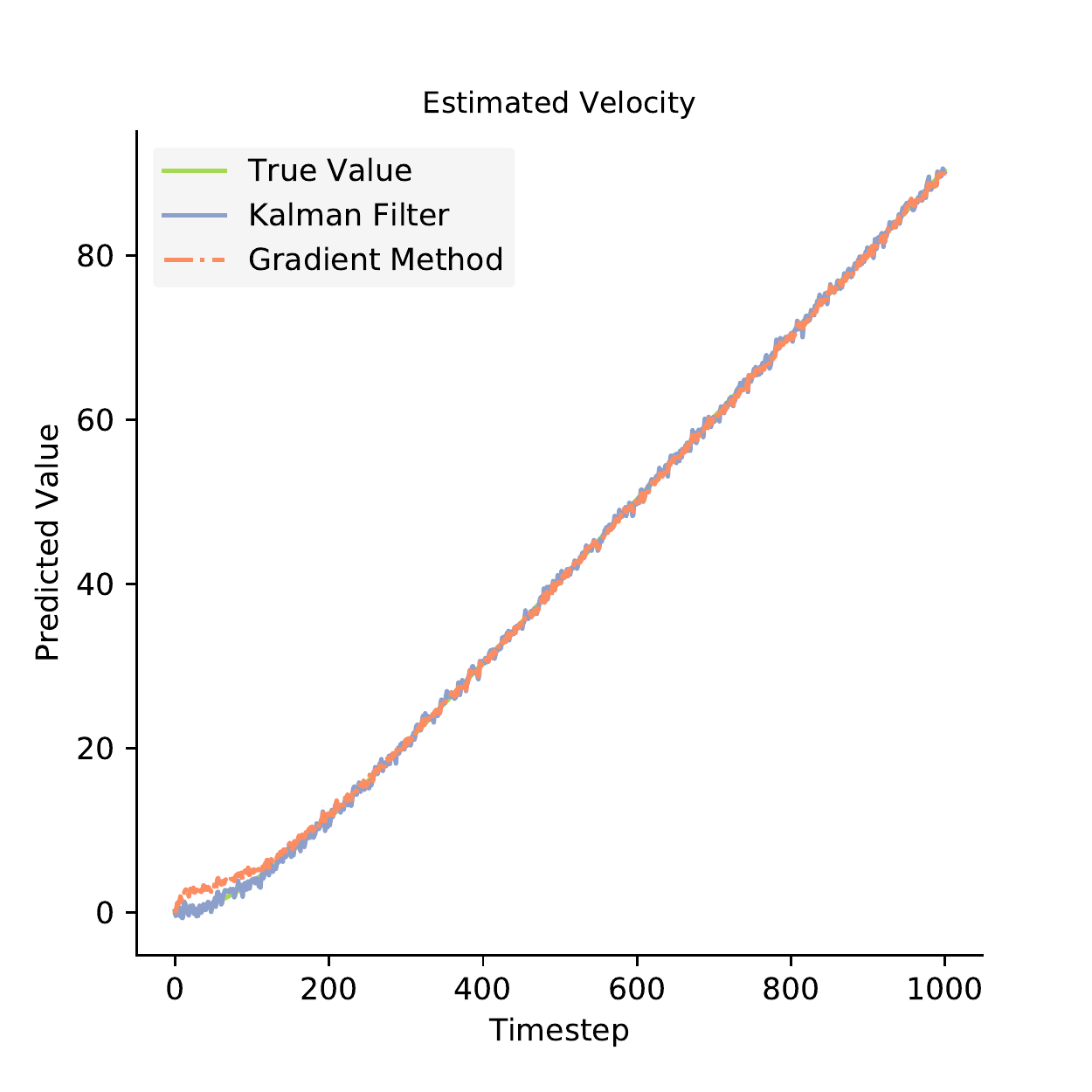}
 \caption{Velocity}
 \end{subfigure}
 \begin{subfigure}{0.49\textwidth}\quad
 \centering
 \includegraphics[width=.95\linewidth]{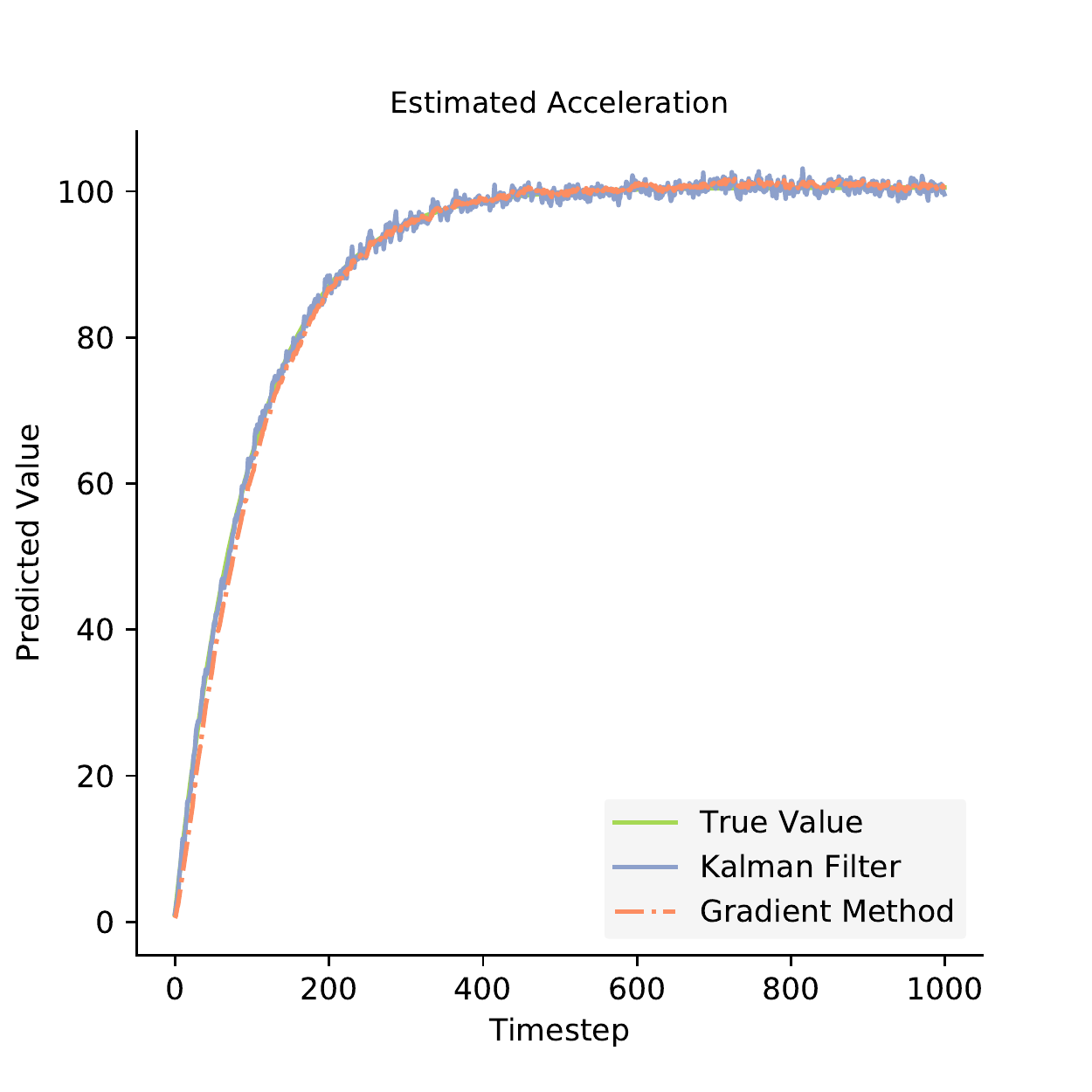}
 \caption{Acceleration}
 \end{subfigure}
 \caption{Tracking performance of our gradient filter compared to the true values and the analytical Kalman Filter.We show the tracking over 2000 timesteps.}
\end{figure}
\begin{figure}
 \begin{subfigure}{0.49\textwidth}
 \centering
 \includegraphics[width=.95\linewidth]{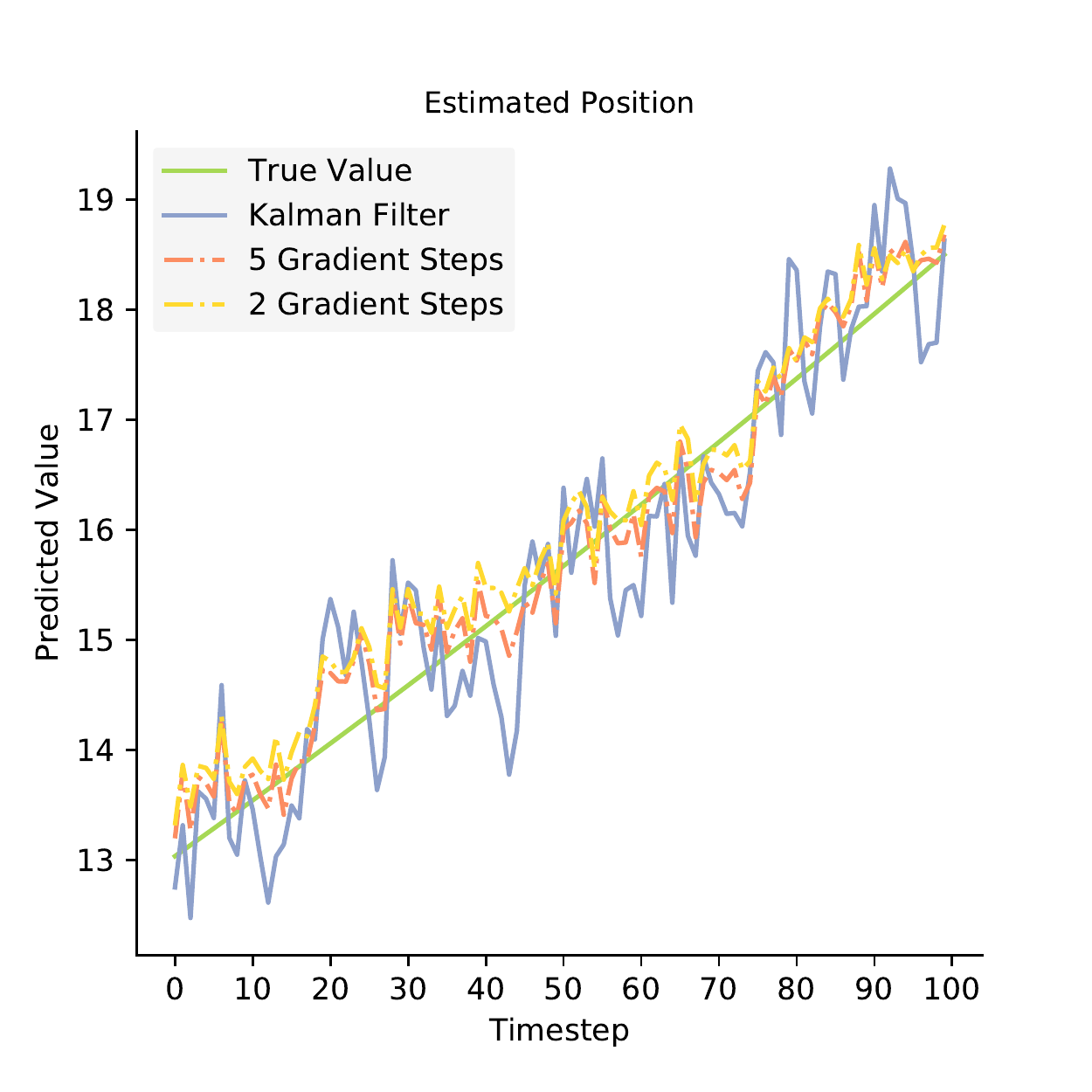}
 \caption{Position}
 \end{subfigure}
 \begin{subfigure}{0.49\textwidth}
 \centering
 \includegraphics[width=.95\linewidth]{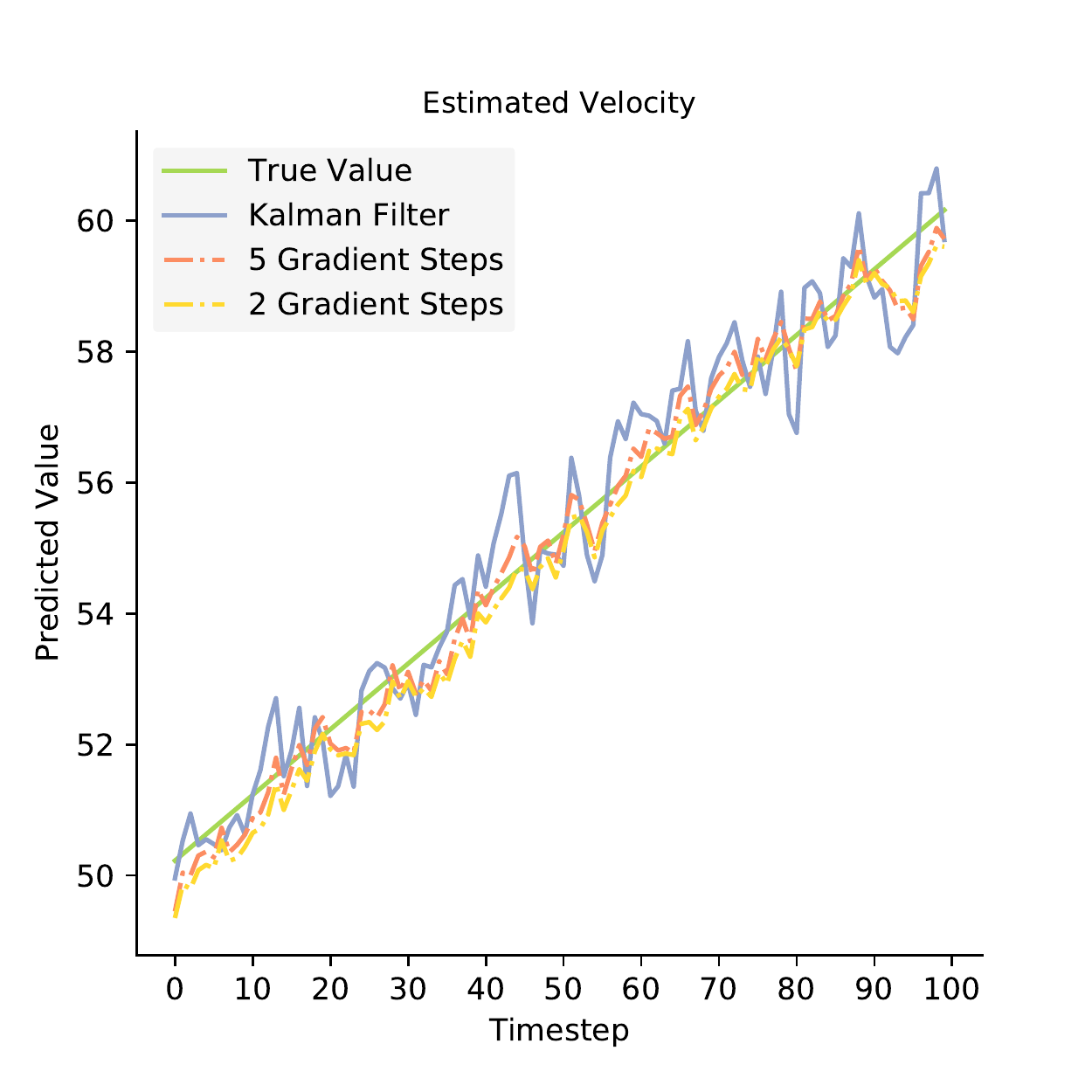}
 \caption{Velocity}
 \end{subfigure}
 \begin{subfigure}{0.49\textwidth}
 \centering
 \includegraphics[width=.95\linewidth]{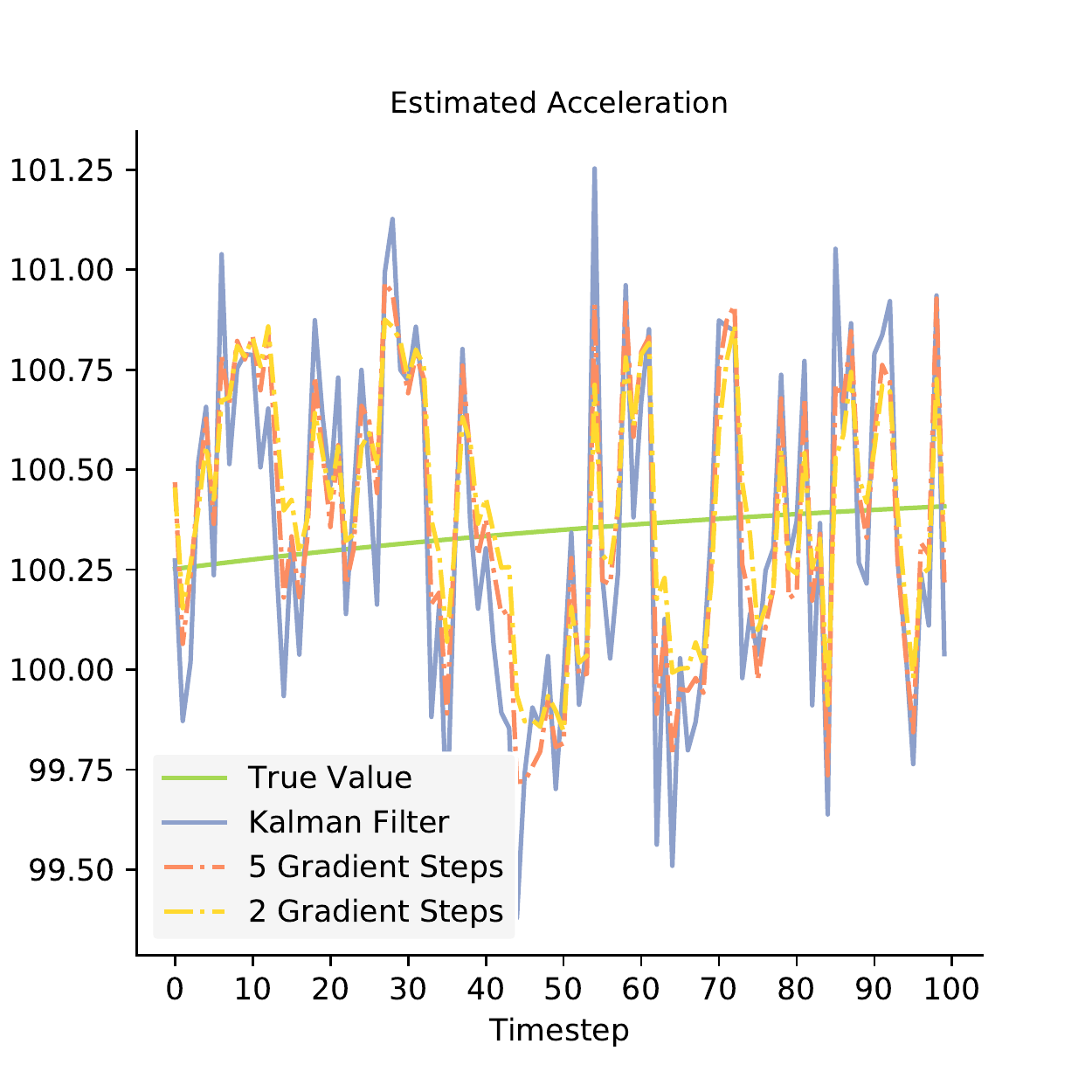}
 \caption{Acceleration}
 \end{subfigure}
 \caption{Here, we zoom in on 100 timestep period to demonstrate tracking performance in miniature and the effect of few gradient updates. In this case, we plotted the estimates after only 5 steps. Even two steps often suffice (with a large learning rate) to provide very accurate estimates}
 
\label{KF_tracking}
\end{figure}

The performance of the analytical Kalman filter which computed updates using equations 1-3 is compared with that of our neural Kalman filter using gradient descent dynamics.\footnote{The code used for these simulations is freely available and online at $https://github.com/Bmillidgework/NeuralKalmanFiltering$} In this comparison the A, B, and C matrices are fixed to their correct values and only the estimated mean is inferred according to Equation \ref{KF_mu}. Comparisons are provided for a number of different gradient steps. As can be seen in Figure \ref{KF_tracking}, only a small number (5) of gradient descent steps are required to obtain performance very closely matching the analytical result. This is likely due to the convexity of the underlying optimization problem, and means that using gradient descent for "online" inference is not prohibitively slow. The simulation also shows the estimate for too few (2) gradient steps for which results are similar, but the estimate may be slightly smoother.

\begin{figure}
 \begin{subfigure}{0.49\textwidth}
 \centering
 \includegraphics[width=.95\linewidth]{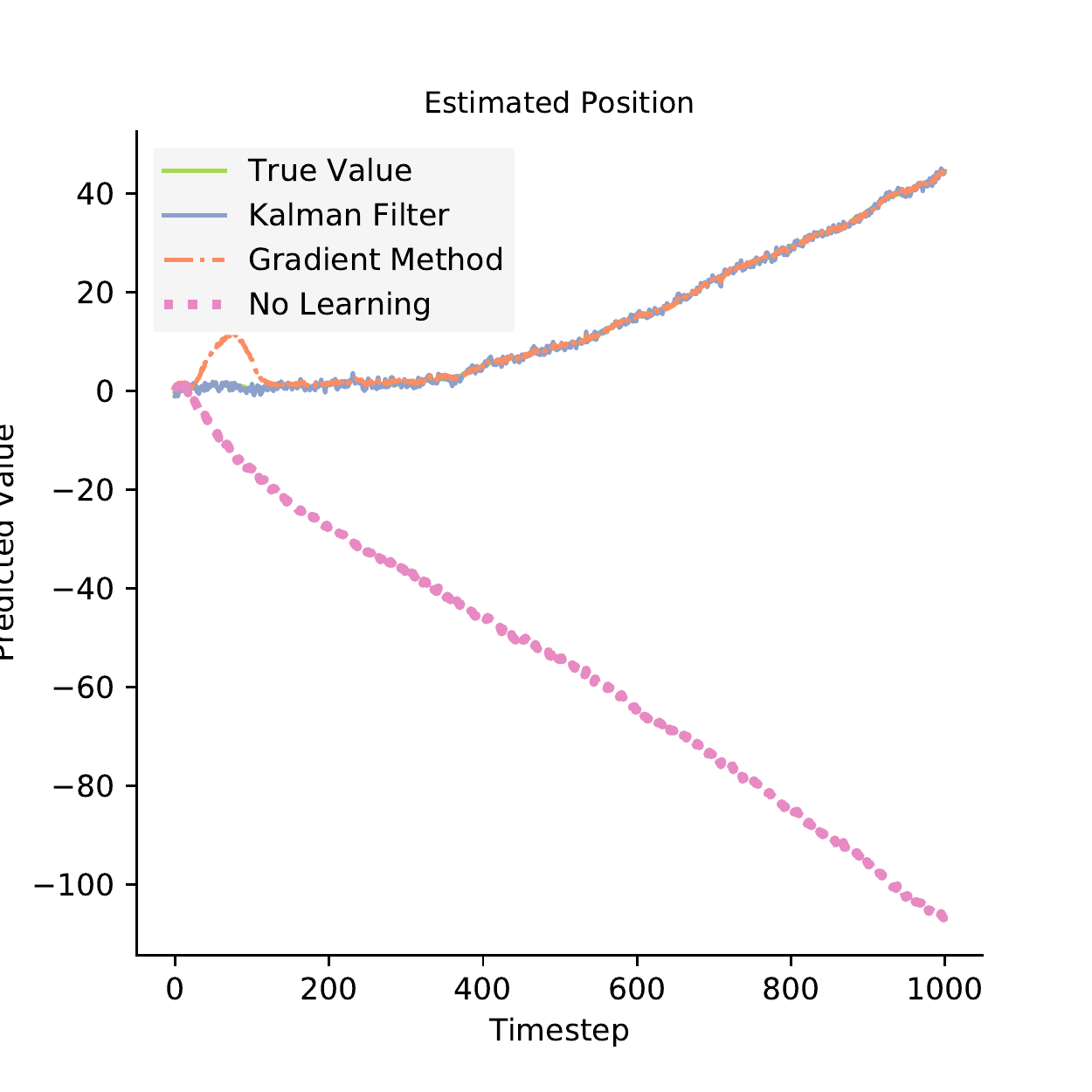}
 \caption{Position: learnt A and B matrices}
 \end{subfigure}
 \begin{subfigure}{0.49\textwidth}
 \centering
 \includegraphics[width=.95\linewidth]{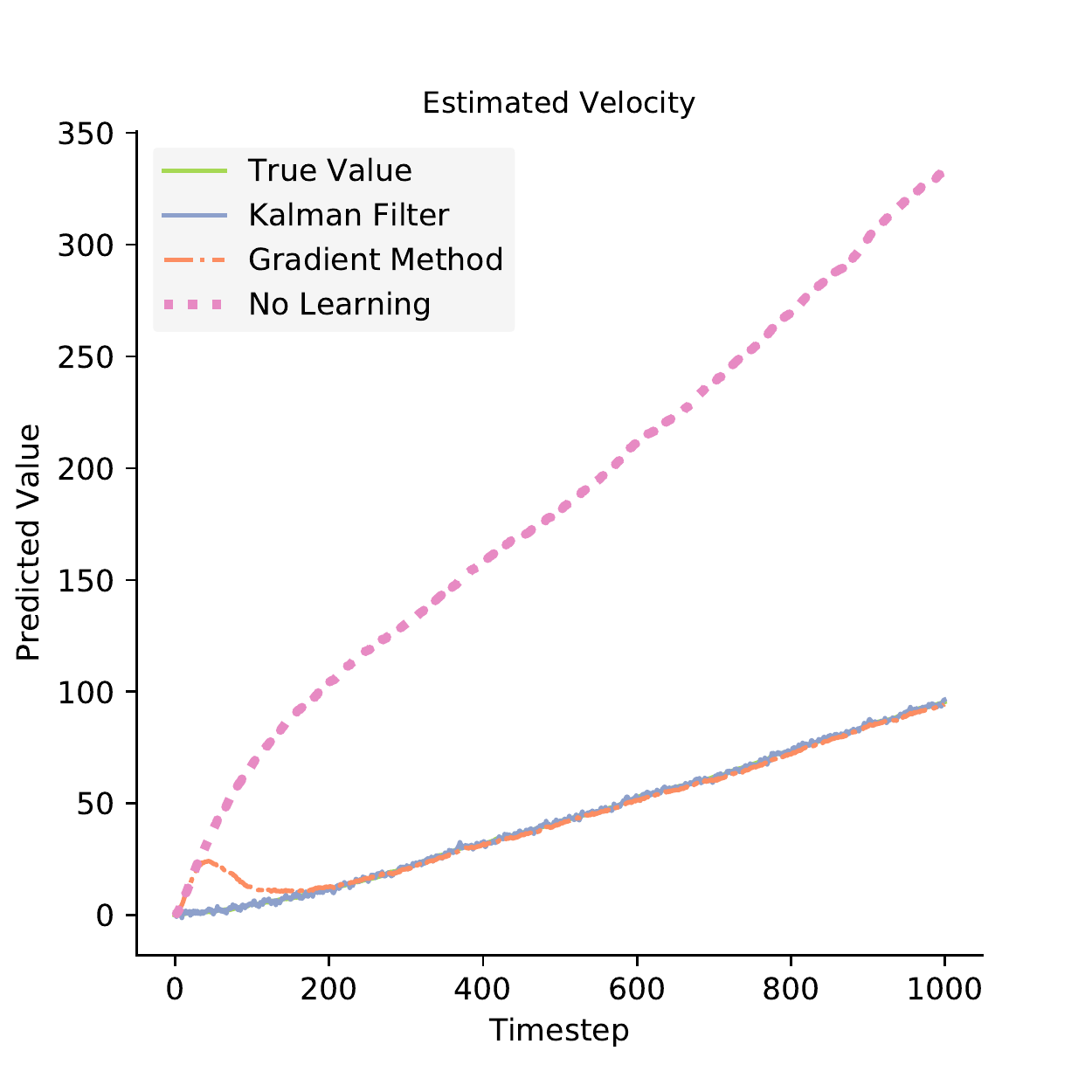}
 \caption{Velocity: learnt A and B matrices}
 \end{subfigure}
 \begin{subfigure}{0.49\textwidth}
 \centering
 \includegraphics[width=.95\linewidth]{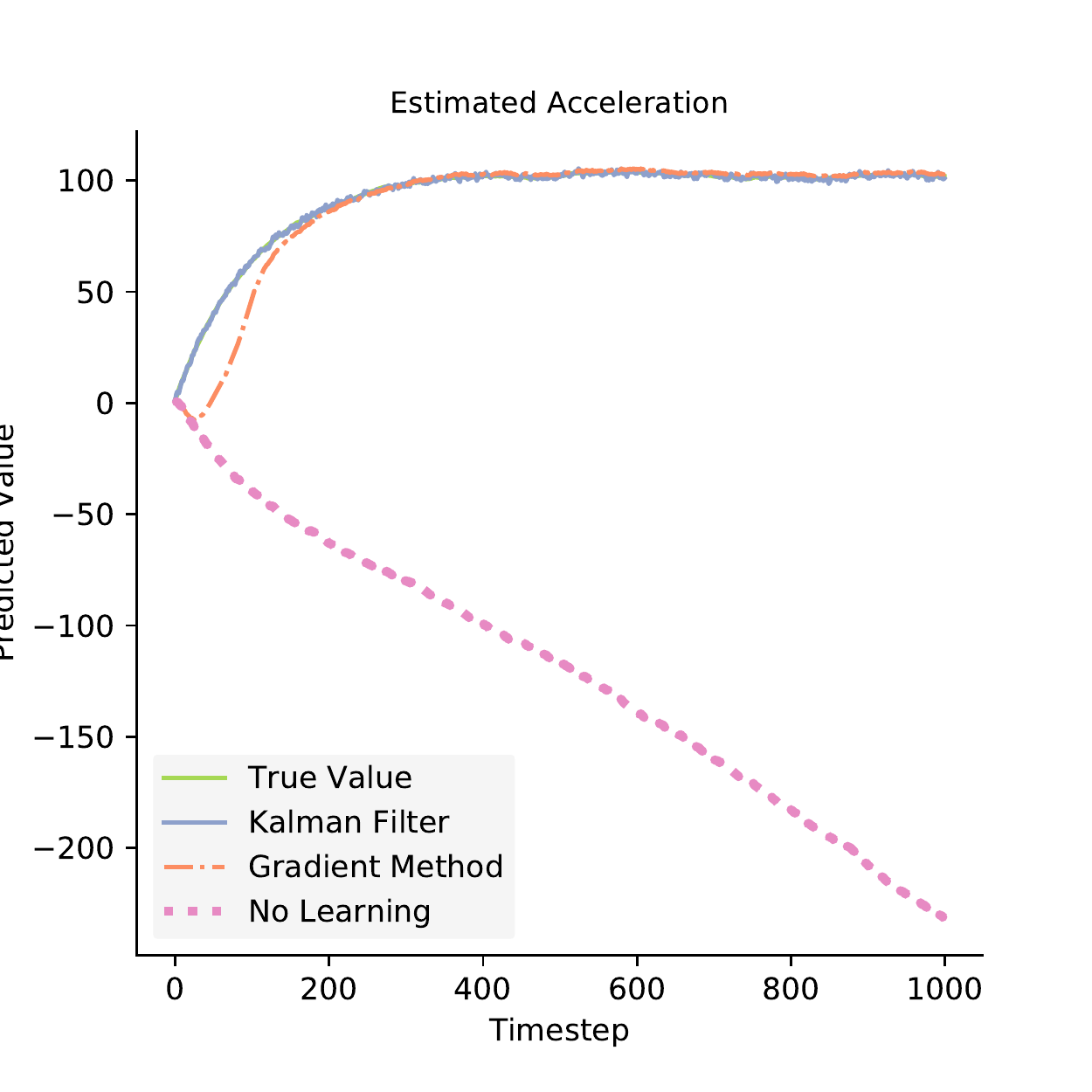}
 \caption{Acceleration: learnt A and B matrices}
 \end{subfigure}
 \caption{Filtering performance for adaptively learning both the A and B matrices in concert (second row). The filtering behaviour of the of the randomly initialized filters without adaptive learning is also shown. Importantly, with learning the estimated position, velocity, and accelerations track their true values precisely while the estimates without learning and just randomly initialized A or A and B matrices rapidly diverge from the truth.}
\label{KF_learn_AB_figure}
\end{figure}

Next, we demonstrate the adaptive capabilities of our algorithm. In Figure \ref{KF_learn_AB_figure}, we show the performance of our algorithm in predicting the position, velocity, and acceleration of the body when provided with a faulty A matrix. Using Equation \ref{KF_A}, our model learns the A matrix online via gradient descent. To ensure numerical stability, a very small learning rate of $10^{-5}$ must be used. The entries of the A matrix given to the algorithm were initialized as random Gaussian noise with a mean of 0 and a standard deviation of 1. The performance of the algorithm without learning the A matrix is also shown, and estimation performance is completely degraded without the adaptive learning. The learning process converges remarkably quickly. It is interesting, moreover, to compare the matrix coefficients learned through the Hebbian plasticity to the known true coefficients. Often they do not match the true values, and yet the network is able to approximate Kalman filtering almost exactly. Precisely how this works is an area for future exploration. 
If a system similar to this is implemented in the brain, then this could imply that the dynamics model inherent in the synaptic weight matrix should not necessarily be interpretable.

We also show (second row of Figure \ref{KF_learn_AB_figure}) that, perhaps surprisingly, both the A and B matrix can be learned simultaneously. In the simulations presented below, either only the A matrix, or both the A and the B matrix were initialized with random Gaussian coefficients, and the network learned to obtain accurate estimates of the hidden state in these cases. \footnote{The results of only learning the B matrix were extremely similar for that of the A matrix. For conciseness, the results were not included. Interested readers are encouraged to look at the $NKF_AB_matrix.ipynb$ file in the online code where these experiments were run.}

\begin{figure}[H]
 \begin{subfigure}{0.49\textwidth}
 \centering
 \includegraphics[width=.95\linewidth]{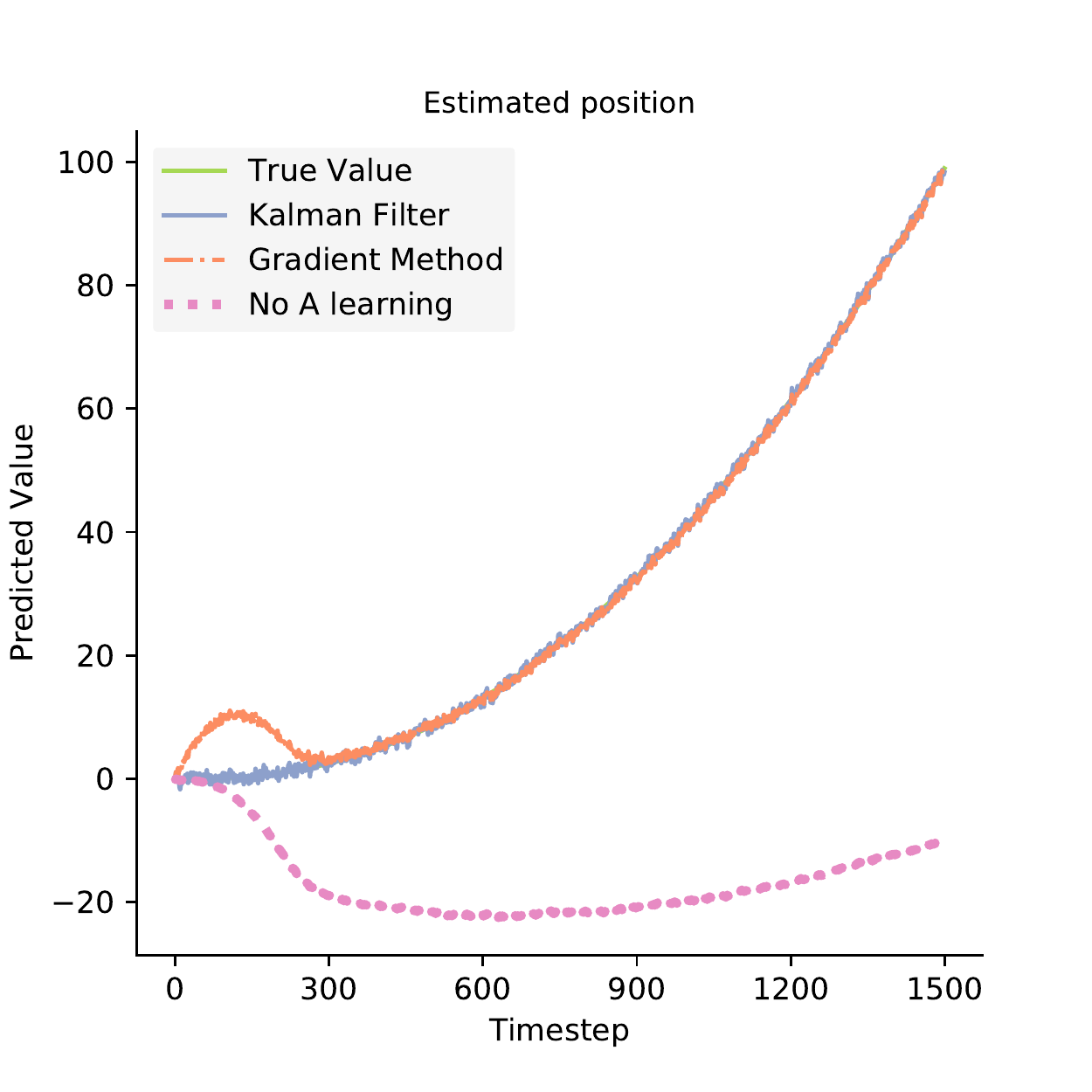}
 \caption{Position for learnt A matrix}
 \end{subfigure}%
 \begin{subfigure}{0.49\textwidth}
 \centering
 \includegraphics[width=.95\linewidth]{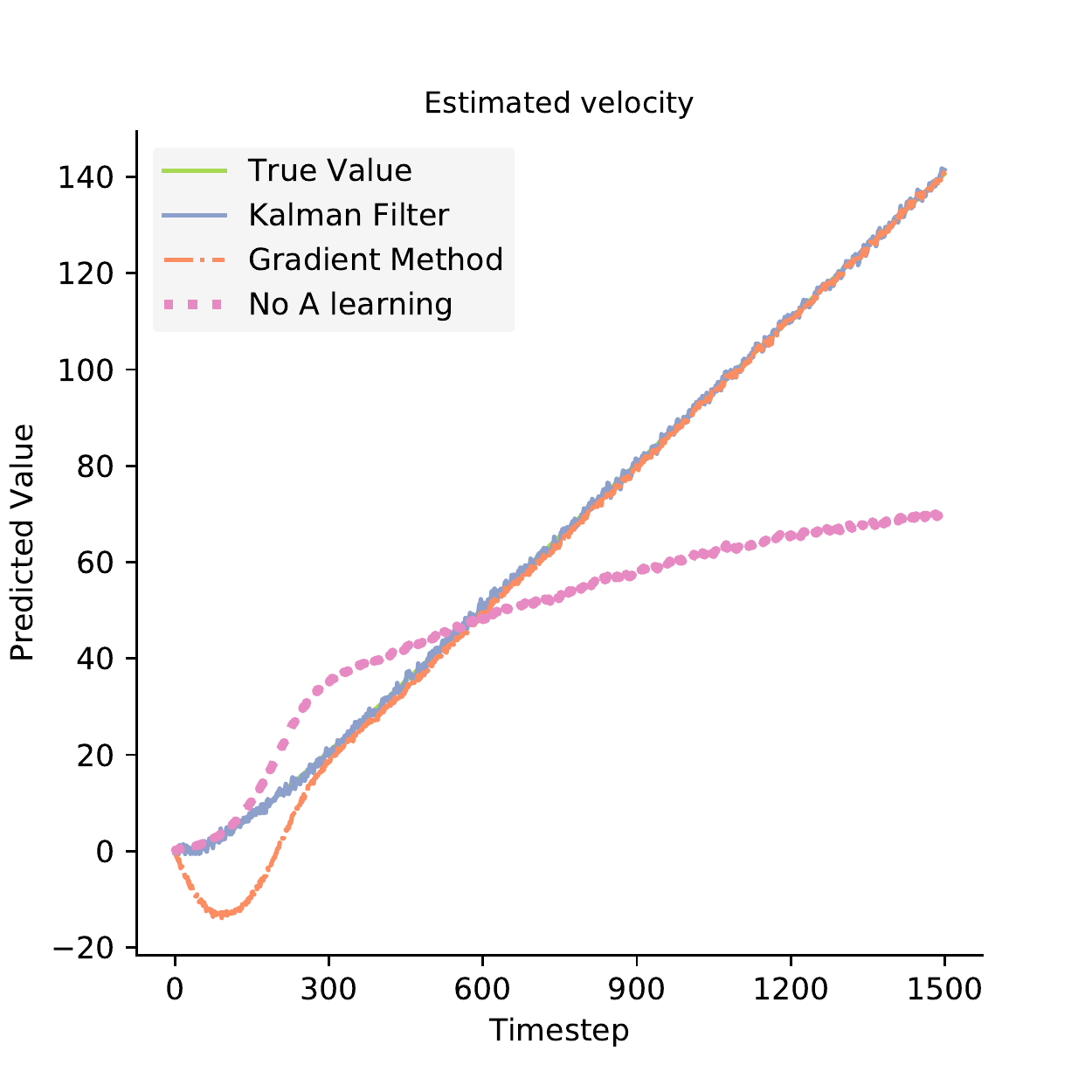}
 \caption{Velocity for learnt A matrix}
 \end{subfigure}
 \begin{subfigure}{0.49\textwidth}\quad
 \centering
 \includegraphics[width=.95\linewidth]{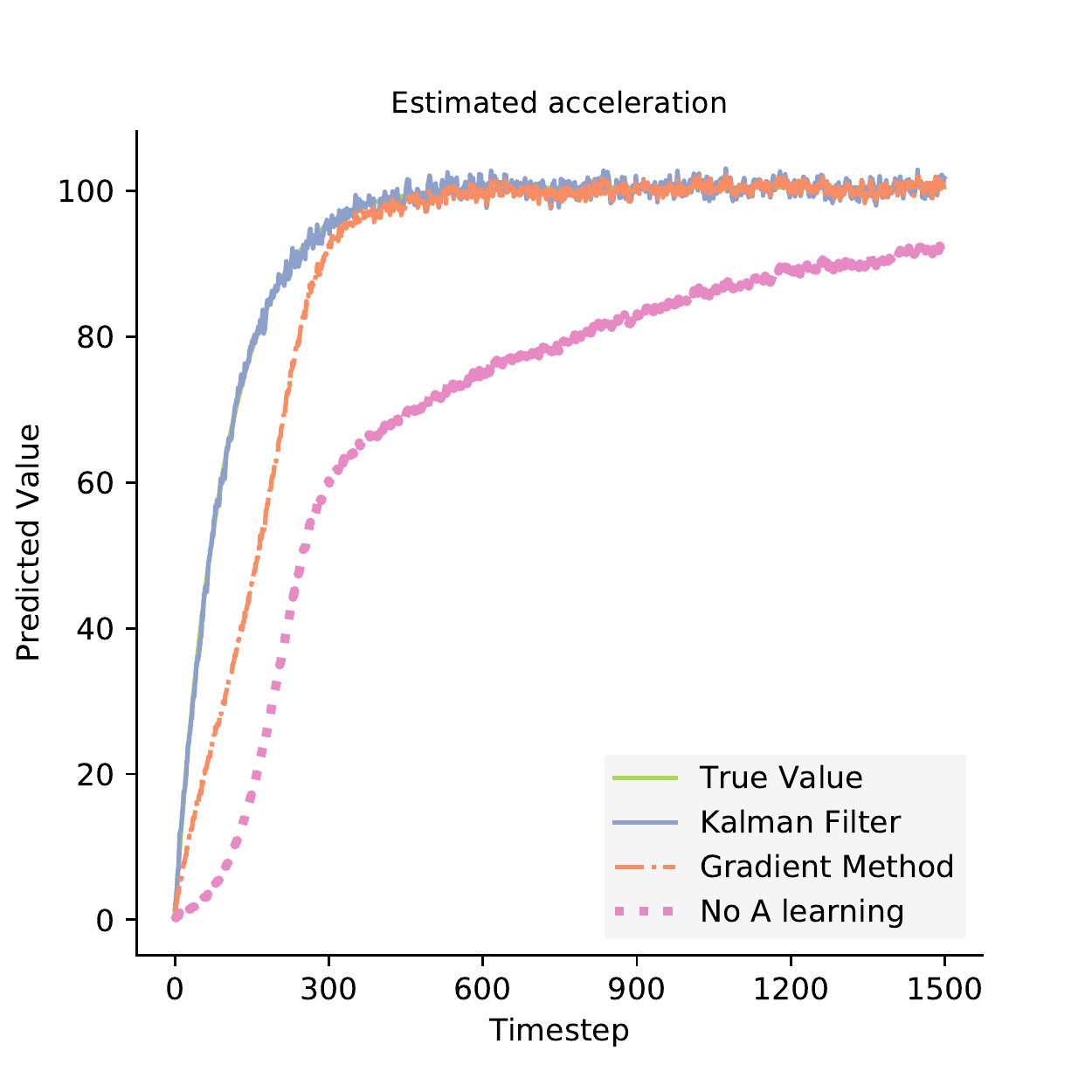}
 \caption{Filtering performance for adaptively learning just the A matrix.}
 \end{subfigure}
 \end{figure}

\begin{figure}
 \begin{subfigure}{0.49\textwidth}
 \centering
 \includegraphics[width=.95\linewidth]{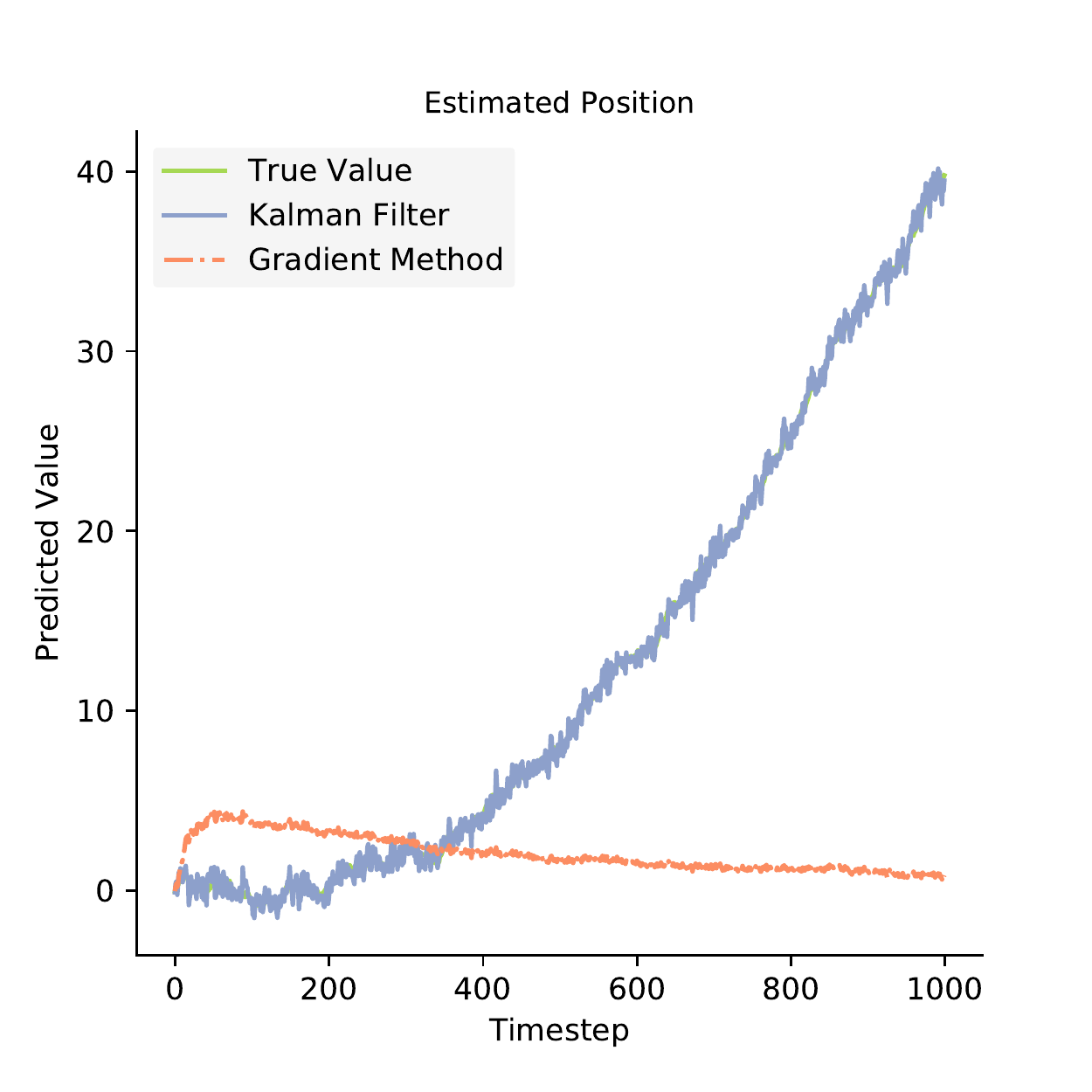}
 \caption{Position: learnt C matrix}
 \end{subfigure}%
 \begin{subfigure}{0.49\textwidth}
 \centering
 \includegraphics[width=.95\linewidth]{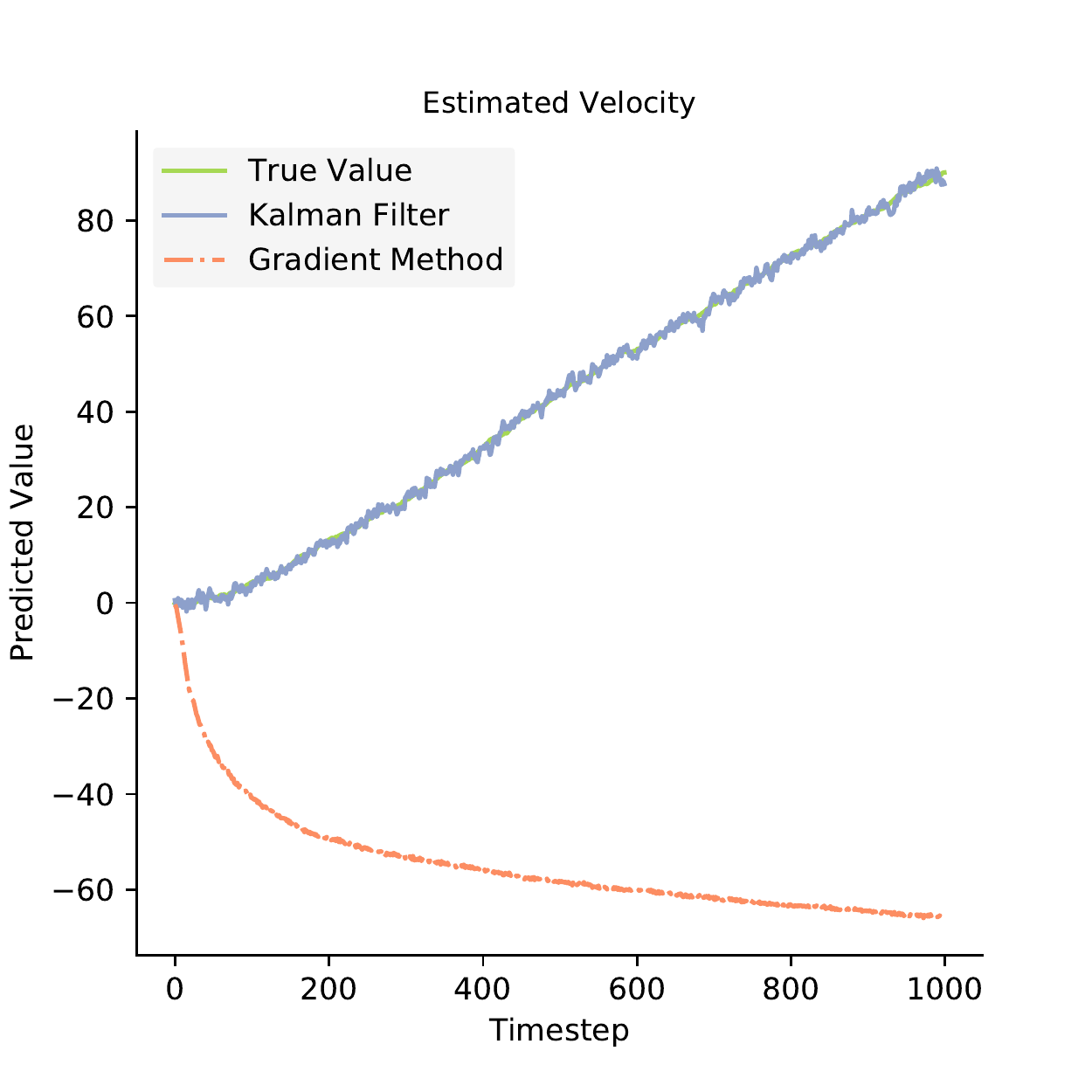}
 \caption{Velocity: learnt C matrix}
 \end{subfigure}
 \begin{subfigure}{0.49\textwidth}\quad
 \centering
 \includegraphics[width=.95\linewidth]{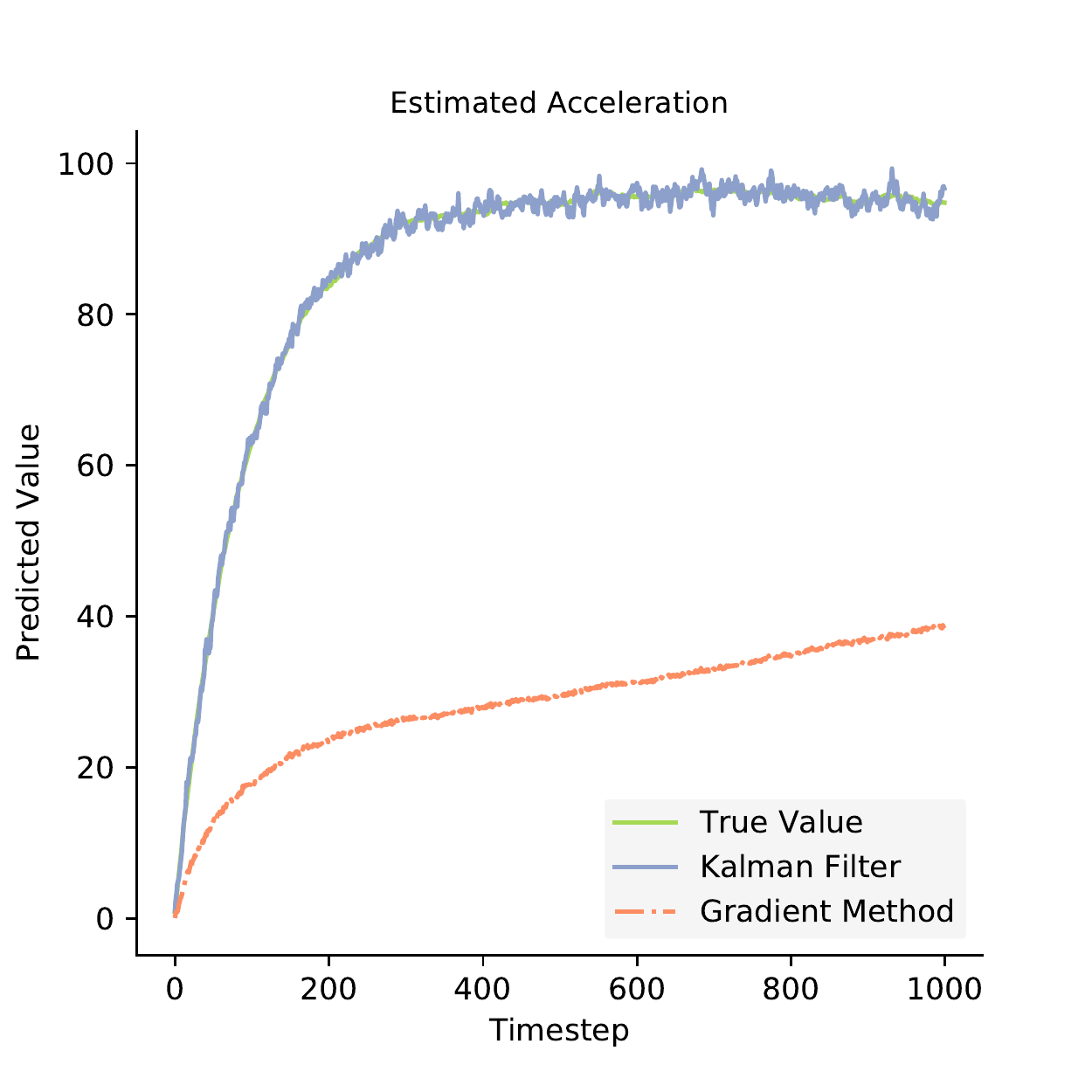}
 \caption{Acceleration: learnt C matrix}
 \end{subfigure}
 \begin{subfigure}{0.49\textwidth}
 \centering
 \includegraphics[width=.95\linewidth]{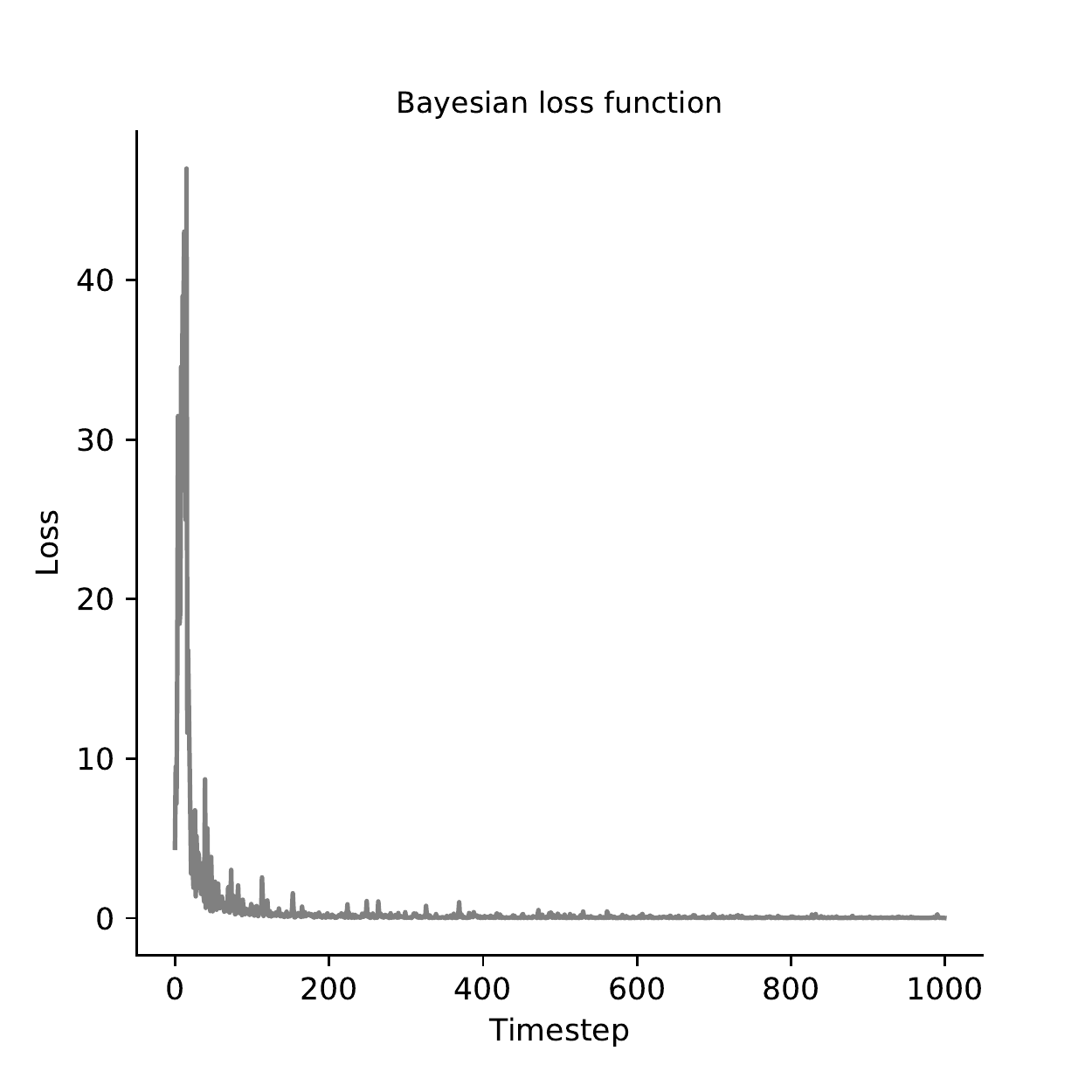}
 \caption{Loss function over timesteps}
 \end{subfigure}
 
 \caption{Very poor tracking behaviour with a learnt C matrix. This is despite the fact that the Bayesian loss function rapidly decreases to a minimum. This shows that the filter can find a prediction-error minimizing "solution" which almost arbitrarily departs from reality if the C-matrix is randomized. Panel D shows the loss computed by the network which rapidly declines, even while the predictions rapidly diverge from the truth (Panels a,b,c)}
 
\label{KF_learn_C_figure}
\end{figure}

We also tried adaptively learning the $C$ matrix using Equation \ref{KF_C}, but all attempts to do so failed. Although the exact reason is unclear, we hypothesise that an incorrect $C$ matrix corrupts the observations which provides the only "source of truth" to the system. If the dynamics are completely unknown but observations are known, then the true state of the system must be at least approximately near that implied by the observations, and the dynamics can be inferred from that. On the other hand, if the dynamics are known, but the observation mapping is unknown, then the actual state of the system could be on any of a large number of possible dynamical trajectories, but the exact specifics of which are underspecified. Thus the network learns a C matrix which corresponds to some dynamical trajectory, which succeeds in minimizing the loss function, but which is completely dissimilar to the actual trajectory the system undergoes. This can be seen by plotting the loss obtained according to Equation \ref{KF_MAP} in Figure \ref{KF_learn_C_figure}, which rapidly decreases, although the estimate diverges from the true values.

\subsection{Discussion}

Here we have elucidated the precise relationship between Kalman filtering -- an optimal linear Bayesian filtering algorithm -- and predictive coding -- a neurophysiologically realistic theory of cortical function. Specifically, that they both optimize the same objective function -- a Bayesian MAP filtering objective -- while the Kalman filtering solves the resulting optimization problem analytically, predictive coding approaches derive their dynamics from a gradient descent on the same objective. "This result provides a new perspective on predictive coding; especially if we make the simplifying assumption that the precisions are not optimised with respect to variational free energy – or, in the examples above, we assume the conditional covariance is zero. This reduces variational inference to a MAP optimisation problem. This follows due to the Laplace approximation, which effectively means the variational precision can be derived analytically from the expectation (from the curvature of the log likelihood at the expectation) – see \citep{friston2007free} for details. Alternatively, we can just ignore the conditional uncertainty as in the MAP optimisation perspective.


Our work also demonstrates how straightforwardly predictive coding can be applied to solve Bayesian \emph{filtering} problems rather than simply static Bayesian inference problems. Since the brain is enmeshed in continuous sensory exchange with a constantly moving world, filtering is a much more realistic challenge to solve than pure inference on a static dataset. It thus seems likely that the neural circuitry dedicated to perception is specialized for solving precisely these sorts of filtering problems. Moreover, due to predictive coding's biologically realistic properties, our results provide a powerful biologically plausible approach for how the brain might solve such filtering problems.

Nevertheless, there remain several deficiencies of our algorithm (and predictive coding more generally) in terms of biological plausibility which it is important to state. Our model assumes full connectivity for the `diffuse' connectivity required to implement matrix multiplications. Additionally in other cases it requires one-to-one excitatory connectivity, both constraints which are not fully upheld in neural circuitry. Additionally, in one case (that of the "C matrix" between the populations of neurons representing the estimate and the sensory prediction errors), we have assumed a complete symmetry of backward and forward weights, such that the connections which embody the C matrix downwards also implement the $C^T$ matrix when traversing upwards. This is also a constraint not satisfied within the brain. Additionally, our model can represent negative numbers in states or prediction errors, which rate-coded neurons cannot. Several of these implausibilities will be directly addressed in the context of (static) predictive coding later in this chapter.

We believe, however, that despite some lack of biological plausibility, our model is useful in that it shows how a standard engineering algorithm can be derived in a way more amenable to neural computation, and provides a sketch at how it could be implemented in the brain. Moreover, we hope to draw attention to Bayesian filtering algorithms and how they can be implemented neurally, instead of just Bayesian inference on static posteriors. 

Finally, while our algorithm and experiments have only considered the linear case, it can be straightforwardly extended to the nonlinear case, where it results in standard nonlinear predictive coding as discussed previously. Explicitly and empirically comparing the performance of our algorithm against nonlinear extensions to the Kalman filter such as extended or unscented \citep{wan2000unscented} Kalman filtering are an important and exciting avenue for future work.


\section{Relaxed Predictive Coding}

In the literature predictive coding has been proposed as a general theory of cortical function \citep{friston2003learning,friston2005theory,friston2008hierarchical,kanai2015cerebral,spratling2008reconciling}. There is additionally a small literature of process theories which try to translate the mathematical formalism into purported neural circuitry \citep{bastos2012canonical,keller2018predictive,kanai2015cerebral}, and some of the predictions of predictive coding have been extensively compared and evaluated against neurophysiological data \citep{walsh2020evaluating,friston2008hierarchical,huang2011predictive,clark2015surfing,aitchison2017or}. Despite the general acceptance of predictive coding as a biologically plausible algorithm which could in theory be implemented in the brain, there nevertheless are several highly implausible aspects of the core algorithm that have been largely glossed over in the formulations of the process theories, which focused primarily on macro-scale connectivity constraints instead of the precise mathematical form of the learning and update rules in the algorithm \citep{bastos2012canonical}. Here, we introduce three potentially severe biological implausibilities which emerge directly from the form of the predictive coding algorithm and demonstrate empirically how, with some ingenuity and adaptation of the algorithm, these implausible assumptions can be `relaxed' without major damage to the empirical performance of predictive coding networks on object recognition tasks. This work is based on \citep{millidge2020relaxing}

Recall, that the core of the predictive coding formalism is three key relationships. First, the concept of prediction error as the difference between the activity of the neurons in a layer and the top-down predictions from higher layers. Second, the update rule for the activities of a layer, which minimizes both the prediction errors at its own layer, as well as the layer below. And thirdly, the learning rule for the weights, as a local Hebbian function of the prediction errors at their own layer \citep{friston2005theory}.
\begin{align*}
 \label{relaxed_PC_rules}
 \epsilon_l &= \mu_l - f(\theta_{l+1} \mu_{l+1}) \\
 \frac{d\mu_l}{dt} &= -\epsilon_l + {\theta_l}^T \epsilon^{l-1} f'(\theta_l \mu_l) \\
 \frac{d\theta_l}{dt} &= \epsilon^{l-1} f'(\theta_l \mu_l) {\mu_l}^T \numberthis
\end{align*}

where $f'(\theta_l \mu_l)$ represents the partial derivative of the post-activations with respect to either the $\mu$ or the $\theta$ depending upon the update rule. Equation\ref{relaxed_PC_rules} states that prediction errors are computed as a simple subtraction of the value neurons at a layer and the prediction from the layer above. The vector $\mu_l$ represents the activity of the value neurons at a specific level $l$. The vector $\epsilon_l$ is a vector of the activity of the error neurons at a level l. Predictions are mapped down from the higher layers through a set of weights, denoted $W$ which is an $M \times N$ matrix where M is the number of neurons at level $l$ and N is the number of neurons at level $l+1$. $f(x)$ is a nonlinear activation function applied to the outputs of a neuron and $f'(x) = \frac{\partial f(x)}{\partial x}$ is the pointwise derivative of the activation function \footnote{Here we have specialized the predictive coding update rules somewhat from any arbitrary function $f$ to an elementwise nonlinearity followed by a multiplication with a weight matrix $\theta$. This is because in this section we consider the application of predictive coding to train artificial neural networks with this specific type of structure}.

Equation \ref{PC_hierarchical_mu} specifies the update rule for the $\mu_l$ at a specific layer. The update is equal to the sum of the prediction errors projected up from the layer below, multiplied by the top down predictions and projected back through the weight matrix and subtracted from the prediction errors at the current layer. This is a biologically plausible learning rule as it is a simple sum of multiplication of locally available information. Note: the the update includes prediction error terms from both the current layer and the layer below. This equation is why it is necessary to transmit prediction errors upwards.

Equation \ref{PC_hierarchical_theta} is the update rule for the weights $\theta$. This obeys Hebbian plasticity since it is simply a multiplication of the two quantities available at each end of the synaptic connection -- the prediction error of the layer below and the value neurons at the current layer. The only slight difficulty is the derivative of the nonlinear activation function of the prediction. While this information is locally available in principle, it requires a somewhat more complex neural architecture and it is not certain that the derivatives of activation functions can be computed straightforwardly by neurons. Luckily, we show below that this term is not needed for successful operation of the learning rule.

Importantly, we additionally recall that predictive coding can be considered to be a variational inference algorithm, as Equations \ref{PC_hierarchical_mu} and \ref{PC_hierarchical_theta} can be directly derived as a gradient descent upon the variational free energy $\mathcal{F} = \sum_{i=0}^L \epsilon_i^2$ which (under Gaussian assumptions) takes the form of a simple sum of squared prediction errors at each layer. We additionally ignore the precision parameters $\Sigma_l$ in this analysis since in general their biological plausibility has not been strongly analyzed in the literature (although there are some speculative suggestions linking them to either lateral connectivity \citep{friston2005theory} or else subcortical activity in the pulvinar \citep{kanai2015cerebral}).

Specifically, we focus on three important implausibilities. The first is the problem of weight symmetry, or the required equality of forward and backward weights. This problem is often called the \emph{weight transport problem} in the literature \citep{lillicrap2016random,crick1989recent,lillicrap2020backpropagation}. The learning rules in these networks require information to be sent `backwards' through the network. Since synaptic connections are generally assumed to be uni-directional, in practice this means that these backward messages need to be sent through a second set of backwards connections with the exact same synaptic weights as the forward connections. Clearly, expecting the brain to have an identical copy of forward and backward weights is infeasible. Mathematically, this problem arises from the $\theta_l^T$ term in the dynamics equation for the $\mu$s (Equation \ref{PC_hierarchical_mu}), since this weight transpose uses the forward weight matrix $\theta$ but instead maps the bottom-up prediction error to the level above, thus requiring information to be sent `backwards' or `upwards' through `downwards' connections. In the brain, this would require information to propagate backwards from the soma of the post-synaptic neuron, back through the axon and to the soma of the pre-synaptic cell -- a possibility which is considered to be extremely implausible \citep{lillicrap2014random}. Here, we address this problem in predictive coding networks by using a separate set of randomly initialized backwards weights trained with a separate Hebbian learning rule, which also only requires local information. This removes the necessity of beginning with symmetrical or identical weights and proposes a biologically plausible method of learning good backward weights from scratch in an unsupervised fashion. In the brain this would be implemented as a reciprocal set of `backwards' connections going from the lower-layers to higher layers, which are definitely present in the brain \citep{grill2004human}.

The weight transport problem is also present in neural implementations of the backpropagation of error algorithm from machine learning, and there exists a small literature addressing it within this context. A key paper \citep{lillicrap2014random,lillicrap2016random} shows that simply using random backwards weights is sufficient for some degree of learning. This method is called feedback alignment (FA) since during training the feedforward weights learn to align themselves with the random feedback weights so as to transmit useful gradient information. A variant of this -- direct feedback alignment (DFA) \citep{nokland2016direct} has been shown that direct forward-backward connectivity is not necessary for successful learning performance. Instead, all layers can receive backwards feedback directly from the output layer. It has also been shown \citep{liao2016important} that performance with random weights is substantially improved if the forward and backward connections share merely the same sign, which is less of a constraint than the exact value. One further possibility is to learn the backwards weights with an independent learning rule. This has been proposed independently in \citet{amit2019deep} and \citet{akrout2019deep} who initialize the backwards weights randomly, but train them with some learning rule. Our work here differs primarily in that we show that this learning rule works for predictive coding networks while they only apply it to deep neural networks learnt with backprop. Moreover, our Hebbian learning rule can be straightforwardly derived in a mathematically principled manner as part of the overarching variational framework of predictive coding. 

The second problem is one of backward nonlinear derivatives. In predictive coding networks (along with backprop), the update and learning rules require the pointwise derivatives of the activation function to be computed at each neuron. Mathematically, this is the $f'(\theta_l \mu_l)$ term. For individual biological neurons, while a nonlinear forward activation function is generally assumed, the ability to compute the derivative of the activation function is not known to be straightforward. While in some cases this issue can be ameliorated by a judicious choice of activation function -- for instance the pointwise derivative of a rectified linear unit is simply 0 or 1, and is a simple step function of the firing rate -- the problem persists in the general case. Here, we show that, somewhat surprisingly, it is possible to simply ignore these pointwise derivatives with relatively little impact on learning performance, despite the update rules now being mathematically incorrect. This may free the brain of the burden of having to compute these quantities.

A third issue, specific to frameworks that explicitly represent prediction errors, is the requirement of one-to-one connections between activation units and their corresponding error units. While not impossible, this precise, one-to-one connectivity pattern is likely difficult for the brain to create and maintain throughout development and learning. One possibility, as explored by \citet{sacramento2018dendritic} is that prediction errors and predictions may be housed in separate dendritic compartments on a single neuron, thus potentially obviating this issue (although their scheme relied on another set of one-to-one connections between pyramidal cells and inhibitory interneurons). However, the plausibility of this idea in terms of actual dendritic morphology and neurophysiology is unclear. Instead, here we present a network-level solution and show that learning can continue unaffected despite random connectivity patterns between value and error units as long as these connection weights can also be learned. We propose a further Hebbian and biologically plausible learning rule to update these weights which also only requires local information. Finally, we experiment with combining our solutions to all of these problems together to produce a fully relaxed predictive coding architecture. Importantly, this architecture possesses a simple bipartite but otherwise fully connected connectivity pattern with separate learnable weight matrices covering every connection, all of which are updated with local Hebbian learning rules. We show that despite the simplicity of the resulting relaxed scheme, that it can still be trained to high classification accuracy comparable with standard predictive coding networks and ANNs using backpropagation.

\subsection{Methods}

To test the performance of the predictive coding network under various relaxations, we utilize the canonical MNIST and FashionMNIST \citep{xiao2017fashion} benchmark datasets. Since this is a supervised classification task, we follow the approach of \citet{whittington2017approximation} and \citet{millidge2020predictive}, who utilized a `reverse' predictive coding architecture where the inputs were presented to the top layer of the network and the labels were predicted at the bottom. This formulation allows for the straightforward representation of supervised learning problems in predictive coding. In effect, the network tries to generate the label from the image. We utilized a 4-layer predictive coding network consisting of 784, 300, 100, and 10 neurons in each layer respectively. We tested both rectified-linear (relu) and hyperbolic tangent (tanh) activation functions, which are the most common activation functions used in machine learning. During training the $\mu$s were updated for 100 iterations using Equation \ref{PC_hierarchical_mu} with both the input and labels held fixed. After the iterations of the $\mu$s, the remaining prediction errors in the network were used to update the weights according to Equation \ref{PC_hierarchical_theta}. 
At test time, a digit image was presented to the network, and the top-down predictions of the network were propagated downwards to produce a prediction at the lowest layer, which was compared to the true label to obtain the test accuracy. 

The network was trained and tested on the MNIST and FashionMNIST datasets. MNIST is a dataset of 60,000 $28 \times 28$ grayscale images of handwritten digits from 0 to 9. The goal is to predict the digit from the image. The FashionMNIST dataset contains images of different types of clothing, which must be sorted into ten classes. FashionMNIST is a more challenging classification dataset, while also serving as a drop-in replacement for MNIST since its data is in exactly the same format. The input images were flattened into a 784x1 vector before being fed into the network. Labels were represented as one-hot vectors and were smoothed using a value of 0.1 for the incorrect labels. The dataset was split into a training set of 50000 and a test set of 10000 images. All weight matrices were initialized as draws from a multivariate Gaussian with a mean of 0 and a variance of 0.05. It is likely, given the large literature on how to best initialize deep neural networks, that there exist much better initialization schemes for predictive coding networks as well, however we did not investigate this here. All results presented were averaged over 5 random seeds. We plot error bars around the means as the standard deviation of the seeds.

\subsection{Results}

\subsubsection{Weight Transport}

\begin{figure}[h]
 \begin{center}
 \includegraphics[width=13cm, height=8cm]{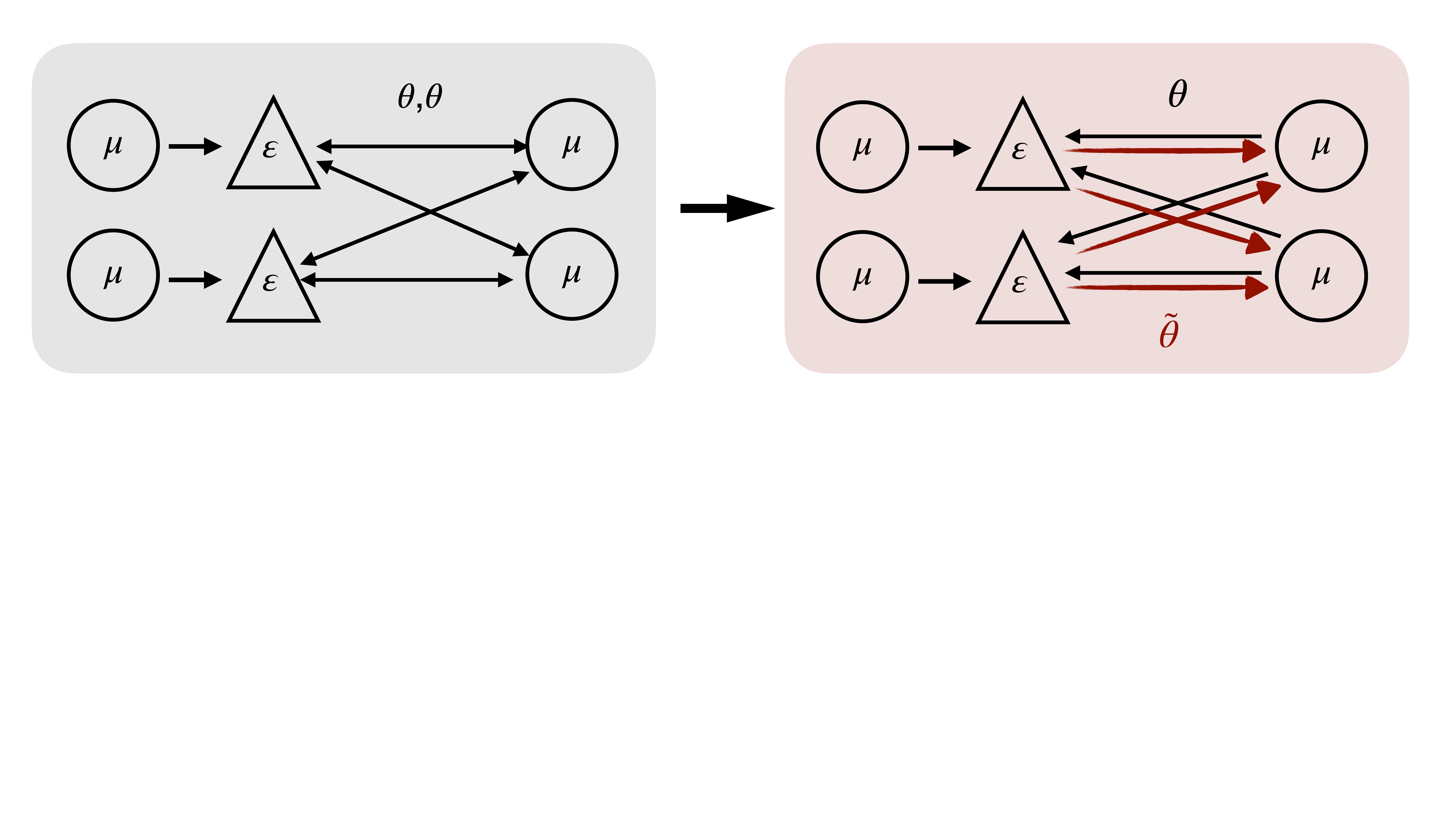}
 \end{center}
 \vspace{-3.9cm}
 \caption{The weight transport problem and our solution. On the left is the standard predictive coding network architecture. Our diagram represents the prediction errors $\epsilon$ of one layer receiving predictions and transmitted prediction errors to the value neurons of the layer above. Prediction errors are transmitted upwards using the same weight matrix $\theta^T$ as the predictions are transmitted downwards. On the right, our solution eschews this biological implausibility by proposing a separate set of backwards weight $\tilde{\theta}$ (in red), which are learned separately using an additional Hebbian learning rule.}
 
 \label{Backward_Weight_Diagram}
 \end{figure}

Mathematically, the weight transport problem is caused by the $\theta^T$ term in Equation \ref{PC_hierarchical_theta}. In neural circuitry this weight transpose corresponds to transmitting the message backwards through the same connections or, alternatively, an identical copy of the backward weights. We wish to replace this copy of the forward weights with an alternative, unrelated set of weights $\tilde{\theta}$. Unlike FA or DFA methods, which simply use random backwards weights, we propose to learn the backwards weights through a simple synaptic plasticity rule. 
\begin{align*}
 \frac{d\tilde{\theta^l}}{dt} = \mu^l (f'(\theta^l \mu^l) \epsilon^{l-1})^T
\end{align*}
This rule is Hebbian since it is just the multiplication of the activities of the units at each end of the connection. The backwards pointwise derivative poses a slight problem in that it is first multiplied with the errors of the level below. However, as we show below, pointwise nonlinear derivatives are not actually needed for good learning performance, so the problem is surmounted. This rule is simply the transposed version of the original weight update rule (Equation \ref{PC_hierarchical_theta}). And thus, if the forward and backwards weights are initialized to the same value, barring numerical error, they will stay the same throughout training. Importantly, we demonstrate here that this rule allows rapid and effective learning of the weights even if the forward and backwards matrices are initialized in a completely independent (and random) fashion. This means that in the brain the forwards and backwards connections originate completely independently, and that, moreover, if we accept the forward weight update rule as plausible, we should accept the backwards weight update as well, thus leading to no greater demands of biological plausibility for this backwards weight update.

\begin{figure}[ht] 
 \begin{subfigure}[b]{0.5\linewidth}
 \centering
 \includegraphics[width=0.75\linewidth]{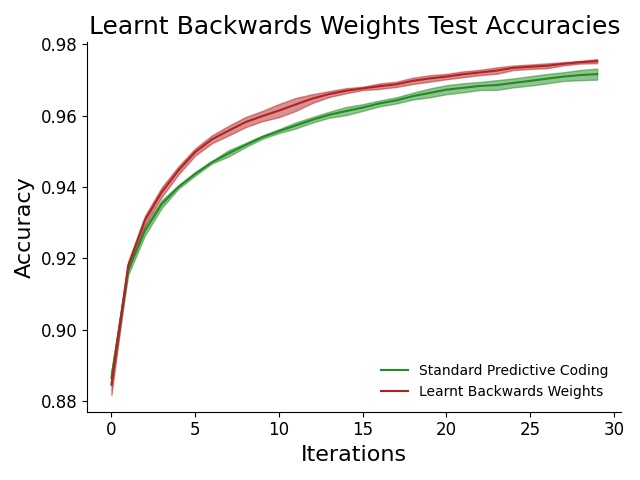} 
 \caption{\small MNIST dataset; tanh activation} 
 \vspace{4ex}
 \end{subfigure}
 \begin{subfigure}[b]{0.5\linewidth}
 \centering
 \includegraphics[width=0.75\linewidth]{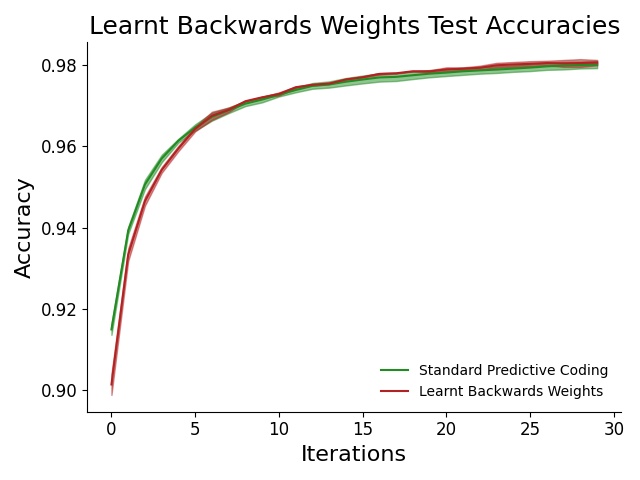} 
 \caption{\small MNIST dataset; relu} 
 \vspace{4ex}
 \end{subfigure} 
 \begin{subfigure}[b]{0.5\linewidth}
 \centering
 \includegraphics[width=0.75\linewidth]{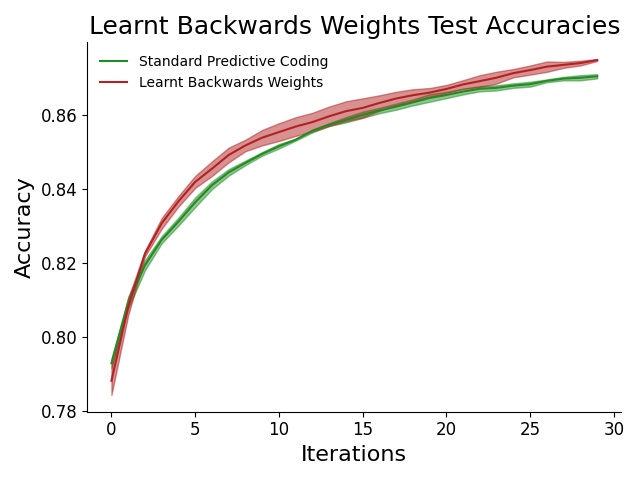} 
 \caption{\small Fashion dataset; tanh} 
 \end{subfigure}
 \begin{subfigure}[b]{0.5\linewidth}
 \centering
 \includegraphics[width=0.75\linewidth]{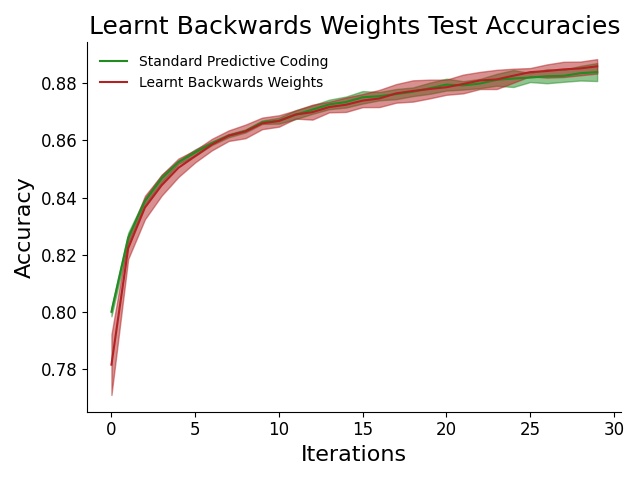} 
 \caption{\small Fashion dataset; relu} 
 \end{subfigure} 
 \caption{Test accuracy of predictive coding networks with both learnt backwards weights, and the ideal weight transposes with both relu and tanh activation functions on the MNIST and FashionMNIST datasets. Both networks obtain almost identical learning curves, thus suggesting that learnt backwards weights allow for a solution to the weight-transport problem.}
 
\label{learnt_backwards_weights_figure}
\end{figure}

This procedure allows us to begin with a randomly initialized set of backwards weights, and then applying the learning rule to these weights allows us to very quickly recover performance equal to the identical backwards weights. As shown in Figure \ref{learnt_backwards_weights_figure}, performance both with and without the learnt backwards weights is almost identical for both the relu and tanh nonlinearities and the MNIST and FashionMNIST datasets, thus suggesting that this approach of simply learning an independent set of backwards weights is a highly effective and robust method for tackling the weight transport problem.

\subsubsection{Backwards nonlinear derivatives}

The second remaining biological implausibility is that of the backwards nonlinear derivatives. Note that in Equations \ref{relaxed_PC_rules}, an $f'$ term regularly appears denoting the pointwise derivative of the nonlinear activation function. Since these are pointwise derivatives, when the mathematics is translated to neural circuitry, these derivatives need to be computed at each individual neuron. It is not clear whether neurons are capable of easily computing with the derivative of their own activation function. We apply a straightforward remedy to address this. We simply experiment with removing the pointwise derivatives from the update rules. For instance Equation \ref{PC_hierarchical_mu} would become just $\frac{d\mu^l}{dt} = -\epsilon^l + {\theta^l}^T \epsilon^{l-1}$. Perhaps surprisingly we found that this modification, although not mathematically correct, did little to impair performance of the model at classification tasks, except perhaps for the hyperbolic tangent nonlinearity in the FashionMNIST case.

\begin{figure}[ht] 
 \begin{subfigure}[b]{0.5\linewidth}
 \centering
 \includegraphics[width=0.75\linewidth]{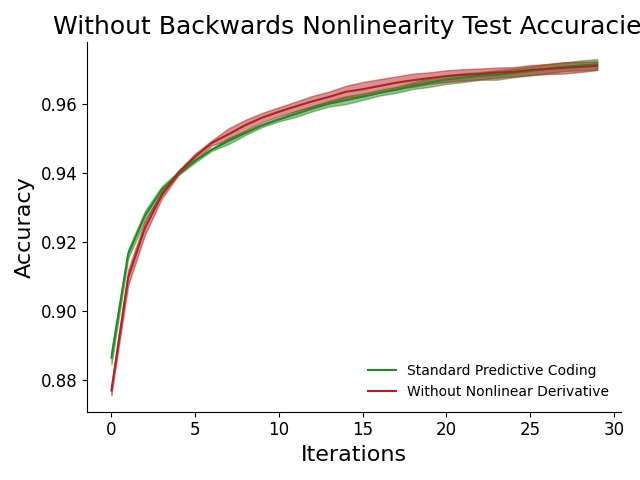} 
 \caption{\small MNIST dataset; tanh activation} 
 \vspace{4ex}
 \end{subfigure}
 \begin{subfigure}[b]{0.5\linewidth}
 \centering
 \includegraphics[width=0.75\linewidth]{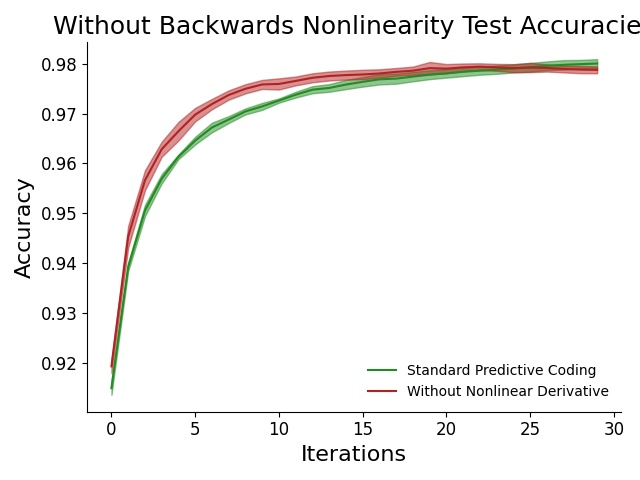} 
 \caption{\small MNIST dataset; relu activation} 
 \vspace{4ex}
 \end{subfigure} 
 \begin{subfigure}[b]{0.5\linewidth}
 \centering
 \includegraphics[width=0.75\linewidth]{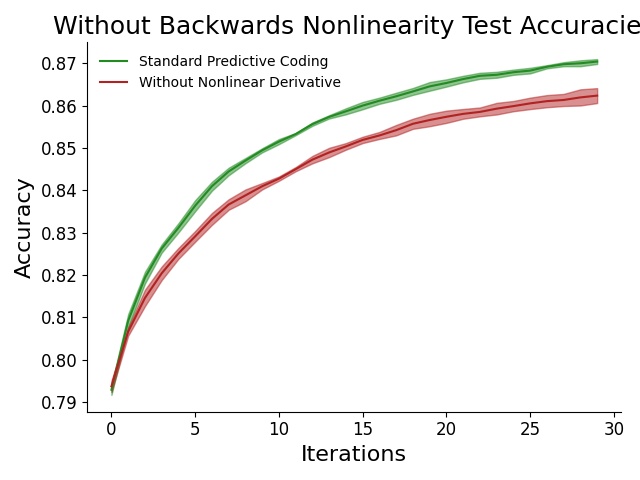} 
 \caption{\small Fashion dataset; tanh activation} 
 \end{subfigure}
 \begin{subfigure}[b]{0.5\linewidth}
 \centering
 \includegraphics[width=0.75\linewidth]{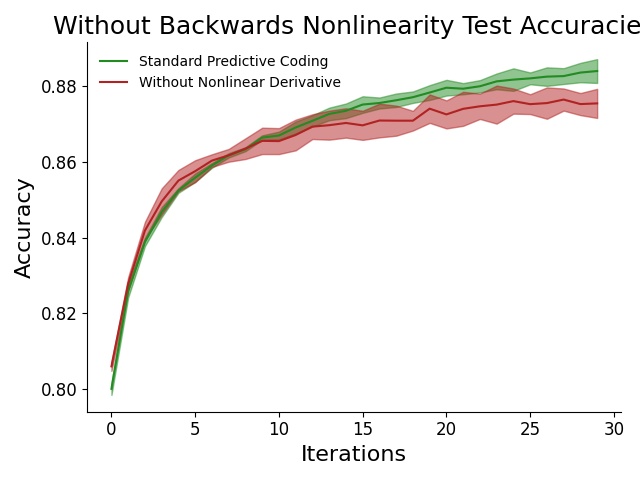} 
 \caption{\small Fashion dataset; relu activation} 
 \end{subfigure} 
 \caption{Test accuracy of predictive coding networks with and without the nonlinear derivative term, using relu and tanh activation functions on the MNIST and FashionMNIST datasets. We find that on the MNIST dataset performance is similar, while on the FashionMNIST dataset and the tanh activation function, the lack of the nonlinear derivative appears to slightly hurt performance.}
\end{figure} 
 
In effect, by removing the pointwise nonlinear derivatives, we have made the gradient updates linear in the parameters. Since the real updates are nonlinear, our update rules are simply the projection of the nonlinear update rules onto a linear subspace. However, using a similar argument to that in feedback alignment, we hypothesize that it is likely that the linear projection of the nonlinear updates are quite close in angle to the nonlinear updates, so the direction of the linear gradient, averaged over many batches and update steps, is sufficiently close to the true gradient as to allow for learning in this model. An alternative option is to note that the derivatives of the nonlinearity depend closely upon the value of the function. In the case of the relu nonlinearity, the derivative is only different from 1 when the activity is 0 (the neuron does not fire). When there is no firing, there can still be updates to the dynamics -- which depend on the prediction errors and not on the actual firing rate -- and so there is still error when dropping the derivative term. However, if most activations are greater than 0, the error should be minimal, which is what we appear to observe. Similarly, in the hyperbolic tangent nonlinearity, the region of activation between $-1$ and $1$ is broadly linear, and thus we should expect the dropping of the nonlinear derivative term in this region to have relatively little effect. The robustness and relatively little impact on training therefore suggest that the activations of the predictive coding network largely remain within this stable regime over the course of training -- an intriguing and important finding given that we made no efforts (such as regularization) to specifically encourage this outcome. If brains operated in a similar regime, it may mean that explicit computation of the activity derivatives is unnecessary, which would make credit assignment substantially easier (if approximate).

\subsubsection{Error connections}
 
\begin{figure}
 \begin{center}
 \includegraphics[width=16cm, height=8cm]{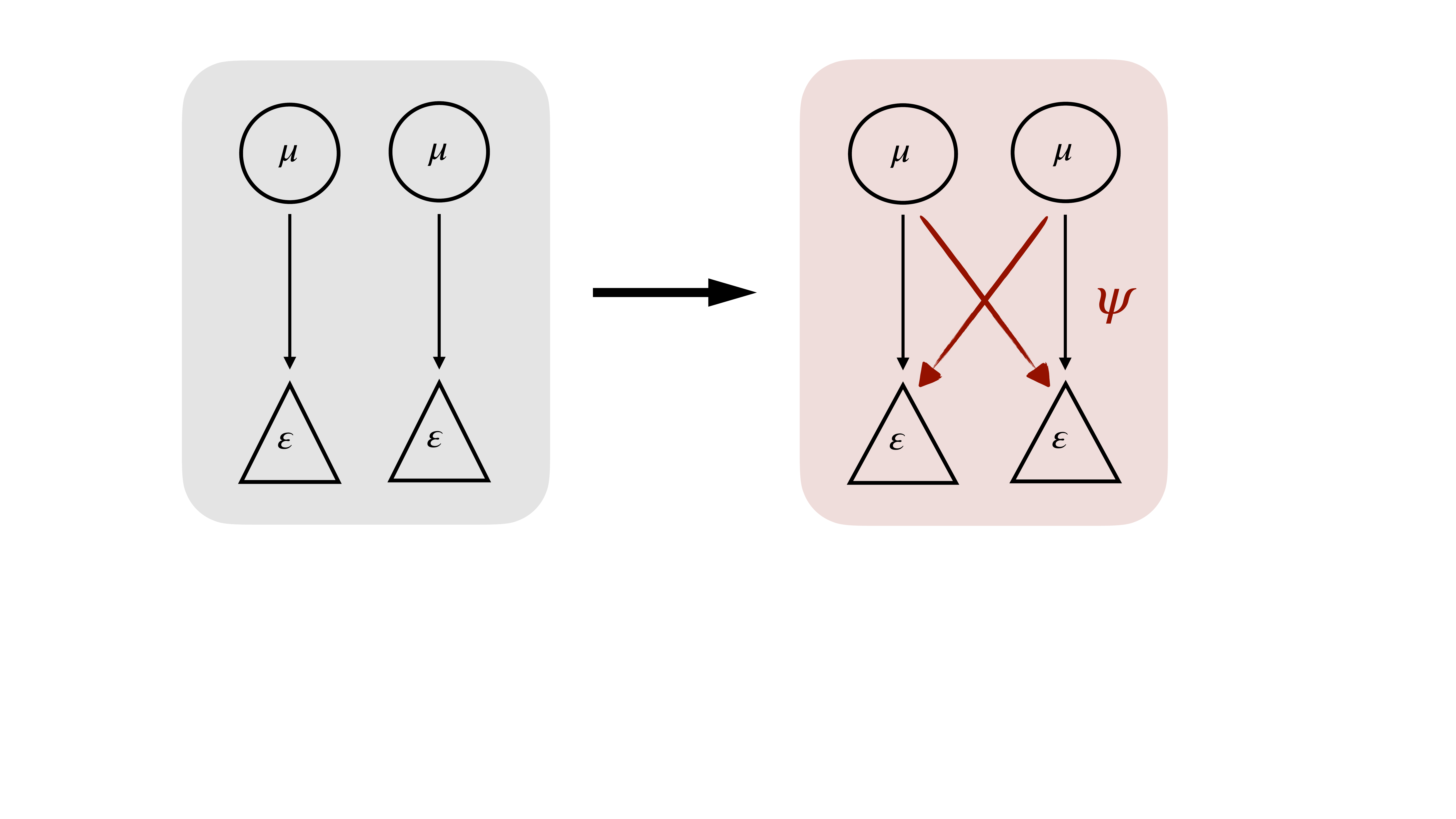}
 \end{center}
 \vspace{-3cm}
 \caption{The error-connectivity problem and our solution. On the left, the biologically implausible one-to-one connectivity between value and error nodes required by the standard predictive coding theory. On the right, our solution to replace these one to one connections by a fully connected connectivity matrix $\psi$. By learning $\psi$ with a Hebbian learning rule we are able to achieve comparable performance to the one-to-one connections with a fully dispersed connectivity matrix.}
 
 \label{Error_Connection_Diagram}
 \end{figure}

The third and final biological implausibility that we address in this section is that of the one-to-one connections between value and the error units at a given layer. This can be seen directly for the prediction errors in Equation \ref{relaxed_PC_rules}, but broken down into individual components (or neurons).
\begin{align*}
 \epsilon_i^l = \mu_i^l - f(\sum_j \theta^{l+1}_{i,j} \mu^{l+1}_{j}) \numberthis
\end{align*}

We see that the activity of the error unit vector (i.e. each error neuron $\epsilon_i$) is driven by a one-to-one connection from its matching value neuron $\mu_i$. By contrast, the top-down predictions have a diffuse connectivity pattern, where every value neuron $\mu_j$ in the layer above affects each error neuron $\epsilon_i$ through the synaptic weight $\theta_{i,j}$. A one-to-one connectivity structure is a highly precise and sensitive pattern and it is difficult to see how it could first develop and then be maintained in the brain throughout the course of an organisms life. Additionally, while precise connectivity can exist in theory, there is little evidence neurophysiologically \citep{bastos2012canonical,walsh2020evaluating} for the kind of regular and repeatable one-to-one connectivity patterns that predictive coding would require in the brain. Moreover, if predictive coding were implemented throughout the cortex, this one-to-one connectivity should be highly visible to neuroscientists. To relax this one-to-one connectivity constraint, we postulate a diffuse connectivity pattern between them, mediated by a set of connection weights $\psi$. The new equation for the prediction errors becomes:
\begin{align*}
 &\epsilon_i^l = \sum_k \psi_{i,k}^l \mu_k^l - f(\sum_j \theta^{l+1}_{i,j} \mu^{l+1}_{j}) \\
 &\implies \epsilon^l = \psi^l \mu^l - f(\theta^{l+1} \mu^{l+1}) \numberthis
\end{align*}

\begin{figure}[ht] 
 \begin{subfigure}[b]{0.5\linewidth}
 \centering
 \includegraphics[width=0.75\linewidth]{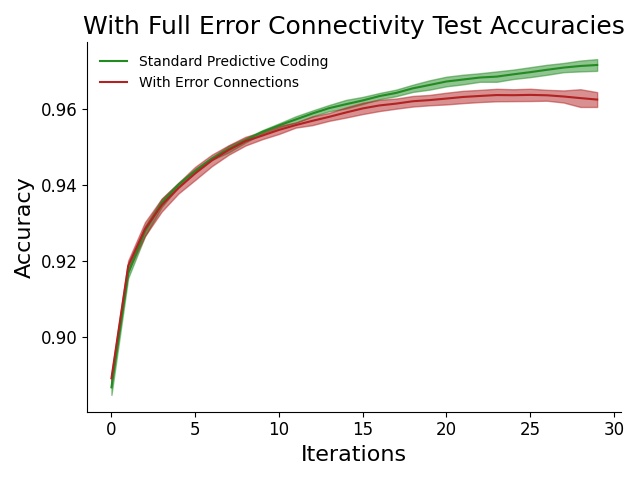} 
 \caption{\small MNIST dataset; tanh activation} 
 \vspace{4ex}
 \end{subfigure}
 \begin{subfigure}[b]{0.5\linewidth}
 \centering
 \includegraphics[width=0.75\linewidth]{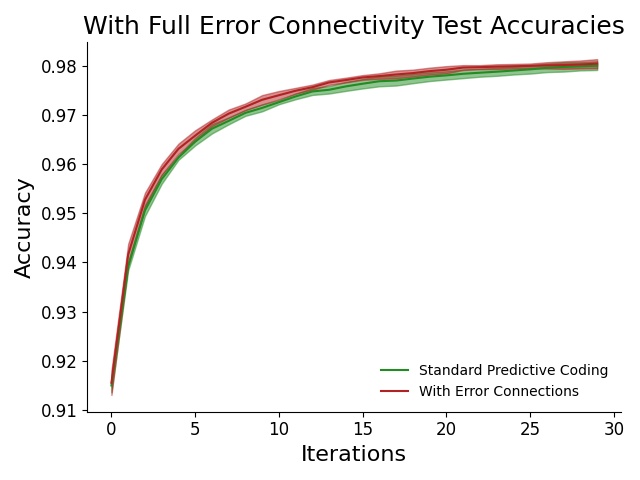} 
 \caption{\small MNIST dataset; relu activation} 
 \vspace{4ex}
 \end{subfigure} 
 \begin{subfigure}[b]{0.5\linewidth}
 \centering
 \includegraphics[width=0.75\linewidth]{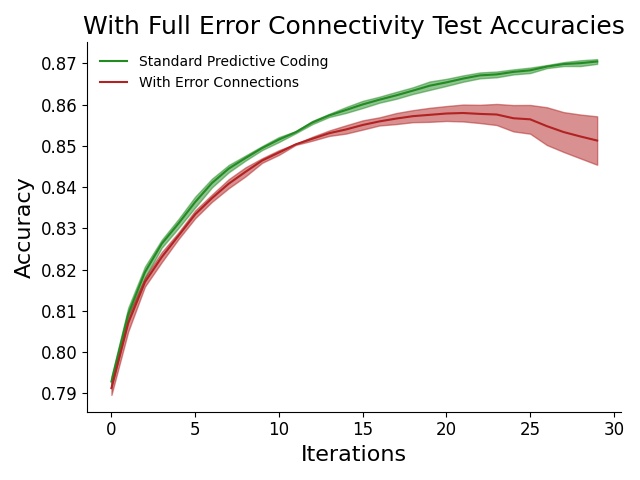} 
 \caption{\small Fashion dataset; tanh activation} 
 \end{subfigure}
 \begin{subfigure}[b]{0.5\linewidth}
 \centering
 \includegraphics[width=0.75\linewidth]{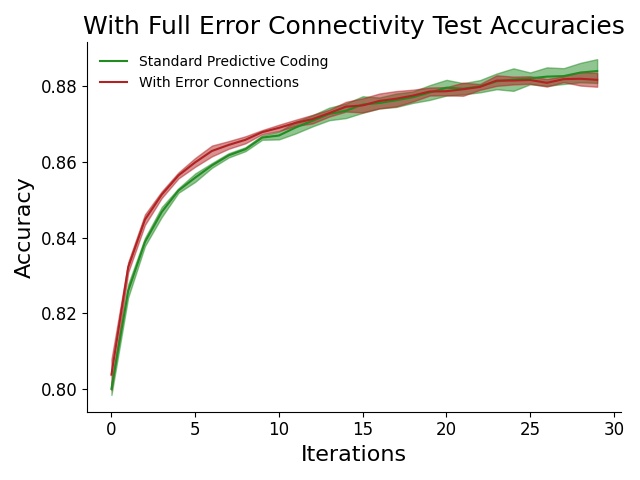} 
 \caption{\small Fashion dataset; relu activation} 
 \end{subfigure} 
 \caption{Test accuracy of predictive coding networks with and without learnable error connections for both relu and tanh activation functions on the MNIST and FashionMNIST datasets. We see that, interestingly, using learnt error weights decreased performance only with the tanh but not the relu nonlinearity, and then only slightly in the FashionMNIST case.}
 
\label{learning_error_weights}
\end{figure}

While using randomly initialized weights $\psi$ completely destroys learning performance, it is possible to learn these weights in an online unsupervised fashion using another Hebbian learning rule. The learning rule for the error weights $\psi$ can be derived as a gradient descent on the variational free energy function, whereby now the prediction errors include the error weights,
\begin{align*}
 \frac{d\psi^l}{dt} = -\frac{\partial \mathcal{F}}{\partial \psi} &= -\epsilon_l \frac{\partial \epsilon_l}{\partial \psi} \\
 &= -\epsilon_l \mu_l^T \numberthis
\end{align*}
This rule is completely linear and Hebbian since it is simply a multiplication of the activations at the two endpoints of the connection. We show in Figure \ref{learning_error_weights} that using this rule allows equivalent learning performance to the one-to-one case as the required weight values are rapidly learnt.

We see that, overall, training performance can be maintained even with learnt error connections, a perhaps surprising result given how key the prediction errors are in driving learning. Interestingly, we see a strong effect of activation function on performance. Performance is indistinguishable from baseline with a relu activation function but asymptotes at a slightly lower value than the baseline with tanh. Investigating the reason for the better performance of the relu nonlinearity would be an interesting task for future work.

\subsubsection{Combining Relaxations}

\begin{figure}[H]
 \centering
 \subfloat[\centering Standard PC]{{\includegraphics[width=0.40\textwidth]{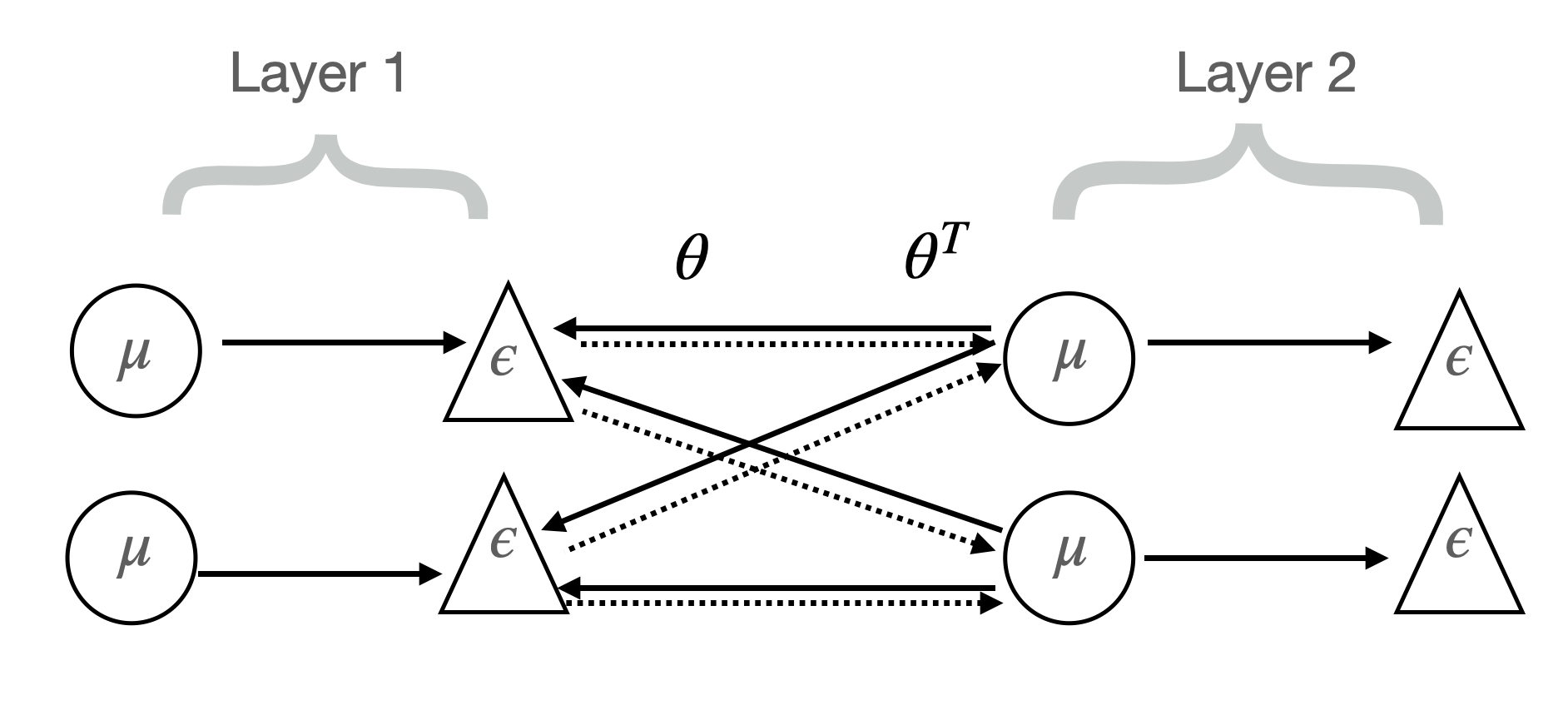} }}%
 \qquad
 \subfloat[\centering Relaxed PC]{{\includegraphics[width=0.40\textwidth]{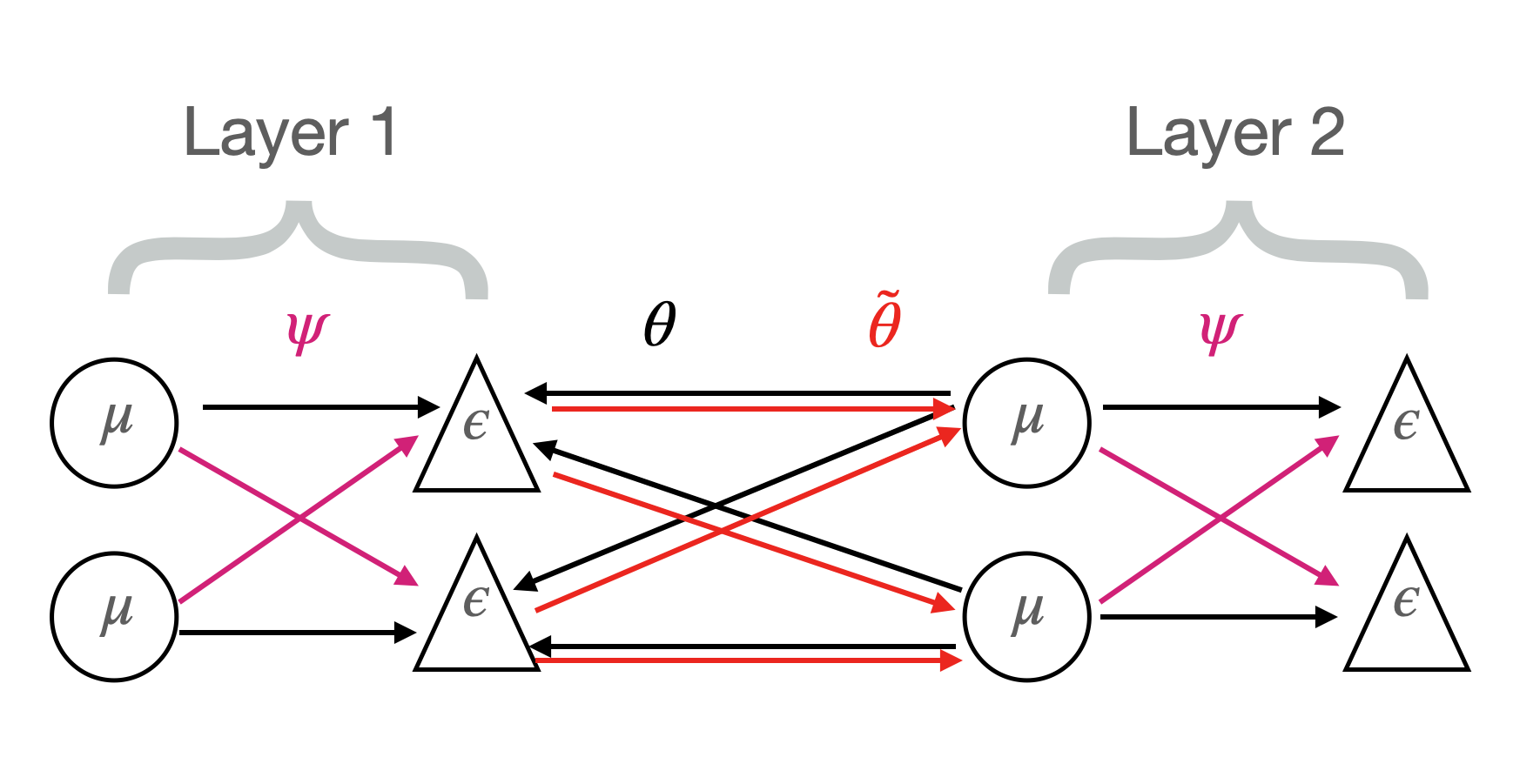} }}%
 \caption{Schematic representations of the architecture across two layers of a.) the standard predictive coding architecture and b.) The fully relaxed architecture. Importantly, this architecture has full connectivity between all nodes and also non-symmetric forward and backwards connectivity in all cases. In effect, this architecture only maintains a bipartite graph between error and value neurons, but no other clear structure}%
 
\label{fully_relaxed_figure}
\end{figure}

It is also possible to \emph{combine} all of the above relaxations together in parallel, to create a network architecture which is avoids many of the major biological plausibility pitfalls of predictive coding. A schematic representation of this combined architecture compared to the standard predictive coding architecture is shown below (Figure \ref{fully_relaxed_figure}).

\begin{figure}[ht]
 \begin{subfigure}[b]{0.5\linewidth}
 \centering
 \includegraphics[width=0.75\linewidth]{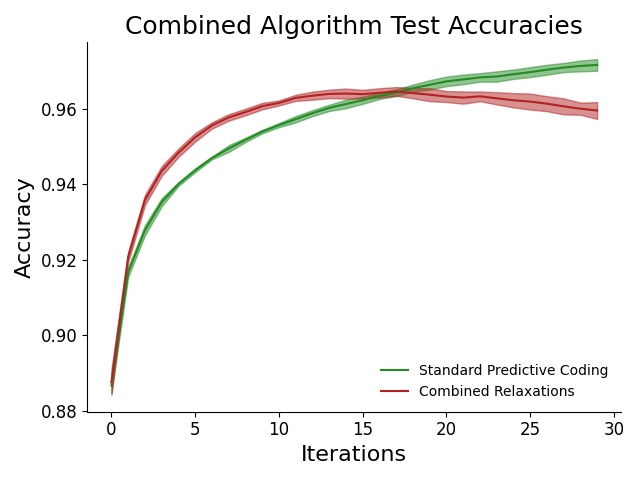} 
 \caption{\small MNIST dataset; tanh activation} 
 \vspace{4ex}
 \end{subfigure}
 \begin{subfigure}[b]{0.5\linewidth}
 \centering
 \includegraphics[width=0.75\linewidth]{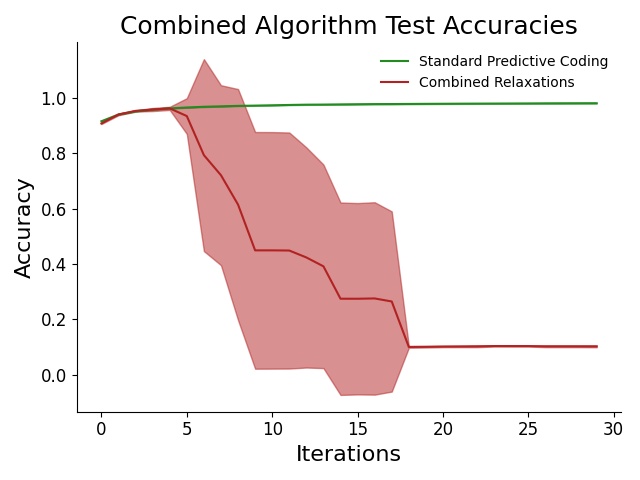} 
 \caption{\small MNIST dataset; relu activation} 
 \vspace{4ex}
 \end{subfigure} 
 \begin{subfigure}[b]{0.5\linewidth}
 \centering
 \includegraphics[width=0.75\linewidth]{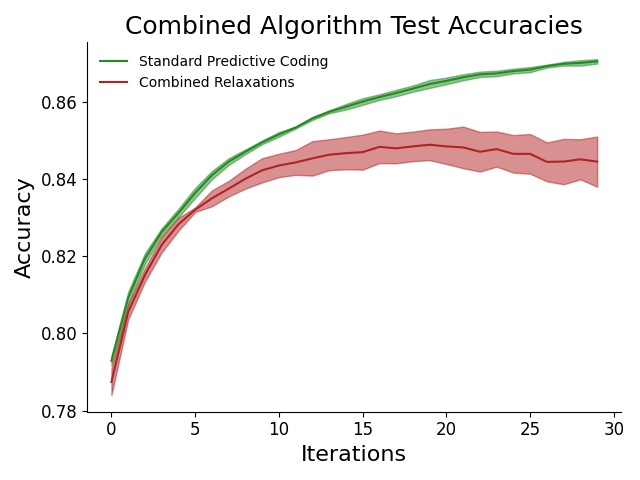} 
 \caption{\small Fashion dataset; tanh activation} 
 \end{subfigure}
 \begin{subfigure}[b]{0.5\linewidth}
 \centering
 \includegraphics[width=0.75\linewidth]{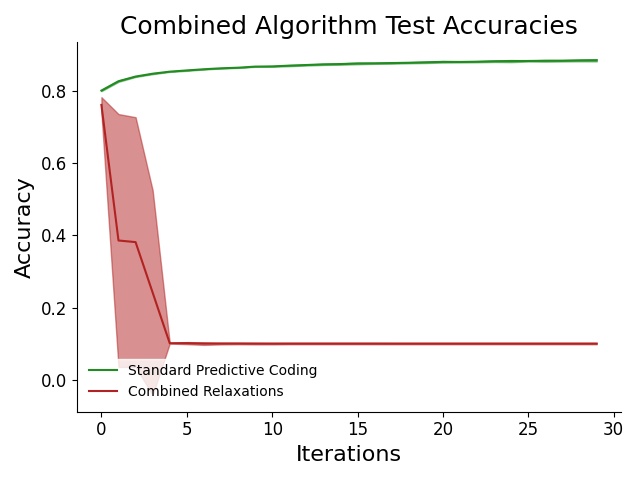} 
 \caption{\small Fashion dataset; relu activation} 
 \end{subfigure} 
 \caption{Test accuracy standard and fully relaxed predictive coding networks (the combined algorithm), for both relu and tanh activation functions on the MNIST and FashionMNIST datasets. We see that, interestingly,performance is degraded in all cases and that the relu networks are especially affected -- with catastrophic declines in performance to become almost random. The reasons for this are currently unknown and will be investigated in future work.}
 
\label{fully_relaxed_results_figure}
\end{figure} 

The relaxed architecture no longer has any one-to-one connectivity patterns, which have been replaced with full connectivity matrices parametrised by the error connection weights $\psi$. Moreover, the forwards and backwards weights are separated into two separate and independent weight matrices $\theta$ and $\tilde{\theta}$ while the standard predictive coding model uses the true weight transpose $\theta^T$, which requires copying the weights. In essence, the fully relaxed architecture simply consists of two bipartite populations of neurons, which only synapse onto the other population. Beyond this there is no special connectivity structure required. Nevertheless, we show that even this very simple architecture, with only Hebbian learning rules, can still be trained to perform well at supervised classification.

We tested the classification ability of the fully relaxed architecture with both hyperbolic tangent and rectified linear activation functions, and on the MNIST and FashionMNIST datasets, and the results are shown in Figure \ref{fully_relaxed_results_figure}. Overall we found that another strong effect of activation function where this time training was unstable and diverged when using rectified linear units but not when using tanh neurons. We hypothesize that this could be due to the rectified linear units not having a saturation point unlike tanh and thus being more prone to exploding gradients. We found, on the other hand, that while performance of the fully relaxed network asymptotically tended to be slightly worse than the standard, it was still very high on both the MNIST and fashion MNIST datasets, thus showing that even highly relaxed and extremely local networks with an extremely generic, essentially fully connected, connectivity pattern can be trained to very high accuracies using this predictive coding algorithm which only requires Hebbian updates. 

\subsection{Discussion}

We have shown that it is possible to surmount three key biological implausibilities of the predictive coding learning rules, and thus strengthen the case that predictive coding may be implemented in cortical circuitry. In the weight transport and error connections case, the solution has been to propose a separate set of weights which are themselves learnable by a Hebbian rule. In the backwards nonlinearities case, it suffices simply to ignore the biologically implausible terms in the rule. Moreover, we have shown that performance, at least under the hyperbolic tangent nonlinearity, is still stable and roughly comparable with the baseline when all the relaxations are combined together, thus resulting in an extremely local,straightforward, and biologically plausible architecture. Overall, we believe these results show that predictive coding offers a surprisingly robust model of learning and inference in the brain and that it can survive often severe perturbations to its basic equations. Through this work we have substantially diminished the constraints a neurophysiologically realistic process theory of predictive coding must satisfy. In so doing, it is possible that predictive coding may now fit a greater part of neurophysiological data, while opening the way to potentially constructing novel microcircuit designs which implement relaxed forms of predictive coding.
There may also be gains from applying these heuristics to other biologically plausible approximations to backprop. We show in the final chapter on biologically plausible credit assignment in the brain, that many of these techniques also work for other algorithms, thus hinting at potentially general properties of perturbational robustness of these neurally inspired learning algorithms which may be of significant theoretical interest. 

An additional theoretical note is the power of the assumption of variational optimality. Like a Lagrangian in physics, the variational free energy $\mathcal{F}$ enables potentially complex `laws of motion' to be derived through a simple mathematical apparatus and which can be extended to lead to otherwise-difficult insights. For instance, the learning rules for the error-weights $\psi$ can be derived straightforwardly in the variational framework as simple gradient descents on $\mathcal{F}$ given an augmented generative model containing the $\psi$ term. Regardless of one's theoretical or ontological commitments to variational inference in the brain, the mathematical reformulation of neural activity as encoding solutions to a variational inference problem allows considerable modelling flexibility and mathematical insight through which results can be easily derived which would be much harder to achieve through other means. 

Having removed the symmetric backwards weights and the one-to-one error connectivity scheme, we are left with an essentially bipartite graph. There are connections between the value and error units of the same level, and the error units of one level and the value units of the level above, but crucially there are no direct connections between the value or error units of one layer and those of the layer above. Stepping out of the predictive coding framework, we have effectively shown that a simple bipartite connectivity structure and Hebbian learning rules suffices to learn complex input-output mappings, and may mathematically approximate backpropagation. This is a surprising result given that previously Hebbian learning has not generally been thought to be sufficient to learn complex representations in the brain \citep{baldi2016theory}. This shows that perhaps it is possible for the brain to go further with clever connectivity patterns and Hebbian learning than previously thought.

It is also important to note here that while we have resolved several biological implausibilities of predictive coding, there are still several other difficulties that must be faced before a direct implementation of predictive coding is possible given what we currently know about neural circuitry. A key challenge is the simple problem of negative prediction errors and activations -- in the mathematical formalism, prediction errors and values are real numbers which can be both positive and negative, and to any degree of accuracy. In the brain, however, we assume that these numbers are represented by average firing rates, which cannot go negative, and additionally have a degree of accuracy constrained by the intrinsic noise levels of the brain and the integration windows over which post-synaptic neurons can listen. While it seems likely that a lack of numerical accuracy is not that significant for neural networks in general, given recent results in machine learning demonstrating that only 16 bit floats are necessary at most \citep{gupta2015deep}, the issue of negative numbers is substantial since negative prediction errors are absolutely necessary for the functioning of the algorithm. One possibility is that the brain could maintain a high default firing rate, and treat deviations below this default as negative. However, the maintenance of a sufficiently high default firing rate would be energy inefficient, and there is much evidence that neurons primarily maintain low tonic firing rates \citep{walsh2020evaluating}. Another option could be to utilize separate populations of `positive' and `negative' neurons, perhaps excitatory and inhibitory neurons, however this would require a precise connectivity scheme to integrate these two contributions together, which has largely not yet been worked out in the context of predictive coding. 

An additional limitation of our work is that we have only tested the performance of the relaxations on relatively small networks and using the relatively simple MNIST and Fashion-MNIST datasets which are simple enough that even fairly non-scalable methods can work on them. A prime example of this is the study by \citet{bartunov2018assessing} who showed that many of the methods in the literature for alternative biologically plausible methods for credit assignment, although performing well on MNIST, generally performed poorly on more challenging datasets such as CIFAR10, CIFAR100, and ImageNet. A key task must be to investigate the scaling properties of these relaxations to more challenging tasks and datasets, as well as different network architectures such as convolutional neural networks. Some preliminary, but supportive results come from \citet{millidge2020investigating} (discussed in chapter 6) where we show that the learnable backwards weights and dropping the nonlinear derivatives do in fact scale to larger scale CNN networks.

\section{Conclusion}

In this chapter, we have studied the application of the free energy principle to perception -- specifically by investigating and proposing substantial extensions to the predictive coding process theory \citep{friston2003learning,friston2005theory,friston2008hierarchical} as well as testing the performance of large-scale implementations of the theory. Overall, we believe that our work in this chapter has make substantial improvements to the theory and practice of predictive coding.

Specifically, for the first time, we have implemented and tested predictive coding models on a larger scale than previously -- and have compared them against machine learning approaches on standard machine learning datasets such as MNIST. We have demonstrated that predictive coding networks are able to successfully reconstruct digits successfully, interpolate between them, and separate out different digit representations in the learnt latent space despite being trained with an entirely unsupervised objective. Moreover, we have demonstrated that predictive coding can achieve this in hierarchical and dynamical setups with randomized initial weights which are then learned, in contrast to prior work \citep{friston2008DEM,friston2008hierarchical,friston2005theory} which primarily focus on the inference capabilities of predictive coding and provide a-priori the correct generative model. We also implemented and demonstrated that dynamical, and both hierarchical and dynamical predictive coding models can function well on simple toy tasks and can very quickly learn various challenging wave-forms.

Secondly, we have investigated the use of predictive coding algorithms for filtering tasks (as opposed to static inference). In filtering, the aim is to infer an entire trajectory of states given a trajectory of observations, instead of simply inferring a single hidden state given a single observation. We make precise, for the first time, the precise relationship between Kalman filtering -- a ubiquitous algorithm in classical control and filtering theory -- and predictive coding which is that the predictive coding dynamics can be derived as a gradient descent on the Gaussian maximum-a-posteriori objective, which the Kalman filter equations can be derived as an analytical solution to. We then demonstrate the successful filtering capabilities of our predictive coding filtering algorithm, and demonstrate how the broader variational approach allows us to successfully also learn the parameters of the generative model online for filtering tasks -- thus performing double deconvolution where we infer both states and parameters simultaneously. 

Finally, we have investigated and improve the biological plausibility of the predictive coding process theory which, after all, is often explicitly proposed as a neuroscientific theory of cortical function \citep{friston2003learning,bastos2012canonical}. Here we focus on and present solutions to three outstanding issues of biological implausibility with the standard predictive coding dynamics. First, we address the weight transport problem -- the need to transmit activities backwards -- by proposing a set of independent backwards weights which are initialized randomly, and then can also be learnt with an independent and biologically plausible Hebbian update rule. Secondly, we address the problem of nonlinear derivatives by showing that in many cases these derivatives can be dropped from the update rules with relatively small performance penalties, and thirdly, we address the issue of needing precise one-to-one error to value neuron connectivity by proposing instead fully distributed connectivity between error and value neurons, but with an additional learnable weight matrix which can be additionally optimized with another Hebbian rule. We show that these relaxations can substantially improve the biological plausibility of the predictive coding algorithm while only causing relatively small degradations of performance on machine learning benchmark classification tasks.

Overall, therefore, we believe that in this chapter we have made significant contributions to the theory and practice of predictive coding. On the theoretical level, we have demonstrated its relationship to Kalman filtering, and we have addressed several outstanding challenges of biological implausibility that the standard theory faces. On an implementational and practical level, we have empirically investigated for the first time the performance of predictive coding networks within the machine learning paradigm on machine learning benchmark tasks, and especially in cases where the true generative model is not provided to the network a-priori. Additionally, through our work, we have substantially scaled up predictive coding approaches to handle significantly larger and more challenging tasks than previously, and have made these implementations available to the community through a number of open-source software projects \footnote{See: https://github.com/BerenMillidge/PredictiveCodingBackprop, https://github.com/BerenMillidge/RelaxedPredictiveCoding, https://github.com/BerenMillidge/NeuralKalmanFiltering}

In the next two chapters, we advance from the problem of perception, to consider the problem of action selection through the lens of the free energy principle, and its concomitant process theory \emph{active inference}. In some ways this problem is more challenging than pure perception, since it requires the modelling of entire trajectories of observations, states, and actions, in order to make the best long term decisions which will, over time, outperform locally greedy options. In the next Chapter (Chapter 4), we focus primarily on scaling up existing active inference models using deep neural networks to match the performance of state of the art reinforcement learning algorithms. In the chapter after that (Chapter 5), we aim to provide a deep mathematical investigation and, ultimately, insight into the nature of objective functionals which combine both reward-seeking and information-seeking imperatives.

%% file: chap4.tex
\chapter{Scaling Active Inference}

\section{Introduction}

In this chapter, we consider the application of the free energy principle to action selection, or control, problems. While in the previous chapter on perception, we focused on the process theory of predictive coding, here we focus on the process theory of active inference, and are especially inspired by the discrete state-space active inference theory introduced in Chapter 2. Here, we aim to solve a key limitation of those methods -- their scalability. We propose to do so by parametrizing the key densities of the generative model and recognition distribution by deep neural networks, and then utilizing the tools of deep reinforcement learning to allow active inference agents to scale to levels comparably achieved by contemporary machine learning.

This chapter comprises multiple sections. At the beginning, we give a detailed introduction to reinforcement learning \citep{sutton2018reinforcement}, and especially deep reinforcement learning, as well as the control of inference framework \citep{rawlik2013probabilistic,levine2018reinforcement} from reinforcement learning which also frames the control problem as one of inference. Then we present two studies where we demonstrate that active inference approaches can scale up to be comparable with contemporary deep reinforcement methods. We achieve this scaling by first parametrizing the key distributions in the active inference model by deep neural networks trained through gradient descent, and secondly by approximating the (exponential time) computation of the path integral of the expected free energy either with an amortized neural network prediction, or else through monte-carlo trajectory sampling using a continuous action planning algorithm. In doing so, we create algorithms that are comparable in scalability and performance to current methods in deep reinforcement learning. Additionally, we demonstrate that oftentimes these algorithms can outperform their reinforcement learning counterparts due to the unique properties and insights active inference brings to the table.

In the second section of this chapter, we focus a little more abstractly in trying to understand the difference between model-free and model-based reinforcement learning approaches in terms of inference, and determine that this difference is primarily due to the difference between what we call \emph{iterative} variational inference -- where the parameters of the variational distribution are directly optimized -- and \emph{amortized} inference -- where instead the parameters of a function which outputs the parameters of the variational distribution are optimized. Given this distinction, we first use it to present a taxonomy of a wide range of current reinforcement learning algorithms using a simple two-dimensional quadrant, and secondly, we derive novel algorithms which emerge by combining both iterative and amortized inference together -- an approach we call \emph{hybrid} inference -- which results in powerful algorithms which combine the benefits of each approach, while ameliorating their respective weaknesses.

\subsection{Reinforcement Learning}

The reinforcement learning, or control, problem is one of the most fundamental problems in artificial intelligence and in engineering adaptive systems \citep{sutton1990integrated,sutton1998introduction,kaelbling1996reinforcement,dayan1997using}, and concerns the computation of optimal action \citep{wolpert1997computational,todorov2008general}. The problem is simple. We assume that there is some kind of agent in some kind of environment, and that the agent can take actions which affect the environment \citep{sutton1998introduction}.  Suppose that the agent has some kind of notion of goals or desires that it wants to achieve -- whether these are encoded as a desire distribution, as an objective function, or as rewards given by the environment. The control problem is to compute the optimal action schedule to fulfill the agent's goals. As might be expected, this question has huge applications and implications for an extremely wide range of fields, from machine learning and artificial intelligence (understanding how to make artificial agents act to achieve their goals) \citep{sutton1998introduction,mnih2013playing,silver2016mastering,schrittwieser2019mastering,schulman2015trust} to cognitive science and economics (understanding how humans implement action strategies to achieve their goals) \citep{todorov2008general,wolpert1997computational,dayan2008decision,daw2006cortical} to biology (how do all sorts of biological systems act adaptively) \citep{dayan2009goal,mehlhorn2015unpacking,krebs1978test,pyke1984optimal} to control theory (how to design and program systems which can adaptively regulate and control their environments) \citep{kirk2004optimal,kwakernaak1972linear,sethi2000optimal,kalman1960contributions,johnson2005pid,kappen2005path}.

While this provides an intuitive specification of the control problem, to make real progress we must make it precise mathematically. First, we assume that the environment has states which we denote $x$ and that the agent can emit actions $a$. Secondly, we assume that there exists some reward function which emits rewards $r$ dependent on environmental states $r = f(x)$. The only thing the agent has control over are its actions $a$ which can affect the environment to give it more rewards. We assume that the agent optimizes over \emph{trajectories} of states and actions going into the future, which we denote as $\tilde{x}$, $\tilde{r}$, and $\tilde{a}$. For the moment, to retain full generality, we remain indifferent to whether the agent considers time continuous or discrete, so that $\tilde{x} = \sum_t x(t) = \int dt x(t)$. Finally, we assume that the environmental dynamics and the rewards granted can both be stochastic and can thus be mathematically formalized in terms of probability distributions $p(\tilde{x} | \tilde{a})$ and $p(\tilde{r} | \tilde{x})$. A key advantage of this probabilistic formalism is that it allows us to represent (and remain agnostic between) intrinsic stochasticity in the environment, and the agent's uncertainty about the environment. If the environment or rewards are in fact deterministic and known, we can simply set the distributions to be dirac deltas to recover a deterministic framework. Under this formalism, the objective of the control problem is to maximize \footnote{Here, following the convention in reinforcement learning and economics, we are optimistic and we talk about reward (or equivalently utility) maximization. Control theory, on the other hand, takes a more depressive interpretation and works in terms of minimizing costs. Mathematically, these two formulations are completely equivalent.},
\begin{align*}
\label{control_objective}
\mathcal{L}_{control} &= \underset{a}{argmax} \, \,  \int d\tilde{x} \, p(\tilde{r} | \tilde{x}, \tilde{a})p(\tilde{x} | \tilde{a})p(\tilde{a}) \\
&= \underset{a}{argmax} \, \, \mathbb{E}_{p(\tilde{x} | \tilde{a})p(\tilde{a})}[ \ln p(\tilde{r} | \tilde{x}, \tilde{a})] \numberthis
\end{align*}

Where we can safely take the log of the reward function since log is a monotonic function and does not impact the optimum of the optimization process, but tends to make things nicer numerically. Essentially, what this states is that the control objective is simply to maximize the probability or amount of reward expected under the trajectory of environment states given the agent's trajectory of actions. From this, we can see that to first get a handle on the control problem, we need to understand firstly the environmental dynamics $p(\tilde{x} | \tilde{a})$ and the reward or utility function $p(\tilde{r} | \tilde{x}, \tilde{a})$. We typically represent these dynamics as stochastic differential (or difference) equations depending on whether time is discrete or continuous, as follows,
\begin{align*}
\frac{dx}{dt} = f(x_{1:t},a_{1:t}, \omega) \numberthis
\end{align*}
for continuous time, where $\omega$ is some kind of noise or,
\begin{align*}
x_{t+1} = f(x_{1:T}, a_{1:T}, \omega) \numberthis
\end{align*}
for discrete time. One simplifying assumptions we often make is that the dynamics are Markovian, meaning that the state at time $t+1$ can be computed solely in terms of the state at time $t$ and the action at time $t$, thus that the dynamics become,
\begin{align*}
x_{t+1} = f(x_t, a_t, \omega) \numberthis
\end{align*}
This approach simplifies the analysis considerably. An additional assumption, which is often made, is that the rewards depend only on the state and actions at the current time $p(\tilde{r} | \tilde{x}, \tilde{a}) = \Pi_t p(r(t) | s(t), a(t))$. Under these assumptions the environment of the control problem can be considered to be a Markov Decision Process (MDP). It is also sometimes the case that we assume we do not know the true state of the environment, but are only given access to partial observations $\tilde{o}$ which may not be Markov, even though the hidden states are Markov. The observations are related to the states through a likelihood mapping $p(\tilde{o} | \tilde{x})$. This type of environment is called a Partially-Observed Markov Decision Process (POMDP) \citep{kaelbling1996reinforcement} and is substantially harder to solve optimally than an MDP due to the need to correctly infer the hidden states $\tilde{x}$ from the observations $\tilde{o}$. Nevertheless, the POMDP model has a great deal of generality since, as the state is hidden, it can be whatever is necessary to preserve Markovian dynamics, thus enabling any non-Markovian environment to be written in terms of a Markovian POMDP.

Early approaches to the control problem tried to use methods in variational calculus to directly find analytical solutions to the control problem. Such approaches yielded success in some simple but important, cases, such as Markov linear Gaussian dynamics and quadratic costs. These conditions correspond to dynamics which can be specified as (assuming discrete time),
\begin{align*}
x_{t+1} &= A x_t + B a_t + \omega \\
r_t &= x_t^T Q x_t + a_t^T R a_t \numberthis
\end{align*}
 
Where A, B, Q, and R, are known matrices and $\omega \sim \mathcal{N}(0,\mathbb{I})$ is white Gaussian Wiener noise \citep{wiener2019cybernetics}. In this case, an analytical solution exists in both continuous and discrete time which gives rise to the linear quadratic regulator \citep{kirk2004optimal,kalman1960contributions,kalman1960new}, a centerpiece of modern control theory which is remarkably effective for controlling even complex systems, and for which control solutions can be computed very relatively cheaply and in real time. While the linear dynamics and quadratic costs conditions are quite restrictive (especially the linear dynamics), the linear quadratic regulator approach can be extended somewhat to nonlinear dynamics by simply using a local linearity approximation at every timestep and applying model-predictive control. This iterative LQR \citep{li2004iterative} algorithm, is quite robust and can achieve significant feats of nonlinear control, including use in controlling industrial robotics \citep{feng2014optimization}. Other variational approaches have also been applied and can be quite effective in many cases. For instance, Pontryagin's maximimum principle \citep{kopp1962pontryagin,kirk2004optimal} or `bang-bang' control can be applied productively to find optimal policies in many settings. Recently, there has been advances using path integral methods and the Feynman-Kac lemma to find control solutions for certain classes of nonlinear dynamics \citep{kappen2005path,kappen2012optimal,williams2017information}.

Another approach to the control problem, which can work for arbitrary dynamics, is to simply optimize the control function by gradient descent with respect to the actions. Such an approach goes by the name of policy gradients, since given a policy function $a = f_\phi(s)$ parametrised by parameters $\phi$, we can simply compute gradients of the control problem loss $\frac{\partial \mathcal{L}_{control}}{\partial \phi}$ and optimize the parameters by stochastic gradient descent. The chief difficulty is to propagate gradients through the potentially nondifferentiable and unknown expectation under the environmental dynamics $\mathbb{E}_{p(\tilde{x} | \tilde{a})}$. Luckily, this is achievable through the policy gradient theorem \citep{williams1989learning, sutton1998introduction}
\begin{align*}
    \frac{\partial \mathcal{L}_{control}}{\partial \phi} &= \frac{\partial}{\partial \phi} \int d\tilde{x} \, p(\tilde{x}, \tilde{a}) \ln p(\tilde{r} | \tilde{x}, \tilde{a}) \\
    &=  \int d\tilde{x} \, \frac{\partial}{\partial \phi} p(\tilde{x}, \tilde{a}) \ln p(\tilde{r} | \tilde{x}, \tilde{a}) \\
    &= \int d\tilde{x} \, p(\tilde{x}, \tilde{a}) \frac{\partial \ln p(\tilde{x}, \tilde{a})}{\partial \phi}  \ln p(\tilde{r} | \tilde{x}, \tilde{a}) \\
    &= \mathbb{E}_{p(\tilde{x}, \tilde{a})}[ \ln p(\tilde{r} | \tilde{x}, \tilde{a}) \frac{\partial \ln p(\tilde{x}, \tilde{a})}{\partial \phi}] \numberthis
\end{align*}

Which allows an estimate of the gradient to be computed simply through averages of environmental dynamics. Importantly, this approach does not require knowledge of the true environmental dynamics at all (unlike classical control theory), since we only require samples from the environment which can be obtained simply through interacting with it. Intuitively, we can think of this theorem as saying that the gradient of the control objective is simply the average gradient of the policy, weighted by the rewards received. 

While this method works, the downside is that since the expectation is effectively computed through Monte-Carlo sampling (and generally relatively few samples at that), the gradient estimates generally have a very high variance, which makes learning troublesome and slow. A number of baseline approaches have been invented to try to deal with this problem \citep{sutton2018reinforcement} and to make it more tractable. Nevertheless, policy gradient approaches can be scaled up and applied successfully in challenging tasks, especially continuous control tasks, by parametrizing the policy $f_\phi(x)$ with a deep neural network and directly applying the policy gradient theorem with some additional tricks \citep{schulman2015trust,schulman2017proximal}.

Another approach to the control problem in Markov conditions (but arbitrary dynamics as long as they are Markov) is to use a recursive solution method pioneered by Richard Bellman \citep{bellman1952theory}. He noticed that optimal solutions to the control problem satisfy an interesting recursive relationship -- that the optimal path to the goal at a timestep $t$, must include the optimal path to the goal at a later timestep $t+1$. This property allows you to build up a backwards recursion where you start at the goal at the end and then work backwards, constructing the optimal path in a piecewise fashion from the previous optimal path. The key mathematical quantity, here, is the cost-to-go, which intuitively is the cost of the optimal trajectory from the current position to the goal. In modern reinforcement learning parlance, this cost-to-go is called the optimal value function of a state, and is conversely the expected reward which would be attained from a given state assuming the optimal policy is followed. Written out mathematically, this approach gives rise to the recursive Bellman equation,
 \begin{align*}
\mathcal{V}^*(x_t) = r(x_t) + \mathbb{E}_{p(x_{t+1} | a_t})[\mathcal{V}^*(x_{t+1})] \numberthis
 \end{align*}
which simply states that the optimal value function of a current state, is the reward of the current state plus the maximum average value function of the next state. In effect, if iterated backwards from the end (where the optimal value function is simply $r(x_T)$ and assumed known) this recursion allows you to build up the optimal path by working backwards. If all environmental states and actions are known and finite, then this algorithm can be run explicitly to compute the optimal solution in polynomial time (as opposed to the exponential time approach of just trying all possible paths and picking the best). This is thus a dynamic programming algorithm which is equivalent to other standard dynamic programming algorithms in computer science such as Dijkstra's algorithm for the shortest paths.

Importantly, the Bellman recursion holds not just for the optimal policy and value function, but indeed for \emph{any} policy and value function, this allows solution methods using this recursion to apply even when the state and action space is too large to represent explicitly. In this case, we can write the Bellman recursion as,

 \begin{align*}
 \label{value_update}
\mathcal{V}(x_t) = r(x_t) + \mathbb{E}_{p(x_{t+1} | a_t,x_ts})[\mathcal{V}(x_{t+1})] \numberthis
 \end{align*}

and, without working backwards from the end, we can simply estimate the value functions $\mathcal{V}_\pi(x)$ of a given policy $\pi$ by moving around in our environment, computing rewards and storing the state and the next state, and applying Equation \ref{value_update}. If we do this sufficiently for a given policy, we can then form an estimate of the global value function $\mathcal{V}_\pi(x)$ for all x. With this value function, we can then improve the policy, by simply defining a new policy that takes $\pi = max(r(x_t,a_t) + \mathbb{E}_{p(x_{t+1} |x_t, a_t)}[\mathcal{V}(x_{t+1})]$. Somewhat surprisingly, it has been proven that if the estimated value function is accurate, then the new policy defined in such a manner is necessarily the same or better (in terms of average reward obtained from the MDP) than the previous policy, and that if we iterate the process of sampling new states to estimate the value function, and then improving the policy, then we will converge upon the optimal policy $\pi^*$ \citep{sutton2018reinforcement}. This approach, called policy iteration, is the cornerstone of classic reinforcement learning algorithms such as temporal difference learning \citep{sutton1988learning}, and SARSA \citep{sutton1996generalization,singh1996reinforcement}.

 A closely related approach is called Q-learning \citep{watkins1992q}, which instead of using the value function, instead maintains an estimate of the state-action value function $Q(x,a)$, which is called the Q-function for historical reasons. The Q function satisfies a similar recursive relationship to the value function,
\begin{align*}
\mathcal{Q}(x_t,a_t) = r(x_t,a_t) + \underset{a_t}{argmax} \mathbb{E}_{p(x_{t+1},a_{t+1} | a_t,x_t})[\mathcal{Q}(x_{t+1},a_{t+1})]
\end{align*}

 Given an estimate of the Q function, the policy improvement step is simple. $\pi^{t+1} = \underset{a}{arg max} \, Q(s,a)$. The Q-learning algorithm combines both policy evaluation (estimating the value or Q function) and policy improvement into one continuous algorithm, which can be simply defined as,
 \begin{align*}
 a(x_t) &= max_{a_t} Q(x_t,a_t) \\
Q(x_t,a_t) &= r(x_t,a_t) + max_{a_{t+1}} \mathbb{E}_{p(x_{t+1} | x_t,a_t)}[\mathcal{Q}(x_{t+1},a_{t+1})] \numberthis
 \end{align*}
The Q-learning algorithm is extremely popular and effective and has been central to many major successes of reinforcement learning, from playing backgammon \citep{tesauro1994td} to Atari \citep{mnih2013playing,schrittwieser2019mastering}. The Q function and value function can be related straightforwardly by $\mathcal{V}(x) = \int da \mathcal{Q}(x,a)$ -- i.e. the value function is simply the Q function averaged over all actions. Another approach is to write everything instead in terms of an \emph{advantage function} $\mathcal{A}(x,a) = \mathcal{Q}(x,a) - \mathcal{V}(x)$ which simply subtracts the action-independent value function baseline from the Q-value, effectively normalizing it, since the only important thing from the perspective of Q learning is the \emph{relative values} of each action. This approach reduces the gradient variance when trying to estimate the Q function and often makes the resulting algorithms more stable \citep{hessel2018rainbow}.

 In classical reinforcement learning we typically represent the Q and value function explicitly in discrete-state and discrete-action environments. For instance, with discrete states, the value function $\mathcal{V}(x)$ would simply be a vector of length $\mathbb{S}$ where $\mathbb{S}$ is the number of distinct states. The Q function $\mathcal{Q}(x,a)$ is simply an $\mathbb{S} \times \mathbb{A}$ matrix where $\mathbb{A}$ is the action dimension. 

 Another useful representation is the successor representation \citep{dayan1997using}. This approach rewrites the value function in terms of the instantaneous reward $r(x)$ and a successor matrix $\mathcal{M}(x, x')$ which is a $\mathbb{S} \times \mathbb{S}$ matrix which represents the average transition probabilities for a given policy from state $x$ to state $x'$. This matrix $\mathcal{M}$ can be thought of as the stationary transition distribution of the Markov chain for a given policy. The value function can be decomposed into,
 \begin{align*}
\mathcal{V}(x) = \mathcal{M}(x,x')r(x) \numberthis
 \end{align*}

 The successor representation, crucially, by separating out the policy-dependent component (M) from the reward $r$ allows for computation of different value functions for a given policy rapidly under different reward functions. Thus this representation allows for very flexible changes of behaviour given a change in reward. Of course the optimal policy also changes under a change of reward function, and this change cannot be straightforwardly determined solely by the successor representation.

\subsection{Deep Reinforcement Learning}                                                                                  

While classical reinforcement learning approaches typically represent the value or Q functions explicitly as vectors and matrices, such methods only work for relatively small and discrete state and action spaces, and cannot easily scale to the extremely large state and action spaces required for playing complex games as well as for complex continuously valued action spaces such as in robotics, where simply discretizing the space with a sufficiently fine grid will simply result in too many states to handle. Therefore, to maintain scalability, instead of explicitly representing the state and value functions, it becomes necessary to approximate them with powerful function approximators. While other approaches using linear features \citep{baird1995residual,gordon1995stable}, or nonlinear basis function kernels \citep{doya2000reinforcement} are possible, recent systems have overwhelmingly used deep neural networks to approximate the value or Q functions directly. To make this explicit, instead of representing the value function $\mathcal{V}[x]$ as a vector, we instead represent it as a function $\mathcal{V}_\psi(x)$ which takes a state $x$ and maps it to a scalar value. This function is implemented by a deep neural network with parameters $\psi$. We can then learn these parameters using stochastic gradient descent on a loss function which is a modified version of the Bellman recursion,
\begin{align*}
\label{valuenet_equation}
\mathcal{L}_{Valuenet}(x) = (\mathcal{V}_\psi(x) - r(x) - max_a \mathbb{E}_{p(x' | x,a)}[\mathcal{V}_\psi(x')]))^2 \numberthis
\end{align*}
which is the squared residual between the value predicted for a state $x$ by the value network, and the value predicted by the Bellman recurrence relation. By using this approach, therefore, we utilize the intrinsic generalization capabilities of deep neural networks to allow us to estimate value or Q functions for an otherwise intractably large space. This approach can be straightforwardly extended to Q-learning and other Bellman based methods. Similarly, applying policy gradients with deep neural networks is even simpler -- we simply parametrise the policy $q_\psi(a | s)$ with a deep neural network with parameters $\psi$ and train it directly using stochastic gradient descent on the policy gradient objective \citep{schulman2015trust,schulman2017proximal}. 

To get this approach working well in practice, however, requires quite a number of tricks. For instance, the neural networks cannot be trained simply through continuous interaction with the environment, as this gives rise to correlated data which leads to overfitting and catastrophic forgetting within the value network \citep{mnih2013playing}. Thus, instead, it is necessary to make the data fed into the network as $i.i.d$ as possible by using a memory replay buffer, which stores all experience the agent has encountered over its lifetime and then replays it at random to be optimized according to Equation \ref{value_update} \citep{mnih2013playing}.  Similarly, note that in the value network update equation (Equation \ref{valuenet_equation}), the parameters of the value network appears twice -- once computing the value of $x$ and again computing the values of $x'$. It has been found empirically, that this leads to instabilities in the optimization process which often destroy learning, since the optimization process is effectively chasing a set of moving targets. To resolve this problem, the value estimates of $x'$ are often computed using a `frozen' value network which is not optimized directly, but is a copy of the value network from some number of iterations past. The frozen network is then updated to match the current value network every given number of iterations \citep{mnih2015human}. Another issue is that the value estimates are often skewed extremely positive due to the max operator in the objective interacting inaccurate value function estimates. This can be ameliorated empirically by simply training two (or many) value networks in parallel, and then choosing the smallest value estimates out of all of them, a technique known as dueling value networks \citep{wang2016dueling}. For a thorough review of tricks and tips for training deep reinforcement learning agents, we suggest \citet{fujimoto2018addressing,hessel2018rainbow}.

Nevertheless, once all of these instabilities have been addressed, the result is an extremely powerful and general learning technique which has been demonstrated empirically to scale up to solve very challenging tasks such as Atari games \citep{mnih2015human,mnih2014neural} and Go \citep{silver2016mastering,silver2017mastering}, Starcraft II \citep{vinyals2019grandmaster}, and ultimately very challenging tasks in robotics \citep{nagabandi_neural_2017,nagabandi2019deep,chua_deep_2018,williams2017model}. 

\subsection{Model-free vs Model-based}

It is important to note that all the methods we have discussed so far require samples from the true environmental dynamics to approximate the expectation $\mathbb{E}_{p(\tilde{x}, \tilde{a})}$ from the ultimate loss function (Equation \ref{control_objective}. While these samples are easy to acquire in the case where interacting with the environment is cheap and easy, such as when the environment is a simulation such as a game or an OpenAI gym environment \citep{brockman2016openai}, in many real world tasks this is not the case. For instance, in robotics, interacting with the real environment is often slow (the real robot has to actually move or do things in physical space) and costly (this movement requires power and also induces wear and tear on the robot. In extreme situations, bad policies may actually result in the robot damaging itself). In this case, it is often better if we can somehow eschew interacting with the real environment in favour of a model of the real environment. Having a `world model' \citep{ha_recurrent_2018} of the environment allows the agent to plan and test different potential courses of action without having to sustain costly and slow interactions with the real world. The utility of models does not just extend to the environmental dynamics. It is often the case that the actual \emph{reward function} of the agent is unknown. This is not generally true in reinforcement learning and control theory, which typically have well specified rewards, but is often true in the case of biological organisms encountering novel contingencies, where it is not necessarily known a-priori if a situation is good or bad. Thus, we can also learn a model of the reward function as well.

Mathematically, we can formalize this property of having a model of the transition dynamics or reward functions by postulating additional probability densities which the agent possesses $q_\phi(\tilde{x} | \tilde{a})$ and $q_\theta(\tilde{r} | \tilde{x}, \tilde{a})$ which represent the agent's model of the true environmental dynamics $p(\tilde{x} | \tilde{a})$ and true reward function $p(\tilde{r} | \tilde{x}, \tilde{a})$. Then, using importance sampling, we can introduce these models into the previous loss function,
\begin{align*}
\mathcal{L}_{control} &= \underset{a}{argmax} \, \mathbb{E}_{p(\tilde{x} | \tilde{a})p(\tilde{a})}[ \ln p(\tilde{r} | \tilde{x}, \tilde{a})] \\
&= \underset{a,\phi,\theta}{argmax} \, \mathbb{E}_{p(\tilde{x} | \tilde{a})p(\tilde{a})} \frac{q(\tilde{x} | \tilde{a})}{q(\tilde{x} | \tilde{a})}    [ \ln p(\tilde{r} | \tilde{x}, \tilde{a}) \frac{q(\tilde{r} | \tilde{x}, \tilde{a})}{q(\tilde{r} | \tilde{x}, \tilde{a})}] \\
&= \underset{a,\phi,\theta}{argmax} \, \mathbb{E}_{q(\tilde{x} | \tilde{a})p(\tilde{a})} \frac{p(\tilde{x} | \tilde{a})}{q(\tilde{x} | \tilde{a})}    [ \ln q(\tilde{r} | \tilde{x}, \tilde{a}) \frac{p(\tilde{r} | \tilde{x}, \tilde{a})}{q(\tilde{r} | \tilde{x}, \tilde{a})}] \\
&= \underset{a,\phi, \theta}{argmax} \, \mathbb{E}_{q(\tilde{x} | \tilde{a})p(\tilde{a})} \ln \frac{p(\tilde{x} | \tilde{a})}{q(\tilde{x} | \tilde{a})}    [ \ln q(\tilde{r} | \tilde{x}, \tilde{a}) \frac{p(\tilde{r} | \tilde{x}, \tilde{a})}{q(\tilde{r} | \tilde{x}, \tilde{a})}] \\
&= \underset{a,\phi, \theta}{argmax} \, \underbrace{\mathbb{E}_{q(\tilde{x} | \tilde{a})p(\tilde{a})} [ \ln q(\tilde{r} | \tilde{x}, \tilde{a}) ]}_{\text{Reward Maximization}}  - \underbrace{KL[q(\tilde{x} | \tilde{a}) || p(\tilde{x} | \tilde{a})]}_{\text{System Identification}}  -  \underbrace{\mathbb{E}_{q(\tilde{x} | \tilde{a})p(\tilde{a})} \big[ \frac{q(\tilde{r} | \tilde{x}, \tilde{a})}{p(\tilde{r} | \tilde{x}, \tilde{a})}}\big]_{\text{Reward Model Identification}} \numberthis \\
\end{align*}

Here we see that the objective can be partitioned into three terms. The first, reward maximization, is equivalent to the original control objective except it utilizes the model environmental dynamics and reward function instead of the true environmental dynamics and reward function. The second term, `system identification', encodes the KL divergence between the true and modelled environmental dynamics, and is minimized with respect to the parameters of the model of the dynamics. Optimizing this term encourages the agent to learn an accurate dynamics model of the world. The third term, `reward model identification' encourages the reward model to match the true reward function and optimizing this term with respect to the parameters of the reward model minimizes the difference between the model reward distribution and the true reward distribution.

Interestingly, mathematically, the minimization of all of the three parameters is over all three terms. This leads to, for instance, the parameters of the reward and dynamics model to be optimized over the reward maximization term, which effectively encourages the dynamics and reward models to try to learn positively biased dynamics and reward models which are encouraged to give out larger rewards. Conversely, the divergence terms between true and modelled environmental dynamics and reward function are also being optimized with respect to action, so the optimisation process also encourages action to make the true dynamics and true reward function correspond more closely to the modelled ones. These interactions between the optimizations of the different terms often have strange and deleterious effects on agent behaviour, especially learning the dynamics to maximize the reward, which can give agents a highly dysfunctional `optimism bias' \citep{levine2018reinforcement}. As such, in practice, this optimization is often split into three separate and independent optimizations for each set of parameters respectively,
\begin{align*}
& \underset{a}{argmax} \, \, \mathbb{E}_{q(\tilde{x} | \tilde{a})p(\tilde{a})} [ \ln q(\tilde{r} | \tilde{x}, \tilde{a})  \numberthis \\
& \underset{\phi}{argmin} \, \,  KL[q_\phi(\tilde{x} | \tilde{a}) || p(\tilde{x} | \tilde{a})]  \\
& \underset{\theta}{argmin} \, \, \mathbb{E}_{q(\tilde{x} | \tilde{a})p(\tilde{a})} \frac{q(\tilde{r} | \tilde{x}, \tilde{a};\theta)}{p(\tilde{r} | \tilde{x}, \tilde{a})} \numberthis
\end{align*}

These three optimization processes correspond to maximizing the reward (standard reinforcement learning), learning the environmental dynamics model, and learning the reward function, respectively. In control theory, the process of learning a dynamics model is called system identification. 

Given good reward and dynamics models, it is then possible to utilize standard model-free reinforcement learning approaches to estimate Q and value functions, or estimate policy gradients directly through simulated samples of the dynamics model. With these it is then possible to learn policies in the usual way without ever having to interact with the environment (except to learn the models in the first place). This approach, first introduced in the Dyna architecture \citep{sutton1991dyna} has been studied in the literature for a long time, and is generally more sample efficient than pure model-free reinforcement learning which learns directly from environmental transitions, as learning dynamics models of the world is typically faster than learning reward models, since the prediction errors in the state of the world is a richer informational signal than the reward prediction error since the reward is usually scalar while the world-state is typically very high dimensional.

Another approach, once you have a model is to skip the model-free methods and move directly to planning using the model. In the simplest case, this can be done by sampling different action trajectories, simulating their consequences and rewards using the dynamics and reward models, and then simply choosing the action trajectory with the best estimated rewards. More advanced methods include the cross-entropy method, which tries to fit a probabilistic (Gaussian) action distribution to maximize rewards, and the path-integral method, which fits a Boltzmann distribution of action trajectories over rewards \citep{kappen2012optimal,rawlik2013probabilistic,theodorou2010reinforcement,williams2017model,williams2018predictive}. This kind of model-based planning is often combined with Model-Predictive-Control, which is simply where you re-plan fully at every time-step, and has been used to reach state of the art performance on a wide variety of reinforcement learning \citep{nagabandi2019deep} and robotics tasks \citep{williams2016aggressive,williams2017information}, all while requiring substantially fewer environmental interactions than model-free reinforcement learning approaches.

\subsection{Exploration and Exploitation}

An interesting question that arises in the control problem, wherever there are any unknowns, either of the optimal value function and policy or the environmental dynamics and reward function, is the exploration-exploitation tradeoff \citep{cohen2007should,dayan2008decision,sutton1998introduction,kaelbling1998planning,mobbs2018foraging,berger2014exploration}. This trade-off arises because in order to obtain a more accurate estimate, it is usually necessary to explore new regions of the state-space away from the locally optimal location. However, by ignoring the locally optimum course of action, you incur an opportunity cost equal to the difference between the (usually worse) reward from the exploring compared to the local optimum. Conversely, by only ever exploiting the local optimum and never exploring, if there are actually better optima out there, which you have simply not found, you incur a constant opportunity cost of the distance to the true optimum every single time you exploit your local optimum. Thus, to find the truly optimal behaviours, it is necessary to explore widely, and not just be sucked into whatever the closest local optima you find. However, exploration has an intrinsic cost associated with it, since assuming you have a halfway decent local optimum, almost everything you explore will be worse than that, and thus exploration must be kept to a minimum to maximize return, even in the long run.

The exploration-exploitation tradeoff is also central to the behaviour and performance of reinforcement learning agents, especially deep reinforcement learning agents which cannot explicitly represent every contingency in the state-space. Empirically, it has been found that except in extremely simple tasks, or where the reward function is a smooth gradient to the global optimum, some forms of exploration are necessary to achieve good performance with deep reinforcement learning techniques. A large number of heuristic exploration techniques have been developed in the literature which work well for many tasks. Perhaps the simplest of these is the $\epsilon$-greedy approach \citep{sutton1998introduction}, which simply takes the greedy action $(1 - \epsilon)$ amount of the time, and takes a random action $\epsilon$ percent of the time, where $\epsilon$ is a small number -- typically about $0.02-0.05$. This method essentially bakes in a certain degree of random exploration into the method so that it will (eventually) explore all contingencies purely due to its random actions. In many control tasks this method generates sufficient exploration to enable good performance. More sophisticated variants anneal the value of $\epsilon$ over time; working on the idea that at the beginning when little is known you should explore more, and then later when you already have a pretty good policy you should explore less. Another approach in algorithms like Q learning is that instead of simply taking the maximum value, take actions with a probability equal to their softmaxed Q-values -- $q(a | x) = \frac{\beta exp(-\mathcal{Q}(a | x))}{\sum_a exp(-\mathcal{Q}(a | x))}$ where $\beta$ is a parameter which controls the spread or entropy of this distribution. When $\beta$ is large, then the distribution is highly peaked and tends towards the max. When $\beta$ is small, then the distribution tends towards a uniform over all action possibilities. This method is called Boltzmann exploration \citep{cesa2017boltzmann}, since the action distribution is equivalent to the Boltzmann distribution of statistical mechanics with the $\beta$ parameter functioning as an inverse temperature.

Another approach, which we will explore in detail in the next section, is to add an entropy maximization term to the objective \citep{levine2018reinforcement}. Thus, instead of simply maximizing the reward, the goal is to maximize the reward while keeping the entropy of the policy as great as possible, and thus making action as random as possible \footnote{This idea is similar to, but distinct from, ideas such as upper-confidence-bound sampling which assign optimism bonuses to states in proportion to the inverse degree to which they have been sampled. The key difference is that maximum entropy approaches provide bonuses to actions based on the overall entropy of the action distribution while UCB algorithms provide bonuses to states based on their epistemic uncertainty.} While still a random kind of exploration, this performs better than $\epsilon$ greedy approaches since it explicitly trades off randomness and reward maximization in the objective, rather than as an $\epsilon$ parameter that needs to be tuned by hand.

While all these exploration methods work well in practice in many benchmark environments for deep reinforcement learning, they all fundamentally utilize \emph{random exploration} -- i.e. actions are selected randomly in order to drive exploration. It is important to note, however, that this strategy is necessarily very inefficient, since even with purely random actions, a random walk will explore slowly. Indeed, these methods tend to perform rather poorly in more challenging large environments where all useful contingencies cannot realistically be explored with a random walk in a reasonable amount of time, and tend to perform especially poorly in \emph{sparse reward} environments, where rewards are hard to obtain and often require a pretty good policy to even get any reward at all \citep{tschantz2020reinforcement}. A good example of such an environment is many games, where you are only rewarded a 1 if you win the game. However, to even win any games at all requires some skill at playing \footnote{This is often addressed in practice by using reward shaping, where typically either the agent designer or the environment designer will create a `proxy' reward function for the real one which is less sparse. For instance, in a game like chess, instead of simply rewarding winning or losing the game, the agent might get rewards for taking pieces, or gaining positional advantage. While this approach works very well in practice, it requires human intervention for every task the agent tries to accomplish, and is often tricky to design a proxy reward which successfully leads to the correct ultimate behaviour. Ultimately, we want agents to be able to interact with the world fully autonomously, which means that they should not need special human-designed reward functions to handle any new situation, and so we do not consider reward shaping further}. To address the shortcomings of random exploration, a significant amount of work has been done on \emph{directed}, or information-seeking, exploration. Here, exploratory actions are not completely random but instead directed at some exploratory goal -- usually to accumulate information or to resolve uncertainty about the world. This makes sense as in a stationary environment there is little point in repeatedly exploring bad options which are known to be bad. The key is to explore where there is remaining resolvable uncertainty. 

A number of different objectives, or `intrinsic measures' \citep{oudeyer2009intrinsic} have been proposed to achieve this. These include prediction error minimization \citep{pathak2017curiosity}, ensemble divergence \citep{chua_deep_2018}, explicit information gain \citep{shyam_model-based_2019,tschantz2020reinforcement,sun_planning_2011}, and empowerment \citep{klyubin2005empowerment}. Typically such approaches postulate a separate `exploration objective' and then either operate in two phases whereby first the model optimizes the exploration objective, and then it switches to optimizing the greedy objective \citep{shyam_model-based_2019}, or alternatively, the exploration and greedy objectives are added together to form a unified objective function which is minimized throughout \citep{tschantz2020reinforcement}. Such approaches therefore encourage the agent to seek a balance between its exploratory and reward-maximization imperatives. This has the theoretical advantage of encouraging the agent to only explore regions of the state-space which combine both a high expected reward \emph{and} much resolvable uncertainty, as opposed to simply resolving uncertainty for its own sake. In the literature, most of these exploratory objectives are simply postulated and argued for on intuitive grounds and then empirically compared. However, with the exception of the entropy maximization discussed above, which we shall see arises from explicitly considering control as a variational inference problem, the mathematically principled origins of these additional exploratory terms, especially information-gain and empowerment terms remains mysterious. Chapter 5 is dedicated to deriving the mathematical basis for such objectives.

\subsection{Control as Inference}

Since we have been so interested in understanding brain function through the lens of Bayesian (variational) inference in this thesis, a natural question arises as to whether the control problem as discussed previously can be cast in such an inference framework. It turns out that this is indeed the case, and is quite straightforward to achieve. Starting from \citep{attias2003planning} and then developed by \citep{toussaint2006probabilistic,todorov2008general,rawlik2013probabilistic,kappen2012optimal,theodorou2010generalized,levine2018reinforcement}, a small line of the literature has worked on investigating and developing the close connections between the control problem and Bayesian inference.
\begin{figure}
    \begin{center}
          \includegraphics[width=0.35\textwidth]{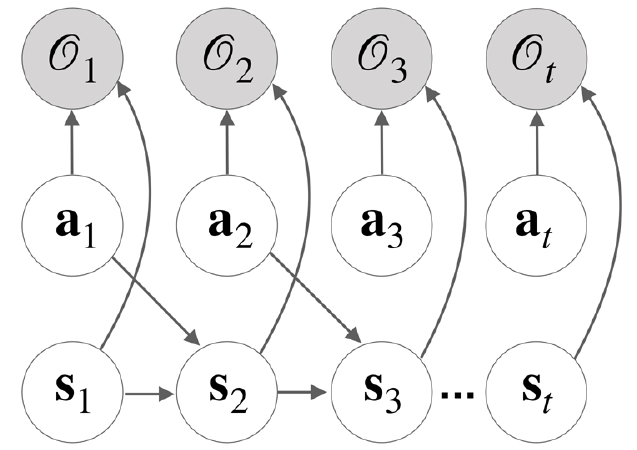}
    \end{center}
    \caption{Graphical model for control as inference, with optimality variables $\Omega$. Other than the optimality variables, the graphical model takes the form of a Markov Decision Process with actions $a$ and states $s$. The state of a specific timestep depends on the action and state of the last time-step. By writing out an explicit graphical model like this, we can apply a whole field's worth of inference algorithms on graphical models like this to solve control problems.}
\label{fig:graphical-model}
\end{figure}

While the control objective (Equation \ref{control_objective} is a probabilistic objective, it is not yet an \emph{inference} objective. There is nothing there to be inferred. The key step is to define dummy variables $\Omega_{1:T}$ which are binary random variables which simply whether a given trajectory timestep is optimal or not. $\Omega_t = 1$ if the timestep is optimal and $\Omega_t = 0$ if it is not optimal. Given this dummy variable, the task of inferring the optimal policy can be written simply as finding the distribution $p(\tilde{a}, \tilde{x} | \tilde{\Omega} = 1)$. To begin inferring this distribution, it is first necessary to make one more assumption about the dummy variables $\Omega$, in order to operationalize the notion of optimality. We define $p(\Omega_t = 1 | a_t, x_t) \propto exp(-r(x_t, a_t))$ such that the probability of optimality is proportional to the exponentiated reward. Intuitively, this can be seen as a mathematical trick allowing the `log-likelihood of optimality' to be equal to the reward, thus allowing us to cast reward maximization as a process of maximum likelihood estimation.

One way we can find the crucial distribution is simply to directly compute it via Bayes rule. First, we write out Bayes rule explicitly,
\begin{align*}
p(a_t | x_t, \tilde{\Omega}) &= \frac{p(\tilde{\Omega}, x_t, a_t)}{p(\tilde{\Omega}, x_t)} \\
&= \frac{p(\tilde{\Omega}| x_t, a_t)p(a_t | x_t)p(x_t)}{p(\tilde{\Omega} | x_t)p(x_t)} \\
&= \frac{p(\tilde{\Omega}| x_t, a_t)p(a_t | x_t)}{p(\tilde{\Omega} | x_t)} \\
&\approx \frac{p(\tilde{\Omega}| x_t, a_t)}{p(\tilde{\Omega} | x_t)} \numberthis
\end{align*} 
Where, in the final line, we assume that the action prior $p(a_t | x_t)$ is uniform. Now, looking at the two terms we have left, we can intuitively think of the numerator $p(\tilde{\Omega}| x_t, a_t)$ as representing the probability of optimality of all future states, given the current state and action. However, we have another term for this -- the cost-to-go, or the Q-function. In effect, we obtain the Bellman recursive relationship directly from Bayes rule. Secondly, the denominator $p(\tilde{\Omega} | x_t) = \int d a_t \, \, p(\tilde{\Omega}| x_t, a_t)$ clearly corresponds to the value function.

We can from this directly derive recursive relationships among these terms,
\begin{align*}
    p(\Omega_{t:T}| x_t, a_t) = \int dx_{t+1} da_{t+1} \, p(\Omega_{t+1:T} | x_{t+1}, a_{t+1}) p(x_{t+1}, a_{t+1} | x_t, a_t) p(\Omega_t | x_t, a_t) \numberthis
\end{align*}
We can thus see, that due to our definition of optimality that $p(\Omega_t | x_t, a_t) = exp(-r(a_t, x_t)$, that to obtain the correspondence to the value and Q function requires the \emph{log} of the optimality probability. We thus have,
\begin{align*}
    & \mathcal{Q}_{CAI} = \ln p(\Omega_{t:T}| x_t, a_t) \\
    & \mathcal{V}_{CAI} = \ln p(\Omega_{t:T}| x_t) \numberthis
\end{align*}
Where, by taking the log of the integral and exponential, we effectively have the log-softmax function instead of the max in the Bellman equations. This corresponds to a `soft' maximum instead of the hard maximum used in the traditional Bellman recursion. However, other than that, our new definitions of the value and Q function satisfy the standard Bellman recursive relationship, and as such can be used to derive all the traditional reinforcement learning algorithms such as Q-learning, temporal difference learning, and SARSA \citep{sutton1996generalization}.

Another approach is to use variational inference and attempt to approximate the true posterior distribution given optimality $p(\tilde{a}, \tilde{x} | \tilde{\Omega})$ with a variational distribution $q(\tilde{x}, \tilde{a})$. To make this approximation accurate, we thus wish to minimize the divergence between the two distribution.
\begin{align*}
\mathcal{L}_{CAI} &= \KL[q(\tilde{x}, \tilde{a}) || p(\tilde{x}, \tilde{a} | \tilde{\Omega})] \\
&= \KL[q(\tilde{x}, \tilde{a}) || \frac{p(\tilde{x}, \tilde{a}, \tilde{\Omega})}{\tilde{\Omega}}] \\
&= \underbrace{\KL[q(\tilde{x}, \tilde{a}) || p(\tilde{x}, \tilde{a}, \tilde{\Omega})]}_{\text{ELBO}}+ \ln \tilde{\Omega} \numberthis
\end{align*}

Where we only need to optimize the Evidence Lower Bound (ELBO) term since $\ln \tilde{\Omega}$ is constant with respect to the variational density $q(\tilde{x}, \tilde{a})$. Crucially, we can then split up the ELBO term as follows,
\begin{align*}
\mathcal{L}_{CAI} &= \KL[q(\tilde{x}, \tilde{a}) || p(\tilde{x}, \tilde{a}, \tilde{\Omega})] \\
&= \KL[q(\tilde{a} | \tilde{x})q(\tilde{x}) || p(\tilde{\Omega} | \tilde{x}, \tilde{a})p(\tilde{a} | \tilde{x})p(\tilde{x})] \\ 
&= \underbrace{\mathbb{E}_{q(\tilde{x},\tilde{a})}[\ln p(\tilde{\Omega} | \tilde{x}, \tilde{a})]}_{\text{Reward Maximization}} + \underbrace{\KL[q(\tilde{a} | \tilde{x}) || p(\tilde{a} | \tilde{x})]}_{\text{Action Divergence}} + \underbrace{\KL[q(\tilde{x}) || p(\tilde{x})]}_{\text{Dynamics Divergence}} \numberthis
\end{align*}

If we then assume that the variational dynamics $q(\tilde{x})$ are equal to the true environmental dynamics $p(\tilde{x})$ (the agent cannot change the dynamics except through action) then the action divergence term disappears. Additionally, if we use the fact, defined earlier, that $p(\tilde{\Omega} | \tilde{x}, \tilde{a}) = \prod_t exp(-r(a_t, x_t))$, then we obtain an objective which looks substantially more similar to traditional reinforcement learning objectives except for an additional action divergence term between the variational action distribution (the policy) and a prior action distribution.
\begin{align*}
\mathcal{L}_{CAI} &= \underbrace{\mathbb{E}_{q(\tilde{a} | \tilde{a})p(\tilde{x})}[\prod_t^{T} r(x_t, a_t)]}_{\text{Reward Maximization}} + \underbrace{\KL[q(\tilde{a} | \tilde{x}) || p(\tilde{a} | \tilde{x})]}_{\text{Action Divergence}} \numberthis
\end{align*}

If we further assume a uniform action prior, then the control as inference objective reduces to,
\begin{align*}
\mathcal{L}_{CAI} &= \mathbb{E}_{q(\tilde{a} | \tilde{a})p(\tilde{x})}[\prod_t^{T} r(x_t, a_t)] - \mathbb{H}[q(\tilde{a} | \tilde{x})] \\
&= \mathbb{E}_{q(\tilde{a} | \tilde{a})p(\tilde{x})}[\prod_t^{T} r(x_t, a_t) - \ln q(a_t | x_t)] \numberthis
\end{align*}

Which is simply reward maximization while also simultaneously maximizing the entropy of the policy $q(\tilde{a} | \tilde{x})$. This objective has been utilized in a number of recent works \citep{rawlik2013stochastic,haarnoja2017reinforcement,haarnoja2018soft,abdolmaleki2018maximum}, and forms the basis of the soft-actor critic architecture \citep{haarnoja2018soft}, which simply optimizes a relatively standard actor-critic architecture on this objective. It has been found to reach state-of-the-art performance for model-free reinforcement learning on a wide range of challenging continuous control tasks \citep{hessel2018rainbow,haarnoja2018applications}. Moreover, the simplicity and robustness of this algorithm allow it to serve as an influential benchmark for the field. This CAI objective can be straightforwardly optimized by taking gradients of $\mathcal{L}_{CAI}$ against whatever parameters there are. For instance, optimizing the control-as-inference objective is identical to standard policy gradients except that each reward the agent receives has $\ln q(a_t | x_t)$ subtracted from it. This differentiates control as inference from previous works \citep{o2017deep}, which heuristically used an entropy regulariser to prevent policy collapse, but computed the entropy directly outside of the expression for the reward. The control-as-inference objective is both simpler to implement and more robust than this method.

Importantly, the control as inference objective, as it is a variational bound, can be derived directly from, and as a lower bound on the marginal likelihood of optimality $\ln p(\tilde{\Omega})$. The derivation is straightforward and goes as follows,

\begin{align*}
\ln p(\tilde{\Omega}) &= \ln \int p(\tilde{\Omega}, \tilde{x}, \tilde{a}) \frac{q(\tilde{a} | \tilde{x})}{q(\tilde{a}, \tilde{x})} \\
&\leq \mathbb{E}_{q(\tilde{a}, \tilde{x})}[\frac{p(\tilde{\Omega}, \tilde{x}, \tilde{a})}{q(\tilde{a}, \tilde{x})}] \\
&\leq \mathbb{E}_{q(\tilde{a} | \tilde{x})p(\tilde{x})}[\frac{p(\tilde{\Omega} | \tilde{x}, \tilde{a})p(\tilde{a} | \tilde{x})}{q(\tilde{a}| \tilde{x})}] \\
&\leq \mathbb{E}_{q(\tilde{a} | \tilde{x})p(\tilde{x})}[\prod_t^T r(x_t, a_t)] - \KL[q(\tilde{x} | \tilde{a}) || p(\tilde{a} | \tilde{x})] \\
&\leq \mathcal{L}_{CAI} \numberthis
\end{align*}

Under the assumption that the variational and generative dynamics are the same. Overall, the control as inference approach demonstrates that it is possible, even relatively straightforward, to derive a range of reinforcement learning algorithms from a variational approach on an MDP graphical model augmented with additional optimality variables. Doing so results in the standard reward maximization objective plus a regularisation term which tries to keep the learned policy $q(\tilde{a} | \tilde{x})$ as close as possible to some action prior $p(\tilde{a} | \tilde{x})$. If the action prior is set to be uniform, then this regularising KL divergence simply reduces to the maximization of the entropy of the action policy. This action entropy maximization term functions as a powerful regulariser and implicit exploratory drive, which aims to keep the policy as random as possible while still maintaining performance. This term is especially powerful and important for preventing policy collapse, a well-known phenomenon in policy gradient and actor-critic methods \citep{fujimoto2018addressing}, in which the probabilistic policy of an agent typically collapses to some deterministic policy which may not even be very good. Once it is in this state, it is very difficult for the agent to continue to explore to find better policies, since it has minimal probability of taking other actions. The theoretical benefits of this action entropy term have been demonstrated empirically in the literature, where `soft' control as inference approaches have generally shown to outperform classical reinforcement approaches as well as being more stable and easy to train. However, it is important to note that although the action entropy term is effective at maintaining exploration, it only encourages \emph{random} exploration, or random walk behaviour in action-space. To solve sparse reward tasks in a reasonable amount of time, it is possible that a more intelligent, \emph{directed} exploration strategy is needed, which focuses explicitly on minimizing resolvable uncertainty about the world. Such information-seeking exploration objectives are a major focus and benefit of active inference, and understanding their mathematical nature and origins is the major task of Chapter 5 of this thesis.

\section{Deep Active Inference}

Active inference and reinforcement learning both purport to solve the same fundamental problem -- that of adaptive action selection to maximize some notion of rewards or desires, given uncertainty about the world and about the optimal policy to take. While active inference arises from the paradigm of variational Bayesian inference with posterior policy distributions and complex generative world models \citep{friston_active_2015,friston2017process}, reinforcement learning arises primarily from the Bellman equation and the recursive properties of optimality \citep{sutton2018reinforcement,kaelbling1996reinforcement}, and utilizes constructs such as value functions and policy networks to learn adaptive behaviour even on challenging and complex control tasks.

Despite the very different origins of the two fields, since they are fundamentally trying to solve the same problem, it seems likely that there is much each field can learn from the other, since they both illuminate difference facets of the same reality. Specifically, the discrete-state-space active inference models introduced previously, in Chapter 2, suffer from many limitations of scale. Specifically, they represent the core distributions as discrete categorical variables, which require a relatively small, discrete and known state-space to function. Moreover, active inference models typically assume knowledge of the true generative process (i.e. the likelihood and prior matrices (A and B)), which often cannot simply be assumed in more realistic control tasks. \footnote{There is some work empirically investigating learning the A matrix using dirichlet hyperpriors over its values \citep{schwartenbeck_computational_2019}, and the rules for learning the B matrix are also straightforward. However, most of the active inference literature eschews these methods in favour of hand-designed likelihood and transition matrices \citep{friston_active_2015,friston2012active,parr2017uncertainty,friston_deep_2018}, and large scale studies of the effectiveness of these learning algorithms has not been ascertained at scale.}. An additional, and serious obstacle to the scalability of classical active inference methods is the computation of the policy prior, which is often taken to be the softmax of the expected free energy over all policies. This is typically computed explicitly and exactly in the literature \citep{da2020active,friston_active_2015}, and requires an explicit enumeration of every single policy and its associated trajectory for which the expected free energy can then be computed. In computational complexity terms, this results in exponential complexity in both the time horizon and the size of the discrete state-space, which clearly poses a significant computational scaling issue even for relatively small state-spaces and time-horizons. It is largely this obstacle which has prevented the application of truly large scale active inference models and limited most studies to toy tasks. There have been several methods in the literature proposed to somewhat ameliorate the computational expense of the expected free energy, notably by pruning away policies which have an a-priori likelihood less than some threshold \citep{friston2020sophisticated}. However, although such approaches enable scaling to slightly larger tasks, they do not attack the fundamentally exponential complexity of the algorithm, rather they simply reduce the exponential coefficient.

Indeed, all the scaling limitations of active inference are almost identical to those of tabular reinforcement learning with explicitly represented state and action value functions. In reinforcement learning, this scaling barrier was removed through the use of deep neural networks as flexible function approximators, to learn, via gradient descent, to approximate the required constructs by training on a dataset of environmental interactions \citep{sutton2018reinforcement}. We propose a similar approach may prove equally useful for scaling up active inference models. In the next two sections we present two studies which attempt to scale up active inference by using deep neural networks to flexibly approximate key densities in the active inference equation, as well as utilize methods from deep reinforcement learning to approximate the evaluation of the expected free energy over policies, which is fundamental to action selection in active inference. 

In the first study, which is based on the paper \citep{millidge_deep_2019}, we are heavily inspired by model-free deep reinforcement learning algorithms. We represent the transition dynamics, observation likelihood, and variational action distribution as neural networks trained to jointly minimize the variational free energy. Similarly, we utilize a bootstrappped value network to approximate the expected-free energy value function. We show that this model-free deep active inference approach can scale to perform equivalently, if not sometimes superiorly to contemporary deep reinforcement learning approaches.

In the second study, which is based on the paper \citep{tschantz2020reinforcement}, which was a joint collaboration with Alexander Tschantz at the University of Sussex, we utilize a scheme inspired by model-based active inference for action selection. Specifically, we use a model-based iterative planner to estimate the variational action distribution, and estimate the expected free energy value function based on simulated rollouts within the planner. We focus more heavily on the exploratory nature of the behaviour furnished by the expected free energy (and free energy of the expected future -- to be discussed in chapter 4) objectives and demonstrate that optimizing these objectives leads to empirically better performance, especially on sparse-reward tasks which require substantial amounts of exploration.

\subsection{Model-Free: Active Inference as Variational Policy Gradients}

\subsubsection{Derivation}
The fundamental idea of active inference is to reformulate the control problem as a variational inference one, and then use variational methods to solve it. Specifically, we wish to recast the problem of control into one of inferring the optimal state and action distribution. The formal setup we use to describe this is a discrete-time Partially Observed Markov Decision Process (POMDP) model. In this model, the agent receives observations $o_{1:T} \in \mathbb{O}$, which are generated by some hidden environmental state $x_{1:T} \in \mathbb{X}$ which satisfies the Markov property. The observations themselves do not necessarily have to be Markov. The agent can then emit actions $a_{1:T} \in \mathbb{A}$ which can alter the latent state of the environment and thus generate new observations. We assume that the agent maintains a desire distribution $\tilde{p}(o_{1:T}) = \prod_{t:T} exp(-r(o_t))$ over observations such that it most desires to be experiencing high rewards. Observations, states, and actions are optimized over full discrete-time trajectories from times $t=0$ to a given time horizon $T$. Given these assumptions, we can write down a factorization of the environmental POMDP as follows,
\begin{align*}
p_{env}(o_{1:T}, x_{1:T} | a_{1:T}) = p_{env}(x_1)p_{env}(o_1 | x_1)\prod_{t=2}^T p_{env}(o_t | x_t)p(x_t | x_{t-1}, a_{t-1}) \numberthis
\end{align*}

We then assume that the agent knows the basic POMDP structure and factorisation properties of the agent (although not necessarily any details about the precise distributions involved), and maintains an additional generative distribution over actions which specify its ideal action generating process. We can thus write the agent's generative model as,
\begin{align*}
p_{agent}(o_{1:T}, x_{1:T}, a_{1:T}) = p_{agent}(o_1, x_1)p_{agent}(a_1) \prod_{t=2}^T p_{agent}(o_t | x_t)p_{agent}(x_t | x_{t-1}, a_{t-1}) p_{agent}(a_t | x_t) \numberthis
\end{align*}

From now on, since the true environmental generative process is never known, we do not refer to it, only the generative model of the agent. Thus, for notational convenience, we denote $p_{agent}$ simply as $p$. The inference problem we wish to solve, is to infer the optimal action and state distribution given observations. That is, the key idea in active inference is to infer the distribution,
\begin{align*}
p(x_{1:T}, a_{1:T} | o_{1:T}) \numberthis
\end{align*}
Note that unlike in control as inference approaches which encode reward directly into the inference process by performing inference on a graphical model augmented with additional optimality nodes, here we encode rewards or goals into the model through the action prior, as we shall see later. In most situations, a direct computation of this posterior distribution is intractable, so we resort to a variational approximation. We define the variational distribution $q(a_{1:T}, x_{1:T} | o_{1:T})$ which is under the control of the agent, and then try to minimize the divergence between the true and approximate posteriors,
\begin{align*}
 \mathcal{L} = \underset{a_{1:T}}{argmin} \, \, \KL[q(a_{1:T}, x_{1:T} | o_{1:T}) || p(x_{1:T}, a_{1:T} | o_{1:T})] \numberthis
\end{align*}
This divergence is still intractable since it contains the intractable posterior, however we can derive a computable bound on this divergence known as the variational free energy, which we can then optimize,
\begin{align*}
\mathcal{F}(o_{1:T}) &= \KL[q(a_{1:T}, x_{1:T} | o_{1:T}) || p(x_{1:T}, a_{1:T}, o_{1:T})] \\ 
&= KL[q(a_{1:T}, x_{1:T} | o_{1:T}) || p(x_{1:T}, a_{1:T} |  o_{1:T})] + \ln p(o_{1:T}) \\
&\geq  KL[q(a_{1:T}, x_{1:T} | o_{1:T}) || p(x_{1:T}, a_{1:T} |  o_{1:T})] \numberthis
\end{align*}

Now, if we study the expression for the variational free energy $\mathcal{F}$ in some detail, we can see that it can be split up into three interpretable terms,
\begin{align*}
\label{AIVPG_F}
\mathcal{F}(o_{1:T}) &= \KL[q(a_{1:T}, x_{1:T} | o_{1:T}) || p(x_{1:T}, a_{1:T}, o_{1:T})] \\ 
&= -\underbrace{\mathbb{E}_{q(a_{1:T}, x_{1:T} | o_{1:T})}[\ln p(o_{1:T} | x_{1:T})]}_{\text{Reconstruction Error}} + \underbrace{\mathbb{E}_{q(a_{1:T} | x_{1:T})}\KL[q(x_{1:T} | o_{1:T}) || p(x_{1:T})]}_{\text{State Divergence}} \\ &+ \underbrace{\KL[q(a_{1:T} | x_{1:T}) || p(a_{1:T} | x_{1:T})]}_{\text{Action Divergence}} \numberthis
\end{align*}

If we apply the Markov assumption in the generative model, and assume that the variational posterior factorises across time such that $q(a_{1:T}, x_{1:T} | o_{1:T}) = \prod_{t=1}^T q(a_t | x_t)q(x_t | o_t)$, then we find that the previous derivation (Equation \ref{AIVPG_F}) in terms of full trajectories simplifies considerably into a sum of individual timesteps. Thus, we can write,
\begin{align*}
\label{AIVPG_full_F}
\mathcal{F}(o_{1:T}) &= \sum_{t=0}^T \mathcal{F}_t(o_t) = \sum_{t=0}^T  \KL[q(a_t, x_t | o_t) || p(x_t, a_t, o_t)] \\
&= \sum_{t=0}^T -\underbrace{\mathbb{E}_{q(a_t, x_t | o_t}[\ln p(o_t | x_t)]}_{\text{Reconstruction Error}} + \underbrace{\mathbb{E}_{q(a_t | x_t)}\KL[q(x_t | o_t) || p(x_t | x_{t-1}, a_{t-1})]}_{\text{State Divergence}}\\  &+ \underbrace{\KL[q(a_t | x_t) || p(a_t | x_t)]}_{\text{Action Divergence}} \numberthis
\end{align*}

Examining these terms, we can see several familiar objectives from the machine learning literature. For instance, the first reconstruction error term is simply just the log-likelihood of observations expected under the trajectory belief distribution. This term is commonly optimized in all sorts of machine learning tasks, of especial interest here is its use as part of the objective of the variational autoencoder \citep{kingma_auto-encoding_2013} If the likelihood term $p(o_t | x_t)$ can be thought of as the decoder of a variational autoencoder, then conversely the $q(x_t | o_t)$ term can be thought of as the encoder. Similarly, the state divergence term is often used as a regulariser or a method of training a transition model in model-based reinforcement learning. Indeed, we can see the transition dynamics term $p(x_t | x_{t-1}, a_{t-1})$ as encoding a direct model of the transition dynamics. Thus, it is straightforward to parameterise these distributions using deep neural networks. The $q(x_t | o_t)$ and $p(o_t | x_t)$ distributions are parametrised by the encoder and decoder of a variational autoencoder respectively, which can be trained through a reconstruction loss on the environmental observations. The $p(x_t | x_{t-1}, a_{t-1})$ distribution can be encoded as a deep neural network trained on the transition dynamics of the environment. With this, we turn to the two terms in the action divergence. The first, $q(a_t | x_t)$, can be thought of as a parametrized policy network, of the kind used in policy gradients or actor critic methods in reinforcement learning. Specifically, it can be thought of as a simple mapping between a state and the correct action to output from this state. So far the inference procedure we have written has no notion of rewards or goals. To achieve adaptive reward-sensitive action inference, this must be added somewhere. Following common practice in the active inference literature, we encode goals or rewards into the
the action prior by assuming that it is equal to the softmax of the expected free energies of future trajectories $p(a_t | x_t) = \sigma(\gamma \mathcal{G}(x_{t:T}, a_{t:T}, o_{t:T}))$ where $\mathcal{G}$ is the expected free energy functional from active inference, $\gamma$ is a precision parameter which controls the entropy or `temperature' of the softmax, and $\sigma(x) = \frac{exp(-x)}{\int dx exp(-x)}$ denotes the softmax function. Intuitively, we can consider this agent trying to optimize the sum over time of its expected free energy (and thus the extrinsic and intrinsic value components of the EFE), and then selecting actions with a probability proportional to the relative value of each choice. Such an action prior effectively implements a Boltzmann action distribution, which has been empirically studied in human and animal choice behaviours \citep{daw2006cortical}. The action divergence term them simply tries to minimize the divergence between the variational action policy $q(a_t | x_t)$ parametrised by a deep neural network, and the `ideal' action distribution $p(a_t | x_t)$.

The key difficulty, then, is the computation of the softmaxed expected free energy, as this is a path integral of the expected free energy of a trajectory into the future. Unlike in tabular active inference approaches, we cannot simply enumerate all possible future trajectories and evaluate them. Instead, we make use of a trick from deep reinforcement learning, called bootstrapping, which takes advantage of the recursive nature of the Bellman equation. Here, we first note that the expected free energy, since it can be simply written as a path integral through time, obeys a similar recursive relationship,
\begin{align*}
&\mathcal{G}(o_{t:T}, x_{t:T}) = \sum_t^T \mathcal{G}_t(o_t, x_t) \\
&\implies \mathcal{G}(o_{t:T}, x_{t:T}) = \mathcal{G}_t(o_t, x_t) + \mathbb{E}_{p(o_{t+1}, x_{t+1} | x_t, a_t)}[\mathcal{G}(o_{t+1:T}, x_{t+1:T})] \numberthis
\end{align*}
From here, we can expand the expected free energy term using its standard definition \citep{friston_active_2015} to obtain,
\begin{align*}
\mathcal{G}_t(o_t, x_t) &= \mathbb{E}_{q(o_t,x_t)}[\ln q(x_t) - \ln \tilde{p}(o_t, x_t)] \\
&\approx \mathbb{E}_{q(o_t,x_t)}[\ln q(x_t) - \ln q(x_t | o_t) - \ln \tilde{p}(o_t)] \\
&\approx -\mathbb{E}_{q(o_t,x_t)}[\ln \tilde{p}(o_t)] - \mathbb{E}_{q(o_t)}\KL[q(x_t | o_t) || q(x_t)] \\
&\approx -\underbrace{\mathbb{E}_{q(o_t,x_t)}[r(o_t)]}_{\text{Reward Maximization}} - \underbrace{\mathbb{E}_{q(o_t)}\KL[q(x_t | o_t) || q(x_t)]}_{\text{Information Gain}} \numberthis
\end{align*}

Here, we see that the expected free energy can be (approximately) decomposed into a reward maximization term and also an information gain term to be maximized. This information gain term, here between posterior and prior expectations over states, can be seen as an exploration-inducing `intrinsic reward' inherent to active inference agents, which furnishes them with greater \emph{directed} exploration capabilities than baseline reinforcement learning agents which lack this additional term and only focus on greedy reward maximization.

Putting this all together, we realize that we can express the path integral of the expected free energy recursively as,
\begin{align*}
\mathcal{G}(o_{t:T} x_{t:T}) = \mathbb{E}_{q(o_t, x_t)}[r(o_t)] - \mathbb{E}_{q(o_t)}\KL[q(x_t | o_t) || q(x_t)] + \mathbb{E}_{q(x_{t+1}, o_{t+1} | x_t, a_t)q(a_t | x_t)}[\mathcal{G}(o_{t+1:T}, x_{t+1:T})] \numberthis
\end{align*}

To efficiently approximate this, we can then utilize the bootstrapping method used in deep Q learning. Here we explicitly train a neural network to predict the expected free energy value function $\mathcal{G}(o_{t:T}, x_{t:T})$. While it may seem like this requires explicit computation of the path integral, to produce correct targets for the network, in fact it does not due to the recursive nature of the expected free energy equation. Let's denote the expected free energy value function predicted by the network as $\mathcal{G}_\phi(o_t,x_t)$ which takes as inputs only the current state and observation and uses it to predict the full path integral. The value network has parameters $\phi$. We can then approximate the true EFE recursive relationship,
\begin{align*}
\hat{\mathcal{G}}(o_{t:T} x_{t:T}) = \mathbb{E}_{q(o_t, x_t)}[r(o_t)] - \mathbb{E}_{q(o_t)}\KL[q(x_t | o_t) || q(x_t)] + \mathbb{E}_{q(x_{t+1}, o_{t+1} | x_t, a_t)q(a_t | x_t)}[\mathcal{G}_\phi(o_t, x_t)] \numberthis
\end{align*}
where we have replaced the recursive computation of the future expected free energies with the prediction from the value network, which thus defines the value `estimate' $\hat{\mathcal{G}}$. Then, to train the value network, we simply minimize the squared difference between the estimated EFE and the predicted EFE,
\begin{align*}
\label{AIVPG_valuenet}
\mathcal{L}_{value}(o_t, x_t) = || \hat{\mathcal{G}}(o_t, x_t) - \mathcal{G}_\phi(o_t, x_t) ||^2 \numberthis
\end{align*}
This method is referred to as bootstrapping because the real expected free energy path integral is never computed. Instead the targets are also constructed from the value network which are then used to train the value network. While this circular relationship may make it unintuitive that this method can work, we can understand why it does work empirically through the fact that at each step of the optimization, some local information about the EFE is fed into the network through the (true) computation of $\mathcal{G}_t(o_t, x_t)$. Thus over time, this local information builds up and allows the network to converge to the correct value function. Importantly, in Q learning, there are also several tricks that have been found to be necessary to ensure a stable convergence. Principally, using the same value network to update both the targets and the predictions simultaneously does lead to instabilities and possible divergence. To ameliorate this, we use a `frozen' copy of the value network to compute the targets $\mathcal{G}_\phi(o_t, x_t)$ and hold it fixed while the true value network is updated for $N$ steps. After the $N$ steps we replace the frozen value network with another frozen copy of the newly trained value network. In this way, the targets do not change rapidly over the course of optimization but nevertheless will slowly converge towards the correct targets. Empirically, as in Q-learning, we found the use of a target network a necessity for stable learning of the value function.  Given a trained value network which can output a prediction for the expected free energy value function, we can then compute the action prior $p(a_t | x_t)$ for any given state and observation. We have thus made concrete every single one of the distributions in the variational free energy functional $\mathcal{F}$, so that it can be explicitly evaluated. Once it can be evaluated, its gradients can be computed through automatic differentiation techniques, and we can train all the parameters of the agent jointly through stochastic gradient descent.

\subsubsection{Model}

At this point, to avoid losing sight of the wider picture, it is worthwhile to take a step back and look at the model as a whole. We propose to represent and train the variational posterior $q(x_t | o_t)$ and the observation likelihood $p(o_t | x_t)$ as a variational autoencoder. We additionally represent the dynamics distribution $p(x_t | x_{t-1}, a_{t-1})$ as a deep neural network transition model. We represent the variational action posterior $q(a_t | x_t)$ as a deep neural network `policy network' and can estimate the action prior $p(a_t | x_t)$ using the softmaxed predictions of an expected free energy value network $\mathcal{G}_\phi(o_t,x_t)$. All of these terms are combined according to Equation \ref{AIVPG_full_F} to compute the total variational free energy which forms the unified loss function of the model. Then, all the parameters of each network, except for the value network, are optimized according to a gradient descent on this total loss. The value network, by contrast, is trained using its own separate bootstrapping loss.

The full deep active inference algorithm is presented below,

\begin{algorithm}[H]
\SetAlgoLined
\DontPrintSemicolon
Initialize Observation Networks $Q_\theta(s|o), p_\theta(o|s)$ with parameters $\theta$. 
\BlankLine
Initialize State Transition Network $p_\phi(s|x_{t-1},a_{t-1})$ with parameters $\phi$ 
\BlankLine
Initialize policy network $Q_\xi(a|s)$ with parameters $\xi$
\BlankLine
 Initialize bootstrapped EFE-network $G_\psi(s,a)$ with parameters $\psi$ \\
Receive prior state $s_1$ \\
Take prior action $a_1$ \\
Receive initial observation $o_1$ \\
Receive initial reward $r_1$ \\
\While{$t < T$ }{
    $\hat{x_t} \leftarrow Q_\theta(s|o)(o_t)$
    $\widehat{x_{t+1}} \leftarrow p_\phi(s|x_{t-1},a_{t-1})(\hat{s})$
    $a_t \sim  Q_\xi(a|s)$
    Receive observation $o_{t+1}$
    Receive reward $r_{t+1}$
    $\widehat{G(s,a)} \leftarrow r_{t+1} + E_{Q(x_{t+1}}[log\widehat{x_{t+1}} - log\hat{x_{t+1}}] + E_{Q(x_{t+1},a_{t+1})}[G_\psi(x_{t+2},a_{t+2})]$
    $F \leftarrow E_{Q(s)}[logp(o|s)] + KL[\hat{x_{t+1}}||\widehat{x_{t+1}}] + E_{Q(s)}[\int da Q_\xi(a|s)\sigma(-\gamma G_\psi(s,a)(x_{t+1})) + H(Q_\xi(a|s))]$
    $\theta \leftarrow \theta + \alpha \frac{dF}{d\theta}$
    $\phi \leftarrow \phi + \alpha \frac{dF}{d\phi}$
    $L \leftarrow ||\widehat{G(s,a)} - G_\psi(s,a)||^2$
    $\psi \leftarrow \psi + \alpha \frac{dL}{d\psi}$
}
\caption{Deep Active Inference}
\end{algorithm}

While all the derivations are straightforwardly presented for a full POMDP model, in our experiments we only utilized simple MDP examples with observable state. This was because the main difficulty and contribution of this work is the action selection mechanism, and that the extension to POMDPs is straightforward by just training a separate variational autoencoder to estimate the latent state given the observations. Because of this the two terms $q(x_t | o_t)$ and $p(o_t | x_t)$ are superfluous and not computed. This leaves the transition model $p(x_t | x_{t-1}, a_{t-1})$, the policy network $q(a_t | x_t)$ and the value network $\mathcal{G}_\phi(x_t, o_t)$. The transition model is necessary to compute the exploratory information gain term in the expected free energy.

We represented the policy network, transition model, and value network each as two-layer fully connected neural networks with relu activation functions and a hidden size of 200 neurons. All networks, except the value network, were optimized by minimizing jointly the variational free energy, through stochastic gradient descent with the ADAM optimizer and a learning rate of 1e-4. The value network was trained on the bootstrapping objective (Equation \ref{AIVPG_valuenet}) with the same learning rate and optimizer. A memory buffer was used to store all the agent's experience which was replayed to the agent for training at random in minibatches of size 64. No preprocessing was done on the input data for any of the experiments. The models were implemented and automatic differentiation was performed in the Flux.jl machine learning framework \footnote{The code to reproduce all experiments can be found at: https://github.com/BerenMillidge/DeepActiveInference.}. We used a target network to stabilize the learning of the value network. The target network was copied from the value network every 50 episodes. Each episode consisted of a full iteration through the replay buffer. Additionally, as is common in reinforcement learning tasks, we utilized a temporal discount on the reward. Our temporal discount factor was 0.99 for all models.

\subsubsection{Results}
We compared the performance of the active inference agent to two strong model-free reinforcement learning baselines. Deep-Q learning \citep{mnih2013playing,mnih2015human}, and an actor-critic \citep{mnih2016asynchronous} architecture. Deep Q learning parametrises the Q function using a deep neural network with a similar bootstrapping objective to the EFE value function that we computed earlier, except only using reward. For the Q-learning agent we utilized the same hyperparameters and value network size that we used for the active inference agent, to enable a fair comparison. Moreover, we implemented the Q-learning agent with Boltzmann exploration \citep{cesa2017boltzmann}, which is very similar to the softmax function used to compute the action prior. 

We also compared the deep active inference agent to an actor-critic architecture. Unlike a Q-learning agent which computes actions directly by maxing over the Q-function, the actor-critic architecture maintains a separate `actor' or policy network which is trained on a policy gradient objective based on value function estimates learnt by the `critic' -- a value function estimator trained through bootstrapping. The policy network and value network of the actor-critic architectures were trained using the same hyperparameters and architecture as the active inference agent, to enable a fair comparison of the methods.

We compared the performance of the active inference agent on three continuous control environments from the OpenAI Gym \citep{brockman2016openai}. These were CartPole, Acrobot, and LunarLander. The cartpole environment is simple and requires the agent to balance a pole atop a cart by pushing the cart either to the left or to the right. The state-space of the cartpole environment is a four dimensional vector, comprising the cart position angle and velocity, as well as the angle and velocity of the pole. A reward of +1 is given for each timestep the episode does not end up to 200 steps. The episode will end early if the cart is more than 2.4 units from the centre (the cart has left the screen), or else the pole angle is more than 15 degrees from vertical (the pole has fallen down). The acrobot environment requires the agent to learn to swing up and balance a triple jointed pendulum. It has a state-space of 6 dimensions which represent the angles and velocities of the joints. The action space is a three dimensional vector corresponding to the force the agent wishes to exert on each joint. The reward schedule is -1 for every timestep the pendulum is above the horizontal, and 0 if it is above horizontal. The acrobot is a challenging task for exploration, since purely random actions are very unlikely to lead to any reward.  The lunarlander environment requires the agent to learn to land a simulated spacecraft on a surface within a target region in a Newtonian physics environment. It has a 8-dimensional state-space and a four-dimensional action space, with actions corresponding to fire left engine, fire right engine, fire upwards engine, and extend docking legs. The agent receives a reward of 100 for landing on the launchpad (located at (0,0)), and a -0.3 reward for every time step the rocket's engines are firing. It does not receive a penalty for simply doing nothing.

We compare the performance of the active inference, Q-learning, and actor-critic agent, in terms of pure reward obtained in each environment \footnote{Note this comparison based purely on reward actually penalizes the active inference agent to some degree, since it does not simply optimize the reward, but also by satisfying the epistemic drives furnished by the EFE objective function}. We ran for 15000 episodes over 20 random seeds for each agent and plotted mean rewards obtained below,

\begin{figure}[H]
    \centering
    \includegraphics[scale=0.4]{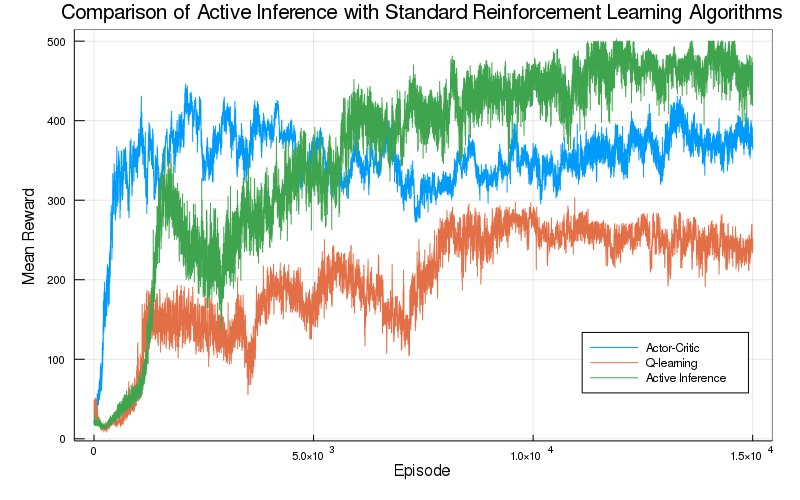}
    \caption{Comparison of the mean reward obtained by the Active Inference agent compared to two reinforcement learning baseline algorithms -- Actor-Critic and Q learning on the CartPole environment. We demonstrate the learning curves over 2000 episodes, averaged over 5 different seeds. 500 is the maximum possible reward. We see that while the vanilla actor critic agent initially learns faster, over a long time horizon, the active inference agent outperforms it -- and both perform better than the vanilla Q learning agent.}
    \label{CartPole Comparison}
\end{figure}

\begin{figure}[H]
\centering
    \includegraphics[scale=0.08]{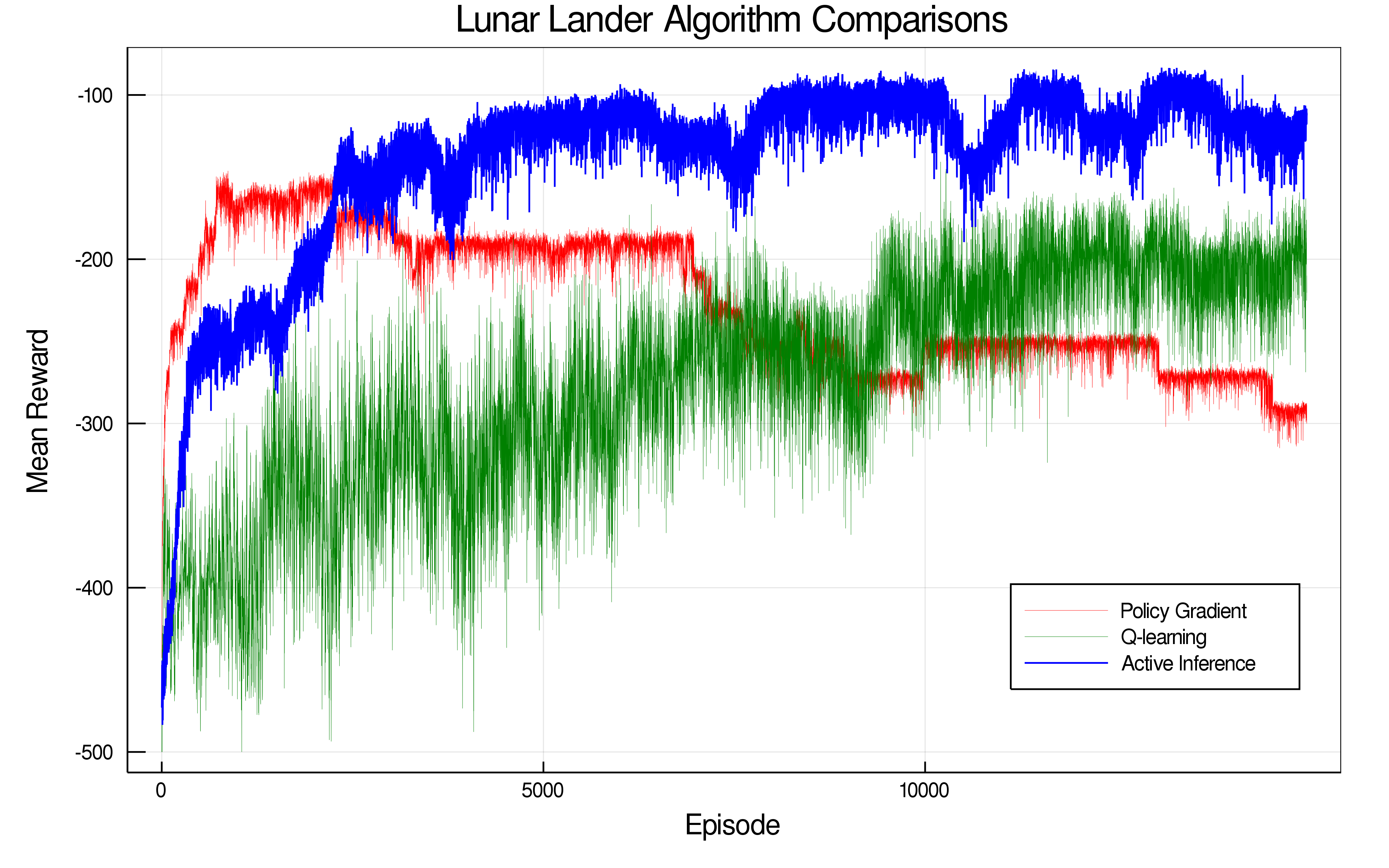}
    \caption{Comparison of Active Inference with standard reinforcement learning algorithms on the Acrobot environment. Here we see the learning curves plotted over five seeds over 20000 episodes. The maximum possible reward in this environment was 0, so no agents are optimal. We see again that active inference outperforms the other two methods consistently.}
\end{figure}
\bigskip
\begin{figure}[H]
    \centering
    \includegraphics[scale=0.08]{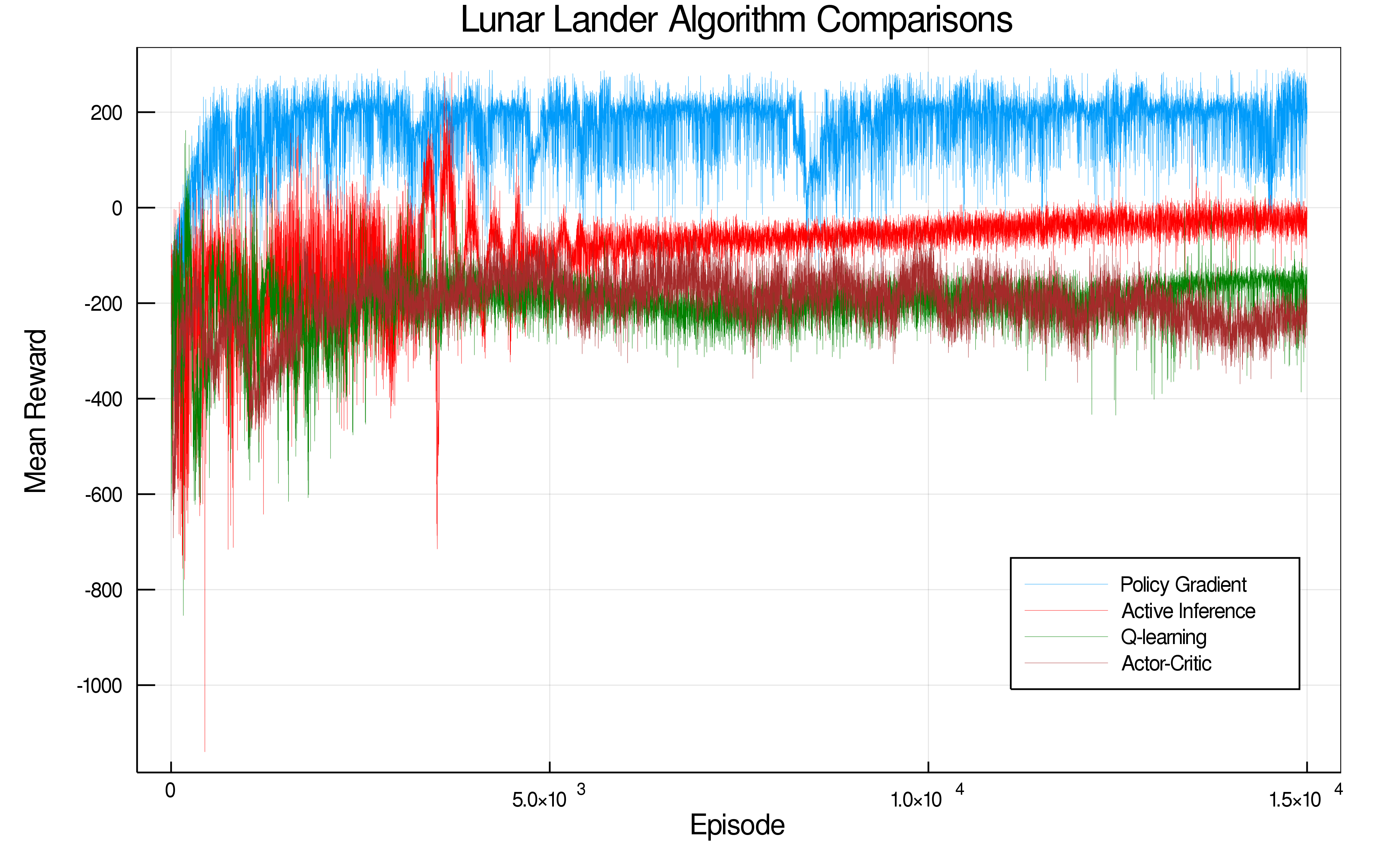}
    \caption{Comparison of Active Inference with reinforcement learning algorithms on the Lunar-Lander environment. Learning curves presented over 15000 episodes, averaged over 5 seeds. Here the vanilla policy gradient algorithm strongly outperforms the others, for unclear reasons, although active inference is still comparable with the other standard reinforcement learning algorithms. A score of 200 is optimal.}
\end{figure}

Since the active inference agent possesses several distinct features beyond the standard actor critic architectures, we performed an ablation study to understand whence its boost in performance arose.

\begin{figure}[H]
    \centering
    \includegraphics[scale=0.4]{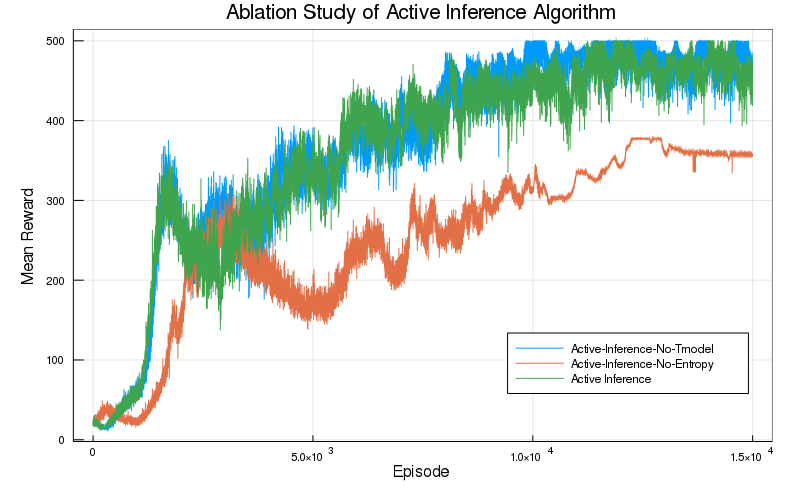}
    \caption{We compare the full Active Inference agent (entropy regularization + transition model) with an Active Inference agent without the transition model, and without both the entropy term and the transition model). We see that while removing the transition model appears to have little effect, removing the entropy regularisation term substantially impairs performance. This may be due to the entropy term aiding in staving off policy collapse.}
    \label{Active Inference Ablation}
\end{figure}

\begin{figure}[H]
    \centering
    \includegraphics[scale=0.4]{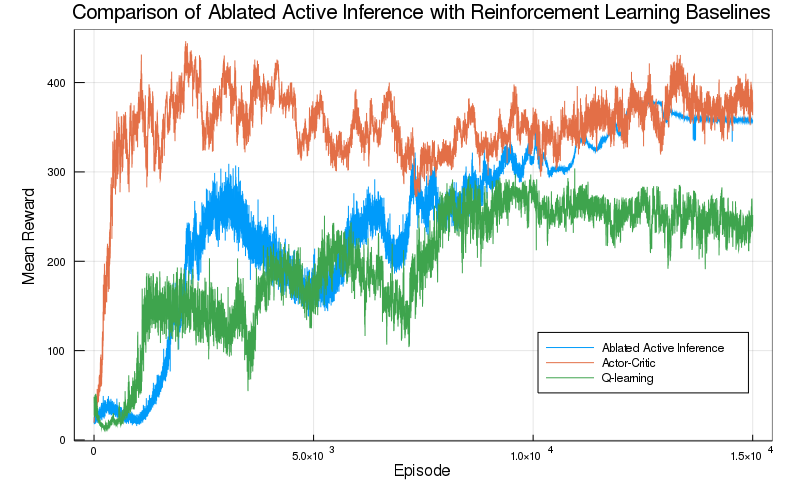}
    \caption{Comparison of the rewards obtained by the fully ablated Active Inference agent with standard reinforcement-learning baselines of Q-learning and Actor-Critic on the CartPole environment. Learning curves are averaged over 5 seeds. We see that despite being fully ablated, the active inference agent continous to perform comparably with standard reinforcement learning agents.}
    \label{Ablation Comparison}
\end{figure}

 We see that the key factor in the superior performance of the active inference agent is the additional action entropy term that is additionally optimized. This provides additional empirical confirmation to the success of control-as-inference approaches in the reinforcement literature which similarly utilize such an entropy regularisation term. Perhaps surprisingly, we found little effect of the epistemic terms in the EFE on total performance. We hypothesise that this was for two reasons. Firstly, the tasks that were tested possessed a relatively dense reward structure, sufficient to be learned by standard reinforcement learning agents utilizing only random exploration strategies, and thus that more advanced and powerful exploration strategies are likely unnecessary for such tasks. Secondly, the epistemic action term was the information gain between the prior and posterior states, which is effectively a measure of the predictive success of the transition model. Importantly, we found that the transition model very rapidly converged during training, much faster than the policy or value network, and thus that throughout most of the training period, the exploration term was thus negligible.

\subsubsection{Interim Discussion}

In this section, we have demonstrated how active inference approaches can be straightforwardly scaled up by parametrizing the likelihood, inference, and transition distributions with deep neural networks, and then additionally approximating the path integral of the expected free energy with an amortized value network. This is possible because the expected free energy satisfies a similar Bellman-like recursion to the reward in reinforcement learning, because it is factorizable across time into a sum of independent time-steps. Importantly, we have demonstrated that by taking this approach, our active inference agent can handle complex machine learning benchmark tasks just as well as several core deep reinforcement learning approaches, thus rendering it significantly more scalable than previous efforts in the literature which have generally been restricted to small, discrete, and straightforwardly enumerable state spaces.

Moreover, we have shown that our algorithm is competitive, and in some cases superior, to standard baseline reinforcement learning agents on a suite of reinforcement learning benchmark tasks from OpenAI Gym \citep{brockman2016openai}. While our active inference agent performed worse than direct policy gradients on the Lunar-Lander task, we believe this is due to the inaccuracy of the expected-free energy-value-function estimation network, since the policy gradient method used direct and unbiased monte-carlo samples of the reward rather than a bootstrapping estimator. Since the performance of Active Inference, at least in the current incarnation, is sensitive to the successful training of the EFE-network, we believe that improvements here could substantially aid performance. Moreover, it is also possible to forego or curtail the use of the bootstrapping estimator and use the generative model to directly estimate future states and the expected-free energy thereof, at the expense of greater computational cost. We take this approach of using the transition model to generate sample rollouts and using these to compute a Monte-Carlo estimate of the EFE path integral in the next section, where these estimates are used to inform model-predictive planning.

An additional advantage of our approach is that due to having the transition model, it is possible to predict future trajectories and rewards N steps into the future instead of just the next time-step. These trajectories can then be sampled from and used to reduce the variance of the bootstrapping estimator, which should work as long as the transition model is accurate. This $N$ could perhaps even be adaptively updated given the current accuracy of the transition model and the variance of the gradient updates. This is a way of controlling the bias-variance trade-off in the estimator, since the future samples should reduce bias while increasing the variance of the estimate, and also the computational cost for each update. 
\newline

Another important parameter in active inference is the precision \citep*{feldman2010attention,kanai2015cerebral}, which in the discrete-state-space paradigm corresponds to the inverse temperature parameter in the softmax and so controls the stochasticity of action selection \footnote{In the continuous predictive coding paradigm the precision modulates the `importance' of the prediction errors.}. In all simulations reported above we used a fixed precision of 1. However, in the discrete state-space case, the precision is often explicitly optimized against the variational free energy, and the same can be done in our deep active inference algorithm. In fact, the derivatives of the precision parameter can be computed automatically using automatic differentiation. Determining the impact of precision optimization on the performance of these algorithms is a potentially worthwhile avenue for future work. 

While we did not find that using the epistemic reward helped improve performance on our benchmarks, this could be due to the simplicity of the tasks we were trying to solve, for which random exploration is sufficient. In the next section, we demonstrate that the epistemic affordances engendered by the use of the EFE value function prove instrumental in attaining high performance in sparse-reward tasks.

The entropy regularization term which emerges directly from the mathematical formulation of active inference proved to be extremely important, and was often the factor causing the superior performance of our active inference agent to the reinforcement learning baselines. This entropy term is interesting, since it parallels similar developments in reinforcement learning, which have also found that adding an entropy term to the standard sum of discounted returns objective improves performance, policy stability and generalizability \citep*{haarnoja2017reinforcement,haarnoja2018acquiring}. This is of even more interest given that these algorithms can be derived from a similar variational framework which also casts control as inference \citep*{levine2018reinforcement}. Later (in Chapter 5), we discuss in significant detail how such paradigms relate to active inference. Additionally, many of the differences between active inference and the standard policy gradients algorithm -- such as the expectation over the action, and the entropy regularization term  -- have been independently proposed to improve policy gradient and actor critic methods \citep{fujimoto2018addressing}. The fact that these improvements fall naturally out of the active inference framework could suggest that there is deeper significance to the probabilistic inference formulation espoused by active inference. The other key difference between policy gradients and active inference is the optimization of the policy probabilities versus the log policy probabilities, and multiplying by the log of the probabilities of the estimated values, rather than the estimated values directly. It is currently unclear precisely how important these differences are to the performance of the algorithm, and their effect on the numerical stability or conditioning of the respective algorithms, and this is also an important avenue for future research. However, the comparable performance of active inference to actor-critic and policy gradient approaches in our results suggest that the effect of these differences may be minor.

\subsection{Model-based: Reinforcement Learning through Active Inference}

While the last section focused on scaling active inference using model-free reinforcement learning methods, here we focus on scaling active inference in a way inspired by model-based reinforcement learning methods.
Model-based reinforcement learning is perhaps a better fit for the central ideas in the classical active inference. Tabular active inference, after all, is a model-based algorithm which explicitly evaluates and optimizes future plans using explicit models of the environmental dynamics. Moreover, as tabular active inference explicitly replans at every time-step, it can be considered to be a model-predictive control algorithm. In fact, the explicit enumeration and evaluation of every possible policy can perhaps best be thought of as a truly exhaustive planning algorithm.

Similar to our previous approach described earlier, we propose to develop deep active inference methods which utilize deep neural networks to parametrize key distributions from the active inference framework. Specifically, we maintain deep neural network representations of the observation likelihood distribution $p(o_t | x_t)$ as well as the transition model which parametrises the dynamics model of the environment $p(x_t | x_{t-1}, a_{t-1})$. The major difference is how the action policy $q(a_t | x_t)$ is handled. In the previous model-free approach, this distribution was represented as an independent policy neural network which was trained against the action prior which represented the softmax of the expected free energy value function in an actor-critic like fashion. Here, we treat the action posterior as the output of a model-based planning algorithm.

Specifically, and quite elegantly, we can show that under certain conditions of the generative model of the future, that we can derive the optimal plan as a softmax over the expected free energy in the future, (which was merely assumed to be the action prior in the model-free case). Moreover, we then show that this path integral can be approximated by monte-carlo sampling in the form of a model-based planning algorithm which samples and evaluates given potential future trajectories using the transition and reward models possessed by the agent.

\begin{align*}
          \mathcal{\tilde{F}}_{\pi} &= \KL \Big( q(o_t, x_t, \theta , \pi ) \Vert \tilde{p}(o_t, x_t, \theta) \Big) \\
         &= \mathbb{E}_{q(o_t, x_t, \theta ,\pi)}[\log q(o_t, x_t, \theta |\pi) + \log q(\pi) - \log \tilde{p}(o_t, x_t, \theta,  \pi)] \\
         &=  \mathbb{E}_{q(\pi)} \Big[ \mathbb{E}_{q(o_t, x_t, \theta| \pi)}[ \log q(\pi) - [\log \tilde{p}(o_t, x_t, \theta) - \log q(o_t, x_t, \theta |\pi)]\Big] \\
         &= \mathbb{E}_{q(\pi)} \Big[\log q(\pi) - \mathbb{E}_{q(o_t, x_t, \theta| \pi)}[\log \tilde{p}(o_t, x_t, \theta) - \log q(o_t, x_t, \theta |\pi)]\Big] \\
         &= \mathbb{E}_{q(\pi)} \Big[\log q(\pi) - \big[-\mathbb{E}_{q(o_t, x_t, \theta| \pi)}[\log q(o_t, x_t, \theta |\pi) - \log \tilde{p}(o_t, x_t, \theta)]\big]\Big] \\
         &= \mathbb{E}_{q(\pi)} \Big[\log q(\pi) - \log e^{- \big[-\mathbb{E}_{q(o_t, x_t, \theta| \pi)}[\log q(o_t, x_t, \theta |\pi) - \log \tilde{p}(o_t, x_t, \theta)]\big]} \Big] \\
         &= \mathbb{E}_{q(\pi)} \Big[\log q(\pi) - \log e^{-\KL \big( q(o_t, x_t, \theta | \pi) \Vert \tilde{p}(o_t, x_t, \theta) \big)}\Big] \\
         &= \KL \Big( q(\pi)  \Vert  e^{- \KL \big( q(o_t, x_t, \theta | \pi) \Vert \tilde{p}(o_t, x_t, \theta) \big)} \Big) \\
         &= \KL \Big( q(\pi)  \Vert  e^{-\mathcal{\tilde{F}_{\pi}}} \Big) \numberthis
\end{align*}

It is important to note that here we do not use the expected free energy as our objective, unlike in standard active inference. Instead, we use the recently introduced objective: the free energy of the expected future (FEEF). This objective maintains the exploratory information gain terms of the traditional expected free energy while possessing a clear mathematical origin with strong intuitive grounding. Specifically, the FEEF can be defined as,

\begin{align*}
    \mathbb{FEEF} = \tilde{\mathcal{F}}_\pi &= \KL[q(o_t, x_t, \theta | \pi) || \tilde{p}(o_t, x_t, \theta) \\
    &\approx \underbrace{\mathbb{E}_q(x_t | \pi)\KL [q(o_t | x_t) || \tilde{p}(o_t)]}_{Extrinsic Value} - \underbrace{\mathbb{E}_{q(o_t; \theta)}\KL[q(x_t | o_t, \theta) || q(x_t)]}_{\text{State Information Gain}} \\ &- \underbrace{\KL [q(\theta | x_t) || q(\theta)]}_{\text{Parameter Information Gain}} \numberthis
    &
\end{align*}
Where we can see that the FEEF can be split into approximately three terms -- the likelihood divergence term which measures how much the expected observations diverge from the desired observations, and which effectively encodes reward or utility seeking behaviour, and an information gain term to be maximized which induces exploratory, uncertainty reducing behaviour. Since the FEEF objective includes both latent states $x$ and parameters $\theta$, we actually obtain two separate information gain terms, one for the states and one for the parameters. The core finding and argument of this part of the chapter, and an example of what the theory of active inference can bring to contemporary deep reinforcement learning, is that the exploratory information-seeking terms furnished by active inference objectives such as the expected free energy, or free energy of the expected future, by inducing \emph{purposeful} and \emph{goal-directed} exploratory behaviour, they can outperform traditional random exploration on a number of challenging reinforcement learning tasks, and their advantages become especially apparent in the case of sparse rewards where random exploration is often simply insufficient to find any good solutions in a reasonable time. Moreover, recent work in the literature, which utilizes exploratory objectives, but in a first exploratory, and then exploitatory phase, we argue that it is necessary to combine the two objectives to be jointly optimized. In this way the agent is furnished with a desire for \emph{goal directed exploration}. It is not rewarded simply for reducing any uncertainty, but only uncertainty that also exists in rewarding regions in the state-space. In this way, agents explore precisely only as much as needed, thus providing a step towards a practical solution to the exploration-exploitation tradeoff.

\subsubsection{Model} 

As in previous work, we extended active inference by using deep neural networks to parametrize key densities. Our model utilized a neural network transition model to model the distribution $p(x_t | x_{t-1, a_{t-1})})$. Since the FEEF objective requires the evaluation of an information gain term over the parameters (denoted $\theta$) of the transition model, we maintained an approximate distribution over the parameters $\theta$ using an ensemble of transition models with independently initialized parameters, and trained on different batches from the replay buffer. This ensemble approach has been found to be widely useful in model-based reinforcement learning and to offer a superior representation of the true posterior over the parameters than competing methods such as Bayesian neural networks. Importantly, utilizing an explicit ensemble of transition models allows the estimated posterior over the parameters to be multimodal, as opposed to the unimodal Gaussian assumption implicit in the Bayesian neural networks approach \citep{tran2018Bayesian,gal2016improving}. Moreover, an ensemble of models has been found empirically to help avoid overfitting in low-data regimes, which are also when the advantages of model-based reinforcement learning are most apparent.
Each element of the transition model ensemble was implemented as an independent neural network with two hidden layers with 400 neurons each. The networks used the swish activation function. The transition networks predicted the \emph{difference} in the next state \citep{shyam_model-based_2019} instead of the next state, as this has been found to help capture environmental dynamics more accurately in practice.
Since we are evaluating future simulated rollouts, we cannot simply rely on environmentally provided `true rewards'. While many methods in the literature assume the existence of a known reward function which can be queried even for counterfactual or simulated trajectories \citep{chua_deep_2018,hafner2018learning}, we do not, as such a reward oracle is unrealistic in many if not most situations. Instead, we learnt a reward model based upon previous interactions with the real environment, and then used the reward model to score proposed trajectories. The reward model was parametrised by a two layer multi-layer perceptron network with 400 units in the hidden layer and a relu activation function. The reward model was trained on a mean-square error loss between actually observed rewards for a given state, and the reward predicted by the reward model.
Importantly, the FEEF objective defines the extrinsic reward to be the KL divergence between the observation likelihood and the desired observation distribution. Since our model was situated purely in an MDP setting with fully observed state, the only observation was the reward, and thus the reward model doubled as the likelihood model. We set the desired reward observations to be a Gaussian distribution with a variance of 1 centred at the maximum possible reward for the environment. Since we interpret the predictions of the reward model as representing the mean of a Gaussian distribution, we can analytically calculate the KL divergence term. This allowed us to straightforwardly compute and optimize the reward maximization part of the FEEF objective.
Similarly, to evaluate the information gain terms of the FEEF objective, we can rewrite it in a more tractable way,

\begin{align*}
        & \E_{q(x_t | \theta)}\KL \big(q(\theta | x_t) \Vert q(\theta) \big) \\
        &= \E_{q(x_t | \theta)q(\theta | rs)}\big [ \log q(\theta | x_t) - \log q(\theta) \big] \\
        &= \E_{q(x_t, \theta)}\big [ \log q(x_t | \theta) + \log q(\theta) - \log q(x_t) - \log q(\theta) \big] \\
        &= \E_{q(x_t, \theta)}\big [ \log q(x_t | \theta)  - \log q(x_t) \big] \\
        &= \E_{q(\theta)q(x_t | \theta)}\big [\log q(x_t | \theta) \big]  -\E_{q(\theta)q(x_t|\theta)} \big[ \log \E_{q(\theta)} q(x_t | \theta) \big] \\
        &= -\E_{q(\theta)}\mathcal{H}\big[ q(x_t | \theta) \big] + \mathcal{H} \big[ \E_{q(\theta)}q(x_t | \theta) \big] \numberthis
\end{align*}

In effect, the parameter information gain term decomposes into an entropy of the average state minus the average of the entropy. The average of the entropy can be computed semi-analytically since each transition model ensemble models a Gaussian distribution, which has a known and analytically calculable entropy. Then, the average can simply be performed by directly averaging the entropies of each member of the ensemble together. The entropy of the average term is more difficult, since the average of many different Gaussian distributions is not necessarily Gaussian. As such we approximate it with a nearest neighbour entropy approximation \citep{mirchev_approximate_2018} which we found worked well in practice.

To train the model, we optimized the reward and transition models on data taken from a replay buffer. They were trained with stochastic gradient descent on their respective loss functions using a negative log likelihood loss. We cold-started the training of the agent at the end of every episode, as we found that this led to more consistent behaviour and performance. We initialized each episode with a dataset $\mathcal{D}$ taken from an agent with a random policy, to ensure some degree of transition and reward model accuracy before beginning with the model-based planning.

To compute actions we used a model-based planner (CEM) \citep{rubinstein1997optimization} within the model-predictive control paradigm. The CEM planning algorithm generates and evaluates the reward of a large number of action trajectories, then takes the mean and variance of some elite set of actions (usually the top 10) of the actions and restarts the evaluation using actions sampled from a Gaussian distribution with this mean and variance. For every timestep, a 30 step planning horizon was used resulting in a 30-step action plan, of which the first action was executed. In accordance with model-predictive control, we replan at each step. For the CEM algorithm, we used 700 candidate action sequences to be evaluated in each iteration, for 7 iterations. To train the transition and reward models we used the ADAM optimizer with a learning rate of 1e-4 and trained for 100 epochs.

\begin{algorithm}[H]
\label{algo:exps}
\SetAlgoLined
   \DontPrintSemicolon
   \textbf{Input:} Planning horizon $H$ | Optimisation iterations $I$ | Number of candidate policies $J$ | Current state $s_t$ | Likelihood $p(o_\tau|s_\tau)$ |  Transition distribution $p(s_\tau|s_{\tau-1}, \theta, \pi)$ | Parameter distribution $P(\theta)$ | Global prior $\tilde{p}(o_\tau)$
   \BlankLine
   Initialize factorized belief over action sequences $q(\pi) \leftarrow \mathcal{N}(0,\mathbb{I})$.
   
   \For{$\mathrm{optimisation \ iteration} \ i = 1...I$}{
      Sample $J$ candidate policies from $q(\pi)$ \;
      \For{$\mathrm{candidate \ policy} \ j = 1...J$}{
      $\pi^{(j)} \sim q(\pi)$ \;
      $-\mathcal{\tilde{F}}_{\pi}^j = 0$ \;
      \For{$\tau = t...t+H$}{
         $q(s_\tau| s_{\tau-1}, \theta, \pi^{(j)}) = \E_{q(s_{\tau-1}|\theta, \pi^{(j)})} \big[p(s_\tau|s_{\tau-1}, \theta, \pi^{(j)})\big]$ \;
         $q(o_\tau| s_\tau, \theta, \pi^{(j)}) = \E_{q(s_\tau|\theta, \pi^{(j)})} \big[p(o_\tau|s_\tau)\big]$ \;
         $-\mathcal{\tilde{F}}_{\pi}^j \leftarrow -\mathcal{\tilde{F}}_{\pi}^j + E_{q(s_\tau,\theta | \pi^{(j)})} \big[ \KL \big( q(o_\tau | s_\tau, \theta, \pi^{(j)}) \Vert \tilde{p}(o_\tau) \big) \big]  
          +\mathbf{H}[q(s_{\tau}| s_{\tau-1}, \theta, \pi^{(j)})] - \mathbb{E}_{q(\theta)}\big[\mathbf{H}[q(s_{\tau}| s_{\tau-1}, \pi^{(j)}, \theta)]\big]$
      }
   }
   $q(\pi) \leftarrow \mathrm{refit}(-\mathcal{\tilde{F}}_{\pi}^j)$
}
\textbf{return} $q(\pi)$
\caption{Inference of $q(\pi)$}
\end{algorithm}

\subsubsection{Results}

We tested the performance of our algorithm against strong model-free and model-based baselines on a number of challenging control tasks \footnote{Different tasks are utilized in this section compared to the previous one because here we are dealing with continuous actions while in the previous section we dealt only with learning discrete-action controllers.}. We utilized first the mountain-car task from OpenAI Gym, which requires the agent to steer a car on a 1D line to a goal. This task is difficult because the agent must first move the car away from the goal up a hill, to build up momentum to be able to get over the larger hill. This poses a difficult exploration problem which purely random exploration agents struggle to solve. By contrast, our agent can solve this task instantaneously, within only a single episode, due to its goal directed exploration.

\begin{figure}[h]
   \centering 
      \includegraphics[width=15cm]{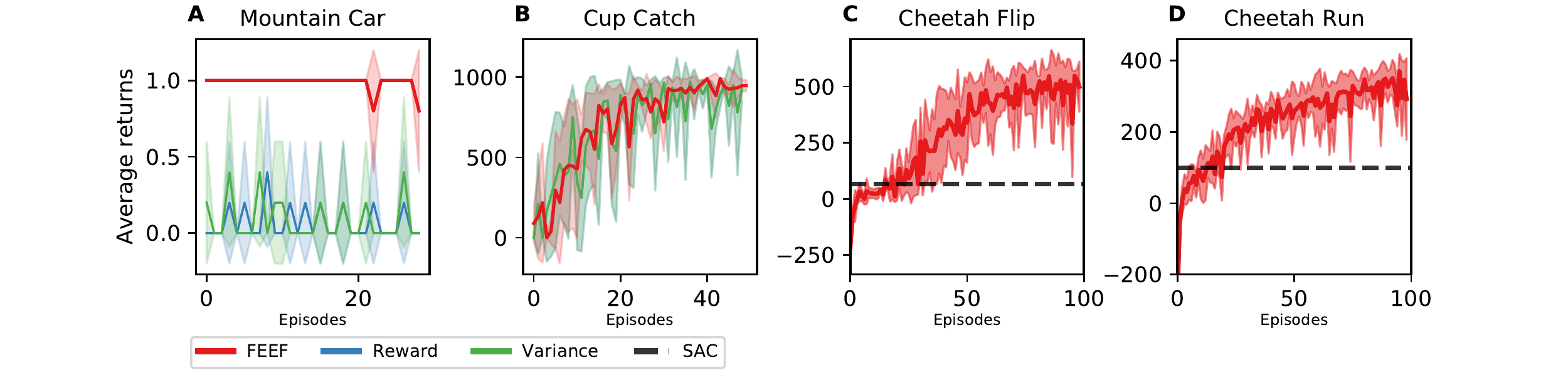}
      \caption{\textbf{(A) Mountain Car:} Average return after each episode on the sparse-reward Mountain Car task. Our algorithm achieves optimal performance in a single trial. \textbf{(B) Cup Catch:} Average return after each episode on the sparse-reward Cup Catch task. Here, results amongst algorithms are similar, with all agents reaching asymptotic performance in around 20 episodes. \textbf{(C \& D) Half Cheetah:} Average return after each episode on the well-shaped Half Cheetah environment, for the running and flipping tasks, respectively. We compare our results to the average performance of SAC after 100 episodes learning, demonstrating our algorithm can perform successfully in environments which do not require directed exploration. Each line is the mean of 5 seeds and filled regions show +/- standard deviation.}
   \end{figure}

Since the mountain car environment possesses only a two dimensional state space, we can explicitly plot and compare the degree of the space covered by the active inference agent vs the greedy reward maximizing reinforcement learning agent.

\begin{figure}[h]
   \centering 
   \includegraphics[width=12cm, height=4cm]{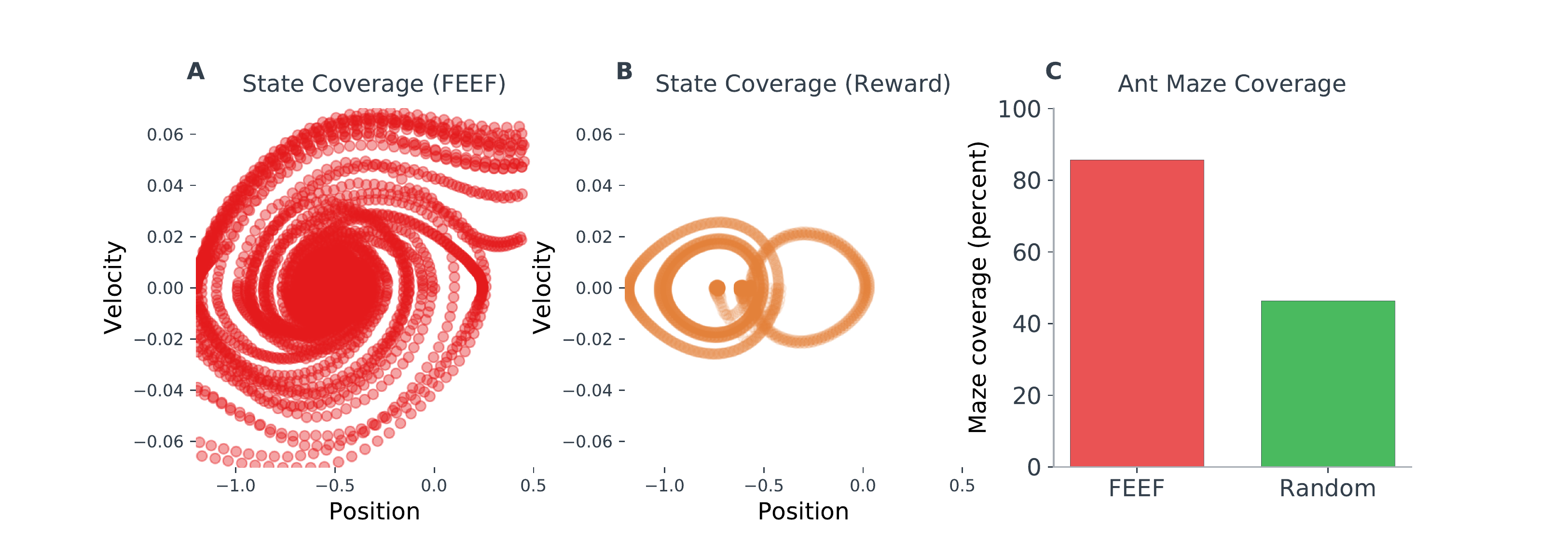}
\caption{\textbf{(A \& B) Mountain Car state space coverage}: We plot the points in state-space visited by two agents - one that minimizes the free energy of the expected future (FEEF) and one that maximises reward. The plots are from 20 episodes and show that the FEEF agent searches almost the entirety of state space, while the reward agent is confined to a region that can be reached with random actions. \textbf{(C) Ant Maze Coverage}: We plot the percentage of the maze covered after 35 episodes, comparing the FEEF agent to an agent acting randomly. These results are the average of 4 seeds.}
\end{figure}

We thus see that the exploratory drives inherent in the active inference agent propel it to explore a significantly larger fraction of the state-space than the reward maximizing agent, and it is this exploration which allows it to stumble upon the goal and thus rapidly learn to solve the task. By contrast, the random exploration reward-maximizing agent is unable to escape the local minimum at its start location by a purely random walk, since to do so requires too many correct moves in a sequence than can be generated within a reasonable time.
We also tested our agent on sparse reward versions of two challenging control tasks. In the first task, Cup Catch, the agent must actuate a cup to catch a ball thrown at it. The agent receives a reward of 1 if it catches the ball, and a reward of 0 if it does not. Similarly, in the Half-Cheetah environment, the agent takes control of the limbs of a running planar cheetah in a semi-realistic physics simulation. Its goal is to maximize the velocity at which the cheetah moves forward, ideally by running. We also experimented with a no-reward environment to test the pure exploration capabilities of our agent. For this we utilized the ant-maze environment in which the agent must actuate an ant-like robot in order to explore as much of a maze as possible. The agent receives no extrinsic rewards at any point in this task.

Overall, we see that the active inference and reward maximization agent performs similarly in the cup-catch environment. We hypothesise that this is because, even with the sparse reward, the reward is easy enough to obtain even with purely random behaviour. Similarly, on the half-cheetah benchmark, our model performs significantly better than the model-free SAC agent, due to the superior sample efficiency of model-based over model-free methods, however it also does not show much significant improvement compared to reward maximizing model-based baselines. However in the ant-maze environment, our agent evinces significantly more exploration capacity and explores a substantially larger proportion of the total state space than any other agent, once again demonstrating its superior exploratory capabilities.

\subsubsection{Interim Discussion}

So far, in this chapter, we have applied an active inference perspective to reinforcement learning and recast the traditional RL objective into a more active-inference-inspired one; reformulating the reward maximization objective as that as minimizing the divergence between predicted and desired probabilistic futures. From this starting point, we can derive a novel algorithm that exhibits high performance as well as robustness, flexibility, and sample efficiency in a number of environments that are known to be exceptionally challenging for traditional reinforcement learning methods, while also performing comparably in environments where standard RL methods do well.

We believe that through these two studies, we have convincingly demonstrated that active inference approaches can be successfully scaled to levels equal to contemporary reinforcement learning methods in both model-free and model-based paradigms. Moreover, we hope to have shown some ways in which the integration of active inference and reinforcement learning can provide novel and useful perspectives to inform and inspire work in the deep reinforcement learning community. We demonstrate that the exploration-inducing properties of active inference objective functionals such as the Expected Free Energy and the Free Energy of the Expected Future are highly beneficial especially in more challenging tasks with sparse or no rewards, while also performing comparably to pure reward maximization approaches on dense reward tasks that can be solved with purely random exploration. Moreover, combining both exploratory and reward maximizing terms in a single objective function and jointly optimizing them is crucial to derive algorithms which can simply learn to solve a task without separate exploratory and exploitatory phases as in much of the literature \citep{shyam_model-based_2019}, although the idea of inducing exploration by optimizing an epistemic term \citep{oudeyer2009intrinsic,schmidhuber2007simple,pathak2017curiosity} has been applied previously in reinforcement learning, and that can be competitive directly with both purely exploratory and purely exploitatory tasks in the regions where each of these methods excel. 

An additional idea, inspired by active inference, which could inform reinforcement learning perspective is the idea of representing preferences, instead of scalar reward values, as a distribution over observations. We believe that this modelling choice could enable greater flexibility in learning non-scalar, non-monotonic reward functions, as well as providing a natural, Bayesian framework for handling the case of unknown, uncertain, or nonstationary reward distributions. We believe that in many naturalistic settings, especially for biological organisms, rewards are not simply given a-priori by some known oracle, but are task-dependent, contingent, often highly uncertain, and nonstationary. Active inference provides a straightforward Bayesian account to handle precisely such conditions. Although in this work, and in much of the literature, we instead take extremely simplifying assumptions such as $\tilde{p}(o) = exp(-r(o))$ to make active inference as close as possible to reinforcement learning, future work should instead head in the opposite direction and try to deliberately explore the regions where active inference approaches offer greater flexibility than the traditional reinforcement learning paradigm, and thus demonstrate the advantages of active inference approaches there.

\subsection{Related Work}

Before our work in deep active inference there was a small amount of prior work which is important to review. The seminal paper which began this field is \emph{Deep Active Inference} \citep{ueltzhoffer_deep_2018}, which initiated the idea of using deep neural networks to approximate key densities within the active inference paradigm. This paper uses small neural networks to parametrise the transition dynamics and likelihood in active inference, and uses genetic algorithms to directly optimize a policy module of the expected free energy in a black-box fashion. This is because, under their problem setup, the expected free energy depends on the environmental dynamics and is thus nondifferentiable, assuming the environment itself is unknown. They produced a simple agent which can learn to solve the mountain car problem from OpenAI gym after many iterations. This paper was my inspiration to dive deeper into trying to understand the commonalities and differences between deep active inference and deep reinforcement learning.

Another piece of work, arising contemporaneously with my own initial work \citep{millidge2019combining,millidge_deep_2019}, was that by \citet{catal_Bayesian_2019}. They also parametrised the likelihood and transition dynamics using deep neural networks, and additionally explicitly utilized an expected free energy value function. However, instead of directly solving the sparse-reward challenge implicit in the mountain-car environment they tested their agent in, they instead constructed a hand-crafted `state-prior' generated from expert-rollouts which already directly solved the task, thus providing an effectively dense reward signal for this sparse reward problem.

Some other related work is that of \citet{cullen2018active} who also applied active inference to more complex non-toy environments. They trained an active inference agent to play a subset of DOOM -- the `take-cover' environment in OpenAI Gym. However, they still fundamentally utilised the discrete-state-space active inference formulation by discretising the continuous DOOM environment into 8 discrete states using the Harris Corner detection algorithm, and then applying discrete-state active inference onto the discrete states. 

Just after my initial work came similar work by \citep{tschantz_scaling_2019}, who instead applied active inference in a model-based fashion. Their model parametrises the transition dynamics using a deep neural network, and then uses model based planning (using the CEM algorithm) to optimize the expected free energy over time. We jointly extended their model in \citep{tschantz2020reinforcement} to investigate explicitly the exploratory effects of the EFE or FEEF objective, and whether such methods can be further scaled through learning the transition and reward model.

After our work, several recent approaches have scaled active inference further. \citet{ccatal2020learning}, situate deep active inference within a purely POMDP setting, using a VAE encoder and decoder to parametrize the likelihood and state-posterior mappings, and then explicitly compute an action search tree using their transition model to approximate the path integral over the expected free energy through time. Similarly, \citet{fountas2020deep}, also explicitly compute a model-based EFE search tree in their tasks, while simultaneously approximating the output of the action planner with a model-free `habitual' policy network.

\subsection{Iterative and Amortised Inference}

Now that we firmly understand the notion of implementing control as an inference procedure, it is worth recapping a fundamental distinction between two different \emph{types} of inference, which are important and implicit in the literature, but rarely well explained. The crucial distinction is between what we call \emph{iterative} and \emph{amortised} inference. Iterative inference is the kind that arises directly from a naive application of Bayes Rule, and was the standard inference approach used until the rise of deep learning very recently \citep{wainwright2008graphical,jordan1998introduction,beal2003variational}. Almost all `classical' variational or Bayesian inference methods are iterative. Amortised inference only became prominent with the advent of the variational autoencoder \citep{kingma_auto-encoding_2013}, but has since become the dominant approach, especially within machine learning. The key distinction is that iterative inference directly optimizes the \emph{parameters} of the variational distribution. For instance, suppose we assume our variational distribution is Gaussian, than iterative inference tries to optimize the means and variance $\phi = \{\mu, \sigma \}$ of this Gaussian to fit some datapoint. This is what is implied by the standard reading of the ELBO or variational free energy equation \citep{hinton1994autoencoders,beal2003variational}.
\begin{align*}
    Iterative = \underset{\phi}{argmax} \, \, \mathbb{E}_{q(x; \phi)}[\ln q(x;\phi) - \ln p(o,x)] \numberthis
\end{align*}

Amortised inference, by contrast, does not \emph{directly} optimize the parameters of the variational distribution. Rather, it \emph{learns} the parameters of a function that \emph{maps data to parameters of the variational distribution}. Effectively, the variational parameters themselves are never optimized directly, they are simply spit out of the amortised function $f$, which is then learned. $\phi = f_\psi(D)$. Rather, it is the parameters of this amortisation function $f_\psi$ that are learned. Importantly, these functions are optimized not just against a single data-point but across the whole dataset $DR$. Once learned, the amortisation function $f$ can be quickly used to compute estimated variational parameters $\hat{\phi}$ for \emph{any} data-point, thus \emph{amortising} the cost of inference across the whole dataset. By contrast, iterative inference must start from scratch from each individual data-point given and optimize the variational parameters afresh. While it is often written the same way, to make the notation very explicit, we write the amortisation objective to be optimized as,

\begin{align*}
    Amortised = \underset{\psi}{argmax} \, \, \mathbb{E}_{p(D)}\Big[ \mathbb{E}_{q(x; \hat{\phi} = f_\psi(D))}[\ln q(x; \hat{\phi} = f_\psi(D)) - \ln p(o,x)] \Big] \numberthis
\end{align*}

The reason that amortised inference has risen to such popularity and ubiquity lately is due to the fact that the amortisation function $f_\psi$ is straightforward to implement as a deep neural network, where $\psi$ are the neural network weights which can be trained by a gradient descent on the ELBO or variational free energy. For instance, in a variational autoencoder, $f_\psi$ is effectively implemented by the encoder, which maps the data directly to the variational parameters (the mean and variance of the Gaussian). The iterative approach, by contrast, would forego the encoder and run gradient descent directly on the mean and variance themselves for each data-points. We thus see why amortised methods are preferred. The amortised method can infer the mean and variance quickly, in one feedforward pass of the network, while gradient descent training is split over an entire dataset. By contrast, the iterative approach would require a gradient descent for every inference that the network wishes to make. Importantly, however, the variational parameters found by the amortised methods are in general worse estimates than those found by iterative inference. This is simply because the iterative inference method optimizes the parameters afresh with each data-point, while amortised inference must try to estimate them given a general function which must work for every datapoint. Thus the amortisation function $f_\psi$ must generalize in a way that iterative methods do not have to, and thus any generalization error will cause the amortised method to perform worse. This difference in performance is called the \emph{amortisation gap}. 

Interestingly, it has recently been shown \citep{marino2018iterative}, that you can gain performance improvements by \emph{combining} iterative and amortised inference together. For instance, in a variational autoencoder, if you first perform the amortised mapping to obtain initial estimates of the variational parameters $\phi$ but then run several iterative descent steps directly on your initial estimates of the parameters, this can improve the inference accuracy and reduce the amortisation gap compared to the pure amortisation approach with only a relatively small computational penalty for each inference.

Given that we know we can understand control problems in terms of inference, it is also interesting to consider whether the type of inference applied in control as inference can be best understood as iterative or amortised inference. Indeed, we argue that this distinction between iterative and amortised inference maps rather cleanly (although not perfectly) to the distinction between model-based and model-free reinforcement learning. Where model-free RL can be thought of as amortised inference and model-based as iterative inference. The reasoning here is straightforward but requires some subtly about what exactly is being inferred. 

The key quantity to be inferred in control as inference approaches is the variational distribution over actions $q(a | x)$. Model-free approaches which try to maintain a constant estimate of the value function, Q function or advantage function using the Bellman equation can be understood as amortised inference. This is most explicit in the case of actor-critic or policy gradient methods which explicitly maintain an amortised policy $q_\psi(a | s)$, which is implemented as a deep neural network with weights $\psi$ where the weights are not optimized separately for each data-point, but rather across all data-points. Approaches based purely on value function learning, such as Q learning, can also be expressed in such a manner, because here the optimal policy depends in a straightforward way upon the amortised value function. For standard deterministic Q learning we have that $q_\psi(a | s) = \delta(a - max_a Q_\psi(s,a))$, or that the action distribution is a dirac-delta over the maximum value of the Q function, which is itself amortised and implemented as a deep neural network. In soft methods, the delta-max is replaced by a softmax over all action values, so that actions are selected with probability proportional to their relative exponentiated magnitudes. 

Model-based methods, by contrast, appear to correspond to iterative inference approaches. The key to understanding this is that it is the planner which matters and is effectively doing the inference, not anything to do with the model -- i.e. the transition model -- in model-based methods, which  is often amortised. We can treat the varieties of planning algorithms such as CEM and path integral control as optimizing the actions or action sequences directly over the course of multiple iterations, and thus corresponds to iterative inference. Indeed \citet{okada_variational_2019} have shown how these standard planning algorithms can be derived as variational inference algorithms themselves under certain conditions.

To support these identifications intuitively model-free methods share the same advantages and disadvantages as amortised inference -- that they are trained across a whole dataset but fast to compute for any individual instance, and less sample efficient, since the amortisation function can only be learnt across a wide range of experience to enable good generalization. Model-based methods are the opposite and share the properties of iterative inference approaches. They are very sample efficient and perform well with very small amounts of data (since planning occurs for each datapoint independently, the only need for data is in the amortised training of the transition model). However, they are much more computationally expensive per datapoint, since they must undertake an iterative planning process for each state, instead of directly mapping a state to an action, as an amortised policy does. Interestingly, however, for model-free vs model-based approaches, the amortisation gap is often the other way around. Currently, model-free amortised policies generally achieve a \emph{higher} asymptotic accuracy than do model-based planners \citep{hafner2019dream,shyam_model-based_2019,haarnoja2018soft}. This is for two reasons. Firstly, there is an additional distinction which must be made between inferring a single action, as is done by model free policies, and inferring a whole sequence of actions (an action plan) which is what is typically done by model-based planners (although often this whole sequence is discarded and recomputed every time, an approach which is called model predictive control). Inferring a full plan is almost always harder than inferring a single action to take immediately, and this may be the cause of some of the reverse amortisation gap. An additional and potentially more serious issue is that current planning methods are generally quite crude and cannot represent expressive distributions over action plans. For instance the cross-entropy method can only represent single unimodal Gaussian plans, and similarly path integral control, the other state of the art method \citep{williams2017information,williams2017model,williams2018predictive,theodorou2010reinforcement,theodorou2012relative} suffers from similar constraints. While there has been some recent work in improving the expressivity of planning methods, such as multimodal CEM \citep{okada2020planet}, much work remains to be done here to be able to match the expressive capabilities of deep neural network policies.

Finally, it is important to note that the above distinction has revealed an additional orthogonal dimension of whether it is single actions that are inferred or whole action plans. We thus see that we can plot reinforcement learning and control methods into a quadrant with two orthogonal dimensions -- whether iterative or amortised inference is used, and whether full action plans or just single actions are inferred. We thus see that the standard distinctions of `model-free' vs `model-based' themselves map onto the diagonal of the quadrant. Model-free reinforcement learning is amortised inference of single actions, while the standard model-based methods correspond to iterative inference of full action plans. Importantly, there are several methods on the off diagonal, such as iLQR \citep{li2004iterative} which infers single actions in an iterative fashion. Understanding and plotting reinforcement learning methods in such a way reveals the full space of methods and which areas are potentially underexplored. For instance, we immediately see that there are very few, if any, methods which utilize amortised plans, even though learning amortised plans could well be straightforward and may even be beneficial. This would then be a fertile area for future work.

\begin{figure}
    \begin{center}
          \includegraphics[width=0.8\textwidth]{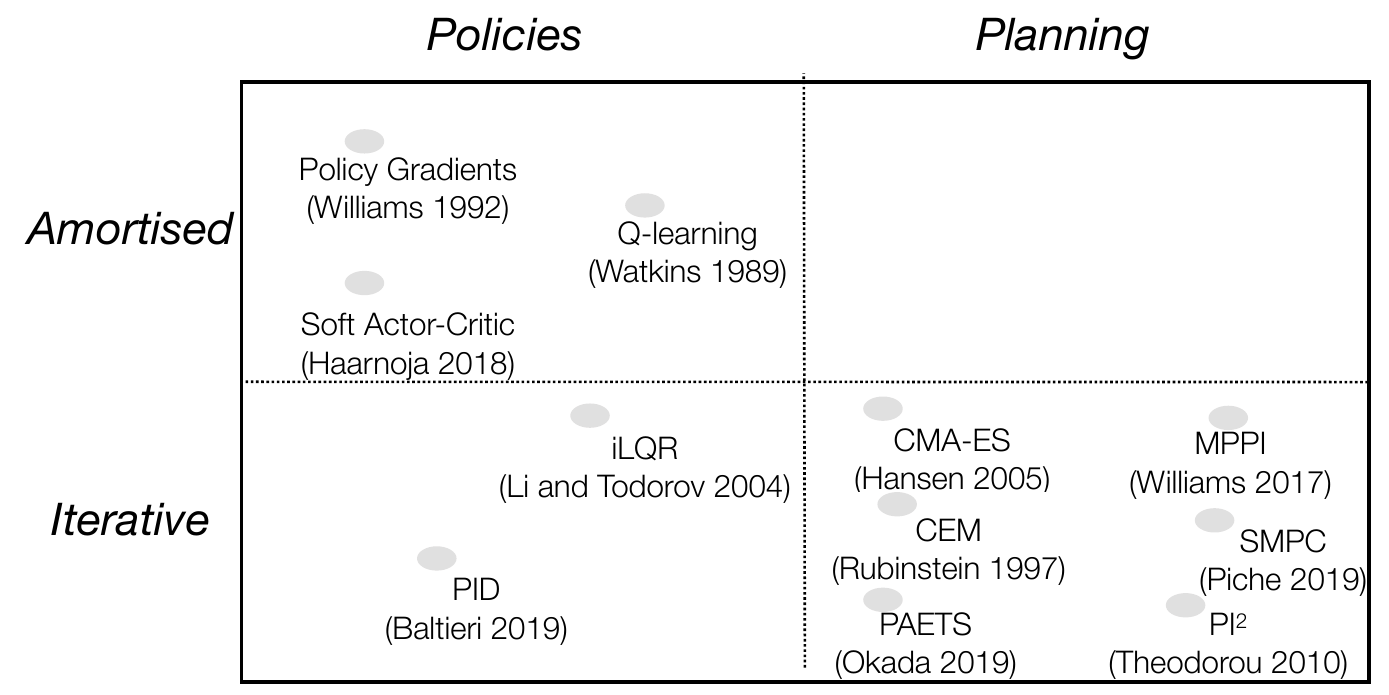}
    \end{center}
    \caption{Overview of classic RL and control algorithms in our scheme. Standard model-free RL corresponds to amortised policies, planning algorithms are iterative planning, and control theory infers iterative policies. The amortised plans quadrant is empty, perhaps suggesting room for novel algorithms. The position of the algorithms within the quadrant is not meaningful.}
    \label{fig:quad}
\end{figure}

\subsection{Control as Hybrid Inference}

\subsubsection{Introduction}
Building on the observation that iterative and amortised inference can be \emph{combined} to construct an iterative-amortised inference scheme which combines the advantages and ameliorates the respective disadvantages of both iterative and amortised inference -- enabling rapid and flexible inference with a high asymptotic performance, and an adaptive scheme which can leverage additional computing power only where it is most needed. In this section, we apply iterative-amortised combination to reinforcement learning which can be seen as combining model-based and model-free RL, using the identification previously developed.

Specifically, variants of model-based planning algorithms can be derived as variational inference algorithms using mirror descent to optimize the variational free energy in an iterative fashion \citep{okada_variational_2019}.
Additionally, as discussed previously, model-free reinforcement learning methods such as Q-learning, policy gradients, and actor-critic can be cast as optimizing another variational free energy bound, but rather this time in an amortised fashion. Given this, there are multiple potential ways to combine model-based and model-free reinforcement learning approaches. Perhaps one of the simplest approaches, which we apply in this study, is to use the model-free policy as an initialization of the model-based planner. Model-based planning algorithms such as CEM or MPPI, require an initial action distribution $p_1(a_{1:T}) = \prod_{t=0}^T p_1(a_t)$ to begin with, which they they proceed to optimize. Usually this initial action distribution is set to some simple known distribution such as a zero-centred normal distribution $p_1(a_t) = \mathcal{N}(a; 0, \sigma_a)$ with a variance parameter $\sigma_a$ which becomes a hyperparameter of the planning algorithm.

Instead, we propose to initialize the planner with the results of the amortised model-free policy network $p_1(a_t) = q_\phi(x_t)$. To do this for a potential action trajectory requires knowledge of future states to feed into the model-free policy network. However, conveniently, the model-based planner also has access to a transition model which is used to generate these simulated state trajectories, given the actions output by the policy network. 

\subsubsection{Model and Hyperparameter Details}

To make our model concrete, we need to instantiate many distributions such as the transition models $p_\theta(x_{t+1} | x_t, a_t)$, the parameterised policy $q_\phi(a_{t:T} | x_{t:T})$ and the variational iterative planning algorithm which instantiates $q(a_{t:T} | x_t; \psi)$.

The transition model was instantiated as an ensemble of three layer multi-layer perceptron networks with a hidden size dimension of size 250, which was trained to output a Gaussian distribution (mean and variance) over the \emph{change in environment state}. That is, rather than explicitly model $p(x_{t+1} |x_t, a_t)$, we instead modelled $p(x_{t+1} - x_t | x_t, (x_t - x_{t-1}), a_t)$, which we could then use to reconstruct the next predicted environmental state as $\hat{s}_{t+1} = x_t + (x_{t+1} - x_t)$. Training the transition model to predict state differences instead of the states directly is a common trick used in model-based reinforcement learning which has been found to significantly improve modelling performance by incorporating explicit information about the derivatives of the states, which is hard to derive solely from the states themselves. To obtain a measure of uncertainty over the transition model parameters $\theta$, which can be utilized to drive information-gain maximizing exploration, we maintained an ensemble of 5 transition models which were each trained on independently sampled batches of transition data. Each ensemble possessed independent randomly and uniformly initialized weights.

For the amortised action policy, we utilized the soft-actor-critic architecture (SAC) \citep{haarnoja2018soft}, with a policy-network which consisted of a three-layer MLP model with a hidden dimension of 256. We did not use an adaptive $\alpha$ parameter for the SAC agent but set it to 0.2 throughout.

For the iterative planner, we used the standard and powerful CEM algorithm \citep{de2005tutorial}, with a time-horizon of 7, a number of iterations of 10, and a trajectory sample size of 500. For each generated trajectory, to encourage more exploration, we added additional action noise sampled independently from $\sim \mathcal{N}(0, 0.3)$.

We maintained a memory buffer of all environmental interactions seen by the agent, and used various samples to train the transition model and amortised policy network. We trained the model over 10 epochs which iterated over the full memory buffer, with a batch size of 50. The full control as hybrid inference algorithm is defined as follows,
\newline 

\begin{algorithm}[H]
  \label{algo:chi}
  \SetAlgoLined
     \DontPrintSemicolon
     \textbf{Input:} Planning horizon $H$ | Optimisation iterations $I$ | Number of samples $K$ | Current state $s_t$ |  Transition distribution $p_{\lambda}(s_{t+1}|s_t, a_t)$ | Amortisation function $f_{\phi}(\cdot)$
     \BlankLine
     \textbf{Amortised Inference}: \\
    $p_{\phi}(\tau) = \delta(s_t) \prod_{t'=t}^T p_{\lambda}(s_{t'+1}|s_{t'}, a_{t'}) q_{\phi}(a_{t'}|s_{t'}; \theta)$ \\
     Extract $\theta^{(1)} = \{\mu_{t:T}, \sigma_{t:T}^2 \}$ from $p_{\phi}(\tau) $ \\
     Initialise $q(a; \theta)$ with parameters $\theta^{(1)}$
     \BlankLine
     \textbf{Iterative Inference}: \\
     \For{$\mathrm{optimisation \ iteration} \ i = 1...I$}{
        Sample $K$ action sequences $\{(a)_k \sim q(a; \theta)\}^K_{k=1}$ \\
        Initialise particle weights $\mathbf{W}^{(i)} := \{w_{k}^{(i)}\}^{K}_{k=1}$ \\
        \For{$\mathrm{action \ sequence} \ k = 1...K$}{
         $ w_k^{(i+1)} \leftarrow \frac{\mathcal{W}\big((a)_k\big) \cdot q^{(i)}\big((a)_k; \theta \big)}{\sum_{j = 1}^K \Big[\mathcal{W}\big((a)_j\big)\ \cdot q^{(i)}\big((a)_j; \theta \big)\Big]}$ \\
        $\theta^{(i+1)} \leftarrow \text{refit}\big(\mathbf{W}^{(i+1}\big)$
     }
  }
  \BlankLine
  Extract $\mu_{t:T}$ from $q(a; \theta)$ \\
  \textbf{return} $\mu_t$
  \caption{Inferring actions via CHI}
\end{algorithm}

\subsubsection{Related Work}

There has been a small amount of prior work aiming at combining model-free and model-based 
\citep{li2020robot,che2018combining}. For instance, a strand of research has focused on using a learned transition model to generate additional simulated data which can then be used to train a model-free policy `offline'. This approach was pioneered with the Dyna architecture \citep{sutton1991dyna}, but has also been extended and applied in more modern deep reinforcement learning settings \citep{gu2016continuous}. Conversely, in \citep{farshidian2014learning} and \citep{nagabandi2018neural} a model-based planner was used to initialize a model-free policy -- the opposite direction to our model. Our approach does share similarities with the approach used in AlphaGo \citep{silver2017mastering} which used learned amortized policy networks to generate proposals for the monte-carlo-tree-search (MCTS) used to select moves in that approach. However, their approach was justified on heuristic grounds and they did not consider how their approach corresponds to a mathematically principled combination of iterative and amortised variational inference. Indeed, it is not yet clear if the MCTS algorithm can be cast as performing some kind of variational inference or not.

While we are the first to consider the combination of amortised and iterative inference in reinforcement learning, and to make the connection to model-based and model-free methods, there is a line of work combining the two approaches to inference in the context of unsupervised generative modelling, typically using variational autoencoders. \citep{kim_emi:_2018}, employ amortised inference in a VAE to initialize the set of variational parameters which are then optimized directly against the ELBO. A similar approach was taken by Marino \citep{marino2018iterative}, who showed that by repeatedly encoding the gradients and optimizing the variational parameters against the ELBO, which was found empirically to improve performance and help narrow the amortisation gap.

\subsubsection{Results}

We tested our hybrid agent first on a didactic toy continuous control task. The goal of this task was to simply explore how the iterative and amortised control schemes interact. The environment was a simple 2-D planar environment, where the agent began in the bottom-left corner, and where the goal was to arrive in the top-right corner. The reward signal was a smooth gradient field leading to the top-right which was implemented as $r(x,y) = 1 - (|| (x,y) - (g_x, g_y) ||^2)$ where $g_x$ and $g_y$ represent the x and y coordinates of the goal state. In the centre of the environment there was an impassable wall except for a small opening through which the agent could pass. The agent could control its x and y velocity -- $a = (\dot{x}, \dot{y})$ with a maximum velocity of 0.05 and a minimum velocity of -0.05.

\begin{figure*}[t!]
  \centering
  \includegraphics[width=\textwidth]{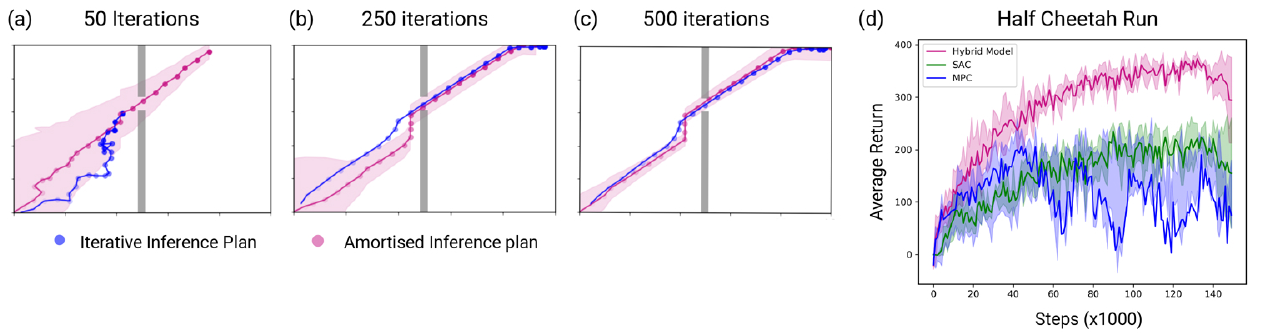}
  \caption{\textbf{(a - c)}: Amortised predictions of $q_{\phi}(a|s; \theta)$ are shown in red, where $\bullet$ denote the expected states, shaded areas denote the predicted actions variance at each step, and the expected trajectory recovered by iterative inference is shown in blue.
  At the onset of learning (\textbf{a}), the amortised predictions are highly uncertain, and thus have little influence on the final approximate posterior. As the amortised model $f_{\phi}(\cdot)$ learns (\textbf{b}), the certainty of the amortised predictions increase, such that the final posterior remains closer to the initial amortised guess. At convergence, (\textbf{c}), the iterative phase of inference has negligible influence on the final distribution, suggesting convergence to a model-free algorithm. \textbf{(d)} Here, we compare our algorithm to its constituent components -- the soft-actor critic (SAC) and an MPC algorithm based on the cross-entropy method (CEM). These results demonstrate that the hybrid model significantly outperforms both of these methods.}
\label{hybrid_performance_figure}
\end{figure*}

The graph shows the evolution of the agent's iterative and amortised policies as it learns to complete the task. As can be seen, the iterative policy starts out highly uncertain, with a high variance. As the amortised policy is slowly learnt, the variance of the iterative policy shrinks, and the resulting policy closely matches the amortised policies. This immediately suggests an adaptive method of saving computation -- when the variance of the iterative policy is small, or the iterative policy is very close to the amortised policy, rely solely on the computationally cheap amortised policy only. Conversely, when the amortised policy or the iterative policy is highly uncertain (as at the beginning of training), then the computationally expensive model-predictive-control of the iterative policy should be utilized. In this way, the agent can attain the impressive sample efficiency and rapid performance of model-based planning at the beginning of training, when the amortised policy is poor, but then once the amortised policy becomes good, the agent can simply rely on that and thus achieve the high asymptotic performance and relative computational cheapness of model-free RL.

We also compared the hybrid agent on a challenging continuous control task -- HalfCheetah Run $\mathbb{S}^{17}, \mathbb{A}^6$. This environment requires the agent to take control of a bipedal simulated cheetah in a planar environment with semi-realistic physics. The agent's goal is to move the cheetah's limbs in such a way as to maximize the overall velocity of the cheetah, while simultaneously minimizing the total action applied. The reward function for the task was $r = v - 0.1a^2$ where $v$ denotes the overall velocity of the cheetah. The hybrid agent was evaluated against strong model-free (SAC) and model-based (CEM) baselines. As can be seen from Figure \ref{hybrid_performance_figure} the hybrid agent significantly outperforms both baselines and simultaneously achieves the sample efficiency of model-based methods with superior performance to the model-free SAC agent.

\subsubsection{Interim Discussion}

Empirically, we find that the hybrid agent performs well, and that the interaction of the iterative and amortised inference components allow for a natural adaptive scheme to switch between and apportion computation in a way that maximizes the computational resources available to the agent. Moreover, the use of an amortised policy to initialize the iterative planner cuts significantly down on the computational expense of the planner and tends to stabilize performance. Additionally, the use of the iterative planner at the start means that the agent rapidly discovers highly rewarding trajectories which are then used to train the SAC agent, and thus provides a powerful source of implicit exploration for the model-free SAC agent. Interestingly, however, this highly rewarding data generated by the iterative planner comes with a cost -- it is heavily biased towards positive trajectories, and thus the SAC agent, as it is not exposed directly to negative trajectories in the real world, simply does not learn them and thus learns a highly optimistic value function, which performs poorly when interacting with the real environment. We call this the \emph{data-bias} issue and left unchecked it inhibited the performance and learning of the algorithm.

To ameliorate the data-bias issue, we instead train the SAC agent from the simulated rollouts of the iterative planner. These rollouts, especially in the early stages of iteration, contain many examples of (predicted) negative trajectories, which thus helps render the dataset fed to the model-free SAC agent less positively biased. These rollouts do have their own difficulties -- namely that they are fundamentally from a simulation and thus may be a poor representation of the actual dynamics of the world (especially when the transition model is poor). Secondly, the rollouts are still biased to some degree by the operation of the iterative planner even in the early stages of iteration. Additionally, this becomes more acute as the model-free policy becomes better, as it learns to avoid negative contingencies and its action predictions are then used to initialize the planner, thus creating a compounding positive bias to the data that is fed into the SAC agent. Nevertheless, we find empirically that this solution suffices to train a high performance model-free policy network.

On a more theoretical level, it is important to note that we chose a relatively straightforward scheme of combination -- using the amortised policy to simply initialize the model-based planner. A more involved, but slightly more principled method may be to set the action prior of the iterative planner (which is currently assumed uniform, as is standard in the control as inference framework) to the amortised policy $p(a | s) = q_\phi(a | s)$. Using this method, instead of a direct initialization, the amortised policy would serve as a regularizer on the iterative model-based planner, ensuring that the resulting iterative policy is penalized for its divergence from the model-free amortised policy. Such regularisation methods have been found to be beneficial for the stability of learning in a number of reinforcement learning algorithms and especially in policy gradient methods, where methods such as PPO \citep{schulman2017proximal} reach state of the art performance.

\section{Conclusion}

In this chapter, we have investigated the application of methods derived from the free energy principle -- specifically \emph{active inference} -- to the general problem of optimal action selection and control. We have focused especially on a core limitation of current active inference methods: their scalability. We have demonstrated how many of the key distributions which arise out of the free energy objective can be parametrized using deep neural networks, to derive schemes which can look very similar to contemporary deep reinforcement methods -- both model-free and model-based methods. We show that these methods -- which we call \emph{deep} active inference approaches -- can perform comparably and often better to their deep reinforcement learning counterparts. 

Specifically, in the first study presented in this chapter, we showcase how active inference can be interpreted through the lens of model-free reinforcement learning. In this case, we use a learnt action policy $q(a | s)$ and parametrize the action prior using an amortized expected-free energy value network, to approximate the required path integral over the expected free energy. The resulting algorithm looks very similar to actor-critic methods in model-free reinforcement learning, but using the expected free energy instead of the reward. Additionally, we find that this approach also utilizes additional entropy regularisation terms which can be shown to substantially improve the stability and the performance of the resulting algorithm -- thus demonstrating how insights and the mathematical formalism of active inference can also lead to improvements in reinforcement learning algorithms.

In the second study, we instead approximate the path integral of the expected free energy, with monte-carlo sampling of trajectories and ultimately use a model-predictive control planning algorithm to compute optimal trajectories instead of an amortised policy network. This simple change is sufficient to move us into the realm of model-based reinforcement learning. Here, we show that active inference can again attain the performance of comparable model-based deep reinforcement learning algorithms, and can be applied to solve challenging continuous control tasks. Additionally, here the superior exploratory capabilities of the expected-free energy functional come into play, and we see that they allow the construction of powerfully exploratory \emph{goal-directed}, \emph{information-seeking} behaviours, which can solve very challenging sparse reward tasks, such as the mountain car, with ease. This demonstrates another way in which insights from active inference can aid the development of deep reinforcement learning algorithms.

We then turn to a more abstract consideration of the difference between model-based and model-free reinforcement learning in terms of inference -- an insight which is heavily enabled by the active inference formulation of action selection as fundamentally an inference problem. We demonstrate that we can see the distinction between model-free and model-based as simply that of iterative vs amortized inference, where iterative variational inference directly optimizes the parameters of the variational distribution, while amortized inference instead optimizes the parameters of a mapping function which maps observations directly to variational parameters. We then show how there is a separate dichotomy between whether policies or plans are inferred, and that this provides us with a simple two dimensional quadrant scheme upon which to place all major reinforcement learning algorithms. It also demonstrates that there are several areas which are underexplored in the literature -- especially the direct computation of amortized plans.

Finally, we use this insight into the nature of model-based and model-free reinforcement learning in terms of iterative and amortized inference to ask how these two approaches can be \emph{combined}. We show that this can yield powerful algorithms which can combine both the sample-efficiency and rapid learning of model-based planning with the asymptotic performance and computational cheapness of model-free reinforcement learning. Importantly, investigations into this field of combined or hybrid reinforcement learning algorithms are only just beginning, and there are many design choices left to be extensively investigated in future work.

Overall, crucially, we have shown that the free energy principle and active inference can be successfully applied and scaled up to handle large and challenging control tasks and to create algorithms which perform comparably with state of the art methods in deep reinforcement learning.

In the next chapter, we extend the intuitions provided here about the importance of combining exploration and exploitation and turn to a more abstract and mathematical analysis of what kind of mathematical procedure gives rise to the combination of exploratory and epistemic action that characterise such objective functionals as the Expected Free Energy and the Free Energy of the Expected Future. 

%% file: chap5.tex
\chapter{The Mathematical Origins of Exploration}

\section{Introduction} 

In the previous chapter, we have seen the importance and benefits of \emph{information-seeking} as opposed to \emph{random} exploration for reinforcement learning tasks. Information-seeking exploration, which explicitly aims to reduce uncertainty about either the environment or the agent's model of the environment, provides a powerful exploration strategy that allows the rapid and efficient exploration of an environment, as opposed to the random walk strategy employed by random exploration. Moreover, when combined in a single loss function with a reward maximizing term, this combination results in \emph{goal-oriented exploration} where the agent is only driven to explore contingencies which are also likely to lead to high reward. We have seen that this goal-directed, or goal-oriented, exploration mechanism has performed well in model-based reinforcement learning tasks including sparse-reward environments which are challenging for standard reinforcement learning agents. Moreover, this kind of exploration is almost certainly necessary for biological organisms in more ecologically valid tasks, where rewards are often very sparse and environments are typically very large compared to those in mainstream reinforcement learning benchmarks.

In this chapter, we take a more abstract perspective, and study in depth the question of the \emph{mathematical origin and meaning} of such goal-directed exploration objectives which unite both reward seeking and information maximizing terms in a single objective. Specifically, we seek to understand whence they arise, and what the mathematical formulation which can give rise to them is. While for practical purposes and engineering applications it is often sufficient to glue different terms together in an ad-hoc way to construct an objective which gives rise to some desired behaviour, we wish to probe the deeper theory underlying such functionals which has so far remained mostly mysterious. It is the hope that by mathematically understanding the origin and nature of such objectives, as well as their properties, we can illuminate a swathe of current methods in reinforcement learning, cognitive sciences, decision theory, and behavioural economics, as well as deeply understanding how they interrelate to one another. Moreover, it seems likely, given the generally productive dialectic between theory and practice in all of these fields, that by contributing to the underlying theory of such objectives, we can ultimately contribute to the design of more powerful objectives and methods than are currently used in the literature.

To begin, we wish to deeply examine the origin and nature of the \emph{Expected Free Energy} (EFE) functional. The EFE is central to the theory of active inference, where it is proposed that all agents under the free energy principle, which must seek to minimize the long term path integral of their surprise must choose policies that minimize the EFE. The EFE has been widely used in almost all models in discrete-time active inference \citep{friston_active_2015,friston2017active,friston2018deep,friston2017process,da2020active} with the exception of the later development of the generalized free energy \citep{friston2015active,parr2017uncertainty,parr2017active}.
  Despite this ubiquitous use within the active inference community, the precise mathematical origin and nature of the EFE functional have remained unclear. In the literature, the EFE is often justified through a \emph{reductio-ad-absurdum} argument \citep{friston2015active} which runs as follows -- since (we assume under the FEP) all agents minimize free energy, then they must think they will minimize free energy in the future. Since the future is uncertain, instead of the standard variational free energy (VFE), they must instead minimize their \emph{expected} free energy (EFE), else they are not a free-energy minimizing agent (disproven conclusion) \footnote{Technically this is more of an induction argument than a reductio-ad-absurdum, but we still refer to it as such due to its description as a reductio in the literature \citep{friston2015active}}. Central to this logic is the claim that the EFE is the `natural' extension of the VFE to account for uncertain futures. 

In the first section of this chapter, we investigate this claim in detail. Specifically, we argue that the EFE is not necessarily the only way to extend the VFE into the future, and that there are in fact other, more straightforward extensions, such as an objective we call the free energy of the future (FEF). We then perform a direct side-by-side comparison of the EFE and FEF functionals and comment on their similarities and differences, and discuss their respective bounding behaviour on the expected free energy. We then discuss how active inference approaches are related to the control as infrerence framework, and decide upon two key differences -- the objective functional utilized for action selection, where active inference uses the EFE, and control as inference uses the FEF, and secondly the encoding of value or goals into the inference process, where active inference directly encodes values into the generative model through the use of a biased desire distribution $\tilde{p}(o)$, control as inference instead uses independent optimality variables $p(\Omega | o)$ \footnote{We also denote any distribution involving a desire distribution with a $\tilde{p}$ and, for instance, refer to $\tilde{p}(o,x)$ as a \emph{biased generative model})}. Finally, we then introduce a second objective functional, which we call the free energy of the Expected Future (FEEF) which combines both an intuitively grounded starting point with the same exploration seeking term as is present in the EFE, and which was investigated previously in Chapter 4. We discuss the nature of different possible objective functionals for control.

In the second half of this chapter, we retreat from the specifics of the EFE, active inference, and control as inference, to instead define a general framework for understanding the origin of information seeking exploration terms in control functionals. We argue that the key distinction, is that between evidence objectives, which maximize the likelihood of achieving a desire distribution, with divergence functionals which try to minimize the divergence between a predicted and desire functional. Specifically, divergence objectives give rise to information gain terms while evidence objectives do not. We trace this capacity to the fact that divergence objectives implicitly maximize the entropy of the agent's future, in a close connection to empowerment objectives, while evidence objectives do not. Finally, we put all this together into a coherent framework which can be used to understand the full landscape of variational objective functionals for control tasks.

The material in this chapter is heavily based on three first-author papers. \emph{Whence the expected free energy} \citep{millidge2020whence} (published in neural computation), \emph{On the relationship between active inference and control as inference} (published at the IEEE international workshop on active inference) \citep{millidge2020relationship}, and \citep{millidge2021understanding} \emph{Understanding the Origin of Information-Seeking Exploration in Probabilistic Objectives for Control}, \emph{Arxiv} (to be submitted to Royal Society Interface).

\section{Origins of the Expected Free Energy}

Here we investigate the origins of the expected free energy (EFE) term within active inference. It is often claimed that the reason active inference agents minimize this term is that free energy minimizing agents must minimize the variational free energy (VFE) into the future which, since the future is uncertain, constitutes the \emph{expected} free energy. To make this claim precise, we need to understand exactly the variational free energy `into the future' should consist of. We argue that it must satisfy two conditions, which are both satisfied by the variational free energy, and which are crucial for that objective functions to operate. First, we argue that, like the VFE, the VFE extended into the future should be a divergence between a variational approximate posterior and a generative model of future states. Secondly, we argue that, again like the VFE, any free energy of the future should additionally be a bound on the log model evidence of future observations. These conditions are both important precisely because they define why the variational free energy is useful. Minimizing the divergence between the posterior and the generative model is useful since it implicitly makes the variational posterior a good approximation. Conversely, bounding the log model evidence is useful since the log-model evidence provides a very general measure of how `good' a specific model is, which can be used for Bayesian model-comparison or even just to understand the amount of inherent information in the data. Moreover, the log model evidence has especial import for methods under the aegis of the free energy principle since the log model evidence is simply the surprisal $-\ln p(o)$ which is the basic quantity which is minimized throughout the theory.

First, we need to define precisely the mathematical setup of the problem. We assume that our agent exists in a POMDP environment with states $x$, observations $o$, policies (sequences of actions) $\pi = [a_1, a_2 \dots a_T]$. The agent maintains a variational distribution over the states and actions and a generative model over the states, observations, and policies. Specifically, although we are technically interested in the functionals over a full trajectory $o_{1:T}$, in practice the functional decomposes into a sum of independent functionals for each timestep. Thus, for understanding the behaviour of agents optimizing the functional, it suffices to consider only a single timestep of the functional at $o_t$.

We argue that the expected free energy does not fulfil these conditions, but rather another objective functional does, which we call the free energy of the future (FEF). We define the FEF to be,
\begin{align*}
\mathbb{FEF}_t(\pi) &=  \mathbb{E}_{q(o_t, x_t | \pi)}[\ln q(x_t | o_t) - \ln \tilde{p}(o_t,x_t)] \\
&= \mathbb{E}_{q(o_t)}\KL[q(x_t | o_t) || \tilde{p}(o_t, x_t)] \numberthis
\end{align*}

which is simply the KL divergence between the approximate posterior generative model over future states, averaged under the expected future observations $q(o_t)$. This trivially satisifes the first condition, since it is a KL divergence between the variational posterior, and the generative model, as is the VFE. Next, we show that this functional is a bound on the expected log model evidence in the future.

\begin{align*}
    \label{FEF_bound}
    \numberthis
    - \mathbb{E}_{q(o_t | \pi)}\big[ \ln \tilde{p}(o_t) \big] &= - \mathbb{E}_{q(o_t | \pi)}\big[ \ln \int dx_t \tilde{p}(o_t,x_t) \big] \\
    &= - \mathbb{E}_{q(o_t | \pi)}\big[ \ln \int dx_t \tilde{p}(o_t,x_t) \frac{q(x_t | o_t)}{q(x_t | o_t)} \big] \\
    &\leq - \mathbb{E}_{q(o_t | \pi)}\int dx_t q(x_t | o_t) \big[ \ln  \frac{\tilde{p}(o_t,x_t)}{q(x_t | o_t)} \big] \\
    &\leq - \mathbb{E}_{q(o_t,x_t | \pi)} \big[ \ln  \frac{\tilde{p}(o_t,x_t)}{q(x_t | o_t)} \big] \\
    &\leq \mathbb{E}_{q(o_t,x_t | \pi)} \big[ \ln  \frac{q(x_t | o_t)}{\tilde{p}(o_t,x_t)} \big] \\
    &\leq \mathbb{E}_{q(o_t | \pi)}\mathbb{D}_{KL}[q(x | o_t)||\tilde{p}(o_t,x_t | \pi)] = \mathbb{FEF}(\pi) \numberthis
\end{align*}

Crucially, we can see that this is an \emph{upper bound} on the log model evidence, and thus minimizing the FEF will tend to decrease the gap between the FEF and the expected log-model evidence. This functional thus exhibits identical behaviour to the VFE. To gain a better understanding of the key differences between the EFE and the FEF, we can exhibit them side by side.

\begin{align*}
    \mathbb{FEF} &= \mathbb{E}_{q(o_t,x_t | \pi)}[\ln q(x_t | o_t) - \ln \tilde{p}(o_t,x_t)] \\
    \mathbb{EFE} &= \mathbb{E}_{q(o_t,x_t | \pi)}[\ln q(x_t | \pi) - \ln \tilde{p}(o_t,x_t)] \numberthis
\end{align*}

While the two formulations may look very similar, the key distinction is that the FEF optimizes the divergence between the variational \emph{posterior} $q(x_t | o_t)$ and the generative model while the EFE minimizes the variational \emph{prior} $q(x_t)$. While this difference may seem small, we see that it has a significant impact when it comes to the resulting interpretable terms from the decomposition of the two functionals,

\begin{align*}
    \label{FEFEFEComparison}
    \mathbb{FEF} &= -\underbrace{\mathbb{E}_{q(o_t,x_t | \pi)} \big[ \ln \tilde{p}(o_t | x_t) \big]}_{\text{Extrinsic Value}} + \underbrace{\mathbb{E}_{q(o_t | \pi)}\mathbb{D}_{KL}[q(x_t | o_t)||q(x_t | \pi)]}_{\text{Epistemic Value}} \numberthis \\
    \mathbb{EFE} &= \underbrace{-\mathbb{E}_{q(o_t,x_t | \pi)}\big[ \ln \tilde{p}(o_t) \big]}_{\text{Extrinsic Value}} -  \underbrace{\mathbb{E}_{q(o_t | \pi)}\mathbb{D}_{KL}[q(x_t | o_t)||q(x_t | \pi)]}_{\text{Epistemic Value}} \numberthis
\end{align*}

Specifically, we see that while both the FEF and the EFE can be split into `extrinsic' and `intrinsic' value terms, the intrinsic value term in the FEF is positive while in the EFE it is negative. Specifically this means that the FEF tries to \emph{minimize} exploration and keep the posterior and prior as close together as possible. This minimizing information gain term is analogous to the complexity term in the VFE which functions as a regulariser which attempts to keep the posterior as close to the prior as possible, while still fitting the data. Here, we see that the goal of the FEF is to, in effect, maximize reward, while trying to learn as little about the environment as possible. While this may seem to be an unfortunate objective, in some small cases it may be beneficial, especially when in the case of offline reinforcement learning, where there is no continual interaction with an environment, only trying to learn an optimal policy from a given dataset of interactions (\citep{levine2018reinforcement}. In such cases, failures of generalization and extrapolation can often result in poor results whenever the learned policy is moved even slightly off the data manifold, and this kind of conservative regularisation of the learning process can prove highly beneficial \citep{levine2020offline}. By contrast, the EFE \emph{maximizes} the information gain term, since it is negative, and tries to drive the posterior and prior as far apart as possible. This results in information-seeking exploration. 

However, while the EFE has an intuitively better exploratory grounding, it is not a bound on the log model evidence, as the FEF is. We can show this straightforwardly by noting that the `extrinsic value' term of the EFE simply \emph{is} the log model evidence,
\begin{align*}
    \mathbb{EFE} &= \mathbb{E}_{q(o_t,x_t | \pi)} [\ln q(x_t | \pi) - \ln \tilde{p}(o_t,x_t)] \\
    &\approx \mathbb{E}_{q(o_t,x_t | \pi)} [\ln q(x_t | \pi) - \ln q(x_t |o_t) - \ln \tilde{p}(o_t)] \\
    &\approx \underbrace{-\mathbb{E}_{q(o_t | \pi)} [ \ln \tilde{p}(o_t)]}_{\text{Negative Expected Log Model Evidence}} -  \underbrace{\mathbb{E}_{q(o_t | \pi)}\mathbb{D}_{KL}[ q(x_t | o_t) \Vert q(x_t | \pi)]|}_{\text{Information Gain}} \numberthis
\end{align*}
and that thus by the non-negativity of KL divergences, the EFE is a \emph{lower bound} on the log model evidence. This bound is in the wrong direction, since to make it tight, the EFE should be \emph{maximized} instead of minimized.

Importantly, in the definition of the EFE there is an approximation step where we have approximated $p(x_t | o_t)$ with the approximate posterior $q(x_t | o_t)$. If we make this approximation explicit, we can write the EFE as,

\begin{align*}
    \mathbb{EFE} &= \mathbb{E}_{q(o_t,x_t | \pi)} [\ln q(x_t | \pi) - \ln \tilde{p}(o_t,x_t)] \\
    &\approx \mathbb{E}_{q(o_t,x_t | \pi)} [\ln q(x_t | \pi) - \ln p(x_t |o_t) - \ln \tilde{p}(o_t)] \\
    &\approx \mathbb{E}_{q(o_t,x_t | \pi)} [\ln q(x_t | \pi) - \ln p(x_t |o_t) - \ln \tilde{p}(o_t) +\ln q(x_t |o_t) - \ln q(x_t | o_t)] \\
    &\approx \underbrace{\underbrace{-\mathbb{E}_{q(o_t | \pi)} [ \ln \tilde{p}(o_t)]}_{\text{Negative Expected Log Model Evidence}} +  \underbrace{\mathbb{E}_{q(o_t | \pi)}\mathbb{D}_{KL}[ q(x_t | o_t) \Vert p(x_t|o_t)]|}_{\text{Posterior Approximation Error}}}_{\text{FEF}}  \\ &-  \underbrace{\mathbb{E}_{q(o_t | \pi)}\mathbb{D}_{KL}[ q(x_t | o_t) \Vert q(x_t | \pi)]|}_{\text{Information Gain}} \numberthis
\end{align*}

Where we see that the EFE can be both an upper and lower bound on the log model evidence depending on whether the information  gain term or the posterior divergence term is larger. We can thus see that the likely time-course of the EFE is to cycle around the bound over the course of inference until, potentially, it reaches it. This is because, at the start of training, when inference is poor, the posterior divergence is likely greater than the information gain, so the EFE functions as an upper bound and minimizing it gets us closer to the true log model evidence. This effect likely quickly fades away as the information gain term becomes bigger, and here the EFE minimizing agent will preferentially explore its environment in an information-seeking fashion, driving the EFE \emph{away} from the real log model evidence for the environment. Finally, if there are no residual sources either of posterior divergence (so that the true and approximate posteriors are in the same class), or information gain (so that the agent has a perfect model of the environment, and the environment has no intrinsic stochasticity which gives rise to aleatoric uncertainty), then both the posterior divergence and the information gain terms will be 0, and the EFE will finally converge to exactly the log model evidence. While this behaviour of the EFE functional may lead to adaptive behaviour, it is not particularly mathematically principled as an extension to the VFE, and thus it is not necessarily clear why the EFE should be considered to be a better extension of the VFE than the FEF.

This derivation also reveals an interesting connection between the EFE and the FEF. Specifically, it is revealed that the EFE is simply the FEF minus an additional information gain term, thus effectively comprising the free energy into the future (FEF), with an additional exploratory information gain term. This derivation can also be derived straightforwardly from a direct comparison of the two functionals,

\begin{align*}
    \mathbb{FEF}_t(\pi) - \mathbb{IG}_t &= \mathbb{E}_{q(o_t,x_t | \pi)}\ln(\frac{q(x_t | o_t)}{\tilde{p}(o_t,x_t)}) - \mathbb{E}_{q(o_t,x_t | \pi)}\ln(\frac{q(x_t | o_t)}{q(x_t | \pi)}) \\
    &=  \mathbb{E}_{q(o_t,x_t | \pi)}\ln(\frac{q(x_t|o_t)q(x_t | \pi)}{\tilde{p}(o_t,x_t)q(x_t|o_t)}) \\ 
    &=  \mathbb{E}_{q(o_t,x_t | \pi)}\ln(\frac{q(x_t | \pi)}{\tilde{p}(o_t,x_t)}) \\
    &= \mathbb{EFE}(\pi)_t \numberthis
\end{align*}

We can thus understand the origin of the information gain term in the EFE -- it is simply the FEF into the future minus the information gain exploration term. This means that, in effect, the  exploratory properties of the EFE are simply present by construction. Is it possible, then, to derive mathematically or intuitively principled objectives which maintain the information seeking properties of the EFE?

While this question is definitively answered later in this chapter, here we present a hint of the solution. We propose a novel objective, which we call the free energy of the expected future (FEEF), which can be characterised simply as the divergence between the expected beliefs about future observations and states $q(o_t, x_t)$ and the desired distribution $\tilde{p}(o_t, x_t)$. The FEEF objective can be written as,

\begin{align*}
    \pi^* = \underset{\pi}{\mathrm{arg min}} \, \, \mathbb{D}_{KL}[q(o_{t:T},x_{t:T}|\pi)||\tilde{p}(o_{t:T},x_{t:T})] \numberthis
\end{align*}

In effect, this objective can be understood as compelling an agent to bring a predicted (variational) world and a desired (generative) distribution into alignment. This objective has a strong intuitive basis for understanding adaptive action, since we are simply trying to minimize the difference between our veridical beliefs about the future and our desires. Since the desire distribution is assumed fixed, the only way to maximize their alignment is to take action to force the predicted belief distribution towards the desired distribution. If the belief distribution is accurate, then this will result in trajectories that really do take the agent towards its desired distribution. Crucially, we can then decompose this objective into an extrinsic and intrinsic information seeking term, just like the EFE.
\begin{align*}
    \mathbb{FEEF}(\pi)_t &= \mathbb{E}_{q(o_t,x_t|\pi)} \ln \big[ \frac{ q(o_t,x_t |\pi)}{\tilde{p}(o_t,x_t )} \big] \\
    &\approx \underbrace{\mathbb{E}_{q(x_t | \pi)} \mathbb{D}_{KL} \big[ q(o_t | x_t) \Vert \tilde{p}(o_t) \big]}_{\text{Extrinsic Value}} - \underbrace{\mathbb{E}_{q(o_t | \pi)} \mathbb{D}_{KL} \big[ q(x_t | o_t) \Vert q(x_t | \pi) \big]}_{\text{Intrinsic Value}} \numberthis
\end{align*}

Here, we see that the epistemic information seeking term is identical to that of the EFE, and thus we would expect FEEF and EFE minimizing agents to show similar exploratory behaviour. The key difference between these objectives lies in the extrinsic value term. While the EFE simply aims to maximize the likelihood of the desire distribution under the variational belief distribution, the FEEF explicitly tries to minimize the KL divergence between them, and thus try to match the two distributions.

Another way of looking at the same thing is to consider this straightforward relationship between the EFE and the FEEF,
\begin{align*}
    \mathbb{FEEF}(\pi)_t &= \mathbb{D}_{KL} \big[ q(o_t, x_t) \Vert \tilde{p}(o_t,x_t) \big] \\
    &= \underbrace{\mathbb{E}_{q(o_t, x_t)}\big[ \ln q(o_t | x_t)]}_{\text{Observation Likelihood}} + \underbrace{\mathbb{E}_{q(o_t, x_t)}\big[ \ln \tilde{p}(o_t | x_t)] - \mathbb{E}_{q(o_t | \pi)} \mathbb{D}_{KL} \big[ q(x_t | o_t) \Vert q(x_t | \pi) \big]}_{\text{EFE}} \numberthis
\end{align*}

We can thus see, that the FEEF is simply the EFE plus an observation likelihood entropy term. This term is to be maximized and thus effectively provides an additional source of random exploration to the FEEF agent rather than the EFE. In effect, the FEEF agent optimizes the EFE while trying to keep its observation mapping as random as possible. Another advantage of the FEEF, is that it is equivalent to the VFE at the present time. This is because the observation entropy term is constant since it cannot be affected by future observations, and thus the expression as a whole reduces to the VFE. This means that the FEEF can be used as a unified objective for both perception and action, while the EFE can only be used for control. Due to this, a FEEF agent can have all distributions trained jointly on the FEEF objective while for an active inference agent, typically, if the transition and likelihood matrices are learnt, they are optimized against the VFE, while only action selection takes place using the EFE. This adds an additional degree of simplicity and elegance to FEEF-minimzing agents while they retain the same exploratory behaviours as active inference agents.

\subsection{Control as Inference and Active Inference}

This relationship between the FEF and the EFE sheds light upon the relationship between active inference and control as inference approaches to control. While the formulations at an abstract level are very similar -- both attempt to solve the control problem by deriving inference algorithms which operate on graphical models, and usually utilize the machinery of variational inference to do so -- at a lower detailed level the theories appear quite different and are presented with substantially differing motivations and notation. Using our newfound understanding of variational objective functionals of the future, such as the EFE and the FEF, here we pin down what exactly the relationship between control as inference and active inference is. 

First, we note that there are many straightforward notational differences between the theories which can be overcome. One obvious difference is that active inference is primarily concerned with the inferring of policies (or action sequences) while control as inference concerns itself with simply inferring policies, or single actions for a given timestep. It is important to note, however, that it is possible to reformulate active inference so that it infers policies, and, conversely, to reformulate control as inference so that it infers full action plans.  A second distinction is that active inference is typically formulated for POMDPs while control as inference only for MDPs. It is straightforward, however, to extend control as inference to the POMDP setup, which results in the following objective,
\begin{align*}
    \mathcal{L}(\phi)_{CAI} &= \KL \Big(q_\phi(x_t, a_t) \Vert p(x_t, a_t,o_t,\Omega_t) \Big) \\
    &= \underbrace{-\mathbb{E}_{q_\phi(x_t, a_t)}\big[ \ln p(\Omega_ | x_t, a_t) \big]}_{\text{Extrinsic Value}} + \underbrace{\KL \Big( q(x_t) \Vert p(x_t | x_{t-1},a_{t-1}) \Big)}_{\text{State divergence}} \\
    & \ \ + \underbrace{\mathbb{E}_{q(x_t)} \big[\KL \Big( q_\phi(a_t | x_t) \Vert p(a_t | x_t) \Big) \big]}_{\text{Action Divergence}} - \underbrace{\mathbb{E}_{q_\phi(x_t, a_t)}\big[ \ln p(o_t | x_t)\big]}_{\text{Observation Ambiguity}} \numberthis
\end{align*}
Here we have used notation standard in control as inference derivations, namely $q_\phi(a_t | x_t)$ is an amortized state-action policy and $\Omega_t$ is the `optimality variable'.
Importantly, this novel extension of control-as-inference to a POMDP setting leads directly to a straightforward implementation in terms of deep reinforcement learning, similar to the approaches in Chapter 4. Specifically, this objective can either be expressed directly as a sum over trajectories, and thus optimized by planning algorithms using model-based deep reinforcement learning or, alternatively, it can be expressed recursively and computed using a value or Q function approach which lends itself naturally to model-free deep reinforcement learning approaches. The key extension would be learning a probabilistic encoder-decoder model, most likely a VAE, to infer the distributions $q(x_t | o_t)$ and $p(o_t | x_t)$ and then to optimize the entropy of the VAE decoder in the control objective, in accordance with this objective function. Empirically investigating the performance of this method, and the impact of the observation ambiguity term, has not, to my knowledge, been explored in the literature, and would be an interesting avenue for further work. 

Secondly, we can similarly reformulate the control as inference approach to infer plans instead of policies. This is done by extending the generative model to cover whole trajectories instead of single observations, states, or actions. We then infer a constant random variable $\pi$ the policy for the whole trajectory. Written out explicitly, from this generative model you can derive a variant of the active inference optimal plan derivation to discover that the optimal plan under the control as inference is simply the softmax path integral over the variational free energy (VFE), augmented with optimality variables, and extended into the future.

\begin{align*}
    \mathcal{L}_{CAI} &= \KL \Big( q(x_{t:T}, \pi) \Vert p(x_{t:T}, \pi, o_{t:T}, \Omega_{t:T}) \Big) \\
    &= \KL \Big(q(\pi) \prod_t^T q(x_t | \pi) \Vert p(\pi) \prod_{t}^T p(\Omega_t | x_t, \pi)p(o_t | x_t)p(x_t | x_{t-1}, \pi) \Big)\\
    &= \KL \Big( q(\pi) \sum_t^T \KL \big[ q(x_t | \pi) \Vert p(\Omega_t | x_t, \pi)p(o_t | x_t)p(x_t | x_{t-1}, \pi) \big]\Vert p(\pi) \Big) \\
    &= \KL \Big( q(\pi)\Vert p(\pi) \exp(- \sum_t^T \mathcal{L}_t(\pi) ) \Big) 
    \implies q^*(\pi) = \sigma \Big(p(\pi) - \sum_t^T \mathcal{L}_t(\pi)\Big)  \numberthis
\end{align*}
We can decompose this VFE functional into the future as, 

\begin{align*}
    \mathcal{L}_t(\pi)_{CAI} &= \mathbb{E}_{q(x_t | \pi)}\big[ \ln q(x_t|\pi) - \ln p(x_t, \pi, o_t, \Omega_t)] \\
    &= -\underbrace{\mathbb{E}_{q(x_t | \pi)}\big[ \ln p(\Omega_t | x_t, \pi) \big]}_{\text{Extrinsic Value}} + \underbrace{\KL \Big(q(x_t | \pi) \Vert p(x_t | x_{t-1}, \pi) \Big)}_{\text{State divergence}} 
     - \underbrace{\mathbb{E}_{q(x_t | \pi)}\big[ \ln p(o_t | x_t) \big]}_{\text{Observation Ambiguity}}  \numberthis
\end{align*}

Which we can see is equivalent to the standard control as inference POMDP objective, except that it is missing the action divergence terms. The action divergence terms are missing simply because full policies are inferred instead of individual actions. Conversely, we can rederive active inference to infer individual actions rather than full policies. To do so, we simply need to add individual actions into the generative model and variational density and then crank through the derivation,

\begin{align*}
    -\mathcal{F}_t(\phi) &= \mathbb{E}_{q(o_t,x_t,a_t)}\Big[\ln q_\phi(a_t, x_t) - \ln \tilde{p}(x_t, o_t, a_t) \Big] \\
    &= -\underbrace{\mathbb{E}_{q(o_t|a_t)}\big[\ln \tilde{p}(o_t|a_t) \big]}_{\text{Extrinsic Value}} - \underbrace{\mathbb{E}_{q(o_t,a_t  | x_t)}\Big[ \KL \big( q(x_t | o_t, a_t) \Vert q(x_t|a_t) \big) \Big]}_{\text{Intrinsic Value}}  \\ &+ \underbrace{\mathbb{E}_{q(x_t)}\Big[ \KL \big( q_\phi(a_t | x_t) \Vert p(a_t | x_t) \big) \Big]}_{\text{Action Divergence}} \numberthis
\end{align*}
Here we see that the expression to be optimized with respect to the policy parameters $\phi$ is simply the standard active inference objective with an additional action-divergence term. If we assume the action prior $p(a_t | x_t)$ is uniform, then we regain the well known control as inference policy entropy term. Now that we have extended the theories to allow for a direct side-by-side comparison, we can see that the two major differences lie in the information gain term for active inference as opposed to the complexity `state-divergence' term for control as inference, and secondly that the control as inference approach contains an additional `likelihood entropy' term in its objective which active inference lacks. We know now that the information gain term in active inference arises directly from the definition of the EFE functional, which is not an intrinsic part of active inference and may not be theoretically justified. Indeed, if we replace the EFE in the active inference derivation with the FEF, we can obtain an objective identical to the control as inference approach except that it has no likelihood entropy term.

\begin{align*}
    -\hat{\mathcal{F}_t}(\phi) &= \mathbb{E}_{q_\phi(x_t, o_t, a_t)}\big[\ln q_\phi(x_t, a_t) - \ln \tilde{p}(o_t, x_t, a_t) \big]  \\
    &=  \underbrace{-\mathbb{E}_{q_\phi(x_t, a_t)}\big[ \ln \tilde{p}(o_t | x_t) \big]}_{\text{Extrinsic Value}} + \underbrace{\KL \Big( q(x_t) \Vert p(x_t | x_{t-1},a_{t-1}) \Big)}_{\text{State divergence}}  + \underbrace{\KL \Big( q_\phi(a_t | x_t) \Vert p(a_t | x_t) \Big)}_{\text{Action Divergence}} \numberthis
\end{align*}

This means that, effectively, we can consider control as inference as active inference with a FEF objective or, conversely, that active inference is simply control as inference with a nonstandard EFE objective. The question then remains, why does the control as inference objective possess an additional likelihood entropy term which active inference does not, since it is not due to a difference in objective function. We demonstrate that the difference actually arises due to the way goals or likelihoods are encoded in the inference procedure. Specifically, active inference encodes goals or rewards \emph{directly} into the generative model, so that active inference agents function with a biased generative model which blends reward-driven and veridical perception. By contrast, control as inference, keeps veridical inference and reward computation entirely separate, and instead includes rewards into the inference procedure through the use of exogenous optimality variables. The two methods thus solve subtly different inference problems due to these distinctions in how they encode rewards. Put simply, active inference encodes rewards and goals through priors; control as inference encodes them through conditioning.  Phrased intuitively, we can think of control as inference as solving the inference problem: \emph{Assuming that I have acted optimally in the future, what actions do I infer I will have taken}, while active inference solves the inference problem: \emph{Infer my most likely actions, given that I strongly believe in the future I will observe highly desirable states}. In sum, control as inference maintains a strict distinction between a veridical perceptual generative model, which generates objectively likely future outcomes for a given action sequence, while active inference maintains a biased generative model which preferentially predicts desired outcomes. Control as inference conditions this accurate generative model on observing high reward contingencies, and thus maintains an adaptive action plan. Active inference, on the other hand, simply maximizes the likelihood of this biased generative model, thus also inferring adaptive actions.

While this distinction may seem arcane, it actually speaks to deep philosophical differences in perspectives between the two theories. Control as inference arises from the cognitivist and logical views of artificial intelligence, which maintains that in some sense intelligence is pure thought, which can then be unbiasedly applied to inferring action. Control as inference maintains the modularity thesis, which is widespread throughout artificial intelligence and engineering fields which is the strict separation of perception and control. Perception aims to build up an accurate world model, while control uses this world model to compute the best actions. Active inference, by contrast, has much more in common with enactivist, embodied, and cybernetical views of control, which see the agent and environment inextricably enmeshed in a fundamental feedback process -- the perception action loop. Here, it is unnecessary to maintain a strict separation between perception and control. Indeed, veridical perception is unnecessary since the ultimate aim of perception, in this view, is not the unbiased modelling of the world, but rather to subserve adaptive action. 

The mathematical effect of this distinction is that control as inference maintains two separate likelihoods - a veridical perceptual likelihood $p(o_t | x_t)$, and a reward-encoding optimality likelihood $p(\Omega | x_t, a_t)$, while active inference only maintains a biased observational likelihood $\tilde{p}(o_t)$. Due to this, control as inference approaches possess an additional likelihood entropy term which active inference ones lack. 

\section{Evidence and Divergence Objectives}

Given that we now see that the control as inference framework, even when extended to full action policies and partially observable environments, does not manifest an information-seeking exploration term while the expected free energy does and, moreover, that conversely the EFE does not function as a bound on a known quantity like the expected model evidence which is used in variational inference, we are left with the question of trying to understand whether and how information-seeking objectives can be derived in a mathematically principled manner. Here, we argue that objective functionals for control existing in the literature can be sensibly split into two separate classes -- \emph{evidence} objectives, which typically arise from a direct variational inference approach, and which do not manifest information gain terms, and \emph{divergence objectives} which arise out of directly minimizing a KL divergence between two models, and which do give rise to information seeking terms. We then show how well known objectives in a variety of literatures can be understood in our scheme. 

First, we need to formalize our mathematical setup. We assume that we have an environment which, from the agent's perspective, is an unknown black box. The agent inputs actions $a_{1:T}$ to the environment and it outputs observations $o_{1:T}$ in return. The exact process that produces these observations in the environment is forever unknown. The agent can \emph{assume} that the observations in the environment are produced by some form of POMDP model with hidden latent states $x_{1:T}$, but this is merely the agent's model of the environment. The agent can also parametrize its model with learnable parameters $\theta$, which it assumes are either constant throughout its entire lifetime, or else change on a much slower timescale than the latent states $x_{1:T}$. Next, to formalize a control, as opposed to just an inference problem, we need some notion of the reward or goal that the agent wishes to achieve. We formalize this by specifying that the agent possesses an additional \emph{desire distribution} $\tilde{p}(o_{1:T})$ over the observations it receives. This distribution encodes the `ideal world' of the agent in the sense that these are precisely the observations it aims to achieve. To map this to a standard reinforcement learning setting, with rewards, we can perform the now familiar trick of defining, $\tilde{p}(o_{1:T}) = exp(-r(o_{1:T}))$ where $r(o_{1:T})$ is the total reward achieved across a given observation trajectory. Importantly, this equivalence to standard reinforcement methodology is only a special case, and that this formulation in terms of a desire distribution is more flexible in its specification of the rewards. Specifically, here we assume no specific form of the desire distribution, nor any factorization of desires across timesteps, for instance, so the desired observations in the future can depend on the desired observations at the present in arbitrarily complex ways. It is also straightforward to extend the framework to a desired distributions over actions as well, by defining $\tilde{p}(o_{1:T}, a_{1:T})$. This is useful for instance if we want to penalize the costs (i.e. energetic for biological organisms or robotic systems) of action. For instance, to define a quadratic cost in the action magnitude, we can define $\tilde{p}(a_{1:T}) = \prod_t^T \mathcal{N}(a; 0, \sigma)$, or a Gaussian distribution in the action centered around 0. The variance of the Gaussian ($\sigma$) effectively defines the weighting coefficient which scales the size of the penalty. Here, to keep notation simple, we do not discuss action penalties and only work with a desired observation distribution $\tilde{p}(o_{1:T})$, but the extension to actions is entirely straightforward.

The general objective of the control problem is to compute an action trajectory $a_{1:T}$ which results in the realization of your goals, which are defined through the desire distribution $\tilde{p}(o_{1:T})$. We also assume, that if the agent actually executes a given action trajectory, it will receive a real trajectory of observations from the environment according to some distribution $p(o_{1:T} | a_{1:T})$. This distribution can be approximated either by sampling real trajectories from the environment, as is implicitly the case in model-free reinforcement learning methods, or else the agent can explicitly \emph{model} this distribution as $p(o_{1:T} | a_{1:T}; \theta)$, with parameters $\theta$, which may be implemented as a deep neural network, for example. We call this distribution the predicted distribution, since in some sense it is what the agent \emph{predicts} will occur if it executes a given action trajectory. 

 With the preliminaries settled, we can formally define the \emph{evidence} and \emph{divergence} objectives. Evidence objectives try to maximize the likelihood of the desire distribution averaged under the predicted distribution. Intuitively, we wish to find the action trajectory that maximizes the expected likelihood of the desire distribution. Mathematically, we can represent this as,

 \begin{align*}
\mathcal{L}_{evidence} = \underset{a_{1:T}}{argmax} \, \, \mathbb{E}_{p(o_{1:T} | a_{1:T})}[\ln \tilde{p}(o_{1:T})] \numberthis
 \end{align*}

Where here, as is usual, we optimize the \emph{log} of the desire distribution instead of the desire distribution itself. Since the log is a monotonic function, this does not affect the actual optimum of the problem and since we often factorize the desire distribution into products, the log will turn that into a sum, which is usually much better behaved numerically. 

Conversely, for a \emph{divergence} objective, instead of maximizing the likelihood of the desires under the predicted distribution, we instead with to directly \emph{minimize} the \emph{divergence} of the desire and predicted distribution. In effect, we want to make the desire distribution and the predicted distribution match. Mathematically, we can write this as,
\begin{align*}
\mathcal{L}_{divergence} = \underset{a_{1:T}}{argmin} \, \, \KL[p(o_{1:T}) | a_{1:T}) || \tilde{p}(o_{1:T})] \numberthis
\end{align*}
Where here we use the KL divergence $\KL[Q||P] = \mathbb{E}_Q \ln \frac{Q}{P}$. Intuitively, we can think of the distinction between evidence and divergence objectives being that the evidence objective seeks to match the predicted distribution to the \emph{mode} of the desire distribution -- i.e. it focuses all effort onto the most desired observations, while neglecting the lesser desired observations. This is why evidence objectives typically arise from direct maximizing principles such as utility maximization, or variational inference. Meanwhile, divergence objectives seek to precisely match the predicted distribution and the desired distribution, so that if some observation is not the most preferred one, but has some moderate level of preference, the divergence objective would seek to have that observation manifest some amount of time in proportion to its relative preferredness. This means that in general, if the desire distribution is spread out, then agents will seek to realize a spread-out distribution of predicted observations. In contrast, under an evidence objective, even if the desire distribution is broad, the agent will continue to place all effort to keep the predicted distribution a peak around the mode of the desire distribution.

\begin{figure}[H]
\centering
\begin{subfigure}{.5\textwidth}
  \centering
  \includegraphics[width=.9\linewidth]{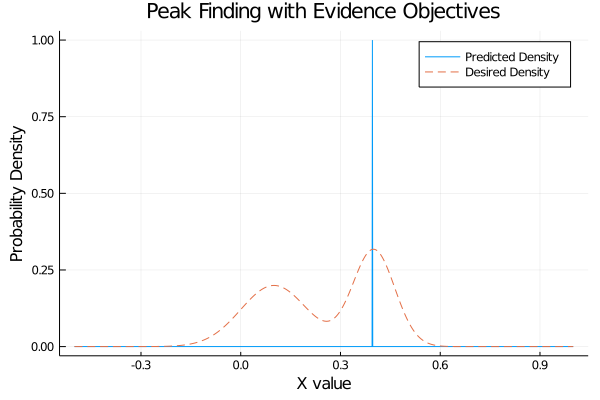}
  \caption{Optimizing with an Evidence Objective}
\end{subfigure}%
\begin{subfigure}{.5\textwidth}
  \centering
  \includegraphics[width=.9\linewidth]{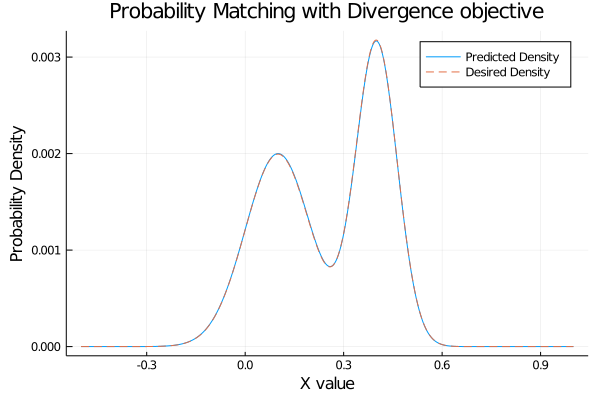}
  \caption{Optimizing with a Divergence Objective}
\end{subfigure}
\caption{Numerical illustration of optimizing a multimodal desired distribution with an Evidence objective (Panel A) vs a Divergence Objective (panel B). The desire distribution consisted of the sum of two univariate Gaussian distributions, with means of $1$ and $4$ and variances of $1$ and $0.4$ respectively. We then optimized an expected future distribution, which also consisted of two Gaussians with free means and variances using both an Evidence and a Divergence objective. As can be seen, optimizing the Evidence Objective results in the agent fitting the predicted future density entirely to an extremely sharp peak around the mode of the desired distribution. Conversely, optimizing a divergence objective leads to a precise match of the predicted and desired distributions (panel B shows the two distributions almost precisely on top of one another). As a technical note, to be able to see both the evidence and deisre distributions on the same scale, for the evidence objective the predicted distribution is normalized but the desired distribution is not. Code for these simulations can be found at: https://github.com/BerenMillidge/origins\_information\_seeking\_exploration.}
\end{figure}
Another way we can interpret the difference between the objectives is in terms of the effect of the objective upon the predicted distribution $p(o_{1:T} | a_{1:T})$. Specifically, we can think of the divergence objective as a balance between trying to maximize the likelihood of the desired distribution under the predicted distribution, and trying to \emph{maximize} the entropy of the predicted distribution. We can think of this as the divergence objective as effectively saying `try to maximize your desires or utility while also keeping the future as broad as possible' -- i.e. keeping your options open. This can be demonstrated mathematically through the simple definition of the KL divergence,
\begin{align*}
\mathcal{L}_{divergence} &= \underset{a_{1:T}}{argmin} \, \KL[p(o_{1:T}) | a_{1:T}) || \tilde{p}(o_{1:T})] \\
&= \underset{a_{1:T}}{argmin} \, \underbrace{-\mathbb{H}[p(o_{1:T} | a_{1:T})]}_{\text{Predicted Entropy}} - \underbrace{\mathbb{E}_{p(o_{1:T} | a_{1:T})}[\ln \tilde{p}(o_{1:T})]}_{\text{Evidence Objective}}
\end{align*}

Here we see that we can express the divergence objective simply as the evidence objective plus the maximization of the entropy of the predicted distribution. In effect, the divergence objective simply includes an entropy regulariser to the standard evidence objective. Conversely, we can also express the relationship between the objectives in the other way. We can think of the evidence objective as simply trying to match the predicted and desire distribution while simultaneously \emph{minimizing} the entropy of the predicted distribution. This is straightforward to show mathematically,
\begin{align*}
\mathcal{L}_{Evidence} &= \underset{a_{1:T}}{argmax} \, \mathbb{E}_{p(o_{1:T} | a_{1:T})}[\ln \tilde{p}(o_{1:T})] \\
    &= \underset{a_{1:T}}{argmax} \, \mathbb{E}_{p(o_{1:T} | a_{1:T})}[\ln \tilde{p}(o_{1:T})\frac{p(o_{1:T} | a_{1:T})}{p(o_{1:T} | a_{1:T})}] \\
    &= \underset{a_{1:T}}{argmax} \, -\underbrace{\KL[p(o_{1:T} | a_{1:T}) || \tilde{p(o_{1:T})}]}_{\text{Divergence}} - \underbrace{\mathbb{H}[p(o_{1:T} | a_{1:T})]}_{\text{Expected Future Entropy}} \numberthis
\end{align*}

This formulation gives a straightforward intuition for the `mode-seeking' behaviour of the evidence objective. The evidence objective seeks to match the predicted and desire distribution, while also being penalized for the breadth of the predicted distribution. The best way to resolve this tension is by forming a highly peaked predicted distribution around the mode of the desire distribution so that it can cover as much probability mass as possible. 

Interestingly, differences between the two formulations generally only emerge when the desire distribution is broad and complex. Here, the divergence objective will tend to force the predicted distribution and desire distribution to match in their complexity, while the evidence distribution will seek out and focus around its mode. Another intuitive way of thinking about the distinction is that the divergence objective implicitly maximizes some sort of future empowerment, by implicitly trying to keep all future options open, by seeking to make future observations as entropic as possible. Evidence objectives, by contrast, seek the opposite. They aim for `precise futures' where the amount of future variability is as low as possible. It is straightforward to show that, unlike evidence objectives, divergence objectives can be immediately decomposed into an `extrinsic' value divergence term, and an information-seeking exploratory term,

\begin{align*}
    \KL[p(o_{1:T} | a_{1:T}) || \tilde{p}(o_{1:T})] &= \mathbb{E}_{p(o_{1:T} | a_{1:T})}[\ln \frac{p(o_{1:T} | a_{1:T})}{\tilde{p}(o_{1:T})}] \\
 &= \mathbb{E}_{p(o_{1:T} | a_{1:T})}[\ln \frac{p(o_{1:T} | a_{1:T})}{\tilde{p}(o_{1:T})}] \\
 &= \mathbb{E}_{p(o_{1:T} | a_{1:T})}[\ln \frac{p(o_{1:T},x_{1:T} | a_{1:T})}{\tilde{p}(o_{1:T})p(x_{1:T} | o_{1:T})}] \\
 &= \underbrace{\mathbb{E}_{p(x_{1:T})}\KL[p(o_{1:T} | x_{1:T})||\tilde{p}(o_{1:T})]}_{\text{Desire Divergence}} \\ &- \underbrace{\mathbb{E}_{p(o_{1:T} | a_{1:T})}\KL[p(x_{1:T} | o_{1:T})||p(x_{1:T})]}_{\text{Information Gain}} \numberthis
\end{align*}

Importantly, the property that divergence objectives give rise to directed information-seeking exploratory terms, arises directly from the previously discussed intuition that they attempt to maximize the entropy of future observations. Such information gain terms arise whenever the predicted distribution is extended to model additional latent variables or parameters. The proof of this is straightforward,
\begin{align*}
    \mathbb{H}[p(o_{1:T} | a_{1:T})] &= \mathbb{E}_{p(o_{1:T},x_{1:T} | a_{1:T})}[\ln p(o_{1:T} | a_{1:T})] \\
    &= \mathbb{E}_{p(o_{1:T},x_{1:T} | a_{1:T})}[\ln \frac{p(o_{1:T},x_{1:T} | a_{1:T})}{p(x_{1:T} | o_{1:T})}] \\ 
    &=- \underbrace{E_{p(x_{1:T})}\mathbb{H}[p(o_{1:T} | x_{1:T})]}_{\text{Likelihood Entropy}} - \underbrace{\mathbb{E}_{p(o_{1:T} | a_{1:T})}\KL[p(x_{1:T} | o_{1:T}) || p(x_{1:T})]}_{\text{Expected Information Gain}} \numberthis
\end{align*}

Put verbally, this relationship shows that to maximize the entropy of a distribution, if the distribution can be understood in terms of a latent set of variables, requires both maximizing the conditional entropy of the variable given the latent variables, while simultaneously maximizing the mutual information of the latent variables between the observed and latent variables. Intuitively, within our context, this means that to maximize the entropy of future observations, it is necessary to successfully model the relationship between these future observations and their latent states, which entails maximizing the mutual information between the latents and the observations. It is the maximization of this mutual information which undergirds the exploratory information-seeking behaviour which is manifested by divergence objectives. Conversely, the fact that evidence objectives seek to \emph{minimize} the entropy of the predicted distribution means that they implicitly seek to minimize the amount of mutual information between observation and latent variables. We can think of this as divergence objectives aim to reach a given set of goals while also learning as much as possible about the environment, in order to precisely match the two distributions. Evidence objectives, on the other hand, seek to reach their goals while learning as little as possible about the environment, and keeping the environment as regular and predictable as possible. 

Now that we have proposed and given considerable intuition for our dichotomy between evidence and divergence objectives, we look to see where these objectives appear in the literature, and how they can explain differences in exploratory behaviour between differing paradigms.

\subsection{Control as Inference}

It is straightforward to show that the control as inference, and variational inference objectives are bounds upon the evidence objective. Put simply, we have that control as inference aims to solve the inference problem of inferring an action distribution given a desired set of observations. Specifically, we seek to obtain the distribution $p(a_{1:T} | \tilde{o}_{1:T})$ where we use $\tilde{o}$ to denote a set of hypothetical `optimal' actions which have been conditioned upon. To find this posterior, we use a variational approximation by defining the variational density $q(a_{1:T})$ and optimizing the following variational lower bound,
\begin{align*}
    \KL[q(a_{1:T} || p(a_{1:T} | \tilde{o}_{1:T})] &= \KL[q(a_{1:T} || \tilde{p}(o_{1:T}, a_{1:T})] - \ln p(o_{1:T}) \\
    &\geq \underbrace{\KL[q(a_{1:T} || \tilde{p}(o_{1:T}, a_{1:T})]}_{\text{CAI Objective}}
\end{align*}
Which serves as the control as inference objective. It is then straightforward to show that this objective serves as a bound on an evidence objective,
\begin{align*}
    \underset{a_{1:T}}{argmax} \, \ln \tilde{p}(o_{1:T}) &= \underset{a_{1:T}}{argmax} \, \ln \int dx \, \tilde{p}(o_{1:T} x_{1:T}) \\
    &= \underset{a_{1:T}}{argmax} \, \ln \int dx \,  \frac{\tilde{p}(o_{1:T} x_{1:T})q(a_{1:T})}{q(a_{1:T})} \\
    &\geq \underset{a_{1:T}}{argmax} \, \mathbb{E}_{q(a_{1:T})}[\ln \frac{\tilde{p}(o_{1:T},a_{1:T})}{q(a_{1:T})}] \\
    &\geq \underset{a_{1:T})}{argmin} \, - \KL[q(a_{1:T}) || \tilde{p}(o_{1:T}, a_{1:T})] \\
    &= \geq \mathcal{L}_{CAI} \numberthis
\end{align*}

And thus we can see that the control as inference framework optimizes a variational bound on the evidence objective. This straightforwardly explains why control as inference approaches do not give rise to directed, information-seeking exploration, but rather instead only induce \emph{random} action entropy maximizing exploration terms. This \emph{random} exploration, while highly efficient in many contemporary dense-reward reinforcement learning benchmark tasks, where a random policy often suffices to cover enough of the state space, it is increasingly ineffective in extremely high dimensional, and sparse reward environments.

\subsection{KL Control}

Another control method in the literature that has been applied, and studied fairly extensively is KL control. Although not as widely used in reinforcement learning, it is often implicit optimized in control tasks, and has deep relationships with the beginning of the control as inference approach, as well as more esoteric path integral methods. Moreover, it has recently seen renewed application in deep reinforcement learning approaches such as state-marginal matching of \citep{lee2019efficient}. KL control, as the name implies, chooses control to minimize the KL divergence between the current state and a set of desired states, leading to the following objective function,
\begin{align*}
    \mathcal{L}_{KL} = \underset{a_{1:T}}{argmin} \, \, \KL[p(x_{1:T}) || \tilde{p}(x_{1:T})] \numberthis
\end{align*}
Here we have used $x$ instead of $o$ to denote that KL control has typically only been applied to fully-observed Markovian MDP environments as opposed to full POMDP dynamics. As such, while the KL control objective is clearly just the divergence objective,  its superior exploratory capabilities have not been significantly explored in the literature due to the only applications currently being in fully observable environments while the information-seeking exploratory terms require the extension to hidden variable models. Another interesting point is that the objective in continuous time active inference in predictive coding models can also be as a KL divergence between a desired `set-point' and a currently observed point, and is thus an instance of KL control. However, these models also do not handle latent variable models, and thus also do not manifest the full exploratory capabilities of divergence objectives.

\subsection{Active Inference}

Given that we know from previously, that the expected free energy contains an information gain term, which gives rise to the information-seeking exploratory behaviour of active inference agents, it is worth investigating the relationship of the EFE to evidence and divergence functionals. Recall, from the previous section that the EFE formed neither an upper nor a lower bound upon the log model evidence, but instead formed an upper bound when the posterior divergence was greater than the information gain term, and a lower bound otherwise, with the goal of eventually converging directly to the log model evidence in the case that both of these terms become zero. Noting that the log model evidence simply, as the name suggests, is the evidence, objective, we can rewrite this in terms of our new understanding as,

\begin{align*}
    \underbrace{\mathbb{E}_{q(o,x)}[\ln \tilde{p}(o)]}_{\text{Evidence Objective}} &= \underbrace{\mathbb{E}_{q(o,x)}[\ln q(x) - \ln \tilde{p}(o,x)]}_{\text{EFE}} +  \underbrace{\mathbb{E}_{q(o)}\KL[q(x | o) || q(x)]}_{\text{Information Gain}} - \underbrace{\mathbb{E}_{q(o)}\KL[q(x | o) || p(x | o)]}_{\text{Posterior Divergence}} \\
    &\implies \underbrace{\mathbb{E}_{q(o,x)}[\ln \tilde{p}(o)]}_{\text{Evidence Objective}} \geq \underbrace{\mathbb{E}_{q(o,x)}[\ln q(x) - \ln \tilde{p}(o,x)]}_{\text{EFE}} \\
    &\text{If} \, \,  \underbrace{\mathbb{E}_{q(o)}\KL[q(x | o) || q(x)]}_{\text{Information Gain}} \geq \underbrace{\mathbb{E}_{q(o)}\KL[q(x | o) || p(x | o)]}_{\text{Posterior Divergence}} \numberthis
\end{align*}
so that the EFE does not stand in a straightforward relationship as bound in any specific direction on the evidence objective. Nevertheless, since the EFE contains an information gain term to be maximized, as do divergence objectives, we might expect to obtain a straightforward relationship between the EFE and the divergence objective. We can write out the relationship between the divergence objective and the EFE as follows,

\begin{align*}
    \underbrace{\KL[p(o) || \tilde{p}(o)]}_{\text{Divergence Objective}} &= \mathbb{E}_{p(o)}[\ln \frac{ \int dx p(o,x)}{\tilde{p}(o)}] \\
    &= \mathbb{E}_{p(o)}[\ln \frac{ \int dx p(o,x)q(x | o)q(o,x)}{\tilde{p}(o)q(x | o)q(o,x)}] \\
    &\leq \mathbb{E}_{p(o)}[\ln \frac{ \int dx p(o,x)q(o,x)}{\tilde{p}(o)q(x | o)q(o,x)}] \\
    &\leq \mathbb{E}_{p(o)}[\ln \frac{ \int dx p(o,x)q(o | x)q(x)}{\tilde{p}(o)q(x | o)q(x | o)q(o)}] \\
    &\leq \underbrace{\mathbb{E}_{p(o)q(x | o)}[\ln q(x) - \ln q(x | o)- \ln \tilde{p}(o)]}_{\text{EFE}}- \underbrace{\mathbb{E}_{p(o)}\KL[q(x | o)||p(o,x)]}_{\text{VFE}} \\ &+\underbrace{\mathbb{E}_{q(x | o)p(o)}\KL[q(x | o) || q(x)]}_{\text{Information Gain}} \numberthis
\end{align*}

Here we see that the EFE can be expressed in terms of the divergence objective, an information gain term, and, interestingly, the variational free energy.  In effect, the divergence objective consists of the EFE, the VFE, and an information gain term. Specifically, the EFE becomes an upper bound on the divergence objective when the information gain term is greater than the variational free energy. This is similar to previously where we saw that the EFE became a bound on the evidence when the information gain is less than the posterior divergence. It is thus clear that the EFE objective does not serve as a valid and consistent bound on either of the divergence or the evidence objectives. Similarly, we can express the EFE directly in terms of the divergence objective as follows,

\begin{align*}
    \underbrace{\KL[p(o) || \tilde{p}(o)]}_{\text{Divergence Objective}} &= \mathbb{E}_{q(x | o)p(o)}\KL[p(o)q(o,x) || \tilde{p}(o)q(o,x)] \\
    &= \mathbb{E}_{q(x | o)p(o)}\KL[p(o)q(o|x)q(x) || \tilde{p}(o)q(x | o)q(o)] \\
    &= \underbrace{\mathbb{E}_{q(x | o)p(o)}[\ln q(x) - \ln \tilde{p}(o) - \ln q(x | o)]}_{\text{EFE}} + \underbrace{\mathbb{E}_{p(o)}\KL[q(x | o) || q(x)]}_{\text{Information Gain}} - \underbrace{\mathbb{H}[p(o)]}_{\text{Marginal Entropy}} \numberthis
\end{align*}
Where we see that the divergence objectives simply is the EFE plus an information gain term, minus the marginal or predicted entropy term. As such, even within this framework, the mathematical origin and the behaviour of the EFE remains unclear since the EFE does not form consistent bounds on either objective, but instead oscillates above and below both.

\subsection{Action and Perception as Divergence Minimization}
A recent framework, inspired by active inference and advances in deep reinforcement learning, which aims to unify perception and action under a single framework is Action and Perception as Divergence Minimization \citep{hafner2020action} (APDM). This framework proposes that both action and perception can be modelled as an agent trying to mininimize a divergence functional between two distributions an `actual' distribution $A(x,o)$, and a target distribution $T(x,o)$. 

\begin{align*}
    \mathcal{L}_{APDM} &= \KL[A(x,o) || T(x,o)] \\
    &= \underbrace{\mathbb{E}_{A(x)}\KL[A(o | x) || T(o)]}_{\text{Realizing Latent Preferences}} - \underbrace{\mathbb{E}_{A(x,o)}[\ln T(x | o) - \ln A(x)]}_{\text{Information Bound}} \numberthis
\end{align*}

\begin{align*}
    -\underbrace{\mathbb{E}_{A(x,o)}[\ln T(x | o) - \ln A(x)]}_{\text{Information Bound}} &= -\mathbb{E}_{A(x,o)}[\ln T(x | o) - \ln A(x) + \ln A(x | o) - \ln A(x | o)] \\
    &= - \underbrace{\mathbb{E}_{A(o)}\KL[A(x | o) || A(x)]}_{\text{Information Gain}} + \underbrace{\mathbb{E}_{A(o)}\KL[A(x |o) || T(x | o)]}_{\text{Posterior Divergence}} \numberthis
\end{align*}
By expressing this bound explicitly, we can see how it is an upper bound on the information gain, since the posterior divergence is always positive. The tightness of the bound then depends on how closely the actual and target distributions match. In general, we can use this approach to write out a full expression for the divergence objective between two joint distributions over both observations and latent variables.
\begin{align*}
    \mathcal{L}_{joint} &= \underset{a}{argmin} \, \KL[p(o,x) || \tilde{p}(o,x)] \\
    &= \underbrace{\mathcal{E}_{p(x)}\KL[p(o| x) || \tilde{p}(o)]}_{\text{Likelihod Divergence}} - \underbrace{\mathbb{E}_{p(o,x)}[\ln \tilde{p}(x | o) - \ln p(x)]}_{\text{Information Bound}} \\
    &= \underbrace{\mathbb{E}_{p(x)}\KL[p(o| x) || \tilde{p}(o)]}_{\text{Likelihod Divergence}} -  \underbrace{\mathbb{E}_{p(o)}\KL[p(x | o) || p(x)]}_{\text{Information Gain}} + \underbrace{\mathbb{E}_{p(o)}\KL[p(x |o) || \tilde{p}(x | o)]}_{\text{Posterior Divergence}} \numberthis
\end{align*}
 
In effect, we see that minimizing the divergence between two joint distributions requires the minimizations of both the likelihood divergence \emph{and} the posterior divergence, while also requiring the maximization of the information between posterior and prior of the first term in the joint KL.

It is also straightforward to relate this joint divergence to the divergence objective, which is the divergence between marginals instead of joints. 
\begin{align*}
    \underbrace{\KL[p(o,x) || \tilde{p}(o,x)]}_{\text{Joint Divergence}} &= \KL[p(o)p(x|o)||\tilde{p}(x |o)\tilde{p}(x|o)] \\
    &= \underbrace{\KL[p(o) || \tilde{p}(o)]}_{\text{Divergence Objective}} + \underbrace{\mathbb{E}_{p(o)}\KL[p(x | o)||\tilde{p}(x|o)]}_{\text{Posterior Divergence}} \\
    &\geq \underbrace{\KL[p(o) || \tilde{p}(o)]}_{\text{Divergence Objective}} \numberthis
\end{align*}
Since the posterior divergence is always positive (as a KL divergence), we observe that the joint divergence is simply an upper bound on the divergence objective. Since the divergence is minimized, this bound is in the correct direction, and thus minimizing the joint divergence is a reasonable proxy for minimizing the marginal divergence objective. By minimizing the joint, it implicitly encourages agents to minimize both the marginal divergence as well as the divergence between the predicted and desired posterior distributions.

While the generic APDM divergence, as just a divergence of joints, is straightforwardly an upper bound on the divergence objective, we show that under the common definitions of the actual and target distributions, the APDM divergence can also be understood as a lower bound on the evidence objective, thus providing a bridge between the two objectives. Although the actual and target distributions can be defined differently depending on the objective you desire to reproduce, one canonical form of the actual and target distributions, which can reproduce control as inference as well as variational perceptual inference is as follows. We define the actual distribution to be the combination of the `real' data distribution $p(o)$ and also a variational belief distribution $q(x | o)$ such that $A(o,x) = q(x | o)p(o)$. Similarly, we define the target distribution to be the product of the agent's veridical generative model $p(o,x)$ and the exogenous desire distribution $\tilde{p}(o)$ such that $T(o,x) = p(o,x)\tilde{p}(o)$. This target distribution is valid as long as the ultimate objective is optimized via gradients of the divergence, which does not require that the target distribution be normalized. Under this definition of the actual and target distributions, the APDM objective becomes,

\begin{align*}
    \mathcal{L}_{APDM} = \KL[q(x | o)p(o) || p(o,x)\tilde{p}(o)] \numberthis
\end{align*}
In the case of known observations in the past, we assume that the data distribution becomes points around the actually observed observations $p(o) = \delta(o = \hat{o})$ while the desire distribution becomes uniform -- as there is little use for control in having desires about the unalterable past. Under these assumptions, the APDM objective simply becomes the ELBO or the negative free energy, thus replicating perceptual inference. However, on inputs in the future, the data distribution becomes a function of action (since actions can change future observations) and the desire distribution becomes relevant, thus allowing the minimization of the APDM functional to underwrite control. To gain a better intuition for the interplay of perception and control in the APDM functional, we showcase the following decomposition,
\begin{align*}
    \mathcal{L}_{APDM} &= \KL[q(x | o)p(o) || p(o,x)\tilde{p}(o)] \\
    &= \underbrace{\mathbb{E}_{p(o)}\KL[q(x | o) || p(o,x)]}_{\text{ELBO}} + \underbrace{\KL[p(o)||\tilde{p}(o)]}_{\text{Divergence Objective}} \numberthis
\end{align*}

which demonstrates that the APDM objective effectively unifies action and perception by summing together a perceptual objective (VFE) with the divergence objective for control. This confirms the previous finding that the APDM objective forms an upper bound on the divergence objective since the ELBO, as a KL divergence, is bounded below by 0. We also observe that this form of the APMD objective is also approximately a lower bound on the expected evidence objective, thus providing a link between the two objectives.
\begin{align*}
    \mathbb{E}_{p(o | a)}[\ln \tilde{p}(o)] &= \mathbb{E}_{p(o | a)}[\ln \int dx \, \tilde{p}(o,x)] \\
    &= \mathbb{E}_{p(o | a)}[\ln \int dx \, \frac{\tilde{p}(o,x)q(x|o)p(o,x)}{q(x | o)p(o,x)}] \\
    &\geq \mathbb{E}_{p(o)q(x | o)}[\ln \frac{\tilde{p}(o,x)p(o,x)}{q(x|o)p(o,x)}] \\
    &\geq \mathbb{E}_{p(o)q(x | o)}[\ln \frac{\tilde{p}(o)\tilde{p}(x|o)p(o,x)}{q(x|o)p(o)p(x|o)}] \\
    &\geq -\underbrace{\mathbb{E}_{p(o)q(x | o)}[\KL[q(x|o)p(o)||p(o,x)\tilde{p}(o)]}_{\text{APDM Objective}} +\underbrace{\mathbb{E}_{p(o)q(x|o)}[\ln \tilde{p}(x|o) - \ln p(x|o)]}_{\text{Posterior Divergence Bound}}\\
    &\approx \geq \underbrace{\mathbb{E}_{p(o)q(x | o)}[\KL[q(x|o)p(o)||p(o,x)\tilde{p}(o)]}_{\text{APDM Objective}} \numberthis
\end{align*}

Which is approximately equal to the APDM objective under the condition that the posterior divergence bound between desire posterior $\tilde{p}(x | o)$ and true posterior $p(x | o)$ is small.

\section{Towards a General Theory of Mean-Field Variational Objectives for Control}

Now that we understand the division of objectives for control into two classes of evidence and divergence functionals, we can start to try to understand the full possibilities of the space of potential objectives. Here, in this final section of Chapter 5, we try to present precisely such a taxonomy. We focus specifically on `mean-field' variational objective functionals, meaning that we can split the objective into a number of independent objectives for each time-step of a trajectory which can be minimized independently. Crucially, such a mean-field assumption is also made in the traditional reinforcement learning paradigm, where it is a necessary precondition for the Bellman equation, and also is standard in the control as inference framework as well.

While the division into divergence and evidence objectives is clearly important, it cannot be the full story. Recall that one of the key differences between control as inference and active inference discussed earlier, was not just the objective of the EFE vs the FEF, but also the way value was encoded into the inference procedure. Control as inference encoded value via an additional set of `optimality variables' which were augmentations to the graphical model, and did not interact with any previously existing variables. Active inference, by contrast, encoded value directly through a biased generative model of the observations \footnote{Active inference can also be formulated with biased states, see  \citep{da2020active}}. We call the method used by control as inference, which does not affect any currently existing variable an \emph{exogenous} encoding of value, while the method used by active inference, since it encodes value directly into the model itself, we call an \emph{endogenous} encoding. Exogenous and Endogenous encodings of value provide an orthogonal dimension of objective variablility on top of the evidence-divergence dichotomy, since clearly one can have an evidence, or a divergence objective with both an endogenous or an exogenous value encoding. Finally, the actual specifics of the generative model used clearly affects the variational objective irrespective of whether it is an evidence or divergence objective, or uses an endogenous or exogenous value encoding. For instance, whether we consider latent states, or various different types of model parameters in the generative model leads to a different objective functional. 

We thus see that we can break down the landscape of potential mean field objective functionals for control into three orthogonal dimensions. 
\begin{itemize}
    \item Whether an evidence or divergence functional is used.
    \item Whether value is encoded exogenously, or endogenously.
    \item The generative model underlying the objective functional.
\end{itemize}

Under different values for each of these dimensions, the objective functional that is specified changes in a straightforward and principled manner. Thus, our scheme allows the direct derivation of any given functional, and an understanding of its decompositions, and hence the behaviour it induces, for any choice of these variables. While we have covered the effect of choosing an evidence or a divergence functional previously, we have not yet been explicit about the effect of the other two dimensions. In this section, we explore the effects of these additional dimensions of design choice in more detail.

\subsection{Encoding Value}

However, the need to encode goals or desires into the inference procedure immediately introduces design choices of how exactly this is to be done. We argue that these design choices can first be split along two orthogonal axes -- firstly, whether goals are encoded exogenously as an additional input to the inference process, or endogenously through fundamentally biasing one or more aspects of the inference model. The second axis of variation is whether goals and desires enter the inference procedure through the generative model or the variational distribution. Making different design choices here produces a variety of different variational algorithms for control.

If goals are encoded through the generative model, then whether the goals are encoded exogenously or endogenously is the primary distinction between the formalisms of control-as-inference and active inference. On the other hand, encodings goals through the variational model instead leads to novel algorithms which generally have not been much explored in the literature. Encoding goals exogenously through the variational distribution leads to a variational bound similar to control-as-inference but with an extrinsic value term with a reversed-KL-divergence which thus exhibits mode-seeking rather than mean-seeking behaviour, which is related to pseudolikelihood methods \citep{peters2007reinforcement} in variational reinforcement learning. Encoding endogenous goals through the variational distribution leads to a novel class of reverse-active-inference algorithms which minimize a variational divergence between a biased approximate posterior and a veridical generative model. 

\subsubsection{Exogenous Value: Maximum-Entropy RL}

To encode goals exogenously into the generative model, we must augment the POMDP graphical model with additional optimality variables $\Omega_{t:T}$. The idea here is that the optimality variables are binary bernoulli variables which mark whether a trajectory is optimal from the current state where the probability of optimality is often set to the exponentiated reward $p(\Omega_t=1) \propto exp(r_t)$ so that $\ln p(\Omega_t = 1) = r_t$. Adaptive actions are then inferred by first assuming that the agent has acted optimally into the future, and then inferring the actions that would be consistent with that belief -- i.e. we wish to infer the posterior $p(a_{t:T} | x_{t:T}, \Omega_{t:T}=1)$. This posterior can then be approximated by minimizing the augmented variational bound:
\begin{align*}
    \mathcal{F}_\Omega &= \KL \Big( q(x_t, a_t | o_t) \Vert p(o_t, x_t, a_t, \Omega_t) \Big) \geq \KL \Big(q(x_t, a_t | o_t) \Vert p(o_t, x_t, a_t | \Omega_{t:T}) \Big) \numberthis
\end{align*}

By splitting apart this bound into its constituent parts, we can investigate the expected behaviour of agents which act so as to minimize the bound.
\begin{align*}
    \mathcal{F}_\Omega &= \KL \Big(q(x_t, a_t | o_t) \Vert p(o_t, x_t, a_t, \Omega_t) \Big) \\
    &= \KL \Big( q(a_t | x_t)q(x_t | o_t) \Vert p(\Omega_t | x_t, a_t)p(o_t | x_t)p(a_t | x_t)p(x_t | x_{t-1}, a_{t-1}) \Big) \\ 
    &= \underbrace{- \mathbb{E}_{q(x_t,a_t | o_t)}\big[\ln p(\Omega_t | x_t, a_t) \big]}_{\text{Extrinsic Value}} - \underbrace{\mathbb{E}_{q(x_t,a_t | o_t)}\big[\ln p(o_t | x_t) \big]}_{\text{Observation Ambiguity}} + \underbrace{\mathbb{E}_{q(x_t | o_t)}\big[ \KL \Big(q(a_t | x_t) \Vert p(a_t | x_t) \Big) \big]}_{\text{Action Divergence}} \\ &+ \underbrace{\KL \Big(q(x_t | o_t) \Vert p(x_t | x_{t-1}, a_{t-1}) \Big)}_{\text{State Divergence}} \numberthis
\end{align*}

The bound thus splits into four separate and identifiable terms -- extrinsic value, observation ambiguity, action divergence, and state divergence. The first extrinsic value term corresponds to the expected external rewards given by the environment. This is due to the definition of optimality that $\ln p(\Omega_t | x_t, a_t) \triangleq r(x_t, a_t)$ so that the extrinsic value is simply the expected reward of a given state-action pair. By minimizing the negative expected reward, we wish to maximize the expected reward on a given time-step. This is identical to the standard reinforcement learning objective of reward maximization so that if $\mathcal{F}_\Omega$ only contained the extrinsic value term, it would be exactly equivalent to reinforcement learning except that the expectation is taken with respect to the agent's beliefs over states and actions rather than the true environmental transition dynamics.

The observation ambiguity term $\mathbb{E}_{q(x_t,a_t | o_t)}\big[\ln p(o_t | x_t) \big]$ grants a bonus to agents for reaching areas of state-space with a high expected likelihood, that is areas where the state-observation mapping is highly precise. In effect, if the agent must learn a likelihood mapping,  this discourages exploration by granting bonuses for staying in regions already well characterised. This term only arises in the POMDP setting due to the addition of a likelihood term in the generative model. The third term is the action divergence which is to be minimized, and penalizes the agent for the divergence between its variational policy $q(a_t | x_t)$ and its prior policy $p(a_t | x_t)$. If the prior policy is assumed to be uniform such that $p(a_t | x_t) \triangleq \frac{1}{|\mathcal{A}|}$ where $|\mathcal{A}|$ is the cardinality of the action space in discrete action-spaces, and $p(a_t | x_t) \triangleq \textit{Unif}(a_{min},a_{max})$ in continuous action spaces with a minimum and maximum action value, then this action divergence term reduces to the negative expected action entropy $-\mathbb{E}_{q(x_t | o_t)}\big[\mathcal{H}[q(a_t | x_t)]\big]$. The final term is the state divergence, so that the agent tries to minimize the divergence between its variational posterior over the state and the prior state expected under the generative model. If the transition model is learnt, this leads it to prioritising transitions with known dynamics, again causing the agent to primarily confine itself to regions of the state-space it has already modelled well. In the MDP setting, without observations, this term vanishes since the variational posterior $q(x_t | o_t)$ becomes the variational prior $q(x_t)$ which is often assumed equal to the generative prior $p(x_t | x_{t-1},a_{t-1})$. Thus in the case of an MDP with a uniform action prior we obtain the maximum-entropy RL objective:
\begin{align*}
    \mathcal{F}_{maxent} = - \mathbb{E}_{q(x_t,a_t)}\big[\ln p(\Omega_t | x_t, a_t) \big] - \mathbb{E}_{q(x_t)}\big[\mathcal{H}[q(a_t | x_t)]\big] \numberthis
\end{align*}

Which induces agents both to maximize expected rewards while also maximizing the policy entropy. Intuitively this means that the agent should try to maximize rewards while acting as randomly as possible (maximizing entropy). This policy entropy term thus provides a crude bonus for random exploration and often helps prevents the commonly-observed phenomenon of `policy collapse' \citep{fujimoto2018addressing} whereby reinforcement learning agents will often rapidly learn to put all probability mass onto a single action, thus preventing other actions from being taken, hindering exploration and long-term performance. Interestingly, when extended to the POMDP case with learnt transition and likelihood models, this formalism gives rise to additional observation-ambiguity and state-divergence terms which serve to further disincentivise exploration by penalising moving too far from the prior predicted state and giving bonuses for highly predictable likelihoods. Intuitively we can think of these extra terms as trying to do away with the additional uncertainty induced by the POMDP setting, so the agent confines itself to the region which is as close to an MDP as possible.

\subsubsection{Endogenous Value: Active Inference}

While maximum entropy reinforcement methods can be derived by encoding goals exogenously to the generative model through the use of additional `optimality variables', it is also possible to encode goals \textit{endogenously} by directly biasing some aspect of the generative model towards preferred outcomes. Intuitively, we can think of the difference as follows. With exogenous optimality variables, we possess a veridical generative model outputting the likely and unbiased trajectories for a given series of actions. We then `shift' these trajectories to converge on the goal by conditioning on the optimality variables, and then infer the actions consistent with the shifted trajectories. By contrast, active inference endogenously encodes goals by biasing the model so that instead of a veridical generative model which generates trajectories that are then shifted, instead we have a \emph{biased} generative model which directly outputs a biased trajectory of observations $\Tilde{o}_{t:T}$ converging on the goal, which can then be used to infer the actions consistent with this biased trajectory. In this manner, instead of proposing additional optimality variables, we directly posit a biased generative model $\tilde{p}(o_t, x_t, a_t)$ and optimise the biased variational bound:
\begin{align*}
    \mathcal{F}_{likelihood-AIF} &= \KL \Big( q(x_t, a_t | o_t) \Vert \tilde{p}(o_t, x_t, a_t) \Big) \\
    &= \KL \Big( q(a_t | x_t)q(x_t | o_t) \Vert \tilde{p}(o_t | x_t) p(x_t | x_{t-1},a_{t-1})p(a_t | x_t)  \Big) \\
    &= - \underbrace{\mathbb{E}_{q(x_t | o_t)} \big[\ln \tilde{p}(o_t | x_t) \big]}_{\text{Extrinsic Value}} + \underbrace{\mathbb{E}_{q(x_t | o_t)}\big[ \KL \Big(q(a_t | x_t) \Vert p(a_t | x_t) \Big) \big]}_{\text{Action Divergence}} \\ &+ \underbrace{\KL \Big(q(x_t | o_t) \Vert p(x_t | x_{t-1}, a_{t-1}) \Big)}_{\text{State Divergence}} \numberthis
\end{align*}

This variational bound decomposes into three terms as can be seen above. The first, extrinsic value, is such because it is the biased probability of observations expected under the variational distribution. If we assume that the biased generative model is influenced by the external rewards in the environment, to allow for consistency with reinforcement learning and the maximum-entropy RL framework such that $\ln \tilde{p}(o_t | x_t) \propto exp(r(x_t)$ then this first term reduces to minimizing the expected sum of negative rewards, or maximising expected rewards. The other two action-divergence and state-divergence terms are equivalent to the terms in the maximum entropy bound above, except that the active inference bound is lacking the `observation ambiguity' term. This is because exogenously encoding goals adds an additional set of optimality variables to the variational bound, thus maintaining a distinction between veridical observations and biased optimality variables, while endogenously encoding goals needs to `hijack' at least one degree of freedom in the bound in order to encode the goals directly. Here the observation ambiguity term has effectively been hijacked by being biased with reward to instead encode the extrinsic value.

Interestingly, the choice of encoding goals endogenously also gives an additional set of choices of how to bias the generative model. The key choice is between having a biased likelihood, as done above, or a biased marginal and posterior. This decomposition is shown below:
\begin{align*}
     \mathcal{F}_{marginal-AIF} &= \KL \Big( q(x_t, a_t | o_t) \Vert \tilde{p}(o_t, x_t, a_t) \Big) \\
    &= \KL \Big( q(a_t | x_t)q(x_t | o_t) \Vert \tilde{p}(o_t)\tilde{p}(x_t | o_t) p(a_t | x_t)  \Big) \\
    &= - \underbrace{\mathbb{E}_{q(x_t | o_t)} \big[\ln \tilde{p}(o_t) \big]}_{\text{Extrinsic Value}} + \underbrace{\mathbb{E}_{q(x_t | o_t)}\big[ \KL \Big(q(a_t | x_t) \Vert p(a_t | x_t) \Big) \big]}_{\text{Action Divergence}} + \underbrace{\KL \Big(q(x_t | o_t) \Vert \tilde{p}(x_t | o_t) \Big)}_{\text{Biased State Divergence}} \numberthis
\end{align*}

Here, we assume that $\ln \tilde{p}(o_t) \propto exp(r(o_t))$ such that the extrinsic value term is again directly equal to the rewards. An interesting difference is that the biased state-divergence is now between the variational posterior and the biased state posterior, which represents the posterior over states that would be observed given the biased observations. This gives this second term also the flavour of an extrinsic value as the goal is not only to maximize rewards but match the veridicial variational distribution to the biased state distribution induced by the desired observations

Alternatively, it is also possible to consider encoding the bias into the \emph{states} $\tilde{p}(x_t)$ instead of the observations. In this case, we obtain the following objective functional:
\begin{align*}
\label{state_AIF}
         \mathcal{F}_{state-AIF} &= \KL \Big( q(x_t, a_t | o_t) \Vert \tilde{p}(o_t, x_t, a_t) \Big) \\
    &= \KL \Big( q(a_t | x_t)q(x_t | o_t) \Vert p(o_t | x_t) \tilde{p}(x_t) p(a_t | x_t)  \Big) \\
    &= \underbrace{\KL \Big( q(x_t | o_t) \Vert \tilde{p}(x_t) \Big)}_{\text{Extrinsic Value}} - \underbrace{\mathbb{E}_{q(x_t,a_t | o_t)}\big[\ln p(o_t | x_t) \big]}_{\text{Observation Ambiguity}} + \underbrace{\mathbb{E}_{q(x_t | o_t)}\big[ \KL \Big(q(a_t | x_t) \Vert p(a_t | x_t) \Big) \big]}_{\text{Action Divergence}} \numberthis
\end{align*}

In this case, we have regained the observation-ambiguity term and instead the state-divergence term has been `hijacked' by the rewards to become the extrinsic value term, which has become the divergence between predicted and desired states. 

Given that there exist these two design choices of whether to encode goals exogenously or endogenously and which lead to subtly different objectives, a natural question to ask is what are the relative advantages and disadvantages of the two methods? What trade-offs exist and where might each be useful? In general, the primary practical difference between the methods is that by endogenously encoding rewards, the agent loses a degree of freedom, which manifests itself as the loss of an observation ambiguity term if the goals are encoded through biased observations, or the loss of the state-divergence term if the goals are encoded as a desired state distribution. These additional terms tend to discourage exploration by causing the agent to remain in areas with known mappings, so that by disabling them endogenous goals tend to encourage exploration. On the other hand, by keeping to well-known regions where the POMDP behaves like an MDP, the exogenous goals method enforces a greater conservative bias towards safety, which may be especially useful in settings where exploration is costly, satisficing policies are easy to find, or in policies learnt in an offline or imitation-learning setting where venturing outside of the training distribution can have deleterious effects on the policy.

There are also important philosophical and representational differences between the two. On a representational note, although in the derivations above both the optimality variable and the biased generative model have been defined in terms of exponentiated reward, this is not necessarily the case. Since in the endogenous encoding case, the goals are encoded directly as prior distributions into the generative model, it is possible to model complex and potentially nonstationary goals distributions in this manner. However, due to the optimality variables being binary, and conditioned upon, this may constrain their representational power compared to the endogenous method, although in practice this difference may be negligible as although the variables themselves are binary, the probability of optimality being one, can be defined as an arbitrary function of the states, actions or observations. Thus in practice, there may be little representational difference between the two methods, except that the endogenous case has a slightly simpler intuitive justification as directly specifying a desired distribution over states or observations.

The difference between exogenous and endogenous encodings of value also has significant philosophical import. Exogenous encodings, by adding desires on top of an unbiased generative model maintain a clean distinction between veridical perception and action selection, and goals. This maintains the core modularity thesis of much work in artificial intelligence that perception, and action selection should be kept separate such that first a veridical world-model is constructed which tries to accurately model the world, then given a set of arbitrary goals, a general-purpose planner or policy can be utilised or learnt to enable the agent to achieve these goals. This approach corresponds to the classical perceive-(value)-plan-act cycle in cognitive science and maintains separate modules for a goal-agnostic perceptual system, a goal-agnostic planner or action-selection mechanism, and then a set of goals which are not intrinsic to the agent but which are constructed or handed-down from on high. Endogenous encoding methods, by contrast, tend to blur the boundaries between these sytems, since goals are encoded and adaptive actions are selected through a process of \emph{biased} perception and inference whereby an agent does not first infer a true trajectory, compares it to its goals, and then tries to match the too, instead it simply sees a biased trajectory leading to its goals and then acts consonantly with what it sees. This view has close links to embodied and enactivist views in philosophy and cognitive science which stress that rather than distinct modular systems of perception, valuation, and action there is instead a single combined system or sensorimotor loop which directly acts on sensorimotor contingencies in an adaptive fashion \citep{baltieri2018modularity}.

\subsubsection{Encoding Value into the Variational Distribution}

Previously, we have encoded values either exogenously or endogenously into the \emph{generative model} of the agent. In the case of endogenous encodings,  this means that the agent makes biased predictions, rather than forming biased inferences. However, it is also possible to consider and investigate what happens if instead values were encoded into the variational distribution so that the agent's inference procedure rather than model is biased. We first consider the case of exogenously encoded goals. This requires that the variational distribution is augmented with binary optimality variables, just like the generative model was previously giving $q(a_t, x_t,\Omega_t | o_t)$. From this we can write the relevant free energy functional:
\begin{align*}
\label{exogenous_goals_equation}
    \mathcal{F}_{q\Omega} &= \KL \Big( q(a_t, x_t, \Omega_t | o_t) \Vert p(o_t, x_t, a_t) \Big) \\
    &= \KL \Big( q(\Omega_t | x_t, a_t)q(a_t | x_t)q(x_t | o_t) \Vert p(o_t | x_t)p(a_t | x_t)p(x_t | x_{t-1},a_{t-1}) \Big) \\
    &=  \underbrace{\mathbb{E}_{q(x_t,a_t | o_t)}\big[\ln q(\Omega_t | x_t, a_t) \big]}_{\text{Extrinsic Value}} - \underbrace{\mathbb{E}_{q(x_t,a_t | o_t)}\big[\ln p(o_t | x_t) \big]}_{\text{Observation Ambiguity}} + \underbrace{\mathbb{E}_{q(x_t | o_t)}\big[ \KL \Big(q(a_t | x_t) \Vert p(a_t | x_t) \Big) \big]}_{\text{Action Divergence}} \\ &+ \underbrace{\KL \Big(q(x_t | o_t) \Vert p(x_t | x_{t-1}, a_{t-1}) \Big)}_{\text{State Divergence}} \numberthis
\end{align*}

We see that the resulting functional, under the assumption that $\ln q(\Omega | x_t, a_t) \propto exp(r(x_t, a_t)$, is exactly equivalent to the functional obtained for the exogenously encoded generative model, up to a sign difference in the extrinsic term which can be finessed without loss of generality by inverting the sign of the reward function. We thus see that, if values are exogenously encoded, it does not matter which distribution they are primarily encoded through. Intuitively this is because since the veridical distributions are maintained through endogenous coding, they are unaffected by the encoding of value into them, thus which one to choose has no effect overall up to a trivial sign difference which can be easily finessed through negatively encoding reward.

When rewards are endogenously encoded, however, the resulting functionals are not equivalent. We first show this with a biased-state functional which should be compared to Equation \ref{exogenous_goals_equation}, where we directly bias the state-inference part of the variational functional so as to preferentially infer being in desired states from a given observation $\tilde{q}(a_t, x_t | o_t) \triangleq q(a_t | x_t) \tilde{q}(x_t | o_t)$. Through this decomposition we obtain the functional:
\begin{align*}
    \mathcal{F}_{state-q-AIF} &= \KL \Big( \tilde{q}(x_t, a_t | o_t) \Vert p(o_t, x_t, a_t) \Big) \\
    &= \KL \Big( q(a_t | x_t)\tilde{q}(x_t | o_t) \Vert p(o_t | x_t) p(x_t | x_{t-1}, a_{t-1}) p(a_t | x_t)  \Big) \\
    &= \underbrace{\KL \Big( \tilde{q}(x_t | o_t) \Vert p(x_t | x_{t-1}, a_{t-1}) \Big)}_{\text{Extrinsic Value}} - \underbrace{\mathbb{E}_{q(x_t,a_t | o_t)}\big[\ln p(o_t | x_t) \big]}_{\text{Observation Ambiguity}} \\ &+ \underbrace{\mathbb{E}_{q(x_t | o_t)}\big[ \KL \Big(q(a_t | x_t) \Vert p(a_t | x_t) \Big) \big]}_{\text{Action Divergence}} \numberthis
\end{align*}

which is very similar to the corresponding state extrinsic value functional in Equation \ref{state_AIF}, except that in the extrinsic value term the divergence is between the biased variational posterior and a veridical generative prior, rather than a veridical variational posterior and a biased generative prior. By having the biases occur on the left side of the KL, we are essentially minimizing the reverse-KL compared to when value is encoded into the generative model. This gives the resulting agents a mode-seeking rather than a mean-seeking behaviour, since agents optimizing under the reverse KL will suffer a large penalty if the desires are in regions with a very low veridical probability. Moreover, there is a further subtle difference in the POMDP case here since we are contrasting the biased distribution with a prior rather than a posterior. However, in the MDP case this difference vanishes, and we obtain the simplified functional:
\begin{align*}
    \mathcal{F}_{qMDP} &= \KL \Big( \tilde{q}(a_t, x_t) \Vert p(x_t, a_t) \Big) \\
    &= \KL \Big( \tilde{q}(a_t | x_t)\tilde{q}(x_t) \Vert p(a_t | x_t)p(x_t | x_{t-1},a_{t-1}) \Big) \\
        &= \underbrace{\KL \Big( \tilde{q}(x_t) \Vert p(x_t | x_{t-1}, a_{t-1}) \Big)}_{\text{Extrinsic Value}} + \underbrace{\mathbb{E}_{q(x_t | o_t)}\big[ \KL \Big(q(a_t | x_t) \Vert p(a_t | x_t) \Big) \big]}_{\text{Action Divergence}} \numberthis
\end{align*}

Which is lacking the observation ambiguity term due to being an MDP, and also the extrinsic value has become the divergence between desired variational states and predicted generative states. This functional is exactly equivalent to the alternate formula with values encoded endogenously into the generative model except that it uses the reverse KL divergence. Moreover these functionals are deeply related to KL control with an additional action divergence term, and thus when value is instead encoded into the variational distribution we have reverse-KL control which uses the reverse KL divergence and is very closely related to pseudo-likelihood methods in reinforcement learning \citep{abdolmaleki2018maximum,peters2007reinforcement}.

Overall then we have explored two orthogonal axes of variation for the problem of how to encode a notion of value, reward, or desires into an otherwise value-agnostic variational inference procedure in order to be able to infer adaptive actions. We have shown that the first question of whether value is to be encoded exogenously or endogenously makes subtle but significant differences in the resulting functionals. Specifically, by requiring the utilisation and biasing of one of the variables in the model to encode value, endogenous encoding tends to lose one degree of freedon in its functional compared to exogenous encoding. In the POMDP setting this is typically the observation-ambiguity term if goals are encoded into observations, or the state-divergence term if goals are encoded into states. Beyond this, we see that these different means of encoding value also has significant philosophical importance as to the nature of perception, action and value. Exogenously encoding goals supports a modular description of these three functions as independent systems which are each agnostic with respect to the outputs of the others, while endogenously encoding goals merges them all together into a mixed system where action and perception are intrinsically biased by goals towards adaptive outcomes. Moreover, this dichotomy of exogenous or endogenous encoding is the primary difference between variational control frameworks arising from reinforcement learning and from active inference, and this fact speaks to the difference in underlying cognitive philosophy between these theories where reinforcement learning draws heavily from cognitivist and representational traditions in artificial intelligence which prize principled, independent, and modular systems, while active inference comes from a heavily embodied and enactive viewpoint influenced by dynamical systems theory whereby systems are seen primarily in terms of their situatedness within a action-perception sensorimotor loop, and there is not necessarily any clean distinction between phases or subsystems of this loop.

Finally, we have seen that the second axis of variation is whether the goals are encoded into the generative model or the variational distribution. The effects of this difference are subtler than for the exogenous vs endogenous encoding dichotomy. Variational and generative encodings are equivalent up to a sign difference in the exogenous case since the encoding does not affect the veridicality of the distribution, while in the endogenous case this causes subtle differences such that the generative and variational encodings are typically equivalent up to the extrinsic value terms which becomes the reverse-KL for the variational encoding relative to the generative. This leads to a close connection with pseudolikelihood methods in reinforcement learning \citep{abdolmaleki2018maximum} and in fact provides a generalisation of these methods to the POMDP setting.

\subsection{General Graphical Models}

In all previous work, we have primarily focused on how the algorithms and functionals differ when placed into a POMDP setting with visible observations and unobserved (Markovian) hidden states, and actions which affect the hidden states. However this is fundamentally a modelling choice. For instance, if state information is perfectly observed, as is often assumed in RL, then the problem reduces to a simple MDP formulation. The MDP formulation with exogenous rewards was addressed in Equation \ref{exogenous_goals_equation}, however when encoding rewards endogenously, this must be encoded into the generative or variational distributions, so that with the FEEF objective functional the extrinsic term becomes simply a state divergence.
\begin{align*}
    FEEF_{MDP} &= \KL \Big( q(x_t, a_t) \Vert \tilde{p}(x_t, a_t) \Big) 
    \\ &= \underbrace{\KL \Big( q(x_t) \Vert \tilde{p}(x_t) \Big)}_{\text{Extrinsic Value}} + \underbrace{\mathbb{E}_{q(x_t)}\big[ \KL \Big( q(a_t | x_t) \Vert p(a_t | x_t) \Big) \big]}_{\text{Action Divergence}} \numberthis
\end{align*}

Here no information gain is possible since states are directly observed, thus the only exploration is through random entropy maximizing terms. Importantly, if the goals were instead encoded into the variational distribution, the only effect this would have would be to flip the extrinsic value KL into a reverse-KL and thus induce mode-seeking rather than mean-seeking distribution matching behaviour. In the MDP setting, then, we obtain a maximum-entropy KL-control objective.

While it is possible to restrict the generative models only to MDPs, we can also consider extending them to also explicitly model the prior and posterior distributions of parameters underlying the variational and generative distributions. For instance, we have implicitly been utilising a variational posterior $q(x_t | o_t)$, a transition model $p(x_t | x_{t-1},a_{t-1})$, a likelihood model $p(o_t | x_t)$, an action policy $q(a_t | x_t)$ and an action prior $p(a_t  |x_t)$. All of these distributions may have parameters, or be parameterised by a flexible function approximator such as a neural network, which itself has parameters. These parameters can be be included in the variational inference procedure by adding a generative model and variational distribution over the parameters. For instance, supposing we have some set of parameters $\theta$ which parametrise the transition model such that the transition model becomes $p(x_t | x_{t-1},a_{t-1};\theta)$, then we can pose an inference problem and write down a variational free energy functional which also includes a model over parameters:
\begin{align*}
    \mathcal{F}_\theta &= \KL \Big( q(x_t,a_t, \theta_t | o_t) \Vert \tilde{p}(o_t, x_t, a_t, \theta_t) \Big) \\
&= \KL \Big( q(a_t | x_t)q(\theta_t | x_t)q(x_t | o_t) \Vert \tilde{p}(o_t | x_t)p(a_t | x_t)p(x_t | x_{t-1},a_{t-1},\theta_t)p(\theta_t) \Big) \\
&= - \underbrace{\mathbb{E}_{q(x_t | o_t)} \big[\ln \tilde{p}(o_t | x_t) \big]}_{\text{Extrinsic Value}} + \underbrace{\mathbb{E}_{q(x_t | o_t)}\big[ \KL \Big(q(a_t | x_t) \Vert p(a_t | x_t) \Big) \big]}_{\text{Action Divergence}} \\ &+ \underbrace{\mathbb{E}_{q(\theta_t)} \big[ \KL \Big(q(x_t | o_t) \Vert p(x_t | x_{t-1}, a_{t-1},\theta) \Big) \big]}_{\text{State Divergence}} + \underbrace{\KL \Big( q(\theta_t | x_t) \Vert p(\theta_t) \Big)}_{\text{Parameter Divergence}} \numberthis
\end{align*}

We thus see that minimizing the variational free energy directly, requires minimizing the divergence between posterior and prior beliefs over the parameters, thus implicitly penalising updates which cause large parameter updates, and thus disincentivising exploration. In general, by adding additional variables to the variational free energy, we simply obtain additional divergence terms to be minimized as above, essentially penalising deviations between the posterior and prior beliefs for that variable. However, and analogously with the states, when we utilize a FEEF objective functional, we can obtain a parameter information gain term as well as a countervailing parameter approximation error term as shown below:
\begin{align*}
        FEEF_\theta &= \KL \Big( q(o_t, x_t, a_t,\theta_t) \Vert \tilde{p}(o_t, x_t, a_t,\theta_t) \Big) \\
    &= \KL \Big( q(o_t | x_t)q(a_t | x_t)q(x_t | \theta)q(\theta_t) q(x_t | o_t)q(\theta_t | x_t) \\ &\Vert \tilde{p}(o_t | x_t) p(a_t | x_t)p(x_t | x_{t-1},a_{t-1},\theta_t) p(\theta_t) q(x_t | o_t) q(\theta_t | x_t) \Big) \\
    &= \underbrace{\mathbb{E}_{q(a_t, x_t)}\big[ \KL \Big( q(o_t | x_t) \Vert \tilde{p}(o_t | x_t) \Big) \big]}_{\text{Extrinsic Value}} + \underbrace{\mathbb{E}_{q(o_t, x_t)}\big[ \KL \Big( q(a_t | x_t) \Vert p(a_t | x_t) \Big) \big]}_{\text{Action Divergence}} \\ &- \underbrace{\mathbb{E}_{q(o_t)}\big[ \KL \Big( q(x_t | o_t) \Vert q(x_t) \Big) \big]}_{\text{Expected Information Gain}} \\ &+ \underbrace{\mathbb{E}_{q(o_t)q(\theta_t)}\big[ \KL \Big( q(x_t | o_t) \Vert p(x_t | x_{t-1},a_{t-1},\theta_t) \Big) \big]}_{\text{Expected Posterior Divergence}} - 
    \underbrace{\KL \Big( q(\theta_t | x_t) \Vert q(\theta_t) \Big)}_{\text{Parameter Information Gain}} \\ &+ \underbrace{\KL \Big( q(\theta_t | x_t) \Vert p(\theta_t) \Big)}_{\text{Parameter Divergence}} \numberthis
\end{align*}

In the general case adding additional variables to the FEEF objective will create an information gain term in that variable as well as the counteracting posterior divergence term. Analogously, adding additional variables to the class of divergence functionals will also create information gain terms in those variables without the posterior divergence. The reason for this behaviour can ultimately be derived directly from the variational marginal entropy by considering augmenting it with parameters $\theta$:
\begin{align*}
    VME &= \mathbb{E}_{q(o_t)}\big[ \ln q(o_t) \big] = \mathbb{E}_{q(o_t, x_t, \theta_t)}\big[ \ln q(o_t) \big] \\
    &=  \mathbb{E}_{q(o_t, x_t, \theta_t)}\big[ \ln \frac{q(o_t, x_t, \theta_t)}{q(x_t, \theta_t | o_t)}\big] \\
    &= \mathbb{E}_{q(o_t, x_t, \theta_t)}\big[ \ln \frac{q(o_t | x_t)q(x_t | \theta_t)q(\theta_t)}{q(x_t | o_t)q(\theta_t | x_t)}\big]
    \\
    &= -\underbrace{\mathcal{H}\big[q(o_t | x_t) \big]}_{\text{Likelihood Entropy}} - \underbrace{\mathbb{E}_{q(o_t)q(\theta_t)}\big[ \KL \Big( q(x_t | o_t) \Vert q(x_t | \theta_t) \Big) \big]}_{\text{Expected State Information Gain}}
     - \underbrace{\KL \Big( q(\theta_t | x_t) \Vert q(\theta_t) \Big)}_{\text{Parameter Information Gain}} \numberthis
\end{align*}

Thus we have seen that by implicitly maximizing the marginal entropy, we in fact are implicitly optimising the information gain for any latent variables in the model. Similarly, as discussed previously, the primary difference between exogenous and endogenous encoding of value is that endogenous encodings lack a degree of freedom that exogenous encodings can make use of. In general therefore, as we augment the graphical models these functionals are derived from with additional variables, we see that the exogenous encoding is roughly equivalent to the endogenous encoding with an additional degree of freedom. This thus derives two `scaling laws' for our families of functionals as additional sets of variables are added to the functional -- that the exogenous encoding will approximate the endogenous encoding with an additional degree of freedom, and that with VFE derived functionals we will obtain divergence terms to be minimized between the posterior and prior for the variable, with a FEEF functional we will obtain an information gain term and a posterior approximation error term, while with a divergence based functional we will obtain an information gain term only in the new variable.

\section{Discussion}

In this chapter, we have answered and discussed two key questions. Firstly, we now understand the mathematical origin of information-seeking exploration terms in variational objectives for control. While this appears arcane, this is actually a question with deep implications both philosophically, as well as for applications. On a philosophical and mathematical note, we have uncovered the principled mathematical origin of information-maximizing exploration. We see that it arises from minimizing \emph{divergences} between a predicted and a desired distribution, and specifically from the predicted entropy maximization half of the divergence objective. In effect, we see that it is by trying to maximize the entropy of future observations, that induces information seeking exploration in latent space, whenever any latent states or parameters are added to the generative model. This makes intuitive sense -- if the future is broad and entropic, it contains much information, and to maximize that breadth entails learning about the world in order to ensure a precise match everywhere between predicted distribution and a complex desire distribution. Conversely, we have seen that standard evidence objectives used in variational inference or control result in effectively trying to match the predicted and desired distribution while trying to \emph{minimize} the entropy of future states or, alternatively, to make the future maximially predictable as well as in conformity to the agent's desires. As such, the agent's goal is to effectively minimize the amount of information it receives about the world, learning only as much as is sufficient for control, and thus not giving rise to any kind of information-seeking exploratory behaviour. We can thus understand precisely why standard objectives such as control as inference are insufficient to obtain information seeking behavioural objectives. Similarly, our approach allows us to understand and rationalize a number of approaches in the literature \citep{sun_planning_2011,oudeyer2009intrinsic,klyubin2005empowerment} which add additional exploratory terms to their reward-maximizing agents, with the heuristic justification that they should increase exploration. Indeed, many such approaches are implicitly minimizing a divergence objective without realizing it. Our advances here make this practice explicit and allows one to understand the precise mathematical nature of the objective being optimized. Moreover, this distinction also sheds light on the possible objective functions used by biological creatures such as humans in psychological or behavioural economics tasks, for instance, one can understand the otherwise-puzzling phenomenon of probability matching, as the inevitable outcome of optimizing a divergence functional, and this elegantly explains both why it is present and additionally, why it is beneficial -- since optimizing this objective in more complex environments naturally leads to exploratory information-seeking behaviour which often will outperform pure reward maximization even on its own terms.

This approach is also useful for applications, since we can begin by deriving a variety of methods using divergence objectives and understanding their exploratory behaviour. While little work has yet been done on explicitly divergence minimizing agents, we believe that this will be an important area in the future, as successful protocols and algorithms for exploration in reinforcement learning will become increasingly important as the sparsity, dimensionality, and difficulty of RL benchmark tasks increases.

Secondly, we understand the relationship between control as inference and active inference. We know that the key difference is simply a difference in objective function between the two (the EFE vs the FEF), and secondly a distinction between them is their encodings of value -- that active inference uses an endogenous, while control as inference uses an exogenous encoding. Moreover, understanding the distinctions between the two theories, as well as the general and broad distinction between evidence and divergence objectives, then allows us to raise our eyes and perceive a much larger vista of the full landscape of potential objective functionals for control. In the latter half of this chapter, we have seen that these functionals can vary along two orthogonal dimensions -- whether an evidence or divergence functional is used, as well as the nature of their encoding of value (endogenous or exogenous) as well as whether value is encoded into the variational or generative distribution. Moreover, we have derived a good understanding of the impact of different definitions of the generative model upon the objective functionals that result. This allows us a broad and unique understanding of the possible space of objective functionals, as well as the design choices which influence which one to choose. Future work in this area should investigate the actual impacts of different choices on agent behaviour, both in simple toy environments where the effects can be easily understood, as well as in more complex and difficult benchmark tasks where using different objectives may well give rise to algorithms which are more effective than current agents. As an example, while only using random exploration, control as inference inspired approaches such as the soft-actor-critic have given rise to state of the art performance. We see no reason why more exploratory information-seeking agents, powered by divergence objectives, should not lead to similar gains in performance, especially on challenging sparse-reward tasks. While we present some preliminary results to this effect in Chapter 4, much more work remains to be done to pin down what gains, if any, are possible by this approach.

Finally, our approach directly gives us the objective functionals  utilizing different generative models. For instance, if you want to extend your reinforcement learning algorithms to POMDPs, or POMDPs with hierarchical levels of latent states, or to explicitly model distributions over different model parameters, or to explicitly model a reward distribution or reward model, then our framework provides a recipe for precisely and immediately deriving the necessary objective to optimize. 

In the next chapter, we move on to consider applications of the free energy principle to learning, where we focus on deriving novel algorithms which can perform credit assignment in neural networks in a biologically plausible fashion.

%% file: chap6.tex
\chapter{Credit Assignment in the Brain}

\section{Introduction}

In this chapter we shift gears again and now consider applications of the free energy principle to the problems of \emph{learning} in the brain. Specifically, here we aim to understand the nature of credit assignment in the brain, and focus on how and whether the backpropagation of error algorithm which underpins all the recent successes of machine learning in training deep artificial neural networks, could potentially be implemented in the brain. In this chapter, we present the fruit of our work investigating this extremely important question for the case of rate-coded integrate and fire neurons engaged in a static task (such as object recognition), where there is only a feedforward pass to be concerned with and all backpropagation is through space, and not time. While this setting is considerably simplified from the one the brain faces in reality, it is also much more tractable and well-understood and solving the problem in this domain may provide vital clues into the full solution. 

This chapter is split into four relatively independent sections. In the first section, we provide a general introduction and mini literature review on backpropagation and previous attempts to derive biologically plausible algorithms to implement backpropagation in the brain.  In the next two sections, we  then present our work deriving new algorithms for biologically plausible approximations to the backpropagation of error algorithm. 

In the first section, we show that predictive coding -- the free energy process theory from chapter 3 -- can, if set-up correctly, exactly approximate the backpropagation of error algorithm along arbitrary computation graphs. This result is fascinating since predictive coding has a long history and well-developed literature on its properties, performance, and especially its biological plausibility, as well as possessing several well-developed theoretical neural implementations \citep{bastos2012canonical,keller2018predictive,kanai2015cerebral}. We empirically validate this approximation to backprop and showcase that predictive coding can perform equally to backprop at training complex machine learning architectures such as CNNs and LSTMs. 

Secondly, we develop a novel algorithm -- \emph{Activation Relaxation} (AR) -- which also can asymptotically converge to the required backpropagation error gradients using only local connectivity -- and which does not require two separate population of value and error neurons. We empirically show that this algorithm can train complex machine learning architectures with performance equal to backprop and, additionally, demonstrate that the same relaxations shown in Chapter 3 for predictive coding -- such as using learnable backwards weights to overcome the weight transport problem, and dropping the nonlinear derivatives also work for the AR algorithm, thus importantly both substantially improving the overall biological plausibility of the AR algorithm as well as demonstrating the generalizability of the results in Chapter 3 to other algorithms \footnote{For this thesis, the AR algorithm is not directly related to the FEP, although it was initially inspired by research into predictive coding which is a process theory of the FEP}.

Finally, in the third section, we have included a more speculative discussion on a potential further algorithm for solving backprop in rate-coded neurons directly, instead of in an iterative fashion. We discuss the possible limitations of this algorithm as well as the required neural circuitry and, for the first time, begin to precisely understand what the key problems are for the rate-coded sense and also what a real solution would look like.

\subsection{Backpropagation in the Brain}

Due to the immense success of machine learning approaches based upon connectionist deep neural networks trained upon the backpropagation of error algorithm, our paradigms of how the brain functions is also shifting. Specifically, the paradigm that the brain, or at least the neocortex, is fundamentally a blank-slate learning machine which uses general purpose learning algorithms to handle inputs, akin to a deep neural network is becoming increasingly influential in neuroscience, partially displacing older views that the brain consists of a series of separate `modules' \citep{fodor1983modularity,pinker2003language}, each of which performs a single specialized function using what are effectively specific and pre-set algorithms, hard-coded over the course of evolutionary history. While functional specialisation is an extremely notable characteristic of the brain, it is more widely believed that this specialisation, especially in the cortex, is due to differences in input and small inductive biases shaping the nature and output of a very general learning algorithm which is implemented throughout the cortex, rather than each functional module possessing its own independent and isolated suite of algorithms. This view is supported by evidence of a remarkable uniformity of cortical cytoarchitecture and neuroanatomy, belied by the heterogeneity observed in subcortical areas \citep{bear2020neuroscience}.

While, for a long time, it was empirically unclear whether the fundamentals of intelligence -- such as robust and generalizable perception, natural language capabilities, and adaptive action planning -- could emerge solely from learning algorithms with relatively few inductive biases, applied to vast amounts of data, the past decade and its immense advances in machine learning are suggestive that this may be possible after all. Modern machine learning, effectively, represents the culmination and empirical verification of earlier connectionist theory \citep{rumelhart1986learning}.

This view places the central focus on learning. Since the backpropagation of error algorithm has proven so immensely successful in machine learning, to train an extremely wide variety of architectures to perform an impressive array of tasks \citep{krizhevsky2012imagenet,goodfellow2014generative,radford2019language,schrittwieser2019mastering,schmidhuber1999artificial}, and given that the brain itself faces an almost identical credit assignment problem in having to adjust synaptic strengths to allow for learning to occur, it is a very interesting and important question to ask whether the learning algorithm implemented in the brain could simply be backprop. If this question were conclusively answered in the affirmative, it would represent an enormous conceptual breakthrough in neuroscience since it would provide, for the first time, a general and powerful organizing principle for the brain (or at least the cortex) as a whole, it would allow the importation directly into neuroscience of a large quantity of results from machine learning. Such an answer would additionally have deep philosophical implications. It would imply that there is very little effective difference between current machine learning methods and the kinds of learning, inference, and planning algorithms that are implemented in the brain to give rise to undeniably intelligent and apparently conscious behaviour, and as such would imply that the current paradigm in machine learning suffices, with more scale and potentially more expressive architectures, to create fully general intelligences akin to humans or beyond \citep{bostrom2017superintelligence}.

On the other hand, if it were shown that the brain were conclusively not doing backprop, then this would also be an advance, although a lesser one, in neuroscience. Such a conclusion would necessarily shed light upon the actual algorithms utilized by the brain for credit assignment and learning, which would provide both a general principle for understanding the function and operation of the brain over time, as well as undoubtedly provide key insights for machine learning in developing, improving, and scaling up current methods. 

The theory and algorithm for backpropagation of error (Backprop) emerges in the 1970s \citep{linnainmaa1970representation}, and by the 1980s was widely used for training connectionist neural networks \citep{rumelhart1985feature,rumelhart1986learning,griewank1989automatic} Already in the 1980s, researchers had proposed that backprop could be implemented in the brain, and tried to find commonalities between then contemporary neuroscience and progress in connectionist modelling using neural networks \citep{rumelhart1986learning}. However, a number of articles argued convincingly that a direct implementation of backpropagation is biologically implausible \citep{crick1989recent}, which dampened down potential interest in this connection considerably until the question was re-raised by the successes of machine learning in the 2010s. Before investigating the ways in which backpropagation appears biologically implausible, and how these issues might be addressed, we first give a detailed introduction to the backprop algorithm.

First we must decide on some terminology. Neural networks are typically trained to minimize some loss function $\mathcal{L}$. 
The credit assignment problem concerns the \emph{computation of the derivatives} of the loss with respect to every parameter of the network $\frac{\partial \mathcal{L}}{\partial W}$. Backpropagation of error is an algorithm that solves the credit assignment problem exactly using the technique of \emph{Automatic Differentiation} (AD) \citep{griewank1989automatic,baydin2017automatic,van2018automatic,paszke2017automatic}. \footnote{There are other methods for computing derivatives, such as finite differences, but they are less accurate and more computationally costly than AD and are not generally used in machine learning} . Given these derivatives, the network can be trained with the stochastic gradient descent algorithm $W_{t+1} = W_t + \eta \frac{\partial \mathcal{L}}{\partial W}$. However, given the gradients computed by backprop, other gradient algorithms are possible including a variety of modified descent procedures such as Nesterov Momentum \citep{nesterov27method}, RMS-prop \citep{hinton2012neural} and Adam \citep{kingma2014adam} second order methods such as natural gradients \citep{amari1995information} and Gauss-Newton optimization, as well as stochastic sampling methods such as stochastic langevin dynamics \citep{welling2011Bayesian}, and Hamiltonian MCMC \citep{neal2011MCMC}.

The fundamental mathematics underlying backprop is the chain rule of calculus. Essentially, if we have a function -- such as a neural network -- which consists of the composition of many differentiable functions, then we can express the derivative of a complex composite function as a product of the derivatives of all the component functions with respect to one another. Suppose we have the forward function and loss function,
\begin{align*}
&\hat{y} = f^L(W^L f^{L-1}(W^{l-1} f^{L-2}(\dots f^0(W^0 x)))) \\
& \mathcal{L} = g(\hat{y}, t) \numberthis
\end{align*}
where $\hat{y}$ is the prediction outputted by the neural network, $L$ is the number of layers, $[f^L \dots f^0]$ is the activation functions for each layer and $[W^L \dots W^0]$ is the weights or parameters for each layer. $\mathcal{L}$ is the overall loss function, $t$ is the desired target outputs and $g$ is the loss function. With this forward function, we can compute the derivative of the loss with respect to any parameter set -- for instance $W^0$ using the chain rule,
\begin{align*}
\label{chain_rule_equation}
\frac{\partial \mathcal{L}}{\partial W^0} = \frac{\partial \mathcal{L}}{\partial \hat{y}}\frac{\partial \hat{y}}{\partial y^{L}} \big[ \prod_{l=L}^{l=1} \frac{\partial y^l}{\partial y^{l-1}} \big] \frac{\partial y^1}{\partial W^0} \numberthis
\end{align*}
which allows us to express the derivative of a product of all the derivatives of the intermediate component functions. In general, this is possible for any function as long as every component function is differentiable. We can represent any function as a \emph{computation graph} which is a graph where each intermediate step in the computation is a vertex and each component function is an edge. While this example, and most simple neural network, is simply a chain graph, other more complex graphs are possible. If there are multiple paths through the graph, the derivatives of the paths are summed together. As long as every component function is differentiable, every one of the derivatives in Equation \ref{chain_rule_equation} can be explicitly computed and evaluated, thus allowing the full derivative $\frac{\partial \mathcal{L}}{\partial W^0}$ to be computed explicitly. Importantly, the computational cost of such an evaluation is generally linear in the number of component functions -- and thus is of approximately the same complexity as simply evaluating the function in the first place. AD approaches like this, then, allow for the evaluation of derivatives of any differentiable computation for a low and constant additional computational cost -- allowing for their widespread use within machine learning.

There are two approaches in AD to computing the chain of derivatives as in Equation \ref{chain_rule_equation}, which are called `forward-mode' and `reverse-mode' AD. The difference between these methods is effectively whether the product of derivatives is computed from right to left (forward-mode) or left to right (reverse mode). Forward-mode effectively accumulates the gradients starting with the initial jacobian $\frac{\partial y^1}{\partial W^0}$ and then moving leftwards down the chain, in the same direction as the original function evaluation. This allows derivative evaluation to take place in parallel with original function evaluation through the use of `dual numbers' \citep{griewank1989automatic} which extend every number with an `derivative part' analogous to how complex numbers extend real numbers with a complex part. Since dual-numbers can be evaluated in parallel with the original function evaluation, the computational cost of forward-mode AD is of the order of the input dimension and it has a constant memory cost, since no intermediate products need to be stored in memory.

Reverse-mode AD, conversely, evaluates the chain of derivatives from left to right. That is, it starts with and accumulates onto the vector $\frac{\partial \mathcal{L}}{\partial \hat{y}}$, which is also called the \emph{adjoint} or \emph{pullback} \footnote{In continuous time, Equation \ref{reverse_mode_equation} becomes the adjoint ODE}. It then iterates recursively `backwards' through the chain using the following equation,
\begin{align*}
\label{reverse_mode_equation}
    \frac{\partial \mathcal{L}}{\partial y^l} = \sum_j \frac{\partial \mathcal{L}}{\partial y_j^{l+1}} \frac{\partial y_j^{l+1}}{\partial y^l} \numberthis
\end{align*}
where the sum simply states that if there are multiple potential paths in the graph, they should be summed together. Since it starts at the `end' and works backwards, reverse-mode AD requires the function to be evaluated first after which the derivatives can begin to be computed in a backwards sweep, thus leading reverse-mode AD to use characteristic forward (function evaluation) and backwards (derivative evaluation) sweeps. Since all intermediate activities must be stored in memory, reverse-mode AD has a memory cost linear in the number of component functions, as well as a computational cost which scales with the dimension of the output. Since neural networks typically have a scalar output loss and very high dimensional inputs (such as image pixels), reverse-mode AD is typically computationally cheaper and is the method used in practice for training deep networks. Reverse mode AD also has the advantage that it backpropagates the gradients back to where the weights are directly, while forward-mode AD does compute the weights, but they are all bunched up at the end of the graph by the loss, and therefore would need, in a physical system such as the brain, to be transmitted back to where the weights were originally as well.

Here, we focus primarily on the implementation of reverse-mode AD in the brain due to its generally superior computational capabilities (and avoidance of this weight locality problem for forward mode AD). It is possible, however, that the brain may use some combination of forward and reverse mode in practice. One especially appealing method is to use reverse-mode AD to handle hierarchical networks -- i.e. nonlocality in space, while using forward mode AD to handle recurrent credit assignment through time. Here forward mode AD has the clear advantage that the gradients move forward in time at the same rate as the weights themselves, so that there is no locality problem here. In fact, the locality problem now afflicts reverse-mode AD which, in this circumstance, needs to backpropagate gradients \emph{backwards through time}, which is problematic. Here we do not address the temporal credit assignment problem and focus entirely on backpropagation through space, where we assume reverse-mode AD is the best approach.

Although reverse-mode AD is a well characterised algorithm, it is not at all clear whether it can be implementated in the brain. Specifically, backpropagation has three principal problems which make its apparent biological plausibility dubious -- firstly, backpropagation appears to require non-local information transfer, since the gradient of the synaptic weights depends on activity from the rest of the network which ultimately leads to the final loss outcome. Secondly, if we look at the specific case of a rate-coded standard integrate and fire model, whereby the output is a function of the input and the synaptic weights,
\begin{align*}
    y^{l+1} = f(W^l y^l) \numberthis
\end{align*}
then the derivative of the loss with respect to the pre-activations becomes,
\begin{align*}
    \frac{\partial \mathcal{L}}{\partial y_l} = \frac{\partial \mathcal{L}}{\partial y^{l+1}} \frac{\partial f(W^l y^l)}{\partial y^l} {W^l}^T \numberthis
\end{align*}
which requires both the derivative of the activation function, which may or may not be easy to compute locally, and also the transpose of the feedforward weights ${W^l}^T$. This transpose is problematic since effectively it requires the backwards pass activations to be sent `backwards' through the forwards weights -- a process which is biologically implausible. This problem is called the `weight transport problem'. Finally, if we look at the update rule for the weights themselves,
\begin{align*}
\frac{\partial \mathcal{L}}{\partial W^l} = \frac{\partial \mathcal{L}}{\partial y^{l+1}} \frac{\partial f(W^l y^l)}{\partial W^l} y^L \numberthis
\end{align*}
which while it does depend on the pre-synaptic activations $y^l$ also depends on the adjoint vector $\frac{\partial \mathcal{L}}{\partial y^{l+1}} $ which is generally non-local. Interestingly, the update rule specifically does \emph{not} depend at all on the post-synaptic activity, in contrast to the widely accepted view of Hebbian plasticity being implemented in the brain, although it does depend on the \emph{derivative} of the post-synaptic activity with respect to the weights $\frac{\partial f(W^l y^l)}{\partial W^l}$, which may or may not be difficult to compute. The issue first of computing the adjoint vector $\frac{\partial \mathcal{L}}{\partial y^{l+1}}$ in a biologically plausible manner, and then transmitting it to the required synapses, since it is non-local, thus form the core issue standing in the way of any biologically plausible implementation of reverse-mode AD. A final issue relates to the need, in reverse-mode AD, for separate forward and backwards phases, while the brain presumably needs to operate continuously in time. It has been suggested that the brain's rhythmic oscillations \citep{buzsaki2006rhythms} may allow it to `multiplex' forward and backward passes together, although this intuition has not yet, to my knowledge, been made precise in the literature.

Due to a general understanding that backpropagation is biologically implausible, much research has focused on other potentially more biologically plausible methods by which the brain might learn. A large amount of attention has focused on Hebbian update rules \citep{gerstner2002mathematical}, which only utilize the post and pre synaptic activities, based on the initial intuitions of Donald Hebb \citep{hebb1949first}. A number of variants of Hebbian learning have been proposed, and the full class of potential algorithms has been exhaustively analysed in \citep{baldi2016theory}. However, a key issue with Hebbian learning is that, since it can only use local information in the form of pre and post synaptic activities, it cannot incorporate information about the distant loss function, and thus allow for the precise goal-directed learning that backprop is capable of. In effect, Hebbian learning can only capture and strengthen the local correlations of firing rates across a network and thus, while it can often be used to solve tasks in shallow networks with just a single or a few hidden layers, it fails to scale successfully to deep layers \citep{lillicrap2019backpropagation}.

A second approach is to use global neuromodulators as a part of a `three-factor' learning rule which includes contributions from both the pre-synaptic and post-synaptic activity as well as this `third-factor' \citep{gershman2018uncertainty} which is often conceptualised to be dopamine, in light of the fact that dopaminergic connections from the mid-brain are well-known to innervate large parts of the cortex \citep{daw2006cortical}. These dopaminergic neurons could be signalling some kind of global reward signal, inducing all neurons to increase their weight when a positive reward is encountered and decrease it when a negative reward occurs \citep{seung2003learning,roelfsema2005attention,lillicrap2020backpropagation}, in a procedure which is effectively equivalent to policy gradient methods from reinforcement learning \citep{williams1989experimental}. While such approaches can, asymptotically, learn complex functions in deep neural networks, their key limitation is that the gradient estimates they compute have extremely high variance, since the global neuromodulator cannot distinguish whether a \emph{particular} weight helped give rise to a reward or not, and thus cannot provide precise feedback like backprop. Instead it takes a substantial amount of trials for the random noise provided by the contributions of all the other neurons in the brain to be averaged out to get at the contribution of just a single synaptic weight, which leads to slow and unstable learning in complex tasks with deep networks \citep{lillicrap2019backpropagation}. Additionally, this approach implicitly assumes that the entire brain is optimized end-to-end for reward, however this seems potentially unlikely given that large parts of the cortices deal with aspects like sensory stimuli which are distant from reward and most likely use auxiliary losses such as their own immediate prediction errors.  Nevertheless, if it turns out that backpropagation is not, in fact, used in the brain, then global neuromodulatory rules like this may be the second best option. Importantly, even if it is the case that the brain does backprop, it likely \emph{also} uses this global neuromodulatory approach in a modulatory function to bias learning towards high reward contingencies and perhaps also to adaptively tune learning rates throughout the cortex so that highly valenced experiences (either positive or negative) have a strong effect on plasticity throughout the brain.

Finally, there is also a small but growing literature attempting to understand how and whether backpropagation can be directly implemented in the brain -- namely whether it is possible to design neural circuits to work around the key limitations proposed earlier. An important line of work tackles the weight transport problem. \citet{lillicrap2016random} demonstrate that in reality precise copying of the forward and backwards weights is not necessary, and in fact random fixed backward weights suffice due to the phenomenon of `feedback alignment' whereby the random feedback weights effectively force the forward weights to align with the backward ones to be able to learn. This approach can be improved by ensuring that the sign of the elements of the feedback weight matrix matches the sign of the forward weight matrix (potentially a more biologically plausible constraint than exact copying of values) \citep{liao2016important}, or else learning the backwards weights using an additional plasticity rule \citep{amit2019deep,akrout2019deep,millidge2020relaxing}. While the feedback alignment technique does not typically scale to deep architectures, as the feedback path becomes increasingly corrupted \citep{bartunov2018assessing}, an approach called direct feedback alignment (DFA) \citep{nokland2016direct}, whereby all layers are directly connected to the output layer through random backwards weights, has been shown to be able to scale to large deep architectures \citep{launay2019principled}, although not the usual convolutional neural networks used in vision. While an impressive result, DFA itself violates known neural connectivity constraints which feature reciprocal connectivity between regions and not every layer receiving direct backwards connections from the `output'. 

Secondly, another line of work has focused on the algorithm of target-propagation \citep{bengio2015early,lee2015difference} which is similar to backprop except that instead of providing gradients back to the weights at each layer it provides targets which can then be optimized locally. These targets are produced by mapping backwards the output of the network `nudged' towards the true target through an inverse mapping at each layer. The intuition, is that we want to find the targets which, if the lower layers had matched their activations, would have produced a final output closer to the target. The past year has seen substantial advances in the theoretical analysis of target-prop, where it is now recognised to not approximate backprop, but instead be performing a hybrid form of Gauss-Newton optimization \citep{bengio2020deriving,meulemans2020theoretical}. While promising, large-scale studies on the stability and scalability of target-prop learning have not yet been done, although initial results are promising \citep{bartunov2018assessing}. Additionally, target-prop does not provide solutions to the weight transport problems (now complicated by the additional necessity of learning the backwards inverse weights), and the necessity of storing and comparing the forward and backward information across phases. The intuition of using the final loss function to compute layer-wise targets has also been applied, with variations, in other works \citep{ororbia2019biologically,ororbia2017learning,kaiser2020synaptic}.

Another approach is to use only local information at the synapses, but use a backwards phase which instead of being purely sequential is treated as a dynamical systems which undergoes multiple iterations. Using this approach, while it takes longer than a sequential backwards pass, also allows using only local information, since the information about the loss can be `leaked' slowly backwards using the dynamics over time instead of having to be explicitly transmitted backwards. This allows the brain to operate using the same dynamical rules at all times, in general, instead of sequential forwards and backwards transmission of information. One key example of such a framework is the algorithm of Equilibrium Propagation \citep{bengio2017stdp,scellier2017equilibrium,scellier2018extending,scellier2018generalization}, which uses two dynamical phases -- a \emph{free phase} where the dynamics of the system are allowed to evolve without any influence of the targets, and a \emph{clamped phase} in which the output units are held at a value nudged towards the targets, which destabilizes the free-phase equilibrium and instead sends the network towards a different clamped equilibrium state. It turns out that the difference between these two states corresponds closely to the gradients which would otherwise have been backpropagated through the network and can thus be used to adjust the synaptic weights \citep{scellier2017equilibrium}. Equilibrium propagation has been extensively tested on small datasets like MNIST and CIFAR, although it is not yet known how well the method scales. Additional problems with EP are its necessary use of two distinct backwards phases and, crucially, the storage of information (the equilibrium in the free-phase, throughout the entirety of the clamped phase before their subtraction to obtain the gradients. Another iterative algorithm which has been shown to approximate backprop is predictive coding \citep{whittington2017approximation}.

In this thesis chapter, we make two contributions to the theory of iterative algorithms for approximating backprop. Firstly, we extend work by \citet{whittington2017approximation}, showing that predictive coding can approximate backprop by making this claim precise and exact, and extending it to arbitrary computation graphs \citep{millidge2020predictive}. Specifically, we show that predictive coding provides a fully general iterative approach to approximating reverse-mode automatic differentiation through an identification of the equilibrium prediction error with the adjoint term. This allows us to define predictive coding networks which can train any contemporary machine learning architecture with accuracy equivalent to backprop in a local and biologically plausible (ish) manner. We demonstrate this capability on CNNs and LSTMs, thus substantially extending the range and scale of architectures to which predictive coding has been applied. 

Secondly, we propose a novel iterative algorithm -- Activation Relaxation \citep{millidge2020activation}-- which converges precisely to the exact backprop gradients, while also considerably simplifying the predictive coding update rules and obviating the need for separate populations of error and `value' neurons which predictive coding possesses. Additionally, we demonstrate that certain remaining implausibilities in the algorithm such as the weight transport problem and the nonlinear derivatives problem can be `relaxed' while retaining learning performance almost equivalent to backprop, even on challenging and large-scale computer vision tasks.

A final issue which undermines many of the proposed learning rules for both sequential methods like target-prop and iterative ones like EP or predictive coding, is the necessity of three-factor learning rules with precise vector feedback, like backprop. It is still fairly unclear whether such rules can actually be implemented in the brain, although there has been some work showing that prediction error, or gradient like quantities could in theory be transmitted backwards through the network using segregated dendrites \citep{sacramento2018dendritic} which may help maintain separate error representations independently of the firing rates of the rest of the somatic neuron. While the actual biological plausibility of this approach is unclear, in the last section of this chapter, we will speculate that if it is plausible, and the brain can maintain and update separate error and value representations on single neurons, or indeed on separate populations but with precise three-factor learning rules, then that is all that is necessary for a direct biologically plausible implementation of backprop, especially given recent research showing that the weight transport problem can be largely overcome through learning the backwards weights.  We present a theoretical algorithm, extremely similar to target-prop, and with three-factor learning rules, which precisely corresponds to backprop in the brain, which is mathematically equivalent to backprop. Thus, if three-factor learning rules are possible in the brain, either through segregated dendrites, or else precise interneuron connectivty, then exact backpropagation can be performed as simply as target-prop or any other algorithm.

\section{Predictive Coding Approximates Backprop Along Arbitrary Computation Graphs}

Here we demonstrate that predictive coding can approximate backpropagation on arbitrary computation graphs. As we recall from chapter 5, predictive coding arises from a variational inference algorithm on the activations on each layer of the hierarchy, whereby the predictive coding update rules can be derived as a gradient descent on the variational free energy $\mathcal{F}$. To showcase how this methodology extends to arbitrary graphs, we must define the variational inference problem to be solved on an arbitrary computation graph. First, we must make the notion of a computation graph explicit. 

\begin{figure}[ht]
    \centering
    \includegraphics[width=\textwidth]{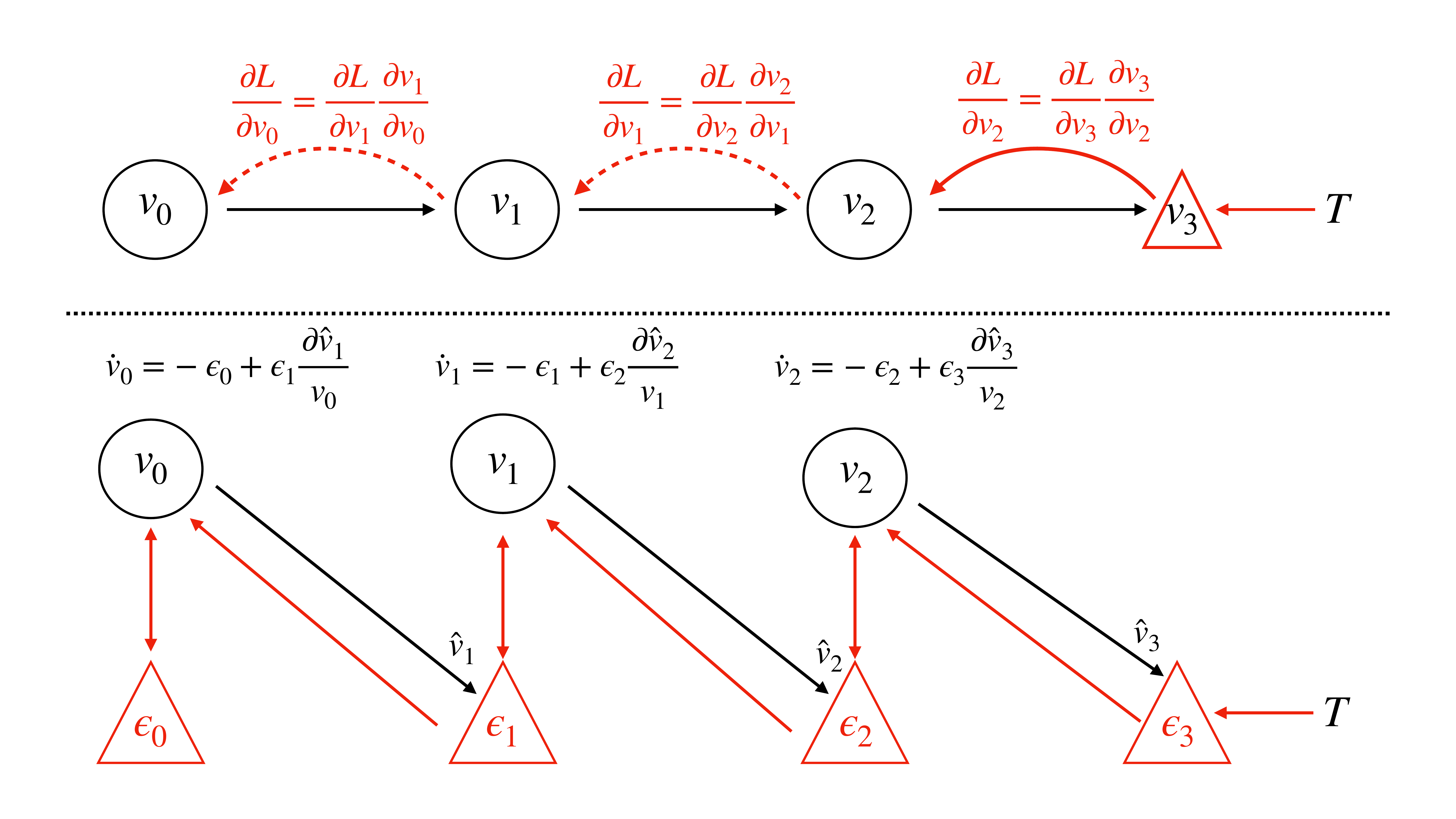}
    \caption{Top: Backpropagation on a chain. Backprop proceeds backwards sequentially and  explicitly computes the gradient at each step on the chain. Bottom: Predictive coding on a chain.  Predictions, and prediction errors are updated in parallel using only local information. Importantly, while the original computation graph (black lines) must be a DAG, the augmented predictive coding graph is cyclic, due to the backwards (red) prediction error connections.}
\label{main_PCBP_schematic}
\end{figure}

A computation graph $\mathcal{G} = \{\mathbb{E},\mathbb{V}\}$ is a directed acyclic graph (DAG) which can represent the computational flow of essentially any program or computable function as a composition of elementary functions. Each edge $e_i \in \mathbb{E}$ of the graph corresponds to an intermediate step -- the application of an elementary function -- while each vertex $v_i \in \mathbb{V}$ is an intermediate variable computed by applying the functions of the edges to the values of their originating vertices. $v_i$ denotes the vector of activations within a layer and we denote the set of all vertices as $\{v_i\}$. Effectively, computation flows `forward' from parent nodes to all their children through the edge functions until the leaf nodes give the final output of the program as a whole (see Figure \ref{main_PCBP_schematic} and \ref{PCBP_numerical_results} (top) for an example). Given a target $T$ and a loss function $L = g(T, v_{out})$, the graph's output can be evaluated and, and if every edge function is differentiable, automatic differentiation can be performed on the computation graph.

We can extend predictive coding to arbitrary computation graphs in a supervised setting by defining the inference problem to be solved as that of inferring the vertex value $v_i$ of each node in the graph given fixed start nodes $v_0$ (the data), and end nodes $v_N$ (the targets). We define a generative model which parametrises the value of each vertex given the feedforward prediction of its parents, $p(\{v_i\}) = p(v_0 \dots v_N) = \prod_i^N p(v_i | \mathcal{P}(v_i) )$ \footnote{This includes the prior $p(v_0$), which simply has no parents.}, and a factorised, variational posterior $Q(\{v_i\} | v_0, v_N) = Q(v_1 \dots v_{N-1} | v_0, v_N) = \prod_i^N Q(v_i | \mathcal{P}(v_i), \mathcal{C}(v_i))$, where $\mathcal{P}(v_i)$ denotes the set of parents and $\mathcal{C}(v_i)$ denotes the set of children of a given node $v_i$. From this, we can define a suitable objective functional, the variational free energy $\mathcal{F}$ (VFE), which acts as an upper bound on the divergence between the true and variational posteriors.
\begin{equation}
\begin{aligned}
    \mathcal{F} &= KL[(Q(v_1 \dots v_{N-1} | v_0, v_N) \Vert p(v_0 \dots v_N)] \geq  KL[(Q(v_1 \dots v_{N-1}) | v_0, v_N) \Vert p(v_1 \dots v_{N-1}| v_0, v_N)] \\
    &\approx \sum_{i=0}^N \epsilon_i^T \epsilon_i 
\end{aligned}
\end{equation}

Under Gaussian assumptions for the generative model $p(\{v_i\}) = \prod_i^N \mathcal{N}(v_i ; \hat{v_i}, \Sigma_i)$, and the variational posterior $Q(\{v_i\}) = \prod_i^N \mathcal{N}(v_i)$, where the `predictions' $\hat{v_i} = f(\mathcal{P}(v_i); \theta_i)$ are defined as the feedforward value of the vertex produced by running the graph forward, and all the precisions, or inverse variances, $\Sigma^{-1}_i$ are fixed at the identity, we can write $\mathcal{F}$ as simply a sum of prediction errors (see chapter 5), with the prediction errors defined as $\epsilon_i = v_i - \hat{v}_i$. Since $\mathcal{F}$ is an upper bound on the divergence between true and approximate posteriors, by minimizing $\mathcal{F}$, we reduce this divergence, thus improving the quality of the variational posterior and approximating exact Bayesian inference. Predictive coding minimizes $\mathcal{F}$ by employing the Cauchy method of steepest descent to set the dynamics of the vertex variables $v_i$ as a gradient descent directly on $\mathcal{F}$

\begin{equation}
\begin{aligned}
\label{PCBP_v_update}
    \frac{dv_i}{dt} = \frac{\partial \mathcal{F}}{\partial v_i} = \epsilon_i - \sum_{j \in \mathcal{C}(v_i)} \epsilon_j \frac{\partial \hat{v}_j}{\partial v_i} 
    \end{aligned}
\end{equation}
The dynamics of the parameters of the edge functions $W$ such that $\hat{v_i} = f(\mathcal{P}(v_i); W)$, can also be derived as a gradient descent on $\mathcal{F}$. In a neural network model, the parameters correspond to the synaptic weights of each layer of the neural network. Importantly these dynamics require only information (the current vertex value, prediction error, and prediction errors of child vertices) locally available at the vertex.
\begin{align*}
\label{PCBP_weight_update}
    \frac{dW_i}{dt} &= \frac{\partial \mathcal{F}}{\partial W_i} = \epsilon_i \frac{\partial \hat{v}_i}{\partial W_i} \numberthis
\end{align*}

To run generalized predictive coding on a given computation graph $\mathcal{G} = \{\mathbb{E},\mathbb{V}\}$, we augment the graph with error units $\epsilon \in \mathcal{E}$ to obtain an augumented computation graph $\tilde{\mathcal{G}} = \{\mathbb{E},\mathbb{V},\mathcal{E} \}$. The predictive coding algorithm then operates in two phases -- a feedforward sweep and a backwards iteration phase. In the feedforward sweep, the augmented computation graph is run forward to obtain the set of predictions $\{ \hat{v_i} \}$, and prediction errors $\{\epsilon_i \} = \{ v_i - \hat{v_i} \}$ for every vertex. To achieve exact equivalence with the backprop gradients computed on the original computation graph, we initialize $v_i = \hat{v}_i$ in the initial feedforward sweep so that the output error computed by the predictive coding network and the original graph are identical -- an assumption we call the \emph{fixed prediction} assumption.

In the backwards iteration phase, the vertex activities $\{ v_i \}$ and prediction errors $\{\epsilon_i \}$ are updated with Equation \ref{PCBP_v_update} for all vertices in parallel until the vertex values converge to a minimum of $\mathcal{F}$. After convergence the parameters are updated according to Equation \ref{PCBP_weight_update}. Note we also assume, following \citet{whittington2017approximation}, that the predictions at each layer are fixed at the values assigned during the feedforward pass throughout the optimisation of the $v$s. This is the \emph{fixed-prediction assumption}. In effect, by removing the coupling between the vertex activities of the parents and the prediction at the child, this assumption separates the global optimisation problem into a local one for each vertex. We implement these dynamics with a simple forward Euler integration scheme so that the update rule for the vertices became $v_i^{t+1} \leftarrow v_i^t - \eta \frac{d\mathcal{F}}{dv_i^t}$
where $\eta$ is the step-size parameter. Importantly, if the edge function linearly combines the activities and the parameters followed by an elementwise nonlinearity, then both the update rule for the vertices (Equation \ref{PCBP_v_update}) and the parameters (Equation \ref{PCBP_weight_update}) become Hebbian. Specifically, the update rules for the vertices and weights become $\frac{dv_i}{dt} = \epsilon_i - \sum_j \epsilon_j f'(\theta_j \hat{v_j}) \theta_j^T$ and $\frac{d\theta_i}{dt} = \epsilon_i f'(\theta_i \hat{v_i}) \hat{v_i}^T$, respectively.

\begin{algorithm}[H]
\SetAlgoLined
\DontPrintSemicolon
\textbf{Dataset:} $\mathcal{D} = \{\mathbf{X},\mathbf{L}\}$, Augmented Computation Graph $\tilde{\mathcal{G}} = \{\mathbb{E},\mathbb{V},\mathcal{E}\}$, inference learning rate $\eta_v$, weight learning rate $\eta_\theta$
\BlankLine
    \For{$(x,L) \in \mathcal{D}$}{
    $\hat{v_0} \leftarrow x$ \\
    \For{$\hat{v}_i \in \mathbb{V}$}{
    $\hat{v}_i\leftarrow f(\{\mathcal{P}(\hat{v}_i); \theta)$ \\
    }
    $\epsilon_L \leftarrow L - \hat{v}_L$ \\
        \While{not converged}{
        \For{$(v_i, \epsilon_i) \in \tilde{\mathcal{G}}$}{
        $\epsilon_i \leftarrow v_i - \hat{v}_i$ \\
        $v_i^{t+1} \leftarrow v_i^t + \eta_v \frac{d\mathcal{F}}{dv_i^t}$
        }
        }
    \For{$\theta_i \in \mathbb{E}$}{
        $\theta_i^{t+1} \leftarrow \theta_i^t + \eta_\theta \frac{d\mathcal{F}}{d\theta_i^t}$
    }
}

\caption{Generalized Predictive Coding \label{general_alg_pseudocode}}
\end{algorithm}

\subsection{Methods and Results}

To demonstrate that this predictive coding scheme approximates the backpropagation of error algorithm at convergence is fairly straightforward. The key step is to show that the recursion relationship of the adjoint term for the equilibrium of the prediction errors in predictive coding is identical to that in reverse-mode AD, even for arbitrary computation graphs. To make this clear more intuitively, we first demonstrate that, at the equilibrium of the dynamics, the prediction errors $\epsilon^*_i$ converge to the correct backpropagated gradients $\frac{\partial \mathcal{L}}{\partial v_i}$, and consequently the parameter updates (Equation \ref{PCBP_weight_update}) become precisely those of a backprop trained network.

\begin{figure*}
        \centering
        \begin{subfigure}[b]{0.475\textwidth}
            \centering
            \includegraphics[width=\textwidth]{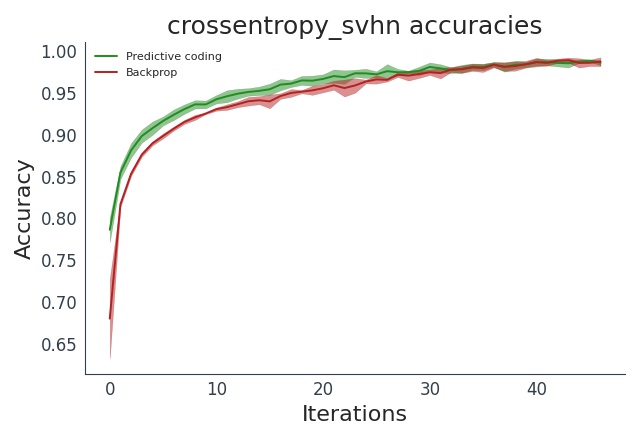}

            {{\small SVHN training accuracy }}    
        \end{subfigure}
        \hfill
        \begin{subfigure}[b]{0.475\textwidth}  
            \centering 
            \includegraphics[width=\textwidth]{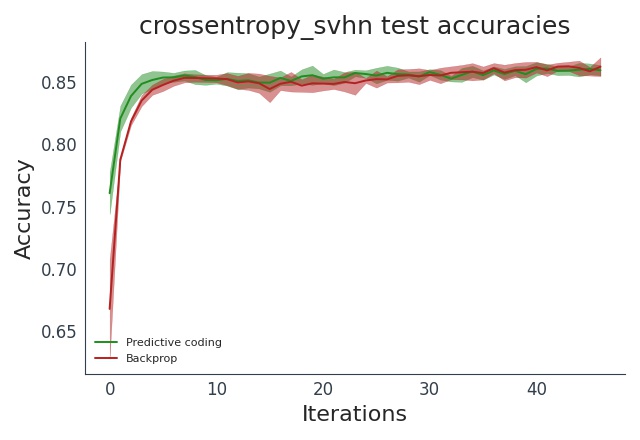}
            \caption[]%
            {{\small SVHN test accuracy}}    
        \end{subfigure}
        \vskip\baselineskip
        \begin{subfigure}[b]{0.475\textwidth}   
            \centering 
            \includegraphics[width=\textwidth]{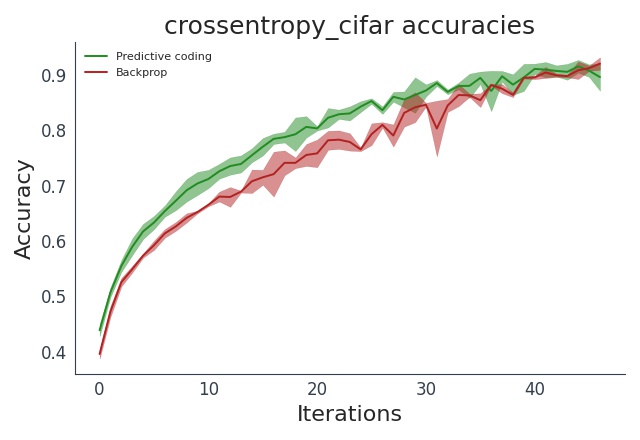}
            \caption[]%
            {{\small CIFAR training accuracy}}    
        \end{subfigure}
        \quad
        \begin{subfigure}[b]{0.475\textwidth}   
            \centering 
            \includegraphics[width=\textwidth]{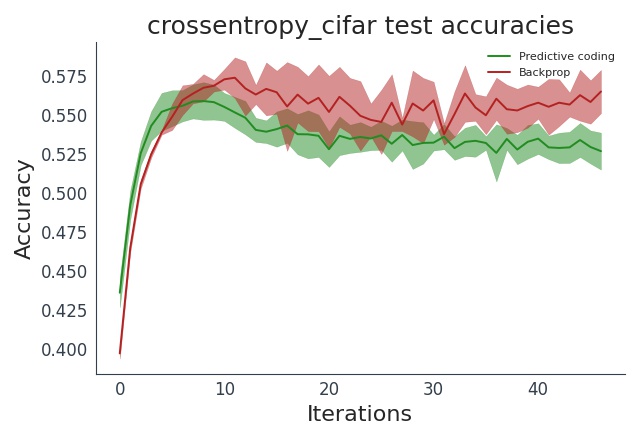}
            \caption[]
            {{\small CIFAR test accuracy}}    
        \end{subfigure}
        \caption{\small Training and test accuracies of the CNN network on the SVHN and CIFAR datasets using the cross-entropy loss. As can be seen performance remains very close to backprop, thus demonstrating that our predictive coding algorithm can be used with different loss functions, not just mean-squared-error.} 
       
        \label{PCBP_CNN_acc}
    \end{figure*}

First, under the fixed prediction assumption, we can directly solve for the equilibrium of the dynamics by setting the time derivative to 0,
\begin{align*}
\label{PCBP_error_equilibrium}
    \epsilon_i^* = \sum_{j \in \mathcal{C}(v_i)} \epsilon^*_j \frac{\partial \hat{v}_i}{\partial v_j}
    \numberthis
\end{align*}
If we compare this to the recursive relationship inherent to reverse-mode AD (Equation \ref{reverse_mode_equation}), we can see that the prediction errors satisfy the same recursive relationship. Since this relationship is recursive, all that is needed for the prediction errors throughout the graph to converge to the backpropagated derivatives is for the prediction errors at the final layer to be equal to the output gradient: $\epsilon^*_L = \frac{\partial \mathcal{L}}{\partial \hat{v}_L}$. 
To see this explicitly, consider a mean-squared-error loss function \footnote{While the mean-squared-error loss function fits most nicely with the Gaussian generative model, other loss functions can be used in practice. If the loss function can be represented as a log probability distribution, then the generative model can be amended to simply set the output distribution to that distribution. If not, then there is no fully consistent generative model (although all nodes except the output remain Gaussian), but the algorithm will still work in practice. See Figure \ref{PCBP_CNN_acc} for results for CNNs trained with a crossentropy loss.}. at the output layer $L = \frac{1}{2}(T - \hat{v}_L)^2$ with T as a vector of targets, and defining $\epsilon_L = T - \hat{v}_L$. We then consider the equilibrium value of the prediction error unit at a penultimate vertex $\epsilon_{L-1}$. By Equation \ref{PCBP_error_equilibrium}, we can see that at equilibrium,
\begin{align*}
    \epsilon^*_{L-1} &= \epsilon^*_L \frac{\partial \hat{v}_L}{\partial v_{L-1}} = (T - \hat{v}_L^*)\frac{\partial \hat{v}_L}{\partial v_{L-1}} \numberthis
\end{align*}
since,  $(T - \hat{v}_L) = \frac{\partial \mathcal{L}}{\partial \hat{v}_L}$, we can then write,
\begin{align*}
     \epsilon^*_{L-1}  &=  \frac{\partial \mathcal{L}}{\partial \hat{v}_L} \frac{\partial \hat{v}_L}{\partial v_{L-1}} =\frac{\partial \mathcal{L}}{\partial v_{L-1}} \numberthis
\end{align*}
Thus the prediction errors of the penultimate nodes converge to the correct backpropagated gradient. Furthermore, recursing through the graph from children to parents allows the correct gradients to be computed\footnote{Some subtlety is needed here since $v_{L-1}$ may have many children which each contribute to the loss. However, these different paths sum together at the node $v_{L-1}$, thus propagating the correct gradient backwards.}. Thus, by induction, we have shown that the fixed points of the prediction errors of the global optimization correspond exactly to the backpropagated gradients. Intuitively, if we imagine the computation-graph as a chain and the error as `tension' in the chain, backprop loads all the tension at the end (the output) and then systematically propagates it backwards. Predictive coding, however, spreads the tension throughout the entire chain until it reaches an equilibrium where the amount of tension at each link is precisely the backpropagated gradient.

By a similar argument, it is apparent that the dynamics of the parameters $\theta_i$ as a gradient descent on $\mathcal{F}$ also exactly match the backpropagated parameter gradients. 
\begin{align*}
    \frac{dW_i}{dt} &= \frac{\partial \mathcal{F}}{\partial W_i} = \epsilon^*_i \frac{\partial \epsilon_i^*}{\partial W_i} \\ 
    &= \frac{\partial \mathcal{L}}{\partial \hat{v}_i} \frac{\partial \hat{v}_i}{\partial  W_i}  
    = \frac{\partial \mathcal{L}}{\partial W_i} \numberthis
\end{align*}
 Which follows from the fact that $\epsilon^*_i = \frac{dL}{d\hat{v}_i}$ and that $\frac{d\epsilon^*_i}{d\theta} = \frac{d \hat{v}_i}{d\theta_i}$.
 
\begin{figure}
\begin{subfigure}{\linewidth}
\centering
\includegraphics[scale=0.15]{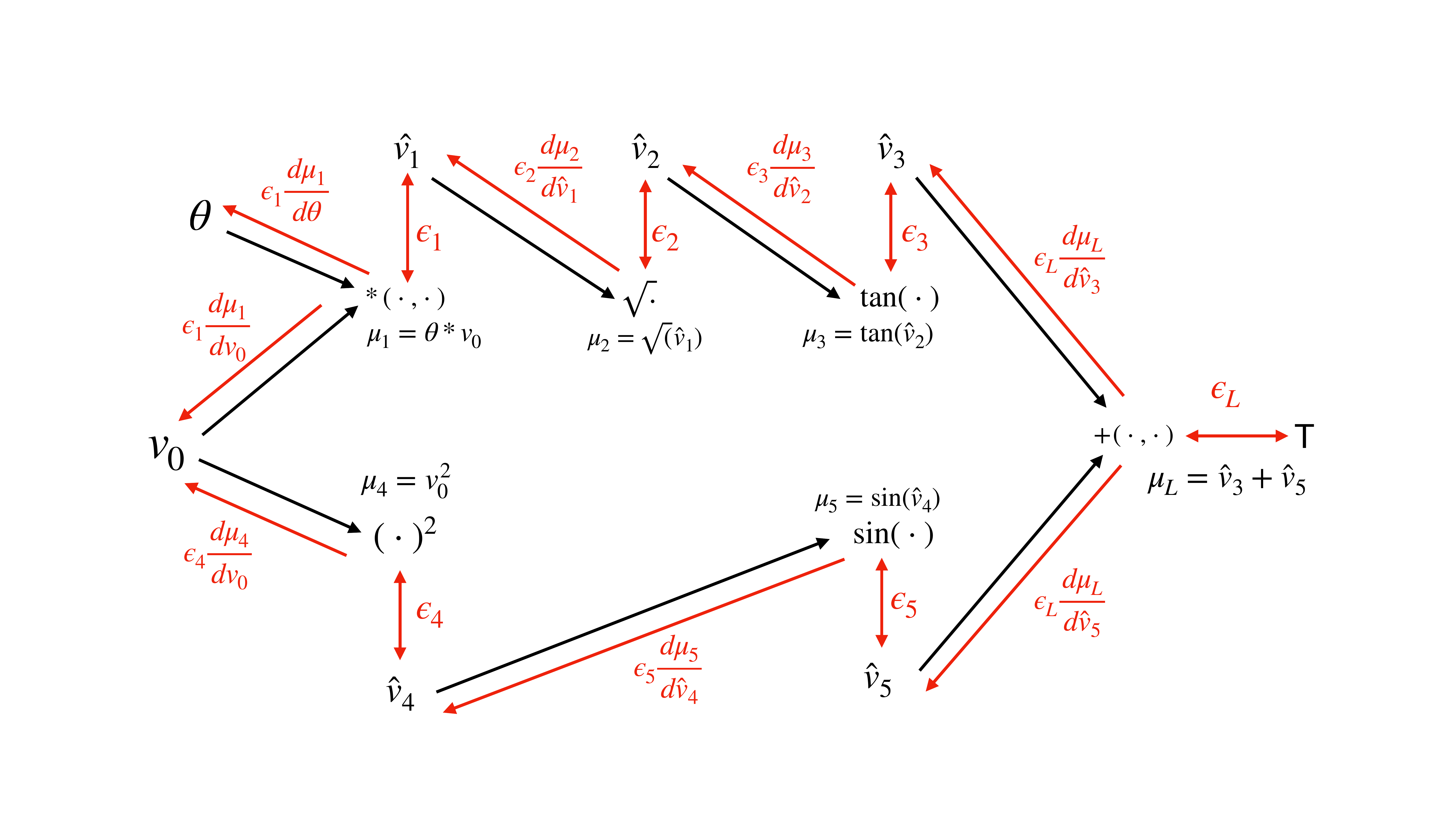}
\end{subfigure}
\begin{subfigure}{.5\linewidth}
\centering
\includegraphics[scale=0.25]{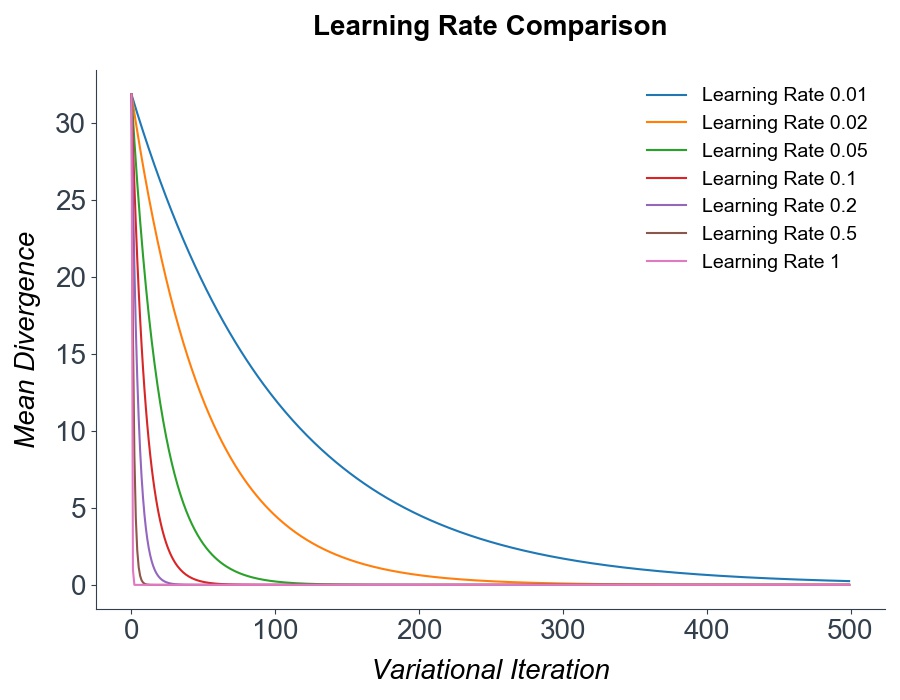}
\end{subfigure}%
\begin{subfigure}{.5\linewidth}
\centering
\includegraphics[scale=0.25]{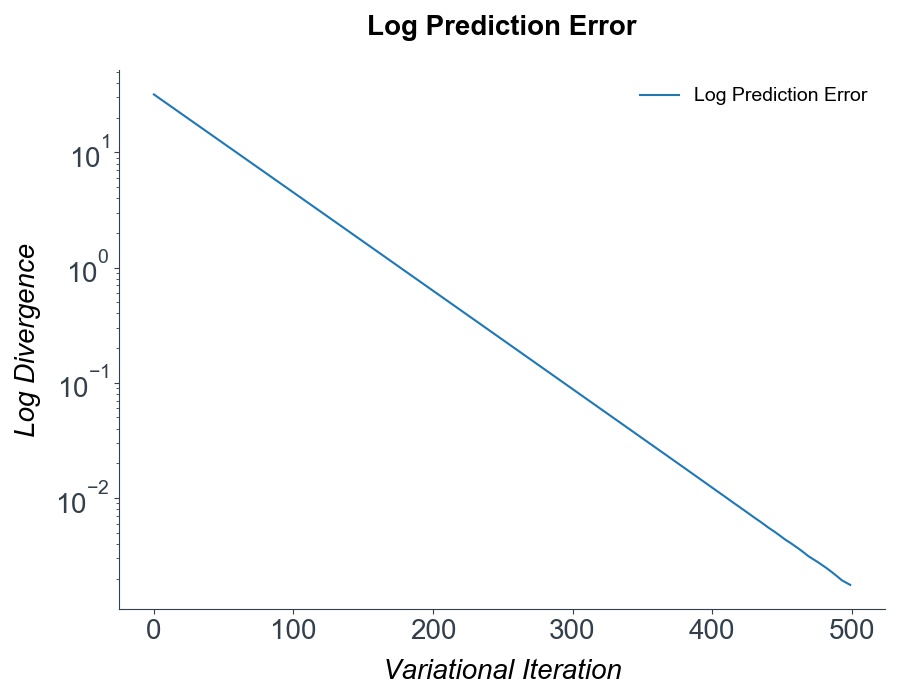}
\end{subfigure}\\[1ex]
\caption{Top: The computation graph of the nonlinear test function $v_{L} = \tan(\sqrt{\theta v_0}) + \sin(v_0^2)$. Bottom: graphs of the log mean divergence from the true gradient and the divergence for different learning rates. Convergence to the exact gradients is exponential and robust to high learning rates.}

 \label{PCBP_numerical_results}
\end{figure}

 To demonstrate empirically that this approach works, we present a numerical test in the simple scalar case, where we use predictive coding to derive the gradients of an arbitrary, highly nonlinear test function $v_{L} = \tan(\sqrt{\theta v_0}) + \sin(v_0^2)$ where $\theta$ is an arbitrary parameter. For our tests, we set $v_0$ to 5 and $\theta$ to 2. The computation graph for this function is presented in Figure \ref{PCBP_numerical_results}. Although simple, this is a good test of predictive coding because the function is highly nonlinear, and its computation graph does not follow a simple layer structure but includes some branching. An arbitrary target of $T = 3$ was set at the output and the gradient of the loss $L= (v_L-T)^2$ with respect to the input $v_0$ was computed by predictive coding. We show (Figure \ref{PCBP_numerical_results}) that the predictive coding optimisation rapidly converges to the exact numerical gradients computed by automatic differentiation, and that moreover this optimization is very robust and can handle even exceptionally high learning rates (up to 0.5) without divergence. 

  \vspace{-0.2cm}

Secondly, we wish to show empirically that this approach can be used to train deep neural network architectures, of the kind used in machine learning, up to high levels of performance equivalent to those trained with backprop. First, we demonstrate a predictive coding CNN model. Convolutional neural networks have been a cornerstone of machine learning since the pioneering demonstrations of their power on ImageNet by \citep{krizhevsky2012imagenet}, and are still widely used as state of the art models for image processing. 

The key concept in a CNN is that of an image convolution, where a small weight matrix is slid (or convolved) across an image to produce an output image. Each patch of the output image only depends on a relatively small patch of the input image. Moreover, the weights of the filter stay the same during the convolution, so each pixel of the output image is generated using the same weights. The weight sharing implicit in the convolution operation enforces translational invariance, since different image patches are all processed with the same weights.

The forward equations of a convolutional layer for a specific output pixel
\begin{align*}
    v_{i,j} = \sum_{k=i-f}^{k=i+f} \sum_{l=j-f}^{l=j+f} \theta_{k,l} x_{i+k,j+l} \numberthis
\end{align*}

Where $v_{i,j}$ is the $(i,j)$th element of the output, $x_{i,j}$ is the element of the input image and $\theta_{k,l}$ is an weight element of a feature map. To set-up a predictive coding CNN, we augment each intermediate $x_i$ and $v_i$ with error units $\epsilon_i$ of the same dimension as the output of the convolutional layer.

Predictions $\hat{v}$ are projected forward using the forward equations. Prediction errors also need to be transmitted backwards for the architecture to work. To achieve this we must have that prediction errors are transmitted upwards by a `backwards convolution'. We thus define the backwards prediction errors $\hat{\epsilon}_j$ as follows:
\begin{align*}
    \hat{\epsilon}_{i,j} = \sum_{k=i-f}^{i+f}\sum_{l=j-f}^{j+f} \theta_{j,i} \tilde{\epsilon}_{i,j} \numberthis
\end{align*}

Where $\tilde{\epsilon}$ is an error map zero-padded to ensure the correct convolutional output size. Inference in the predictive coding network then proceeds by updating the intermediate values of each layer as follows:
\begin{align*}
    \frac{dv_l}{dt} = \epsilon_l - \hat{\epsilon}_{l+1} \numberthis
\end{align*}

The CNN weights can be updated using the simple Hebbian rule of the multiplication of the pre and post synaptic potentials. 
\begin{align*}
    \frac{d\theta_l}{dt} = \sum_{i,j} \epsilon_{l_{i,j}} {v_{l-1}}_{i,j}^T  \numberthis
\end{align*}

In our experiments we used a relatively simple CNN architecture consisting of one convolutional layer of kernel size 5, and a filter bank of 6 filters. This was followed by a max-pooling layer with a (2,2) kernel and a further convolutional layer with a (5,5) kernel and filter bank of 16 filters. This was then followed by three fully connected layers of 200, 150, and 10 (or 100 for CIFAR100) output units. Each convolutional and fully connected layer used the relu activation function, except the output layer which was linear. Although this architecture is far smaller than state of the art for convolutional networks, our primary purpose here is to demonstrate the equivalence of predictive coding and backprop. Further work could investigate scaling up predictive coding to more state-of-the-art architectures.

Our datasets consisted of 32x32 RGB images. We normalised the values of all pixels of each image to lie between 0 and 1, but otherwise performed no other image preprocessing. We did not use data augmentation of any kind. We set the weight learning rate for the predictive coding and backprop networks 0.0001. A minibatch size of 64 was used. These parameters were chosen without any detailed hyperparameter search and so are likely suboptimal. The magnitude of the gradient updates was clamped to lie between -50 and 50 in all of our models. This was done to prevent divergences, as occasionally occurred in the LSTM networks, likely due to exploding gradients. 

First, we compare the test and training accuracy, as well as training loss plots for convolutional CNNs trained on CIFAR10. Figure \ref{PCBP_CNN_acc} shows convincingly that the accuracy and indeed the training dynamics of predictive coding and backprop are the same, to all intents and purposes, thus demonstrating that predictive coding approaches can approximate backprop to a very high accuracy, using only local learning rules. To investigate this further, we explicitly plotted the divergence between the gradient estimates produced by predictive coding and the analytically correct gradients produced by backprop.

 \begin{figure*}
        \centering
        \begin{subfigure}[b]{0.475\textwidth}
            \centering
            \includegraphics[width=\textwidth]{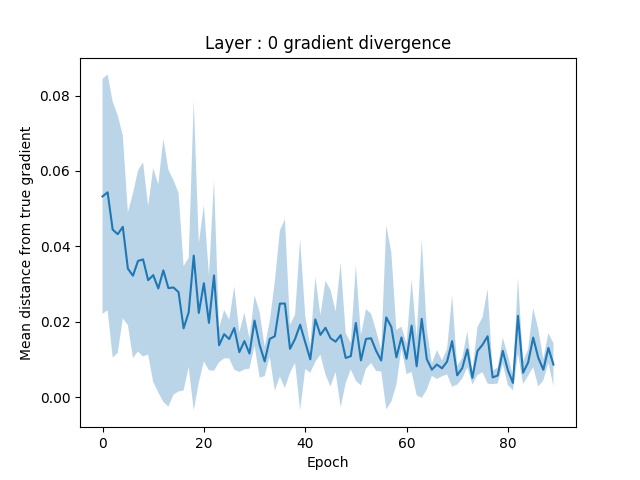}
            \caption[Network2]%
            {{\small Conv Layer 1}}    
        \end{subfigure}
        \hfill
        \begin{subfigure}[b]{0.475\textwidth}  
            \centering 
            \includegraphics[width=\textwidth]{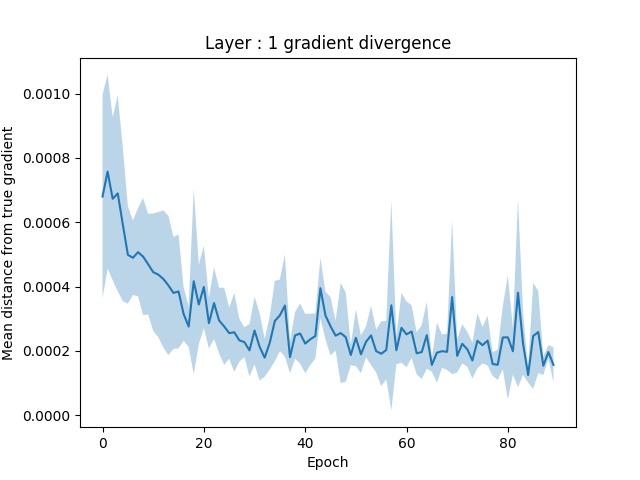}
            \caption[]%
            {{\small Conv Layer 2}}    
        \end{subfigure}
        \vskip\baselineskip
        \begin{subfigure}[b]{0.475\textwidth}   
            \centering 
            \includegraphics[width=\textwidth]{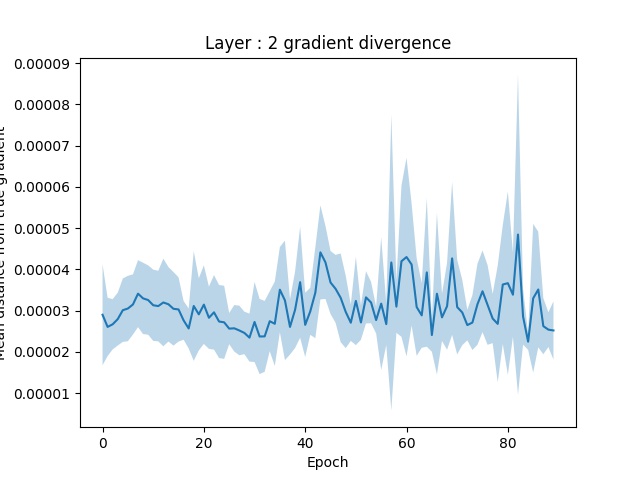}
            \caption[]%
            {{\small FC Layer 1}}    
        \end{subfigure}
        \quad
        \begin{subfigure}[b]{0.475\textwidth}   
            \centering 
            \includegraphics[width=\textwidth]{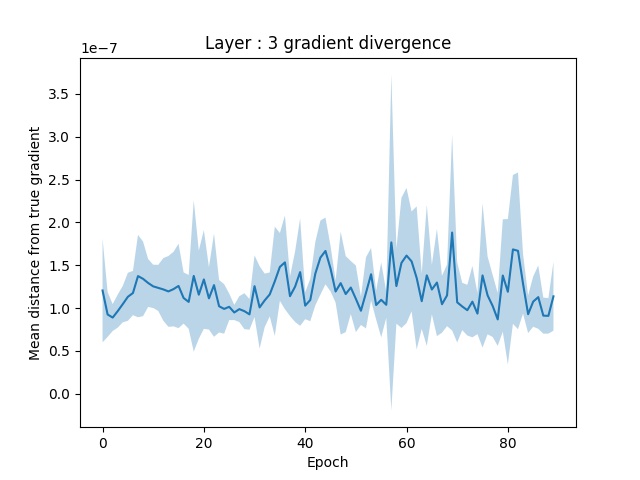}
            \caption[]
            {{\small FC Layer 2}}    
        \end{subfigure}
        \caption{\small Mean divergence between the true numerical and predictive coding backprops over the course of training. In general, the divergence appeared to follow a largely random walk pattern, and was generally neglible. Importantly, the divergence did not grow over time throughout training, implying that errors from slightly incorrect gradients did not appear to compound.} 
       
\label{training_divergence}
    \end{figure*}
    
Importantly, we found that the divergence between the true and predictive coding gradients was extremely small, and remained approximately constant throughout training suggesting that predictive coding networks do not suffer from accumulating errors in their gradient approximation process. Importantly, to achieve this level of convergence required 100 backwards iterations using an inference learning rate of 0.1. This means that the predictive coding has an approximately 100x computational overhead compared to backprop -- largely rendering it uncompetitive for direct competition in serial computers. Nevertheless, this choice of 100 iterations is on the high end of what is necessary, since we are primarily concerned with showing the asymptotic equivalence, and in reality the number of iterations required may be substantially lower. Additionally, predictive coding is a fully parallel algorithm unlike backprop, which must be implemented sequentially and is a better fit for the highly parallel neural circuitry.

It is also important to note that while predictive coding `naturally' uses the mean-squared error loss -- so that the output error is a standard prediction error, other loss functions as possible, such as the widely used cross-entropy loss. Predictive coding can be straightforwardly extended to cover other loss functions by simply replacing the final prediction error $\epsilon_L$ with the gradient of the loss function with respect to the outputs.  Here we demonstrate that predictive coding with a multi-class cross entropy loss also performs equivalently to the network trained with backprop on the CIFAR and SVHN datasets.

\subsection{RNN and LSTM}
\subsubsection{RNN}
We additionally tested a predictive coding RNN and LSTM. To train these recurrent networks with predictive coding, we simply used the approach of using predictive coding to approximate backpropagation through time (BPTT) and applied predictive coding to the unrolled computation graph. With long sequence lengths, this lead to extremely deep graphs for predictive coding to train. Crucially, we demonstrate that predictive coding's ability to train such graphs is not impaired by their depth, meaning that predictive coding as a training algorithm has an exceptional scalability for extremely deep models of the kind increasingly used in contemporary machine learning \citep{radford2019language,he2016deep}

The computation graph on RNNs is relatively straightforward. We consider only a single layer RNN here although the architecture can be straightforwardly extended to hierarchically stacked RNNs. An RNN is similar to a feedforward network except that it possesses an additional hidden state $h$ which is maintained and updated over time as a function of both the current input $x$ and the previous hidden state. The output of the network $y$ is a function of $h$. By considering the RNN at a single timestep we obtain the following equations.
\begin{align*}
    \numberthis
    h_t &= f(\theta_h h_{t-1} + \theta_x x_t) \\
    y_t &= g(\theta_y h_t) \numberthis
\end{align*}
Where f and g are elementwise nonlinear activation functions. And $\theta_h, \theta_x, \theta_y$ are weight matrices for each specific input. To predict a sequence the RNN simply rolls forward the above equations to generate new predictions and hidden states at each timestep.

It is important to note that this is an additional aspect of biological implausibility that we do not address in here. BPTT requires updates to proceed backwards through time from the end of the sequence to the beginning. Ignoring any biological implausibility with the rules themselves, this updating sequence is clearly not biologically plausible as naively it requires maintaining the entire sequence of predictions and prediction errors perfectly in memory until the end of the sequence, and waiting until the sequence ends before making any updates. There is a small literature on trying to produce biologically plausible, or forward-looking approximations to BPTT which does not require updates to be propagated back through time \citep{williams1989learning,lillicrap2019backpropagation,steil2004backpropagation,ollivier2015training,tallec2017unbiased}. While this is a fascinating area, we do not address it here. We are solely concerned with the fact that predictive coding approximates backpropagation on feedforward computation graphs for which the unrolled RNN graph is a sufficient substrate.

To learn a predictive coding RNN, we first augment each of the variables $h_t$ and $y_t$ of the original graph with additional error units $\epsilon_{h_t}$ and $\epsilon_{y_t}$. Predictions $\hat{y}_t, \hat{h}_t$ are generated according to the feedforward rules (16). A sequence of true labels $\{T_1...T_T\}$ is then presented to the network, and then inference proceeds by recursively applying the following rules backwards through time until convergence.
\begin{align*}
    \epsilon_{y_t} &= L - \hat{y}_{t} \\
    \epsilon_{h_t} &= h_t - \hat{h}_{t} \\
    \frac{dh_t}{dt} &= \epsilon_{h_t} - \epsilon_{y_t} \theta_y^T - \epsilon_{h_{t+1}} \theta_h^T \numberthis
\end{align*}

Upon convergence the weights are updated according to the following rules.
\begin{align*}
    \frac{d\theta_y}{dt} &= \sum_{t=0}^T \epsilon_{y_t} \frac{\partial g(\theta_y h_t)}{\partial \theta_y} h_t^T \\
    \frac{d\theta_x}{dt} &= \sum_{t=0}^T \epsilon_{h_t} \frac{\partial f(\theta_h h_{t-1}+\theta_x x_t)}{\partial \theta_x} x_t^T \\
    \frac{d\theta_h}{dt} &= \sum_{t=0}^T \epsilon_{h_t} \frac{\partial f(\theta_h h_{t-1}+\theta_x x_t)}{\partial \theta_h} h_{t+1}^T \numberthis
\end{align*}

Since the RNN feedforward updates are parameter-linear, these rules are Hebbian, only requiring the multiplication of pre and post-synaptic potentials. This means that the predictive coding updates proposed here are biologically plausible and could in theory be implemented in the brain. The only biological implausibility remains the BPTT learning scheme.

Our RNN was trained on a simple character-level name-origin dataset which can be found here: \textit{https://download.pytorch.org/tutorial/data.zip}. The RNN was presented with sequences of characters representing names and had to predict the national origin of the name -- French, Spanish, Russian, etc. The characters were presented to the network as one-hot-encoded vectors without any embedding. The output categories were also presented as a one-hot vector. The RNN has a hidden size of 256 units. A \textit{tanh} nonlinearity was used between hidden states and the output layer was linear. The network was trained on randomly selected name-category pairs from the dataset. 

We first present the training and test accuracy for the backprop RNNs, averaged over five seeds. In general, performance between the backprop-trained and predictive coding networks was indistinguishable on this task.

\begin{figure}[ht]
\vspace{-0.3cm}
\hspace{-0.6cm}
\begin{subfigure}{.5\textwidth}
  \centering
  \includegraphics[width=1\linewidth]{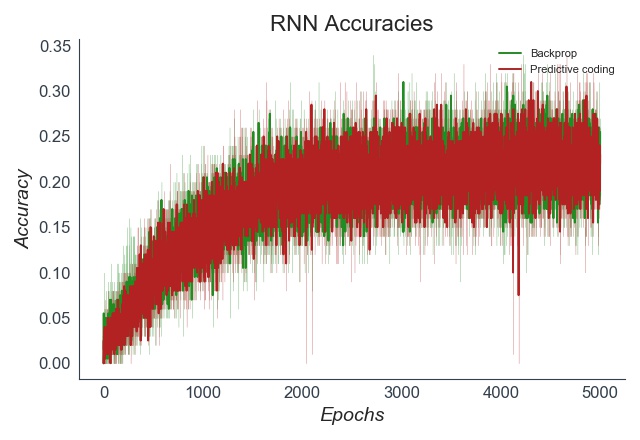}  
\end{subfigure}
\begin{subfigure}{.5\textwidth}
  \centering
  \includegraphics[width=1\linewidth]{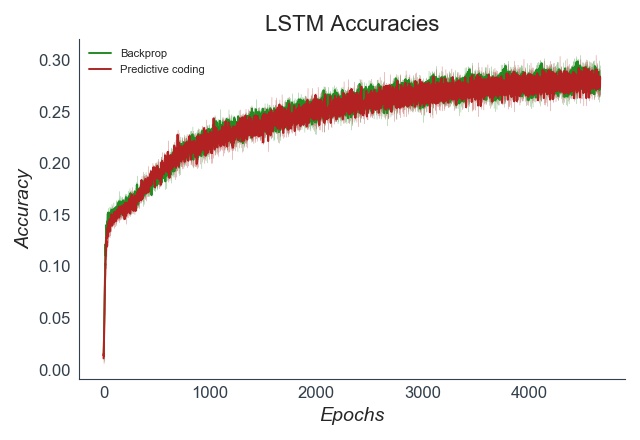}  
\end{subfigure}
\caption{Test accuracy plots for the Predictive Coding and Backprop RNN and LSTM on their respective tasks, averaged over 5 seeds. Performance is again indistinguishable from backprop.}

\label{pc_rnn_lstm_results_figure}
\vspace{-0.3cm}
\end{figure}

\begin{figure}[ht]
  \centering
  \includegraphics[width=.7\linewidth]{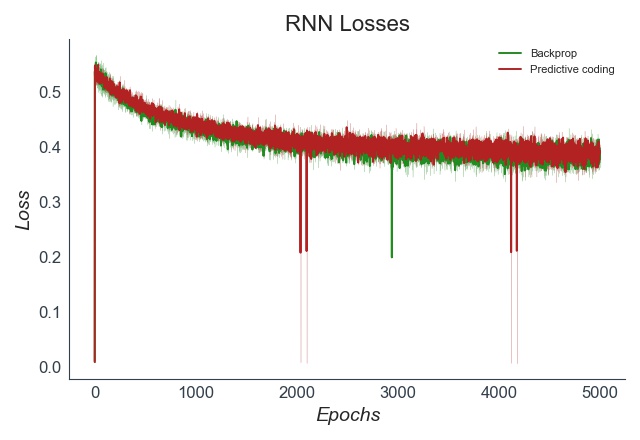}  
\caption{Training losses for the predictive coding and backprop RNN. As expected, they are effectively identical.}

\label{rnn_losses}
\end{figure}

Additionally, The training loss for the predictive coding and backprop RNNs, averaged over 5 seeds is presented below (Figure \ref{rnn_losses}).

\subsubsection{LSTM}
\begin{figure}[ht]
  \centering
  \includegraphics[width=1\linewidth]{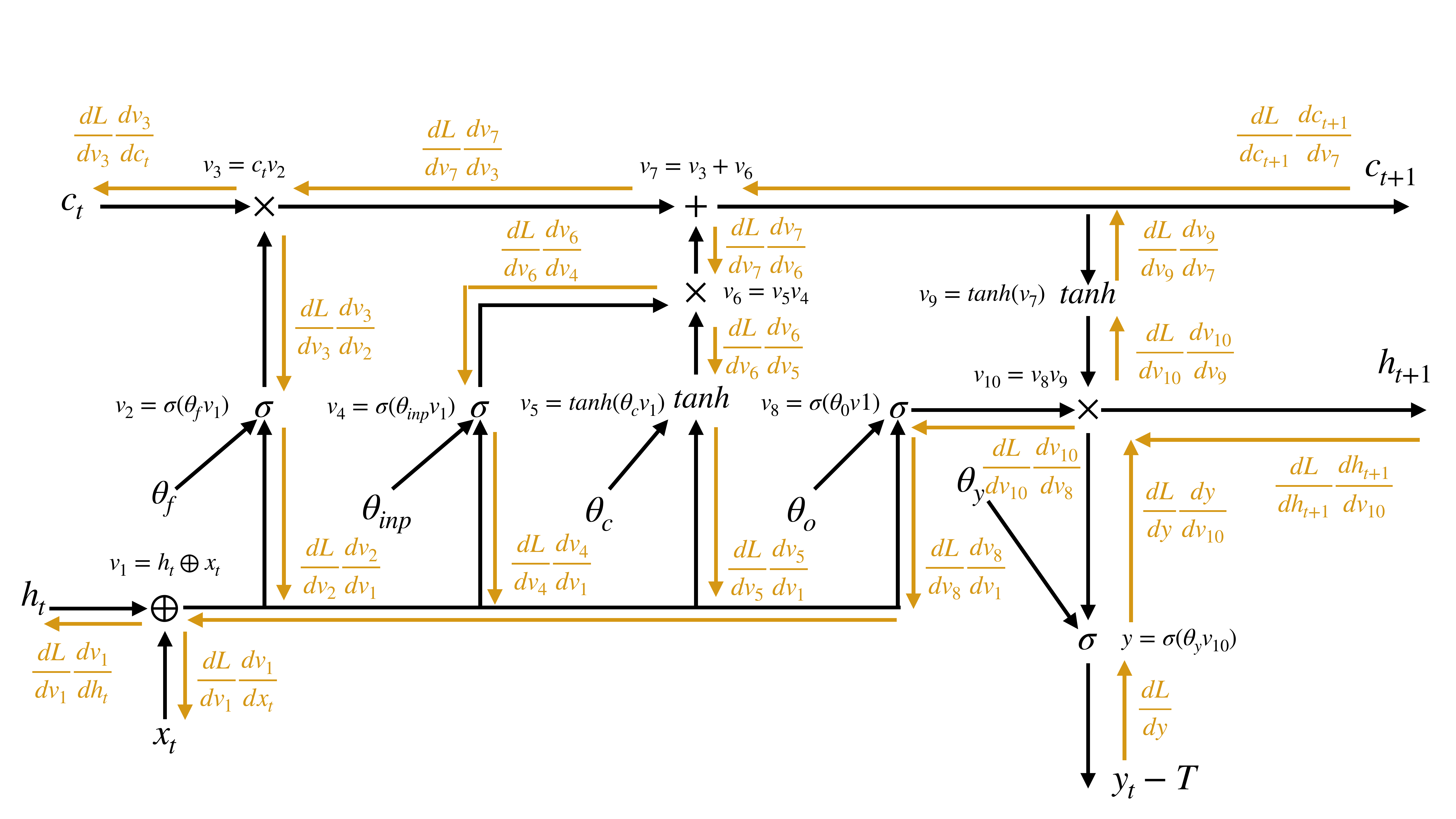}  
\caption{Computation graph and backprop learning rules for a single LSTM cell. Inputs to the LSTM cell are the current input $x_t$ and the previous embedding $h_t$. These are then passed through three gates -- an input, forget, and output gate, before the output of the whole LSTM cell can be computed}

\label{PC_LSTM}
\end{figure}
\begin{figure}
  \centering
  \includegraphics[width=1\linewidth]{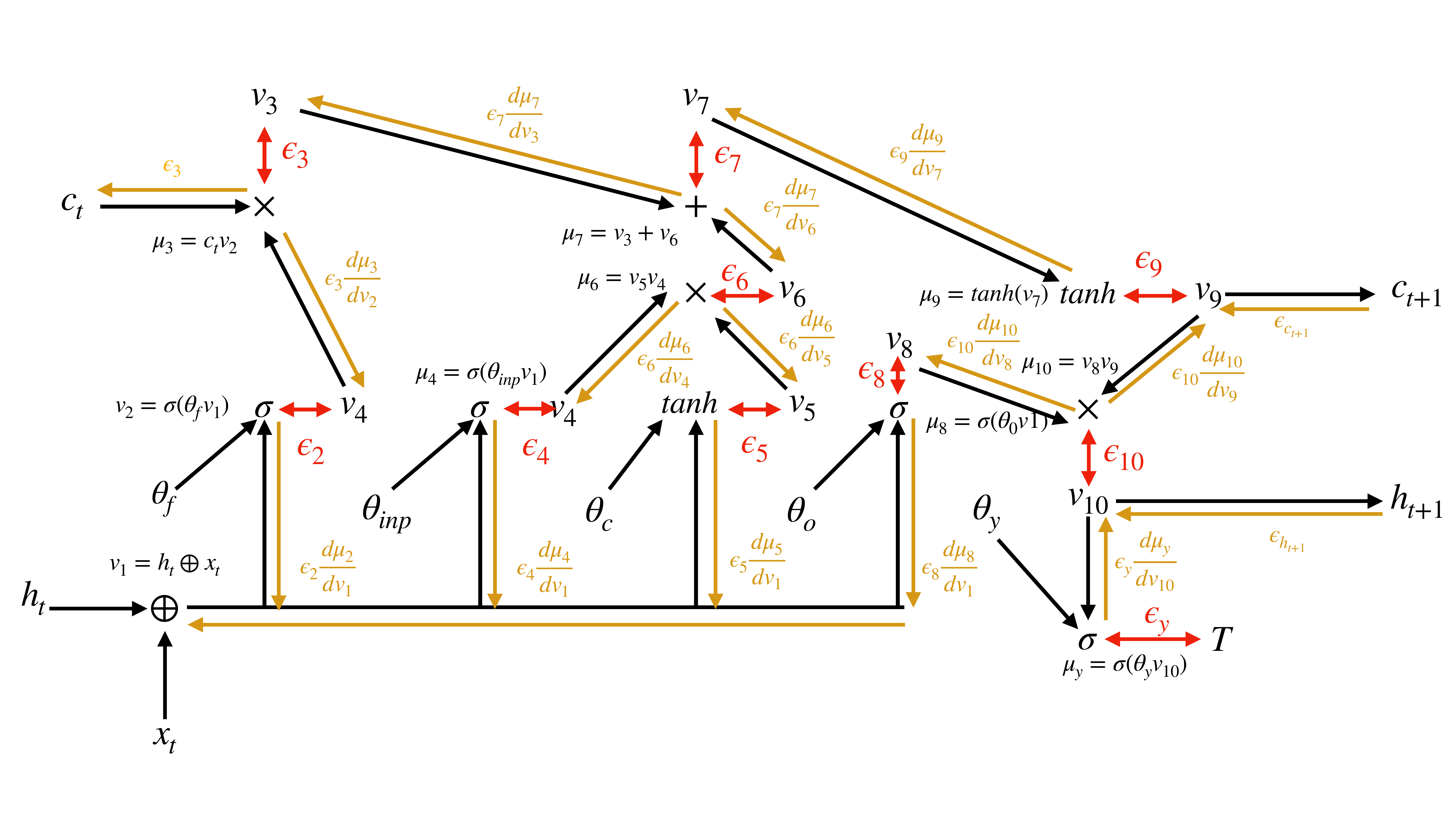}  
\caption{The LSTM cell computation graph augmented with error units, evincing the connectivity scheme of the predictive coding algorithm. The key move is to associate each intermediate node in the computation graph with its own prediction error unit}
\label{pc_lstm}
\end{figure}

Unlike the other two models, the LSTM possesses a complex and branching internal computation graph, and is thus a good opportunity to make explicit the predictive coding `recipe' for approximating backprop on arbitrary computation graphs. The computation graph for a single LSTM cell is shown (with backprop updates) in Figure \ref{pc_lstm}. Prediction for the LSTM occurs by simply rolling forward a copy of the LSTM cell for each timestep. The LSTM cell receives its hidden state $h_t$ and cell state $c_t$ from the previous timestep. During training we compute derivatives on the unrolled computation graph and receive backwards derivatives (or prediction errors) from the LSTM cell at time $t+1$. For a full and detailed set of equations specifying the complex LSTM cell, see Appendix B.

The recipe to convert this computation graph into a predictive coding algorithm is straightforward. We first rewire the connectivity so that the predictions are set to the forward functions of their parents. We then compute the errors between the vertices and the predictions. 

During inference, the inputs $h_t$,$x_t$ and the output $y_t$ are fixed. The vertices and then the prediction errors are updated. This recipe is straightforward and can easily be extended to other more complex machine learning architectures. The full augmented computation graph, including the vertex update rules, is presented in Figure \ref{PC_LSTM}.

For the LSTM we also observed a close correspondence between the performance (in terms of training and test accuracy) between the predictive coding and backpropagation networks, thus demonstrating that predictive coding can converge to the exact backprop gradients even on exceptionally deep and complex computation graphs such as the LSTM

\begin{figure}[ht]
  \centering
  \includegraphics[width=0.7\linewidth]{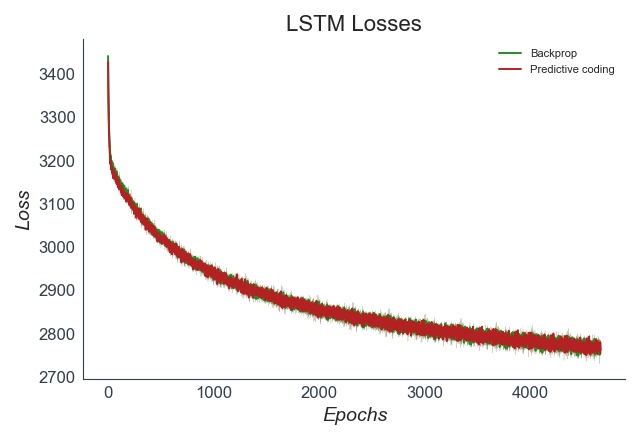}  
\caption{Training losses for the predictive coding and backprop LSTMs averaged over 5 seeds. The performance of the two training methods is effectively equivalent.}
\label{LSTM_losses}
\end{figure}

Importantly, we observed rapid convergence to the exact backprop gradients even in the case of very deep computation graphs (as is an unrolled LSTM with a sequence length of 100). Although convergence was slower than was the case for CNNs or lesser sequence lengths, it was still straightforward to achieve convergence to the exact numerical gradients with sufficient iterations.

Below we plot the mean divergence between the predictive coding and true numerical gradients as a function of sequence length (and hence depth of graph) for a fixed computational budget of 200 iterations with an inference learning rate of 0.05. As can be seen, the divergence increases roughly linearly with sequence length. Importantly, even with long sequences, the divergence is not especially large, and can be decreased further by increasing the computational budget. As the increase is linear, we believe that predictive coding approaches should be scalable even for backpropagating through very deep and complex graphs.

\begin{figure}[ht]
  \centering
  \includegraphics[width=.9\linewidth]{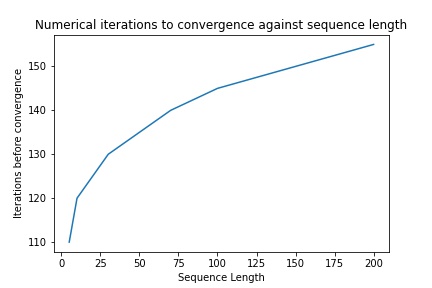}  
\caption{Number of iterations to reach convergence threshold as a function of sequence length. Importantly, the number of iterations required to converge appears to grow sublinearly with sequence length, again implying that convergence is not computationally unattainable even with very long sequences.}
\label{num_iterations_to_converge}
\end{figure}

We also plot the number of iterations required to reach a given convergence threshold (here taken to be 0.005) as a function of sequence length (Figure \ref{num_iterations_to_converge}). We see that the number of iterations required increases sublinearly with the sequence length, and likely asymptotes at about 300 iterations. Although this is a lot of iterations, the sublinear convergence nevertheless shows that the method can scale to even extremely deep graphs.

\begin{figure}[ht]
  \centering
  \includegraphics[width=.9\linewidth]{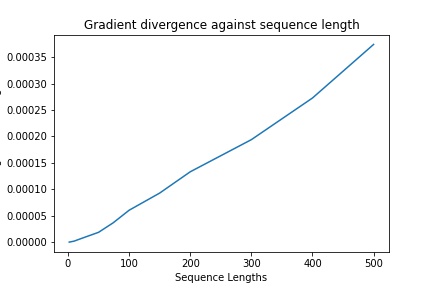}  
\caption{Divergence between predictive coding and the correct backprop gradients as a function of sequence length. Crucially, this divergence only increases linearly in the sequence length, allowing for very accurate gradient computation even with long sequences.}
\label{sequence_length_effect}

\end{figure}

\begin{figure}[ht]
  \centering
  \includegraphics[width=0.7\linewidth]{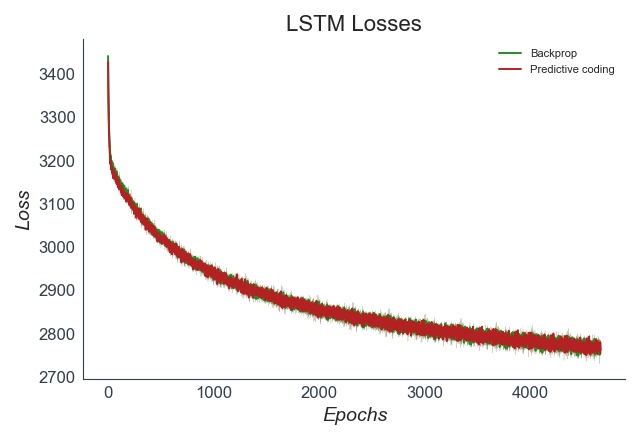}  
\caption{Training losses for the predictive coding and backprop LSTMs averaged over 5 seeds. The performance of the two training methods is effectively equivalent.}
\label{pcbp_lstm_losses}
\end{figure}

Our architecture consisted of a single LSTM layer (more complex architectures would consist of multiple stacked LSTM layers).
The LSTM was trained on a next-character character-level prediction task. The dataset was the full works of Shakespeare, downloadable from Tensorflow. The text was shuffled and split into sequences of 50 characters, which were fed to the LSTM one character at a time. The LSTM was trained then to predict the next character, so as to ultimately be able to generate text. The characters were presented as one-hot-encoded vectors. The LSTM had a hidden size and a cell-size of 1056 units. A minibatch size of 64 was used and a weight learning rate of 0.0001 was used for both predictive coding and backprop networks. To achieve sufficient numerical convergence to the correct gradient, we used 200 variational iterations with an inference learning rate of 0.1. This rendered the predictive LSTM approximately 200x as costly as the backprop LSTM to run. A graph of the LSTM training loss for both predictive coding and backprop LSTMs, averaged over 5 random seeds, can be found below (Figure \ref{pcbp_lstm_losses}). 

\section{Interim Discussion}

Here we have shown that predictive coding can be applied directly to arbitrary computation graphs and can rapidly and effectively converge to the exact gradients required for the backpropagation of error algorithm. We have demonstrated this on deep and state of the art machine learning architectures, thus achieving significantly greater scale than previous works using predictive coding \citep{millidge2019implementing,orchard2019making,whittington2017approximation}. Moreover, the predictive coding learning rule uses only local learning dynamics and Hebbian weight updates in the case of the usual feedforward neural networks (although they differ somewhat for the LSTM). Weights are updated using only local prediction errors. 

This approach also is innovative in that it phrases backprop in terms of variational inference on the values of the nodes in the computation graph. While it may seem just like a mathematical convenience, it actually has deep implications. It draws another link between the processes of optimization and variational inference, in a rather different manner from that which has largely been explored before. Instead of conceptualising inference as optimization, as is typically done in variational inference, we instead conceptualize a core component of optimization -- credit assignment -- purely in terms of inference. While this duality has been considered before for two layer networks \citep{amari1995information}, our approach is substantially more powerful and general, by showcasing that it holds for arbitrary computation graphs. Additionally, our approach provides an avenue for interesting generalizations of backprop through the use of precision parameters in predictive coding. Note that in our analysis, we have implicitly assumed that all of the precision parameters are set to the identity $\Sigma = \mathbb{I}$, as they do not feature in the learning and update rules, as they do in Chapter 3. If we reintroduce precision in this context, we see that it has the role of modulating gradient magnitudes -- effectively implementing an adaptive learning rate. What this means, intuitively, is that we can think about precision weighting in this case as enabling an \emph{uncertainty aware backprop}, which specifically weights gradients by how uncertain they are -- or by their variance. Effectively, this method, with learnable precisions, would down-weight highly variable and uncertain gradients while upweighting those known to be certain. When applied to the input this would mimic features of attention, by downweighting noisy or otherwise uncertain inputs and having them play little role in learning. While such an adaptive modulatory role for precision may bring learning benefits, this must be explored further, as the author hopes to do in future work.

Finally, it is worth discussing several drawbacks of the method. The key one is its computational cost. The networks presented here were trained with 100 dynamical iterations to converge to the prediction error equilibrium before each weight update, giving predictive coding an approximately 100x computational cost compared to backprop. This is obviously highly significant and renders these approaches unusable for large scale networks on serial Von-Neumann computers. While the brain utilizes highly parallel circuitry, and may therefore be more suited to such an iterative algorithm, there are still issues with requiring a dynamical iteration for convergence. Specifically, such iterations still require time and discrete phases so that the system cannot likely simply operate in continuous time. Moreover, if too many iterations are required, since the brain must respond to a continually changing world instead of just single images presented in isolation, it may become overwhelmed by events and fail computationally, if the input changes faster than it can dynamically converge to a solution. While not necessarily as severe as needing dynamical approaches for \emph{inference}, which would entirely hamstring any response, here the dynamics are only required for learning and weight updates, this may nevertheless prove to be a substantial drawback of the method. Our method, like most others, also requires two distinct phases which must either be somehow coordinated explicitly, or multiplexed in the brain.

Additionally, the fixed-prediction assumption embedded in the model requires maintaining the stored memory of the feedforward pass values somewhere in the network throughout the backwards dynamical phase which is potentially problematic in neural circuitry. Finally, although the predictive coding learning rules are local, in the sense that they only require information from the same layer, they still require information from the prediction error units to be transmitted to the activity units, where we assume the synaptic weights are located (although in a segregated dendrite model they could just be in different dendrites \citep{sacramento2018dendritic}).

\section{Activation Relaxation}
Here we introduce a second iterative algorithm which approximates the exact backpropagation gradients asymptotically at the equilibrium of a dyanmical system. We call this algorithm \emph{Activation Relaxation} (AR) because of the nature of the update rules, which iteratively update the \emph{activations} of neurons in the iterative phase rather than prediction errors. Crucially, the update rules proposed by AR are exceedingly simple and elegant, and do not require additional populations of `error neurons' as in predictive coding, or multiple backwards phases as in Equilibirium-prop. AR arises quite straightforwardly by trying to take a first principles approach to the iterative backprop approximation schemes.

To establish notation, we consider the simple case of a fully-connected deep multi-layer perceptron (MLP) composed of $L$ layers of rate-coded neurons trained in a supervised setting. The firing rates of these neurons are represented as a single scalar value $x^l_i$, referred to as the neurons activation,  and a vector of all activations at given layer is denoted as $x^l$. The activations of the hierarchically superordinate layer are a function of the hierarchically subordinate layers activations $x^{l+1} = f(W^l x^l)$, where $W^l \in \Theta$ is the set of synaptic weights, and the product of activation and weights is transformed through a nonlinear activation function $f$. The final output $x^L$ of the network is compared with the desired targets $T$, according to some loss function $\mathcal{L}(x^L, T)$. In this work, we take this loss function to be the mean-squared-error (MSE) $\mathcal{L}(x^L, T) = \frac{1}{2}\sum_i (x^L_i - T_i)^2$, although the algorithm applies to any other loss function without loss of generality (see Appendix B). We denote the gradient of the loss with respect to the output layer as $\frac{dL}{dx^L}$. In the case of the MSE loss, the gradient of the output layer is just the prediction error $\epsilon^L = (x^L - T)$.

Firstly, we know that the key quantity we wish to approximate is the adjoint term $\frac{\partial \mathcal{L}}{\partial x_l}$ for a given layer $l$. If we know this adjoint, and have it present somewhere in the local environment, then the gradient with respect to the weights can be computed using only locally available information. In predictive coding, we compute this adjoint term using the recursive relationship of the prediction errors. Here, we take a different approach. Instead, we ask  what is the simplest possible dynamical system which can converge to the exact adjoint term at the equilibrium. After some thought, we emerge at a straightforward leaky integrator model.
\begin{align*}
\label{AR_initial_equation}
    \frac{d x^l}{dt} &= -x^l + \frac{\partial \mathcal{L}}{\partial x^l} \numberthis 
 \end{align*}
    which, at equilibrium, converges to \vspace{-0.2cm}
\begin{align*}
\label{AR_equilibrium_equation}
     \frac{dx^l}{dt} =0 &\implies {x^*}^l = \frac{\partial \mathcal{L}}{\partial x^l} \numberthis
\end{align*}

This update rule includes the very adjoint term we are trying to compute, however, so these dynamics are not immediately computable. To make them so, we first split up the adjoint. By the chain rule, we can write Equation \ref{AR_initial_equation} as,
\begin{align*}
    \frac{dx^l}{dt} = -x^l + \frac{\partial \mathcal{L}}{\partial x^{l+1}}\frac{\partial x^{l+1}}{\partial x^l} \Bigr|_{x^l=\bar{x}^l} \numberthis
\end{align*}
where $\bar{x}^l$ is the value of $x^l$ computed in the forward pass. Next, we note that if we use the \emph{activations of the neurons at each layer} instead of prediction errors to accumulate the adjoint, we can express this in terms of the equilibrium activation of the superordinate layer,
    \begin{align*}
    \frac{dx^l}{dt} =  -x^l + {x^*}^{l+1} \frac{\partial x^{l+1}}{\partial x^l} \Bigr|_{x^l=\bar{x}^l} \numberthis
\end{align*}
However, to achieve these dynamics exactly in a multilayered network would require the sequential convergence of layers, as each layer must converge to equilibrium before the dynamics of the layer below can operate. This sequential convergence would make the algorithm no better than the sequential backwards sweep of backprop. However, if we approximate the equilibrium activations of the layer with the current activation, this allows us to run all layers in parallel, yielding,
\begin{align*}
\label{AR_main_equation}
    \frac{dx^l}{dt} &=  -x^l + {x^*}^{l+1} \frac{\partial x^{l+1}}{\partial x^l}\Bigr|_{x^l=\bar{x}^l} \\
    &\approx -x^l + x^{l+1}\frac{\partial x^{l+1}}{\partial x^l} \Bigr|_{x^l=\bar{x}^l} \numberthis \\
    &\approx -x^l + x^{l+1}f'(W^l, \bar{x}^l) {W^l}^T  \numberthis
\end{align*}
where $f' = \frac{\partial f'(W^l, \bar{x}^l)}{\partial \bar{x}^l}$ represents the partial derivative of the postsynaptic activation with respect to the presynaptic activation. Despite this approximation, we argue that the system nevertheless converges to the same optimum as Equation \ref{AR_equilibrium_equation}. Specifically, because we evaluate $\frac{\partial x^{l+1}}{\partial x^l}$ at the feedforward pass value $\bar{x}^l$, this term remains constant throughout the relaxation phase \footnote{The need to keep this term fixed throughout the relaxation phase does present a potential issue of biological plausibility. In theory it could be maintained by short-term synaptic traces, and for some activation functions such as rectified linear units it is trivial. Moreover, later we show that this term can be dropped from the equations without apparent ill-effect}. Keeping this term fixed effectively decouples each layer from any bottom-up influence. If the top-down input is also constant, because it has already converged so that $x^{l+1} \approx {x^{l+1}}^*$, then the dynamics become linear, and the system is globally stable due to possessing a Jacobian which is everywhere negative-definite. The top-layer is provided with the stipulatively correct gradient, so it must converge. Recursing backwards through each layer, we see that once the top-level has converged, so too must the penultimate layer, and so through to all layers.

Crucially, Equation \ref{AR_main_equation}, which is core to the AR algorithm is extremely simple and biologically plausible. It only requires that the activations of a given layer are sensitive to the difference between their own activity and that of the layer above mapped through the backwards weights, and modulated by the nonlinear derivative of the postsynaptic potential. This update rule thus functions as a kind of prediction error, but one that emerges between layers, rather than being represented by specific prediction error units at a given layer.

Computationally, the AR algorithm proceeds as follows. First, a standard forward pass computes the network output, which is compared with the target to calculate the top-layer error derivative $\epsilon_L$ and thus update the activation of the penultimate layer. \footnote{This top-layer error is simply a prediction error for the MSE loss, but may be more complicated and less biologically-plausible for arbitrary loss functions}. Then, the network enters into a relaxation phase where Equation \ref{AR_main_equation} is iterated globally for all layers until convergence for each layer. 
Upon convergence, the activations of each layer are precisely equal the backpropagated derivatives, and are used to update the weights (via Equation (\ref{BP_weights_equation}).
\begin{align*}
\label{BP_weights_equation}
    \frac{\partial L}{\partial W^l} &= \frac{\partial L}{\partial x^{l+1}}\frac{\partial x^{l+1}}{\partial W^l} \\
    &= \frac{\partial L}{\partial x^{l+1}}f'(W^l x^l) {x^L}^T \numberthis
\end{align*}
\newline
\begin{algorithm}[H]
\DontPrintSemicolon
\SetAlgoLined
\textbf{Dataset} $\mathcal{D} = \{\mathbf{X},\mathbf{T}\}$, parameters $\Theta = \{W^0 \dots W^L\}$, inference learning rate $\eta_x$, weight learning rate $\eta_\theta$.
\BlankLine
\For{$(x^0, t \in \mathcal{D})$}{
    \For{$(x^l,W^l)$ for each layer}{
        $x^{l+1} = f(W^l, x^l)$ 
    }
    \While{not converged}{
        $\epsilon^L = T - x^L $ \\ 
        $ dx^L= -x^L +  \epsilon^L \frac{\partial \epsilon^L}{\partial x^L} $ \\
        \For{$x^l, W^l, x^{l+1}$ for each layer}{
         $dx^l = -x^l +  x^{l+1} \frac{\partial x^{l+1} }{\partial x^l}$ \\
        ${x^l}^{t+1} \leftarrow {x^l}^t + \eta_x dx^l $
        }
    }
    \For{$W^l \in \{W^0 \dots W^L \}$}{
        ${W^l}^{t+1} \leftarrow {W^l}^t + \eta_\theta x^l \frac{\partial x^l}{\partial W^l}$ 
    }
    }
\caption{Activation Relaxation}
\end{algorithm} 

\subsection{Method and Results}

We first demonstrate that our algorithm can train a deep neural network with equal performance to backprop. For training, we utilised the MNIST and Fashion-MNIST \citep{xiao2017online} datasets. The MNIST dataset consists of 60000 training and 10000 test 28x28 images of handwritten digits, while the Fashion-MNIST dataset consists of 60000 training and 10000 test 28x28 images of clothing items. The Fashion-MNIST dataset is designed to be identical in shape and size to MNIST while being harder to solve. We used a 4-layer fully-connected multi-layer perceptron (MLP) with rectified-linear activation functions and a linear output layer. The layers consisted of 300, 300, 100, and 10 neurons respectively. In the dynamical relaxation phase, we integrate  Equation \ref{AR_main_equation} with a simple first-order Euler integration scheme. ${x^l}^{t+1} = {x^l}^t - \eta_x \frac{dx^l}{dt}$ where $\eta_x$ was a learning rate which was set to $0.1$. The relaxation phase lasted for 100 iterations, which we found sufficient to  closely approximate the numerical backprop gradients. After the relaxation phase was complete, the weights were updated using the standard stochastic gradient descent optimizer, with a learning rate of 0.0005. The weights were initialized as draws from a Gaussian distribution with a mean of 0 and a variance of 0.05.  Hyperparameter values were chosen based on initial intuition and were not found using a grid-search. 
The AR algorithm was applied to each minibatch of 64 digits sequentially. The network was trained with the mean-squared-error loss.
\begin{figure}
\centering
\subfloat[MNIST train accuracy]{\includegraphics[width=0.3\linewidth]{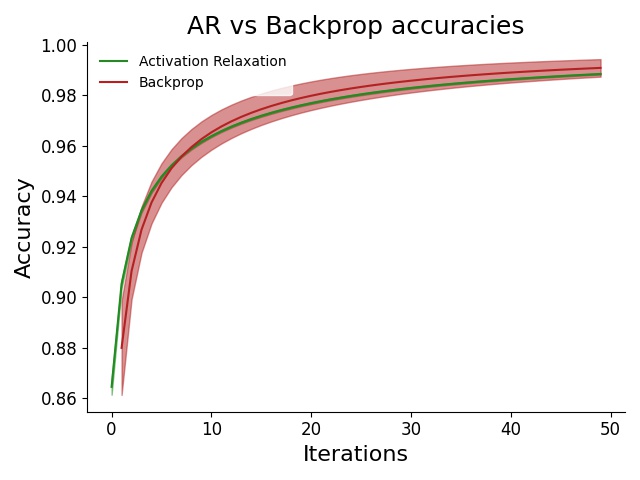}}\hfil
\subfloat[MNIST test accuracy]{\includegraphics[width=0.3\linewidth]{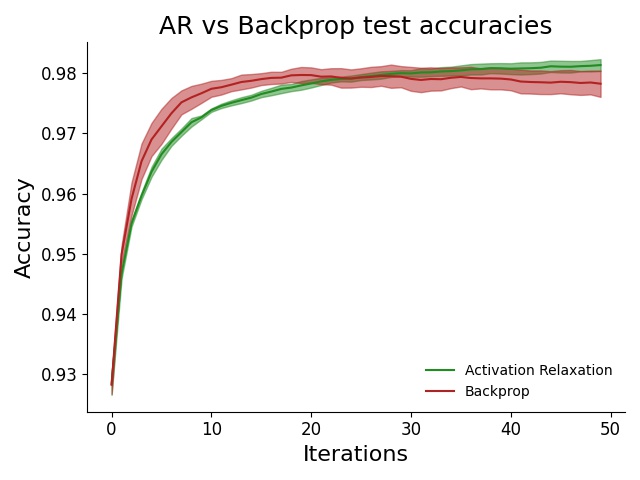}}\hfil 
\subfloat[MNIST gradient angle]{\includegraphics[width=0.3\linewidth]{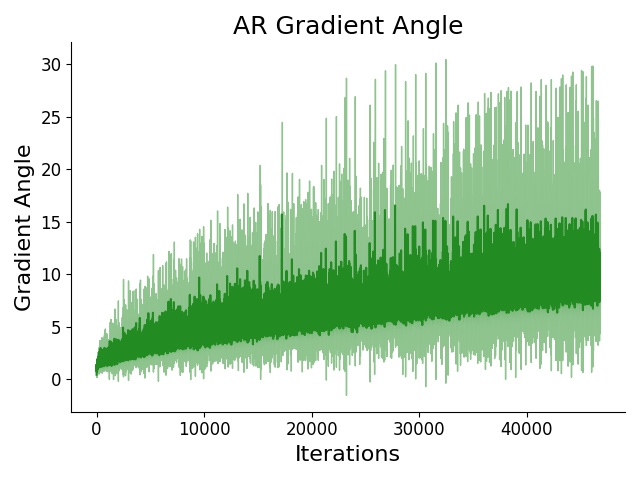}} 

\subfloat[Fashion train accuracy]{\includegraphics[width=0.3\linewidth]{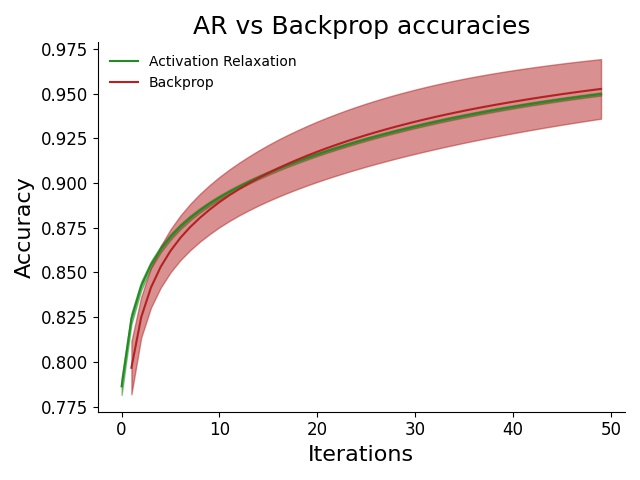}}\hfil
\subfloat[Fashion test accuracy]{\includegraphics[width=0.3\linewidth]{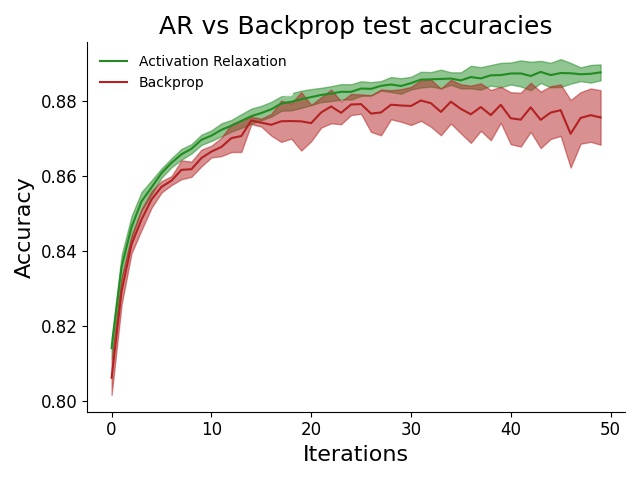}}\hfil AR/
\subfloat[Fashion gradient angle]{\includegraphics[width=0.3\linewidth]{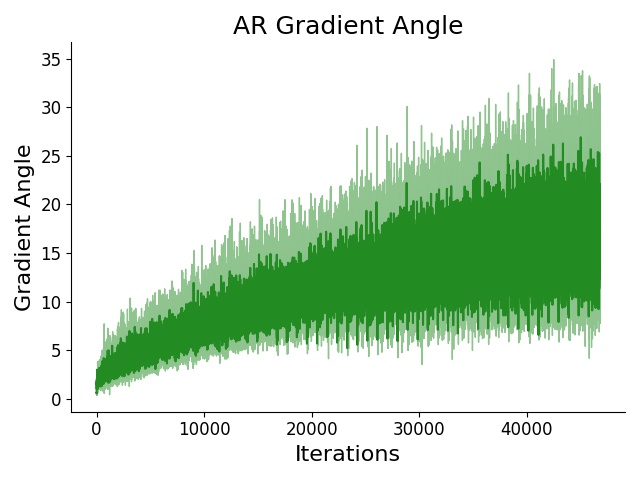}}
\caption{Train and test accuracy and gradient angle (cosine similarity) for AR vs backprop for MNIST and Fashion-MNIST datasets. Importantly the training and test accuracies are virtually identical between the AR-trained and backprop-trained networks. Additionally, the gradient angle between the AR update and backprop is always substantially less than the 90 degrees required to allow for learning.}
\label{AR_mnist_results}
\end{figure}

In Figure \ref{AR_mnist_results} we show that the training and test performance of the network trained with activation-relaxation is nearly identical to that of the network trained with backpropagation, thus demonstrating that our algorithm can correctly perform credit assignment in deep neural networks with only local learning rules. We also empirically investigate the angle between the AR-computed gradient updates and the true backpropagated updates. The gradient angle $\mathcal{A}$ was computed using the cosine similarity metric $\mathcal{A}(\nabla_\theta) =  \cos^{-1} \frac{\nabla_\theta^T \nabla_\theta^*}{||\nabla_\theta|| ||\nabla_\theta^*}$, where $\nabla_\theta$ was the AR-computed gradients and $\nabla_\theta^*$ were the backprop gradients. To handle the fact that we had gradient matrices while the cosine similarity metric only applies to vectors, following \citep{lillicrap2016random}, we simply flattened the gradient matrices into vectors before performing the computation. We see that the updates computed by AR are very close in angle to the backprop updates (under 10 degrees), although the angle increases slightly over the course of training. The convergence in training and test accuracies between the AR and backprop shows that this slight difference in gradient angle is not enough to impede effective credit assignment and learning in AR. In Appendix C, we take a step towards demonstrating the scalability of this algorithm, by showing preliminary results that indicate that AR, including with the biologically plausible simplifications introduced below, can scale to deeper CNN architectures and more challenging classification tasks.

\subsection{Loosening Constraints}

While the AR algorithm as above precisely approximates adjoint term $\frac{\partial \mathcal{L}}{\partial x^l}$ central to backprop, using only local learning rules, it still retains a number of biological implausibilities. The core implausibility is the weight transport problem, which is still present due to the weight transpose present in Equation \ref{AR_main_equation}.  Following our previous work on relaxed predictive coding (Chapter 3), we demonstrate how the same remedies can be directly applied to the AR algorithm without jeopardising learning performance.

To address the weight transport problem, we take inspirations from the approaches of feedback alignment \citep{lillicrap2016random}, and \citep{millidge2020relaxing}. We postulate an independent separate set of backwards weights $\psi^l$, so that the update rule for the activations becomes,
\begin{align*}
    \frac{dx^l}{dt} = x^l - x^{l+1}f'(W^l x^l) \psi^l \numberthis
\end{align*} 
Then, following our work on relaxed preditive coding in Chapter 5, we learn these backwards weights with the following Hebbian update,
\begin{align*}
\label{AR_psi_equation}
    \frac{d\psi^l}{dt} = {x^{l+1}}^T f'(W^l x^l) x^l \numberthis
\end{align*}

\begin{figure}
\centering
\subfloat[MNIST backwards weights]{\includegraphics[width=0.3\linewidth]{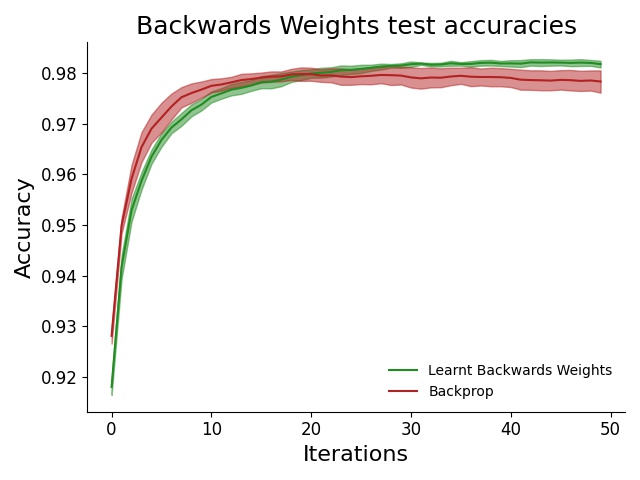}}\hfil
\subfloat[MNIST nonlinear derivatives]{\includegraphics[width=0.3\linewidth]{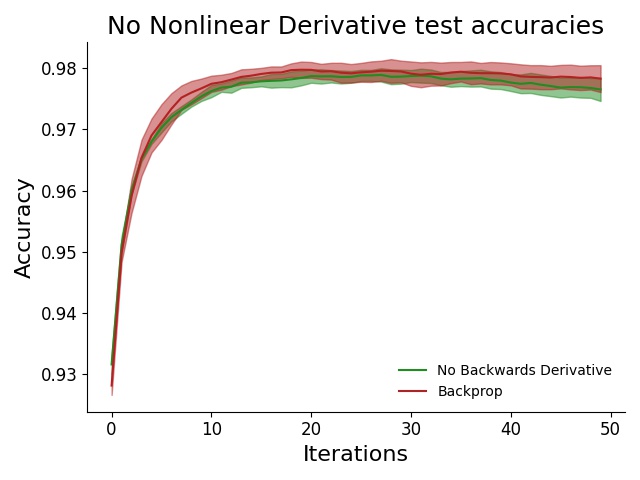}}\hfil 
\subfloat[MNIST combined]{\includegraphics[width=0.3\linewidth]{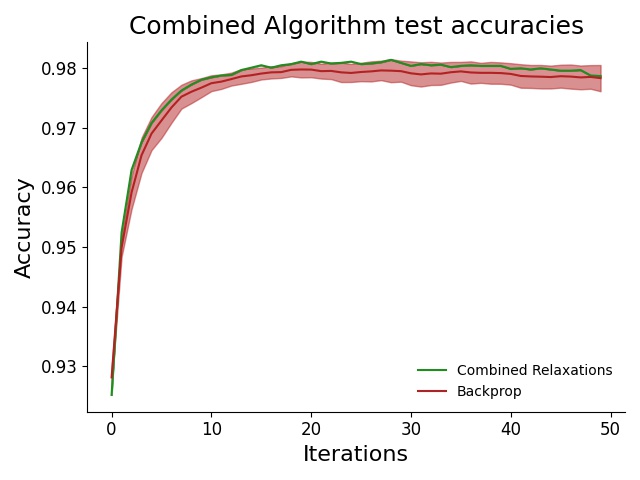}} 

\subfloat[Fashion backwards weights]{\includegraphics[width=0.3\linewidth]{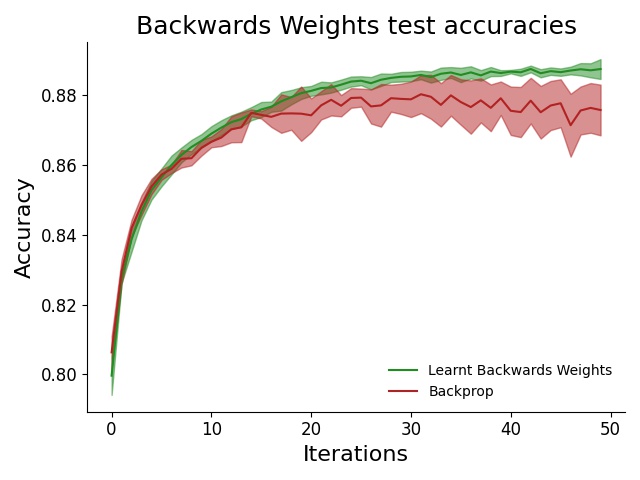}}\hfil
\subfloat[Fashion nonlinear derivatives]{\includegraphics[width=0.3\linewidth]{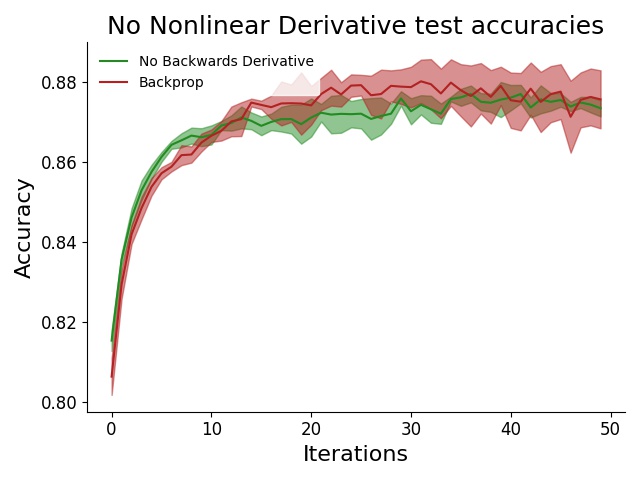}}\hfil 
\subfloat[Fashion combined]{\includegraphics[width=0.3\linewidth]{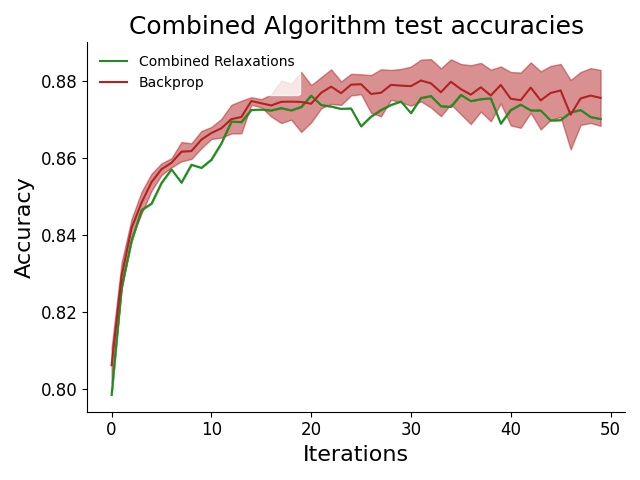}} 
\caption{Train and test accuracy and gradient angle (cosine similarity) for AR vs backprop for MNIST and Fashion-MNIST datasets. Importantly, we see that even with the additional relaxations, the AR trained algorithm performs comparably to backprop}
\label{AR_results_figures}
\end{figure}

The backwards weights were initialized as draws from a 0 mean, 0.05 variance Gaussian. In Figure \ref{AR_results_figures} we show that strong performance is obtained with the learnt backwards weights. We found that using random feedback weights without learning (i.e. feedback alignment), typically converged to a lower accuracy and had a tendency to diverge, which may be due to a simple Gaussian weight initialization used here. Nevertheless, when the backwards weights are learnt, we find that the algorithm is stable and can obtain performance comparable with using the exact weight transposes (Figure \ref{AR_results_figures}).  This is a very strong and encouraging result. First that this learning rule enables performance with exact weight transposes is impressive, since it implies that the Hebbian update rule on the backwards weights works and is highly effective, even early on in training. Secondly, the generalizability of this remedy for weight transport, from predictive coding networks, deep neural networks \citep{amit2019deep,akrout2019deep} with backprop, and now AR suggests that the backwards weights may be able to be independent and robustly learned from scratch in the brain, thus largely resolving the weight transport problem altogether.

\begin{figure}
\centering
\subfloat[Backwards weight angle]{\includegraphics[width=0.3\linewidth]{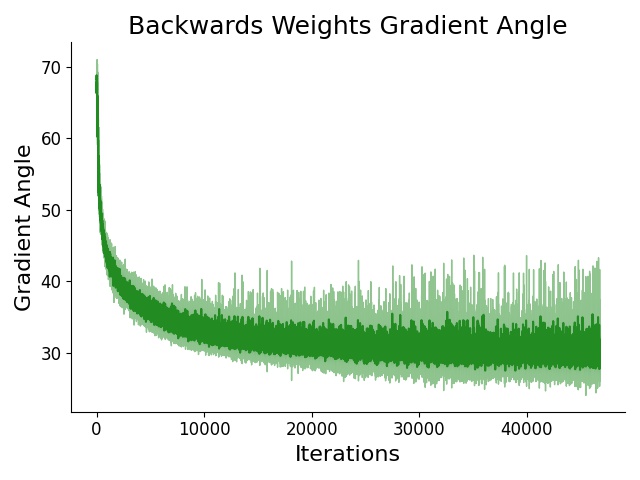}}\hfil
\subfloat[Nonlinear derivatives angle]{\includegraphics[width=0.3\linewidth]{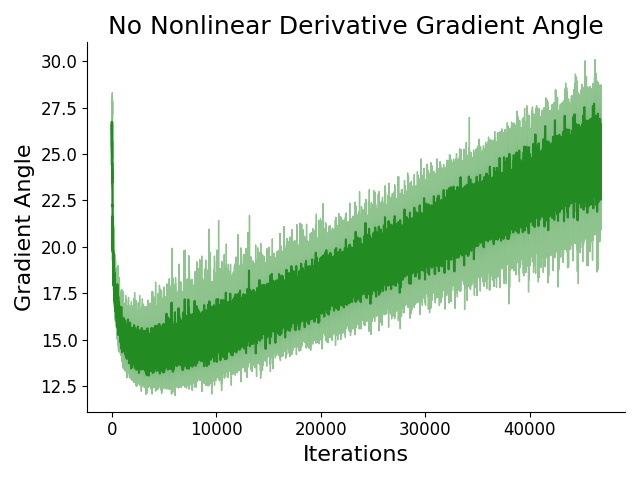}}\hfil 
\subfloat[Combined angles]{\includegraphics[width=0.3\linewidth]{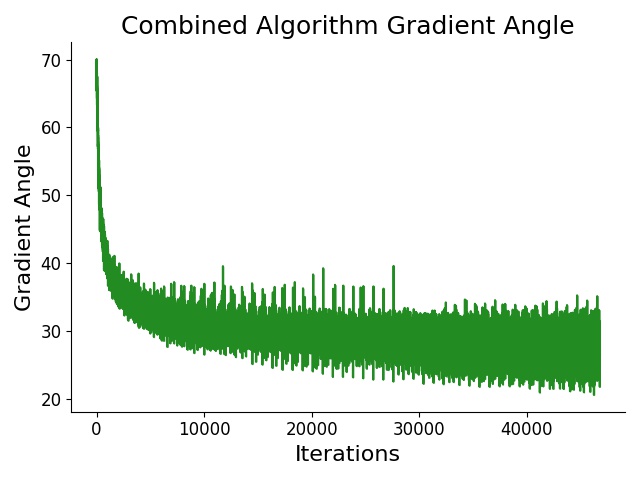}} 
\caption{Angle between the AR and backprop updates in the learnable backwards weights, no nonlinear derivatives, and the combined conditions. At all times this angle remains under 90 degrees and is steady for all cases apart from the no nonlinear derivatives cases, where it appears to increase over time. Interestingly, this does not appear to hinder learning performance noticeably, and may simply reflect the angle getting increasingly worse as the network converges.}
\label{AR_grad_angle_figure}
\end{figure}

We additionally  plot the angle between the AR with learnable backwards weights and the true BP gradients (Figure \ref{AR_grad_angle_figure}). The angle starts out very large (about 70 degrees) since the backwards weights are randomly initialized but then rapidly decreases to about 30 degrees as the backwards weights are learnt, which seems empirically to be sufficient to enable strong learning performance. 

An additional potential biological implausibility to address is the nonlinear derivative problem, which consists of the $f'(W^l, \bar{x}^l) {W^l}^T$ term. The biological plausibility of this term depends heavily upon the activation function used in the network. For instance, in a relu network, this term is trivial, being 0 if the postsynaptic output is greater than 0, and 1 if it is. However, other commonly used activation functions like tanh, sigmoid, and especially softmax are more complex and may be challenging to compute locally in the brain. Here, we experiment with simply dropping the nonlinear derivative term from the update rule, which results in the following dynamics,

\begin{align*}
\frac{dx^l}{dt} = x^l - x^{l+1}{W^l}^T \numberthis
\end{align*}
Although the gradients do no longer match backprop, we show in Figure \ref{AR_results_figures} that learning performance against the standard model is relatively unaffected, showing that the influence of the nonlinear derivative is small. We hypothesise that by removing the nonlinear derivative, we are effectively projecting the backprop update onto the closest linear subspace, which is still sufficiently close in angle to the true gradient that it can support learning. Alternatively, it could be that in the regime of standard activity values prevailing throughout the network during training, the nonlinear derivatives generally are close to 1, and thus have little effect on the update rules in any case. If this is the case, then given that we made no particular effort to constrain the activities of the network, it supposes that this property, if it exists, may be highly beneficial for simplifying the computations in the brain.

By explicitly plotting the angle (Figure \ref{AR_grad_angle_figure}), we see that it always remains under about 30 degrees, sufficient for learning, although the angle appears to rise over the course of training, potentially due to the gradients becoming smaller and more noisy as the network gets closer to convergence.

Moreover, we can \emph{combine} these two changes of the algorithm such that there is both no nonlinear derivative and also learnable backwards weights. Perhaps surprisingly, when we do this  we retain equivalent performance to the full AR algorithm (see Figure \ref{AR_results_figures}), and therefore a valid approximation to backprop in an extremely simple and biologically plausible form. The activation update equation for the fully simplified algorithm is: 
\vspace{-0.2cm}
\begin{align*}
\label{fully_relaxed_AR_update}
    \frac{dx^l}{dt} = x^l - x^{l+1}\psi \numberthis
\end{align*}
which requires only locally available information and is mathematically very simple. In effect, each layer is only updated using its own activations and the activations of the layer above mapped backwards through the feedback connections, which are themselves learned through a local and Hebbian learning rule. This rule maintains high training performance and a close angle between its updates and the true backprop updates (Figure \ref{AR_grad_angle_figure}), and is, at least in theory, relatively straightforward to implement in neural or neuromorphic circuitry.

Finally, we note that the AR update rules require the nonlinear derivative $f'(W^l x^l)$ to be evaluated with the activity $x^l$ evaluated at its feedforward pass value $x^l = \bar{x}^l$. Additionally, in the weight update, the value of the activity $x^l$ needs to be evaluated at its feedforward pass value
\begin{align*}
    \frac{\partial \mathcal{L}}{\partial W^l} = \frac{\partial \mathcal{L}}{\partial x^{l+1}}\frac{\partial x^{l+1}}{\partial W^l} \Bigr|_{x^{l+1}=\bar{x}^{l+1}}  = {x^{l+1}}^* \bar{x}^T {\frac{\partial f(W^l \bar{x}^l)}{\partial \bar{x}}}^T \numberthis
\end{align*}
We call this the \emph{frozen feedforward pass} assumption, and it is very closely related to the fixed-prediction assumption in predictive coding or, similarly the requirement in equilibrium-propagation to store all the intermediate equilibrium values of the free phase. Here we investigate to what extent this assumption can also be relaxed. 

We evaluate whether the nonlinear derivative term can be unfrozen so that it uses the current value of the activity in a.) the function derivative $f'$ in Equation \ref{AR_main_equation}, b.) in the weight update equation (Equation \ref{BP_weights_equation}), and c.) we investigate whether the activation value itself can be replaced in the weight update equation.
\begin{figure}[htb]
\centering
  \begin{subfigure}[b]{0.4\linewidth}
    \centering
    \includegraphics[width=0.75\linewidth]{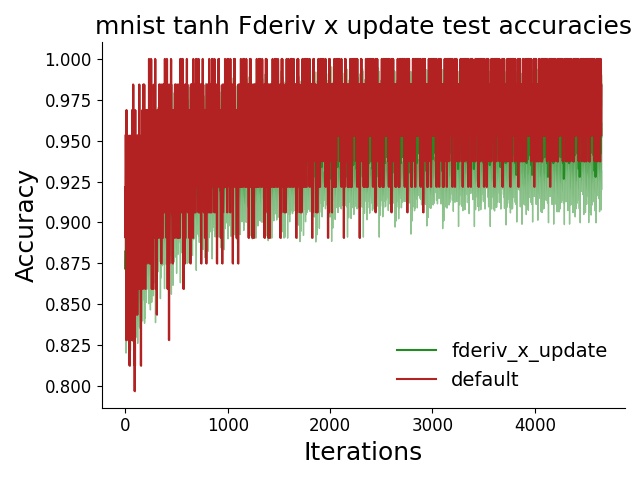} 
    \caption{MNIST nonlinear derivative relaxation update} 
    \vspace{4ex}
  \end{subfigure}
  \begin{subfigure}[b]{0.4\linewidth}
    \centering
    \includegraphics[width=0.75\linewidth]{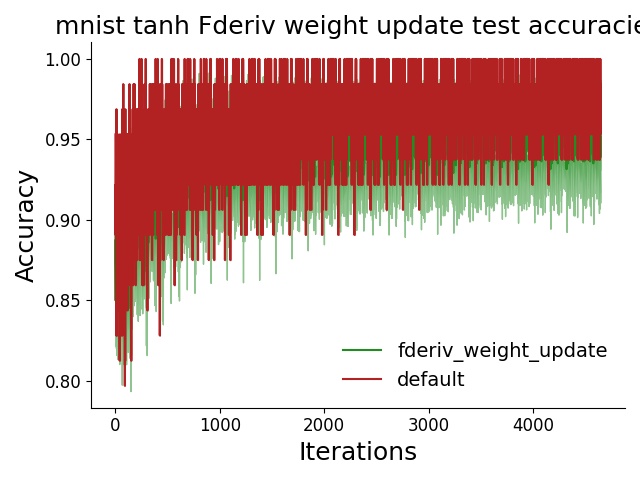} 
    \caption{MNIST nonlinear derivative weight update} 
    \vspace{4ex}
  \end{subfigure} 
  \begin{subfigure}[b]{0.4\linewidth}
    \centering
    \includegraphics[width=0.75\linewidth]{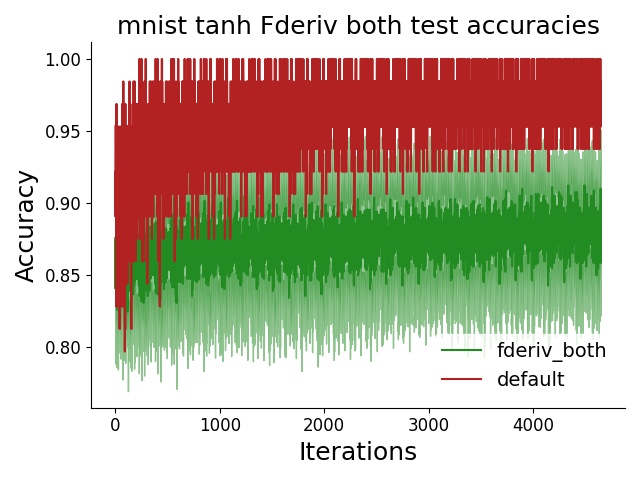} 
    \caption{MNIST both nonlinear derivative} 
  \end{subfigure}
  \begin{subfigure}[b]{0.4\linewidth}
    \centering
    \includegraphics[width=0.75\linewidth]{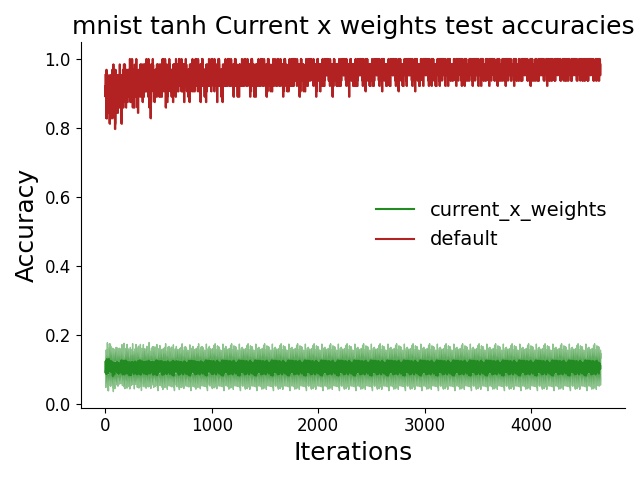} 
    \caption{MNIST current x weight update} 
  \end{subfigure} 
  \caption{Assessing whether the frozen feedforward pass assumption can be relaxed. We show the resulting performance (test accuracy) against baseline of relaxing this assumption on the MNIST dataset. All results averaged over 10 seeds. These results show clearly that the frozen feedforward pass assumption can be relaxed for the nonlinear derivative and in the weight update nonlinear derivative, but not both nonlinear derivatives simultaneously, and definitely not using the $x^T$ term in the weight update}
  \label{AR_no_frozen_pass}
\end{figure} 

In Figure \ref{AR_no_frozen_pass}, we see that the frozen feedforward pass assumption can be relaxed in the case of the nonlinear derivatives for both the AR update and the weight update equation. However, relaxing it in the case of the weight update equation destroys performance. This means that ultimately, a direct implementation of AR in biological circuitry would require neurons to store the feedforward pass value of their own activations. 

All the experiments so far have been done on the relatively simple and straightforward MNIST dataset with small MLP models. However, it is also important to verify the scalability of this method. Here we demonstrate that AR can be used to train large CNNs on challenging image recognition datasets (SVHN, CIFAR10, and CIFAR100) and, moreover, that the previous loosenings of the biologically implausible constraints on the algorithm still do not appear to degrade performance unduly. This extension to CNNs is especially important because other biologically plausible schemes such as feedback alignment \citep{lillicrap2016random,lillicrap2014random}, and directed feedback alignment \citep{nokland2016direct}, have been shown to struggle with the CNN116architectures \citep{launay2019principled}. We tested the simplifications (dropping nonlinearity or learning backwards weights) on just the convolutional layers of the network, just the fully-connected layers of the network, or both together. We found that ultimately performance was largely maintained even when both convolutional and fully connected layers in the network used learnable backwards weights or had their nonlinear derivatives dropped from both the update and weight equations. These results speak to the scalability and generalisability of these relaxations, and the general robustness of the AR algorithm. We implemented the learnable backwards weights of the CNN by applying Equation \ref{BP_weights_equation} to the flattened form of the CNN filter kernel weights.

Our CNN consisted of a convolutional layer followed by a max-pooling layer, followed by an additional convolutional layers, then two fully connected layers. The convolutional layers had 32 and 64 filters respectively, while the FC layers had 64,120, and 10 neurons respectively. For CIFAR100 there are 100 output classes so the final layer had 100 neurons. The labels were one-hot-encoded and fed to the network. All input images were normalized so that their pixel values lay in the range $[0,1]$ but no other preprocessing was undertaken. We used hyperbolic tangent activations functions at every layer except the final layer which was linear. The network was trained on a mean-square-error loss function. 

 \begin{figure}[htb]
    \centering 
\begin{subfigure}{0.25\textwidth}
  \includegraphics[width=\linewidth]{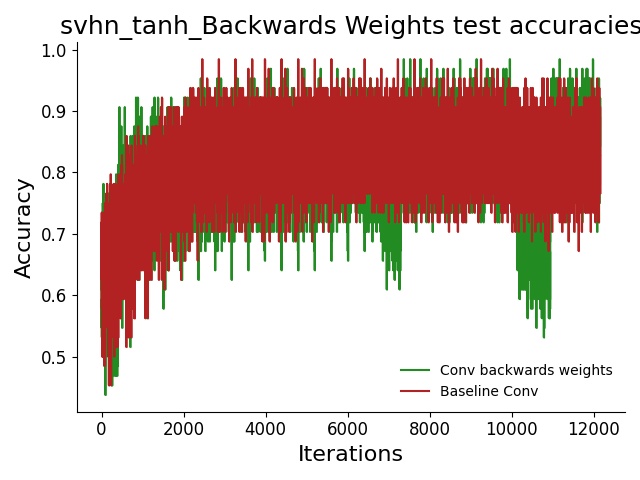}
  \caption{Conv backwards weights}
\end{subfigure}\hfil 
\begin{subfigure}{0.25\textwidth}
  \includegraphics[width=\linewidth]{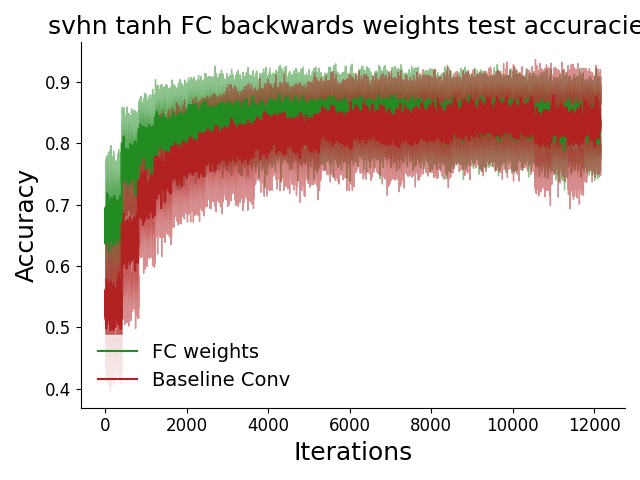}
  \caption{FC backwards weights}
\end{subfigure}\hfil 
\begin{subfigure}{0.25\textwidth}
  \includegraphics[width=\linewidth]{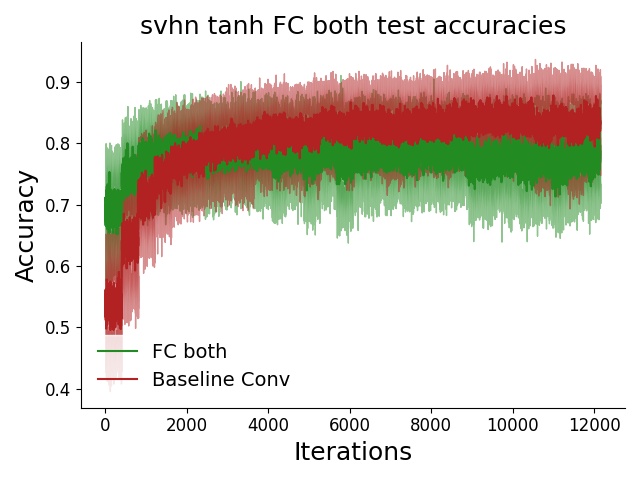}
  \caption{Both backwards weights}
\end{subfigure}

\medskip
\begin{subfigure}{0.25\textwidth}
  \includegraphics[width=\linewidth]{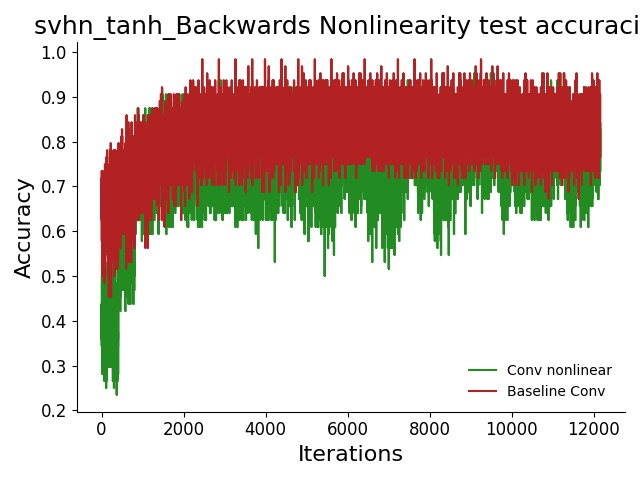}
  \caption{Conv no nonlinear derivative}
\end{subfigure}\hfil 
\begin{subfigure}{0.25\textwidth}
  \includegraphics[width=\linewidth]{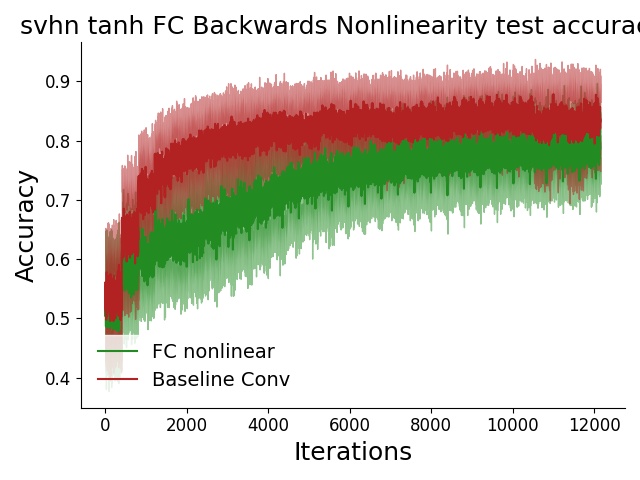}
  \caption{FC no nonlinear derivative}
\end{subfigure}\hfil 
\begin{subfigure}{0.25\textwidth}
  \includegraphics[width=\linewidth]{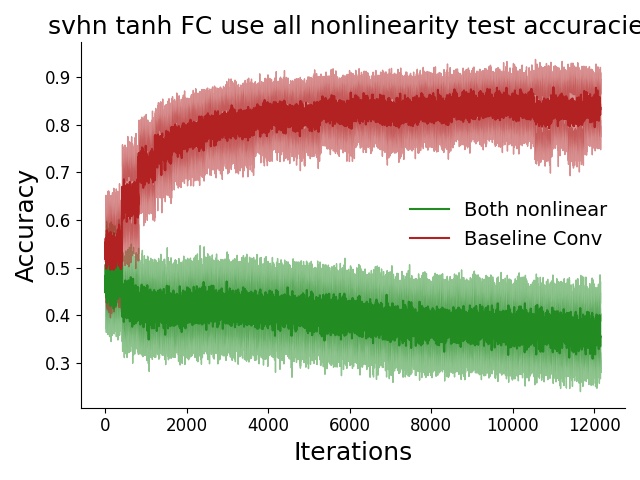}
  \caption{Both no nonlinear derivative}
\end{subfigure}
\end{figure}

We see that the simplifications also scale to the CNN for the SVHN dataset, although, interestingly, performance is degraded on this dataset when both convolutional and FC nonlinearities are dropped. However, since this does not occur in the other, more challenging, CIFAR datasets, we take this result to be an anomaly.

 \begin{figure}[htb]
    \centering 
\begin{subfigure}{0.3\textwidth}
  \includegraphics[width=\linewidth]{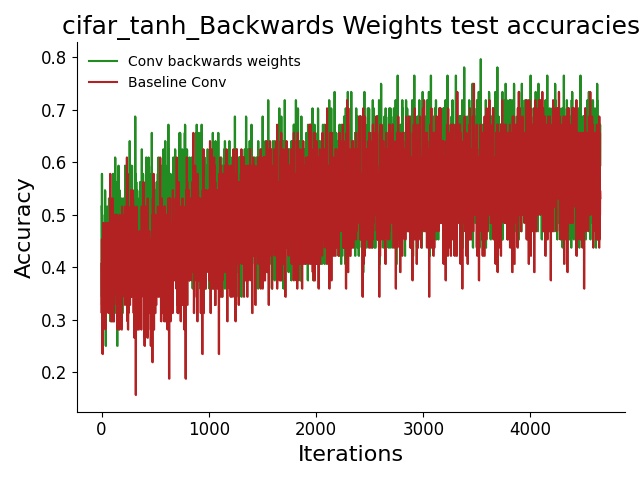}
  \caption{Conv backwards weights}
\end{subfigure}\hfil 
\begin{subfigure}{0.3\textwidth}
  \includegraphics[width=\linewidth]{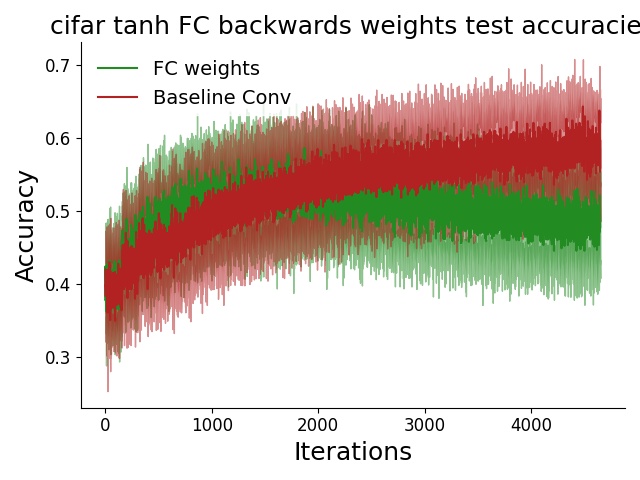}
  \caption{FC backwards weights}
\end{subfigure}\hfil 
\begin{subfigure}{0.3\textwidth}
  \includegraphics[width=\linewidth]{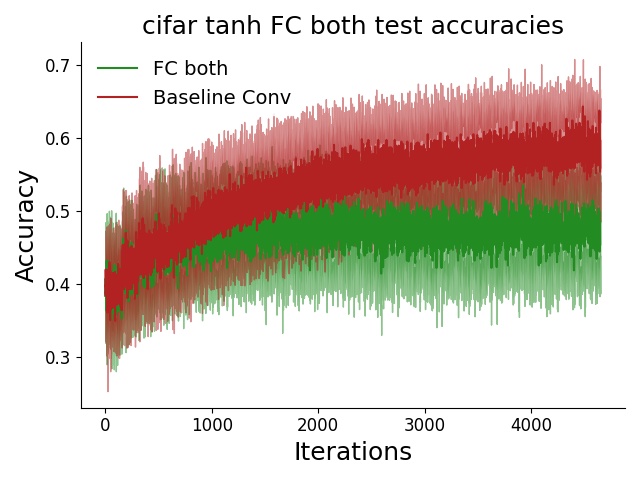}
  \caption{Both backwards weights}
\end{subfigure}

\medskip
\begin{subfigure}{0.3\textwidth}
  \includegraphics[width=\linewidth]{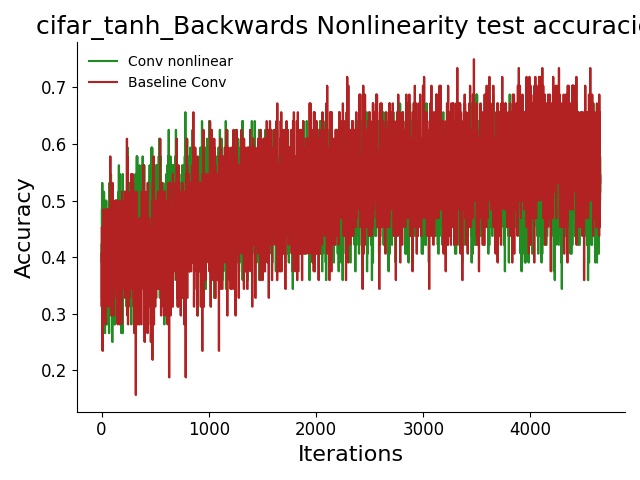}
  \caption{Conv no nonlinear derivative}
\end{subfigure}\hfil 
\begin{subfigure}{0.3\textwidth}
  \includegraphics[width=\linewidth]{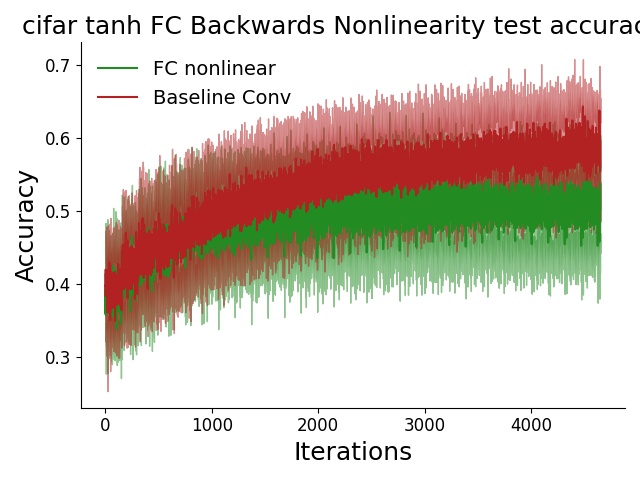}
  \caption{FC no nonlinear derivative}
\end{subfigure}\hfil 
\begin{subfigure}{0.3\textwidth}
  \includegraphics[width=\linewidth]{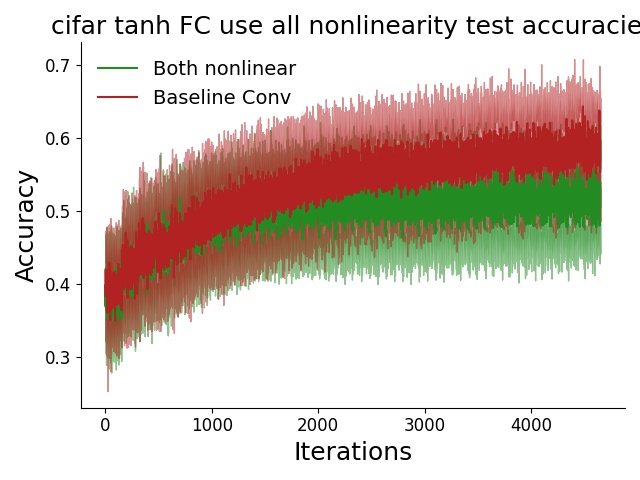}
  \caption{Both no nonlinear derivative}
\end{subfigure}
\caption{Performance (test accuracy), averaged over 10 seeds, on CIFAR10 demonstrating the scalability of the learnable backwards weights and dropping the nonlinear derivatives in a CNN architecture, compared to baseline AR without simplifications. Performance is equivalent throughout.}
\end{figure}

 \begin{figure}[htb]
    \centering 
\begin{subfigure}{0.25\textwidth}
  \includegraphics[width=\linewidth]{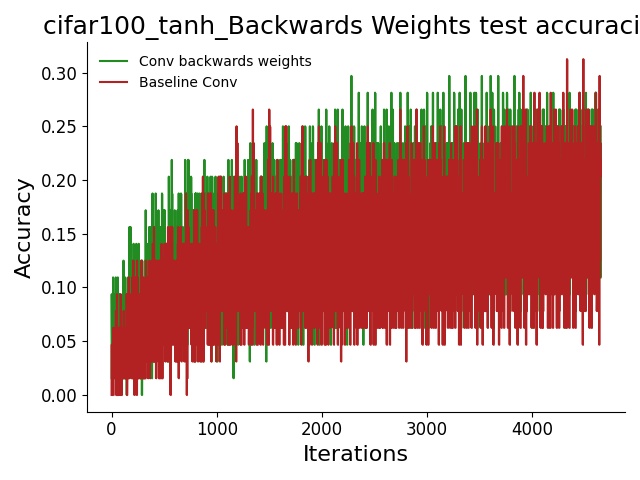}
  \caption{Conv backwards weights}
\end{subfigure}\hfil 
\begin{subfigure}{0.25\textwidth}
  \includegraphics[width=\linewidth]{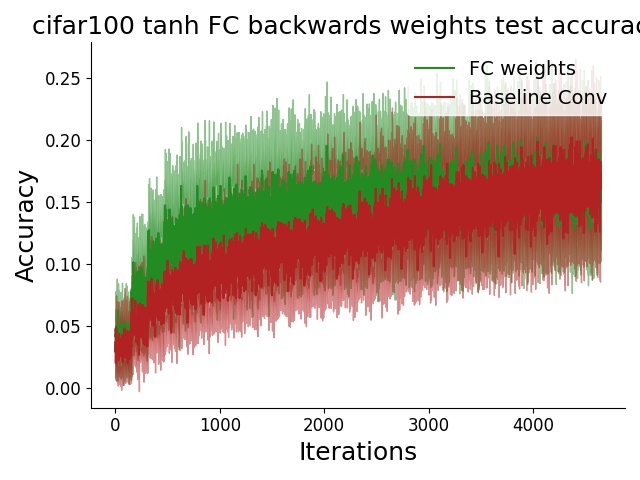}
  \caption{FC backwards weights}
\end{subfigure}\hfil 
\begin{subfigure}{0.25\textwidth}
  \includegraphics[width=\linewidth]{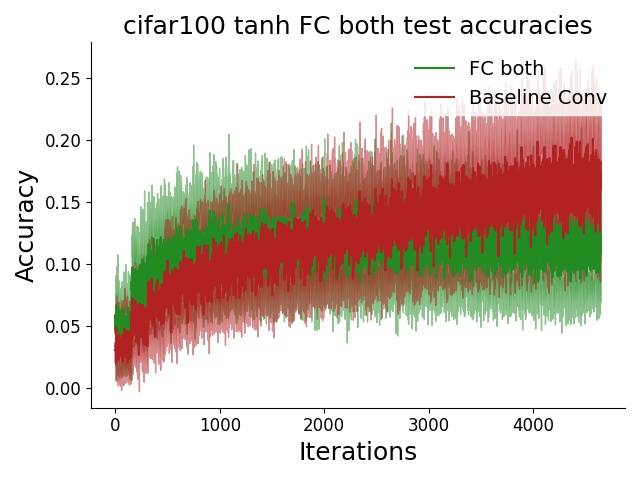}
  \caption{Both backwards weights}
\end{subfigure}

\medskip
\begin{subfigure}{0.25\textwidth}
  \includegraphics[width=\linewidth]{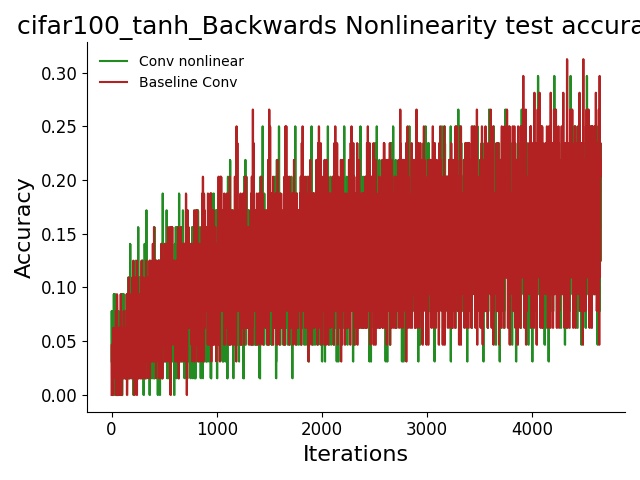}
  \caption{Conv no nonlinear derivative}
\end{subfigure}\hfil 
\begin{subfigure}{0.25\textwidth}
  \includegraphics[width=\linewidth]{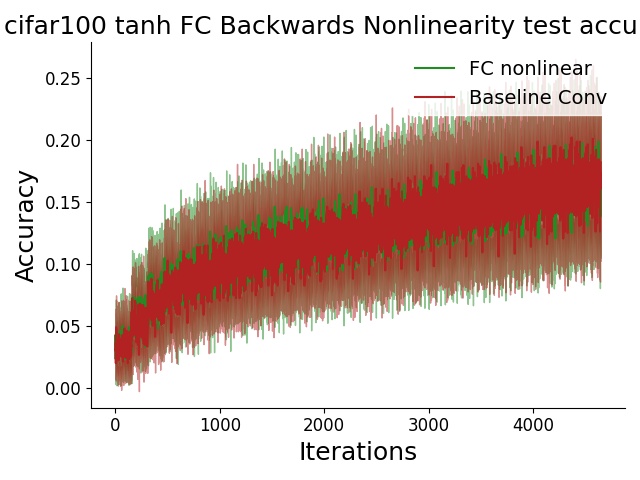}
  \caption{FC no nonlinear derivative}
\end{subfigure}\hfil 
\begin{subfigure}{0.25\textwidth}
  \includegraphics[width=\linewidth]{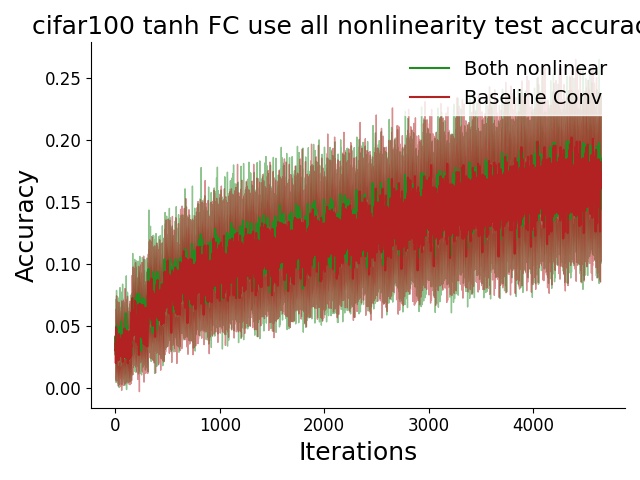}
  \caption{Both no nonlinear derivative}
\end{subfigure}
\end{figure}

\subsection{Interim Discussion}

In sum the AR algorithm uses only simple learning rules to asymptotically approximate the adjoint terms of backprop over the course of multiple dynamical iterations. We have demonstrated that the AR algorithm can apply to arbitrary computation graphs, as can predictive coding, and can be used to train deep CNN models on challenging object recognition tasks with a performance equivalent to backprop. Moreover, AR eschews much of the complexity of competing schemes such as predictive coding, by not requiring two separate populations of value and error neurons, and equilibrium propagation by not needing two separate backwards phases -- a free phase and a clamped phase. Additionally, we have demonstrated that some of the remaining biological implausibilities of AR, such as the weight transport and backwards nonlinearities problems, can be successfully ameliorated through the right extensions to the algorithm such as learnable backwards weights with minimal effect on overall performance. Other constraints, such as the necessity to use the feedforward pass activities in the dynamics instead of the current activities cannot be relaxed without catastrophically damaging overall performance.

Like predictive coding, a limitation of this method is its intrinsically iterative nature. This iteration scheme means that it is at least several times more costly than standard backpropagation of error -- for instance, in these experiments we used 100 iterations to reach exact convergence to the backpropagated gradients, although this is not strictly necessary for good performance. While some of this computational cost may be ameliorated by the intrinsic parallelism of neural circuitry, there is nevertheless a timing issue if convergence is required to be complete before the next sensory datum arrives, and issues of gradient interference if it is not. A speculative solution to this could be synchronization mediated by the brain's alpha/beta or gamma band frequencies in the cortex, where one dynamical phase iterating to convergence would correspond to one wavelength of the band. Such an identification is highly speculative however, and the computational function of such oscillations in the brain are still largely mysterious and highly controversial \citep{buzsaki2006rhythms}.

One additional and important drawback of the AR algorithm is that it requires keeping a memory of the feedforward pass activations throughout the backwards dynamical phase. This is because the activations swap their purpose from representing feedforward pass values in the forward phase, to representing gradients in the backwards phase. Taken literally, this requires that the learning rules become non-local in time, although in practice the feedforward pass activations are just stored. While this is the most substantial drawback to the biological plausibility of the algorithm, it is important to note that this drawback is also shared with every other iterative algorithm in the literature. Predictive coding similarly requires the fixed-prediction assumption, which is effectively the same, and equilibrium-propagation requires that all the activations at the equilibrium of the free phase are stored while the clamped phase converges. We suggest that this necessity for storage in iterative algorithms to approximate backprop is universal and it arises for a simple reason. Namely, that the backpropagated gradients themselves fundamentally only depend upon the feedforward pass values, since we are backpropagating only on the feedforward pass and not through the dynamics themselves. Since the gradients ultimately depend only on the feedforward pass, any dynamical scheme to approximate them must `remember' what these values are, somehow. In theory, it may be possible to design algorithms such that this memory is implicit and thus not explicitly necessary, but no such algorithms have, to my knowledge, been found so far.

This issue of memory is also closely related to the core computational properties of reverse-mode AD. Specifically, that it requires storage in memory of all intermediate activations in the forward pass. This means that the memory issue also closely applies to sequential backward algorithms such as target-propagation and, indeed, backprop itself. The fundamental fact is that backwards pass can only take place \emph{after} the forward pass is complete, and that it requires knowledge of forward pass activities. This fact cannot be ignored or cleverly wished away by any algorithm but must simply be addressed. If the brain is performing reverse-mode AD, then it simply must have some way to store, implicitly or explicitly, the feedforward pass values. The question then becomes how can this be done in the brain? Some possibilities are multiplexing using the brain's own intrinsic rhythms \citep{buzsaki2006rhythms}, or some kind of special parallel class or neurons to maintain activity explicitly \citep{o1999biologically}, or else storage at the synaptic level through mechanisms like eligibility traces \citep{bellec2020solution}. While we here remain ambiguous on the means by which such storage is achieved, we have reached a point of conceptual clarity in knowing that there \emph{must be storage}. The only question is how.

\section{Three-Factor Learning Rules and a Direct Implementation}

After having gone through several iterative algorithms for approximating the backpropagation of error algorithm, it is worth taking a step back to understand what has been shown and what is truly necessary. Here, we argue that in fact, despite strong claims that backprop is biologically implausible, in fact an actual implementation of backprop in the brain could be surprisingly simple and biologically plausible. Moreover, that many of the algorithms, including the iterative algorithms previously, may simply be complicating matters. The material in this section is speculative and early stage, and will be investigated further in future work.

First, we begin by stating, somewhat boldly, that in general the weight transport problem is solved. That is, there is now fairly strong evidence in the literature \citep{lillicrap2016random,amit2019deep,akrout2019deep,millidge2020relaxing} firstly that a precise equality of forward and backwards weights is not necessary for good learning performance, and secondly, that competitive performance with backprop can be maintained, even for deep networks, through learnable backwards weights which update with a fairly straightforward Hebbian learning rule. If we assume that the weight transport problem is solved, then there is only the locality issues remaining with a direct implementation of backprop.

Firstly, we note that if we are implementing reverse-mode AD, and we find a way to compute the adjoint $\frac{\partial \mathcal{L}}{\partial x^l}$ locally at a layer then we actual weight update $\frac{\partial \mathcal{L}}{\partial W^l} = \frac{\partial \mathcal{L}}{\partial x^l}\frac{\partial x^l}{\partial W^l}$ requires only local information. This means that almost the entire challenge of backprop is computing the adjoint term locally.

Secondly, we notice that the standard forward pass of an artificial neural network $x^l = f(W^l x^{l-1})$ does not actually correspond to how it would work in the brain since here the activation function (which is typically assumed to be applied through the threshold for making an action potential in the cell soma) is applied \emph{after} the weights, while in the brain the synaptic weights are on the dendrites of the post-synaptic neuron, and thus occur after the action potential. This means that instead we should use the more biologically plausible forward pass as,
\begin{align*}
\label{forward_equation}
    x^l = W^l f(x^{l-1}) \numberthis
\end{align*}
Where the order of the weights and the activation function are switched. Specifically, this means that the activation function only applies to the presynaptic activity and not the weights. This approach considerably simplifies the requisite gradients, so we get the following expressions for the adjoint and the weight update
\begin{align*}
    \frac{\partial \mathcal{L}}{\partial x^l} &=\frac{\partial \mathcal{L}}{\partial x^{l+1}} {W^{l+1}}^T f'(x^l) \\
    \frac{\partial x^l}{\partial W^l} &=  \frac{\partial \mathcal{L}}{\partial x^l} f(x^l) \numberthis
\end{align*}

Specifically, the adjoint recursion is simply the adjoint of the layer above, multiplied through the backwards weights and the derivative of the activation function of the pre-synaptic activity. The weight update is substantially simpler since, because the forward pass is linear in the weights, it is simply a multiplication of the adjoint with the presynaptic activity which is essentially Hebbian except with the adjoint replacing the post-synaptic term. Interestingly, this applies that in this case of backprop, the post-synaptic term should have \emph{no} effect on the synaptic weights, which is a very strong and counterintuitive empirical prediction of backprop. 

While it may seem that the switch in the order of the weights and the activation function might have some serious impact on the expressive power of the network, we argue that it likely does not. In fact, the two formulations are equivalent in deep neural networks except for the beginning and ending layers. To see this, we can simply explicitly write out the expression for the function computed by a 4 layer neural network,
\begin{align*}
    y = f(W^4 f(W^3 f(W^2 f(W^1 x)))) \approx W^4 f(W^3 f(W^2 f(W^1 f(x)))) \numberthis
\end{align*}
For the vast majority of layers, these expressions are the same up to a re-bracketing. The only difference is the last layer has no activation function using the more neural forward pass (but often in neural networks we use a linear last layer anyway), and that the input is first passed to an activation function activation function where this is not common in standard ANNs -- but this function could always be set to the identity to make the equivalence exact. In general, the effect of these differences will be minor for deep networks.

\begin{figure}
    \centering
    \includegraphics[width=\linewidth]{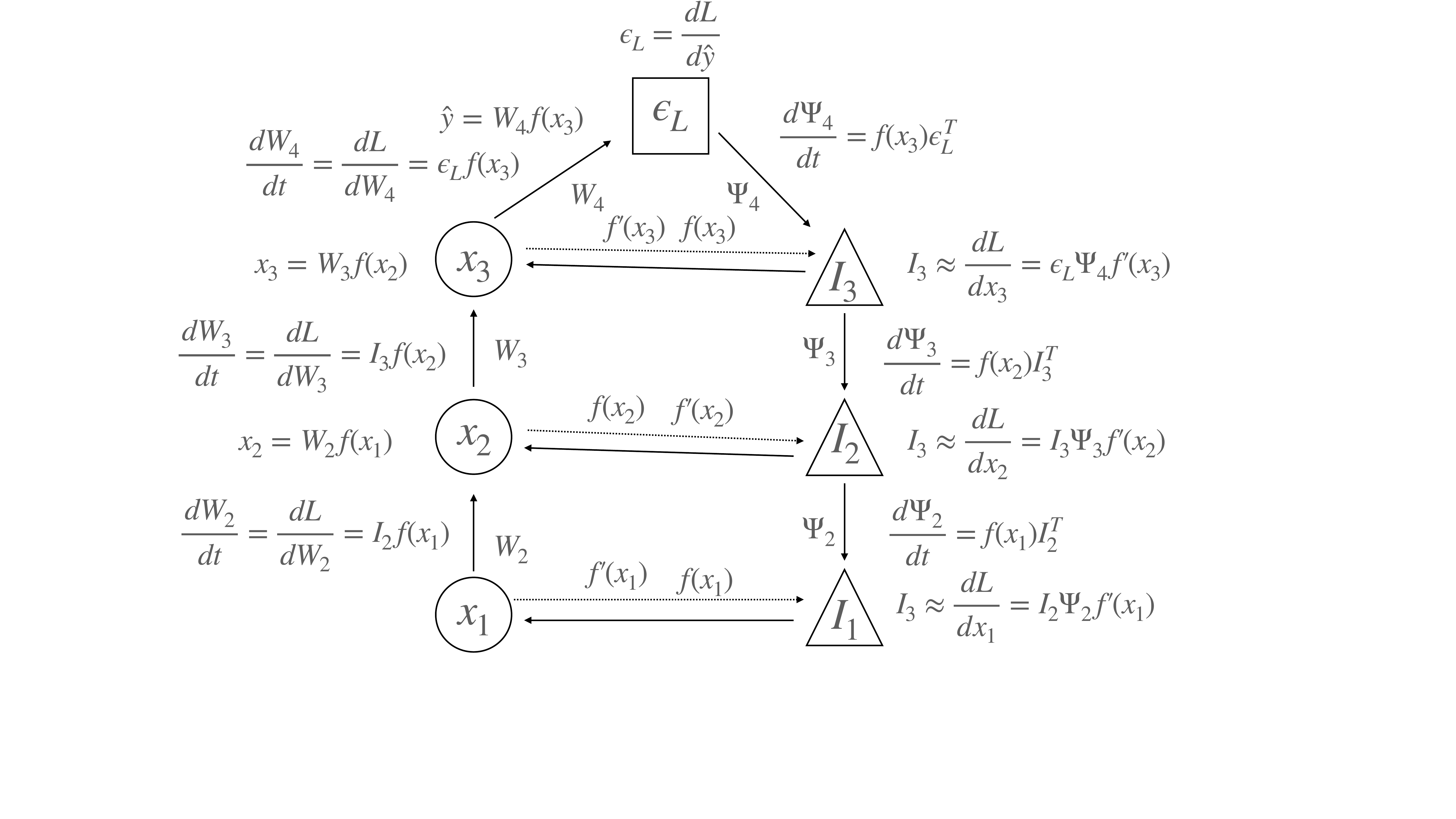}
    \caption{Potential schematic for a direct implementation of backprop in the brain. All that is necessary for this to be plausible is three-factor learning rules.}
\label{continuous_bp_figure}
\end{figure}

Now we note the crucial point, that the only biological implausibilities in these learning rules is the necessity to have the adjoint value present at the synapse for the weight update rule, since it is just the multiplication of the adjoint and the presynaptic activation. The recursive computation of the adjoint itself (Equation \ref{reverse_mode_equation} is relatively plausible, since the only difficulty is the nonlinear derivative term $f'(x)$ which now is only the derivative of the activation function applied to the pre-synaptic input. Importantly, for spiking neurons, this derivative is trivial as is essentially consists of a spike when the neuron fires and not when it doesn't. As such, this rule effectively says that the adjoint should only be updated when there is presynaptic firing. Putting this all together, we can imagine implementing this in the brain in a fairly direct forward-backward scheme as in Figure \ref{continuous_bp_figure}

Specifically, if we assume that the weight transport problem can be solved with independently learnable backwards weights, then the recursion for the adjoint becomes simply,
\begin{align*}
    \frac{\partial \mathcal{L}}{\partial x^l} &=\frac{\partial \mathcal{L}}{\partial x^{l+1}} \psi^T f'(x^l) \\
    &\approx \frac{\partial \mathcal{L}}{\partial x^{l+1}} \psi^T \numberthis
\end{align*}
where the second line is the recursion if we simply ignore the nonlinear derivative term, as we have also found does not hinder learning in practice \citep{millidge2020relaxing,millidge2020investigating,millidge2020activation,ororbia2019biologically}. This recursion is extremely simple and is just the adjoint mapped through the backwards weights to the layer below. Thus, we can imagine keeping separate forward and backward passes whereby the adjoint is represented by interneurons $I^l$. Here, the update rules simply become,
\begin{align*}
\label{continous_bp_backward_equation}
       I^l &=I^{l+1}\psi ^T  \\
    \frac{\partial \mathcal{L}}{\partial W^l} &=  I^l f(x^l) \numberthis
\end{align*}

The simplicity of these update rules implies that the only potential biological implausibility is in the weight update \ref{BP_weights_equation} where we have transmitted the value of the adjoint `interneurons' to the synapses of the post-synaptic neuron, and used them to update those weights. Through a careful analysis, we have revealed this question to be the ultimate crux of whether backprop in the brain is plausible or not. Importantly, this ability to use the adjoints as part of a `local' learning rule is crucial to every purported `biologically-plausible' method in the literature -- from predictive coding, to target-propagation, to equilibrium-prop and AR. 

Whether or not the adjoint can be transmitted to the synaptic weights in the brain is currently a controversial and unresolved question, and to my knowledge, has not been studied directly. While it may seem fairly obscure, this analysis suggests that this question is absolutely crucial to our understanding of whether the brain can directly implement backprop or not. An important possibility is that of segregated dendrites \citep{sacramento2018dendritic}. Having separate dendritic compartments would, presumably, straightforwardly enable the broadcast of the adjoint values to the synaptic weights, since they would be located in the dendritic tree of the same neuron. The key question would then become whether information transmitted to the dendrites could remain segregated. That is, could both Equation \ref{forward_equation} and Equation \ref{continous_bp_backward_equation} be implemented using the same neuron. If this is the case then it would suggest an incredibly simple biological implementation for backpropagation, needing only a single type of neuron which would both send forward connections and reciprocally receive backwards connections from neurons in the layer above. Such a simple architecture would lend great support to the idea that the brain can indeed implement backprop, and perhaps that backpropagation is so straightforward that it may even function as a computational primitive in neural circuitry.

\section{Discussion}

Overall, in this chapter, we have shown that predictive coding, as an approximation to the backpropagation of error algorithm, can be extended to arbitrary computation graphs and we have applied predictive coding to large-scale machine learning architectures such as CNNs and LSTMs and demonstrated that they perform comparably to backprop trained networks. Moreover, we have posited a novel iterative algorithm -- Activation Relaxation -- that also converges to the exact backprop gradients, does not require two separate populations of predictions and prediction error units, and uses extremely simple and elegant learning and update rules. We have shown that AR also scales to large-scale CNN neural network models and is competitive with backprop trained networks at scale. Finally, taking experience from our previous work in this field, we have re-analyzesd the problem of backpropagation in the brain from first principles and discovered, somewhat surprisingly, that if we assume that the weight transport problem is solved, the only major issue of biological implausibility is whether recursively computed adjoints can be `transferred' onto synapses to be able to form part of the synaptic weight updates. If they can, and it seems likely that this is possible through a mechanism of segregated dendrites or, alternatively, backpropagating action potentials \citep{stuart1997action}, then we can be fairly certain that propagation in the brain is at least theoretically achievable. This is a remarkable turn-around from the consensus only five years ago that it was completely biologically implausible, and speaks to the rapid development and advances in this field.

Additionally, while this thesis chapter has presented a substantial extension to an existing algorithm (predictive coding), and an entirely novel algorithm for credit assignment in the brain (activation relaxation), we also wish to highlight the conceptual contributions we have made while thinking about these issues deeply. In my opinion, these are perhaps the most important sections of this work. Namely, firstly, the issue of memory and time in any implementation or approximation to reverse-mode AD. Specifically, that any biologically plausible algorithm, whether sequential or iterative, due to the very nature of reverse-mode AD \emph{must} store the values of the feedforward pass throughout the backwards sweep or phase, either implicitly or explicitly. While the rationale for this seems obvious in retrospect, it was not clear beforehand, and is still not at all clear in the literature. Indeed, the dependence of almost all of these biologically plausible algorithms on the memory of forward pass values is generally obfuscated or presented as a minor hindrance, when in fact it is an absolutely irrevocable fact of the comptuation these algorithms are trying to render biologically plausible. Secondly, and perhaps most importantly, we have reached the crux of the issue of whether backpropagation in the brain is plausible -- namely whether adjoints, which must remain separate from the post-synaptic activation -- can modulate synaptic weight updates. If they can, then very biologically plausible and elegant schemes exist for a direct implementation of backpropagation in the brain (see Equations \ref{continous_bp_backward_equation}, \ref{BP_weights_equation}). If it is not possible, then it seems likely, given that the adjoint equation is absolutely fundamental to reverse-mode AD, that backpropagation in the brain is not plausible or, at least, is explicitly achievable except via some roundabout method. Moreover, if the transport of the adjoint onto the synaptic weight terminals is possible, then it must be supported by some kind of dedicated neurophysiological mechanism, which can and must be studied in detail if we are to understand the explicit details of this key aspect of credit assignment in the brain. Nevertheless, I now believe that the key mathematical and conceptual issues in the question of whether the brain can do backpropagation (in this rate-coded static model) have been largely worked out and depend now solely on details of neurophysiology.

The crucial caveat to this response is that we have only worked out credit assignment in an incredibly simplified model of what occurs in the brain -- namely with rate-coded integrate and fire neurons -- on static inputs. Both of these assumptions, however, are false in the brain. Firstly, neurons are spiking networks which may communicate using precise spike timings to convey information. Understanding how to perform credit assignment in such spiking networks is still a young and open field, although there has been much recent progress \citep{schiess2016somato,zenke2018superspike,kaiser2020synaptic,neftci2019surrogate}. Moreover, and crucially, the key problem the brain faces is not just backpropagation through space (i.e. layers), but backpropagation through time. The brain must be able to assign credit correctly to temporally distant events from the synaptic weight values that, ultimately caused them. While a mathematical formulation of reverse-mode AD can be directly formulated by simply performing backprop on a computation graph `unrolled through time', in practice this means that in the backwards phase that \emph{time must run backwards} or, alternatively that the network must store not only the feedforward pass, but its \emph{entire history}, which is definitely biologically implausible. The key question is thus how to implement backpropagation through time in a biologically plausible manner. There has been much recent progress in this field, also combined with spiking networks such as \citep{zenke2018superspike,bellec2020solution}. However, the innovative approach engendered by Eligibility Propagation (E-prop) only applies to single layer recurrent networks, leaving open the question of how to marry backpropagation through space and backpropagation through time.

An additional interesting consideration is that throughout, and generally in the literature, only reverse-mode AD is considered to be a contender for the credit assignment algorithm implemented in the brain. This is due to the historical use and generally better computational properties of reverse-mode for artificial neural networks \citep{griewank1989automatic,baydin2017automatic} and has thus become the dominant paradigm \citep{goodfellow2016deep,rumelhart1985feature,silver2017mastering}. However, this is not necessarily the case in highly parallel architectures like the brain, for which additional forward computation cost may be effectively negligible due to the degree of parallelization. Forward-mode AD can be implemented through dual numbers -- which directly pair activations with their derivatives, and which it is interesting to think about how this could relate to neural activity and synapses. Moreover, a key computational advantage of forward-mode AD is that it imposes no memory cost, since it is entirely online and requires no storage of intermediate activations, thus entirely obviating the memory issues inherent in implementations of reverse-mode AD. The disadvantage, however, with forward-mode AD in the brain is that it dislocates the derivative computation from the physical location of the synapses. The computations and derivatives `move forwards' up through higher layers and levels of processing while the synapses remain firmly put. it is necessary, then, whenever the computations and derivatives reach the end of the process to transmit the fully computed derivatives back to where they originated. Precisely working out this process has, to my knowledge, not yet been done, but it may result in a practicable algorithm. One important case where forward-mode AD makes sense is in backpropagation through time since, as the computation moves forward in time, so do the synapses themselves. Thus, the correct derivatives are always locally available precisely when they are needed. Forward-mode AD through time is known as real-time-recurrent learning (RTRL) \citep{williams1989experimental}, and is potentially a good algorithm for the brain to solve recurrence, although it is extremely computationally expensive, rendering it uncompetitive with reverse-mode BPTT for training large neural networks. Moreover, by taking various sparse approximations to RTRL, it is possible to reduce the computational cost at the cost of somewhat reduced learning performance. Algorithms such as eligibility-prop essentially try to make RTRL updates biologically plausible, with some success.

\section{Conclusion}

In this chapter, we have proposed two novel biologically plausible algorithms for credit assignment in the brain. Firstly, we demonstrate that predictive coding, under the fixed prediction assumption, and set-up in a `reverse mode' naturally computed the gradients required for the backpropagation of error algorithm, as its dynamics satisfy the same recursive structure of the adjoint equation and thus, the fixed points of the prediction errors, upon convergence, equal the backpropagated error gradients which can then be used to perform backprop. We have extensively empirically validated this correspondence and used it to train large-scale and complex machine learning architectures such as CNNs and LSTMs with performance equal to those trained by backprop. 

Secondly, we have utilized the insights gained by our work with predictive coding to derive a new, and much simpler algorithm which we call \emph{Activation Relaxation} (AR). Here, instead of using separate prediction error neurons, we simply update the activation of the value neurons themselves to become equal to the backpropagated errors during the backwards iteration phase. While this eschews the fixed feedforward pass assumption required for predictive coding, it introduces a similar requirement of storing the initial feedforward pass value throughout the backwards iteration phase, so that they can then be used during the weight updates. We also empirically validate this correspondence and demonstrate that AR can be used to train machine learning architectures with the same performance as backpropagation. Importantly, we also investigate the potential for applying the same biologically plausible relaxations to the AR algorithm as we applied to predictive coding in Chapter 3, and show that the relaxations perform just as well in this new setting -- speaking to robustness and generalizability of these relaxations. Overall, we believe the AR algorithm is simpler, more elegant, and more biologically plausible than competing iterative backprop schemes such as predictive coding and equilibrium-propagation. However, it suffers from the limitations inherent in all iterative approaches -- the necessity to somehow store the feedforward pass values throughout the backwards pass. A clear understanding of this limitation, then opens the way for future work to try to remedy it or propose a different method entirely for solving backpropagation in the brain.

Finally, we have included some current (and unpublished) speculations on the potential solution to backpropagation in the brain for simple feedforward networks of rate-coded integrate and fire neurons, and have constructed a relatively direct method of implementing backpropagation with only a few moving components. Importantly, this construction relies heavily first on its nonstandard definition of the forward pass, using $x^{l+1} = W f(x^l)$  -- or the weights after the activation function, rather than inside of it -- which is non-standard for artificial neural networks, but is actually more biologically plausible, and secondly on the solution to the weight transport problem to allow for learnable backward weights. With these issues circumvented, we believe that biologically plausible backpropagation for rate-coded integrate and fire neurons actually turns out to be relatively straightforward. The key next move for future work, now that this base of understanding is established, is to start to attack the substantially harder problem of biologically plausible implementations of backpropagation through time, as well as with spiking neuron models.

Overall, in this chapter, we believe that we have made several clear contributions towards understanding the biological plausibility of backpropagation in the brain -- firstly by providing and empirically validating two new iterative algorithms (predictive coding and activation relaxation) and secondly by coming to a much clearer understanding of what exactly the remaining stumbling blocks to a biological implementation are.

%% file: chap7.tex
\chapter{Discussion}

The computer scientist and mathematician Richard Hamming in his insightful essay \emph{You and Your Research} describes how he would pose the following question to his colleagues at Bell Labs -- \emph{`What is the most important question in your field, and why aren't you working on it?'}. As might be expected, this made him unpopular with many of his colleagues. However, it speaks an important truth -- the absolute and overriding importance of posing and working on the right questions. An important question that is impossible is futile. A tractable but unimportant question is useless. Throughout my PhD, I have endeavoured to orient by Hamming's maxim; to find the most important yet solvable question within my field and to try to answer it. This thesis, then, can be seen as a concatenation of three questions of progressively (in my opinion) increasing scope and importance.

The first question is local to the active inference community, but is very important within it. Namely, can active inference be combined with contemporary deep reinforcement learning methods, and thus be scaled to the kind of tasks that can be handled by contemporary deep reinforcement learning? Conversely, does the theory of active inference itself contain any insights which can be useful for machine learning theorists and practitioners? I believe that through my work \citep{millidge_deep_2019,millidge2019combining,tschantz2020reinforcement,millidge2020relationship,tschantz2020control} and others \citep{tschantz_scaling_2019,fountas2020deep,ueltzhoffer_deep_2018,ccatal2020learning}, both sides of this question have been definitively answered in the affirmative. Active inference can be straightforwardly scaled up using artificial neural networks and the techniques of deep reinforcement learning while, conversely, active inference has many interesting properties and ideas which could be of use to the deep reinforcement learning community. In this thesis, we have explored several of these ideas and primarily focused on how deep active inference and deep reinforcement learning can be merged. Now that this has been answered, future work should focus on the converse -- how deep active inference \emph{differs} from deep reinforcement learning and the extent to which it can inform and lead to novel and performant algorithms in deep reinforcement learning.  

The second question is more broadly targeted to the reinforcement learning and cognitive science communities, and concerns the \emph{mathematical origins of exploration}. This question is central to a number of related disciplines such as reinforcement learning \citep{sutton2018reinforcement}, decision theory \citep{daw2006cortical}, control theory \citep{kalman1960contributions}, and behavioural economics \citep{tversky1974judgment}, which all share the same fundamental object of study -- adaptive decision-making under uncertainty. Where there is uncertainty so that the true dynamics of the environment and/or the value of each possible contingency are not known, then the optimal policy cannot straightforwardly be computed, and agents are necessarily faced with the exploration-exploitation trade-off. This trade-off arises because new information can generally only be obtained by trying new courses of action or venturing into new regions of the state-space. However, to \emph{explore} new regions necessarily has an opportunity cost of not doing what you thought to be the current best option, which could instead have been \emph{exploited}. Given that to succeed at complex tasks, it is almost always necessary to explore, it is important to figure out how to explore in the most efficient manner. Specifically, we wish to design algorithms that can acquire the information necessary for success as rapidly as possible while incurring the minimum opportunity cost. In the literature, it has been discovered that a very good heuristic for doing this is simply to optimize a combination of the greedy reward maximization objective \emph{exploit} with an additional information gain exploration term \emph{explore} \citep{shyam_model-based_2019,schmidhuber2007simple,tschantz2020reinforcement}. Specifically, the reward maximization part of the objective ensures that agents do not spend large amounts of time exploring informative but barren regions, while the information gain terms help the agent not to get stuck in locally greedy, but globally poor optima. While this objective works well in practice, its mathematical origin and nature remains obscure. In the literature, this approach is often described intuitively as simply adding an additional exploratory term to the loss function. While random entropy-maximizing exploration can be derived straightforwardly from variational inference approaches to action \citep{levine2018reinforcement}, the mathematical origin and commitments of specifically optimizing information-seeking exploration terms has remained mysterious. This is the second question we set out to answer in this thesis -- mathematically, from what sort of fundamental objectives do information-gain exploration terms arise, and how can we characterise the possible space of such objectives?

In Chapter 5, we answer this question. We show that information gain maximizing exploration arises from minimizing the divergence between two distributions -- a predicted or expected distribution over likely states, given actions, and a desired distribution over states, which encodes the goals of the agent. This differs crucially from \emph{evidence} objectives, which are typically used in control as inference schemes \citep{levine2018reinforcement,rawlik2013stochastic}, which only seek to maximize the likelihood of the desired states, rather than explicitly match the two distributions. This finding has important implications for a wide range of fields. Specifically, we argue that any kind of information-maximizing exploratory behaviour can be seen as implicitly aiming for a matching of two distributions rather than a likelihood maximization. This, for instance, can explain several phenomenon, such as the probability matching behaviour that is regularly observed in human participants in cognitive science and behavioural economics tasks \citep{vulkan2000economist,daw2006cortical,west2003probability,shanks2002re}, which are puzzling under the presumption of evidence maximization \citep{tversky1974judgment,gaissmaier2008smart}. Moreover, by understanding the origin of information-seeking behaviour as emerging directly from divergence objectives, it provides us with a greater and deeper understanding of what agents which optimize these information-seeking terms are actually doing, while the ensuing mathematical understanding may allow us to manipulate these terms more confidently into more easily computable or tractable versions which could aid implementations directly.

The third question is one with the greatest scope and importance. This question is how can credit assignment be implemented in the brain? And, specifically, whether and how (if it does) the brain can implement the backpropagation of error algorithm. The solution to such a question would be of great importance to neuroscience, since it would provide a unifying view and mechanistic, algorithmic explanation of at least part of cortical function. Moreover, it would explain at a detailed level one of the core functionalities of the brain, and the one that underpins almost all adaptive behaviour. Credit assignment is crucial to any kind of long-range learning of the kind that \emph{must} be occurring in the brain. It is crucial for everything from learning the best way to form and interpret sensory representations, to action selection operations to, potentially long term memory and complex cognitive processing. While the brain undoubtedly performs a substantial amount of top-down contextual feedback processing as well as various kinds of homeostatic plasticity, which both remain poorly understood, we also know from the stunning success of machine learning in the last decade that simple feedforward passes on large neural networks trained with the backpropagation of error algorithm can accomplish tasks such as visual object recognition \citep{krizhevsky2012imagenet,child2020very}, generating realistic images from text inputs (\citep{radford2021learning}, human-passable natural language generation \citep{radford2019language}, and playing at a superhuman level games such as Go \citep{silver2017mastering}, Atari \citep{mnih2015human,mnih2013playing,schrittwieser2019mastering}, and Starcraft \citep{vinyals2019grandmaster}, which ten years ago were thought to be extremely challenging, if not impossible for computers to accomplish. Credit assignment, then, must be one of the core operations in the brain, and if we can understand this then it is possible we may obtain a grasp on how known computational algorithms are implemented in the brain which allows us to grapple tractably with its immense complexity. 

While I, and the field as a whole, have taken steps towards addressing and answering this question, we are still a long way from a viable solution for the brain. Nevertheless, I feel that, in general, with the profusion of algorithms addressing issues such as the weight transport problem \citep{lillicrap2016random,akrout2019deep,nokland2016direct}, and well as addressing issues of locality \citep{ororbia2019biologically,whittington2017approximation,scellier2016towards}, the field is close to a good solution for the case of rate-coded neurons on a temporally static graph. However, a solution under these constraints is fundamentally only an abstraction of a much messier reality, where neurons in the brain are not rate-coded but spiking, and must also achieve credit assignment not just across space (layers) but across time \citep{lillicrap2020backpropagation}. While there are some approaches which grapple with these additional, and harder problems \citep{zenke2018superspike,bellec2020solution}, there are not many and we are far from a viable global solution to this problem. I suspect it is into these new domains that future research should primarily be directed, and where important advances will be made. 

\section{Question 1: Scaling Active Inference}

It turns out the active inference approaches can be quite straightforwardly merged with those used in deep reinforcement learning and can thus straightforwardly be `scaled up' to achieve performance comparable with the state of the art. Moreover, different choices lead directly to different schools of model-free or model-based reinforcement learning. Specifically, active inference fundamentally operates on several core probabilistic distributions and objectives. The key distributions are the likelihood distribution $p(o | x)$, and the transition distribution $p(x_t | x_{t-1},a_t)$. While standard discrete-state-space active inference approaches parametrize these explicitly with categorical distributions, and optimization of the variational free energy exactly using analytical solutions resulting in a fixed-point iteration algorithm, deep reinforcement learning algorithms instead amortize these distributions with artificial neural networks, and instead optimize the variational free energy with respect to the amortised parameters (the weights of the artificial neural networks). In pure inference terms this can be seen as an E-M algorithm, with a trivial E-step (amortised inference as a forward pass through the networks), and then an iterative M-step which corresponds to a gradient step of stochastic gradient descent on the weights of the neural networks. 

The next translation is to identify the value function in deep reinforcement learning with the path integral of the expected free energy over time in active inference. Therefore the derivation of the optimal policy under active inference $q(\pi) = \sigma(\int dt \mathcal{G}(\pi)_t)$ can be seen as a design choice in active inference to perform what is effectively Thompson sampling over the softmaxed value function in reinforcement learning, a choice which is often, but not necessarily made in comparable reinforcement learning algorithms \citep{osband2015bootstrapped}. Finally, the sole remaining question remains how to compute or approximate the path integral of the expected free energy, or the value function. While the discrete state-space active inference literature typically only deals with short time horizons and small state-spaces where this integral can be exhaustively computed \citep{da2020active}, or else by simply enumerating and pruning unlikely paths \citep{friston2020sophisticated}, the statespaces and time horizons in deep reinforcement learning problems are typically large enough that this approach becomes infeasible. 

The sole remaining question remains how to approximate this path integral, and here we can augment active inference with methods well used in the deep reinforcement learning community. The approach taken by model-free reinforcement learning is to utilize the iterative and recursive nature of the Bellman equation to maintain and update at all times a bootstrapped estimate of the value function \citep{kaelbling1996reinforcement,mnih2013playing}. This approach was pioneered through the temporal-difference \citep{sutton1988learning}, and Q-learning algorithms \citep{watkins1992q}, and gives rise to the model-free family of reinforcement learning algorithms. Translating this into the terms of active inference is quite straightforward. Since the expected free energy objective can be factorised into separate independent contributions for each timestep, the path integral satisfies a similar recursive Bellman-like equation. This equivalence has been recently used to prove the similarities between active inference and reinforcement learning \citep{da2020relationship}. This approach allows us to straightforwardly define Q-learning and actor-critic like active inference approaches, as was pioneered in my paper \citep{millidge2019deep}. One minor distinction is that the computation of the expected free energy contains an information gain term which necessitates a model of the states or dynamics of the world, which would not be necessary when using standard reinforcement learning approaches, but this information gain yields superior exploratory capabilities and ultimately performance.

While, in \citet{millidge_deep_2019} we explicitly included action within the generative model in the agent, so that the action prior $p(\pi)$ becomes the path integral of the expected free energy, and the variational policy posterior $q(\pi)$ becomes the independently trained policy, thus recapitulating and providing a variational inference gloss on standard actor-critic algorithms, this is fundamentally a design choice. If we instead ignore the policy prior $p(\pi)$ and define the variational policy posterior $q(\pi)$ directly in terms of the value function, we obtain an algorithm very similar to soft-Q-learning from deep reinforcement learning \citep{haarnoja2018soft}, except it optimizes a value functional of the expected free energy instead of the reward.

Conversely, we can take the approach used in deep reinforcement learning to approximate the value function at every timestep through samples of model rollouts. This is straightforward because the value function is just fundamentally the expected value of the reward across all possible trajectories under a given policy. Using model-based rollouts to approximate this is simply taking a monte-carlo approximation of an expectation where the real environmental dynamics are approximated by the model's transition dynamics. By using importance sampling on this objective, we can unsurprisingly see that the goodness of this approximation depends crucially on the match between the true and modelled dynamics.

If we are equipped with a model of the transition dynamics of the world $p(x_t | x_{t-1}, a_{t-1})$, we can approximate the path integral of the expected free energy over time in a similar way. By simulating rollouts through the transition model under a given policy, and then averaging together the path integral of the expected free energy across rollouts, we form a monte carlo estimate of the expected free energy value function. This can then be used to directly compute the posterior distribution over policies, or else can be fed into an iterative planning algorithm such as path integral control or CEM which can then be used to obtain an action plan. Using model-predictive control (replanning at every step), then allows for the creation of flexible plans for any given situation. Due to utilizing a transition model and simulated rollouts to estimate the local value function, instead of bootstrapping from previous experience, this model-based approach is substantially more sample efficient than the model-free alternative. In the \citet{tschantz2020reinforcement} paper, we took this approach and demonstrated performance comparable to or superior to standard model-based benchmarks. Additionally, as before, the exploratory properties of the expected free energy functional lead to improved performance.

Given that we thus know that active inference can relatively straightforwardly be mapped to existing algorithms in deep reinforcement learning, we now turn to the other face of the question -- whether deep reinforcement learning can learn anything from active inference. We again argue in the affirmative. Namely that active inference, through the expected free energy functional, provides superior exploratory capabilities of active inference agent which, in challenging sparse-reward tasks are necessary to obtain intelligent behaviour where random exploration is simply not sufficient. While in reinforcement learning this is a small literature on training agents with additional exploratory loss functions \citep{pathak2017curiosity,shyam_model-based_2019,still2012information,chua_deep_2018,nagabandi2019deep,klyubin2005empowerment}, active inference provides a mathematically principled and unified way of looking at this, rather than simply postulating ad-hoc additional loss functions \citep{oudeyer2009intrinsic}. Moreover, active inference also provides a theory of reinforcement learning deeply grounded in variational inference, and can thus, for instance, be straightforwardly extended to POMDP models in a way that is nontrivial for standard reinforcement learning algorithms. Finally, by proposing a unified objective of the expected free energy, active inference allows, in principle, all hyperparameters of the algorithm to be optimized directly by gradient descents against this objective, thus theoretically obviating the need for expensive hyperparamter sweeps and tuning. 

Although the close connection between deep active inference and deep reinforcement learning is now understood, and we know it is possible to scale up active inference to the level of deep reinforcement learning, there still remains much work to be done actually realizing this connection and constructing deep active inference agents, using the insights of active inference, which can compete head-to-head with the state of the art in deep reinforcement learning and, potentially, exceed it. I believe that especially as the field moves towards facing more challenging environments with sparse rewards, the exploratory drives implicitly embedded within active inference agents will become increasingly important and impactful, since many current environments possess straightforward dense rewards which provide a continuous reward gradient from any initial condition to a successful final policy. In such environments purely random exploration suffices for learning effective policies. However, such environments are not in general representative of the kind of environments that face biological organisms in the real world, and thus to model their behaviour additional exploratory instincts appear to be required. 

Furthermore, on a general note, while current work has been focused on trying to maximize the commonalities between deep active inference and deep reinforcement learning, as the goal has been to establish the connection and derive proof of principle scaled up active inference models, later work should go the other way, and try to retain the scalability of deep reinforcement learning while maximizing on what is unique about active inference. 

One intriguing possibility in this direction is to experiment with more complex distributions of rewards. While active inference can be formulating in a reward maximizing way, the real object active inference handles is the biased generative model $\tilde{p}(o,x)$, or the desire distribution $\tilde{p}(o)$. While this can be defined to be equal to reward maximization by simply defining the desire distribution to be a Boltzmann distribution over the realized rewards $\tilde{p}(o) = exp(-r(o))$ \citep{friston2012active}, this is not the only way to do it. Indeed, more complex and potentially multimodal reward distributions could be defined and optimized directly in the algorithm. In theory this could lead to more flexible behaviour or, alternatively, being able to model more complex, context-sensitive or contingent rewards naturally within the framework. There has been very little work done in this direction, and it remains an exciting avenue for future work.

Another interesting direction is to explicitly model the generative processes producing action within an inference framework. For instance, search algorithms such as Monte-Carlo-Tree-Search have been vital in the success in key reinforcement learning tasks such as playing Go and Chess \citep{silver2017mastering}, and can theoretically be written in a probabilistic generative model. Doing this would then allow them to be combined productively with all the standard tools of Bayesian and variational inference and could potentially lead to substantially more flexible algorithms for action selection which would provide extremely powerful inductive biases over standard MLP policy modules which would allow fast and very effective learning. In a similar vein, continuous-action planning algorithms, which are currently rather primitive (such as CEM \citep{rubinstein1997optimization} and path integral control \citep{kappen2007introduction}) and can only explicitly model unimodal policies, are generally quite ineffective. \citet{okada_variational_2019} has shown how many of these algorithms can be directly modelled as part of a generative model and directly used in variational inference, and they use this result to derive more effective algorithms such as multimodal CEM \citep{okada2020planet}. Further extending this work may lead to the derivation of highly effective and efficient planning algorithms for continuous control, which are currently sorely lacking and which would lead to a substantial improvement in the abilities of model-based continuous control. 

Another interesting avenue, which has begun to be explored in the literature \citep{friston2020sophisticated}, which may lead to more nuanced and effective forms of exploration, is to explicitly model, in model rollouts, the change in its own beliefs the agent expects to encounter. If this is explicitly modelled, then the agent can design exploration strategies specifically to test hypotheses and explore different strategies. In short, this kind of meta self-knowledge of the likely changes of one's own beliefs are vital for the kind of scientific and experimental thinking that often characterises humans' phenomenological experience of the planning process. Such agents would be able to intelligently consider and compute the value of information both now and the expected value of information in the future under their expected future beliefs. Basic models of this have been explored using the discrete-state-space paradigm \citep{friston2020sophisticated,hesp2020sophisticated}, however figuring out how to implement this within the deep reinforcement learning paradigm in a computationally tractable and efficient way, as well as to test its performance on tasks which require such nuanced exploration strategies remains a serious challenge and a worthy research project.

Finally the perspective of active inference -- that action and control are merely inference problems over a graphical model which includes action variables -- naturally lends itself to an understanding of different kinds of inference -- specifically amortised and iterative inference \citep{millidge2020reinforcement,kim2018semi,marino2018iterative}. Understanding how these different types of action can be combined and merged together, to inherit the strengths of both and ameliorate the weaknesses of each other, is ultimately going to be very important in designing algorithms which can initially learn rapidly from data and then slowly converge to a high asymptotic performance. There is also strong, but circumstantial evidence, that a system like this, which combines iterative and amortised inference, takes place in the brain. For instance, whenever you start learning a new skill, it takes a lot of thought and explicit mental planning, but you can learn quickly without needing an extremely large number of interactions with the environment. However, as you continue to practice, slowly your skills become habitual. They do not need mental effort and can occur almost automatically, allowing you to focus on other things while they are occurring. This distinction between conscious, effortful action and unconscious, effortless habit is precisely the distinction between iterative and amortised inference. The benefits of having such a hybrid system are obvious. It is quick to learn new skills since it can explicitly plan with them, while once a skill has been practiced many times it can be offloaded onto a computationally cheap habit system. This eliminates the necessity, in current model-based reinforcement learning systems, to undertake expensive model-predictive control and planning on every single timestep, even when the system has practiced a given contingency many many times, and also where computing the best action is actually extremely straightforward.

While we have undertaken some preliminary work in this direction, as is reviewed in this thesis, really the combination of the two to design systems with explicit planning and habit systems is just beginning and there are very many questions which remain unanswered. For instance, what is the best way to train the habitual system -- should it be trained to mimic the decisions of the explicit planner, or be completely independently trained on the reward, or both (i.e. by using the output of the planner as a regulariser of some kind)? Should the habit system and explicit planner optimize separate reward functions (for instance, should the planner be more exploratory and the habit system just optimize rewards?)? Should the habit system be used by the planner in any manner -- for instance habit value functions could be used to endstop the model rollouts of the planner to provide better local value function estimates? Should the habit policy be used to initialize the planner? How should these systems interact to produce actual output actions -- should the output be the sum of both systems? or should there be some gating mechanism which selects one or the other to produce the output? If there is such a gating mechanism, how does it work and how should it compute? How do we know when to turn off the planner and just use the habit system, if ever? The answer to all of these questions is a fascinating combination of engineering practice and machine learning theory, and the end result of getting these questions right will be flexible, robust, adaptable and sample-efficient systems which also have a high asymptotic performance with low latency and computational cost when the habit is established. Such systems could also be used to model similar computations that occur in biological brains and may shed light into the design choices available for such system and, ultimately, their neural implementations.

\section{Question 2: The Mathematical Origins of Exploration}

In Chapter 5, we delved deeply into the mathematical origins of exploratory behaviour. We saw that to obtain information-seeking exploration as a core part of the objective functional, in addition to reward maximization crucially entails minimizing a \emph{divergence} objective instead of an evidence objective. We then related this new dichotomy between divergence and evidence objectives to a wide range of currently used objectives within the reinforcement learning and theoretical neuroscience communities. The importance of this result, really, lies not in the relationship to existing methods, but what it tells us about the deep foundation of exploration. Put simply, we see that extrinsic exploratory drives emerge from trying to match rather than maximize. Matching tries to maintain the complexity of the inputs, so that given a complex desire distribution, agents are driven to stabilize a similarly complex future. Conversely, maximizing implicitly tries to simplify the inputs, ideally a maximizing agent would collapse all future inputs to a dirac delta around the future reward. It is only the extent to which there is uncertainty in the world, or in the reward function, or a lack of controllability in the world which prevents this full maximization. This difference is what gives rise to intrinsic exploratory behaviour in the divergence minimization case, and to effectively anti-exploratory, information-minimizing behaviour in the evidence, reward, or utility maximizing case. We additionally see that reward maximizing agents do not have, and cannot have, any intrinsic exploratory drives, since the very nature of their objective compels them to minimize information gain. Information and learning, to them, is a cost which must be borne, and not a reward to be pursued for its own sake.

This additional information gain term is very important, because it drives agents which optimize it to explore and seek out new contingencies, and to find and update upon resolvable uncertainty in their world. This means that agents which have an expressive desire distribution, will tend to learn faster and better world models, as well as pursue more exploratory policies, which in the long run lead to higher performance than purely reward-maximizing agents, even when judged on rewards alone. This has been investigated by ourselves in Chapter 4, as well as under many other approaches in the literature. What is most interesting here is that our understanding of these objectives as \emph{divergence} objectives provides a precise mathematical characterisation of what these objectives are implicitly doing.

It is important to note that it is possible to derive some form of information gain exploration directly from reward maximization -- in the form of explicitly computing and calculating with the \emph{value of information} \citep{still2012information,schmidhuber2007simple,osband2019deep,tishby2011information}, where we can operationalize the value of information purely in terms of reward as the additional amount of reward expected given better policies as the result of obtaining and integrating the information into your world and policy models. While this value of information computation is theoretically optimal given a reward maximization objective, in practice it is intractable to compute exactly, and there has been little investigation in the literature as to direct approximations of this term. However, there are some heuristic approaches which seek to approximate it, although it is not clear how well. For instance, \citet{osband2019deep} argue that using `optimistic value functions' which automatically up-weight unknown contingencies can lead to a good approximation of the value of information, since in practice, the agent will automatically explore until it has diminished its optimistic bias to the extent that the bias for all contingencies is lower than the current best option. This approach, under the names of the upper-confidence bound is widely used in the multi-armed bandit literature \citep{garivier2011upper} and additionally has been used to great effect in Monte-Carlo Tree Search algorithms \citep{kocsis2006bandit}, and has been investigated to some degree within deep reinforcement learning \citep{silver2017mastering}. Another approach, known as Thompson sampling \citep{russo2016information}, explicitly computes or approximates the posterior distribution over actions given the current history of observations and rewards, and then achieves some degree of exploration by sampling from this posterior -- the idea being that in uncertain regions, the posterior distribution should be fairly uniform, and thus provides effectively random exploration, while when the posterior is sharp then it is likely that the true optimum has been found and thus exploration is unnecessary and costly. 

An important challenge is that inference based approaches such as control as inference do not naturally compute any analogue of the value of information. This is because, ultimately, these approaches take a mean-field factorisation across time and split their objective up across time-steps. This means that information can only flow through time through the transition model, and thus the agent cannot model any kind of learning in the future, where information it may or may not discover in the future leads it to change its model, leading it to perform better (or worse) in the future. Due to this limitation, which is ultimately applied for reasons of computational tractability, approaches like control as inference do not compute any kind of value of information across time-steps. This limitation also applies to the information-seeking methods we discuss which arise from divergence functionals. If these functionals are also mean-field factorized, then agents only seek to maximize the information-gain in the current time-step. Extending these methods by relaxing the temporal mean-field assumptions will likely yield more effective and nuanced forms of exploration, which can induce consistent exploratory behaviour across multiple timesteps and thus handle more complex contingencies. However, designing effective and mathematically tractable algorithms which can do this largely remains an avenue for future work.

However, this information-maximizing exploration with divergence functionals is not done explicitly to gain any kind of future reward. Instead, agents optimizing divergence objectives treat optimizing information gain as an intrinsic good. This is what allows an information gain objective to arise even though the objective satisfies the same mean field assumptions as the control as inference objective. However, this means that to some extent divergence minimizing agents will continue to explore beyond the point which is strictly necessary for reward maximization. This means that, in effect, from the perspective of a purely reward maximizing agent, divergence minimization is just an additional exploratory heuristic by which to approximate the value of information terms within a mean-field formulation. However, in general, it has been empirically found that the information seeking exploration in a mean-field fashion, while not strictly the value of information, gives a good approximation in general and will lead to good performance especially in high dimensional, sparse environments. However, because it explicitly trades off a reward maximizing and an information-seeking objective, it will tend to over-explore relative to pure reward-maximizing value of information computation, by exploring regions with much resolvable uncertainty but relatively little reward, and will continue exploring even when it is likely (but not certain) that the optimal solution has been found. However, as long as all uncertainty in the environment is resolvable, then the divergence objective will eventually converge to the reward maximization objective since the information gain term will eventually become negligible once the agent possesses a very good and accurate world model.

An interesting direction for future work will lie in relaxing the mean field assumptions which currently underpin all of these functionals. While it is likely that a full relaxation will be intractable, there are many intermediate relaxations which have been proposed in the general variational inference literature which could prove highly fruitful within the control task. For instance, the Bethe free energy and related objectives \citep{yedidia2001generalized,pearl2014probabilistic,schwobel2018active}, allow for temporal pairwise correlations to be explicitly considered. Moreover, there is a highly general family of `region graph' approximations \citep{yedidia2005constructing,yedidia2011message} which have been developed within the variational inference literature, and which allow for more complex interactions to be modelled within a relatively tractable computational framework and which have been found to improve inference performance. If there is a way to compute successively better Bayesian approximations to things like the value of information, or to relax the temporal mean-field approximations made in contemporary divergence and evidence objectives for control, it will likely come from a thorough mathematical and experimental investigation of these more advanced and accurate approximation techniques. Understanding how the information gain functionals from divergence objectives function and change as the temporal mean-field approximation is relaxed is especially interesting, since preliminary investigations, even within the mean field paradigm, show that when expressly writing out the objective in terms of entire trajectories, terms similar to empowerment \citep{klyubin2005empowerment} and filtering information gain (backwards in time) result. 

Related to the relaxation of the temporal mean-field approximation, there also needs to be much work allowing agents to explicitly model changes to their own beliefs in the future. This is necessary to truly compute realistic information-seeking objectives when utilizing multi-step planning algorithms, since currently all information gain is computed with respect to the agent's current beliefs, which will not necessarily hold in the future if and when it actually eventually reaches this information. Such a process would enable an agent to understand how various kinds of information would change its own beliefs and policies, and thus be able to plan for multiple sequential `realizations'. Human planning is certainly capable of such introspective capabilities, where we can, for instance, decide to seek out and investigate certain phenomenon in order to be able to understand and better seek out information about another task, and so on. Some fascinating recent work has begun to explore these sorts of metacognitive abilities within the active inference framework \citep{friston2020sophisticated}, but only within a discrete state space and nonscalable task. The true issue with such approaches will be their inherent computational difficulty, since on a naive approach they will require the agent to simulate its own belief updates, requiring it to store and introspect upon a copy of its own models and inference procedures. However, since such metacognitive abilities are likely crucial to effective long range planning, an important strand of future work will be designing approximations and objectives which can accomplish this in a computationally tractable manner.

Additionally, the divergence vs evidence framework presents a new class of objectives which are, in theory, distinct from the usual paradigm of reward or utility maximization. Here, instead we simply seek to minimize divergence to a complex reward or desire distribution. In theory, the idea of having a complex, multimodal and potentially non-scalar desire \emph{distribution} instead of a simply scalar reward function to maximize is that in theory it allows for more complex notions of goals or desires to be implemented and optimized by agents. Specifically, it allows for \emph{vector-valued}, and \emph{multimodal} goals which are not well handled within the standard reward-maximization framework, but which can be straightforwardly handled within our formulation of a desire distribution by both evidence and divergence objectives. While current work has mostly focused on demonstrating the equivalence of reward-maximization, and the desire distribution under certain conditions (the desire distribution being a Boltzmann distribution of the reward), and thus showing that the probabilistic case is a strict generalization of the scalar deterministic case, the real interest of the probabilistic representation is precisely how it can differ from simple reward-maximization. Much work remains to be done to understand the possibilities for more flexible and expressive goal or value representation which is unlocked by this more general formalism, and how it can be leveraged to design artificial systems which perform more capably in practice.

Finally, the idea of divergence minimization also has extremely close links with Markov-Chain-Monte-Carlo inference procedures, where it has recently been realized that much of this paradigm can be expressed within a simple framework of stochastic dynamical systems theory, whereby all the various samplers can be interpreted as implementing a certain stochastic differential equation which explicitly performs a gradient descent on a divergence objective \citep{ma2015complete}, with different algorithms in the literature simply specifying different noise terms and solenoidal flows \citep{yuan2017sde}. Beyond this, the idea of divergence minimization can also intriguingly be linked to recent advances in stochastic non-equilibrium thermodynamics \citep{seifert2012stochastic}, which has developed ways to translate classic thermodynamic notions of entropy and entropy production from properties of large ensembles to properties of individual statistical trajectories \citep{esposito2010three1}. A crucial result in this new formalism of stochastic dynamics is that any system with positive entropy production can be construed as minimizing a divergence between its current state and its ultimate steady state density \citep{esposito2010three1} -- and can thus be interpreted as performing a form of direct divergence minimization -- thus potentially implying that this objective may in some sense be a more natural one for systems and agents to perform than pure evidence maximization, and secondly that the laws of thermodynamics themselves may implicitly require information-maximizing behaviour from disspative non-equilibrium systems. While these links currently remain speculative, and further investigation will likely discover greater nuance and require some qualification of these claims, in my opinion there is significant potential here to link discoveries in stochastic thermodynamics to help us build a fully general picture of the necessary nature of exploratory behaviours in systems evincing the classic action-perception loop.

\section{Question 3: Credit Assignment in the Brain}

By showing that predictive coding can approximate backpropagation along arbitrary computation graphs, instead of just MLPs, we have specifically turned predictive coding from a direct model of brain function or perception, into a learning algorithm which can be applied to arbitrary architectures. This dual perspective, where predictive coding is both a generic learning algorithm, as well as specifically a model of learning and perceptual inference in the brain is most interesting and, as far as I am aware, is unique to predictive coding. Specifically, casting predictive coding as a learning algorithm makes clear several interesting correspondences between inference procedures and learning. Specifically, that we can derive a learning algorithm on a computational graph by trying to infer the values of the nodes in the graph. Predictive coding additionally provides for a straightforward extension to backprop in the form of precisions, which allow for the learnable up or down weighting of certain gradient signals depending on the intrinsic noise of their generating process. Such a system, while not particularly useful in the standard machine learning paradigm of independent, identically distributed datasets, may prove extremely important for learning with more ecologically valid sensory streams which contain various amounts of noise and distractor information which should not be learnt from. This perspective of learning and credit assignment as a kind of inference also immediately lends itself to further applications and extensions beyond just precision. For instance, the effect of different generative models other than Gaussian remain to be determined, as well as potentially different optimization procedures and variational functionals to be optimized. In general, this perspective allows the highly developed machinery of inference in graphical models \citep{pearl2014probabilistic,ghahramani2001propagation,beal2003variational,yedidia2011message} to be deployed to improve credit assignment and optimization algorithms. This area, I believe, is an exciting and potentially highly impactful one for future work since the impact of improving either credit assignment or optimization processes, which are at the heart of all of modern machine learning, will necessarily be substantial.

Furthermore, by demonstrating that many of the biological implausibilities in the predictive coding scheme can be relaxed \citep{millidge2020relaxing}, we, for the first time, demonstrate a biologically plausible local approximation to backprop which does not suffer from either issues of locality or issues of weight transport. Furthermore, we develop a novel and much simplified algorithm -- Activation Relaxation -- which possesses relatively straightforward, local, and elegant update rules when compared with predictive coding which succeeds in approximating the backpropagation of error algorithm to arbitrary accuracy given enough iteration steps. Moreover, we have shown that the same relaxations which work with predictive coding, also work with the activation relaxation algorithm, thus demonstrating the robustness and efficacy both of the AR algorithm and of the methods of relaxation utilized. Crucially, if we look at the final, relaxed AR update rule (Equation \ref{fully_relaxed_AR_update}, we see that the change in activities for a layer only requires the current activation values of the layer above, mapped through the backwards weights which are learnt independently of the forward weights, be subtracted from the current activation of the layer. This is sufficient, over a number of iterations, to allow the activations of each layer of the network to converge to the gradients of backprop. Then, once this is achieved, the weights can be updated. 

While these algorithms have considerable advantages -- they exactly approximate backprop given enough iterations, they require only biologically plausible local update rules, and they are simple and potentially straightforward to implement in neural circuitry, they also have substantial disadvantages. I believe these disadvantages are worth discussing in some depth since they provide a precise specification of the areas for improvement in current algorithms. The fundamental disadvantage of iterative schemes like this is their iterative nature. Specifically, they require separate phases of operation -- a forward phase which is equivalent to a feedforward pass through the network, and a backwards iterative phase of multiple dynamical iterations. A significant issue is that it is unclear whether such phases can realistically exist within the brain. While there is evidence for different oscillatory frequency bands in the brain \citep{buzsaki2006rhythms}, and even in superficial vs deep cortical layers \citep{bastos2020layer,bastos2015visual}, it is unclear whether these rhythms do, or can, coordinate separate feedforward and iterative phases. Such a scheme, if implemented in the brain, would form a kind of clock, only allowing feedforward information to be processed in short bursts in between the iterative phases. While not impossible, this seems at odds with our current understanding of the brain where feedforward and feedback inputs are combined together in real time. Another very straightforward problem is simply the number of iterations these schemes require. The brain cannot wait for tens or hundreds of iterations until convergence, even under generous assumptions about rhythmic activity implementing the phases, while these schemes often require a substantial amount of iterations to converge to nearly exactly the backprop gradients. While this could be ameliorated somewhat with high learning rates, and settling for less than exact convergence to the backprop gradients, it has not yet been extensively investigated whether the algorithms are even stable under such conditions. Understanding and optimizing the number of iterations and various parameters like the learning rate is still an open area of research. Unlike backprop with deep neural networks, where all hyperparameters have been effectively extensively tuned in the literature over a decade of experimentation, good hyperparameter settings for these alternative algorithms have barely been explored at all, and their empirical limits of performance at scale have largely yet to be determined. A final serious difficulty with such approaches is that to match backprop they require some level of nonlocality in time, where information from the feedforward pass is stored and then utilized throughout or at the end of the backwards pass. This storage of information is fundamentally necessary because the backprop gradients depend only on the state of the network in the feedforward pass. If the state changes due to the iterative algorithm, then the old information from the forward pass must be stored somehow to maintain convergence to backprop on the forward pass. This manifests itself in predictive coding as the fixed-prediction assumption, which explicitly assumes that the values of the forward pass are stored. In the AR algorithm, it manifests in the necessity to store the original value of the activity neurons to use to update the weights after the backwards pass is complete. Due to this storage, the actual update equations become nonlocal in time. This shortcoming could potentially be addressed in two ways. Firstly, it might be possible to store the information from the forward pass, for instance in local recurrent units which are insulated from the activity changes in the dynamical iterations, or secondly, if the number of iterations is short enough, such information could be persisted simply through multi-step recurrent connectivity. Nevertheless, even if this could be done, the circuitry to align and ensure the correct time of arrival of all the necessary signals could be quite complex.

In this field, it is important to reflect deeply upon the role and utility of simplified models of neural dynamics. Almost all work in this area operates with very simple models of rate-coded integrate and fire neurons, typically often in a temporally static `instantaneous' computation graph. Biological plausibility within this model, while a somewhat vague concept, is often defined by conditions such as only local connectivity, or that connectivity must be additionally Hebbian, and that neural connectivity cannot be too precise, and that information cannot be directly transmitted backwards along axons. However, this definition necessarily ignores some important aspects of reality. Obviously the brain uses spiking neurons and must also achieve credit assignment through time, but additionally there are even questions about the simplification of neural architecture. For instance, typically such models implicitly assume a multilayer perceptron (MLP) style architecture as just a stack of fully connected layers, however each region of the cortex has an intricate 6-layer structure, and it is not clear whether each such layer in the cortex should be considered as a single layer of the MLP, or only the cortical region. If the former, then the different properties and neurophysiology of each cortical layer is not modelled. If the latter, then it is not at all clear to what extent the entire cortical region \emph{can} be modelled as a simple fully connected layer. An additional interesting neurophysiological fact which is rarely explicitly modelled is the predominance of cortical columns in the visual cortical regions. These columns do not at all correspond to fully connected layers and it is not yet fully clear what their computational role is. A straightforward hypothesis is that they are the brain's way of implementing a local receptive field operation, like convolution in convolutional neural networks, but without the advantage of shared weights across space which is core to the generalization capabilities of the CNN \citep{hawkins2007intelligence}. 

The question remains, however, how applicable are such simplified rate-coded models to the full complexity of credit assignment in spiking networks through time. The hope, ideally, is that by and by large important computational primitives and algorithms which have been developed for biologically plausible credit assignment in rate-coded neural networks remain functional, or only require minor adaptation, to work in a spiking context. This is a strong possibility, but it could also be completely false, and that the brain implements an algorithm which relies heavily on the unique properties of spiking networks for its credit assignment capabilities. Ultimately, before we fully understand the mechanisms of credit assignment in the brain, it will be hard to fully assess the degree to which work on rate-coded models will generalize. There has been some preliminary successes though. For instance, the surrogate gradient technique shows that with only minor tweaking (defining a surrogate gradient to avoid the nondifferentiable threshold spiking function), backprop through time (BPTT) can be straightforwardly used to train spiking neural networks for complex tasks \citep{zenke2018superspike,neftci2019surrogate}. However, these methods currently utilize the biologically implausible BPTT algorithm and it is unclear to what extent biologically plausible alternatives to BPTT can be straightforwardly applied in such a manner. To answer such a research question would be an important and timely research agenda which would substantially advance our understanding of the generalizability of such models.

Fundamentally, the core challenge is that of time. Most models (with some exceptions \citep{bellec2020solution,schiess2016somato}) focus only on backpropagation and credit assignment through space (i.e. layers of a neural network architecture) and not through time. However, time is an inextricable component of the computation in the brain. Fundamentally, perception in the brain is not about handling static $i.i.d$ datasets, but rather \emph{filtering} on constantly changing sensory streams of information. Moreover, credit assignment through time is in some sense a substantially harder problem, in that the key information necessary at each step disappears, while when backproapgating through space the information is always present somewhere in the graph. In the case of BPTT, all the intermediate activations at each timestep are stored and then replayed in backwards sequence once the sequence has ended. However, such an acausal solution is clearly not suitable for computation in the brain. There are alternatives to BPTT which do not require explicitly repeating computations backwards through time, but allow for online integration of gradients. The key such algorithm is the RTRL algorithm which maintains a Jacobian of values at each timestep and iteratively updates them at every timestep. In algorithmic terms, RTRL corresponds to forward-model automatic differentiation while BPTT corresponds to reverse-mode. Unfortunately, however, RTRL is substantially more expensive to implement on digital computers, scaling with $O(n^4)$ where N is the number of parameters, rather than $O(n^2T)$ for BPTT, where T is the time horizon. One key advantage of RTRL however is that unlike BPTT it is not bound by sequence length. While BPTT must choose some point to stop and then backpropagate all the gradients and then truncate gradients from after time T, RTRL can operate on sequences with indefinite length without issues, although credit is slowly diluted away through time. However, RTRL is also likely not neurally plausible and it is not clear how to implement RTRL in systems with multiple recurrent layers interacting without a blowup in the number of learning rules required. My personal hunch is that extending current models of local, biologically plausible credit assignment to spiking neural networks will not be especially challenging, although they might not be the method that the brain uses. However, I believe that credit assignment through time is a fundamentally different and more challenging problem than credit assignment through space, and that ultimately, to solve the problem of credit assignment in the brain we must grapple head on with the problem of local credit assignment through time. 

There are already some methods existing in the literature which accomplish this, specifically eligibility-prop \citep{bellec2020solution} proposes eligibility traces to compute the RTRL algorithm in a local fashion. However the method currently only works for a single recurrent layer while the brain is deep both in time and in space, and that this depth requires new algorithms, or the combination of existing algorithms for backprop in space with those for backprop in time. It is also unclear to what extent the brain computes the full credit assignment mandated by RTRL or else some sparse approximation therein, perhaps aided by the natural sparsity properties of spiking neural networks. 

An additional consideration when thinking about credit assignment in the brain is the question of feedforward vs feedback processing \citep{kriegeskorte2015deep}. While the BPTT is only designed for backpropagating through immediately recurrent feedforward neural networks, the brain actually contains a multitude of long-range recurrent loops mediated by top-down feedback connectivity \citep{felleman1991distributed,grill2004human}. Understanding the properties of these and whether they are used for credit assignment, or whether the brain computes credit backwards through these top-down connections is also crucial for understanding the full picture of credit assignment in the brain. An additional, and almost entirely unanswered question is the role of long term memory in credit assignment. Current methods, including BPTT, typically use some temporal cut-off to stop considering gradient information beyond some time-horizon, and a key flaw in naive recurrent architectures is that information is slowly lost over time \citep{ollivier2015training,hochreiter1997long}. The key innovation in LSTM units is that they contain an explicit forget/remember gate which allows for memories to be stored potentially indefinitely \citep{hochreiter1997long}. There are additionally various modifications for recurrent RNNs which help ameliorate this problem as well \citep{ollivier2015training}. Since (we assume) the brain is equipped with a fairly standard recurrent architecture, the architectural or learning-rule modifications that enable it to avoid this problem and to successfully store and utilize even long term credit assignment remains to be explored and is a very important and fundamental question in understanding credit assignment and learning in the brain. One hypothesis is that synapses in the brain may store a temporal hierarchy of eligibility traces which retain information over progressively longer timescales, thus allowing them to act on information which has been accumulated even in the relatively distant past. However, figuring out the actual specific operation of such a system remains an open research challenge.

A further interesting question, raised by the recent successes of instantaneous feedforward architectures such as transformers in processing sequential data such as natural language text over recurrent architectures such as LSTMs is to what extent recurrence is actually a useful computational primitive for handling sequence data as opposed to one the brain may be forced into due to its inherent computational constraints and need for online processing instead of batch processing at the end of a sequence, as in transformers. A key advantage of attention, as implemented in transformers \citep{vaswani2017attention}, is that it enables arbitrary time-to-time modulation, rather than in recurrent architectures where the immediate past inputs are combined with the present ones to predict the future. In this way, transformers can handle data in a fundamentally acausal manner, with accompanying computational advantages. It remains to be seen whether attention like mechanisms can be implemented in a recurrent way, whether recurrent architectures can be successfully scaled to the level that transformers have done, or whether they remain on an inferior scaling curve \citep{kaplan2020scaling}, and the precise mechanism by which similar sequence computations are implemented in the brain. The key lesson of attention is that recurrence is not the only way to handle intrinsically sequential inputs. It remains to be seen whether other non-recurrent mechanisms are implemented in the brain.

Importantly, although this line of work has focused on determining whether the backpropagation of error algorithm can be implemented in the brain, which is inspired by the impressive success of modern machine learning, which is based on this algorithm, it is not entirely certain that the brain achieves credit assignment through backpropagation at all. There are several alternative methods which are worth discussing in some detail. For instance, it is also possible that the brain may be implementing some more advanced kind of learning algorithm than stochastic gradient descent. \citet{meulemans2020theoretical} showed that target propagation implements an approximate version of a second order gradient descent scheme -- known as Gauss-Newton optimization. Similarly there are other optimization methods, such as conjugate gradients, or coordinate ascent, or fixed-point iteration, as well as a variety of probabilistic message passing schemes which do not require explicit gradients to be computed \citep{yedidia2011message,parr2019neuronal}. Moreover, it is possible that the brain, while needing to compute gradients, does not compute by the backpropagation of error algorithm. For instance, it is possible to compute gradients by finite differences, and although these finite differences perform strictly worse than automatic differentiation techniques computationally, they are very simple to implement in practice. For instance, the brain could easily contain circuits which could compute time derivatives of a constantly varying temporal stream. Then, once we have the time derivatives of two variables, it is possible to directly compute their gradient by
$\frac{dx}{dy} = \frac{dx}{dt} / \frac{dy}{dt}$. Such a circuit would only need local temporal finite differences as well as a divisive feedback connection to combine the two time derivatives, well within the possibilities afforded by known neurophysiological constraints. A further option would be that the brain simply may not compute with derivatives at all, all the while optimizing some objective function. There are many `black box' optimization methods which do not require gradients of the objective. For instance, genetic algorithms \citep{salimans2017evolution}, or other brute-force-esque algorithms may instead be implemented, and indeed it has been shown that in some cases genetic algorithms, when sufficiently scaled are able to compete with backprop on some optimization tasks \citep{salimans2017evolution,such2017deep}. Another possibility, is that the brain could learn solely by global rewards broadcast to all neurons, where learning effectively takes place via the policy gradient theorem \citep{roelfsema2005attention}. There is some circumstantial neurophysiological evidence in favour of this -- specifically the well known role of dopamine as a spur to synaptic plasticity \citep{dayan2009goal,dayan2008decision}, as well as the fact that many cortical pyramidal cells receive global dopaminergic inputs from subcortical regions. There have also been a variety of models \citep{roelfsema2005attention,pozzi2018biologically} proposed of this kind of global reward-driven learning. However, this type of learning suffers from a severe intrinsic flaw that the gradients it estimates for each parameter have extremely high variance. This is because each neuron is only provided with a global reward signal, which is ultimately caused by the interaction of an extremely large number of other neurons. Thus, averaging out all the noise introduced by all the other neurons in the brain requires a very large statistical sample, thus effectively leading to extremely high variance gradients and slow learning, which scales increasingly poorly with network size. This is precisely why backprop is such a useful algorithm, since it provides precise feedback to each neuron about the loss, thus meaning that the only source of noise is minibatch noise rather than intrinsic noise due to the activities of other neurons. This means that backprop computed gradients have much lower variance and can lead to much faster learning. Nevertheless, it remains inarguable that pyramidal cells are generally innervated by dopaminergic inputs and that dopamine, which is released when there is a reward prediction error in the basal ganglia \citep{schultz1998predictive,schultz1998reward,dayan2009goal} can strongly modulate learning. This may imply that there are effectively two learning systems on top of one another in the brain. The first, potentially older, is the global slow dopaminergic system, while the second, entirely cortically based uses precise vector-feedback with some backpropagation like algorithm. The global reward signals, then can potentially modulate various aspects of learning so that contingencies with high reward prediction error are especially salient and may induce larger changes than those without \citep{daw2006cortical,roelfsema2005attention}.

Finally, it is always possible that we have been misled by the contemporary successes of machine learning, and that the core formulation where we perceive the brain implicitly or explicitly optimizing some objective is simply wrong, and that the brain cannot be described in such a way at all. While this is a very general formulation of perception and learning, with support from both machine learning and statistics and control theory in engineering, it nevertheless could not be a productive framework in which to think about the function of the brain. Such a scenario would arise, for instance, if the brain was merely a grab-bag of various heuristics and reflexes, implicitly tuned over the course of evolution, without any serious potential for learning. While this may be true of the brains of some simple animals, it is clearly not for humans and other creatures with complex, learnt cognition. Nevertheless, if this is somehow the case, the question will then become what is a productive mathematical framework in which to think about what the brain is doing, if not in the language of probabilistic models and objective functions. It may be that the answer to this question, if it must be posed, is more interesting and productive than discovering that the brain was simply doing backprop all along. Nevertheless, it is extremely unclear at present which of these possibilities is true. My opinion, given the mathematical elegance and generality, as well as the empirical successes of the objective function viewpoint, is that it is a valuable and most likely correct framework for understanding the operation of the brain. However, it is always worth noting that this is purely a speculative hunch, and there remains little non-circumstantial evidence one way or another.

\section{Closing Thoughts}

This thesis is titled `\emph{Applications of the Free Energy Principle for Machine Learning and Neuroscience}', and throughout the thesis we have endeavoured to demonstrate how to adapt and extend methods from the free energy principle and its primary process theories -- active inference and predictive coding -- to make advances in deep reinforcement learning, and in neuroscientific theories of perception and credit assignment in the brain. Interestingly, the pattern of progress in this thesis is that, in parallel, between the neuroscience and the machine learning, is that first we simply try to adapt and extend the methods of the free energy principle to test their capabilities against other existing methods from the core literatures of these subjects, and then strive to show how process theories like active inference and predictive coding can extend and advance upon the current state of the art. Then, this process inevitably reveals new and interesting questions by itself, which are somewhat separate from the free energy principle and its process theories. These questions are firstly the mathematical origin of exploration (which emerges from considering the nature of the expected free energy objective in active inference), and secondly credit assignment in the brain, which emerges from considering how predictive coding relates to backpropgation of error. In the second set of chapters (5 and 6) we have then tried to address these further questions, using methods \emph{inspired} by the free energy principle, but not necessarily directly deriving from it. We believe that in many ways this reflects the theoretical fertility and utility of the FEP as an abstract principle, that it allows the postulation and exploration of deeper questions than would be possible without it. Indeed, this theoretical fertility may be one of the primary means of judging the utility of the FEP since, as we discussed in Chapter 2, the FEP is technically non-falsifiable and can be considered more of a mathematical principle, or perspective, rather than a theory. If this is the case, then the work in this thesis provides support to the contention that the FEP provides a useful and fruitful perspective to understanding a variety of questions both machine learning and neuroscience.

The phenomenological process of intellectual understanding is an interesting one. At first you appear to spend ages groping around in the dark, hitting various unknown objects in your way, but slowly gathering a picture of the obstacles in your path. Then, at long last, you eventually stumble your way to a light-switch and the whole vista is revealed. What were once unknown lurking obstacles are transformed into precisely specified, and tractable, objects, whose relation to one another can be seen at first glance. In the light, everything is at once clearer, but also \emph{smaller}. Questions and dilemmas which seems huge and irreducibly complex are revealed to be straightforward, even trivial, in the light of understanding. So much so that it is often easy to forget, looking back, how these questions appeared when the answer was unknown. From a personal perspective, and one I have hoped to share in this thesis, I believe I have managed to obtain and present a clear understanding of two topics which were not clear in the literature before my PhD -- whether active inference can be successfully scaled up to compete with modern reinforcement learning methods (and the relationships between the two theories), and the mathematical origins of the exploration term in the expected free energy, and its relationship to more standard functionals such as the variational free energy for perception. I have additionally contributed to the theory and implementation of predictive coding models, as well as algorithms for biologically plausible backprop in the brain. I feel, however, that while I have uncovered certain key aspects and facets of the problem of credit assignment in the brain, which were previously shrouded, I have not yet reached the point where all mystery falls away and the light pours in. Similarly, I hope that the work in this thesis has helped you, the reader, understand some things at least a little more clearly.

To conclude, we offer some ideas of for the future development of the ideas worked out in this thesis. We believe it is clear, we have shown, that active inference can be productively related and merged with the large and extremely powerful set of deep reinforcement learning agents to enable general and flexible learning based algorithms which can succeed in challenging tasks, even those requiring a considerably degree of exploration. Further progress in this field, from the perspective of active inference, should move beyond merely scaling up, and instead focus on the unique insights and ideas that active inference brings to the table. Examples of this include its distinct and exploratory objective functionals such as the Expected Free Energy, or the free energy of the Expected Future. Another potentially interesting avenue lies in active inference's more flexible consideration of reward as a specific prior \emph{distribution}, rather than a scalar value. This could enable more flexible behaviour through better reward speicfication, and include the ability to learn flexible reward distributions on the fly, effectively performing reward shaping in a semi-autonomous manner. A final avenue for exploration is the refinement of the specific generative models used in control agents. A key tenet of active inference is the idea of deriving powerful behaviours from inference on detailed and flexible generative models of the dynamics of the world. While in this thesis, we have focused primarily on simply approximating and parametrizing the distributions in the generative model with deep neural networks, a more refined approach might be to investigate whether the world can be factorized in a tractable manner, and design generative models to explicitly exploit these factorizations to allow for more tractable and efficient inference -- effectively giving the agent just the right inductive biases about the world to subserve its control objectives. Secondly, for the question of credit assignment in the brain, I believe that the key advances will come in understanding the temporal component of credit assignment, and this requires deeply understanding the fundamental computations of the brain as performing dynamical algorithms such as filtering and smoothing rather than merely static Bayesian inference. This line of research would focus attention both on how the brain can achieve backpropagation through time as well as through space. There are also extremely interesting links here with predictive coding, where dynamical approaches using generalized coordinates remain underexplored, and it may be that by further developing the explicitly Bayesian approaches to filtering provided by the dynamical formulations of predictive coding, we can better characterise and understand the temporal nature of the brain's computations. The work in this thesis relating predictive coding and Kalman filtering is only the beginning, there needs to be much work done scaling and precisely characterising the performance of dynamical predictive coding algorithms, understanding whether such recurrent predictive coding networks can be utilized to perform some form of temporal credit assignment, as static predictive coding networks can.

%% file: appendix_A.tex
\chapter{Derivation of Kalman Filtering Equations from Bayes' Rule}

In this appendix we derive the Kalman filtering equations directly from Bayes rule. The first step is to derive the projected covariance,
\begin{align}
    \E[\hat{x}_{t+1}\hat{x}_{t+1}^T] &= \E[(Ax + Bu + \omega)(Ax + Bu + \omega)^T] \\
    &= \E[Ax x^TA^T] + \E[Ax u^TB^T] + \E[Ax\omega^T] + \E[Bu x^TA^T] + \E[Bu\omega^T] + \E[\omega_x^T A^T] \\ &+ \E[\omega u^TB^T] + \E[\omega\omega^T] \\
    &= A \E[xx^T]A^T + \E[\omega\omega^T] \\
    &= A\Sigma_x(t)A^T + \Sigma_\omega 
\end{align}
Step 11 uses the fact that matrices A,B are constant so come out of the expectation operator, and that it is assumed that covariances between the state, the noise, and the control -- $\E[x u^T]$,$\E[x\omega^T]$, $\E[u\omega^T]$ -- are 0. Step 13 uses the fact that $\E[xx^T] = \Sigma_x(t)$ and that $\E[\omega\omega^T] = \Sigma_\omega$.

Next we optimize the following loss function, derived from Bayes' rule above (equation 5).
\begin{flalign}
     L &= -(y - C\mu_{t+1})^T\Sigma_Z(y - C\mu_{t+1}) + (\mu_{t+1} - A\mu_t - Bu_t)^T\hat{\Sigma}_x(\mu_{t+1} - A\mu_t - Bu_t) &
\end{flalign}
To obtain the Kalman estimate for $\mu_{t+1}$ we simply take derivatives of the loss, set it to zero and solve analytically.
\begin{flalign}
    0 &= \frac{dL}{d\mu_{t+1}}[\mu_{t+1}^T[C^T RC + \Sigma_x]\mu_{t+1} - \mu_{t+1}^T[C^T Ry - \Sigma_x A\mu_t - \Sigma_x Bu_t] \\ &- [y^TR - \mu_t^TA^T\Sigma_x - u_t^TB^T \Sigma_x]\mu_{t+1} \\
    &= 2[C^T RC + \Sigma_x]\mu_{t+1} - 2[C^T Ry + \Sigma_x (A \mu_t + Bu_t)] \\
    \mu_{t+1} &= [C^T RC + \Sigma_x^{-1}[C^T Ry + \Sigma_x (A \mu_t + Bu_t] \\
    &= [\Sigma_x^{-1} - \Sigma_x^{-1}C^T[C\Sigma_x C^T + R]^{-1}C\Sigma_x^{-1}[C^T Ry + \Sigma_x (A \mu_t + Bu_t)] \\
    &= [\Sigma_x^{-1} - KC\Sigma_x^{-1}][C^T Ry + \Sigma_x (A \mu_t + Bu_t)] \\
    &= A\mu_t + Bu_t + \Sigma_x^{-1}C^T Ry - KC\Sigma_x^{-1}C^T Ry - KC(A\mu_t + Bu_t) \\
    &= \hat{\mu_{t+1}} - KC\hat{\mu_{t+1}} + [\Sigma_x^{-1}C^TR - KC\Sigma_x^{-1}C^TR]y \\
    &= \hat{\mu_{t+1}} - KC\hat{\mu_{t+1}} + K K^{-1}[\Sigma_x^{-1}C^TR - KC\Sigma_x^{-1}C^TR]y \\
    &= \hat{\mu_{t+1}} - KC\hat{\mu_{t+1}} + K[(C\Sigma_x C^T + R)C^{-T}\Sigma_x[\Sigma_x^{-1}C^TR] - C\Sigma_x C^T R]y \\
    &= \hat{\mu_{t+1}} - KC\hat{\mu_{t+1}} + K[(C\Sigma_x C^T + R^{-1})R - C\Sigma_x C^T R]y \\
    &= \hat{\mu_{t+1}} - KC\hat{\mu_{t+1}} + Ky \\
    &= \hat{\mu_{t+1}} + K[y-C\hat{\mu_{t+1}}]
\end{flalign}
Where $K =\Sigma_x^{-1}C^T[C \Sigma_x C^T + R]^{-1}$ and is the Kalman gain and $\hat{\mu_{t+1}} = A \mu_t + Bu_t$ is the projected mean. 

The first few steps rearrange the loss function into a convenient form and then derive an expression for $\mu_{t+1}$ directly. Step 22 applies the Woodbury matrix inversion lemma to the $[C^TRC + \Sigma_x]^{-1}$ term. The next step rewrites the formula in terms of the Kalman gain matrix K and  multiplies it through. The other major manipulation is the multiplication of the last term of equation 23 by $KK^{-1}$ which is valid since $KK^{-1} = I$.

This derives the optimal posterior mean as the analytical solution to the optimization problem. Deriving the optimal covariance is straightforward and done as follows,

\begin{align}
    \E[\mu_{t+1} \mu_{t+1}^T] &= \E[(\hat{mu_{t+1}} + Ky - KC\hat{mu_{t+1}})(\hat{\mu_{t+1}} + Ky - KC\hat{\mu_{t+1}})^T] \\
    &= \E[\hat{\mu_{t+1}}\hat{\mu_{t+1}}^T] - \E[\hat{\mu_{t+1}}\hat{\mu_{t+1}}^T]C^T K^T - K C \E[\hat{\mu_{t+1}}\hat{\mu_{t+1}}^T] \\ &+ K \E[y y^T]K^T + K C \E[\hat{\mu_{t+1}}\hat{\mu_{t+1}}^T]C^T K^T \\
    &= \Sigma_{\hat{\mu_{t+1}}} - \Sigma_{\hat{\mu_{t+1}}} C^T K^T - KC \Sigma_{\hat{\mu_{t+1}}} + K[R + C\Sigma_{\hat{\mu_{t+1}}}C^T]K^T \\
   &=  \Sigma_{\hat{\mu_{t+1}}} - \Sigma_{\hat{\mu_{t+1}}} C^T K^T - KC \Sigma_{\hat{\mu_{t+1}}} + \Sigma_{\hat{\mu_{t+1}}}C^T[C\Sigma_{\hat{\mu_{t+1}}}C^T +R]^{-1}[R + C\Sigma_{\hat{\mu_{t+1}}}C^T]K^T \\
   &= \Sigma_{\hat{\mu_{t+1}}} - \Sigma_{\hat{\mu_{t+1}}} C^T K^T - KC \Sigma_{\hat{\mu_{t+1}}} +  \Sigma_{\hat{\mu_{t+1}}} C^T K^T \\
   &= \Sigma_{\hat{\mu_{t+1}}} - KC \Sigma_{\hat{\mu_{t+1}}} \\
   &= [I - KC]\Sigma_{\hat{\mu_{t+1}}}
\end{align}

Which is the Kalman update equation for the optimal variance. The second line follows on the assumption that $\E[x y^T] = 0$. On equation 32 the definition of the Kalman gain is substituted back in and the two $C\Sigma_x C^T + R$ terms cancel.

%% file: appendix_B.tex
\chapter{Appendix B: Equations of the LSTM cell}

The equations that specify the computation graph of the LSTM cell are as follows.
\begin{align*}
    v_1 &= h_t \oplus x_t \\
    v_2 &= \sigma(\theta_i v_1) \\ 
    v_3 &= c_t v_2 \\ 
    v_4 &= \sigma (\theta_{inp} v_1) \\ 
    v_5 &= \tanh(\theta_c v_1) \\ 
    v_6 &= v_4 v_5 \\ 
    v_7 &= v_3 + v_6 \\ 
    v_8 &= \sigma (\theta_o v_1)  \\ 
    v_9 &= \tanh(v_7)\\
    v_{10} &= v_8 v_9 \\
    y &= \sigma(\theta_y v_{10})
\end{align*}

The recipe to convert this computation graph into a predictive coding algorithm is straightforward. We first rewire the connectivity so that the predictions are set to the forward functions of their parents. We then compute the errors between the vertices and the predictions. 
\begin{align*}
    \hat{v}_1 &= h_t \oplus x_t \\
    \hat{v}_2 &= \sigma(\theta_i v_1) \\ 
    \hat{v}_3 &= c_t v_2 \\ 
    \hat{v}_4 &= \sigma (\theta_{inp} v_1) \\ 
    \hat{v}_5 &= \tanh(\theta_c v_1) \\ 
    \hat{v}_6 &= v_4 v_5 \\ 
    \hat{v}_7 &= v_3 + v_6 \\ 
    \hat{v} &= \sigma (\theta_o v_1)  \\ 
    \hat{v}_9 &= \tanh(v_7)\\
    \hat{v}_{10} &= v_8 v_9 \\
    \hat{v}_y &= \sigma(\theta_y v_{10}) \\
    \epsilon_1 &= v_1 - \hat{v}_1 \\
    \epsilon_2 &= v_2 - \hat{v}_2 \\
    \epsilon_3 &= v_3 - \hat{v}_3 \\
    \epsilon_4 &= v_4 - \hat{v}_4 \\
    \epsilon_5 &= v_5 - \hat{v}_5 \\
    \epsilon_6 &= v_6 - \hat{v}_6 \\
    \epsilon_7 &= v_7 - \hat{v}_7 \\
    \epsilon_8 &= v_8 - \hat{v}_8 \\
    \epsilon_9 &= v_9 - \hat{v}_9 \\
    \epsilon_{10} &= v_{10} - \hat{v}_{10}
\end{align*}

%% file: Appendix_C.tex
\chapter{Appendix C: Predictive Coding Under the Laplace Approximation}

In the main derivation in Chapter 3 of the variational free energy $\mathcal{F}$, we used the assumption that the variational density is a dirac delta function: $q(x | o; \mu) = \delta(x - \mu)$. However, the majority of derivations, including the original derivations in \citep{friston2005theory} instead applied the Laplace approximation to the variational distribution $q$. This approximation defines $q$ to be a Gaussian distribution with a variance which is a function of the mean $\mu$:  $q(x | o; \mu) = \mathcal{N}(x;\mu, \sigma(\mu))$. Notationally, it is important to distinguish between the generative model $\Sigma$, and the variational distribution $\sigma$. Here we use the lower-case $\sigma$ to denote the parameter of the variational distribution. The lower-case is not meant to imply it is necessarily scalar. As we shall see, the optimal $\sigma$ will be become the inverse-Hessian of the free energy at the mode. 

Intuitively, this is because the curvature at the mode of a Gaussian distribution gives a good indication of the variance of the Gaussian, since a Gaussian with high curvature at the mode (i.e. the mean) will be highly peaked and thus have a small variance, while a Gaussian with low curvature will be broad, and thus have a large variance.  While our derivation using a dirac-delta approximation and the standard derivation using a Laplace approximation obviously differ, they ultimately arrive at the same expression for the variational free energy $\mathcal{F}$. This is because both approximations effectively remove the variational variances from consideration and only use the variational mean in practice. Under the Laplace approximation, the variational coveriance has an analytical optimal form and thus does not need to be optimized, and plays no real role in the optimization process for the $\mu$s either. We chose to present our derivation using dirac-deltas in the interests of simplicity, since as we shall see the Laplace approximation derivation is somewhat more involved.

To begin, we return to the standard energy function in multilayer case, this time under the assuption the Laplace approximation.

\begin{align*}
    \mathcal{F} &= \sum_{i=}^L \underbrace{\mathbb{E}_{q(x | o;\mu)}\big[ \ln p( x_i | x_{i+1}) \big]}_{\text{Energy}} + \underbrace{\mathbb{E}_{q(x | o;\mu)} \big[ \ln q(x | o ;\mu) \big]}_{\text{Entropy}} \\
    &= \sum_{i=}^L \mathbb{E}_{\mathcal{N}(x;\mu, \sigma(\mu))}\big[ \ln \mathcal{N}(x_i;f(\mu_{i+1}, \Sigma(\mu))) \big] - \ln 2 \pi \sigma_i \numberthis
\end{align*}
Where we have used the analytical result that the entropy of a gaussian distribution $\mathbb{H}[\mathcal{N}] = \ln 2 \pi \sigma$.
Then, we apply a Taylor expansion around $x_i = \mu_i$ to each element in the sum,
\begin{align*}
    \mathcal{F} &\propto \sum_{i=}^L \mathbb{E}_q \big[ \ln p(\mu_i | \mu_{i+1}) \big] + \mathbb{E}_q \big[\frac{\partial \ln p(x_i | x{i+1})}{\partial x_i}(x_i - \mu_i) \big] + \mathbb{E}_q \big[\frac{\partial^2 \ln p(x_i | x{i+1})}{\partial x_i^2}(x_i - \mu_i)^2 \big]  - \ln 2 \pi \sigma_i\\
    &= \sum_{i=0}^L \ln p(\mu_i | \mu_{i+1})  +  \frac{\partial^2 \ln p(x_i | x{i+1})}{\partial x_i^2}\sigma_i - \ln 2 \pi \sigma_i \numberthis
\end{align*}
Where in the second line, we have used the fact that $\mathbb{E}_q \big[x_i - \mu_i \big] = \mathbb{E}_q \big[x_i] - \mu_i = \mu_i - \mu_i = 0$ and that $\mathbb{E}_q \big[ (x_i - \mu_i)^2] = \Sigma_i$, which is that the expected squared residual simply is the variance. We also drop the expectation around the first term, since as a function only of $\mu$ and $\mu_{i+1}$, it is no longer a function of $x_i$ which is the variable the expectation is under. We can then differentiate this expression with respect to $\sigma_i$ and solve for 0 to obtain the optimal variance.
\begin{align*}
    \frac{\partial \mathcal{F}}{\partial \sigma_i} &= \frac{\partial^2 \ln p(x_i | x{i+1})}{\partial x_i^2} - \sigma_i^{-1} \\
    & \frac{\partial \mathcal{F}}{\partial \sigma_i} = 0 \implies \sigma_i = {\frac{\partial^2 \ln p(x_i | x{i+1})}{\partial x_i^2}}^{-1} \numberthis
\end{align*}
Given this analytical result, there is no point optimizing $\mathcal{F}$ with respect to the variational variances $\sigma_i$, so our objective simply becomes,
\begin{align*}
    \mathcal{F} = \sum_{i=0}^L \ln p(\mu_i | \mu_{i+1}) \numberthis
\end{align*}
which is exactly the same result as obtained through the dirac delta approximation.

%% file: main.bbl
\begin{thebibliography}{}

\bibitem [\protect \citeauthoryear {%
Abdolmaleki%
\ \protect \BOthers {.}}{%
Abdolmaleki%
\ \protect \BOthers {.}}{%
{\protect \APACyear {2018}}%
}]{%
abdolmaleki2018maximum}
\APACinsertmetastar {%
abdolmaleki2018maximum}%
\begin{APACrefauthors}%
Abdolmaleki, A.%
, Springenberg, J\BPBI T.%
, Tassa, Y.%
, Munos, R.%
, Heess, N.%
\BCBL {}\ \BBA {} Riedmiller, M.%
\end{APACrefauthors}%
\unskip\
\newblock
\APACrefYearMonthDay{2018}{}{}.
\newblock
{\BBOQ}\APACrefatitle {Maximum a posteriori policy optimisation} {Maximum a
  posteriori policy optimisation}.{\BBCQ}
\newblock
\APACjournalVolNumPages{arXiv preprint arXiv:1806.06920}{}{}{}.
\PrintBackRefs{\CurrentBib}

\bibitem [\protect \citeauthoryear {%
Adams%
, Perrinet%
\BCBL {}\ \BBA {} Friston%
}{%
Adams%
\ \protect \BOthers {.}}{%
{\protect \APACyear {2012}}%
}]{%
adams2012smooth}
\APACinsertmetastar {%
adams2012smooth}%
\begin{APACrefauthors}%
Adams, R\BPBI A.%
, Perrinet, L\BPBI U.%
\BCBL {}\ \BBA {} Friston, K.%
\end{APACrefauthors}%
\unskip\
\newblock
\APACrefYearMonthDay{2012}{}{}.
\newblock
{\BBOQ}\APACrefatitle {Smooth pursuit and visual occlusion: active inference
  and oculomotor control in schizophrenia} {Smooth pursuit and visual
  occlusion: active inference and oculomotor control in schizophrenia}.{\BBCQ}
\newblock
\APACjournalVolNumPages{PloS one}{7}{10}{e47502}.
\PrintBackRefs{\CurrentBib}

\bibitem [\protect \citeauthoryear {%
Aitchison%
\ \BBA {} Lengyel%
}{%
Aitchison%
\ \BBA {} Lengyel%
}{%
{\protect \APACyear {2017}}%
}]{%
aitchison2017or}
\APACinsertmetastar {%
aitchison2017or}%
\begin{APACrefauthors}%
Aitchison, L.%
\BCBT {}\ \BBA {} Lengyel, M.%
\end{APACrefauthors}%
\unskip\
\newblock
\APACrefYearMonthDay{2017}{}{}.
\newblock
{\BBOQ}\APACrefatitle {With or without you: predictive coding and Bayesian
  inference in the brain} {With or without you: predictive coding and bayesian
  inference in the brain}.{\BBCQ}
\newblock
\APACjournalVolNumPages{Current Opinion in Neurobiology}{46}{}{219--227}.
\PrintBackRefs{\CurrentBib}

\bibitem [\protect \citeauthoryear {%
Akrout%
, Wilson%
, Humphreys%
, Lillicrap%
\BCBL {}\ \BBA {} Tweed%
}{%
Akrout%
\ \protect \BOthers {.}}{%
{\protect \APACyear {2019}}%
}]{%
akrout2019deep}
\APACinsertmetastar {%
akrout2019deep}%
\begin{APACrefauthors}%
Akrout, M.%
, Wilson, C.%
, Humphreys, P.%
, Lillicrap, T.%
\BCBL {}\ \BBA {} Tweed, D\BPBI B.%
\end{APACrefauthors}%
\unskip\
\newblock
\APACrefYearMonthDay{2019}{}{}.
\newblock
{\BBOQ}\APACrefatitle {Deep learning without weight transport} {Deep learning
  without weight transport}.{\BBCQ}
\newblock
\BIn{} \APACrefbtitle {Advances in Neural Information Processing Systems}
  {Advances in neural information processing systems}\ (\BPGS\ 974--982).
\PrintBackRefs{\CurrentBib}

\bibitem [\protect \citeauthoryear {%
Amari%
}{%
Amari%
}{%
{\protect \APACyear {1995}}%
}]{%
amari1995information}
\APACinsertmetastar {%
amari1995information}%
\begin{APACrefauthors}%
Amari, S\BHBI I.%
\end{APACrefauthors}%
\unskip\
\newblock
\APACrefYearMonthDay{1995}{}{}.
\newblock
{\BBOQ}\APACrefatitle {Information geometry of the EM and em algorithms for
  neural networks} {Information geometry of the em and em algorithms for neural
  networks}.{\BBCQ}
\newblock
\APACjournalVolNumPages{Neural networks}{8}{9}{1379--1408}.
\PrintBackRefs{\CurrentBib}

\bibitem [\protect \citeauthoryear {%
Amit%
}{%
Amit%
}{%
{\protect \APACyear {2019}}%
}]{%
amit2019deep}
\APACinsertmetastar {%
amit2019deep}%
\begin{APACrefauthors}%
Amit, Y.%
\end{APACrefauthors}%
\unskip\
\newblock
\APACrefYearMonthDay{2019}{}{}.
\newblock
{\BBOQ}\APACrefatitle {Deep learning with asymmetric connections and Hebbian
  updates} {Deep learning with asymmetric connections and hebbian
  updates}.{\BBCQ}
\newblock
\APACjournalVolNumPages{Frontiers in computational neuroscience}{13}{}{18}.
\PrintBackRefs{\CurrentBib}

\bibitem [\protect \citeauthoryear {%
Andrews%
}{%
Andrews%
}{%
{\protect \APACyear {2020}}%
}]{%
andrews2020math}
\APACinsertmetastar {%
andrews2020math}%
\begin{APACrefauthors}%
Andrews, M.%
\end{APACrefauthors}%
\unskip\
\newblock
\APACrefYearMonthDay{2020}{}{}.
\newblock
{\BBOQ}\APACrefatitle {The Math is not the Territory: Navigating the Free
  Energy Principle} {The math is not the territory: Navigating the free energy
  principle}.{\BBCQ}
\newblock

\PrintBackRefs{\CurrentBib}

\bibitem [\protect \citeauthoryear {%
Arulampalam%
, Maskell%
, Gordon%
\BCBL {}\ \BBA {} Clapp%
}{%
Arulampalam%
\ \protect \BOthers {.}}{%
{\protect \APACyear {2002}}%
}]{%
arulampalam2002tutorial}
\APACinsertmetastar {%
arulampalam2002tutorial}%
\begin{APACrefauthors}%
Arulampalam, M\BPBI S.%
, Maskell, S.%
, Gordon, N.%
\BCBL {}\ \BBA {} Clapp, T.%
\end{APACrefauthors}%
\unskip\
\newblock
\APACrefYearMonthDay{2002}{}{}.
\newblock
{\BBOQ}\APACrefatitle {A tutorial on particle filters for online
  nonlinear/non-Gaussian Bayesian tracking} {A tutorial on particle filters for
  online nonlinear/non-gaussian bayesian tracking}.{\BBCQ}
\newblock
\APACjournalVolNumPages{IEEE Transactions on Signal
  Processing}{50}{2}{174--188}.
\PrintBackRefs{\CurrentBib}

\bibitem [\protect \citeauthoryear {%
Attias%
}{%
Attias%
}{%
{\protect \APACyear {2003}}%
}]{%
attias2003planning}
\APACinsertmetastar {%
attias2003planning}%
\begin{APACrefauthors}%
Attias, H.%
\end{APACrefauthors}%
\unskip\
\newblock
\APACrefYearMonthDay{2003}{}{}.
\newblock
{\BBOQ}\APACrefatitle {Planning by Probabilistic Inference.} {Planning by
  probabilistic inference.}{\BBCQ}
\newblock
\BIn{} \APACrefbtitle {AISTATS.} {Aistats.}
\PrintBackRefs{\CurrentBib}

\bibitem [\protect \citeauthoryear {%
Auksztulewicz%
\ \BBA {} Friston%
}{%
Auksztulewicz%
\ \BBA {} Friston%
}{%
{\protect \APACyear {2016}}%
}]{%
auksztulewicz2016repetition}
\APACinsertmetastar {%
auksztulewicz2016repetition}%
\begin{APACrefauthors}%
Auksztulewicz, R.%
\BCBT {}\ \BBA {} Friston, K.%
\end{APACrefauthors}%
\unskip\
\newblock
\APACrefYearMonthDay{2016}{}{}.
\newblock
{\BBOQ}\APACrefatitle {Repetition suppression and its contextual determinants
  in predictive coding} {Repetition suppression and its contextual determinants
  in predictive coding}.{\BBCQ}
\newblock
\APACjournalVolNumPages{cortex}{80}{}{125--140}.
\PrintBackRefs{\CurrentBib}

\bibitem [\protect \citeauthoryear {%
Baird%
}{%
Baird%
}{%
{\protect \APACyear {1995}}%
}]{%
baird1995residual}
\APACinsertmetastar {%
baird1995residual}%
\begin{APACrefauthors}%
Baird, L.%
\end{APACrefauthors}%
\unskip\
\newblock
\APACrefYearMonthDay{1995}{}{}.
\newblock
{\BBOQ}\APACrefatitle {Residual algorithms: Reinforcement learning with
  function approximation} {Residual algorithms: Reinforcement learning with
  function approximation}.{\BBCQ}
\newblock
\BIn{} \APACrefbtitle {Machine Learning Proceedings 1995} {Machine learning
  proceedings 1995}\ (\BPGS\ 30--37).
\newblock
\APACaddressPublisher{}{Elsevier}.
\PrintBackRefs{\CurrentBib}

\bibitem [\protect \citeauthoryear {%
Baldi%
\ \BBA {} Sadowski%
}{%
Baldi%
\ \BBA {} Sadowski%
}{%
{\protect \APACyear {2016}}%
}]{%
baldi2016theory}
\APACinsertmetastar {%
baldi2016theory}%
\begin{APACrefauthors}%
Baldi, P.%
\BCBT {}\ \BBA {} Sadowski, P.%
\end{APACrefauthors}%
\unskip\
\newblock
\APACrefYearMonthDay{2016}{}{}.
\newblock
{\BBOQ}\APACrefatitle {A theory of local learning, the learning channel, and
  the optimality of backpropagation} {A theory of local learning, the learning
  channel, and the optimality of backpropagation}.{\BBCQ}
\newblock
\APACjournalVolNumPages{Neural Networks}{83}{}{51--74}.
\PrintBackRefs{\CurrentBib}

\bibitem [\protect \citeauthoryear {%
Baltieri%
\ \BBA {} Buckley%
}{%
Baltieri%
\ \BBA {} Buckley%
}{%
{\protect \APACyear {2017}}%
}]{%
baltieri2017active}
\APACinsertmetastar {%
baltieri2017active}%
\begin{APACrefauthors}%
Baltieri, M.%
\BCBT {}\ \BBA {} Buckley, C\BPBI L.%
\end{APACrefauthors}%
\unskip\
\newblock
\APACrefYearMonthDay{2017}{}{}.
\newblock
{\BBOQ}\APACrefatitle {An active inference implementation of phototaxis} {An
  active inference implementation of phototaxis}.{\BBCQ}
\newblock
\BIn{} \APACrefbtitle {Artificial Life Conference Proceedings 14} {Artificial
  life conference proceedings 14}\ (\BPGS\ 36--43).
\PrintBackRefs{\CurrentBib}

\bibitem [\protect \citeauthoryear {%
Baltieri%
\ \BBA {} Buckley%
}{%
Baltieri%
\ \BBA {} Buckley%
}{%
{\protect \APACyear {2018}}%
}]{%
baltieri2018modularity}
\APACinsertmetastar {%
baltieri2018modularity}%
\begin{APACrefauthors}%
Baltieri, M.%
\BCBT {}\ \BBA {} Buckley, C\BPBI L.%
\end{APACrefauthors}%
\unskip\
\newblock
\APACrefYearMonthDay{2018}{}{}.
\newblock
{\BBOQ}\APACrefatitle {The modularity of action and perception revisited using
  control theory and active inference} {The modularity of action and perception
  revisited using control theory and active inference}.{\BBCQ}
\newblock
\BIn{} \APACrefbtitle {Artificial Life Conference Proceedings} {Artificial life
  conference proceedings}\ (\BPGS\ 121--128).
\PrintBackRefs{\CurrentBib}

\bibitem [\protect \citeauthoryear {%
Baltieri%
\ \BBA {} Buckley%
}{%
Baltieri%
\ \BBA {} Buckley%
}{%
{\protect \APACyear {2019}}%
}]{%
baltieri2019pid}
\APACinsertmetastar {%
baltieri2019pid}%
\begin{APACrefauthors}%
Baltieri, M.%
\BCBT {}\ \BBA {} Buckley, C\BPBI L.%
\end{APACrefauthors}%
\unskip\
\newblock
\APACrefYearMonthDay{2019}{}{}.
\newblock
{\BBOQ}\APACrefatitle {PID control as a process of active inference with linear
  generative models} {Pid control as a process of active inference with linear
  generative models}.{\BBCQ}
\newblock
\APACjournalVolNumPages{Entropy}{21}{3}{257}.
\newblock
\begin{APACrefURL} \url{https://www.mdpi.com/1099-4300/21/3/257}
  \end{APACrefURL}
\PrintBackRefs{\CurrentBib}

\bibitem [\protect \citeauthoryear {%
Baltieri%
\ \BBA {} Buckley%
}{%
Baltieri%
\ \BBA {} Buckley%
}{%
{\protect \APACyear {2020}}%
}]{%
baltieri2020Kalman}
\APACinsertmetastar {%
baltieri2020Kalman}%
\begin{APACrefauthors}%
Baltieri, M.%
\BCBT {}\ \BBA {} Buckley, C\BPBI L.%
\end{APACrefauthors}%
\unskip\
\newblock
\APACrefYearMonthDay{2020}{}{}.
\newblock
{\BBOQ}\APACrefatitle {On Kalman-Bucy filters, linear quadratic control and
  active inference} {On kalman-bucy filters, linear quadratic control and
  active inference}.{\BBCQ}
\newblock
\APACjournalVolNumPages{arXiv preprint arXiv:2005.06269}{}{}{}.
\newblock
\begin{APACrefURL} \url{https://arxiv.org/abs/2005.06269} \end{APACrefURL}
\PrintBackRefs{\CurrentBib}

\bibitem [\protect \citeauthoryear {%
Baltieri%
, Buckley%
\BCBL {}\ \BBA {} Bruineberg%
}{%
Baltieri%
\ \protect \BOthers {.}}{%
{\protect \APACyear {2020}}%
}]{%
baltieri2020predictions}
\APACinsertmetastar {%
baltieri2020predictions}%
\begin{APACrefauthors}%
Baltieri, M.%
, Buckley, C\BPBI L.%
\BCBL {}\ \BBA {} Bruineberg, J.%
\end{APACrefauthors}%
\unskip\
\newblock
\APACrefYearMonthDay{2020}{}{}.
\newblock
{\BBOQ}\APACrefatitle {Predictions in the eye of the beholder: an active
  inference account of Watt governors} {Predictions in the eye of the beholder:
  an active inference account of watt governors}.{\BBCQ}
\newblock
\BIn{} \APACrefbtitle {Artificial Life Conference Proceedings} {Artificial life
  conference proceedings}\ (\BPGS\ 121--129).
\newblock
\begin{APACrefURL}
  \url{https://www.mitpressjournals.org/doi/abs/10.1162/isal_a_00288}
  \end{APACrefURL}
\PrintBackRefs{\CurrentBib}

\bibitem [\protect \citeauthoryear {%
Barlow%
\ \protect \BOthers {.}}{%
Barlow%
\ \protect \BOthers {.}}{%
{\protect \APACyear {1961}}%
}]{%
barlow1961possible}
\APACinsertmetastar {%
barlow1961possible}%
\begin{APACrefauthors}%
Barlow, H\BPBI B.%
\BCBT {}\ \BOthersPeriod {.}
\end{APACrefauthors}%
\unskip\
\newblock
\APACrefYearMonthDay{1961}{}{}.
\newblock
{\BBOQ}\APACrefatitle {Possible principles underlying the transformation of
  sensory messages} {Possible principles underlying the transformation of
  sensory messages}.{\BBCQ}
\newblock
\APACjournalVolNumPages{Sensory communication}{1}{}{217--234}.
\PrintBackRefs{\CurrentBib}

\bibitem [\protect \citeauthoryear {%
Bartunov%
\ \protect \BOthers {.}}{%
Bartunov%
\ \protect \BOthers {.}}{%
{\protect \APACyear {2018}}%
}]{%
bartunov2018assessing}
\APACinsertmetastar {%
bartunov2018assessing}%
\begin{APACrefauthors}%
Bartunov, S.%
, Santoro, A.%
, Richards, B.%
, Marris, L.%
, Hinton, G.%
\BCBL {}\ \BBA {} Lillicrap, T.%
\end{APACrefauthors}%
\unskip\
\newblock
\APACrefYearMonthDay{2018}{}{}.
\newblock
{\BBOQ}\APACrefatitle {Assessing the scalability of biologically-motivated deep
  learning algorithms and architectures} {Assessing the scalability of
  biologically-motivated deep learning algorithms and architectures}.{\BBCQ}
\newblock
\BIn{} \APACrefbtitle {Advances in Neural Information Processing Systems}
  {Advances in neural information processing systems}\ (\BPGS\ 9368--9378).
\PrintBackRefs{\CurrentBib}

\bibitem [\protect \citeauthoryear {%
Bastos%
, Lundqvist%
, Waite%
, Kopell%
\BCBL {}\ \BBA {} Miller%
}{%
Bastos%
\ \protect \BOthers {.}}{%
{\protect \APACyear {2020}}%
}]{%
bastos2020layer}
\APACinsertmetastar {%
bastos2020layer}%
\begin{APACrefauthors}%
Bastos, A\BPBI M.%
, Lundqvist, M.%
, Waite, A\BPBI S.%
, Kopell, N.%
\BCBL {}\ \BBA {} Miller, E\BPBI K.%
\end{APACrefauthors}%
\unskip\
\newblock
\APACrefYearMonthDay{2020}{}{}.
\newblock
{\BBOQ}\APACrefatitle {Layer and rhythm specificity for predictive routing}
  {Layer and rhythm specificity for predictive routing}.{\BBCQ}
\newblock
\APACjournalVolNumPages{Proceedings of the National Academy of
  Sciences}{117}{49}{31459--31469}.
\PrintBackRefs{\CurrentBib}

\bibitem [\protect \citeauthoryear {%
Bastos%
\ \protect \BOthers {.}}{%
Bastos%
\ \protect \BOthers {.}}{%
{\protect \APACyear {2012}}%
}]{%
bastos2012canonical}
\APACinsertmetastar {%
bastos2012canonical}%
\begin{APACrefauthors}%
Bastos, A\BPBI M.%
, Usrey, W\BPBI M.%
, Adams, R\BPBI A.%
, Mangun, G\BPBI R.%
, Fries, P.%
\BCBL {}\ \BBA {} Friston, K.%
\end{APACrefauthors}%
\unskip\
\newblock
\APACrefYearMonthDay{2012}{}{}.
\newblock
{\BBOQ}\APACrefatitle {Canonical microcircuits for predictive coding}
  {Canonical microcircuits for predictive coding}.{\BBCQ}
\newblock
\APACjournalVolNumPages{Neuron}{76}{4}{695--711}.
\PrintBackRefs{\CurrentBib}

\bibitem [\protect \citeauthoryear {%
Bastos%
\ \protect \BOthers {.}}{%
Bastos%
\ \protect \BOthers {.}}{%
{\protect \APACyear {2015}}%
}]{%
bastos2015visual}
\APACinsertmetastar {%
bastos2015visual}%
\begin{APACrefauthors}%
Bastos, A\BPBI M.%
, Vezoli, J.%
, Bosman, C\BPBI A.%
, Schoffelen, J\BHBI M.%
, Oostenveld, R.%
, Dowdall, J\BPBI R.%
\BDBL {}Fries, P.%
\end{APACrefauthors}%
\unskip\
\newblock
\APACrefYearMonthDay{2015}{}{}.
\newblock
{\BBOQ}\APACrefatitle {Visual areas exert feedforward and feedback influences
  through distinct frequency channels} {Visual areas exert feedforward and
  feedback influences through distinct frequency channels}.{\BBCQ}
\newblock
\APACjournalVolNumPages{Neuron}{85}{2}{390--401}.
\PrintBackRefs{\CurrentBib}

\bibitem [\protect \citeauthoryear {%
Baydin%
, Pearlmutter%
, Radul%
\BCBL {}\ \BBA {} Siskind%
}{%
Baydin%
\ \protect \BOthers {.}}{%
{\protect \APACyear {2017}}%
}]{%
baydin2017automatic}
\APACinsertmetastar {%
baydin2017automatic}%
\begin{APACrefauthors}%
Baydin, A\BPBI G.%
, Pearlmutter, B\BPBI A.%
, Radul, A\BPBI A.%
\BCBL {}\ \BBA {} Siskind, J\BPBI M.%
\end{APACrefauthors}%
\unskip\
\newblock
\APACrefYearMonthDay{2017}{}{}.
\newblock
{\BBOQ}\APACrefatitle {Automatic differentiation in machine learning: a survey}
  {Automatic differentiation in machine learning: a survey}.{\BBCQ}
\newblock
\APACjournalVolNumPages{The Journal of Machine Learning
  Research}{18}{1}{5595--5637}.
\PrintBackRefs{\CurrentBib}

\bibitem [\protect \citeauthoryear {%
Beal%
}{%
Beal%
}{%
{\protect \APACyear {2003}}%
}]{%
beal2003variational}
\APACinsertmetastar {%
beal2003variational}%
\begin{APACrefauthors}%
Beal, M\BPBI J.%
\end{APACrefauthors}%
\unskip\
\newblock
\APACrefYear{2003}.
\unskip\
\newblock
\APACrefbtitle {Variational algorithms for approximate Bayesian inference}
  {Variational algorithms for approximate bayesian inference}\
  \APACtypeAddressSchool {\BUPhD}{}{}.
\unskip\
\newblock
\APACaddressSchool {}{UCL (University College London)}.
\PrintBackRefs{\CurrentBib}

\bibitem [\protect \citeauthoryear {%
Bear%
, Connors%
\BCBL {}\ \BBA {} Paradiso%
}{%
Bear%
\ \protect \BOthers {.}}{%
{\protect \APACyear {2020}}%
}]{%
bear2020neuroscience}
\APACinsertmetastar {%
bear2020neuroscience}%
\begin{APACrefauthors}%
Bear, M.%
, Connors, B.%
\BCBL {}\ \BBA {} Paradiso, M\BPBI A.%
\end{APACrefauthors}%
\unskip\
\newblock
\APACrefYear{2020}.
\newblock
\APACrefbtitle {Neuroscience: Exploring the brain} {Neuroscience: Exploring the
  brain}.
\newblock
\APACaddressPublisher{}{Jones \& Bartlett Learning, LLC}.
\PrintBackRefs{\CurrentBib}

\bibitem [\protect \citeauthoryear {%
Bellec%
\ \protect \BOthers {.}}{%
Bellec%
\ \protect \BOthers {.}}{%
{\protect \APACyear {2020}}%
}]{%
bellec2020solution}
\APACinsertmetastar {%
bellec2020solution}%
\begin{APACrefauthors}%
Bellec, G.%
, Scherr, F.%
, Subramoney, A.%
, Hajek, E.%
, Salaj, D.%
, Legenstein, R.%
\BCBL {}\ \BBA {} Maass, W.%
\end{APACrefauthors}%
\unskip\
\newblock
\APACrefYearMonthDay{2020}{}{}.
\newblock
{\BBOQ}\APACrefatitle {A solution to the learning dilemma for recurrent
  networks of spiking neurons} {A solution to the learning dilemma for
  recurrent networks of spiking neurons}.{\BBCQ}
\newblock
\APACjournalVolNumPages{bioRxiv}{}{}{738385}.
\PrintBackRefs{\CurrentBib}

\bibitem [\protect \citeauthoryear {%
Bellman%
}{%
Bellman%
}{%
{\protect \APACyear {1952}}%
}]{%
bellman1952theory}
\APACinsertmetastar {%
bellman1952theory}%
\begin{APACrefauthors}%
Bellman, R.%
\end{APACrefauthors}%
\unskip\
\newblock
\APACrefYearMonthDay{1952}{}{}.
\newblock
{\BBOQ}\APACrefatitle {On the theory of dynamic programming} {On the theory of
  dynamic programming}.{\BBCQ}
\newblock
\APACjournalVolNumPages{Proceedings of the National Academy of Sciences of the
  United States of America}{38}{8}{716}.
\PrintBackRefs{\CurrentBib}

\bibitem [\protect \citeauthoryear {%
Bengio%
}{%
Bengio%
}{%
{\protect \APACyear {2020}}%
}]{%
bengio2020deriving}
\APACinsertmetastar {%
bengio2020deriving}%
\begin{APACrefauthors}%
Bengio, Y.%
\end{APACrefauthors}%
\unskip\
\newblock
\APACrefYearMonthDay{2020}{}{}.
\newblock
{\BBOQ}\APACrefatitle {Deriving differential target propagation from iterating
  approximate inverses} {Deriving differential target propagation from
  iterating approximate inverses}.{\BBCQ}
\newblock
\APACjournalVolNumPages{arXiv preprint arXiv:2007.15139}{}{}{}.
\PrintBackRefs{\CurrentBib}

\bibitem [\protect \citeauthoryear {%
Bengio%
\ \BBA {} Fischer%
}{%
Bengio%
\ \BBA {} Fischer%
}{%
{\protect \APACyear {2015}}%
}]{%
bengio2015early}
\APACinsertmetastar {%
bengio2015early}%
\begin{APACrefauthors}%
Bengio, Y.%
\BCBT {}\ \BBA {} Fischer, A.%
\end{APACrefauthors}%
\unskip\
\newblock
\APACrefYearMonthDay{2015}{}{}.
\newblock
{\BBOQ}\APACrefatitle {Early inference in energy-based models approximates
  back-propagation} {Early inference in energy-based models approximates
  back-propagation}.{\BBCQ}
\newblock
\APACjournalVolNumPages{arXiv preprint arXiv:1510.02777}{}{}{}.
\PrintBackRefs{\CurrentBib}

\bibitem [\protect \citeauthoryear {%
Bengio%
, Mesnard%
, Fischer%
, Zhang%
\BCBL {}\ \BBA {} Wu%
}{%
Bengio%
\ \protect \BOthers {.}}{%
{\protect \APACyear {2017}}%
}]{%
bengio2017stdp}
\APACinsertmetastar {%
bengio2017stdp}%
\begin{APACrefauthors}%
Bengio, Y.%
, Mesnard, T.%
, Fischer, A.%
, Zhang, S.%
\BCBL {}\ \BBA {} Wu, Y.%
\end{APACrefauthors}%
\unskip\
\newblock
\APACrefYearMonthDay{2017}{}{}.
\newblock
{\BBOQ}\APACrefatitle {STDP-compatible approximation of backpropagation in an
  energy-based model} {Stdp-compatible approximation of backpropagation in an
  energy-based model}.{\BBCQ}
\newblock
\APACjournalVolNumPages{Neural Computation}{29}{3}{555--577}.
\PrintBackRefs{\CurrentBib}

\bibitem [\protect \citeauthoryear {%
Berger-Tal%
, Nathan%
, Meron%
\BCBL {}\ \BBA {} Saltz%
}{%
Berger-Tal%
\ \protect \BOthers {.}}{%
{\protect \APACyear {2014}}%
}]{%
berger2014exploration}
\APACinsertmetastar {%
berger2014exploration}%
\begin{APACrefauthors}%
Berger-Tal, O.%
, Nathan, J.%
, Meron, E.%
\BCBL {}\ \BBA {} Saltz, D.%
\end{APACrefauthors}%
\unskip\
\newblock
\APACrefYearMonthDay{2014}{}{}.
\newblock
{\BBOQ}\APACrefatitle {The exploration-exploitation dilemma: a
  multidisciplinary framework} {The exploration-exploitation dilemma: a
  multidisciplinary framework}.{\BBCQ}
\newblock
\APACjournalVolNumPages{PloS One}{9}{4}{e95693}.
\PrintBackRefs{\CurrentBib}

\bibitem [\protect \citeauthoryear {%
M.~Betancourt%
}{%
M.~Betancourt%
}{%
{\protect \APACyear {2017}}%
}]{%
betancourt2017conceptual}
\APACinsertmetastar {%
betancourt2017conceptual}%
\begin{APACrefauthors}%
Betancourt, M.%
\end{APACrefauthors}%
\unskip\
\newblock
\APACrefYearMonthDay{2017}{}{}.
\newblock
{\BBOQ}\APACrefatitle {A conceptual introduction to Hamiltonian Monte Carlo} {A
  conceptual introduction to hamiltonian monte carlo}.{\BBCQ}
\newblock
\APACjournalVolNumPages{arXiv preprint arXiv:1701.02434}{}{}{}.
\PrintBackRefs{\CurrentBib}

\bibitem [\protect \citeauthoryear {%
M\BPBI J.~Betancourt%
}{%
M\BPBI J.~Betancourt%
}{%
{\protect \APACyear {2013}}%
}]{%
betancourt2013generalizing}
\APACinsertmetastar {%
betancourt2013generalizing}%
\begin{APACrefauthors}%
Betancourt, M\BPBI J.%
\end{APACrefauthors}%
\unskip\
\newblock
\APACrefYearMonthDay{2013}{}{}.
\newblock
{\BBOQ}\APACrefatitle {Generalizing the no-U-turn sampler to Riemannian
  manifolds} {Generalizing the no-u-turn sampler to riemannian
  manifolds}.{\BBCQ}
\newblock
\APACjournalVolNumPages{arXiv preprint arXiv:1304.1920}{}{}{}.
\PrintBackRefs{\CurrentBib}

\bibitem [\protect \citeauthoryear {%
Blei%
, Kucukelbir%
\BCBL {}\ \BBA {} McAuliffe%
}{%
Blei%
\ \protect \BOthers {.}}{%
{\protect \APACyear {2017}}%
}]{%
blei2017variational}
\APACinsertmetastar {%
blei2017variational}%
\begin{APACrefauthors}%
Blei, D\BPBI M.%
, Kucukelbir, A.%
\BCBL {}\ \BBA {} McAuliffe, J\BPBI D.%
\end{APACrefauthors}%
\unskip\
\newblock
\APACrefYearMonthDay{2017}{}{}.
\newblock
{\BBOQ}\APACrefatitle {Variational inference: A review for statisticians}
  {Variational inference: A review for statisticians}.{\BBCQ}
\newblock
\APACjournalVolNumPages{Journal of the American statistical
  Association}{112}{518}{859--877}.
\PrintBackRefs{\CurrentBib}

\bibitem [\protect \citeauthoryear {%
Bostrom%
}{%
Bostrom%
}{%
{\protect \APACyear {2017}}%
}]{%
bostrom2017superintelligence}
\APACinsertmetastar {%
bostrom2017superintelligence}%
\begin{APACrefauthors}%
Bostrom, N.%
\end{APACrefauthors}%
\unskip\
\newblock
\APACrefYear{2017}.
\newblock
\APACrefbtitle {Superintelligence} {Superintelligence}.
\newblock
\APACaddressPublisher{}{Dunod}.
\PrintBackRefs{\CurrentBib}

\bibitem [\protect \citeauthoryear {%
Brockman%
\ \protect \BOthers {.}}{%
Brockman%
\ \protect \BOthers {.}}{%
{\protect \APACyear {2016}}%
}]{%
brockman2016openai}
\APACinsertmetastar {%
brockman2016openai}%
\begin{APACrefauthors}%
Brockman, G.%
, Cheung, V.%
, Pettersson, L.%
, Schneider, J.%
, Schulman, J.%
, Tang, J.%
\BCBL {}\ \BBA {} Zaremba, W.%
\end{APACrefauthors}%
\unskip\
\newblock
\APACrefYearMonthDay{2016}{}{}.
\newblock
{\BBOQ}\APACrefatitle {Openai gym} {Openai gym}.{\BBCQ}
\newblock
\APACjournalVolNumPages{arXiv preprint arXiv:1606.01540}{}{}{}.
\PrintBackRefs{\CurrentBib}

\bibitem [\protect \citeauthoryear {%
Brooks%
, Gelman%
, Jones%
\BCBL {}\ \BBA {} Meng%
}{%
Brooks%
\ \protect \BOthers {.}}{%
{\protect \APACyear {2011}}%
}]{%
brooks2011handbook}
\APACinsertmetastar {%
brooks2011handbook}%
\begin{APACrefauthors}%
Brooks, S.%
, Gelman, A.%
, Jones, G.%
\BCBL {}\ \BBA {} Meng, X\BHBI L.%
\end{APACrefauthors}%
\unskip\
\newblock
\APACrefYear{2011}.
\newblock
\APACrefbtitle {Handbook of markov chain monte carlo} {Handbook of markov chain
  monte carlo}.
\newblock
\APACaddressPublisher{}{CRC press}.
\PrintBackRefs{\CurrentBib}

\bibitem [\protect \citeauthoryear {%
Brown%
\ \protect \BOthers {.}}{%
Brown%
\ \protect \BOthers {.}}{%
{\protect \APACyear {2020}}%
}]{%
brown2020language}
\APACinsertmetastar {%
brown2020language}%
\begin{APACrefauthors}%
Brown, T\BPBI B.%
, Mann, B.%
, Ryder, N.%
, Subbiah, M.%
, Kaplan, J.%
, Dhariwal, P.%
\BDBL {}others%
\end{APACrefauthors}%
\unskip\
\newblock
\APACrefYearMonthDay{2020}{}{}.
\newblock
{\BBOQ}\APACrefatitle {Language models are few-shot learners} {Language models
  are few-shot learners}.{\BBCQ}
\newblock
\APACjournalVolNumPages{arXiv preprint arXiv:2005.14165}{}{}{}.
\PrintBackRefs{\CurrentBib}

\bibitem [\protect \citeauthoryear {%
Bruineberg%
, Dolega%
, Dewhurst%
\BCBL {}\ \BBA {} Baltieri%
}{%
Bruineberg%
\ \protect \BOthers {.}}{%
{\protect \APACyear {2020}}%
}]{%
bruineberg2020emperor}
\APACinsertmetastar {%
bruineberg2020emperor}%
\begin{APACrefauthors}%
Bruineberg, J.%
, Dolega, K.%
, Dewhurst, J.%
\BCBL {}\ \BBA {} Baltieri, M.%
\end{APACrefauthors}%
\unskip\
\newblock
\APACrefYearMonthDay{2020}{}{}.
\newblock
{\BBOQ}\APACrefatitle {The Emperor’s New Markov Blankets} {The emperor’s
  new markov blankets}.{\BBCQ}
\newblock

\PrintBackRefs{\CurrentBib}

\bibitem [\protect \citeauthoryear {%
Buckley%
, Kim%
, McGregor%
\BCBL {}\ \BBA {} Seth%
}{%
Buckley%
\ \protect \BOthers {.}}{%
{\protect \APACyear {2017}}%
}]{%
buckley2017free}
\APACinsertmetastar {%
buckley2017free}%
\begin{APACrefauthors}%
Buckley, C\BPBI L.%
, Kim, C\BPBI S.%
, McGregor, S.%
\BCBL {}\ \BBA {} Seth, A\BPBI K.%
\end{APACrefauthors}%
\unskip\
\newblock
\APACrefYearMonthDay{2017}{}{}.
\newblock
{\BBOQ}\APACrefatitle {The free energy principle for action and perception: A
  mathematical review} {The free energy principle for action and perception: A
  mathematical review}.{\BBCQ}
\newblock
\APACjournalVolNumPages{Journal of Mathematical Psychology}{81}{}{55--79}.
\PrintBackRefs{\CurrentBib}

\bibitem [\protect \citeauthoryear {%
Buzsaki%
}{%
Buzsaki%
}{%
{\protect \APACyear {2006}}%
}]{%
buzsaki2006rhythms}
\APACinsertmetastar {%
buzsaki2006rhythms}%
\begin{APACrefauthors}%
Buzsaki, G.%
\end{APACrefauthors}%
\unskip\
\newblock
\APACrefYear{2006}.
\newblock
\APACrefbtitle {Rhythms of the Brain} {Rhythms of the brain}.
\newblock
\APACaddressPublisher{}{Oxford University Press}.
\PrintBackRefs{\CurrentBib}

\bibitem [\protect \citeauthoryear {%
{\c{C}}atal%
, Nauta%
, Verbelen%
, Simoens%
\BCBL {}\ \BBA {} Dhoedt%
}{%
{\c{C}}atal%
\ \protect \BOthers {.}}{%
{\protect \APACyear {2019}}%
}]{%
catal_Bayesian_2019}
\APACinsertmetastar {%
catal_Bayesian_2019}%
\begin{APACrefauthors}%
{\c{C}}atal, O.%
, Nauta, J.%
, Verbelen, T.%
, Simoens, P.%
\BCBL {}\ \BBA {} Dhoedt, B.%
\end{APACrefauthors}%
\unskip\
\newblock
\APACrefYearMonthDay{2019}{}{}.
\newblock
{\BBOQ}\APACrefatitle {Bayesian policy selection using active inference}
  {Bayesian policy selection using active inference}.{\BBCQ}
\newblock
\APACjournalVolNumPages{arXiv preprint arXiv:1904.08149}{}{}{}.
\PrintBackRefs{\CurrentBib}

\bibitem [\protect \citeauthoryear {%
{\c{C}}atal%
, Verbelen%
, Nauta%
, De~Boom%
\BCBL {}\ \BBA {} Dhoedt%
}{%
{\c{C}}atal%
\ \protect \BOthers {.}}{%
{\protect \APACyear {2020}}%
}]{%
ccatal2020learning}
\APACinsertmetastar {%
ccatal2020learning}%
\begin{APACrefauthors}%
{\c{C}}atal, O.%
, Verbelen, T.%
, Nauta, J.%
, De~Boom, C.%
\BCBL {}\ \BBA {} Dhoedt, B.%
\end{APACrefauthors}%
\unskip\
\newblock
\APACrefYearMonthDay{2020}{}{}.
\newblock
{\BBOQ}\APACrefatitle {Learning Perception and Planning with Deep Active
  Inference} {Learning perception and planning with deep active
  inference}.{\BBCQ}
\newblock
\APACjournalVolNumPages{arXiv preprint arXiv:2001.11841}{}{}{}.
\PrintBackRefs{\CurrentBib}

\bibitem [\protect \citeauthoryear {%
Caticha%
}{%
Caticha%
}{%
{\protect \APACyear {2015}}%
}]{%
caticha2015basics}
\APACinsertmetastar {%
caticha2015basics}%
\begin{APACrefauthors}%
Caticha, A.%
\end{APACrefauthors}%
\unskip\
\newblock
\APACrefYearMonthDay{2015}{}{}.
\newblock
{\BBOQ}\APACrefatitle {The basics of information geometry} {The basics of
  information geometry}.{\BBCQ}
\newblock
\BIn{} \APACrefbtitle {AIP Conference Proceedings} {Aip conference
  proceedings}\ (\BVOL\ 1641, \BPGS\ 15--26).
\PrintBackRefs{\CurrentBib}

\bibitem [\protect \citeauthoryear {%
Cesa-Bianchi%
, Gentile%
, Lugosi%
\BCBL {}\ \BBA {} Neu%
}{%
Cesa-Bianchi%
\ \protect \BOthers {.}}{%
{\protect \APACyear {2017}}%
}]{%
cesa2017boltzmann}
\APACinsertmetastar {%
cesa2017boltzmann}%
\begin{APACrefauthors}%
Cesa-Bianchi, N.%
, Gentile, C.%
, Lugosi, G.%
\BCBL {}\ \BBA {} Neu, G.%
\end{APACrefauthors}%
\unskip\
\newblock
\APACrefYearMonthDay{2017}{}{}.
\newblock
{\BBOQ}\APACrefatitle {Boltzmann exploration done right} {Boltzmann exploration
  done right}.{\BBCQ}
\newblock
\BIn{} \APACrefbtitle {Advances in Neural Information Processing Systems}
  {Advances in neural information processing systems}\ (\BPGS\ 6284--6293).
\PrintBackRefs{\CurrentBib}

\bibitem [\protect \citeauthoryear {%
Che%
\ \protect \BOthers {.}}{%
Che%
\ \protect \BOthers {.}}{%
{\protect \APACyear {2018}}%
}]{%
che2018combining}
\APACinsertmetastar {%
che2018combining}%
\begin{APACrefauthors}%
Che, T.%
, Lu, Y.%
, Tucker, G.%
, Bhupatiraju, S.%
, Gu, S.%
, Levine, S.%
\BCBL {}\ \BBA {} Bengio, Y.%
\end{APACrefauthors}%
\unskip\
\newblock
\APACrefYearMonthDay{2018}{}{}.
\newblock
{\BBOQ}\APACrefatitle {Combining Model-based and Model-free RL via Multi-step
  Control Variates} {Combining model-based and model-free rl via multi-step
  control variates}.{\BBCQ}
\newblock

\PrintBackRefs{\CurrentBib}

\bibitem [\protect \citeauthoryear {%
Chen%
, Fox%
\BCBL {}\ \BBA {} Guestrin%
}{%
Chen%
\ \protect \BOthers {.}}{%
{\protect \APACyear {2014}}%
}]{%
chen2014stochastic}
\APACinsertmetastar {%
chen2014stochastic}%
\begin{APACrefauthors}%
Chen, T.%
, Fox, E.%
\BCBL {}\ \BBA {} Guestrin, C.%
\end{APACrefauthors}%
\unskip\
\newblock
\APACrefYearMonthDay{2014}{}{}.
\newblock
{\BBOQ}\APACrefatitle {Stochastic gradient hamiltonian monte carlo} {Stochastic
  gradient hamiltonian monte carlo}.{\BBCQ}
\newblock
\BIn{} \APACrefbtitle {International Conference on Machine Learning}
  {International conference on machine learning}\ (\BPGS\ 1683--1691).
\PrintBackRefs{\CurrentBib}

\bibitem [\protect \citeauthoryear {%
Child%
}{%
Child%
}{%
{\protect \APACyear {2020}}%
}]{%
child2020very}
\APACinsertmetastar {%
child2020very}%
\begin{APACrefauthors}%
Child, R.%
\end{APACrefauthors}%
\unskip\
\newblock
\APACrefYearMonthDay{2020}{}{}.
\newblock
{\BBOQ}\APACrefatitle {Very Deep VAEs Generalize Autoregressive Models and Can
  Outperform Them on Images} {Very deep vaes generalize autoregressive models
  and can outperform them on images}.{\BBCQ}
\newblock
\APACjournalVolNumPages{arXiv preprint arXiv:2011.10650}{}{}{}.
\PrintBackRefs{\CurrentBib}

\bibitem [\protect \citeauthoryear {%
Chua%
, Calandra%
, McAllister%
\BCBL {}\ \BBA {} Levine%
}{%
Chua%
\ \protect \BOthers {.}}{%
{\protect \APACyear {2018}}%
}]{%
chua_deep_2018}
\APACinsertmetastar {%
chua_deep_2018}%
\begin{APACrefauthors}%
Chua, K.%
, Calandra, R.%
, McAllister, R.%
\BCBL {}\ \BBA {} Levine, S.%
\end{APACrefauthors}%
\unskip\
\newblock
\APACrefYearMonthDay{2018}{}{}.
\newblock
{\BBOQ}\APACrefatitle {Deep reinforcement learning in a handful of trials using
  probabilistic dynamics models} {Deep reinforcement learning in a handful of
  trials using probabilistic dynamics models}.{\BBCQ}
\newblock
\BIn{} \APACrefbtitle {Advances in Neural Information Processing Systems}
  {Advances in neural information processing systems}\ (\BPGS\ 4754--4765).
\PrintBackRefs{\CurrentBib}

\bibitem [\protect \citeauthoryear {%
Clark%
}{%
Clark%
}{%
{\protect \APACyear {2013}}%
{\protect \APACexlab {{\protect \BCnt {1}}}}}]{%
clark2013whatever}
\APACinsertmetastar {%
clark2013whatever}%
\begin{APACrefauthors}%
Clark, A.%
\end{APACrefauthors}%
\unskip\
\newblock
\APACrefYearMonthDay{2013{\protect \BCnt {1}}}{}{}.
\newblock
{\BBOQ}\APACrefatitle {Whatever next? Predictive brains, situated agents, and
  the future of cognitive science} {Whatever next? predictive brains, situated
  agents, and the future of cognitive science}.{\BBCQ}
\newblock
\APACjournalVolNumPages{Behavioral and brain sciences}{36}{3}{181--204}.
\newblock
\begin{APACrefURL}
  \url{https://www.cambridge.org/core/journals/behavioral-and-brain-sciences/article/whatever-next-predictive-brains-situated-agents-and-the-future-of-cognitive-science/33542C736E17E3D1D44E8D03BE5F4CD9}
  \end{APACrefURL}
\PrintBackRefs{\CurrentBib}

\bibitem [\protect \citeauthoryear {%
Clark%
}{%
Clark%
}{%
{\protect \APACyear {2013}}%
{\protect \APACexlab {{\protect \BCnt {2}}}}}]{%
clark_whatever_2013}
\APACinsertmetastar {%
clark_whatever_2013}%
\begin{APACrefauthors}%
Clark, A.%
\end{APACrefauthors}%
\unskip\
\newblock
\APACrefYearMonthDay{2013{\protect \BCnt {2}}}{}{}.
\newblock
{\BBOQ}\APACrefatitle {Whatever next? Predictive brains, situated agents, and
  the future of cognitive science} {Whatever next? predictive brains, situated
  agents, and the future of cognitive science}.{\BBCQ}
\newblock
\APACjournalVolNumPages{}{36}{3}{181--204}.
\newblock
\begin{APACrefDOI} \doi{10.1017/S0140525X12000477} \end{APACrefDOI}
\PrintBackRefs{\CurrentBib}

\bibitem [\protect \citeauthoryear {%
Clark%
}{%
Clark%
}{%
{\protect \APACyear {2015}}%
}]{%
clark2015surfing}
\APACinsertmetastar {%
clark2015surfing}%
\begin{APACrefauthors}%
Clark, A.%
\end{APACrefauthors}%
\unskip\
\newblock
\APACrefYear{2015}.
\newblock
\APACrefbtitle {Surfing uncertainty: Prediction, action, and the embodied mind}
  {Surfing uncertainty: Prediction, action, and the embodied mind}.
\newblock
\APACaddressPublisher{}{Oxford University Press}.
\newblock
\begin{APACrefURL}
  \url{https://books.google.co.uk/books?hl=en&lr=&id=TnqECgAAQBAJ&oi=fnd&pg=PP1&dq=andy+clark+surfing+uncertainty&ots=aurm4jE3NO&sig=KxeHGJ6YJJdN9tKyr6snwDyBBKg&redir_esc=y#v=onepage&q=andy%20clark%20surfing%20uncertainty&f=false}
  \end{APACrefURL}
\PrintBackRefs{\CurrentBib}

\bibitem [\protect \citeauthoryear {%
Cohen%
, McClure%
\BCBL {}\ \BBA {} Yu%
}{%
Cohen%
\ \protect \BOthers {.}}{%
{\protect \APACyear {2007}}%
}]{%
cohen2007should}
\APACinsertmetastar {%
cohen2007should}%
\begin{APACrefauthors}%
Cohen, J\BPBI D.%
, McClure, S\BPBI M.%
\BCBL {}\ \BBA {} Yu, A\BPBI J.%
\end{APACrefauthors}%
\unskip\
\newblock
\APACrefYearMonthDay{2007}{}{}.
\newblock
{\BBOQ}\APACrefatitle {Should I stay or should I go? How the human brain
  manages the trade-off between exploitation and exploration} {Should i stay or
  should i go? how the human brain manages the trade-off between exploitation
  and exploration}.{\BBCQ}
\newblock
\APACjournalVolNumPages{Philosophical Transactions of the Royal Society B:
  Biological Sciences}{362}{1481}{933--942}.
\PrintBackRefs{\CurrentBib}

\bibitem [\protect \citeauthoryear {%
Conant%
\ \BBA {} Ross~Ashby%
}{%
Conant%
\ \BBA {} Ross~Ashby%
}{%
{\protect \APACyear {1970}}%
}]{%
conant1970every}
\APACinsertmetastar {%
conant1970every}%
\begin{APACrefauthors}%
Conant, R\BPBI C.%
\BCBT {}\ \BBA {} Ross~Ashby, W.%
\end{APACrefauthors}%
\unskip\
\newblock
\APACrefYearMonthDay{1970}{}{}.
\newblock
{\BBOQ}\APACrefatitle {Every good regulator of a system must be a model of that
  system} {Every good regulator of a system must be a model of that
  system}.{\BBCQ}
\newblock
\APACjournalVolNumPages{International journal of systems
  science}{1}{2}{89--97}.
\PrintBackRefs{\CurrentBib}

\bibitem [\protect \citeauthoryear {%
Crick%
}{%
Crick%
}{%
{\protect \APACyear {1989}}%
}]{%
crick1989recent}
\APACinsertmetastar {%
crick1989recent}%
\begin{APACrefauthors}%
Crick, F.%
\end{APACrefauthors}%
\unskip\
\newblock
\APACrefYearMonthDay{1989}{}{}.
\newblock
{\BBOQ}\APACrefatitle {The recent excitement about neural networks} {The recent
  excitement about neural networks}.{\BBCQ}
\newblock
\APACjournalVolNumPages{Nature}{337}{6203}{129--132}.
\PrintBackRefs{\CurrentBib}

\bibitem [\protect \citeauthoryear {%
Cullen%
, Davey%
, Friston%
\BCBL {}\ \BBA {} Moran%
}{%
Cullen%
\ \protect \BOthers {.}}{%
{\protect \APACyear {2018}}%
}]{%
cullen2018active}
\APACinsertmetastar {%
cullen2018active}%
\begin{APACrefauthors}%
Cullen, M.%
, Davey, B.%
, Friston, K.%
\BCBL {}\ \BBA {} Moran, R\BPBI J.%
\end{APACrefauthors}%
\unskip\
\newblock
\APACrefYearMonthDay{2018}{}{}.
\newblock
{\BBOQ}\APACrefatitle {Active inference in openai gym: A paradigm for
  computational investigations into psychiatric illness} {Active inference in
  openai gym: A paradigm for computational investigations into psychiatric
  illness}.{\BBCQ}
\newblock
\APACjournalVolNumPages{Biological psychiatry: Cognitive Neuroscience and
  neuroimaging}{3}{9}{809--818}.
\PrintBackRefs{\CurrentBib}

\bibitem [\protect \citeauthoryear {%
Da~Costa%
, Parr%
\BCBL {}\ \protect \BOthers {.}}{%
Da~Costa%
, Parr%
\BCBL {}\ \protect \BOthers {.}}{%
{\protect \APACyear {2020}}%
}]{%
da2020active}
\APACinsertmetastar {%
da2020active}%
\begin{APACrefauthors}%
Da~Costa, L.%
, Parr, T.%
, Sajid, N.%
, Veselic, S.%
, Neacsu, V.%
\BCBL {}\ \BBA {} Friston, K.%
\end{APACrefauthors}%
\unskip\
\newblock
\APACrefYearMonthDay{2020}{}{}.
\newblock
{\BBOQ}\APACrefatitle {Active inference on discrete state-spaces: a synthesis}
  {Active inference on discrete state-spaces: a synthesis}.{\BBCQ}
\newblock
\APACjournalVolNumPages{arXiv preprint arXiv:2001.07203}{}{}{}.
\PrintBackRefs{\CurrentBib}

\bibitem [\protect \citeauthoryear {%
Da~Costa%
, Sajid%
, Parr%
, Friston%
\BCBL {}\ \BBA {} Smith%
}{%
Da~Costa%
, Sajid%
\BCBL {}\ \protect \BOthers {.}}{%
{\protect \APACyear {2020}}%
}]{%
da2020relationship}
\APACinsertmetastar {%
da2020relationship}%
\begin{APACrefauthors}%
Da~Costa, L.%
, Sajid, N.%
, Parr, T.%
, Friston, K.%
\BCBL {}\ \BBA {} Smith, R.%
\end{APACrefauthors}%
\unskip\
\newblock
\APACrefYearMonthDay{2020}{}{}.
\newblock
{\BBOQ}\APACrefatitle {The relationship between dynamic programming and active
  inference: The discrete, finite-horizon case} {The relationship between
  dynamic programming and active inference: The discrete, finite-horizon
  case}.{\BBCQ}
\newblock
\APACjournalVolNumPages{arXiv preprint arXiv:2009.08111}{}{}{}.
\newblock
\begin{APACrefURL} \url{https://arxiv.org/abs/2009.08111} \end{APACrefURL}
\PrintBackRefs{\CurrentBib}

\bibitem [\protect \citeauthoryear {%
Daw%
, O'doherty%
, Dayan%
, Seymour%
\BCBL {}\ \BBA {} Dolan%
}{%
Daw%
\ \protect \BOthers {.}}{%
{\protect \APACyear {2006}}%
}]{%
daw2006cortical}
\APACinsertmetastar {%
daw2006cortical}%
\begin{APACrefauthors}%
Daw, N\BPBI D.%
, O'doherty, J\BPBI P.%
, Dayan, P.%
, Seymour, B.%
\BCBL {}\ \BBA {} Dolan, R\BPBI J.%
\end{APACrefauthors}%
\unskip\
\newblock
\APACrefYearMonthDay{2006}{}{}.
\newblock
{\BBOQ}\APACrefatitle {Cortical substrates for exploratory decisions in humans}
  {Cortical substrates for exploratory decisions in humans}.{\BBCQ}
\newblock
\APACjournalVolNumPages{Nature}{441}{7095}{876--879}.
\PrintBackRefs{\CurrentBib}

\bibitem [\protect \citeauthoryear {%
Dayan%
}{%
Dayan%
}{%
{\protect \APACyear {2009}}%
}]{%
dayan2009goal}
\APACinsertmetastar {%
dayan2009goal}%
\begin{APACrefauthors}%
Dayan, P.%
\end{APACrefauthors}%
\unskip\
\newblock
\APACrefYearMonthDay{2009}{}{}.
\newblock
{\BBOQ}\APACrefatitle {Goal-directed control and its antipodes} {Goal-directed
  control and its antipodes}.{\BBCQ}
\newblock
\APACjournalVolNumPages{Neural Networks}{22}{3}{213--219}.
\PrintBackRefs{\CurrentBib}

\bibitem [\protect \citeauthoryear {%
Dayan%
\ \BBA {} Daw%
}{%
Dayan%
\ \BBA {} Daw%
}{%
{\protect \APACyear {2008}}%
}]{%
dayan2008decision}
\APACinsertmetastar {%
dayan2008decision}%
\begin{APACrefauthors}%
Dayan, P.%
\BCBT {}\ \BBA {} Daw, N\BPBI D.%
\end{APACrefauthors}%
\unskip\
\newblock
\APACrefYearMonthDay{2008}{}{}.
\newblock
{\BBOQ}\APACrefatitle {Decision theory, reinforcement learning, and the brain}
  {Decision theory, reinforcement learning, and the brain}.{\BBCQ}
\newblock
\APACjournalVolNumPages{Cognitive, Affective, \& Behavioral
  Neuroscience}{8}{4}{429--453}.
\PrintBackRefs{\CurrentBib}

\bibitem [\protect \citeauthoryear {%
Dayan%
\ \BBA {} Hinton%
}{%
Dayan%
\ \BBA {} Hinton%
}{%
{\protect \APACyear {1997}}%
}]{%
dayan1997using}
\APACinsertmetastar {%
dayan1997using}%
\begin{APACrefauthors}%
Dayan, P.%
\BCBT {}\ \BBA {} Hinton, G.%
\end{APACrefauthors}%
\unskip\
\newblock
\APACrefYearMonthDay{1997}{}{}.
\newblock
{\BBOQ}\APACrefatitle {Using expectation-maximization for reinforcement
  learning} {Using expectation-maximization for reinforcement learning}.{\BBCQ}
\newblock
\APACjournalVolNumPages{Neural Computation}{9}{2}{271--278}.
\PrintBackRefs{\CurrentBib}

\bibitem [\protect \citeauthoryear {%
Dayan%
, Hinton%
, Neal%
\BCBL {}\ \BBA {} Zemel%
}{%
Dayan%
\ \protect \BOthers {.}}{%
{\protect \APACyear {1995}}%
}]{%
dayan1995helmholtz}
\APACinsertmetastar {%
dayan1995helmholtz}%
\begin{APACrefauthors}%
Dayan, P.%
, Hinton, G.%
, Neal, R\BPBI M.%
\BCBL {}\ \BBA {} Zemel, R\BPBI S.%
\end{APACrefauthors}%
\unskip\
\newblock
\APACrefYearMonthDay{1995}{}{}.
\newblock
{\BBOQ}\APACrefatitle {The helmholtz machine} {The helmholtz machine}.{\BBCQ}
\newblock
\APACjournalVolNumPages{Neural Computation}{7}{5}{889--904}.
\PrintBackRefs{\CurrentBib}

\bibitem [\protect \citeauthoryear {%
De~Boer%
, Kroese%
, Mannor%
\BCBL {}\ \BBA {} Rubinstein%
}{%
De~Boer%
\ \protect \BOthers {.}}{%
{\protect \APACyear {2005}}%
}]{%
de2005tutorial}
\APACinsertmetastar {%
de2005tutorial}%
\begin{APACrefauthors}%
De~Boer, P\BHBI T.%
, Kroese, D\BPBI P.%
, Mannor, S.%
\BCBL {}\ \BBA {} Rubinstein, R\BPBI Y.%
\end{APACrefauthors}%
\unskip\
\newblock
\APACrefYearMonthDay{2005}{}{}.
\newblock
{\BBOQ}\APACrefatitle {A tutorial on the cross-entropy method} {A tutorial on
  the cross-entropy method}.{\BBCQ}
\newblock
\APACjournalVolNumPages{Annals of Operations Research}{134}{1}{19--67}.
\PrintBackRefs{\CurrentBib}

\bibitem [\protect \citeauthoryear {%
Dempster%
, Laird%
\BCBL {}\ \BBA {} Rubin%
}{%
Dempster%
\ \protect \BOthers {.}}{%
{\protect \APACyear {1977}}%
}]{%
dempster1977maximum}
\APACinsertmetastar {%
dempster1977maximum}%
\begin{APACrefauthors}%
Dempster, A\BPBI P.%
, Laird, N\BPBI M.%
\BCBL {}\ \BBA {} Rubin, D\BPBI B.%
\end{APACrefauthors}%
\unskip\
\newblock
\APACrefYearMonthDay{1977}{}{}.
\newblock
{\BBOQ}\APACrefatitle {Maximum likelihood from incomplete data via the EM
  algorithm} {Maximum likelihood from incomplete data via the em
  algorithm}.{\BBCQ}
\newblock
\APACjournalVolNumPages{Journal of the Royal Statistical Society: Series B
  (Methodological)}{39}{1}{1--22}.
\PrintBackRefs{\CurrentBib}

\bibitem [\protect \citeauthoryear {%
de Xivry%
, Coppe%
, Blohm%
\BCBL {}\ \BBA {} Lefevre%
}{%
de Xivry%
\ \protect \BOthers {.}}{%
{\protect \APACyear {2013}}%
}]{%
de2013Kalman}
\APACinsertmetastar {%
de2013Kalman}%
\begin{APACrefauthors}%
de Xivry, J\BHBI J\BPBI O.%
, Coppe, S.%
, Blohm, G.%
\BCBL {}\ \BBA {} Lefevre, P.%
\end{APACrefauthors}%
\unskip\
\newblock
\APACrefYearMonthDay{2013}{}{}.
\newblock
{\BBOQ}\APACrefatitle {Kalman filtering naturally accounts for visually guided
  and predictive smooth pursuit dynamics} {Kalman filtering naturally accounts
  for visually guided and predictive smooth pursuit dynamics}.{\BBCQ}
\newblock
\APACjournalVolNumPages{Journal of Neuroscience}{33}{44}{17301--17313}.
\PrintBackRefs{\CurrentBib}

\bibitem [\protect \citeauthoryear {%
Doucet%
, Godsill%
\BCBL {}\ \BBA {} Andrieu%
}{%
Doucet%
\ \protect \BOthers {.}}{%
{\protect \APACyear {2000}}%
}]{%
doucet2000sequential}
\APACinsertmetastar {%
doucet2000sequential}%
\begin{APACrefauthors}%
Doucet, A.%
, Godsill, S.%
\BCBL {}\ \BBA {} Andrieu, C.%
\end{APACrefauthors}%
\unskip\
\newblock
\APACrefYearMonthDay{2000}{}{}.
\newblock
{\BBOQ}\APACrefatitle {On sequential Monte Carlo sampling methods for Bayesian
  filtering} {On sequential monte carlo sampling methods for bayesian
  filtering}.{\BBCQ}
\newblock
\APACjournalVolNumPages{Statistics and Computing}{10}{3}{197--208}.
\PrintBackRefs{\CurrentBib}

\bibitem [\protect \citeauthoryear {%
Doya%
}{%
Doya%
}{%
{\protect \APACyear {2000}}%
}]{%
doya2000reinforcement}
\APACinsertmetastar {%
doya2000reinforcement}%
\begin{APACrefauthors}%
Doya, K.%
\end{APACrefauthors}%
\unskip\
\newblock
\APACrefYearMonthDay{2000}{}{}.
\newblock
{\BBOQ}\APACrefatitle {Reinforcement learning in continuous time and space}
  {Reinforcement learning in continuous time and space}.{\BBCQ}
\newblock
\APACjournalVolNumPages{Neural Computation}{12}{1}{219--245}.
\PrintBackRefs{\CurrentBib}

\bibitem [\protect \citeauthoryear {%
Esposito%
\ \BBA {} Van~den Broeck%
}{%
Esposito%
\ \BBA {} Van~den Broeck%
}{%
{\protect \APACyear {2010}}%
}]{%
esposito2010three1}
\APACinsertmetastar {%
esposito2010three1}%
\begin{APACrefauthors}%
Esposito, M.%
\BCBT {}\ \BBA {} Van~den Broeck, C.%
\end{APACrefauthors}%
\unskip\
\newblock
\APACrefYearMonthDay{2010}{}{}.
\newblock
{\BBOQ}\APACrefatitle {Three faces of the second law. I. Master equation
  formulation} {Three faces of the second law. i. master equation
  formulation}.{\BBCQ}
\newblock
\APACjournalVolNumPages{Physical Review E}{82}{1}{011143}.
\PrintBackRefs{\CurrentBib}

\bibitem [\protect \citeauthoryear {%
Farshidian%
, Neunert%
\BCBL {}\ \BBA {} Buchli%
}{%
Farshidian%
\ \protect \BOthers {.}}{%
{\protect \APACyear {2014}}%
}]{%
farshidian2014learning}
\APACinsertmetastar {%
farshidian2014learning}%
\begin{APACrefauthors}%
Farshidian, F.%
, Neunert, M.%
\BCBL {}\ \BBA {} Buchli, J.%
\end{APACrefauthors}%
\unskip\
\newblock
\APACrefYearMonthDay{2014}{}{}.
\newblock
{\BBOQ}\APACrefatitle {Learning of closed-loop motion control} {Learning of
  closed-loop motion control}.{\BBCQ}
\newblock
\BIn{} \APACrefbtitle {2014 IEEE/RSJ International Conference on Intelligent
  Robots and Systems} {2014 ieee/rsj international conference on intelligent
  robots and systems}\ (\BPGS\ 1441--1446).
\PrintBackRefs{\CurrentBib}

\bibitem [\protect \citeauthoryear {%
Feldman%
\ \BBA {} Friston%
}{%
Feldman%
\ \BBA {} Friston%
}{%
{\protect \APACyear {2010}}%
}]{%
feldman2010attention}
\APACinsertmetastar {%
feldman2010attention}%
\begin{APACrefauthors}%
Feldman, H.%
\BCBT {}\ \BBA {} Friston, K.%
\end{APACrefauthors}%
\unskip\
\newblock
\APACrefYearMonthDay{2010}{}{}.
\newblock
{\BBOQ}\APACrefatitle {Attention, uncertainty, and free-energy} {Attention,
  uncertainty, and free-energy}.{\BBCQ}
\newblock
\APACjournalVolNumPages{Frontiers in human neuroscience}{4}{}{215}.
\newblock
\begin{APACrefURL}
  \url{https://www.frontiersin.org/articles/10.3389/fnhum.2010.00215/full}
  \end{APACrefURL}
\PrintBackRefs{\CurrentBib}

\bibitem [\protect \citeauthoryear {%
Felleman%
\ \BBA {} Van~Essen%
}{%
Felleman%
\ \BBA {} Van~Essen%
}{%
{\protect \APACyear {1991}}%
}]{%
felleman1991distributed}
\APACinsertmetastar {%
felleman1991distributed}%
\begin{APACrefauthors}%
Felleman, D\BPBI J.%
\BCBT {}\ \BBA {} Van~Essen, D\BPBI C.%
\end{APACrefauthors}%
\unskip\
\newblock
\APACrefYearMonthDay{1991}{}{}.
\newblock
{\BBOQ}\APACrefatitle {Distributed hierarchical processing in the primate
  cerebral cortex} {Distributed hierarchical processing in the primate cerebral
  cortex}.{\BBCQ}
\newblock
\BIn{} \APACrefbtitle {Cereb cortex.} {Cereb cortex.}
\PrintBackRefs{\CurrentBib}

\bibitem [\protect \citeauthoryear {%
Feng%
, Whitman%
, Xinjilefu%
\BCBL {}\ \BBA {} Atkeson%
}{%
Feng%
\ \protect \BOthers {.}}{%
{\protect \APACyear {2014}}%
}]{%
feng2014optimization}
\APACinsertmetastar {%
feng2014optimization}%
\begin{APACrefauthors}%
Feng, S.%
, Whitman, E.%
, Xinjilefu, X.%
\BCBL {}\ \BBA {} Atkeson, C\BPBI G.%
\end{APACrefauthors}%
\unskip\
\newblock
\APACrefYearMonthDay{2014}{}{}.
\newblock
{\BBOQ}\APACrefatitle {Optimization based full body control for the atlas
  robot} {Optimization based full body control for the atlas robot}.{\BBCQ}
\newblock
\BIn{} \APACrefbtitle {2014 IEEE-RAS International Conference on Humanoid
  Robots} {2014 ieee-ras international conference on humanoid robots}\ (\BPGS\
  120--127).
\PrintBackRefs{\CurrentBib}

\bibitem [\protect \citeauthoryear {%
Feynman%
}{%
Feynman%
}{%
{\protect \APACyear {1998}}%
}]{%
feynman1998statistical}
\APACinsertmetastar {%
feynman1998statistical}%
\begin{APACrefauthors}%
Feynman, R.%
\end{APACrefauthors}%
\unskip\
\newblock
\APACrefYearMonthDay{1998}{}{}.
\newblock
{\BBOQ}\APACrefatitle {Statistical mechanics: a set of lectures (advanced book
  classics)} {Statistical mechanics: a set of lectures (advanced book
  classics)}.{\BBCQ}
\newblock

\PrintBackRefs{\CurrentBib}

\bibitem [\protect \citeauthoryear {%
Fodor%
}{%
Fodor%
}{%
{\protect \APACyear {1983}}%
}]{%
fodor1983modularity}
\APACinsertmetastar {%
fodor1983modularity}%
\begin{APACrefauthors}%
Fodor, J\BPBI A.%
\end{APACrefauthors}%
\unskip\
\newblock
\APACrefYear{1983}.
\newblock
\APACrefbtitle {The modularity of mind} {The modularity of mind}.
\newblock
\APACaddressPublisher{}{MIT press}.
\PrintBackRefs{\CurrentBib}

\bibitem [\protect \citeauthoryear {%
Fountas%
, Sajid%
, Mediano%
\BCBL {}\ \BBA {} Friston%
}{%
Fountas%
\ \protect \BOthers {.}}{%
{\protect \APACyear {2020}}%
}]{%
fountas2020deep}
\APACinsertmetastar {%
fountas2020deep}%
\begin{APACrefauthors}%
Fountas, Z.%
, Sajid, N.%
, Mediano, P\BPBI A.%
\BCBL {}\ \BBA {} Friston, K.%
\end{APACrefauthors}%
\unskip\
\newblock
\APACrefYearMonthDay{2020}{}{}.
\newblock
{\BBOQ}\APACrefatitle {Deep active inference agents using Monte-Carlo methods}
  {Deep active inference agents using monte-carlo methods}.{\BBCQ}
\newblock
\APACjournalVolNumPages{arXiv preprint arXiv:2006.04176}{}{}{}.
\PrintBackRefs{\CurrentBib}

\bibitem [\protect \citeauthoryear {%
Fox%
\ \BBA {} Roberts%
}{%
Fox%
\ \BBA {} Roberts%
}{%
{\protect \APACyear {2012}}%
}]{%
fox2012tutorial}
\APACinsertmetastar {%
fox2012tutorial}%
\begin{APACrefauthors}%
Fox, C\BPBI W.%
\BCBT {}\ \BBA {} Roberts, S\BPBI J.%
\end{APACrefauthors}%
\unskip\
\newblock
\APACrefYearMonthDay{2012}{}{}.
\newblock
{\BBOQ}\APACrefatitle {A tutorial on variational Bayesian inference} {A
  tutorial on variational bayesian inference}.{\BBCQ}
\newblock
\APACjournalVolNumPages{Artificial intelligence review}{38}{2}{85--95}.
\PrintBackRefs{\CurrentBib}

\bibitem [\protect \citeauthoryear {%
Friston%
}{%
Friston%
}{%
{\protect \APACyear {2003}}%
}]{%
friston2003learning}
\APACinsertmetastar {%
friston2003learning}%
\begin{APACrefauthors}%
Friston, K.%
\end{APACrefauthors}%
\unskip\
\newblock
\APACrefYearMonthDay{2003}{}{}.
\newblock
{\BBOQ}\APACrefatitle {Learning and inference in the brain} {Learning and
  inference in the brain}.{\BBCQ}
\newblock
\APACjournalVolNumPages{Neural Networks}{16}{9}{1325--1352}.
\PrintBackRefs{\CurrentBib}

\bibitem [\protect \citeauthoryear {%
Friston%
}{%
Friston%
}{%
{\protect \APACyear {2005}}%
}]{%
friston2005theory}
\APACinsertmetastar {%
friston2005theory}%
\begin{APACrefauthors}%
Friston, K.%
\end{APACrefauthors}%
\unskip\
\newblock
\APACrefYearMonthDay{2005}{}{}.
\newblock
{\BBOQ}\APACrefatitle {A theory of cortical responses} {A theory of cortical
  responses}.{\BBCQ}
\newblock
\APACjournalVolNumPages{Philosophical Transactions of the Royal Society B:
  Biological sciences}{360}{1456}{815--836}.
\PrintBackRefs{\CurrentBib}

\bibitem [\protect \citeauthoryear {%
Friston%
}{%
Friston%
}{%
{\protect \APACyear {2008}}%
{\protect \APACexlab {{\protect \BCnt {1}}}}}]{%
friston2008hierarchical}
\APACinsertmetastar {%
friston2008hierarchical}%
\begin{APACrefauthors}%
Friston, K.%
\end{APACrefauthors}%
\unskip\
\newblock
\APACrefYearMonthDay{2008{\protect \BCnt {1}}}{}{}.
\newblock
{\BBOQ}\APACrefatitle {Hierarchical models in the brain} {Hierarchical models
  in the brain}.{\BBCQ}
\newblock
\APACjournalVolNumPages{PLoS Computational Biology}{4}{11}{}.
\PrintBackRefs{\CurrentBib}

\bibitem [\protect \citeauthoryear {%
Friston%
}{%
Friston%
}{%
{\protect \APACyear {2008}}%
{\protect \APACexlab {{\protect \BCnt {2}}}}}]{%
friston2008variational}
\APACinsertmetastar {%
friston2008variational}%
\begin{APACrefauthors}%
Friston, K.%
\end{APACrefauthors}%
\unskip\
\newblock
\APACrefYearMonthDay{2008{\protect \BCnt {2}}}{}{}.
\newblock
{\BBOQ}\APACrefatitle {Variational filtering} {Variational filtering}.{\BBCQ}
\newblock
\APACjournalVolNumPages{NeuroImage}{41}{3}{747--766}.
\PrintBackRefs{\CurrentBib}

\bibitem [\protect \citeauthoryear {%
Friston%
}{%
Friston%
}{%
{\protect \APACyear {2009}}%
}]{%
friston2009free}
\APACinsertmetastar {%
friston2009free}%
\begin{APACrefauthors}%
Friston, K.%
\end{APACrefauthors}%
\unskip\
\newblock
\APACrefYearMonthDay{2009}{}{}.
\newblock
{\BBOQ}\APACrefatitle {The free-energy principle: a rough guide to the brain?}
  {The free-energy principle: a rough guide to the brain?}{\BBCQ}
\newblock
\APACjournalVolNumPages{Trends in Cognitive Sciences}{13}{7}{293--301}.
\PrintBackRefs{\CurrentBib}

\bibitem [\protect \citeauthoryear {%
Friston%
}{%
Friston%
}{%
{\protect \APACyear {2010}}%
}]{%
friston2010free}
\APACinsertmetastar {%
friston2010free}%
\begin{APACrefauthors}%
Friston, K.%
\end{APACrefauthors}%
\unskip\
\newblock
\APACrefYearMonthDay{2010}{}{}.
\newblock
{\BBOQ}\APACrefatitle {The free-energy principle: a unified brain theory?} {The
  free-energy principle: a unified brain theory?}{\BBCQ}
\newblock
\APACjournalVolNumPages{Nature reviews neuroscience}{11}{2}{127--138}.
\PrintBackRefs{\CurrentBib}

\bibitem [\protect \citeauthoryear {%
Friston%
}{%
Friston%
}{%
{\protect \APACyear {2012}}%
}]{%
friston2012history}
\APACinsertmetastar {%
friston2012history}%
\begin{APACrefauthors}%
Friston, K.%
\end{APACrefauthors}%
\unskip\
\newblock
\APACrefYearMonthDay{2012}{}{}.
\newblock
{\BBOQ}\APACrefatitle {The history of the future of the Bayesian brain} {The
  history of the future of the bayesian brain}.{\BBCQ}
\newblock
\APACjournalVolNumPages{NeuroImage}{62}{2}{1230--1233}.
\newblock
\begin{APACrefURL}
  \url{https://www.sciencedirect.com/science/article/pii/S1053811911011657}
  \end{APACrefURL}
\PrintBackRefs{\CurrentBib}

\bibitem [\protect \citeauthoryear {%
Friston%
}{%
Friston%
}{%
{\protect \APACyear {2013}}%
}]{%
friston2013life}
\APACinsertmetastar {%
friston2013life}%
\begin{APACrefauthors}%
Friston, K.%
\end{APACrefauthors}%
\unskip\
\newblock
\APACrefYearMonthDay{2013}{}{}.
\newblock
{\BBOQ}\APACrefatitle {Life as we know it} {Life as we know it}.{\BBCQ}
\newblock
\APACjournalVolNumPages{Journal of the Royal Society
  Interface}{10}{86}{20130475}.
\PrintBackRefs{\CurrentBib}

\bibitem [\protect \citeauthoryear {%
Friston%
}{%
Friston%
}{%
{\protect \APACyear {2019}}%
{\protect \APACexlab {{\protect \BCnt {1}}}}}]{%
friston2019particularphysics}
\APACinsertmetastar {%
friston2019particularphysics}%
\begin{APACrefauthors}%
Friston, K.%
\end{APACrefauthors}%
\unskip\
\newblock
\APACrefYearMonthDay{2019{\protect \BCnt {1}}}{}{}.
\newblock
{\BBOQ}\APACrefatitle {A free energy principle for a particular physics} {A
  free energy principle for a particular physics}.{\BBCQ}
\newblock
\APACjournalVolNumPages{arXiv preprint arXiv:1906.10184}{}{}{}.
\PrintBackRefs{\CurrentBib}

\bibitem [\protect \citeauthoryear {%
Friston%
}{%
Friston%
}{%
{\protect \APACyear {2019}}%
{\protect \APACexlab {{\protect \BCnt {2}}}}}]{%
friston2019free}
\APACinsertmetastar {%
friston2019free}%
\begin{APACrefauthors}%
Friston, K.%
\end{APACrefauthors}%
\unskip\
\newblock
\APACrefYearMonthDay{2019{\protect \BCnt {2}}}{}{}.
\newblock
{\BBOQ}\APACrefatitle {A free energy principle for a particular physics} {A
  free energy principle for a particular physics}.{\BBCQ}
\newblock
\APACjournalVolNumPages{arXiv preprint arXiv:1906.10184}{}{}{}.
\newblock
\begin{APACrefURL} \url{https://arxiv.org/pdf/1906.10184.pdf} \end{APACrefURL}
\PrintBackRefs{\CurrentBib}

\bibitem [\protect \citeauthoryear {%
Friston%
}{%
Friston%
}{%
{\protect \APACyear {2019}}%
{\protect \APACexlab {{\protect \BCnt {3}}}}}]{%
friston_free_2019}
\APACinsertmetastar {%
friston_free_2019}%
\begin{APACrefauthors}%
Friston, K.%
\end{APACrefauthors}%
\unskip\
\newblock
\APACrefYearMonthDay{2019{\protect \BCnt {3}}}{}{}.
\newblock
{\BBOQ}\APACrefatitle {A free energy principle for a particular physics} {A
  free energy principle for a particular physics}.{\BBCQ}
\newblock
\APACjournalVolNumPages{arXiv preprint arXiv:1906.10184}{}{}{}.
\PrintBackRefs{\CurrentBib}

\bibitem [\protect \citeauthoryear {%
Friston%
}{%
Friston%
}{%
{\protect \APACyear {2019}}%
{\protect \APACexlab {{\protect \BCnt {4}}}}}]{%
friston2019physics}
\APACinsertmetastar {%
friston2019physics}%
\begin{APACrefauthors}%
Friston, K.%
\end{APACrefauthors}%
\unskip\
\newblock
\APACrefYearMonthDay{2019{\protect \BCnt {4}}}{}{}.
\newblock
{\BBOQ}\APACrefatitle {A free energy principle for a particular physics} {A
  free energy principle for a particular physics}.{\BBCQ}
\newblock
\APACjournalVolNumPages{arXiv preprint arXiv:1906.10184}{}{}{}.
\PrintBackRefs{\CurrentBib}

\bibitem [\protect \citeauthoryear {%
Friston%
\ \BBA {} Ao%
}{%
Friston%
\ \BBA {} Ao%
}{%
{\protect \APACyear {2012}}%
{\protect \APACexlab {{\protect \BCnt {1}}}}}]{%
friston2012free}
\APACinsertmetastar {%
friston2012free}%
\begin{APACrefauthors}%
Friston, K.%
\BCBT {}\ \BBA {} Ao, P.%
\end{APACrefauthors}%
\unskip\
\newblock
\APACrefYearMonthDay{2012{\protect \BCnt {1}}}{}{}.
\newblock
{\BBOQ}\APACrefatitle {Free energy, value, and attractors} {Free energy, value,
  and attractors}.{\BBCQ}
\newblock
\APACjournalVolNumPages{Computational and mathematical methods in
  medicine}{2012}{}{}.
\PrintBackRefs{\CurrentBib}

\bibitem [\protect \citeauthoryear {%
Friston%
\ \BBA {} Ao%
}{%
Friston%
\ \BBA {} Ao%
}{%
{\protect \APACyear {2012}}%
{\protect \APACexlab {{\protect \BCnt {2}}}}}]{%
friston2012ao}
\APACinsertmetastar {%
friston2012ao}%
\begin{APACrefauthors}%
Friston, K.%
\BCBT {}\ \BBA {} Ao, P.%
\end{APACrefauthors}%
\unskip\
\newblock
\APACrefYearMonthDay{2012{\protect \BCnt {2}}}{}{}.
\newblock
{\BBOQ}\APACrefatitle {Free energy, value, and attractors} {Free energy, value,
  and attractors}.{\BBCQ}
\newblock
\APACjournalVolNumPages{Computational and mathematical methods in
  medicine}{2012}{}{}.
\PrintBackRefs{\CurrentBib}

\bibitem [\protect \citeauthoryear {%
Friston%
, Da~Costa%
, Hafner%
, Hesp%
\BCBL {}\ \BBA {} Parr%
}{%
Friston%
, Da~Costa%
, Hafner%
\BCBL {}\ \protect \BOthers {.}}{%
{\protect \APACyear {2020}}%
}]{%
friston2020sophisticated}
\APACinsertmetastar {%
friston2020sophisticated}%
\begin{APACrefauthors}%
Friston, K.%
, Da~Costa, L.%
, Hafner, D.%
, Hesp, C.%
\BCBL {}\ \BBA {} Parr, T.%
\end{APACrefauthors}%
\unskip\
\newblock
\APACrefYearMonthDay{2020}{}{}.
\newblock
{\BBOQ}\APACrefatitle {Sophisticated Inference} {Sophisticated
  inference}.{\BBCQ}
\newblock
\APACjournalVolNumPages{arXiv preprint arXiv:2006.04120}{}{}{}.
\newblock
\begin{APACrefURL} \url{https://arxiv.org/abs/2006.04120} \end{APACrefURL}
\PrintBackRefs{\CurrentBib}

\bibitem [\protect \citeauthoryear {%
Friston%
, Da~Costa%
\BCBL {}\ \BBA {} Parr%
}{%
Friston%
, Da~Costa%
\BCBL {}\ \BBA {} Parr%
}{%
{\protect \APACyear {2020}}%
}]{%
friston2020some}
\APACinsertmetastar {%
friston2020some}%
\begin{APACrefauthors}%
Friston, K.%
, Da~Costa, L.%
\BCBL {}\ \BBA {} Parr, T.%
\end{APACrefauthors}%
\unskip\
\newblock
\APACrefYearMonthDay{2020}{}{}.
\newblock
{\BBOQ}\APACrefatitle {Some interesting observations on the free energy
  principle} {Some interesting observations on the free energy
  principle}.{\BBCQ}
\newblock
\APACjournalVolNumPages{arXiv preprint arXiv:2002.04501}{}{}{}.
\PrintBackRefs{\CurrentBib}

\bibitem [\protect \citeauthoryear {%
Friston%
, Daunizeau%
\BCBL {}\ \BBA {} Kiebel%
}{%
Friston%
\ \protect \BOthers {.}}{%
{\protect \APACyear {2009}}%
}]{%
friston2009reinforcement}
\APACinsertmetastar {%
friston2009reinforcement}%
\begin{APACrefauthors}%
Friston, K.%
, Daunizeau, J.%
\BCBL {}\ \BBA {} Kiebel, S\BPBI J.%
\end{APACrefauthors}%
\unskip\
\newblock
\APACrefYearMonthDay{2009}{}{}.
\newblock
{\BBOQ}\APACrefatitle {Reinforcement learning or active inference?}
  {Reinforcement learning or active inference?}{\BBCQ}
\newblock
\APACjournalVolNumPages{PloS one}{4}{7}{}.
\PrintBackRefs{\CurrentBib}

\bibitem [\protect \citeauthoryear {%
Friston%
, Daunizeau%
, Kilner%
\BCBL {}\ \BBA {} Kiebel%
}{%
Friston%
, Daunizeau%
\BCBL {}\ \protect \BOthers {.}}{%
{\protect \APACyear {2010}}%
}]{%
friston2010action}
\APACinsertmetastar {%
friston2010action}%
\begin{APACrefauthors}%
Friston, K.%
, Daunizeau, J.%
, Kilner, J.%
\BCBL {}\ \BBA {} Kiebel, S\BPBI J.%
\end{APACrefauthors}%
\unskip\
\newblock
\APACrefYearMonthDay{2010}{}{}.
\newblock
{\BBOQ}\APACrefatitle {Action and behavior: a free-energy formulation} {Action
  and behavior: a free-energy formulation}.{\BBCQ}
\newblock
\APACjournalVolNumPages{Biological Cybernetics}{102}{3}{227--260}.
\newblock
\begin{APACrefURL}
  \url{https://link.springer.com/article/10.1007/s00422-010-0364-z}
  \end{APACrefURL}
\PrintBackRefs{\CurrentBib}

\bibitem [\protect \citeauthoryear {%
Friston%
\ \protect \BOthers {.}}{%
Friston%
\ \protect \BOthers {.}}{%
{\protect \APACyear {2007}}%
}]{%
friston2007parcels}
\APACinsertmetastar {%
friston2007parcels}%
\begin{APACrefauthors}%
Friston, K.%
, Fagerholm, E\BPBI D.%
, Zarghami, T\BPBI S.%
, Parr, T.%
, Hip{\'o}lito, I.%
, Magrou, L.%
\BCBL {}\ \BBA {} Razi, A.%
\end{APACrefauthors}%
\unskip\
\newblock
\APACrefYearMonthDay{2007}{}{}.
\newblock
{\BBOQ}\APACrefatitle {Parcels and particles: Markov blankets in the brain}
  {Parcels and particles: Markov blankets in the brain}.{\BBCQ}
\newblock
\APACjournalVolNumPages{Network Neuroscience}{}{Just Accepted}{1--76}.
\PrintBackRefs{\CurrentBib}

\bibitem [\protect \citeauthoryear {%
Friston%
, Fagerholm%
\BCBL {}\ \protect \BOthers {.}}{%
Friston%
, Fagerholm%
\BCBL {}\ \protect \BOthers {.}}{%
{\protect \APACyear {2020}}%
}]{%
friston2020parcels}
\APACinsertmetastar {%
friston2020parcels}%
\begin{APACrefauthors}%
Friston, K.%
, Fagerholm, E\BPBI D.%
, Zarghami, T\BPBI S.%
, Parr, T.%
, Hip{\'o}lito, I.%
, Magrou, L.%
\BCBL {}\ \BBA {} Razi, A.%
\end{APACrefauthors}%
\unskip\
\newblock
\APACrefYearMonthDay{2020}{}{}.
\newblock
{\BBOQ}\APACrefatitle {Parcels and particles: Markov blankets in the brain}
  {Parcels and particles: Markov blankets in the brain}.{\BBCQ}
\newblock
\APACjournalVolNumPages{arXiv preprint arXiv:2007.09704}{}{}{}.
\newblock
\begin{APACrefURL} \url{https://arxiv.org/abs/2007.09704} \end{APACrefURL}
\PrintBackRefs{\CurrentBib}

\bibitem [\protect \citeauthoryear {%
Friston%
, FitzGerald%
, Rigoli%
, Schwartenbeck%
\BCBL {}\ \BBA {} Pezzulo%
}{%
Friston%
, FitzGerald%
\BCBL {}\ \protect \BOthers {.}}{%
{\protect \APACyear {2017}}%
{\protect \APACexlab {{\protect \BCnt {1}}}}}]{%
friston2017active}
\APACinsertmetastar {%
friston2017active}%
\begin{APACrefauthors}%
Friston, K.%
, FitzGerald, T.%
, Rigoli, F.%
, Schwartenbeck, P.%
\BCBL {}\ \BBA {} Pezzulo, G.%
\end{APACrefauthors}%
\unskip\
\newblock
\APACrefYearMonthDay{2017{\protect \BCnt {1}}}{}{}.
\newblock
{\BBOQ}\APACrefatitle {Active inference: a process theory} {Active inference: a
  process theory}.{\BBCQ}
\newblock
\APACjournalVolNumPages{Neural Computation}{29}{1}{1--49}.
\PrintBackRefs{\CurrentBib}

\bibitem [\protect \citeauthoryear {%
Friston%
, FitzGerald%
, Rigoli%
, Schwartenbeck%
\BCBL {}\ \BBA {} Pezzulo%
}{%
Friston%
, FitzGerald%
\BCBL {}\ \protect \BOthers {.}}{%
{\protect \APACyear {2017}}%
{\protect \APACexlab {{\protect \BCnt {2}}}}}]{%
friston2017process}
\APACinsertmetastar {%
friston2017process}%
\begin{APACrefauthors}%
Friston, K.%
, FitzGerald, T.%
, Rigoli, F.%
, Schwartenbeck, P.%
\BCBL {}\ \BBA {} Pezzulo, G.%
\end{APACrefauthors}%
\unskip\
\newblock
\APACrefYearMonthDay{2017{\protect \BCnt {2}}}{}{}.
\newblock
{\BBOQ}\APACrefatitle {Active inference: a process theory} {Active inference: a
  process theory}.{\BBCQ}
\newblock
\APACjournalVolNumPages{Neural Computation}{29}{1}{1--49}.
\PrintBackRefs{\CurrentBib}

\bibitem [\protect \citeauthoryear {%
Friston%
\ \BBA {} Frith%
}{%
Friston%
\ \BBA {} Frith%
}{%
{\protect \APACyear {2015}}%
}]{%
friston2015duet}
\APACinsertmetastar {%
friston2015duet}%
\begin{APACrefauthors}%
Friston, K.%
\BCBT {}\ \BBA {} Frith, C.%
\end{APACrefauthors}%
\unskip\
\newblock
\APACrefYearMonthDay{2015}{}{}.
\newblock
{\BBOQ}\APACrefatitle {A duet for one} {A duet for one}.{\BBCQ}
\newblock
\APACjournalVolNumPages{Consciousness and Cognition}{36}{}{390--405}.
\PrintBackRefs{\CurrentBib}

\bibitem [\protect \citeauthoryear {%
Friston%
\ \BBA {} Kiebel%
}{%
Friston%
\ \BBA {} Kiebel%
}{%
{\protect \APACyear {2009}}%
}]{%
friston2009predictive}
\APACinsertmetastar {%
friston2009predictive}%
\begin{APACrefauthors}%
Friston, K.%
\BCBT {}\ \BBA {} Kiebel, S.%
\end{APACrefauthors}%
\unskip\
\newblock
\APACrefYearMonthDay{2009}{}{}.
\newblock
{\BBOQ}\APACrefatitle {Predictive coding under the free-energy principle}
  {Predictive coding under the free-energy principle}.{\BBCQ}
\newblock
\APACjournalVolNumPages{Philosophical Transactions of the Royal Society B:
  Biological Sciences}{364}{1521}{1211--1221}.
\PrintBackRefs{\CurrentBib}

\bibitem [\protect \citeauthoryear {%
Friston%
, Kilner%
\BCBL {}\ \BBA {} Harrison%
}{%
Friston%
\ \protect \BOthers {.}}{%
{\protect \APACyear {2006}}%
}]{%
friston2006free}
\APACinsertmetastar {%
friston2006free}%
\begin{APACrefauthors}%
Friston, K.%
, Kilner, J.%
\BCBL {}\ \BBA {} Harrison, L.%
\end{APACrefauthors}%
\unskip\
\newblock
\APACrefYearMonthDay{2006}{}{}.
\newblock
{\BBOQ}\APACrefatitle {A free energy principle for the brain} {A free energy
  principle for the brain}.{\BBCQ}
\newblock
\APACjournalVolNumPages{Journal of Physiology-Paris}{100}{1-3}{70--87}.
\PrintBackRefs{\CurrentBib}

\bibitem [\protect \citeauthoryear {%
Friston%
, Levin%
, Sengupta%
\BCBL {}\ \BBA {} Pezzulo%
}{%
Friston%
, Levin%
\BCBL {}\ \protect \BOthers {.}}{%
{\protect \APACyear {2015}}%
}]{%
friston2015knowing}
\APACinsertmetastar {%
friston2015knowing}%
\begin{APACrefauthors}%
Friston, K.%
, Levin, M.%
, Sengupta, B.%
\BCBL {}\ \BBA {} Pezzulo, G.%
\end{APACrefauthors}%
\unskip\
\newblock
\APACrefYearMonthDay{2015}{}{}.
\newblock
{\BBOQ}\APACrefatitle {Knowing one's place: a free-energy approach to pattern
  regulation} {Knowing one's place: a free-energy approach to pattern
  regulation}.{\BBCQ}
\newblock
\APACjournalVolNumPages{Journal of the Royal Society
  Interface}{12}{105}{20141383}.
\newblock
\begin{APACrefURL}
  \url{https://royalsocietypublishing.org/doi/full/10.1098/rsif.2014.1383}
  \end{APACrefURL}
\PrintBackRefs{\CurrentBib}

\bibitem [\protect \citeauthoryear {%
Friston%
, Lin%
\BCBL {}\ \protect \BOthers {.}}{%
Friston%
, Lin%
\BCBL {}\ \protect \BOthers {.}}{%
{\protect \APACyear {2017}}%
}]{%
friston2017curiosity}
\APACinsertmetastar {%
friston2017curiosity}%
\begin{APACrefauthors}%
Friston, K.%
, Lin, M.%
, Frith, C\BPBI D.%
, Pezzulo, G.%
, Hobson, J\BPBI A.%
\BCBL {}\ \BBA {} Ondobaka, S.%
\end{APACrefauthors}%
\unskip\
\newblock
\APACrefYearMonthDay{2017}{}{}.
\newblock
{\BBOQ}\APACrefatitle {Active inference, curiosity and insight} {Active
  inference, curiosity and insight}.{\BBCQ}
\newblock
\APACjournalVolNumPages{Neural Computation}{29}{10}{2633--2683}.
\PrintBackRefs{\CurrentBib}

\bibitem [\protect \citeauthoryear {%
Friston%
, Parr%
\BCBL {}\ \BBA {} Zeidman%
}{%
Friston%
, Parr%
\BCBL {}\ \BBA {} Zeidman%
}{%
{\protect \APACyear {2018}}%
}]{%
friston2018Bayesian}
\APACinsertmetastar {%
friston2018Bayesian}%
\begin{APACrefauthors}%
Friston, K.%
, Parr, T.%
\BCBL {}\ \BBA {} Zeidman, P.%
\end{APACrefauthors}%
\unskip\
\newblock
\APACrefYearMonthDay{2018}{}{}.
\newblock
{\BBOQ}\APACrefatitle {Bayesian model reduction} {Bayesian model
  reduction}.{\BBCQ}
\newblock
\APACjournalVolNumPages{arXiv preprint arXiv:1805.07092}{}{}{}.
\PrintBackRefs{\CurrentBib}

\bibitem [\protect \citeauthoryear {%
Friston%
, Rigoli%
\BCBL {}\ \protect \BOthers {.}}{%
Friston%
, Rigoli%
\BCBL {}\ \protect \BOthers {.}}{%
{\protect \APACyear {2015}}%
{\protect \APACexlab {{\protect \BCnt {1}}}}}]{%
friston2015active}
\APACinsertmetastar {%
friston2015active}%
\begin{APACrefauthors}%
Friston, K.%
, Rigoli, F.%
, Ognibene, D.%
, Mathys, C.%
, Fitzgerald, T.%
\BCBL {}\ \BBA {} Pezzulo, G.%
\end{APACrefauthors}%
\unskip\
\newblock
\APACrefYearMonthDay{2015{\protect \BCnt {1}}}{}{}.
\newblock
{\BBOQ}\APACrefatitle {Active inference and epistemic value} {Active inference
  and epistemic value}.{\BBCQ}
\newblock
\APACjournalVolNumPages{Cognitive Neuroscience}{6}{4}{187--214}.
\PrintBackRefs{\CurrentBib}

\bibitem [\protect \citeauthoryear {%
Friston%
, Rigoli%
\BCBL {}\ \protect \BOthers {.}}{%
Friston%
, Rigoli%
\BCBL {}\ \protect \BOthers {.}}{%
{\protect \APACyear {2015}}%
{\protect \APACexlab {{\protect \BCnt {2}}}}}]{%
friston_active_2015}
\APACinsertmetastar {%
friston_active_2015}%
\begin{APACrefauthors}%
Friston, K.%
, Rigoli, F.%
, Ognibene, D.%
, Mathys, C.%
, Fitzgerald, T.%
\BCBL {}\ \BBA {} Pezzulo, G.%
\end{APACrefauthors}%
\unskip\
\newblock
\APACrefYearMonthDay{2015{\protect \BCnt {2}}}{}{}.
\newblock
{\BBOQ}\APACrefatitle {Active inference and epistemic value} {Active inference
  and epistemic value}.{\BBCQ}
\newblock
\APACjournalVolNumPages{}{6}{4}{187--214}.
\newblock
\begin{APACrefDOI} \doi{10.1080/17588928.2015.1020053} \end{APACrefDOI}
\PrintBackRefs{\CurrentBib}

\bibitem [\protect \citeauthoryear {%
Friston%
, Rosch%
, Parr%
, Price%
\BCBL {}\ \BBA {} Bowman%
}{%
Friston%
, Rosch%
\BCBL {}\ \protect \BOthers {.}}{%
{\protect \APACyear {2018}}%
{\protect \APACexlab {{\protect \BCnt {1}}}}}]{%
friston2018deep}
\APACinsertmetastar {%
friston2018deep}%
\begin{APACrefauthors}%
Friston, K.%
, Rosch, R.%
, Parr, T.%
, Price, C.%
\BCBL {}\ \BBA {} Bowman, H.%
\end{APACrefauthors}%
\unskip\
\newblock
\APACrefYearMonthDay{2018{\protect \BCnt {1}}}{}{}.
\newblock
{\BBOQ}\APACrefatitle {Deep temporal models and active inference} {Deep
  temporal models and active inference}.{\BBCQ}
\newblock
\APACjournalVolNumPages{Neuroscience \& Biobehavioral Reviews}{90}{}{486--501}.
\PrintBackRefs{\CurrentBib}

\bibitem [\protect \citeauthoryear {%
Friston%
, Rosch%
, Parr%
, Price%
\BCBL {}\ \BBA {} Bowman%
}{%
Friston%
, Rosch%
\BCBL {}\ \protect \BOthers {.}}{%
{\protect \APACyear {2018}}%
{\protect \APACexlab {{\protect \BCnt {2}}}}}]{%
friston_deep_2018}
\APACinsertmetastar {%
friston_deep_2018}%
\begin{APACrefauthors}%
Friston, K.%
, Rosch, R.%
, Parr, T.%
, Price, C.%
\BCBL {}\ \BBA {} Bowman, H.%
\end{APACrefauthors}%
\unskip\
\newblock
\APACrefYearMonthDay{2018{\protect \BCnt {2}}}{}{}.
\newblock
{\BBOQ}\APACrefatitle {Deep temporal models and active inference} {Deep
  temporal models and active inference}.{\BBCQ}
\newblock
\APACjournalVolNumPages{}{90}{}{486--501}.
\newblock
\begin{APACrefURL}
  [{2019-11-15}]\url{http://www.sciencedirect.com/science/article/pii/S0149763418302525}
  \end{APACrefURL}
\newblock
\begin{APACrefDOI} \doi{10.1016/j.neubiorev.2018.04.004} \end{APACrefDOI}
\PrintBackRefs{\CurrentBib}

\bibitem [\protect \citeauthoryear {%
Friston%
, Samothrakis%
\BCBL {}\ \BBA {} Montague%
}{%
Friston%
\ \protect \BOthers {.}}{%
{\protect \APACyear {2012}}%
}]{%
friston2012active}
\APACinsertmetastar {%
friston2012active}%
\begin{APACrefauthors}%
Friston, K.%
, Samothrakis, S.%
\BCBL {}\ \BBA {} Montague, R.%
\end{APACrefauthors}%
\unskip\
\newblock
\APACrefYearMonthDay{2012}{}{}.
\newblock
{\BBOQ}\APACrefatitle {Active inference and agency: optimal control without
  cost functions} {Active inference and agency: optimal control without cost
  functions}.{\BBCQ}
\newblock
\APACjournalVolNumPages{Biological Cybernetics}{106}{8-9}{523--541}.
\PrintBackRefs{\CurrentBib}

\bibitem [\protect \citeauthoryear {%
Friston%
\ \protect \BOthers {.}}{%
Friston%
\ \protect \BOthers {.}}{%
{\protect \APACyear {2014}}%
}]{%
friston2014anatomy}
\APACinsertmetastar {%
friston2014anatomy}%
\begin{APACrefauthors}%
Friston, K.%
, Schwartenbeck, P.%
, FitzGerald, T.%
, Moutoussis, M.%
, Behrens, T.%
\BCBL {}\ \BBA {} Dolan, R\BPBI J.%
\end{APACrefauthors}%
\unskip\
\newblock
\APACrefYearMonthDay{2014}{}{}.
\newblock
{\BBOQ}\APACrefatitle {The anatomy of choice: dopamine and decision-making}
  {The anatomy of choice: dopamine and decision-making}.{\BBCQ}
\newblock
\APACjournalVolNumPages{Philosophical Transactions of the Royal Society B:
  Biological Sciences}{369}{1655}{20130481}.
\PrintBackRefs{\CurrentBib}

\bibitem [\protect \citeauthoryear {%
Friston%
, Stephan%
, Li%
\BCBL {}\ \BBA {} Daunizeau%
}{%
Friston%
, Stephan%
\BCBL {}\ \protect \BOthers {.}}{%
{\protect \APACyear {2010}}%
}]{%
friston2010generalised}
\APACinsertmetastar {%
friston2010generalised}%
\begin{APACrefauthors}%
Friston, K.%
, Stephan, K.%
, Li, B.%
\BCBL {}\ \BBA {} Daunizeau, J.%
\end{APACrefauthors}%
\unskip\
\newblock
\APACrefYearMonthDay{2010}{}{}.
\newblock
{\BBOQ}\APACrefatitle {Generalised filtering} {Generalised filtering}.{\BBCQ}
\newblock
\APACjournalVolNumPages{Mathematical Problems in Engineering}{2010}{}{}.
\newblock
\begin{APACrefURL} \url{https://www.hindawi.com/journals/mpe/2010/621670/}
  \end{APACrefURL}
\PrintBackRefs{\CurrentBib}

\bibitem [\protect \citeauthoryear {%
Friston%
\ \BBA {} Stephan%
}{%
Friston%
\ \BBA {} Stephan%
}{%
{\protect \APACyear {2007}}%
}]{%
friston2007free}
\APACinsertmetastar {%
friston2007free}%
\begin{APACrefauthors}%
Friston, K.%
\BCBT {}\ \BBA {} Stephan, K\BPBI E.%
\end{APACrefauthors}%
\unskip\
\newblock
\APACrefYearMonthDay{2007}{}{}.
\newblock
{\BBOQ}\APACrefatitle {Free-energy and the brain} {Free-energy and the
  brain}.{\BBCQ}
\newblock
\APACjournalVolNumPages{Synthese}{159}{3}{417--458}.
\PrintBackRefs{\CurrentBib}

\bibitem [\protect \citeauthoryear {%
Friston%
, Trujillo-Barreto%
\BCBL {}\ \BBA {} Daunizeau%
}{%
Friston%
\ \protect \BOthers {.}}{%
{\protect \APACyear {2008}}%
}]{%
friston2008DEM}
\APACinsertmetastar {%
friston2008DEM}%
\begin{APACrefauthors}%
Friston, K.%
, Trujillo-Barreto, N.%
\BCBL {}\ \BBA {} Daunizeau, J.%
\end{APACrefauthors}%
\unskip\
\newblock
\APACrefYearMonthDay{2008}{}{}.
\newblock
{\BBOQ}\APACrefatitle {DEM: a variational treatment of dynamic systems} {Dem: a
  variational treatment of dynamic systems}.{\BBCQ}
\newblock
\APACjournalVolNumPages{Neuroimage}{41}{3}{849--885}.
\PrintBackRefs{\CurrentBib}

\bibitem [\protect \citeauthoryear {%
Friston%
, Wiese%
\BCBL {}\ \BBA {} Hobson%
}{%
Friston%
, Wiese%
\BCBL {}\ \BBA {} Hobson%
}{%
{\protect \APACyear {2020}}%
}]{%
friston2020sentience}
\APACinsertmetastar {%
friston2020sentience}%
\begin{APACrefauthors}%
Friston, K.%
, Wiese, W.%
\BCBL {}\ \BBA {} Hobson, J\BPBI A.%
\end{APACrefauthors}%
\unskip\
\newblock
\APACrefYearMonthDay{2020}{}{}.
\newblock
{\BBOQ}\APACrefatitle {Sentience and the origins of consciousness: From
  Cartesian duality to Markovian monism} {Sentience and the origins of
  consciousness: From cartesian duality to markovian monism}.{\BBCQ}
\newblock
\APACjournalVolNumPages{Entropy}{22}{5}{516}.
\PrintBackRefs{\CurrentBib}

\bibitem [\protect \citeauthoryear {%
Fujimoto%
, van Hoof%
\BCBL {}\ \BBA {} Meger%
}{%
Fujimoto%
\ \protect \BOthers {.}}{%
{\protect \APACyear {2018}}%
}]{%
fujimoto2018addressing}
\APACinsertmetastar {%
fujimoto2018addressing}%
\begin{APACrefauthors}%
Fujimoto, S.%
, van Hoof, H.%
\BCBL {}\ \BBA {} Meger, D.%
\end{APACrefauthors}%
\unskip\
\newblock
\APACrefYearMonthDay{2018}{}{}.
\newblock
{\BBOQ}\APACrefatitle {Addressing function approximation error in actor-critic
  methods} {Addressing function approximation error in actor-critic
  methods}.{\BBCQ}
\newblock
\APACjournalVolNumPages{arXiv preprint arXiv:1802.09477}{}{}{}.
\PrintBackRefs{\CurrentBib}

\bibitem [\protect \citeauthoryear {%
Gaissmaier%
\ \BBA {} Schooler%
}{%
Gaissmaier%
\ \BBA {} Schooler%
}{%
{\protect \APACyear {2008}}%
}]{%
gaissmaier2008smart}
\APACinsertmetastar {%
gaissmaier2008smart}%
\begin{APACrefauthors}%
Gaissmaier, W.%
\BCBT {}\ \BBA {} Schooler, L\BPBI J.%
\end{APACrefauthors}%
\unskip\
\newblock
\APACrefYearMonthDay{2008}{}{}.
\newblock
{\BBOQ}\APACrefatitle {The smart potential behind probability matching} {The
  smart potential behind probability matching}.{\BBCQ}
\newblock
\APACjournalVolNumPages{Cognition}{109}{3}{416--422}.
\PrintBackRefs{\CurrentBib}

\bibitem [\protect \citeauthoryear {%
Gal%
, McAllister%
\BCBL {}\ \BBA {} Rasmussen%
}{%
Gal%
\ \protect \BOthers {.}}{%
{\protect \APACyear {2016}}%
}]{%
gal2016improving}
\APACinsertmetastar {%
gal2016improving}%
\begin{APACrefauthors}%
Gal, Y.%
, McAllister, R.%
\BCBL {}\ \BBA {} Rasmussen, C\BPBI E.%
\end{APACrefauthors}%
\unskip\
\newblock
\APACrefYearMonthDay{2016}{}{}.
\newblock
{\BBOQ}\APACrefatitle {Improving PILCO with Bayesian neural network dynamics
  models} {Improving pilco with bayesian neural network dynamics
  models}.{\BBCQ}
\newblock
\BIn{} \APACrefbtitle {Data-Efficient Machine Learning workshop, ICML}
  {Data-efficient machine learning workshop, icml}\ (\BVOL~4, \BPG~25).
\PrintBackRefs{\CurrentBib}

\bibitem [\protect \citeauthoryear {%
Garivier%
\ \BBA {} Moulines%
}{%
Garivier%
\ \BBA {} Moulines%
}{%
{\protect \APACyear {2011}}%
}]{%
garivier2011upper}
\APACinsertmetastar {%
garivier2011upper}%
\begin{APACrefauthors}%
Garivier, A.%
\BCBT {}\ \BBA {} Moulines, E.%
\end{APACrefauthors}%
\unskip\
\newblock
\APACrefYearMonthDay{2011}{}{}.
\newblock
{\BBOQ}\APACrefatitle {On upper-confidence bound policies for switching bandit
  problems} {On upper-confidence bound policies for switching bandit
  problems}.{\BBCQ}
\newblock
\BIn{} \APACrefbtitle {International Conference on Algorithmic Learning Theory}
  {International conference on algorithmic learning theory}\ (\BPGS\ 174--188).
\PrintBackRefs{\CurrentBib}

\bibitem [\protect \citeauthoryear {%
Gershman%
}{%
Gershman%
}{%
{\protect \APACyear {2018}}%
}]{%
gershman2018uncertainty}
\APACinsertmetastar {%
gershman2018uncertainty}%
\begin{APACrefauthors}%
Gershman, S\BPBI J.%
\end{APACrefauthors}%
\unskip\
\newblock
\APACrefYearMonthDay{2018}{}{}.
\newblock
{\BBOQ}\APACrefatitle {Uncertainty and exploration} {Uncertainty and
  exploration}.{\BBCQ}
\newblock
\APACjournalVolNumPages{bioRxiv}{}{}{265504}.
\PrintBackRefs{\CurrentBib}

\bibitem [\protect \citeauthoryear {%
Gerstner%
\ \BBA {} Kistler%
}{%
Gerstner%
\ \BBA {} Kistler%
}{%
{\protect \APACyear {2002}}%
}]{%
gerstner2002mathematical}
\APACinsertmetastar {%
gerstner2002mathematical}%
\begin{APACrefauthors}%
Gerstner, W.%
\BCBT {}\ \BBA {} Kistler, W\BPBI M.%
\end{APACrefauthors}%
\unskip\
\newblock
\APACrefYearMonthDay{2002}{}{}.
\newblock
{\BBOQ}\APACrefatitle {Mathematical formulations of Hebbian learning}
  {Mathematical formulations of hebbian learning}.{\BBCQ}
\newblock
\APACjournalVolNumPages{Biological Cybernetics}{87}{5}{404--415}.
\PrintBackRefs{\CurrentBib}

\bibitem [\protect \citeauthoryear {%
Geweke%
}{%
Geweke%
}{%
{\protect \APACyear {2007}}%
}]{%
geweke2007Bayesian}
\APACinsertmetastar {%
geweke2007Bayesian}%
\begin{APACrefauthors}%
Geweke, J.%
\end{APACrefauthors}%
\unskip\
\newblock
\APACrefYearMonthDay{2007}{}{}.
\newblock
{\BBOQ}\APACrefatitle {Bayesian model comparison and validation} {Bayesian
  model comparison and validation}.{\BBCQ}
\newblock
\APACjournalVolNumPages{American Economic Review}{97}{2}{60--64}.
\PrintBackRefs{\CurrentBib}

\bibitem [\protect \citeauthoryear {%
Ghahramani%
\ \BBA {} Beal%
}{%
Ghahramani%
\ \BBA {} Beal%
}{%
{\protect \APACyear {2001}}%
}]{%
ghahramani2001propagation}
\APACinsertmetastar {%
ghahramani2001propagation}%
\begin{APACrefauthors}%
Ghahramani, Z.%
\BCBT {}\ \BBA {} Beal, M\BPBI J.%
\end{APACrefauthors}%
\unskip\
\newblock
\APACrefYearMonthDay{2001}{}{}.
\newblock
{\BBOQ}\APACrefatitle {Propagation algorithms for variational Bayesian
  learning} {Propagation algorithms for variational bayesian learning}.{\BBCQ}
\newblock
\BIn{} \APACrefbtitle {Advances in neural Information Processing Systems}
  {Advances in neural information processing systems}\ (\BPGS\ 507--513).
\PrintBackRefs{\CurrentBib}

\bibitem [\protect \citeauthoryear {%
Ghahramani%
, Beal%
\BCBL {}\ \protect \BOthers {.}}{%
Ghahramani%
\ \protect \BOthers {.}}{%
{\protect \APACyear {2000}}%
}]{%
ghahramani2000graphical}
\APACinsertmetastar {%
ghahramani2000graphical}%
\begin{APACrefauthors}%
Ghahramani, Z.%
, Beal, M\BPBI J.%
\BCBL {}\ \BOthersPeriod {.}\end{APACrefauthors}%
\unskip\
\newblock
\APACrefYear{2000}.
\newblock
\APACrefbtitle {Graphical models and variational methods} {Graphical models and
  variational methods}.
\newblock
\APACaddressPublisher{}{Advanced mean field methods-theory and practice. MIT
  Press}.
\PrintBackRefs{\CurrentBib}

\bibitem [\protect \citeauthoryear {%
Gibson%
}{%
Gibson%
}{%
{\protect \APACyear {2002}}%
}]{%
gibson2002theory}
\APACinsertmetastar {%
gibson2002theory}%
\begin{APACrefauthors}%
Gibson, J\BPBI J.%
\end{APACrefauthors}%
\unskip\
\newblock
\APACrefYearMonthDay{2002}{}{}.
\newblock
{\BBOQ}\APACrefatitle {A theory of direct visual perception} {A theory of
  direct visual perception}.{\BBCQ}
\newblock
\APACjournalVolNumPages{Vision and Mind: selected readings in the philosophy of
  perception}{}{}{77--90}.
\PrintBackRefs{\CurrentBib}

\bibitem [\protect \citeauthoryear {%
Girolami%
\ \BBA {} Calderhead%
}{%
Girolami%
\ \BBA {} Calderhead%
}{%
{\protect \APACyear {2011}}%
}]{%
girolami2011riemann}
\APACinsertmetastar {%
girolami2011riemann}%
\begin{APACrefauthors}%
Girolami, M.%
\BCBT {}\ \BBA {} Calderhead, B.%
\end{APACrefauthors}%
\unskip\
\newblock
\APACrefYearMonthDay{2011}{}{}.
\newblock
{\BBOQ}\APACrefatitle {Riemann manifold langevin and hamiltonian monte carlo
  methods} {Riemann manifold langevin and hamiltonian monte carlo
  methods}.{\BBCQ}
\newblock
\APACjournalVolNumPages{Journal of the Royal Statistical Society: Series B
  (Statistical Methodology)}{73}{2}{123--214}.
\PrintBackRefs{\CurrentBib}

\bibitem [\protect \citeauthoryear {%
Gold%
\ \BBA {} Shadlen%
}{%
Gold%
\ \BBA {} Shadlen%
}{%
{\protect \APACyear {2003}}%
}]{%
gold2003influence}
\APACinsertmetastar {%
gold2003influence}%
\begin{APACrefauthors}%
Gold, J\BPBI I.%
\BCBT {}\ \BBA {} Shadlen, M\BPBI N.%
\end{APACrefauthors}%
\unskip\
\newblock
\APACrefYearMonthDay{2003}{}{}.
\newblock
{\BBOQ}\APACrefatitle {The influence of behavioral context on the
  representation of a perceptual decision in developing oculomotor commands}
  {The influence of behavioral context on the representation of a perceptual
  decision in developing oculomotor commands}.{\BBCQ}
\newblock
\APACjournalVolNumPages{Journal of Neuroscience}{23}{2}{632--651}.
\PrintBackRefs{\CurrentBib}

\bibitem [\protect \citeauthoryear {%
Goodfellow%
, Bengio%
\BCBL {}\ \BBA {} Courville%
}{%
Goodfellow%
\ \protect \BOthers {.}}{%
{\protect \APACyear {2016}}%
}]{%
goodfellow2016deep}
\APACinsertmetastar {%
goodfellow2016deep}%
\begin{APACrefauthors}%
Goodfellow, I.%
, Bengio, Y.%
\BCBL {}\ \BBA {} Courville, A.%
\end{APACrefauthors}%
\unskip\
\newblock
\APACrefYear{2016}.
\newblock
\APACrefbtitle {Deep learning} {Deep learning}.
\newblock
\APACaddressPublisher{}{MIT press}.
\PrintBackRefs{\CurrentBib}

\bibitem [\protect \citeauthoryear {%
Goodfellow%
\ \protect \BOthers {.}}{%
Goodfellow%
\ \protect \BOthers {.}}{%
{\protect \APACyear {2014}}%
}]{%
goodfellow2014generative}
\APACinsertmetastar {%
goodfellow2014generative}%
\begin{APACrefauthors}%
Goodfellow, I.%
, Pouget-Abadie, J.%
, Mirza, M.%
, Xu, B.%
, Warde-Farley, D.%
, Ozair, S.%
\BDBL {}Bengio, Y.%
\end{APACrefauthors}%
\unskip\
\newblock
\APACrefYearMonthDay{2014}{}{}.
\newblock
{\BBOQ}\APACrefatitle {Generative adversarial nets} {Generative adversarial
  nets}.{\BBCQ}
\newblock
\APACjournalVolNumPages{Advances in neural Information Processing
  Systems}{27}{}{2672--2680}.
\PrintBackRefs{\CurrentBib}

\bibitem [\protect \citeauthoryear {%
G\BPBI J.~Gordon%
}{%
G\BPBI J.~Gordon%
}{%
{\protect \APACyear {1995}}%
}]{%
gordon1995stable}
\APACinsertmetastar {%
gordon1995stable}%
\begin{APACrefauthors}%
Gordon, G\BPBI J.%
\end{APACrefauthors}%
\unskip\
\newblock
\APACrefYearMonthDay{1995}{}{}.
\newblock
{\BBOQ}\APACrefatitle {Stable function approximation in dynamic programming}
  {Stable function approximation in dynamic programming}.{\BBCQ}
\newblock
\BIn{} \APACrefbtitle {Machine Learning Proceedings 1995} {Machine learning
  proceedings 1995}\ (\BPGS\ 261--268).
\newblock
\APACaddressPublisher{}{Elsevier}.
\PrintBackRefs{\CurrentBib}

\bibitem [\protect \citeauthoryear {%
N\BPBI J.~Gordon%
, Salmond%
\BCBL {}\ \BBA {} Smith%
}{%
N\BPBI J.~Gordon%
\ \protect \BOthers {.}}{%
{\protect \APACyear {1993}}%
}]{%
gordon1993novel}
\APACinsertmetastar {%
gordon1993novel}%
\begin{APACrefauthors}%
Gordon, N\BPBI J.%
, Salmond, D\BPBI J.%
\BCBL {}\ \BBA {} Smith, A\BPBI F.%
\end{APACrefauthors}%
\unskip\
\newblock
\APACrefYearMonthDay{1993}{}{}.
\newblock
{\BBOQ}\APACrefatitle {Novel approach to nonlinear/non-Gaussian Bayesian state
  estimation} {Novel approach to nonlinear/non-gaussian bayesian state
  estimation}.{\BBCQ}
\newblock
\BIn{} \APACrefbtitle {IEEE proceedings F (Radar and Signal Processing)} {Ieee
  proceedings f (radar and signal processing)}\ (\BVOL~140, \BPGS\ 107--113).
\PrintBackRefs{\CurrentBib}

\bibitem [\protect \citeauthoryear {%
Grewal%
\ \BBA {} Andrews%
}{%
Grewal%
\ \BBA {} Andrews%
}{%
{\protect \APACyear {2010}}%
}]{%
grewal2010applications}
\APACinsertmetastar {%
grewal2010applications}%
\begin{APACrefauthors}%
Grewal, M\BPBI S.%
\BCBT {}\ \BBA {} Andrews, A\BPBI P.%
\end{APACrefauthors}%
\unskip\
\newblock
\APACrefYearMonthDay{2010}{}{}.
\newblock
{\BBOQ}\APACrefatitle {Applications of Kalman filtering in aerospace 1960 to
  the present [historical perspectives]} {Applications of kalman filtering in
  aerospace 1960 to the present [historical perspectives]}.{\BBCQ}
\newblock
\APACjournalVolNumPages{IEEE Control Systems Magazine}{30}{3}{69--78}.
\PrintBackRefs{\CurrentBib}

\bibitem [\protect \citeauthoryear {%
Griewank%
\ \protect \BOthers {.}}{%
Griewank%
\ \protect \BOthers {.}}{%
{\protect \APACyear {1989}}%
}]{%
griewank1989automatic}
\APACinsertmetastar {%
griewank1989automatic}%
\begin{APACrefauthors}%
Griewank, A.%
\BCBT {}\ \BOthersPeriod {.}
\end{APACrefauthors}%
\unskip\
\newblock
\APACrefYearMonthDay{1989}{}{}.
\newblock
{\BBOQ}\APACrefatitle {On automatic differentiation} {On automatic
  differentiation}.{\BBCQ}
\newblock
\APACjournalVolNumPages{Mathematical Programming: recent developments and
  applications}{6}{6}{83--107}.
\PrintBackRefs{\CurrentBib}

\bibitem [\protect \citeauthoryear {%
Grill-Spector%
\ \BBA {} Malach%
}{%
Grill-Spector%
\ \BBA {} Malach%
}{%
{\protect \APACyear {2004}}%
}]{%
grill2004human}
\APACinsertmetastar {%
grill2004human}%
\begin{APACrefauthors}%
Grill-Spector, K.%
\BCBT {}\ \BBA {} Malach, R.%
\end{APACrefauthors}%
\unskip\
\newblock
\APACrefYearMonthDay{2004}{}{}.
\newblock
{\BBOQ}\APACrefatitle {The human visual cortex} {The human visual
  cortex}.{\BBCQ}
\newblock
\APACjournalVolNumPages{Annu. Rev. Neurosci.}{27}{}{649--677}.
\PrintBackRefs{\CurrentBib}

\bibitem [\protect \citeauthoryear {%
Gu%
, Lillicrap%
, Sutskever%
\BCBL {}\ \BBA {} Levine%
}{%
Gu%
\ \protect \BOthers {.}}{%
{\protect \APACyear {2016}}%
}]{%
gu2016continuous}
\APACinsertmetastar {%
gu2016continuous}%
\begin{APACrefauthors}%
Gu, S.%
, Lillicrap, T.%
, Sutskever, I.%
\BCBL {}\ \BBA {} Levine, S.%
\end{APACrefauthors}%
\unskip\
\newblock
\APACrefYearMonthDay{2016}{}{}.
\newblock
{\BBOQ}\APACrefatitle {Continuous deep q-learning with model-based
  acceleration} {Continuous deep q-learning with model-based
  acceleration}.{\BBCQ}
\newblock
\BIn{} \APACrefbtitle {International Conference on Machine Learning}
  {International conference on machine learning}\ (\BPGS\ 2829--2838).
\PrintBackRefs{\CurrentBib}

\bibitem [\protect \citeauthoryear {%
Gupta%
, Agrawal%
, Gopalakrishnan%
\BCBL {}\ \BBA {} Narayanan%
}{%
Gupta%
\ \protect \BOthers {.}}{%
{\protect \APACyear {2015}}%
}]{%
gupta2015deep}
\APACinsertmetastar {%
gupta2015deep}%
\begin{APACrefauthors}%
Gupta, S.%
, Agrawal, A.%
, Gopalakrishnan, K.%
\BCBL {}\ \BBA {} Narayanan, P.%
\end{APACrefauthors}%
\unskip\
\newblock
\APACrefYearMonthDay{2015}{}{}.
\newblock
{\BBOQ}\APACrefatitle {Deep learning with limited numerical precision} {Deep
  learning with limited numerical precision}.{\BBCQ}
\newblock
\BIn{} \APACrefbtitle {International Conference on Machine Learning}
  {International conference on machine learning}\ (\BPGS\ 1737--1746).
\PrintBackRefs{\CurrentBib}

\bibitem [\protect \citeauthoryear {%
Ha%
\ \BBA {} Schmidhuber%
}{%
Ha%
\ \BBA {} Schmidhuber%
}{%
{\protect \APACyear {2018}}%
}]{%
ha_recurrent_2018}
\APACinsertmetastar {%
ha_recurrent_2018}%
\begin{APACrefauthors}%
Ha, D.%
\BCBT {}\ \BBA {} Schmidhuber, J.%
\end{APACrefauthors}%
\unskip\
\newblock
\APACrefYearMonthDay{2018}{}{}.
\newblock
{\BBOQ}\APACrefatitle {World models} {World models}.{\BBCQ}
\newblock
\APACjournalVolNumPages{arXiv preprint arXiv:1803.10122}{}{}{}.
\PrintBackRefs{\CurrentBib}

\bibitem [\protect \citeauthoryear {%
Haarnoja%
}{%
Haarnoja%
}{%
{\protect \APACyear {2018}}%
}]{%
haarnoja2018acquiring}
\APACinsertmetastar {%
haarnoja2018acquiring}%
\begin{APACrefauthors}%
Haarnoja, T.%
\end{APACrefauthors}%
\unskip\
\newblock
\APACrefYear{2018}.
\unskip\
\newblock
\APACrefbtitle {Acquiring Diverse Robot Skills via Maximum Entropy Deep
  Reinforcement Learning} {Acquiring diverse robot skills via maximum entropy
  deep reinforcement learning}\ \APACtypeAddressSchool {\BUPhD}{}{}.
\unskip\
\newblock
\APACaddressSchool {}{UC Berkeley}.
\PrintBackRefs{\CurrentBib}

\bibitem [\protect \citeauthoryear {%
Haarnoja%
, Tang%
, Abbeel%
\BCBL {}\ \BBA {} Levine%
}{%
Haarnoja%
\ \protect \BOthers {.}}{%
{\protect \APACyear {2017}}%
}]{%
haarnoja2017reinforcement}
\APACinsertmetastar {%
haarnoja2017reinforcement}%
\begin{APACrefauthors}%
Haarnoja, T.%
, Tang, H.%
, Abbeel, P.%
\BCBL {}\ \BBA {} Levine, S.%
\end{APACrefauthors}%
\unskip\
\newblock
\APACrefYearMonthDay{2017}{}{}.
\newblock
{\BBOQ}\APACrefatitle {Reinforcement learning with deep energy-based policies}
  {Reinforcement learning with deep energy-based policies}.{\BBCQ}
\newblock
\BIn{} \APACrefbtitle {Proceedings of the 34th International Conference on
  Machine Learning-Volume 70} {Proceedings of the 34th international conference
  on machine learning-volume 70}\ (\BPGS\ 1352--1361).
\PrintBackRefs{\CurrentBib}

\bibitem [\protect \citeauthoryear {%
Haarnoja%
, Zhou%
, Abbeel%
\BCBL {}\ \BBA {} Levine%
}{%
Haarnoja%
, Zhou%
, Abbeel%
\BCBL {}\ \BBA {} Levine%
}{%
{\protect \APACyear {2018}}%
}]{%
haarnoja2018soft}
\APACinsertmetastar {%
haarnoja2018soft}%
\begin{APACrefauthors}%
Haarnoja, T.%
, Zhou, A.%
, Abbeel, P.%
\BCBL {}\ \BBA {} Levine, S.%
\end{APACrefauthors}%
\unskip\
\newblock
\APACrefYearMonthDay{2018}{}{}.
\newblock
{\BBOQ}\APACrefatitle {Soft actor-critic: Off-policy maximum entropy deep
  reinforcement learning with a stochastic actor} {Soft actor-critic:
  Off-policy maximum entropy deep reinforcement learning with a stochastic
  actor}.{\BBCQ}
\newblock
\APACjournalVolNumPages{arXiv preprint arXiv:1801.01290}{}{}{}.
\PrintBackRefs{\CurrentBib}

\bibitem [\protect \citeauthoryear {%
Haarnoja%
, Zhou%
, Hartikainen%
\BCBL {}\ \protect \BOthers {.}}{%
Haarnoja%
, Zhou%
, Hartikainen%
\BCBL {}\ \protect \BOthers {.}}{%
{\protect \APACyear {2018}}%
}]{%
haarnoja2018applications}
\APACinsertmetastar {%
haarnoja2018applications}%
\begin{APACrefauthors}%
Haarnoja, T.%
, Zhou, A.%
, Hartikainen, K.%
, Tucker, G.%
, Ha, S.%
, Tan, J.%
\BDBL {}others%
\end{APACrefauthors}%
\unskip\
\newblock
\APACrefYearMonthDay{2018}{}{}.
\newblock
{\BBOQ}\APACrefatitle {Soft actor-critic algorithms and applications} {Soft
  actor-critic algorithms and applications}.{\BBCQ}
\newblock
\APACjournalVolNumPages{arXiv preprint arXiv:1812.05905}{}{}{}.
\PrintBackRefs{\CurrentBib}

\bibitem [\protect \citeauthoryear {%
Hafner%
, Lillicrap%
, Ba%
\BCBL {}\ \BBA {} Norouzi%
}{%
Hafner%
\ \protect \BOthers {.}}{%
{\protect \APACyear {2019}}%
}]{%
hafner2019dream}
\APACinsertmetastar {%
hafner2019dream}%
\begin{APACrefauthors}%
Hafner, D.%
, Lillicrap, T.%
, Ba, J.%
\BCBL {}\ \BBA {} Norouzi, M.%
\end{APACrefauthors}%
\unskip\
\newblock
\APACrefYearMonthDay{2019}{}{}.
\newblock
{\BBOQ}\APACrefatitle {Dream to Control: Learning Behaviors by Latent
  Imagination} {Dream to control: Learning behaviors by latent
  imagination}.{\BBCQ}
\newblock
\APACjournalVolNumPages{arXiv preprint arXiv:1912.01603}{}{}{}.
\PrintBackRefs{\CurrentBib}

\bibitem [\protect \citeauthoryear {%
Hafner%
\ \protect \BOthers {.}}{%
Hafner%
\ \protect \BOthers {.}}{%
{\protect \APACyear {2018}}%
}]{%
hafner2018learning}
\APACinsertmetastar {%
hafner2018learning}%
\begin{APACrefauthors}%
Hafner, D.%
, Lillicrap, T.%
, Fischer, I.%
, Villegas, R.%
, Ha, D.%
, Lee, H.%
\BCBL {}\ \BBA {} Davidson, J.%
\end{APACrefauthors}%
\unskip\
\newblock
\APACrefYearMonthDay{2018}{}{}.
\newblock
{\BBOQ}\APACrefatitle {Learning latent dynamics for planning from pixels}
  {Learning latent dynamics for planning from pixels}.{\BBCQ}
\newblock
\APACjournalVolNumPages{arXiv preprint arXiv:1811.04551}{}{}{}.
\PrintBackRefs{\CurrentBib}

\bibitem [\protect \citeauthoryear {%
Hafner%
\ \protect \BOthers {.}}{%
Hafner%
\ \protect \BOthers {.}}{%
{\protect \APACyear {2020}}%
}]{%
hafner2020action}
\APACinsertmetastar {%
hafner2020action}%
\begin{APACrefauthors}%
Hafner, D.%
, Ortega, P\BPBI A.%
, Ba, J.%
, Parr, T.%
, Friston, K.%
\BCBL {}\ \BBA {} Heess, N.%
\end{APACrefauthors}%
\unskip\
\newblock
\APACrefYearMonthDay{2020}{}{}.
\newblock
{\BBOQ}\APACrefatitle {Action and perception as divergence minimization}
  {Action and perception as divergence minimization}.{\BBCQ}
\newblock
\APACjournalVolNumPages{arXiv preprint arXiv:2009.01791}{}{}{}.
\PrintBackRefs{\CurrentBib}

\bibitem [\protect \citeauthoryear {%
Harvey%
}{%
Harvey%
}{%
{\protect \APACyear {1990}}%
}]{%
harvey1990forecasting}
\APACinsertmetastar {%
harvey1990forecasting}%
\begin{APACrefauthors}%
Harvey, A\BPBI C.%
\end{APACrefauthors}%
\unskip\
\newblock
\APACrefYear{1990}.
\newblock
\APACrefbtitle {Forecasting, structural time series models and the Kalman
  filter} {Forecasting, structural time series models and the kalman filter}.
\newblock
\APACaddressPublisher{}{Cambridge university press}.
\PrintBackRefs{\CurrentBib}

\bibitem [\protect \citeauthoryear {%
Hawkins%
\ \BBA {} Blakeslee%
}{%
Hawkins%
\ \BBA {} Blakeslee%
}{%
{\protect \APACyear {2007}}%
}]{%
hawkins2007intelligence}
\APACinsertmetastar {%
hawkins2007intelligence}%
\begin{APACrefauthors}%
Hawkins, J.%
\BCBT {}\ \BBA {} Blakeslee, S.%
\end{APACrefauthors}%
\unskip\
\newblock
\APACrefYear{2007}.
\newblock
\APACrefbtitle {On intelligence: How a new understanding of the brain will lead
  to the creation of truly intelligent machines} {On intelligence: How a new
  understanding of the brain will lead to the creation of truly intelligent
  machines}.
\newblock
\APACaddressPublisher{}{Macmillan}.
\PrintBackRefs{\CurrentBib}

\bibitem [\protect \citeauthoryear {%
He%
, Zhang%
, Ren%
\BCBL {}\ \BBA {} Sun%
}{%
He%
\ \protect \BOthers {.}}{%
{\protect \APACyear {2016}}%
}]{%
he2016deep}
\APACinsertmetastar {%
he2016deep}%
\begin{APACrefauthors}%
He, K.%
, Zhang, X.%
, Ren, S.%
\BCBL {}\ \BBA {} Sun, J.%
\end{APACrefauthors}%
\unskip\
\newblock
\APACrefYearMonthDay{2016}{}{}.
\newblock
{\BBOQ}\APACrefatitle {Deep residual learning for image recognition} {Deep
  residual learning for image recognition}.{\BBCQ}
\newblock
\BIn{} \APACrefbtitle {Proceedings of the IEEE conference on computer vision
  and pattern recognition} {Proceedings of the ieee conference on computer
  vision and pattern recognition}\ (\BPGS\ 770--778).
\PrintBackRefs{\CurrentBib}

\bibitem [\protect \citeauthoryear {%
Hebb%
}{%
Hebb%
}{%
{\protect \APACyear {1949}}%
}]{%
hebb1949first}
\APACinsertmetastar {%
hebb1949first}%
\begin{APACrefauthors}%
Hebb, D\BPBI O.%
\end{APACrefauthors}%
\unskip\
\newblock
\APACrefYearMonthDay{1949}{}{}.
\newblock
{\BBOQ}\APACrefatitle {The first stage of perception: growth of the assembly}
  {The first stage of perception: growth of the assembly}.{\BBCQ}
\newblock
\APACjournalVolNumPages{The Organization of Behavior}{4}{}{60--78}.
\PrintBackRefs{\CurrentBib}

\bibitem [\protect \citeauthoryear {%
Heins%
\ \protect \BOthers {.}}{%
Heins%
\ \protect \BOthers {.}}{%
{\protect \APACyear {2020}}%
}]{%
heins2020deep}
\APACinsertmetastar {%
heins2020deep}%
\begin{APACrefauthors}%
Heins, R\BPBI C.%
, Mirza, M\BPBI B.%
, Parr, T.%
, Friston, K.%
, Kagan, I.%
\BCBL {}\ \BBA {} Pooresmaeili, A.%
\end{APACrefauthors}%
\unskip\
\newblock
\APACrefYearMonthDay{2020}{}{}.
\newblock
{\BBOQ}\APACrefatitle {Deep Active Inference and Scene Construction} {Deep
  active inference and scene construction}.{\BBCQ}
\newblock
\APACjournalVolNumPages{Frontiers in Artificial Intelligence}{3}{}{81}.
\PrintBackRefs{\CurrentBib}

\bibitem [\protect \citeauthoryear {%
Helmholtz%
}{%
Helmholtz%
}{%
{\protect \APACyear {1866}}%
}]{%
helmholtz1866concerning}
\APACinsertmetastar {%
helmholtz1866concerning}%
\begin{APACrefauthors}%
Helmholtz, H\BPBI v.%
\end{APACrefauthors}%
\unskip\
\newblock
\APACrefYearMonthDay{1866}{}{}.
\newblock
{\BBOQ}\APACrefatitle {Concerning the perceptions in general} {Concerning the
  perceptions in general}.{\BBCQ}
\newblock
\APACjournalVolNumPages{Treatise on physiological optics,}{}{}{}.
\PrintBackRefs{\CurrentBib}

\bibitem [\protect \citeauthoryear {%
Henderson%
}{%
Henderson%
}{%
{\protect \APACyear {2017}}%
}]{%
henderson2017gaze}
\APACinsertmetastar {%
henderson2017gaze}%
\begin{APACrefauthors}%
Henderson, J\BPBI M.%
\end{APACrefauthors}%
\unskip\
\newblock
\APACrefYearMonthDay{2017}{}{}.
\newblock
{\BBOQ}\APACrefatitle {Gaze control as prediction} {Gaze control as
  prediction}.{\BBCQ}
\newblock
\APACjournalVolNumPages{Trends in Cognitive Sciences}{21}{1}{15--23}.
\PrintBackRefs{\CurrentBib}

\bibitem [\protect \citeauthoryear {%
Hesp%
\ \protect \BOthers {.}}{%
Hesp%
\ \protect \BOthers {.}}{%
{\protect \APACyear {2020}}%
}]{%
hesp2020sophisticated}
\APACinsertmetastar {%
hesp2020sophisticated}%
\begin{APACrefauthors}%
Hesp, C.%
, Tschantz, A.%
, Millidge, B.%
, Ramstead, M.%
, Friston, K.%
\BCBL {}\ \BBA {} Smith, R.%
\end{APACrefauthors}%
\unskip\
\newblock
\APACrefYearMonthDay{2020}{}{}.
\newblock
{\BBOQ}\APACrefatitle {Sophisticated Affective Inference: Simulating
  Anticipatory Affective Dynamics of Imagining Future Events} {Sophisticated
  affective inference: Simulating anticipatory affective dynamics of imagining
  future events}.{\BBCQ}
\newblock
\BIn{} \APACrefbtitle {International Workshop on Active Inference}
  {International workshop on active inference}\ (\BPGS\ 179--186).
\PrintBackRefs{\CurrentBib}

\bibitem [\protect \citeauthoryear {%
Hessel%
\ \protect \BOthers {.}}{%
Hessel%
\ \protect \BOthers {.}}{%
{\protect \APACyear {2018}}%
}]{%
hessel2018rainbow}
\APACinsertmetastar {%
hessel2018rainbow}%
\begin{APACrefauthors}%
Hessel, M.%
, Modayil, J.%
, Van~Hasselt, H.%
, Schaul, T.%
, Ostrovski, G.%
, Dabney, W.%
\BDBL {}Silver, D.%
\end{APACrefauthors}%
\unskip\
\newblock
\APACrefYearMonthDay{2018}{}{}.
\newblock
{\BBOQ}\APACrefatitle {Rainbow: Combining improvements in deep reinforcement
  learning} {Rainbow: Combining improvements in deep reinforcement
  learning}.{\BBCQ}
\newblock
\BIn{} \APACrefbtitle {Thirty-Second AAAI Conference on Artificial
  Intelligence.} {Thirty-second aaai conference on artificial intelligence.}
\PrintBackRefs{\CurrentBib}

\bibitem [\protect \citeauthoryear {%
Hinton%
, Srivastava%
\BCBL {}\ \BBA {} Swersky%
}{%
Hinton%
\ \protect \BOthers {.}}{%
{\protect \APACyear {2012}}%
}]{%
hinton2012neural}
\APACinsertmetastar {%
hinton2012neural}%
\begin{APACrefauthors}%
Hinton, G.%
, Srivastava, N.%
\BCBL {}\ \BBA {} Swersky, K.%
\end{APACrefauthors}%
\unskip\
\newblock
\APACrefYearMonthDay{2012}{}{}.
\newblock
{\BBOQ}\APACrefatitle {Neural networks for machine learning lecture 6a overview
  of mini-batch gradient descent} {Neural networks for machine learning lecture
  6a overview of mini-batch gradient descent}.{\BBCQ}
\newblock
\APACjournalVolNumPages{Powerpoint Presentation}{14}{8}{}.
\PrintBackRefs{\CurrentBib}

\bibitem [\protect \citeauthoryear {%
Hinton%
\ \BBA {} Zemel%
}{%
Hinton%
\ \BBA {} Zemel%
}{%
{\protect \APACyear {1994}}%
}]{%
hinton1994autoencoders}
\APACinsertmetastar {%
hinton1994autoencoders}%
\begin{APACrefauthors}%
Hinton, G.%
\BCBT {}\ \BBA {} Zemel, R\BPBI S.%
\end{APACrefauthors}%
\unskip\
\newblock
\APACrefYearMonthDay{1994}{}{}.
\newblock
{\BBOQ}\APACrefatitle {Autoencoders, minimum description length and Helmholtz
  free energy} {Autoencoders, minimum description length and helmholtz free
  energy}.{\BBCQ}
\newblock
\BIn{} \APACrefbtitle {Advances in neural Information Processing Systems}
  {Advances in neural information processing systems}\ (\BPGS\ 3--10).
\PrintBackRefs{\CurrentBib}

\bibitem [\protect \citeauthoryear {%
Hochreiter%
\ \BBA {} Schmidhuber%
}{%
Hochreiter%
\ \BBA {} Schmidhuber%
}{%
{\protect \APACyear {1997}}%
}]{%
hochreiter1997long}
\APACinsertmetastar {%
hochreiter1997long}%
\begin{APACrefauthors}%
Hochreiter, S.%
\BCBT {}\ \BBA {} Schmidhuber, J.%
\end{APACrefauthors}%
\unskip\
\newblock
\APACrefYearMonthDay{1997}{}{}.
\newblock
{\BBOQ}\APACrefatitle {Long short-term memory} {Long short-term memory}.{\BBCQ}
\newblock
\APACjournalVolNumPages{Neural Computation}{9}{8}{1735--1780}.
\PrintBackRefs{\CurrentBib}

\bibitem [\protect \citeauthoryear {%
Hohwy%
}{%
Hohwy%
}{%
{\protect \APACyear {2016}}%
}]{%
hohwy2016self}
\APACinsertmetastar {%
hohwy2016self}%
\begin{APACrefauthors}%
Hohwy, J.%
\end{APACrefauthors}%
\unskip\
\newblock
\APACrefYearMonthDay{2016}{}{}.
\newblock
{\BBOQ}\APACrefatitle {The self-evidencing brain} {The self-evidencing
  brain}.{\BBCQ}
\newblock
\APACjournalVolNumPages{No{\^u}s}{50}{2}{259--285}.
\PrintBackRefs{\CurrentBib}

\bibitem [\protect \citeauthoryear {%
Hohwy%
, Roepstorff%
\BCBL {}\ \BBA {} Friston%
}{%
Hohwy%
\ \protect \BOthers {.}}{%
{\protect \APACyear {2008}}%
}]{%
hohwy2008predictive}
\APACinsertmetastar {%
hohwy2008predictive}%
\begin{APACrefauthors}%
Hohwy, J.%
, Roepstorff, A.%
\BCBL {}\ \BBA {} Friston, K.%
\end{APACrefauthors}%
\unskip\
\newblock
\APACrefYearMonthDay{2008}{}{}.
\newblock
{\BBOQ}\APACrefatitle {Predictive coding explains binocular rivalry: An
  epistemological review} {Predictive coding explains binocular rivalry: An
  epistemological review}.{\BBCQ}
\newblock
\APACjournalVolNumPages{Cognition}{108}{3}{687--701}.
\PrintBackRefs{\CurrentBib}

\bibitem [\protect \citeauthoryear {%
Huang%
\ \BBA {} Rao%
}{%
Huang%
\ \BBA {} Rao%
}{%
{\protect \APACyear {2011}}%
}]{%
huang2011predictive}
\APACinsertmetastar {%
huang2011predictive}%
\begin{APACrefauthors}%
Huang, Y.%
\BCBT {}\ \BBA {} Rao, R\BPBI P.%
\end{APACrefauthors}%
\unskip\
\newblock
\APACrefYearMonthDay{2011}{}{}.
\newblock
{\BBOQ}\APACrefatitle {Predictive coding} {Predictive coding}.{\BBCQ}
\newblock
\APACjournalVolNumPages{Wiley Interdisciplinary Reviews: Cognitive
  Science}{2}{5}{580--593}.
\newblock
\begin{APACrefURL}
  \url{https://onlinelibrary.wiley.com/doi/pdf/10.1002/wcs.142?casa_token=TJvdr2nDbr8AAAAA:0T3LOAIXt6I7YYpJIqOs204qnwU0FFQiVC976sVifVv0XB4wFlrLZ7WvALY9x_qdoIGciEZWd12hfNQ}
  \end{APACrefURL}
\PrintBackRefs{\CurrentBib}

\bibitem [\protect \citeauthoryear {%
Hubel%
\ \BBA {} Wiesel%
}{%
Hubel%
\ \BBA {} Wiesel%
}{%
{\protect \APACyear {1962}}%
}]{%
hubel1962receptive}
\APACinsertmetastar {%
hubel1962receptive}%
\begin{APACrefauthors}%
Hubel, D\BPBI H.%
\BCBT {}\ \BBA {} Wiesel, T\BPBI N.%
\end{APACrefauthors}%
\unskip\
\newblock
\APACrefYearMonthDay{1962}{}{}.
\newblock
{\BBOQ}\APACrefatitle {Receptive fields, binocular interaction and functional
  architecture in the cat's visual cortex} {Receptive fields, binocular
  interaction and functional architecture in the cat's visual cortex}.{\BBCQ}
\newblock
\APACjournalVolNumPages{The Journal of Physiology}{160}{1}{106}.
\PrintBackRefs{\CurrentBib}

\bibitem [\protect \citeauthoryear {%
Isomura%
, Parr%
\BCBL {}\ \BBA {} Friston%
}{%
Isomura%
\ \protect \BOthers {.}}{%
{\protect \APACyear {2019}}%
}]{%
isomura2019Bayesian}
\APACinsertmetastar {%
isomura2019Bayesian}%
\begin{APACrefauthors}%
Isomura, T.%
, Parr, T.%
\BCBL {}\ \BBA {} Friston, K.%
\end{APACrefauthors}%
\unskip\
\newblock
\APACrefYearMonthDay{2019}{}{}.
\newblock
{\BBOQ}\APACrefatitle {Bayesian filtering with multiple internal models: toward
  a theory of social intelligence} {Bayesian filtering with multiple internal
  models: toward a theory of social intelligence}.{\BBCQ}
\newblock
\APACjournalVolNumPages{Neural Computation}{31}{12}{2390--2431}.
\PrintBackRefs{\CurrentBib}

\bibitem [\protect \citeauthoryear {%
Jaswinski%
}{%
Jaswinski%
}{%
{\protect \APACyear {1970}}%
}]{%
jaswinskistochastic}
\APACinsertmetastar {%
jaswinskistochastic}%
\begin{APACrefauthors}%
Jaswinski, A.%
\end{APACrefauthors}%
\unskip\
\newblock
\APACrefYearMonthDay{1970}{}{}.
\newblock
\APACrefbtitle {Stochastic Processes and Filtering Theory, 1970.} {Stochastic
  processes and filtering theory, 1970.}
\newblock
\APACaddressPublisher{}{Academic Press}.
\PrintBackRefs{\CurrentBib}

\bibitem [\protect \citeauthoryear {%
Johnson%
\ \BBA {} Moradi%
}{%
Johnson%
\ \BBA {} Moradi%
}{%
{\protect \APACyear {2005}}%
}]{%
johnson2005pid}
\APACinsertmetastar {%
johnson2005pid}%
\begin{APACrefauthors}%
Johnson, M\BPBI A.%
\BCBT {}\ \BBA {} Moradi, M\BPBI H.%
\end{APACrefauthors}%
\unskip\
\newblock
\APACrefYear{2005}.
\newblock
\APACrefbtitle {PID control} {Pid control}.
\newblock
\APACaddressPublisher{}{Springer}.
\PrintBackRefs{\CurrentBib}

\bibitem [\protect \citeauthoryear {%
M.~Jordan%
, Ghahramani%
, Jaakkola%
\BCBL {}\ \BBA {} Saul%
}{%
M.~Jordan%
\ \protect \BOthers {.}}{%
{\protect \APACyear {1998}}%
}]{%
jordan1998introduction}
\APACinsertmetastar {%
jordan1998introduction}%
\begin{APACrefauthors}%
Jordan, M.%
, Ghahramani, Z.%
, Jaakkola, T\BPBI S.%
\BCBL {}\ \BBA {} Saul, L\BPBI K.%
\end{APACrefauthors}%
\unskip\
\newblock
\APACrefYearMonthDay{1998}{}{}.
\newblock
{\BBOQ}\APACrefatitle {An introduction to variational methods for graphical
  models} {An introduction to variational methods for graphical models}.{\BBCQ}
\newblock
\BIn{} \APACrefbtitle {Learning in graphical models} {Learning in graphical
  models}\ (\BPGS\ 105--161).
\newblock
\APACaddressPublisher{}{Springer}.
\PrintBackRefs{\CurrentBib}

\bibitem [\protect \citeauthoryear {%
M\BPBI I.~Jordan%
, Ghahramani%
, Jaakkola%
\BCBL {}\ \BBA {} Saul%
}{%
M\BPBI I.~Jordan%
\ \protect \BOthers {.}}{%
{\protect \APACyear {1999}}%
}]{%
jordan1999introduction}
\APACinsertmetastar {%
jordan1999introduction}%
\begin{APACrefauthors}%
Jordan, M\BPBI I.%
, Ghahramani, Z.%
, Jaakkola, T\BPBI S.%
\BCBL {}\ \BBA {} Saul, L\BPBI K.%
\end{APACrefauthors}%
\unskip\
\newblock
\APACrefYearMonthDay{1999}{}{}.
\newblock
{\BBOQ}\APACrefatitle {An introduction to variational methods for graphical
  models} {An introduction to variational methods for graphical models}.{\BBCQ}
\newblock
\APACjournalVolNumPages{Machine learning}{37}{2}{183--233}.
\PrintBackRefs{\CurrentBib}

\bibitem [\protect \citeauthoryear {%
R.~Jordan%
, Kinderlehrer%
\BCBL {}\ \BBA {} Otto%
}{%
R.~Jordan%
\ \protect \BOthers {.}}{%
{\protect \APACyear {1998}}%
}]{%
jordan1998variational}
\APACinsertmetastar {%
jordan1998variational}%
\begin{APACrefauthors}%
Jordan, R.%
, Kinderlehrer, D.%
\BCBL {}\ \BBA {} Otto, F.%
\end{APACrefauthors}%
\unskip\
\newblock
\APACrefYearMonthDay{1998}{}{}.
\newblock
{\BBOQ}\APACrefatitle {The variational formulation of the Fokker--Planck
  equation} {The variational formulation of the fokker--planck
  equation}.{\BBCQ}
\newblock
\APACjournalVolNumPages{SIAM journal on mathematical analysis}{29}{1}{1--17}.
\PrintBackRefs{\CurrentBib}

\bibitem [\protect \citeauthoryear {%
Kaelbling%
, Littman%
\BCBL {}\ \BBA {} Cassandra%
}{%
Kaelbling%
\ \protect \BOthers {.}}{%
{\protect \APACyear {1998}}%
}]{%
kaelbling1998planning}
\APACinsertmetastar {%
kaelbling1998planning}%
\begin{APACrefauthors}%
Kaelbling, L\BPBI P.%
, Littman, M\BPBI L.%
\BCBL {}\ \BBA {} Cassandra, A\BPBI R.%
\end{APACrefauthors}%
\unskip\
\newblock
\APACrefYearMonthDay{1998}{}{}.
\newblock
{\BBOQ}\APACrefatitle {Planning and acting in partially observable stochastic
  domains} {Planning and acting in partially observable stochastic
  domains}.{\BBCQ}
\newblock
\APACjournalVolNumPages{Artificial Intelligence}{101}{1-2}{99--134}.
\PrintBackRefs{\CurrentBib}

\bibitem [\protect \citeauthoryear {%
Kaelbling%
, Littman%
\BCBL {}\ \BBA {} Moore%
}{%
Kaelbling%
\ \protect \BOthers {.}}{%
{\protect \APACyear {1996}}%
}]{%
kaelbling1996reinforcement}
\APACinsertmetastar {%
kaelbling1996reinforcement}%
\begin{APACrefauthors}%
Kaelbling, L\BPBI P.%
, Littman, M\BPBI L.%
\BCBL {}\ \BBA {} Moore, A\BPBI W.%
\end{APACrefauthors}%
\unskip\
\newblock
\APACrefYearMonthDay{1996}{}{}.
\newblock
{\BBOQ}\APACrefatitle {Reinforcement learning: A survey} {Reinforcement
  learning: A survey}.{\BBCQ}
\newblock
\APACjournalVolNumPages{Journal of artificial intelligence
  research}{4}{}{237--285}.
\PrintBackRefs{\CurrentBib}

\bibitem [\protect \citeauthoryear {%
Kaiser%
, Mostafa%
\BCBL {}\ \BBA {} Neftci%
}{%
Kaiser%
\ \protect \BOthers {.}}{%
{\protect \APACyear {2020}}%
}]{%
kaiser2020synaptic}
\APACinsertmetastar {%
kaiser2020synaptic}%
\begin{APACrefauthors}%
Kaiser, J.%
, Mostafa, H.%
\BCBL {}\ \BBA {} Neftci, E.%
\end{APACrefauthors}%
\unskip\
\newblock
\APACrefYearMonthDay{2020}{}{}.
\newblock
{\BBOQ}\APACrefatitle {Synaptic plasticity dynamics for deep continuous local
  learning (DECOLLE)} {Synaptic plasticity dynamics for deep continuous local
  learning (decolle)}.{\BBCQ}
\newblock
\APACjournalVolNumPages{Frontiers in Neuroscience}{14}{}{424}.
\PrintBackRefs{\CurrentBib}

\bibitem [\protect \citeauthoryear {%
Kalman%
}{%
Kalman%
}{%
{\protect \APACyear {1960}}%
}]{%
kalman1960new}
\APACinsertmetastar {%
kalman1960new}%
\begin{APACrefauthors}%
Kalman, R\BPBI E.%
\end{APACrefauthors}%
\unskip\
\newblock
\APACrefYearMonthDay{1960}{}{}.
\newblock
{\BBOQ}\APACrefatitle {A new approach to linear filtering and prediction
  problems} {A new approach to linear filtering and prediction
  problems}.{\BBCQ}
\newblock

\PrintBackRefs{\CurrentBib}

\bibitem [\protect \citeauthoryear {%
Kalman%
\ \BBA {} Bucy%
}{%
Kalman%
\ \BBA {} Bucy%
}{%
{\protect \APACyear {1961}}%
}]{%
kalman1961new}
\APACinsertmetastar {%
kalman1961new}%
\begin{APACrefauthors}%
Kalman, R\BPBI E.%
\BCBT {}\ \BBA {} Bucy, R\BPBI S.%
\end{APACrefauthors}%
\unskip\
\newblock
\APACrefYearMonthDay{1961}{}{}.
\newblock
{\BBOQ}\APACrefatitle {New results in linear filtering and prediction theory}
  {New results in linear filtering and prediction theory}.{\BBCQ}
\newblock
\APACjournalVolNumPages{Journal of Basic Engineering}{83}{1}{95--108}.
\PrintBackRefs{\CurrentBib}

\bibitem [\protect \citeauthoryear {%
Kalman%
\ \protect \BOthers {.}}{%
Kalman%
\ \protect \BOthers {.}}{%
{\protect \APACyear {1960}}%
}]{%
kalman1960contributions}
\APACinsertmetastar {%
kalman1960contributions}%
\begin{APACrefauthors}%
Kalman, R\BPBI E.%
\BCBT {}\ \BOthersPeriod {.}
\end{APACrefauthors}%
\unskip\
\newblock
\APACrefYearMonthDay{1960}{}{}.
\newblock
{\BBOQ}\APACrefatitle {Contributions to the theory of optimal control}
  {Contributions to the theory of optimal control}.{\BBCQ}
\newblock
\APACjournalVolNumPages{Bol. soc. mat. mexicana}{5}{2}{102--119}.
\PrintBackRefs{\CurrentBib}

\bibitem [\protect \citeauthoryear {%
Kanai%
, Komura%
, Shipp%
\BCBL {}\ \BBA {} Friston%
}{%
Kanai%
\ \protect \BOthers {.}}{%
{\protect \APACyear {2015}}%
}]{%
kanai2015cerebral}
\APACinsertmetastar {%
kanai2015cerebral}%
\begin{APACrefauthors}%
Kanai, R.%
, Komura, Y.%
, Shipp, S.%
\BCBL {}\ \BBA {} Friston, K.%
\end{APACrefauthors}%
\unskip\
\newblock
\APACrefYearMonthDay{2015}{}{}.
\newblock
{\BBOQ}\APACrefatitle {Cerebral hierarchies: predictive processing, precision
  and the pulvinar} {Cerebral hierarchies: predictive processing, precision and
  the pulvinar}.{\BBCQ}
\newblock
\APACjournalVolNumPages{Philosophical Transactions of the Royal Society B:
  Biological Sciences}{370}{1668}{20140169}.
\PrintBackRefs{\CurrentBib}

\bibitem [\protect \citeauthoryear {%
J.~Kaplan%
\ \protect \BOthers {.}}{%
J.~Kaplan%
\ \protect \BOthers {.}}{%
{\protect \APACyear {2020}}%
}]{%
kaplan2020scaling}
\APACinsertmetastar {%
kaplan2020scaling}%
\begin{APACrefauthors}%
Kaplan, J.%
, McCandlish, S.%
, Henighan, T.%
, Brown, T\BPBI B.%
, Chess, B.%
, Child, R.%
\BDBL {}Amodei, D.%
\end{APACrefauthors}%
\unskip\
\newblock
\APACrefYearMonthDay{2020}{}{}.
\newblock
{\BBOQ}\APACrefatitle {Scaling laws for neural language models} {Scaling laws
  for neural language models}.{\BBCQ}
\newblock
\APACjournalVolNumPages{arXiv preprint arXiv:2001.08361}{}{}{}.
\PrintBackRefs{\CurrentBib}

\bibitem [\protect \citeauthoryear {%
R.~Kaplan%
\ \BBA {} Friston%
}{%
R.~Kaplan%
\ \BBA {} Friston%
}{%
{\protect \APACyear {2018}}%
}]{%
kaplan2018planning}
\APACinsertmetastar {%
kaplan2018planning}%
\begin{APACrefauthors}%
Kaplan, R.%
\BCBT {}\ \BBA {} Friston, K.%
\end{APACrefauthors}%
\unskip\
\newblock
\APACrefYearMonthDay{2018}{}{}.
\newblock
{\BBOQ}\APACrefatitle {Planning and navigation as active inference} {Planning
  and navigation as active inference}.{\BBCQ}
\newblock
\APACjournalVolNumPages{Biological Cybernetics}{112}{4}{323--343}.
\PrintBackRefs{\CurrentBib}

\bibitem [\protect \citeauthoryear {%
Kappen%
}{%
Kappen%
}{%
{\protect \APACyear {2005}}%
}]{%
kappen2005path}
\APACinsertmetastar {%
kappen2005path}%
\begin{APACrefauthors}%
Kappen, H\BPBI J.%
\end{APACrefauthors}%
\unskip\
\newblock
\APACrefYearMonthDay{2005}{}{}.
\newblock
{\BBOQ}\APACrefatitle {Path integrals and symmetry breaking for optimal control
  theory} {Path integrals and symmetry breaking for optimal control
  theory}.{\BBCQ}
\newblock
\APACjournalVolNumPages{Journal of statistical mechanics: theory and
  experiment}{2005}{11}{P11011}.
\PrintBackRefs{\CurrentBib}

\bibitem [\protect \citeauthoryear {%
Kappen%
}{%
Kappen%
}{%
{\protect \APACyear {2007}}%
}]{%
kappen2007introduction}
\APACinsertmetastar {%
kappen2007introduction}%
\begin{APACrefauthors}%
Kappen, H\BPBI J.%
\end{APACrefauthors}%
\unskip\
\newblock
\APACrefYearMonthDay{2007}{}{}.
\newblock
{\BBOQ}\APACrefatitle {An introduction to stochastic control theory, path
  integrals and reinforcement learning} {An introduction to stochastic control
  theory, path integrals and reinforcement learning}.{\BBCQ}
\newblock
\BIn{} \APACrefbtitle {AIP conference proceedings} {Aip conference
  proceedings}\ (\BVOL~887, \BPGS\ 149--181).
\PrintBackRefs{\CurrentBib}

\bibitem [\protect \citeauthoryear {%
Kappen%
, G{\'o}mez%
\BCBL {}\ \BBA {} Opper%
}{%
Kappen%
\ \protect \BOthers {.}}{%
{\protect \APACyear {2012}}%
}]{%
kappen2012optimal}
\APACinsertmetastar {%
kappen2012optimal}%
\begin{APACrefauthors}%
Kappen, H\BPBI J.%
, G{\'o}mez, V.%
\BCBL {}\ \BBA {} Opper, M.%
\end{APACrefauthors}%
\unskip\
\newblock
\APACrefYearMonthDay{2012}{}{}.
\newblock
{\BBOQ}\APACrefatitle {Optimal control as a graphical model inference problem}
  {Optimal control as a graphical model inference problem}.{\BBCQ}
\newblock
\APACjournalVolNumPages{Machine learning}{87}{2}{159--182}.
\PrintBackRefs{\CurrentBib}

\bibitem [\protect \citeauthoryear {%
Keller%
\ \BBA {} Mrsic-Flogel%
}{%
Keller%
\ \BBA {} Mrsic-Flogel%
}{%
{\protect \APACyear {2018}}%
}]{%
keller2018predictive}
\APACinsertmetastar {%
keller2018predictive}%
\begin{APACrefauthors}%
Keller, G\BPBI B.%
\BCBT {}\ \BBA {} Mrsic-Flogel, T\BPBI D.%
\end{APACrefauthors}%
\unskip\
\newblock
\APACrefYearMonthDay{2018}{}{}.
\newblock
{\BBOQ}\APACrefatitle {Predictive processing: a canonical cortical computation}
  {Predictive processing: a canonical cortical computation}.{\BBCQ}
\newblock
\APACjournalVolNumPages{Neuron}{100}{2}{424--435}.
\PrintBackRefs{\CurrentBib}

\bibitem [\protect \citeauthoryear {%
H.~Kim%
, Kim%
, Jeong%
, Levine%
\BCBL {}\ \BBA {} Song%
}{%
H.~Kim%
\ \protect \BOthers {.}}{%
{\protect \APACyear {2018}}%
}]{%
kim_emi:_2018}
\APACinsertmetastar {%
kim_emi:_2018}%
\begin{APACrefauthors}%
Kim, H.%
, Kim, J.%
, Jeong, Y.%
, Levine, S.%
\BCBL {}\ \BBA {} Song, H\BPBI O.%
\end{APACrefauthors}%
\unskip\
\newblock
\APACrefYearMonthDay{2018}{}{}.
\newblock
{\BBOQ}\APACrefatitle {Emi: Exploration with mutual information} {Emi:
  Exploration with mutual information}.{\BBCQ}
\newblock
\APACjournalVolNumPages{arXiv preprint arXiv:1810.01176}{}{}{}.
\PrintBackRefs{\CurrentBib}

\bibitem [\protect \citeauthoryear {%
Y.~Kim%
, Wiseman%
, Miller%
, Sontag%
\BCBL {}\ \BBA {} Rush%
}{%
Y.~Kim%
\ \protect \BOthers {.}}{%
{\protect \APACyear {2018}}%
}]{%
kim2018semi}
\APACinsertmetastar {%
kim2018semi}%
\begin{APACrefauthors}%
Kim, Y.%
, Wiseman, S.%
, Miller, A\BPBI C.%
, Sontag, D.%
\BCBL {}\ \BBA {} Rush, A\BPBI M.%
\end{APACrefauthors}%
\unskip\
\newblock
\APACrefYearMonthDay{2018}{}{}.
\newblock
{\BBOQ}\APACrefatitle {Semi-amortized variational autoencoders} {Semi-amortized
  variational autoencoders}.{\BBCQ}
\newblock
\APACjournalVolNumPages{arXiv preprint arXiv:1802.02550}{}{}{}.
\PrintBackRefs{\CurrentBib}

\bibitem [\protect \citeauthoryear {%
Kingma%
\ \BBA {} Ba%
}{%
Kingma%
\ \BBA {} Ba%
}{%
{\protect \APACyear {2014}}%
}]{%
kingma2014adam}
\APACinsertmetastar {%
kingma2014adam}%
\begin{APACrefauthors}%
Kingma, D\BPBI P.%
\BCBT {}\ \BBA {} Ba, J.%
\end{APACrefauthors}%
\unskip\
\newblock
\APACrefYearMonthDay{2014}{}{}.
\newblock
{\BBOQ}\APACrefatitle {Adam: A method for stochastic optimization} {Adam: A
  method for stochastic optimization}.{\BBCQ}
\newblock
\APACjournalVolNumPages{arXiv preprint arXiv:1412.6980}{}{}{}.
\PrintBackRefs{\CurrentBib}

\bibitem [\protect \citeauthoryear {%
Kingma%
\ \BBA {} Welling%
}{%
Kingma%
\ \BBA {} Welling%
}{%
{\protect \APACyear {2013}}%
}]{%
kingma_auto-encoding_2013}
\APACinsertmetastar {%
kingma_auto-encoding_2013}%
\begin{APACrefauthors}%
Kingma, D\BPBI P.%
\BCBT {}\ \BBA {} Welling, M.%
\end{APACrefauthors}%
\unskip\
\newblock
\APACrefYearMonthDay{2013}{}{}.
\newblock
{\BBOQ}\APACrefatitle {Auto-encoding variational bayes} {Auto-encoding
  variational bayes}.{\BBCQ}
\newblock
\APACjournalVolNumPages{arXiv preprint arXiv:1312.6114}{}{}{}.
\PrintBackRefs{\CurrentBib}

\bibitem [\protect \citeauthoryear {%
Kirk%
}{%
Kirk%
}{%
{\protect \APACyear {2004}}%
}]{%
kirk2004optimal}
\APACinsertmetastar {%
kirk2004optimal}%
\begin{APACrefauthors}%
Kirk, D\BPBI E.%
\end{APACrefauthors}%
\unskip\
\newblock
\APACrefYear{2004}.
\newblock
\APACrefbtitle {Optimal control theory: an introduction} {Optimal control
  theory: an introduction}.
\newblock
\APACaddressPublisher{}{Courier Corporation}.
\PrintBackRefs{\CurrentBib}

\bibitem [\protect \citeauthoryear {%
Klyubin%
, Polani%
\BCBL {}\ \BBA {} Nehaniv%
}{%
Klyubin%
\ \protect \BOthers {.}}{%
{\protect \APACyear {2005}}%
}]{%
klyubin2005empowerment}
\APACinsertmetastar {%
klyubin2005empowerment}%
\begin{APACrefauthors}%
Klyubin, A\BPBI S.%
, Polani, D.%
\BCBL {}\ \BBA {} Nehaniv, C\BPBI L.%
\end{APACrefauthors}%
\unskip\
\newblock
\APACrefYearMonthDay{2005}{}{}.
\newblock
{\BBOQ}\APACrefatitle {Empowerment: A universal agent-centric measure of
  control} {Empowerment: A universal agent-centric measure of control}.{\BBCQ}
\newblock
\BIn{} \APACrefbtitle {2005 IEEE Congress on Evolutionary Computation} {2005
  ieee congress on evolutionary computation}\ (\BVOL~1, \BPGS\ 128--135).
\PrintBackRefs{\CurrentBib}

\bibitem [\protect \citeauthoryear {%
Kocsis%
\ \BBA {} Szepesv{\'a}ri%
}{%
Kocsis%
\ \BBA {} Szepesv{\'a}ri%
}{%
{\protect \APACyear {2006}}%
}]{%
kocsis2006bandit}
\APACinsertmetastar {%
kocsis2006bandit}%
\begin{APACrefauthors}%
Kocsis, L.%
\BCBT {}\ \BBA {} Szepesv{\'a}ri, C.%
\end{APACrefauthors}%
\unskip\
\newblock
\APACrefYearMonthDay{2006}{}{}.
\newblock
{\BBOQ}\APACrefatitle {Bandit based monte-carlo planning} {Bandit based
  monte-carlo planning}.{\BBCQ}
\newblock
\BIn{} \APACrefbtitle {European Conference on Machine Learning} {European
  conference on machine learning}\ (\BPGS\ 282--293).
\PrintBackRefs{\CurrentBib}

\bibitem [\protect \citeauthoryear {%
Kondepudi%
\ \BBA {} Prigogine%
}{%
Kondepudi%
\ \BBA {} Prigogine%
}{%
{\protect \APACyear {2014}}%
}]{%
kondepudi2014modern}
\APACinsertmetastar {%
kondepudi2014modern}%
\begin{APACrefauthors}%
Kondepudi, D.%
\BCBT {}\ \BBA {} Prigogine, I.%
\end{APACrefauthors}%
\unskip\
\newblock
\APACrefYear{2014}.
\newblock
\APACrefbtitle {Modern thermodynamics: from heat engines to dissipative
  structures} {Modern thermodynamics: from heat engines to dissipative
  structures}.
\newblock
\APACaddressPublisher{}{John Wiley \& Sons}.
\PrintBackRefs{\CurrentBib}

\bibitem [\protect \citeauthoryear {%
Kopp%
}{%
Kopp%
}{%
{\protect \APACyear {1962}}%
}]{%
kopp1962pontryagin}
\APACinsertmetastar {%
kopp1962pontryagin}%
\begin{APACrefauthors}%
Kopp, R\BPBI E.%
\end{APACrefauthors}%
\unskip\
\newblock
\APACrefYearMonthDay{1962}{}{}.
\newblock
{\BBOQ}\APACrefatitle {Pontryagin maximum principle} {Pontryagin maximum
  principle}.{\BBCQ}
\newblock
\BIn{} \APACrefbtitle {Mathematics in Science and Engineering} {Mathematics in
  science and engineering}\ (\BVOL~5, \BPGS\ 255--279).
\newblock
\APACaddressPublisher{}{Elsevier}.
\PrintBackRefs{\CurrentBib}

\bibitem [\protect \citeauthoryear {%
Krebs%
, Kacelnik%
\BCBL {}\ \BBA {} Taylor%
}{%
Krebs%
\ \protect \BOthers {.}}{%
{\protect \APACyear {1978}}%
}]{%
krebs1978test}
\APACinsertmetastar {%
krebs1978test}%
\begin{APACrefauthors}%
Krebs, J\BPBI R.%
, Kacelnik, A.%
\BCBL {}\ \BBA {} Taylor, P.%
\end{APACrefauthors}%
\unskip\
\newblock
\APACrefYearMonthDay{1978}{}{}.
\newblock
{\BBOQ}\APACrefatitle {Test of optimal sampling by foraging great tits} {Test
  of optimal sampling by foraging great tits}.{\BBCQ}
\newblock
\APACjournalVolNumPages{Nature}{275}{5675}{27--31}.
\PrintBackRefs{\CurrentBib}

\bibitem [\protect \citeauthoryear {%
Kriegeskorte%
}{%
Kriegeskorte%
}{%
{\protect \APACyear {2015}}%
}]{%
kriegeskorte2015deep}
\APACinsertmetastar {%
kriegeskorte2015deep}%
\begin{APACrefauthors}%
Kriegeskorte, N.%
\end{APACrefauthors}%
\unskip\
\newblock
\APACrefYearMonthDay{2015}{}{}.
\newblock
{\BBOQ}\APACrefatitle {Deep neural networks: a new framework for modeling
  biological vision and brain information processing} {Deep neural networks: a
  new framework for modeling biological vision and brain information
  processing}.{\BBCQ}
\newblock
\APACjournalVolNumPages{Annual review of vision science}{1}{}{417--446}.
\PrintBackRefs{\CurrentBib}

\bibitem [\protect \citeauthoryear {%
Krizhevsky%
, Sutskever%
\BCBL {}\ \BBA {} Hinton%
}{%
Krizhevsky%
\ \protect \BOthers {.}}{%
{\protect \APACyear {2012}}%
}]{%
krizhevsky2012imagenet}
\APACinsertmetastar {%
krizhevsky2012imagenet}%
\begin{APACrefauthors}%
Krizhevsky, A.%
, Sutskever, I.%
\BCBL {}\ \BBA {} Hinton, G.%
\end{APACrefauthors}%
\unskip\
\newblock
\APACrefYearMonthDay{2012}{}{}.
\newblock
{\BBOQ}\APACrefatitle {Imagenet classification with deep convolutional neural
  networks} {Imagenet classification with deep convolutional neural
  networks}.{\BBCQ}
\newblock
\BIn{} \APACrefbtitle {Advances in neural Information Processing Systems}
  {Advances in neural information processing systems}\ (\BPGS\ 1097--1105).
\PrintBackRefs{\CurrentBib}

\bibitem [\protect \citeauthoryear {%
Kutschireiter%
}{%
Kutschireiter%
}{%
{\protect \APACyear {2018}}%
}]{%
kutschireiter2018nonlinear}
\APACinsertmetastar {%
kutschireiter2018nonlinear}%
\begin{APACrefauthors}%
Kutschireiter, A.%
\end{APACrefauthors}%
\unskip\
\newblock
\APACrefYear{2018}.
\unskip\
\newblock
\APACrefbtitle {Nonlinear filtering in neuroscience: theory and application}
  {Nonlinear filtering in neuroscience: theory and application}\
  \APACtypeAddressSchool {\BUPhD}{}{}.
\unskip\
\newblock
\APACaddressSchool {}{University of Zurich}.
\PrintBackRefs{\CurrentBib}

\bibitem [\protect \citeauthoryear {%
Kutschireiter%
, Surace%
\BCBL {}\ \BBA {} Pfister%
}{%
Kutschireiter%
\ \protect \BOthers {.}}{%
{\protect \APACyear {2020}}%
}]{%
kutschireiter2020hitchhiker}
\APACinsertmetastar {%
kutschireiter2020hitchhiker}%
\begin{APACrefauthors}%
Kutschireiter, A.%
, Surace, S\BPBI C.%
\BCBL {}\ \BBA {} Pfister, J\BHBI P.%
\end{APACrefauthors}%
\unskip\
\newblock
\APACrefYearMonthDay{2020}{}{}.
\newblock
{\BBOQ}\APACrefatitle {The Hitchhiker’s guide to nonlinear filtering} {The
  hitchhiker’s guide to nonlinear filtering}.{\BBCQ}
\newblock
\APACjournalVolNumPages{Journal of Mathematical Psychology}{94}{}{102307}.
\PrintBackRefs{\CurrentBib}

\bibitem [\protect \citeauthoryear {%
Kutschireiter%
, Surace%
, Sprekeler%
\BCBL {}\ \BBA {} Pfister%
}{%
Kutschireiter%
\ \protect \BOthers {.}}{%
{\protect \APACyear {2015}}%
}]{%
kutschireiter2015neural}
\APACinsertmetastar {%
kutschireiter2015neural}%
\begin{APACrefauthors}%
Kutschireiter, A.%
, Surace, S\BPBI C.%
, Sprekeler, H.%
\BCBL {}\ \BBA {} Pfister, J\BHBI P.%
\end{APACrefauthors}%
\unskip\
\newblock
\APACrefYearMonthDay{2015}{}{}.
\newblock
{\BBOQ}\APACrefatitle {The neural particle filter} {The neural particle
  filter}.{\BBCQ}
\newblock
\APACjournalVolNumPages{arXiv preprint arXiv:1508.06818}{}{}{}.
\PrintBackRefs{\CurrentBib}

\bibitem [\protect \citeauthoryear {%
Kwakernaak%
\ \BBA {} Sivan%
}{%
Kwakernaak%
\ \BBA {} Sivan%
}{%
{\protect \APACyear {1972}}%
}]{%
kwakernaak1972linear}
\APACinsertmetastar {%
kwakernaak1972linear}%
\begin{APACrefauthors}%
Kwakernaak, H.%
\BCBT {}\ \BBA {} Sivan, R.%
\end{APACrefauthors}%
\unskip\
\newblock
\APACrefYear{1972}.
\newblock
\APACrefbtitle {Linear optimal control systems} {Linear optimal control
  systems}\ (\BVOL~1).
\newblock
\APACaddressPublisher{}{Wiley-interscience New York}.
\PrintBackRefs{\CurrentBib}

\bibitem [\protect \citeauthoryear {%
Lanczos%
}{%
Lanczos%
}{%
{\protect \APACyear {2012}}%
}]{%
lanczos2012variational}
\APACinsertmetastar {%
lanczos2012variational}%
\begin{APACrefauthors}%
Lanczos, C.%
\end{APACrefauthors}%
\unskip\
\newblock
\APACrefYear{2012}.
\newblock
\APACrefbtitle {The variational principles of mechanics} {The variational
  principles of mechanics}.
\newblock
\APACaddressPublisher{}{Courier Corporation}.
\PrintBackRefs{\CurrentBib}

\bibitem [\protect \citeauthoryear {%
Launay%
, Poli%
\BCBL {}\ \BBA {} Krzakala%
}{%
Launay%
\ \protect \BOthers {.}}{%
{\protect \APACyear {2019}}%
}]{%
launay2019principled}
\APACinsertmetastar {%
launay2019principled}%
\begin{APACrefauthors}%
Launay, J.%
, Poli, I.%
\BCBL {}\ \BBA {} Krzakala, F.%
\end{APACrefauthors}%
\unskip\
\newblock
\APACrefYearMonthDay{2019}{}{}.
\newblock
{\BBOQ}\APACrefatitle {Principled Training of Neural Networks with Direct
  Feedback Alignment} {Principled training of neural networks with direct
  feedback alignment}.{\BBCQ}
\newblock
\APACjournalVolNumPages{arXiv preprint arXiv:1906.04554}{}{}{}.
\PrintBackRefs{\CurrentBib}

\bibitem [\protect \citeauthoryear {%
Lawson%
, Rees%
\BCBL {}\ \BBA {} Friston%
}{%
Lawson%
\ \protect \BOthers {.}}{%
{\protect \APACyear {2014}}%
}]{%
lawson2014aberrant}
\APACinsertmetastar {%
lawson2014aberrant}%
\begin{APACrefauthors}%
Lawson, R\BPBI P.%
, Rees, G.%
\BCBL {}\ \BBA {} Friston, K.%
\end{APACrefauthors}%
\unskip\
\newblock
\APACrefYearMonthDay{2014}{}{}.
\newblock
{\BBOQ}\APACrefatitle {An aberrant precision account of autism} {An aberrant
  precision account of autism}.{\BBCQ}
\newblock
\APACjournalVolNumPages{Frontiers in human neuroscience}{8}{}{302}.
\newblock
\begin{APACrefURL}
  \url{https://www.frontiersin.org/articles/10.3389/fnhum.2014.00302/full}
  \end{APACrefURL}
\PrintBackRefs{\CurrentBib}

\bibitem [\protect \citeauthoryear {%
D\BHBI H.~Lee%
, Zhang%
, Fischer%
\BCBL {}\ \BBA {} Bengio%
}{%
D\BHBI H.~Lee%
\ \protect \BOthers {.}}{%
{\protect \APACyear {2015}}%
}]{%
lee2015difference}
\APACinsertmetastar {%
lee2015difference}%
\begin{APACrefauthors}%
Lee, D\BHBI H.%
, Zhang, S.%
, Fischer, A.%
\BCBL {}\ \BBA {} Bengio, Y.%
\end{APACrefauthors}%
\unskip\
\newblock
\APACrefYearMonthDay{2015}{}{}.
\newblock
{\BBOQ}\APACrefatitle {Difference target propagation} {Difference target
  propagation}.{\BBCQ}
\newblock
\BIn{} \APACrefbtitle {Joint european conference on machine learning and
  knowledge discovery in databases} {Joint european conference on machine
  learning and knowledge discovery in databases}\ (\BPGS\ 498--515).
\PrintBackRefs{\CurrentBib}

\bibitem [\protect \citeauthoryear {%
L.~Lee%
\ \protect \BOthers {.}}{%
L.~Lee%
\ \protect \BOthers {.}}{%
{\protect \APACyear {2019}}%
}]{%
lee2019efficient}
\APACinsertmetastar {%
lee2019efficient}%
\begin{APACrefauthors}%
Lee, L.%
, Eysenbach, B.%
, Parisotto, E.%
, Xing, E.%
, Levine, S.%
\BCBL {}\ \BBA {} Salakhutdinov, R.%
\end{APACrefauthors}%
\unskip\
\newblock
\APACrefYearMonthDay{2019}{}{}.
\newblock
{\BBOQ}\APACrefatitle {Efficient exploration via state marginal matching}
  {Efficient exploration via state marginal matching}.{\BBCQ}
\newblock
\APACjournalVolNumPages{arXiv preprint arXiv:1906.05274}{}{}{}.
\PrintBackRefs{\CurrentBib}

\bibitem [\protect \citeauthoryear {%
Leondes%
}{%
Leondes%
}{%
{\protect \APACyear {1970}}%
}]{%
leondes1970theory}
\APACinsertmetastar {%
leondes1970theory}%
\begin{APACrefauthors}%
Leondes, C\BPBI T.%
\end{APACrefauthors}%
\unskip\
\newblock
\APACrefYearMonthDay{1970}{}{}.
\newblock
\APACrefbtitle {Theory and applications of Kalman filtering} {Theory and
  applications of kalman filtering}\ \APACbVolEdTR{}{\BTR{}}.
\newblock
\APACaddressInstitution{}{Advisory Group for Aerospace Research and Development
  Neuilly-Sur-Seine (France)}.
\PrintBackRefs{\CurrentBib}

\bibitem [\protect \citeauthoryear {%
Levine%
}{%
Levine%
}{%
{\protect \APACyear {2018}}%
}]{%
levine2018reinforcement}
\APACinsertmetastar {%
levine2018reinforcement}%
\begin{APACrefauthors}%
Levine, S.%
\end{APACrefauthors}%
\unskip\
\newblock
\APACrefYearMonthDay{2018}{}{}.
\newblock
{\BBOQ}\APACrefatitle {Reinforcement learning and control as probabilistic
  inference: Tutorial and review} {Reinforcement learning and control as
  probabilistic inference: Tutorial and review}.{\BBCQ}
\newblock
\APACjournalVolNumPages{arXiv preprint arXiv:1805.00909}{}{}{}.
\PrintBackRefs{\CurrentBib}

\bibitem [\protect \citeauthoryear {%
Levine%
, Kumar%
, Tucker%
\BCBL {}\ \BBA {} Fu%
}{%
Levine%
\ \protect \BOthers {.}}{%
{\protect \APACyear {2020}}%
}]{%
levine2020offline}
\APACinsertmetastar {%
levine2020offline}%
\begin{APACrefauthors}%
Levine, S.%
, Kumar, A.%
, Tucker, G.%
\BCBL {}\ \BBA {} Fu, J.%
\end{APACrefauthors}%
\unskip\
\newblock
\APACrefYearMonthDay{2020}{}{}.
\newblock
{\BBOQ}\APACrefatitle {Offline reinforcement learning: Tutorial, review, and
  perspectives on open problems} {Offline reinforcement learning: Tutorial,
  review, and perspectives on open problems}.{\BBCQ}
\newblock
\APACjournalVolNumPages{arXiv preprint arXiv:2005.01643}{}{}{}.
\PrintBackRefs{\CurrentBib}

\bibitem [\protect \citeauthoryear {%
S.~Li%
}{%
S.~Li%
}{%
{\protect \APACyear {2020}}%
}]{%
li2020robot}
\APACinsertmetastar {%
li2020robot}%
\begin{APACrefauthors}%
Li, S.%
\end{APACrefauthors}%
\unskip\
\newblock
\APACrefYearMonthDay{2020}{}{}.
\newblock
{\BBOQ}\APACrefatitle {Robot Playing Kendama with Model-Based and Model-Free
  Reinforcement Learning} {Robot playing kendama with model-based and
  model-free reinforcement learning}.{\BBCQ}
\newblock
\APACjournalVolNumPages{arXiv preprint arXiv:2003.06751}{}{}{}.
\PrintBackRefs{\CurrentBib}

\bibitem [\protect \citeauthoryear {%
W.~Li%
\ \BBA {} Todorov%
}{%
W.~Li%
\ \BBA {} Todorov%
}{%
{\protect \APACyear {2004}}%
}]{%
li2004iterative}
\APACinsertmetastar {%
li2004iterative}%
\begin{APACrefauthors}%
Li, W.%
\BCBT {}\ \BBA {} Todorov, E.%
\end{APACrefauthors}%
\unskip\
\newblock
\APACrefYearMonthDay{2004}{}{}.
\newblock
{\BBOQ}\APACrefatitle {Iterative linear quadratic regulator design for
  nonlinear biological movement systems.} {Iterative linear quadratic regulator
  design for nonlinear biological movement systems.}{\BBCQ}
\newblock
\BIn{} \APACrefbtitle {ICINCO (1)} {Icinco (1)}\ (\BPGS\ 222--229).
\PrintBackRefs{\CurrentBib}

\bibitem [\protect \citeauthoryear {%
Liao%
, Leibo%
\BCBL {}\ \BBA {} Poggio%
}{%
Liao%
\ \protect \BOthers {.}}{%
{\protect \APACyear {2016}}%
}]{%
liao2016important}
\APACinsertmetastar {%
liao2016important}%
\begin{APACrefauthors}%
Liao, Q.%
, Leibo, J\BPBI Z.%
\BCBL {}\ \BBA {} Poggio, T.%
\end{APACrefauthors}%
\unskip\
\newblock
\APACrefYearMonthDay{2016}{}{}.
\newblock
{\BBOQ}\APACrefatitle {How important is weight symmetry in backpropagation?}
  {How important is weight symmetry in backpropagation?}{\BBCQ}
\newblock
\BIn{} \APACrefbtitle {Thirtieth AAAI Conference on Artificial Intelligence.}
  {Thirtieth aaai conference on artificial intelligence.}
\PrintBackRefs{\CurrentBib}

\bibitem [\protect \citeauthoryear {%
Lillicrap%
, Cownden%
, Tweed%
\BCBL {}\ \BBA {} Akerman%
}{%
Lillicrap%
\ \protect \BOthers {.}}{%
{\protect \APACyear {2014}}%
}]{%
lillicrap2014random}
\APACinsertmetastar {%
lillicrap2014random}%
\begin{APACrefauthors}%
Lillicrap, T\BPBI P.%
, Cownden, D.%
, Tweed, D\BPBI B.%
\BCBL {}\ \BBA {} Akerman, C\BPBI J.%
\end{APACrefauthors}%
\unskip\
\newblock
\APACrefYearMonthDay{2014}{}{}.
\newblock
{\BBOQ}\APACrefatitle {Random feedback weights support learning in deep neural
  networks} {Random feedback weights support learning in deep neural
  networks}.{\BBCQ}
\newblock
\APACjournalVolNumPages{arXiv preprint arXiv:1411.0247}{}{}{}.
\PrintBackRefs{\CurrentBib}

\bibitem [\protect \citeauthoryear {%
Lillicrap%
, Cownden%
, Tweed%
\BCBL {}\ \BBA {} Akerman%
}{%
Lillicrap%
\ \protect \BOthers {.}}{%
{\protect \APACyear {2016}}%
}]{%
lillicrap2016random}
\APACinsertmetastar {%
lillicrap2016random}%
\begin{APACrefauthors}%
Lillicrap, T\BPBI P.%
, Cownden, D.%
, Tweed, D\BPBI B.%
\BCBL {}\ \BBA {} Akerman, C\BPBI J.%
\end{APACrefauthors}%
\unskip\
\newblock
\APACrefYearMonthDay{2016}{}{}.
\newblock
{\BBOQ}\APACrefatitle {Random synaptic feedback weights support error
  backpropagation for deep learning} {Random synaptic feedback weights support
  error backpropagation for deep learning}.{\BBCQ}
\newblock
\APACjournalVolNumPages{Nature communications}{7}{1}{1--10}.
\PrintBackRefs{\CurrentBib}

\bibitem [\protect \citeauthoryear {%
Lillicrap%
\ \BBA {} Santoro%
}{%
Lillicrap%
\ \BBA {} Santoro%
}{%
{\protect \APACyear {2019}}%
}]{%
lillicrap2019backpropagation}
\APACinsertmetastar {%
lillicrap2019backpropagation}%
\begin{APACrefauthors}%
Lillicrap, T\BPBI P.%
\BCBT {}\ \BBA {} Santoro, A.%
\end{APACrefauthors}%
\unskip\
\newblock
\APACrefYearMonthDay{2019}{}{}.
\newblock
{\BBOQ}\APACrefatitle {Backpropagation through time and the brain}
  {Backpropagation through time and the brain}.{\BBCQ}
\newblock
\APACjournalVolNumPages{Current Opinion in Neurobiology}{55}{}{82--89}.
\PrintBackRefs{\CurrentBib}

\bibitem [\protect \citeauthoryear {%
Lillicrap%
, Santoro%
, Marris%
, Akerman%
\BCBL {}\ \BBA {} Hinton%
}{%
Lillicrap%
\ \protect \BOthers {.}}{%
{\protect \APACyear {2020}}%
}]{%
lillicrap2020backpropagation}
\APACinsertmetastar {%
lillicrap2020backpropagation}%
\begin{APACrefauthors}%
Lillicrap, T\BPBI P.%
, Santoro, A.%
, Marris, L.%
, Akerman, C\BPBI J.%
\BCBL {}\ \BBA {} Hinton, G.%
\end{APACrefauthors}%
\unskip\
\newblock
\APACrefYearMonthDay{2020}{}{}.
\newblock
{\BBOQ}\APACrefatitle {Backpropagation and the brain} {Backpropagation and the
  brain}.{\BBCQ}
\newblock
\APACjournalVolNumPages{Nature Reviews Neuroscience}{}{}{1--12}.
\PrintBackRefs{\CurrentBib}

\bibitem [\protect \citeauthoryear {%
Linnainmaa%
}{%
Linnainmaa%
}{%
{\protect \APACyear {1970}}%
}]{%
linnainmaa1970representation}
\APACinsertmetastar {%
linnainmaa1970representation}%
\begin{APACrefauthors}%
Linnainmaa, S.%
\end{APACrefauthors}%
\unskip\
\newblock
\APACrefYearMonthDay{1970}{}{}.
\newblock
{\BBOQ}\APACrefatitle {The representation of the cumulative rounding error of
  an algorithm as a Taylor expansion of the local rounding errors} {The
  representation of the cumulative rounding error of an algorithm as a taylor
  expansion of the local rounding errors}.{\BBCQ}
\newblock
\APACjournalVolNumPages{Master's Thesis (in Finnish), Univ.
  Helsinki}{}{}{6--7}.
\PrintBackRefs{\CurrentBib}

\bibitem [\protect \citeauthoryear {%
Lotter%
, Kreiman%
\BCBL {}\ \BBA {} Cox%
}{%
Lotter%
\ \protect \BOthers {.}}{%
{\protect \APACyear {2016}}%
}]{%
lotter2016deep}
\APACinsertmetastar {%
lotter2016deep}%
\begin{APACrefauthors}%
Lotter, W.%
, Kreiman, G.%
\BCBL {}\ \BBA {} Cox, D.%
\end{APACrefauthors}%
\unskip\
\newblock
\APACrefYearMonthDay{2016}{}{}.
\newblock
{\BBOQ}\APACrefatitle {Deep predictive coding networks for video prediction and
  unsupervised learning} {Deep predictive coding networks for video prediction
  and unsupervised learning}.{\BBCQ}
\newblock
\APACjournalVolNumPages{arXiv preprint arXiv:1605.08104}{}{}{}.
\PrintBackRefs{\CurrentBib}

\bibitem [\protect \citeauthoryear {%
Ma%
, Chen%
\BCBL {}\ \BBA {} Fox%
}{%
Ma%
\ \protect \BOthers {.}}{%
{\protect \APACyear {2015}}%
}]{%
ma2015complete}
\APACinsertmetastar {%
ma2015complete}%
\begin{APACrefauthors}%
Ma, Y\BHBI A.%
, Chen, T.%
\BCBL {}\ \BBA {} Fox, E\BPBI B.%
\end{APACrefauthors}%
\unskip\
\newblock
\APACrefYearMonthDay{2015}{}{}.
\newblock
{\BBOQ}\APACrefatitle {A complete recipe for stochastic gradient MCMC} {A
  complete recipe for stochastic gradient mcmc}.{\BBCQ}
\newblock
\APACjournalVolNumPages{arXiv preprint arXiv:1506.04696}{}{}{}.
\PrintBackRefs{\CurrentBib}

\bibitem [\protect \citeauthoryear {%
Marino%
, Yue%
\BCBL {}\ \BBA {} Mandt%
}{%
Marino%
\ \protect \BOthers {.}}{%
{\protect \APACyear {2018}}%
}]{%
marino2018iterative}
\APACinsertmetastar {%
marino2018iterative}%
\begin{APACrefauthors}%
Marino, J.%
, Yue, Y.%
\BCBL {}\ \BBA {} Mandt, S.%
\end{APACrefauthors}%
\unskip\
\newblock
\APACrefYearMonthDay{2018}{}{}.
\newblock
{\BBOQ}\APACrefatitle {Iterative amortized inference} {Iterative amortized
  inference}.{\BBCQ}
\newblock
\APACjournalVolNumPages{arXiv preprint arXiv:1807.09356}{}{}{}.
\PrintBackRefs{\CurrentBib}

\bibitem [\protect \citeauthoryear {%
Marr%
}{%
Marr%
}{%
{\protect \APACyear {1982}}%
}]{%
marr1982vision}
\APACinsertmetastar {%
marr1982vision}%
\begin{APACrefauthors}%
Marr, D.%
\end{APACrefauthors}%
\unskip\
\newblock
\APACrefYearMonthDay{1982}{}{}.
\newblock
{\BBOQ}\APACrefatitle {Vision: A computational investigation into the human
  representation and processing of visual information} {Vision: A computational
  investigation into the human representation and processing of visual
  information}.{\BBCQ}
\newblock

\PrintBackRefs{\CurrentBib}

\bibitem [\protect \citeauthoryear {%
Maturana%
\ \BBA {} Varela%
}{%
Maturana%
\ \BBA {} Varela%
}{%
{\protect \APACyear {2012}}%
}]{%
maturana2012autopoiesis}
\APACinsertmetastar {%
maturana2012autopoiesis}%
\begin{APACrefauthors}%
Maturana, H\BPBI R.%
\BCBT {}\ \BBA {} Varela, F\BPBI J.%
\end{APACrefauthors}%
\unskip\
\newblock
\APACrefYear{2012}.
\newblock
\APACrefbtitle {Autopoiesis and cognition: The realization of the living}
  {Autopoiesis and cognition: The realization of the living}\ (\BVOL~42).
\newblock
\APACaddressPublisher{}{Springer Science \& Business Media}.
\PrintBackRefs{\CurrentBib}

\bibitem [\protect \citeauthoryear {%
Mehlhorn%
\ \protect \BOthers {.}}{%
Mehlhorn%
\ \protect \BOthers {.}}{%
{\protect \APACyear {2015}}%
}]{%
mehlhorn2015unpacking}
\APACinsertmetastar {%
mehlhorn2015unpacking}%
\begin{APACrefauthors}%
Mehlhorn, K.%
, Newell, B\BPBI R.%
, Todd, P\BPBI M.%
, Lee, M\BPBI D.%
, Morgan, K.%
, Braithwaite, V\BPBI A.%
\BDBL {}Gonzalez, C.%
\end{APACrefauthors}%
\unskip\
\newblock
\APACrefYearMonthDay{2015}{}{}.
\newblock
{\BBOQ}\APACrefatitle {Unpacking the exploration--exploitation tradeoff: A
  synthesis of human and animal literatures.} {Unpacking the
  exploration--exploitation tradeoff: A synthesis of human and animal
  literatures.}{\BBCQ}
\newblock
\APACjournalVolNumPages{Decision}{2}{3}{191}.
\PrintBackRefs{\CurrentBib}

\bibitem [\protect \citeauthoryear {%
Metropolis%
, Rosenbluth%
, Rosenbluth%
, Teller%
\BCBL {}\ \BBA {} Teller%
}{%
Metropolis%
\ \protect \BOthers {.}}{%
{\protect \APACyear {1953}}%
}]{%
metropolis1953equation}
\APACinsertmetastar {%
metropolis1953equation}%
\begin{APACrefauthors}%
Metropolis, N.%
, Rosenbluth, A\BPBI W.%
, Rosenbluth, M\BPBI N.%
, Teller, A\BPBI H.%
\BCBL {}\ \BBA {} Teller, E.%
\end{APACrefauthors}%
\unskip\
\newblock
\APACrefYearMonthDay{1953}{}{}.
\newblock
{\BBOQ}\APACrefatitle {Equation of state calculations by fast computing
  machines} {Equation of state calculations by fast computing machines}.{\BBCQ}
\newblock
\APACjournalVolNumPages{The Journal of Chemical Physics}{21}{6}{1087--1092}.
\PrintBackRefs{\CurrentBib}

\bibitem [\protect \citeauthoryear {%
Meulemans%
, Carzaniga%
, Suykens%
, Sacramento%
\BCBL {}\ \BBA {} Grewe%
}{%
Meulemans%
\ \protect \BOthers {.}}{%
{\protect \APACyear {2020}}%
}]{%
meulemans2020theoretical}
\APACinsertmetastar {%
meulemans2020theoretical}%
\begin{APACrefauthors}%
Meulemans, A.%
, Carzaniga, F\BPBI S.%
, Suykens, J\BPBI A.%
, Sacramento, J.%
\BCBL {}\ \BBA {} Grewe, B\BPBI F.%
\end{APACrefauthors}%
\unskip\
\newblock
\APACrefYearMonthDay{2020}{}{}.
\newblock
{\BBOQ}\APACrefatitle {A Theoretical Framework for Target Propagation} {A
  theoretical framework for target propagation}.{\BBCQ}
\newblock
\APACjournalVolNumPages{arXiv preprint arXiv:2006.14331}{}{}{}.
\PrintBackRefs{\CurrentBib}

\bibitem [\protect \citeauthoryear {%
Millidge%
}{%
Millidge%
}{%
{\protect \APACyear {2019}}%
{\protect \APACexlab {{\protect \BCnt {1}}}}}]{%
millidge2019combining}
\APACinsertmetastar {%
millidge2019combining}%
\begin{APACrefauthors}%
Millidge, B.%
\end{APACrefauthors}%
\unskip\
\newblock
\APACrefYearMonthDay{2019{\protect \BCnt {1}}}{}{}.
\newblock
{\BBOQ}\APACrefatitle {Combining active inference and hierarchical predictive
  coding: A tutorial introduction and case study} {Combining active inference
  and hierarchical predictive coding: A tutorial introduction and case
  study}.{\BBCQ}
\newblock

\PrintBackRefs{\CurrentBib}

\bibitem [\protect \citeauthoryear {%
Millidge%
}{%
Millidge%
}{%
{\protect \APACyear {2019}}%
{\protect \APACexlab {{\protect \BCnt {2}}}}}]{%
millidge2019deep}
\APACinsertmetastar {%
millidge2019deep}%
\begin{APACrefauthors}%
Millidge, B.%
\end{APACrefauthors}%
\unskip\
\newblock
\APACrefYearMonthDay{2019{\protect \BCnt {2}}}{}{}.
\newblock
{\BBOQ}\APACrefatitle {Deep active inference as variational policy gradients}
  {Deep active inference as variational policy gradients}.{\BBCQ}
\newblock
\APACjournalVolNumPages{arXiv preprint arXiv:1907.03876}{}{}{}.
\PrintBackRefs{\CurrentBib}

\bibitem [\protect \citeauthoryear {%
Millidge%
}{%
Millidge%
}{%
{\protect \APACyear {2019}}%
{\protect \APACexlab {{\protect \BCnt {3}}}}}]{%
millidge2019implementing}
\APACinsertmetastar {%
millidge2019implementing}%
\begin{APACrefauthors}%
Millidge, B.%
\end{APACrefauthors}%
\unskip\
\newblock
\APACrefYearMonthDay{2019{\protect \BCnt {3}}}{}{}.
\newblock
{\BBOQ}\APACrefatitle {Implementing Predictive Processing and Active Inference:
  Preliminary Steps and Results} {Implementing predictive processing and active
  inference: Preliminary steps and results}.{\BBCQ}
\newblock

\PrintBackRefs{\CurrentBib}

\bibitem [\protect \citeauthoryear {%
Millidge%
}{%
Millidge%
}{%
{\protect \APACyear {2020}}%
}]{%
millidge_deep_2019}
\APACinsertmetastar {%
millidge_deep_2019}%
\begin{APACrefauthors}%
Millidge, B.%
\end{APACrefauthors}%
\unskip\
\newblock
\APACrefYearMonthDay{2020}{}{}.
\newblock
{\BBOQ}\APACrefatitle {Deep active inference as variational policy gradients}
  {Deep active inference as variational policy gradients}.{\BBCQ}
\newblock
\APACjournalVolNumPages{Journal of Mathematical Psychology}{96}{}{102348}.
\PrintBackRefs{\CurrentBib}

\bibitem [\protect \citeauthoryear {%
Millidge%
, Tschantz%
\BCBL {}\ \BBA {} Buckley%
}{%
Millidge%
, Tschantz%
\BCBL {}\ \BBA {} Buckley%
}{%
{\protect \APACyear {2020}}%
{\protect \APACexlab {{\protect \BCnt {1}}}}}]{%
millidge2020predictive}
\APACinsertmetastar {%
millidge2020predictive}%
\begin{APACrefauthors}%
Millidge, B.%
, Tschantz, A.%
\BCBL {}\ \BBA {} Buckley, C\BPBI L.%
\end{APACrefauthors}%
\unskip\
\newblock
\APACrefYearMonthDay{2020{\protect \BCnt {1}}}{}{}.
\newblock
{\BBOQ}\APACrefatitle {Predictive Coding Approximates Backprop along Arbitrary
  Computation Graphs} {Predictive coding approximates backprop along arbitrary
  computation graphs}.{\BBCQ}
\newblock
\APACjournalVolNumPages{arXiv preprint arXiv:2006.04182}{}{}{}.
\PrintBackRefs{\CurrentBib}

\bibitem [\protect \citeauthoryear {%
Millidge%
, Tschantz%
\BCBL {}\ \BBA {} Buckley%
}{%
Millidge%
, Tschantz%
\BCBL {}\ \BBA {} Buckley%
}{%
{\protect \APACyear {2020}}%
{\protect \APACexlab {{\protect \BCnt {2}}}}}]{%
millidge2020whence}
\APACinsertmetastar {%
millidge2020whence}%
\begin{APACrefauthors}%
Millidge, B.%
, Tschantz, A.%
\BCBL {}\ \BBA {} Buckley, C\BPBI L.%
\end{APACrefauthors}%
\unskip\
\newblock
\APACrefYearMonthDay{2020{\protect \BCnt {2}}}{}{}.
\newblock
{\BBOQ}\APACrefatitle {Whence the Expected Free Energy?} {Whence the expected
  free energy?}{\BBCQ}
\newblock
\APACjournalVolNumPages{arXiv preprint arXiv:2004.08128}{}{}{}.
\PrintBackRefs{\CurrentBib}

\bibitem [\protect \citeauthoryear {%
Millidge%
, Tschantz%
, Buckley%
\BCBL {}\ \BBA {} Seth%
}{%
Millidge%
, Tschantz%
, Buckley%
\BCBL {}\ \BBA {} Seth%
}{%
{\protect \APACyear {2020}}%
}]{%
millidge2020activation}
\APACinsertmetastar {%
millidge2020activation}%
\begin{APACrefauthors}%
Millidge, B.%
, Tschantz, A.%
, Buckley, C\BPBI L.%
\BCBL {}\ \BBA {} Seth, A.%
\end{APACrefauthors}%
\unskip\
\newblock
\APACrefYearMonthDay{2020}{}{}.
\newblock
{\BBOQ}\APACrefatitle {Activation Relaxation: A Local Dynamical Approximation
  to Backpropagation in the Brain} {Activation relaxation: A local dynamical
  approximation to backpropagation in the brain}.{\BBCQ}
\newblock
\APACjournalVolNumPages{arXiv preprint arXiv:2009.05359}{}{}{}.
\PrintBackRefs{\CurrentBib}

\bibitem [\protect \citeauthoryear {%
Millidge%
, Tschantz%
, Seth%
\BCBL {}\ \BBA {} Buckley%
}{%
Millidge%
\ \protect \BOthers {.}}{%
{\protect \APACyear {2021}}%
}]{%
millidge2021understanding}
\APACinsertmetastar {%
millidge2021understanding}%
\begin{APACrefauthors}%
Millidge, B.%
, Tschantz, A.%
, Seth, A.%
\BCBL {}\ \BBA {} Buckley, C.%
\end{APACrefauthors}%
\unskip\
\newblock
\APACrefYearMonthDay{2021}{}{}.
\newblock
{\BBOQ}\APACrefatitle {Understanding the origin of information-seeking
  exploration in probabilistic objectives for control} {Understanding the
  origin of information-seeking exploration in probabilistic objectives for
  control}.{\BBCQ}
\newblock
\APACjournalVolNumPages{arXiv preprint arXiv:2103.06859}{}{}{}.
\PrintBackRefs{\CurrentBib}

\bibitem [\protect \citeauthoryear {%
Millidge%
, Tschantz%
, Seth%
\BCBL {}\ \BBA {} Buckley%
}{%
Millidge%
, Tschantz%
, Seth%
\BCBL {}\ \BBA {} Buckley%
}{%
{\protect \APACyear {2020}}%
{\protect \APACexlab {{\protect \BCnt {1}}}}}]{%
millidge2020investigating}
\APACinsertmetastar {%
millidge2020investigating}%
\begin{APACrefauthors}%
Millidge, B.%
, Tschantz, A.%
, Seth, A.%
\BCBL {}\ \BBA {} Buckley, C\BPBI L.%
\end{APACrefauthors}%
\unskip\
\newblock
\APACrefYearMonthDay{2020{\protect \BCnt {1}}}{}{}.
\newblock
{\BBOQ}\APACrefatitle {Investigating the Scalability and Biological
  Plausibility of the Activation Relaxation Algorithm} {Investigating the
  scalability and biological plausibility of the activation relaxation
  algorithm}.{\BBCQ}
\newblock
\APACjournalVolNumPages{arXiv preprint arXiv:2010.06219}{}{}{}.
\PrintBackRefs{\CurrentBib}

\bibitem [\protect \citeauthoryear {%
Millidge%
, Tschantz%
, Seth%
\BCBL {}\ \BBA {} Buckley%
}{%
Millidge%
, Tschantz%
, Seth%
\BCBL {}\ \BBA {} Buckley%
}{%
{\protect \APACyear {2020}}%
{\protect \APACexlab {{\protect \BCnt {4}}}}}]{%
millidge2020relaxing}
\APACinsertmetastar {%
millidge2020relaxing}%
\begin{APACrefauthors}%
Millidge, B.%
, Tschantz, A.%
, Seth, A.%
\BCBL {}\ \BBA {} Buckley, C\BPBI L.%
\end{APACrefauthors}%
\unskip\
\newblock
\APACrefYearMonthDay{2020{\protect \BCnt {4}}}{}{}.
\newblock
{\BBOQ}\APACrefatitle {Relaxing the constraints on predictive coding models}
  {Relaxing the constraints on predictive coding models}.{\BBCQ}
\newblock
\APACjournalVolNumPages{arXiv preprint arXiv:2010.01047}{}{}{}.
\PrintBackRefs{\CurrentBib}

\bibitem [\protect \citeauthoryear {%
Millidge%
, Tschantz%
, Seth%
\BCBL {}\ \BBA {} Buckley%
}{%
Millidge%
, Tschantz%
, Seth%
\BCBL {}\ \BBA {} Buckley%
}{%
{\protect \APACyear {2020}}%
{\protect \APACexlab {{\protect \BCnt {2}}}}}]{%
millidge2020relationship}
\APACinsertmetastar {%
millidge2020relationship}%
\begin{APACrefauthors}%
Millidge, B.%
, Tschantz, A.%
, Seth, A\BPBI K.%
\BCBL {}\ \BBA {} Buckley, C\BPBI L.%
\end{APACrefauthors}%
\unskip\
\newblock
\APACrefYearMonthDay{2020{\protect \BCnt {2}}}{}{}.
\newblock
{\BBOQ}\APACrefatitle {On the Relationship Between Active Inference and Control
  as Inference} {On the relationship between active inference and control as
  inference}.{\BBCQ}
\newblock
\APACjournalVolNumPages{arXiv preprint arXiv:2006.12964}{}{}{}.
\PrintBackRefs{\CurrentBib}

\bibitem [\protect \citeauthoryear {%
Millidge%
, Tschantz%
, Seth%
\BCBL {}\ \BBA {} Buckley%
}{%
Millidge%
, Tschantz%
, Seth%
\BCBL {}\ \BBA {} Buckley%
}{%
{\protect \APACyear {2020}}%
{\protect \APACexlab {{\protect \BCnt {3}}}}}]{%
millidge2020reinforcement}
\APACinsertmetastar {%
millidge2020reinforcement}%
\begin{APACrefauthors}%
Millidge, B.%
, Tschantz, A.%
, Seth, A\BPBI K.%
\BCBL {}\ \BBA {} Buckley, C\BPBI L.%
\end{APACrefauthors}%
\unskip\
\newblock
\APACrefYearMonthDay{2020{\protect \BCnt {3}}}{}{}.
\newblock
{\BBOQ}\APACrefatitle {Reinforcement Learning as Iterative and Amortised
  Inference} {Reinforcement learning as iterative and amortised
  inference}.{\BBCQ}
\newblock
\APACjournalVolNumPages{arXiv preprint arXiv:2006.10524}{}{}{}.
\PrintBackRefs{\CurrentBib}

\bibitem [\protect \citeauthoryear {%
Mirchev%
, Kayalibay%
, Soelch%
, van~der Smagt%
\BCBL {}\ \BBA {} Bayer%
}{%
Mirchev%
\ \protect \BOthers {.}}{%
{\protect \APACyear {2018}}%
}]{%
mirchev_approximate_2018}
\APACinsertmetastar {%
mirchev_approximate_2018}%
\begin{APACrefauthors}%
Mirchev, A.%
, Kayalibay, B.%
, Soelch, M.%
, van~der Smagt, P.%
\BCBL {}\ \BBA {} Bayer, J.%
\end{APACrefauthors}%
\unskip\
\newblock
\APACrefYearMonthDay{2018}{}{}.
\newblock
{\BBOQ}\APACrefatitle {Approximate bayesian inference in spatial environments}
  {Approximate bayesian inference in spatial environments}.{\BBCQ}
\newblock
\APACjournalVolNumPages{arXiv preprint arXiv:1805.07206}{}{}{}.
\PrintBackRefs{\CurrentBib}

\bibitem [\protect \citeauthoryear {%
Mirza%
, Adams%
, Parr%
\BCBL {}\ \BBA {} Friston%
}{%
Mirza%
\ \protect \BOthers {.}}{%
{\protect \APACyear {2019}}%
}]{%
mirza2019impulsivity}
\APACinsertmetastar {%
mirza2019impulsivity}%
\begin{APACrefauthors}%
Mirza, M\BPBI B.%
, Adams, R\BPBI A.%
, Parr, T.%
\BCBL {}\ \BBA {} Friston, K.%
\end{APACrefauthors}%
\unskip\
\newblock
\APACrefYearMonthDay{2019}{}{}.
\newblock
{\BBOQ}\APACrefatitle {Impulsivity and active inference} {Impulsivity and
  active inference}.{\BBCQ}
\newblock
\APACjournalVolNumPages{Journal of Cognitive Neuroscience}{31}{2}{202--220}.
\PrintBackRefs{\CurrentBib}

\bibitem [\protect \citeauthoryear {%
A.~Mnih%
\ \BBA {} Gregor%
}{%
A.~Mnih%
\ \BBA {} Gregor%
}{%
{\protect \APACyear {2014}}%
}]{%
mnih2014neural}
\APACinsertmetastar {%
mnih2014neural}%
\begin{APACrefauthors}%
Mnih, A.%
\BCBT {}\ \BBA {} Gregor, K.%
\end{APACrefauthors}%
\unskip\
\newblock
\APACrefYearMonthDay{2014}{}{}.
\newblock
{\BBOQ}\APACrefatitle {Neural variational inference and learning in belief
  networks} {Neural variational inference and learning in belief
  networks}.{\BBCQ}
\newblock
\APACjournalVolNumPages{arXiv preprint arXiv:1402.0030}{}{}{}.
\PrintBackRefs{\CurrentBib}

\bibitem [\protect \citeauthoryear {%
V.~Mnih%
\ \protect \BOthers {.}}{%
V.~Mnih%
\ \protect \BOthers {.}}{%
{\protect \APACyear {2016}}%
}]{%
mnih2016asynchronous}
\APACinsertmetastar {%
mnih2016asynchronous}%
\begin{APACrefauthors}%
Mnih, V.%
, Badia, A\BPBI P.%
, Mirza, M.%
, Graves, A.%
, Lillicrap, T.%
, Harley, T.%
\BDBL {}Kavukcuoglu, K.%
\end{APACrefauthors}%
\unskip\
\newblock
\APACrefYearMonthDay{2016}{}{}.
\newblock
{\BBOQ}\APACrefatitle {Asynchronous methods for deep reinforcement learning}
  {Asynchronous methods for deep reinforcement learning}.{\BBCQ}
\newblock
\BIn{} \APACrefbtitle {International conference on machine learning}
  {International conference on machine learning}\ (\BPGS\ 1928--1937).
\PrintBackRefs{\CurrentBib}

\bibitem [\protect \citeauthoryear {%
V.~Mnih%
\ \protect \BOthers {.}}{%
V.~Mnih%
\ \protect \BOthers {.}}{%
{\protect \APACyear {2013}}%
}]{%
mnih2013playing}
\APACinsertmetastar {%
mnih2013playing}%
\begin{APACrefauthors}%
Mnih, V.%
, Kavukcuoglu, K.%
, Silver, D.%
, Graves, A.%
, Antonoglou, I.%
, Wierstra, D.%
\BCBL {}\ \BBA {} Riedmiller, M.%
\end{APACrefauthors}%
\unskip\
\newblock
\APACrefYearMonthDay{2013}{}{}.
\newblock
{\BBOQ}\APACrefatitle {Playing atari with deep reinforcement learning} {Playing
  atari with deep reinforcement learning}.{\BBCQ}
\newblock
\APACjournalVolNumPages{arXiv preprint arXiv:1312.5602}{}{}{}.
\PrintBackRefs{\CurrentBib}

\bibitem [\protect \citeauthoryear {%
V.~Mnih%
\ \protect \BOthers {.}}{%
V.~Mnih%
\ \protect \BOthers {.}}{%
{\protect \APACyear {2015}}%
}]{%
mnih2015human}
\APACinsertmetastar {%
mnih2015human}%
\begin{APACrefauthors}%
Mnih, V.%
, Kavukcuoglu, K.%
, Silver, D.%
, Rusu, A\BPBI A.%
, Veness, J.%
, Bellemare, M\BPBI G.%
\BDBL {}others%
\end{APACrefauthors}%
\unskip\
\newblock
\APACrefYearMonthDay{2015}{}{}.
\newblock
{\BBOQ}\APACrefatitle {Human-level control through deep reinforcement learning}
  {Human-level control through deep reinforcement learning}.{\BBCQ}
\newblock
\APACjournalVolNumPages{Nature}{518}{7540}{529--533}.
\PrintBackRefs{\CurrentBib}

\bibitem [\protect \citeauthoryear {%
Mobbs%
, Trimmer%
, Blumstein%
\BCBL {}\ \BBA {} Dayan%
}{%
Mobbs%
\ \protect \BOthers {.}}{%
{\protect \APACyear {2018}}%
}]{%
mobbs2018foraging}
\APACinsertmetastar {%
mobbs2018foraging}%
\begin{APACrefauthors}%
Mobbs, D.%
, Trimmer, P\BPBI C.%
, Blumstein, D\BPBI T.%
\BCBL {}\ \BBA {} Dayan, P.%
\end{APACrefauthors}%
\unskip\
\newblock
\APACrefYearMonthDay{2018}{}{}.
\newblock
{\BBOQ}\APACrefatitle {Foraging for foundations in decision neuroscience:
  insights from ethology} {Foraging for foundations in decision neuroscience:
  insights from ethology}.{\BBCQ}
\newblock
\APACjournalVolNumPages{Nature Reviews Neuroscience}{19}{7}{419--427}.
\PrintBackRefs{\CurrentBib}

\bibitem [\protect \citeauthoryear {%
Mumford%
}{%
Mumford%
}{%
{\protect \APACyear {1992}}%
}]{%
mumford1992computational}
\APACinsertmetastar {%
mumford1992computational}%
\begin{APACrefauthors}%
Mumford, D.%
\end{APACrefauthors}%
\unskip\
\newblock
\APACrefYearMonthDay{1992}{}{}.
\newblock
{\BBOQ}\APACrefatitle {On the computational architecture of the neocortex} {On
  the computational architecture of the neocortex}.{\BBCQ}
\newblock
\APACjournalVolNumPages{Biological Cybernetics}{66}{3}{241--251}.
\PrintBackRefs{\CurrentBib}

\bibitem [\protect \citeauthoryear {%
Munuera%
, Morel%
, Duhamel%
\BCBL {}\ \BBA {} Deneve%
}{%
Munuera%
\ \protect \BOthers {.}}{%
{\protect \APACyear {2009}}%
}]{%
munuera2009optimal}
\APACinsertmetastar {%
munuera2009optimal}%
\begin{APACrefauthors}%
Munuera, J.%
, Morel, P.%
, Duhamel, J\BHBI R.%
\BCBL {}\ \BBA {} Deneve, S.%
\end{APACrefauthors}%
\unskip\
\newblock
\APACrefYearMonthDay{2009}{}{}.
\newblock
{\BBOQ}\APACrefatitle {Optimal sensorimotor control in eye movement sequences}
  {Optimal sensorimotor control in eye movement sequences}.{\BBCQ}
\newblock
\APACjournalVolNumPages{Journal of Neuroscience}{29}{10}{3026--3035}.
\PrintBackRefs{\CurrentBib}

\bibitem [\protect \citeauthoryear {%
Nagabandi%
, Kahn%
, Fearing%
\BCBL {}\ \BBA {} Levine%
}{%
Nagabandi%
\ \protect \BOthers {.}}{%
{\protect \APACyear {2018}}%
}]{%
nagabandi2018neural}
\APACinsertmetastar {%
nagabandi2018neural}%
\begin{APACrefauthors}%
Nagabandi, A.%
, Kahn, G.%
, Fearing, R\BPBI S.%
\BCBL {}\ \BBA {} Levine, S.%
\end{APACrefauthors}%
\unskip\
\newblock
\APACrefYearMonthDay{2018}{}{}.
\newblock
{\BBOQ}\APACrefatitle {Neural network dynamics for model-based deep
  reinforcement learning with model-free fine-tuning} {Neural network dynamics
  for model-based deep reinforcement learning with model-free
  fine-tuning}.{\BBCQ}
\newblock
\BIn{} \APACrefbtitle {2018 IEEE International Conference on Robotics and
  Automation (ICRA)} {2018 ieee international conference on robotics and
  automation (icra)}\ (\BPGS\ 7559--7566).
\PrintBackRefs{\CurrentBib}

\bibitem [\protect \citeauthoryear {%
Nagabandi%
, Konoglie%
, Levine%
\BCBL {}\ \BBA {} Kumar%
}{%
Nagabandi%
\ \protect \BOthers {.}}{%
{\protect \APACyear {2019}}%
}]{%
nagabandi2019deep}
\APACinsertmetastar {%
nagabandi2019deep}%
\begin{APACrefauthors}%
Nagabandi, A.%
, Konoglie, K.%
, Levine, S.%
\BCBL {}\ \BBA {} Kumar, V.%
\end{APACrefauthors}%
\unskip\
\newblock
\APACrefYearMonthDay{2019}{}{}.
\newblock
{\BBOQ}\APACrefatitle {Deep Dynamics Models for Learning Dexterous
  Manipulation} {Deep dynamics models for learning dexterous
  manipulation}.{\BBCQ}
\newblock
\APACjournalVolNumPages{arXiv preprint arXiv:1909.11652}{}{}{}.
\PrintBackRefs{\CurrentBib}

\bibitem [\protect \citeauthoryear {%
Neal%
\ \BBA {} Hinton%
}{%
Neal%
\ \BBA {} Hinton%
}{%
{\protect \APACyear {1998}}%
}]{%
neal1998view}
\APACinsertmetastar {%
neal1998view}%
\begin{APACrefauthors}%
Neal, R\BPBI M.%
\BCBT {}\ \BBA {} Hinton, G.%
\end{APACrefauthors}%
\unskip\
\newblock
\APACrefYearMonthDay{1998}{}{}.
\newblock
{\BBOQ}\APACrefatitle {A view of the EM algorithm that justifies incremental,
  sparse, and other variants} {A view of the em algorithm that justifies
  incremental, sparse, and other variants}.{\BBCQ}
\newblock
\BIn{} \APACrefbtitle {Learning in graphical models} {Learning in graphical
  models}\ (\BPGS\ 355--368).
\newblock
\APACaddressPublisher{}{Springer}.
\PrintBackRefs{\CurrentBib}

\bibitem [\protect \citeauthoryear {%
Neal%
\ \protect \BOthers {.}}{%
Neal%
\ \protect \BOthers {.}}{%
{\protect \APACyear {2011}}%
}]{%
neal2011MCMC}
\APACinsertmetastar {%
neal2011MCMC}%
\begin{APACrefauthors}%
Neal, R\BPBI M.%
\BCBT {}\ \BOthersPeriod {.}
\end{APACrefauthors}%
\unskip\
\newblock
\APACrefYearMonthDay{2011}{}{}.
\newblock
{\BBOQ}\APACrefatitle {MCMC using Hamiltonian dynamics} {Mcmc using hamiltonian
  dynamics}.{\BBCQ}
\newblock
\APACjournalVolNumPages{Handbook of Markov Chain Monte Carlo}{2}{11}{2}.
\PrintBackRefs{\CurrentBib}

\bibitem [\protect \citeauthoryear {%
Neftci%
, Mostafa%
\BCBL {}\ \BBA {} Zenke%
}{%
Neftci%
\ \protect \BOthers {.}}{%
{\protect \APACyear {2019}}%
}]{%
neftci2019surrogate}
\APACinsertmetastar {%
neftci2019surrogate}%
\begin{APACrefauthors}%
Neftci, E\BPBI O.%
, Mostafa, H.%
\BCBL {}\ \BBA {} Zenke, F.%
\end{APACrefauthors}%
\unskip\
\newblock
\APACrefYearMonthDay{2019}{}{}.
\newblock
{\BBOQ}\APACrefatitle {Surrogate gradient learning in spiking neural networks:
  Bringing the power of gradient-based optimization to spiking neural networks}
  {Surrogate gradient learning in spiking neural networks: Bringing the power
  of gradient-based optimization to spiking neural networks}.{\BBCQ}
\newblock
\APACjournalVolNumPages{IEEE Signal Processing Magazine}{36}{6}{51--63}.
\PrintBackRefs{\CurrentBib}

\bibitem [\protect \citeauthoryear {%
Nesterov%
}{%
Nesterov%
}{%
{\protect \APACyear {1983}}%
}]{%
nesterov27method}
\APACinsertmetastar {%
nesterov27method}%
\begin{APACrefauthors}%
Nesterov, Y.%
\end{APACrefauthors}%
\unskip\
\newblock
\APACrefYearMonthDay{1983}{}{}.
\newblock
{\BBOQ}\APACrefatitle {A method of solving a convex programming problem with
  convergence rate O (1/k\^{} 2) O (1/k2)} {A method of solving a convex
  programming problem with convergence rate o (1/k\^{} 2) o (1/k2)}.{\BBCQ}
\newblock
\BIn{} \APACrefbtitle {Sov. Math. Dokl} {Sov. math. dokl}\ (\BVOL~27).
\PrintBackRefs{\CurrentBib}

\bibitem [\protect \citeauthoryear {%
N{\o}kland%
}{%
N{\o}kland%
}{%
{\protect \APACyear {2016}}%
}]{%
nokland2016direct}
\APACinsertmetastar {%
nokland2016direct}%
\begin{APACrefauthors}%
N{\o}kland, A.%
\end{APACrefauthors}%
\unskip\
\newblock
\APACrefYearMonthDay{2016}{}{}.
\newblock
{\BBOQ}\APACrefatitle {Direct feedback alignment provides learning in deep
  neural networks} {Direct feedback alignment provides learning in deep neural
  networks}.{\BBCQ}
\newblock
\BIn{} \APACrefbtitle {Advances in neural Information Processing Systems}
  {Advances in neural information processing systems}\ (\BPGS\ 1037--1045).
\PrintBackRefs{\CurrentBib}

\bibitem [\protect \citeauthoryear {%
Okada%
, Kosaka%
\BCBL {}\ \BBA {} Taniguchi%
}{%
Okada%
\ \protect \BOthers {.}}{%
{\protect \APACyear {2020}}%
}]{%
okada2020planet}
\APACinsertmetastar {%
okada2020planet}%
\begin{APACrefauthors}%
Okada, M.%
, Kosaka, N.%
\BCBL {}\ \BBA {} Taniguchi, T.%
\end{APACrefauthors}%
\unskip\
\newblock
\APACrefYearMonthDay{2020}{}{}.
\newblock
{\BBOQ}\APACrefatitle {PlaNet of the Bayesians: Reconsidering and Improving
  Deep Planning Network by Incorporating Bayesian Inference} {Planet of the
  bayesians: Reconsidering and improving deep planning network by incorporating
  bayesian inference}.{\BBCQ}
\newblock
\APACjournalVolNumPages{arXiv preprint arXiv:2003.00370}{}{}{}.
\PrintBackRefs{\CurrentBib}

\bibitem [\protect \citeauthoryear {%
Okada%
\ \BBA {} Taniguchi%
}{%
Okada%
\ \BBA {} Taniguchi%
}{%
{\protect \APACyear {2020}}%
}]{%
okada_variational_2019}
\APACinsertmetastar {%
okada_variational_2019}%
\begin{APACrefauthors}%
Okada, M.%
\BCBT {}\ \BBA {} Taniguchi, T.%
\end{APACrefauthors}%
\unskip\
\newblock
\APACrefYearMonthDay{2020}{}{}.
\newblock
{\BBOQ}\APACrefatitle {Variational inference mpc for bayesian model-based
  reinforcement learning} {Variational inference mpc for bayesian model-based
  reinforcement learning}.{\BBCQ}
\newblock
\BIn{} \APACrefbtitle {Conference on Robot Learning} {Conference on robot
  learning}\ (\BPGS\ 258--272).
\PrintBackRefs{\CurrentBib}

\bibitem [\protect \citeauthoryear {%
Olah%
, Mordvintsev%
\BCBL {}\ \BBA {} Schubert%
}{%
Olah%
\ \protect \BOthers {.}}{%
{\protect \APACyear {2017}}%
}]{%
olah2017feature}
\APACinsertmetastar {%
olah2017feature}%
\begin{APACrefauthors}%
Olah, C.%
, Mordvintsev, A.%
\BCBL {}\ \BBA {} Schubert, L.%
\end{APACrefauthors}%
\unskip\
\newblock
\APACrefYearMonthDay{2017}{}{}.
\newblock
{\BBOQ}\APACrefatitle {Feature visualization} {Feature visualization}.{\BBCQ}
\newblock
\APACjournalVolNumPages{Distill}{2}{11}{e7}.
\PrintBackRefs{\CurrentBib}

\bibitem [\protect \citeauthoryear {%
Ollivier%
}{%
Ollivier%
}{%
{\protect \APACyear {2019}}%
}]{%
ollivier2019extended}
\APACinsertmetastar {%
ollivier2019extended}%
\begin{APACrefauthors}%
Ollivier, Y.%
\end{APACrefauthors}%
\unskip\
\newblock
\APACrefYearMonthDay{2019}{}{}.
\newblock
{\BBOQ}\APACrefatitle {The Extended Kalman Filter is a Natural Gradient Descent
  in Trajectory Space} {The extended kalman filter is a natural gradient
  descent in trajectory space}.{\BBCQ}
\newblock
\APACjournalVolNumPages{arXiv preprint arXiv:1901.00696}{}{}{}.
\PrintBackRefs{\CurrentBib}

\bibitem [\protect \citeauthoryear {%
Ollivier%
, Arnold%
, Auger%
\BCBL {}\ \BBA {} Hansen%
}{%
Ollivier%
\ \protect \BOthers {.}}{%
{\protect \APACyear {2017}}%
}]{%
ollivier2017information}
\APACinsertmetastar {%
ollivier2017information}%
\begin{APACrefauthors}%
Ollivier, Y.%
, Arnold, L.%
, Auger, A.%
\BCBL {}\ \BBA {} Hansen, N.%
\end{APACrefauthors}%
\unskip\
\newblock
\APACrefYearMonthDay{2017}{}{}.
\newblock
{\BBOQ}\APACrefatitle {Information-geometric optimization algorithms: A
  unifying picture via invariance principles} {Information-geometric
  optimization algorithms: A unifying picture via invariance
  principles}.{\BBCQ}
\newblock
\APACjournalVolNumPages{Journal of Machine Learning Research}{18}{18}{1--65}.
\PrintBackRefs{\CurrentBib}

\bibitem [\protect \citeauthoryear {%
Ollivier%
, Tallec%
\BCBL {}\ \BBA {} Charpiat%
}{%
Ollivier%
\ \protect \BOthers {.}}{%
{\protect \APACyear {2015}}%
}]{%
ollivier2015training}
\APACinsertmetastar {%
ollivier2015training}%
\begin{APACrefauthors}%
Ollivier, Y.%
, Tallec, C.%
\BCBL {}\ \BBA {} Charpiat, G.%
\end{APACrefauthors}%
\unskip\
\newblock
\APACrefYearMonthDay{2015}{}{}.
\newblock
{\BBOQ}\APACrefatitle {Training recurrent networks online without backtracking}
  {Training recurrent networks online without backtracking}.{\BBCQ}
\newblock
\APACjournalVolNumPages{arXiv preprint arXiv:1507.07680}{}{}{}.
\PrintBackRefs{\CurrentBib}

\bibitem [\protect \citeauthoryear {%
Orchard%
\ \BBA {} Sun%
}{%
Orchard%
\ \BBA {} Sun%
}{%
{\protect \APACyear {2019}}%
}]{%
orchard2019making}
\APACinsertmetastar {%
orchard2019making}%
\begin{APACrefauthors}%
Orchard, J.%
\BCBT {}\ \BBA {} Sun, W.%
\end{APACrefauthors}%
\unskip\
\newblock
\APACrefYearMonthDay{2019}{}{}.
\newblock
{\BBOQ}\APACrefatitle {Making Predictive Coding Networks Generative} {Making
  predictive coding networks generative}.{\BBCQ}
\newblock
\APACjournalVolNumPages{arXiv preprint arXiv:1910.12151}{}{}{}.
\newblock
\begin{APACrefURL} \url{https://arxiv.org/abs/1910.12151} \end{APACrefURL}
\PrintBackRefs{\CurrentBib}

\bibitem [\protect \citeauthoryear {%
O'Reilly%
, Wyatte%
\BCBL {}\ \BBA {} Rohrlich%
}{%
O'Reilly%
\ \protect \BOthers {.}}{%
{\protect \APACyear {2017}}%
}]{%
o2017deep}
\APACinsertmetastar {%
o2017deep}%
\begin{APACrefauthors}%
O'Reilly, R\BPBI C.%
, Wyatte, D\BPBI R.%
\BCBL {}\ \BBA {} Rohrlich, J.%
\end{APACrefauthors}%
\unskip\
\newblock
\APACrefYearMonthDay{2017}{}{}.
\newblock
{\BBOQ}\APACrefatitle {Deep predictive learning: a comprehensive model of three
  visual streams} {Deep predictive learning: a comprehensive model of three
  visual streams}.{\BBCQ}
\newblock
\APACjournalVolNumPages{arXiv preprint arXiv:1709.04654}{}{}{}.
\PrintBackRefs{\CurrentBib}

\bibitem [\protect \citeauthoryear {%
Ororbia%
\ \BBA {} Mali%
}{%
Ororbia%
\ \BBA {} Mali%
}{%
{\protect \APACyear {2019}}%
}]{%
ororbia2019biologically}
\APACinsertmetastar {%
ororbia2019biologically}%
\begin{APACrefauthors}%
Ororbia, A\BPBI G.%
\BCBT {}\ \BBA {} Mali, A.%
\end{APACrefauthors}%
\unskip\
\newblock
\APACrefYearMonthDay{2019}{}{}.
\newblock
{\BBOQ}\APACrefatitle {Biologically motivated algorithms for propagating local
  target representations} {Biologically motivated algorithms for propagating
  local target representations}.{\BBCQ}
\newblock
\BIn{} \APACrefbtitle {Proceedings of the AAAI Conference on Artificial
  Intelligence} {Proceedings of the aaai conference on artificial
  intelligence}\ (\BVOL~33, \BPGS\ 4651--4658).
\PrintBackRefs{\CurrentBib}

\bibitem [\protect \citeauthoryear {%
Ororbia~II%
, Haffner%
, Reitter%
\BCBL {}\ \BBA {} Giles%
}{%
Ororbia~II%
\ \protect \BOthers {.}}{%
{\protect \APACyear {2017}}%
}]{%
ororbia2017learning}
\APACinsertmetastar {%
ororbia2017learning}%
\begin{APACrefauthors}%
Ororbia~II, A\BPBI G.%
, Haffner, P.%
, Reitter, D.%
\BCBL {}\ \BBA {} Giles, C\BPBI L.%
\end{APACrefauthors}%
\unskip\
\newblock
\APACrefYearMonthDay{2017}{}{}.
\newblock
{\BBOQ}\APACrefatitle {Learning to adapt by minimizing discrepancy} {Learning
  to adapt by minimizing discrepancy}.{\BBCQ}
\newblock
\APACjournalVolNumPages{arXiv preprint arXiv:1711.11542}{}{}{}.
\PrintBackRefs{\CurrentBib}

\bibitem [\protect \citeauthoryear {%
Osband%
\ \BBA {} Van~Roy%
}{%
Osband%
\ \BBA {} Van~Roy%
}{%
{\protect \APACyear {2015}}%
}]{%
osband2015bootstrapped}
\APACinsertmetastar {%
osband2015bootstrapped}%
\begin{APACrefauthors}%
Osband, I.%
\BCBT {}\ \BBA {} Van~Roy, B.%
\end{APACrefauthors}%
\unskip\
\newblock
\APACrefYearMonthDay{2015}{}{}.
\newblock
{\BBOQ}\APACrefatitle {Bootstrapped thompson sampling and deep exploration}
  {Bootstrapped thompson sampling and deep exploration}.{\BBCQ}
\newblock
\APACjournalVolNumPages{arXiv preprint arXiv:1507.00300}{}{}{}.
\PrintBackRefs{\CurrentBib}

\bibitem [\protect \citeauthoryear {%
Osband%
, Van~Roy%
, Russo%
\BCBL {}\ \BBA {} Wen%
}{%
Osband%
\ \protect \BOthers {.}}{%
{\protect \APACyear {2019}}%
}]{%
osband2019deep}
\APACinsertmetastar {%
osband2019deep}%
\begin{APACrefauthors}%
Osband, I.%
, Van~Roy, B.%
, Russo, D\BPBI J.%
\BCBL {}\ \BBA {} Wen, Z.%
\end{APACrefauthors}%
\unskip\
\newblock
\APACrefYearMonthDay{2019}{}{}.
\newblock
{\BBOQ}\APACrefatitle {Deep Exploration via Randomized Value Functions.} {Deep
  exploration via randomized value functions.}{\BBCQ}
\newblock
\APACjournalVolNumPages{Journal of Machine Learning Research}{20}{124}{1--62}.
\PrintBackRefs{\CurrentBib}

\bibitem [\protect \citeauthoryear {%
Oudeyer%
\ \BBA {} Kaplan%
}{%
Oudeyer%
\ \BBA {} Kaplan%
}{%
{\protect \APACyear {2009}}%
}]{%
oudeyer2009intrinsic}
\APACinsertmetastar {%
oudeyer2009intrinsic}%
\begin{APACrefauthors}%
Oudeyer, P\BHBI Y.%
\BCBT {}\ \BBA {} Kaplan, F.%
\end{APACrefauthors}%
\unskip\
\newblock
\APACrefYearMonthDay{2009}{}{}.
\newblock
{\BBOQ}\APACrefatitle {What is intrinsic motivation? A typology of
  computational approaches} {What is intrinsic motivation? a typology of
  computational approaches}.{\BBCQ}
\newblock
\APACjournalVolNumPages{Frontiers in neurorobotics}{1}{}{6}.
\PrintBackRefs{\CurrentBib}

\bibitem [\protect \citeauthoryear {%
Ovchinnikov%
}{%
Ovchinnikov%
}{%
{\protect \APACyear {2016}}%
}]{%
ovchinnikov2016introduction}
\APACinsertmetastar {%
ovchinnikov2016introduction}%
\begin{APACrefauthors}%
Ovchinnikov, I\BPBI V.%
\end{APACrefauthors}%
\unskip\
\newblock
\APACrefYearMonthDay{2016}{}{}.
\newblock
{\BBOQ}\APACrefatitle {Introduction to supersymmetric theory of stochastics}
  {Introduction to supersymmetric theory of stochastics}.{\BBCQ}
\newblock
\APACjournalVolNumPages{Entropy}{18}{4}{108}.
\PrintBackRefs{\CurrentBib}

\bibitem [\protect \citeauthoryear {%
O’Reilly%
, Braver%
, Cohen%
\BCBL {}\ \protect \BOthers {.}}{%
O’Reilly%
\ \protect \BOthers {.}}{%
{\protect \APACyear {1999}}%
}]{%
o1999biologically}
\APACinsertmetastar {%
o1999biologically}%
\begin{APACrefauthors}%
O’Reilly, R\BPBI C.%
, Braver, T\BPBI S.%
, Cohen, J\BPBI D.%
\BCBL {}\ \BOthersPeriod {.}\end{APACrefauthors}%
\unskip\
\newblock
\APACrefYearMonthDay{1999}{}{}.
\newblock
{\BBOQ}\APACrefatitle {A biologically based computational model of working
  memory} {A biologically based computational model of working memory}.{\BBCQ}
\newblock
\APACjournalVolNumPages{Models of working memory: Mechanisms of active
  maintenance and executive control}{}{}{375--411}.
\PrintBackRefs{\CurrentBib}

\bibitem [\protect \citeauthoryear {%
Palacios%
, Razi%
, Parr%
, Kirchhoff%
\BCBL {}\ \BBA {} Friston%
}{%
Palacios%
\ \protect \BOthers {.}}{%
{\protect \APACyear {2017}}%
}]{%
palacios2017biological}
\APACinsertmetastar {%
palacios2017biological}%
\begin{APACrefauthors}%
Palacios, E\BPBI R.%
, Razi, A.%
, Parr, T.%
, Kirchhoff, M.%
\BCBL {}\ \BBA {} Friston, K.%
\end{APACrefauthors}%
\unskip\
\newblock
\APACrefYearMonthDay{2017}{}{}.
\newblock
{\BBOQ}\APACrefatitle {Biological self-organisation and Markov blankets}
  {Biological self-organisation and markov blankets}.{\BBCQ}
\newblock
\APACjournalVolNumPages{BioRxiv}{}{}{227181}.
\newblock
\begin{APACrefURL}
  \url{https://www.biorxiv.org/content/10.1101/227181v1.abstract}
  \end{APACrefURL}
\PrintBackRefs{\CurrentBib}

\bibitem [\protect \citeauthoryear {%
Parr%
}{%
Parr%
}{%
{\protect \APACyear {2019}}%
}]{%
parr2019computational}
\APACinsertmetastar {%
parr2019computational}%
\begin{APACrefauthors}%
Parr, T.%
\end{APACrefauthors}%
\unskip\
\newblock
\APACrefYear{2019}.
\unskip\
\newblock
\APACrefbtitle {The computational neurology of active vision} {The
  computational neurology of active vision}\ \APACtypeAddressSchool
  {\BUPhD}{}{}.
\unskip\
\newblock
\APACaddressSchool {}{UCL (University College London)}.
\PrintBackRefs{\CurrentBib}

\bibitem [\protect \citeauthoryear {%
Parr%
, Da~Costa%
\BCBL {}\ \BBA {} Friston%
}{%
Parr%
, Da~Costa%
\BCBL {}\ \BBA {} Friston%
}{%
{\protect \APACyear {2020}}%
}]{%
parr2020Markov}
\APACinsertmetastar {%
parr2020Markov}%
\begin{APACrefauthors}%
Parr, T.%
, Da~Costa, L.%
\BCBL {}\ \BBA {} Friston, K.%
\end{APACrefauthors}%
\unskip\
\newblock
\APACrefYearMonthDay{2020}{}{}.
\newblock
{\BBOQ}\APACrefatitle {Markov blankets, information geometry and stochastic
  thermodynamics} {Markov blankets, information geometry and stochastic
  thermodynamics}.{\BBCQ}
\newblock
\APACjournalVolNumPages{Philosophical Transactions of the Royal Society
  A}{378}{2164}{20190159}.
\PrintBackRefs{\CurrentBib}

\bibitem [\protect \citeauthoryear {%
Parr%
\ \BBA {} Friston%
}{%
Parr%
\ \BBA {} Friston%
}{%
{\protect \APACyear {2017}}%
{\protect \APACexlab {{\protect \BCnt {1}}}}}]{%
parr2017active}
\APACinsertmetastar {%
parr2017active}%
\begin{APACrefauthors}%
Parr, T.%
\BCBT {}\ \BBA {} Friston, K.%
\end{APACrefauthors}%
\unskip\
\newblock
\APACrefYearMonthDay{2017{\protect \BCnt {1}}}{}{}.
\newblock
{\BBOQ}\APACrefatitle {The active construction of the visual world} {The active
  construction of the visual world}.{\BBCQ}
\newblock
\APACjournalVolNumPages{Neuropsychologia}{104}{}{92--101}.
\PrintBackRefs{\CurrentBib}

\bibitem [\protect \citeauthoryear {%
Parr%
\ \BBA {} Friston%
}{%
Parr%
\ \BBA {} Friston%
}{%
{\protect \APACyear {2017}}%
{\protect \APACexlab {{\protect \BCnt {2}}}}}]{%
parr2017uncertainty}
\APACinsertmetastar {%
parr2017uncertainty}%
\begin{APACrefauthors}%
Parr, T.%
\BCBT {}\ \BBA {} Friston, K.%
\end{APACrefauthors}%
\unskip\
\newblock
\APACrefYearMonthDay{2017{\protect \BCnt {2}}}{}{}.
\newblock
{\BBOQ}\APACrefatitle {Uncertainty, epistemics and active inference}
  {Uncertainty, epistemics and active inference}.{\BBCQ}
\newblock
\APACjournalVolNumPages{Journal of The Royal Society
  Interface}{14}{136}{20170376}.
\PrintBackRefs{\CurrentBib}

\bibitem [\protect \citeauthoryear {%
Parr%
\ \BBA {} Friston%
}{%
Parr%
\ \BBA {} Friston%
}{%
{\protect \APACyear {2018}}%
{\protect \APACexlab {{\protect \BCnt {1}}}}}]{%
parr2018active}
\APACinsertmetastar {%
parr2018active}%
\begin{APACrefauthors}%
Parr, T.%
\BCBT {}\ \BBA {} Friston, K.%
\end{APACrefauthors}%
\unskip\
\newblock
\APACrefYearMonthDay{2018{\protect \BCnt {1}}}{}{}.
\newblock
{\BBOQ}\APACrefatitle {Active inference and the anatomy of oculomotion} {Active
  inference and the anatomy of oculomotion}.{\BBCQ}
\newblock
\APACjournalVolNumPages{Neuropsychologia}{111}{}{334--343}.
\PrintBackRefs{\CurrentBib}

\bibitem [\protect \citeauthoryear {%
Parr%
\ \BBA {} Friston%
}{%
Parr%
\ \BBA {} Friston%
}{%
{\protect \APACyear {2018}}%
{\protect \APACexlab {{\protect \BCnt {2}}}}}]{%
parr2018anatomy}
\APACinsertmetastar {%
parr2018anatomy}%
\begin{APACrefauthors}%
Parr, T.%
\BCBT {}\ \BBA {} Friston, K.%
\end{APACrefauthors}%
\unskip\
\newblock
\APACrefYearMonthDay{2018{\protect \BCnt {2}}}{}{}.
\newblock
{\BBOQ}\APACrefatitle {The anatomy of inference: Generative models and brain
  structure} {The anatomy of inference: Generative models and brain
  structure}.{\BBCQ}
\newblock
\APACjournalVolNumPages{Frontiers in computational neuroscience}{12}{}{}.
\PrintBackRefs{\CurrentBib}

\bibitem [\protect \citeauthoryear {%
Parr%
\ \BBA {} Friston%
}{%
Parr%
\ \BBA {} Friston%
}{%
{\protect \APACyear {2018}}%
{\protect \APACexlab {{\protect \BCnt {3}}}}}]{%
parr2018computational}
\APACinsertmetastar {%
parr2018computational}%
\begin{APACrefauthors}%
Parr, T.%
\BCBT {}\ \BBA {} Friston, K.%
\end{APACrefauthors}%
\unskip\
\newblock
\APACrefYearMonthDay{2018{\protect \BCnt {3}}}{}{}.
\newblock
{\BBOQ}\APACrefatitle {The computational anatomy of visual neglect} {The
  computational anatomy of visual neglect}.{\BBCQ}
\newblock
\APACjournalVolNumPages{Cerebral Cortex}{28}{2}{777--790}.
\PrintBackRefs{\CurrentBib}

\bibitem [\protect \citeauthoryear {%
Parr%
, Markovic%
, Kiebel%
\BCBL {}\ \BBA {} Friston%
}{%
Parr%
\ \protect \BOthers {.}}{%
{\protect \APACyear {2019}}%
}]{%
parr2019neuronal}
\APACinsertmetastar {%
parr2019neuronal}%
\begin{APACrefauthors}%
Parr, T.%
, Markovic, D.%
, Kiebel, S\BPBI J.%
\BCBL {}\ \BBA {} Friston, K.%
\end{APACrefauthors}%
\unskip\
\newblock
\APACrefYearMonthDay{2019}{}{}.
\newblock
{\BBOQ}\APACrefatitle {Neuronal message passing using Mean-field, Bethe, and
  Marginal approximations} {Neuronal message passing using mean-field, bethe,
  and marginal approximations}.{\BBCQ}
\newblock
\APACjournalVolNumPages{Scientific reports}{9}{1}{1--18}.
\PrintBackRefs{\CurrentBib}

\bibitem [\protect \citeauthoryear {%
Parr%
, Sajid%
\BCBL {}\ \BBA {} Friston%
}{%
Parr%
, Sajid%
\BCBL {}\ \BBA {} Friston%
}{%
{\protect \APACyear {2020}}%
}]{%
parr2020modules}
\APACinsertmetastar {%
parr2020modules}%
\begin{APACrefauthors}%
Parr, T.%
, Sajid, N.%
\BCBL {}\ \BBA {} Friston, K.%
\end{APACrefauthors}%
\unskip\
\newblock
\APACrefYearMonthDay{2020}{}{}.
\newblock
{\BBOQ}\APACrefatitle {Modules or Mean-Fields?} {Modules or
  mean-fields?}{\BBCQ}
\newblock
\APACjournalVolNumPages{Entropy}{22}{5}{552}.
\newblock
\begin{APACrefURL} \url{https://www.mdpi.com/1099-4300/22/5/552}
  \end{APACrefURL}
\PrintBackRefs{\CurrentBib}

\bibitem [\protect \citeauthoryear {%
Paszke%
\ \protect \BOthers {.}}{%
Paszke%
\ \protect \BOthers {.}}{%
{\protect \APACyear {2017}}%
}]{%
paszke2017automatic}
\APACinsertmetastar {%
paszke2017automatic}%
\begin{APACrefauthors}%
Paszke, A.%
, Gross, S.%
, Chintala, S.%
, Chanan, G.%
, Yang, E.%
, DeVito, Z.%
\BDBL {}Lerer, A.%
\end{APACrefauthors}%
\unskip\
\newblock
\APACrefYearMonthDay{2017}{}{}.
\newblock
{\BBOQ}\APACrefatitle {Automatic differentiation in pytorch} {Automatic
  differentiation in pytorch}.{\BBCQ}
\newblock

\PrintBackRefs{\CurrentBib}

\bibitem [\protect \citeauthoryear {%
Pathak%
, Agrawal%
, Efros%
\BCBL {}\ \BBA {} Darrell%
}{%
Pathak%
\ \protect \BOthers {.}}{%
{\protect \APACyear {2017}}%
}]{%
pathak2017curiosity}
\APACinsertmetastar {%
pathak2017curiosity}%
\begin{APACrefauthors}%
Pathak, D.%
, Agrawal, P.%
, Efros, A\BPBI A.%
\BCBL {}\ \BBA {} Darrell, T.%
\end{APACrefauthors}%
\unskip\
\newblock
\APACrefYearMonthDay{2017}{}{}.
\newblock
{\BBOQ}\APACrefatitle {Curiosity-driven exploration by self-supervised
  prediction} {Curiosity-driven exploration by self-supervised
  prediction}.{\BBCQ}
\newblock
\BIn{} \APACrefbtitle {Proceedings of the IEEE Conference on Computer Vision
  and Pattern Recognition Workshops} {Proceedings of the ieee conference on
  computer vision and pattern recognition workshops}\ (\BPGS\ 16--17).
\PrintBackRefs{\CurrentBib}

\bibitem [\protect \citeauthoryear {%
Pearl%
}{%
Pearl%
}{%
{\protect \APACyear {2011}}%
}]{%
pearl2011Bayesian}
\APACinsertmetastar {%
pearl2011Bayesian}%
\begin{APACrefauthors}%
Pearl, J.%
\end{APACrefauthors}%
\unskip\
\newblock
\APACrefYearMonthDay{2011}{}{}.
\newblock
{\BBOQ}\APACrefatitle {Bayesian networks} {Bayesian networks}.{\BBCQ}
\newblock

\PrintBackRefs{\CurrentBib}

\bibitem [\protect \citeauthoryear {%
Pearl%
}{%
Pearl%
}{%
{\protect \APACyear {2014}}%
}]{%
pearl2014probabilistic}
\APACinsertmetastar {%
pearl2014probabilistic}%
\begin{APACrefauthors}%
Pearl, J.%
\end{APACrefauthors}%
\unskip\
\newblock
\APACrefYear{2014}.
\newblock
\APACrefbtitle {Probabilistic reasoning in intelligent systems: networks of
  plausible inference} {Probabilistic reasoning in intelligent systems:
  networks of plausible inference}.
\newblock
\APACaddressPublisher{}{Elsevier}.
\PrintBackRefs{\CurrentBib}

\bibitem [\protect \citeauthoryear {%
Peters%
\ \BBA {} Schaal%
}{%
Peters%
\ \BBA {} Schaal%
}{%
{\protect \APACyear {2007}}%
}]{%
peters2007reinforcement}
\APACinsertmetastar {%
peters2007reinforcement}%
\begin{APACrefauthors}%
Peters, J.%
\BCBT {}\ \BBA {} Schaal, S.%
\end{APACrefauthors}%
\unskip\
\newblock
\APACrefYearMonthDay{2007}{}{}.
\newblock
{\BBOQ}\APACrefatitle {Reinforcement learning by reward-weighted regression for
  operational space control} {Reinforcement learning by reward-weighted
  regression for operational space control}.{\BBCQ}
\newblock
\BIn{} \APACrefbtitle {Proceedings of the 24th International Conference on
  Machine learning} {Proceedings of the 24th international conference on
  machine learning}\ (\BPGS\ 745--750).
\PrintBackRefs{\CurrentBib}

\bibitem [\protect \citeauthoryear {%
Pinker%
}{%
Pinker%
}{%
{\protect \APACyear {2003}}%
}]{%
pinker2003language}
\APACinsertmetastar {%
pinker2003language}%
\begin{APACrefauthors}%
Pinker, S.%
\end{APACrefauthors}%
\unskip\
\newblock
\APACrefYear{2003}.
\newblock
\APACrefbtitle {The language instinct: How the mind creates language} {The
  language instinct: How the mind creates language}.
\newblock
\APACaddressPublisher{}{Penguin UK}.
\PrintBackRefs{\CurrentBib}

\bibitem [\protect \citeauthoryear {%
Pio-Lopez%
, Nizard%
, Friston%
\BCBL {}\ \BBA {} Pezzulo%
}{%
Pio-Lopez%
\ \protect \BOthers {.}}{%
{\protect \APACyear {2016}}%
}]{%
pio2016active}
\APACinsertmetastar {%
pio2016active}%
\begin{APACrefauthors}%
Pio-Lopez, L.%
, Nizard, A.%
, Friston, K.%
\BCBL {}\ \BBA {} Pezzulo, G.%
\end{APACrefauthors}%
\unskip\
\newblock
\APACrefYearMonthDay{2016}{}{}.
\newblock
{\BBOQ}\APACrefatitle {Active inference and robot control: a case study}
  {Active inference and robot control: a case study}.{\BBCQ}
\newblock
\APACjournalVolNumPages{Journal of The Royal Society
  Interface}{13}{122}{20160616}.
\PrintBackRefs{\CurrentBib}

\bibitem [\protect \citeauthoryear {%
Pozzi%
, Boht{\'e}%
\BCBL {}\ \BBA {} Roelfsema%
}{%
Pozzi%
\ \protect \BOthers {.}}{%
{\protect \APACyear {2018}}%
}]{%
pozzi2018biologically}
\APACinsertmetastar {%
pozzi2018biologically}%
\begin{APACrefauthors}%
Pozzi, I.%
, Boht{\'e}, S.%
\BCBL {}\ \BBA {} Roelfsema, P.%
\end{APACrefauthors}%
\unskip\
\newblock
\APACrefYearMonthDay{2018}{}{}.
\newblock
{\BBOQ}\APACrefatitle {A biologically plausible learning rule for deep learning
  in the brain} {A biologically plausible learning rule for deep learning in
  the brain}.{\BBCQ}
\newblock
\APACjournalVolNumPages{arXiv preprint arXiv:1811.01768}{}{}{}.
\PrintBackRefs{\CurrentBib}

\bibitem [\protect \citeauthoryear {%
Prigogine%
}{%
Prigogine%
}{%
{\protect \APACyear {2017}}%
}]{%
prigogine2017non}
\APACinsertmetastar {%
prigogine2017non}%
\begin{APACrefauthors}%
Prigogine, I.%
\end{APACrefauthors}%
\unskip\
\newblock
\APACrefYear{2017}.
\newblock
\APACrefbtitle {Non-equilibrium statistical mechanics} {Non-equilibrium
  statistical mechanics}.
\newblock
\APACaddressPublisher{}{Courier Dover Publications}.
\PrintBackRefs{\CurrentBib}

\bibitem [\protect \citeauthoryear {%
Prigogine%
\ \BBA {} Lefever%
}{%
Prigogine%
\ \BBA {} Lefever%
}{%
{\protect \APACyear {1973}}%
}]{%
prigogine1973theory}
\APACinsertmetastar {%
prigogine1973theory}%
\begin{APACrefauthors}%
Prigogine, I.%
\BCBT {}\ \BBA {} Lefever, R.%
\end{APACrefauthors}%
\unskip\
\newblock
\APACrefYearMonthDay{1973}{}{}.
\newblock
{\BBOQ}\APACrefatitle {Theory of dissipative structures} {Theory of dissipative
  structures}.{\BBCQ}
\newblock
\BIn{} \APACrefbtitle {Synergetics} {Synergetics}\ (\BPGS\ 124--135).
\newblock
\APACaddressPublisher{}{Springer}.
\PrintBackRefs{\CurrentBib}

\bibitem [\protect \citeauthoryear {%
Pyke%
}{%
Pyke%
}{%
{\protect \APACyear {1984}}%
}]{%
pyke1984optimal}
\APACinsertmetastar {%
pyke1984optimal}%
\begin{APACrefauthors}%
Pyke, G\BPBI H.%
\end{APACrefauthors}%
\unskip\
\newblock
\APACrefYearMonthDay{1984}{}{}.
\newblock
{\BBOQ}\APACrefatitle {Optimal foraging theory: a critical review} {Optimal
  foraging theory: a critical review}.{\BBCQ}
\newblock
\APACjournalVolNumPages{Annual review of ecology and
  systematics}{15}{1}{523--575}.
\PrintBackRefs{\CurrentBib}

\bibitem [\protect \citeauthoryear {%
Radford%
\ \protect \BOthers {.}}{%
Radford%
\ \protect \BOthers {.}}{%
{\protect \APACyear {2021}}%
}]{%
radford2021learning}
\APACinsertmetastar {%
radford2021learning}%
\begin{APACrefauthors}%
Radford, A.%
, Kim, J\BPBI W.%
, Hallacy, C.%
, Ramesh, A.%
, Goh, G.%
, Agarwal, S.%
\BDBL {}others%
\end{APACrefauthors}%
\unskip\
\newblock
\APACrefYearMonthDay{2021}{}{}.
\newblock
{\BBOQ}\APACrefatitle {Learning transferable visual models from natural
  language supervision} {Learning transferable visual models from natural
  language supervision}.{\BBCQ}
\newblock
\APACjournalVolNumPages{arXiv preprint arXiv:2103.00020}{}{}{}.
\PrintBackRefs{\CurrentBib}

\bibitem [\protect \citeauthoryear {%
Radford%
\ \protect \BOthers {.}}{%
Radford%
\ \protect \BOthers {.}}{%
{\protect \APACyear {2019}}%
}]{%
radford2019language}
\APACinsertmetastar {%
radford2019language}%
\begin{APACrefauthors}%
Radford, A.%
, Wu, J.%
, Child, R.%
, Luan, D.%
, Amodei, D.%
\BCBL {}\ \BBA {} Sutskever, I.%
\end{APACrefauthors}%
\unskip\
\newblock
\APACrefYearMonthDay{2019}{}{}.
\newblock
{\BBOQ}\APACrefatitle {Language models are unsupervised multitask learners}
  {Language models are unsupervised multitask learners}.{\BBCQ}
\newblock
\APACjournalVolNumPages{OpenAI blog}{1}{8}{9}.
\PrintBackRefs{\CurrentBib}

\bibitem [\protect \citeauthoryear {%
Rao%
\ \BBA {} Ballard%
}{%
Rao%
\ \BBA {} Ballard%
}{%
{\protect \APACyear {1999}}%
}]{%
rao1999predictive}
\APACinsertmetastar {%
rao1999predictive}%
\begin{APACrefauthors}%
Rao, R\BPBI P.%
\BCBT {}\ \BBA {} Ballard, D\BPBI H.%
\end{APACrefauthors}%
\unskip\
\newblock
\APACrefYearMonthDay{1999}{}{}.
\newblock
{\BBOQ}\APACrefatitle {Predictive coding in the visual cortex: a functional
  interpretation of some extra-classical receptive-field effects} {Predictive
  coding in the visual cortex: a functional interpretation of some
  extra-classical receptive-field effects}.{\BBCQ}
\newblock
\APACjournalVolNumPages{Nature Neuroscience}{2}{1}{79--87}.
\PrintBackRefs{\CurrentBib}

\bibitem [\protect \citeauthoryear {%
K.~Rawlik%
, Toussaint%
\BCBL {}\ \BBA {} Vijayakumar%
}{%
K.~Rawlik%
\ \protect \BOthers {.}}{%
{\protect \APACyear {2013}}%
}]{%
rawlik2013stochastic}
\APACinsertmetastar {%
rawlik2013stochastic}%
\begin{APACrefauthors}%
Rawlik, K.%
, Toussaint, M.%
\BCBL {}\ \BBA {} Vijayakumar, S.%
\end{APACrefauthors}%
\unskip\
\newblock
\APACrefYearMonthDay{2013}{}{}.
\newblock
{\BBOQ}\APACrefatitle {On stochastic optimal control and reinforcement learning
  by approximate inference} {On stochastic optimal control and reinforcement
  learning by approximate inference}.{\BBCQ}
\newblock
\BIn{} \APACrefbtitle {Twenty-Third International Joint Conference on
  Artificial Intelligence.} {Twenty-third international joint conference on
  artificial intelligence.}
\PrintBackRefs{\CurrentBib}

\bibitem [\protect \citeauthoryear {%
K\BPBI C.~Rawlik%
}{%
K\BPBI C.~Rawlik%
}{%
{\protect \APACyear {2013}}%
}]{%
rawlik2013probabilistic}
\APACinsertmetastar {%
rawlik2013probabilistic}%
\begin{APACrefauthors}%
Rawlik, K\BPBI C.%
\end{APACrefauthors}%
\unskip\
\newblock
\APACrefYearMonthDay{2013}{}{}.
\newblock
{\BBOQ}\APACrefatitle {On probabilistic inference approaches to stochastic
  optimal control} {On probabilistic inference approaches to stochastic optimal
  control}.{\BBCQ}
\newblock

\PrintBackRefs{\CurrentBib}

\bibitem [\protect \citeauthoryear {%
Roelfsema%
\ \BBA {} Ooyen%
}{%
Roelfsema%
\ \BBA {} Ooyen%
}{%
{\protect \APACyear {2005}}%
}]{%
roelfsema2005attention}
\APACinsertmetastar {%
roelfsema2005attention}%
\begin{APACrefauthors}%
Roelfsema, P\BPBI R.%
\BCBT {}\ \BBA {} Ooyen, A\BPBI v.%
\end{APACrefauthors}%
\unskip\
\newblock
\APACrefYearMonthDay{2005}{}{}.
\newblock
{\BBOQ}\APACrefatitle {Attention-gated reinforcement learning of internal
  representations for classification} {Attention-gated reinforcement learning
  of internal representations for classification}.{\BBCQ}
\newblock
\APACjournalVolNumPages{Neural Computation}{17}{10}{2176--2214}.
\PrintBackRefs{\CurrentBib}

\bibitem [\protect \citeauthoryear {%
Rubinstein%
}{%
Rubinstein%
}{%
{\protect \APACyear {1997}}%
}]{%
rubinstein1997optimization}
\APACinsertmetastar {%
rubinstein1997optimization}%
\begin{APACrefauthors}%
Rubinstein, R\BPBI Y.%
\end{APACrefauthors}%
\unskip\
\newblock
\APACrefYearMonthDay{1997}{}{}.
\newblock
{\BBOQ}\APACrefatitle {Optimization of computer simulation models with rare
  events} {Optimization of computer simulation models with rare events}.{\BBCQ}
\newblock
\APACjournalVolNumPages{European Journal of Operational
  Research}{99}{1}{89--112}.
\PrintBackRefs{\CurrentBib}

\bibitem [\protect \citeauthoryear {%
Rumelhart%
, Hinton%
\BCBL {}\ \BBA {} Williams%
}{%
Rumelhart%
\ \protect \BOthers {.}}{%
{\protect \APACyear {1986}}%
}]{%
rumelhart1986learning}
\APACinsertmetastar {%
rumelhart1986learning}%
\begin{APACrefauthors}%
Rumelhart, D\BPBI E.%
, Hinton, G.%
\BCBL {}\ \BBA {} Williams, R\BPBI J.%
\end{APACrefauthors}%
\unskip\
\newblock
\APACrefYearMonthDay{1986}{}{}.
\newblock
{\BBOQ}\APACrefatitle {Learning representations by back-propagating errors}
  {Learning representations by back-propagating errors}.{\BBCQ}
\newblock
\APACjournalVolNumPages{nature}{323}{6088}{533--536}.
\PrintBackRefs{\CurrentBib}

\bibitem [\protect \citeauthoryear {%
Rumelhart%
\ \BBA {} Zipser%
}{%
Rumelhart%
\ \BBA {} Zipser%
}{%
{\protect \APACyear {1985}}%
}]{%
rumelhart1985feature}
\APACinsertmetastar {%
rumelhart1985feature}%
\begin{APACrefauthors}%
Rumelhart, D\BPBI E.%
\BCBT {}\ \BBA {} Zipser, D.%
\end{APACrefauthors}%
\unskip\
\newblock
\APACrefYearMonthDay{1985}{}{}.
\newblock
{\BBOQ}\APACrefatitle {Feature discovery by competitive learning} {Feature
  discovery by competitive learning}.{\BBCQ}
\newblock
\APACjournalVolNumPages{Cognitive science}{9}{1}{75--112}.
\PrintBackRefs{\CurrentBib}

\bibitem [\protect \citeauthoryear {%
Russo%
\ \BBA {} Van~Roy%
}{%
Russo%
\ \BBA {} Van~Roy%
}{%
{\protect \APACyear {2016}}%
}]{%
russo2016information}
\APACinsertmetastar {%
russo2016information}%
\begin{APACrefauthors}%
Russo, D.%
\BCBT {}\ \BBA {} Van~Roy, B.%
\end{APACrefauthors}%
\unskip\
\newblock
\APACrefYearMonthDay{2016}{}{}.
\newblock
{\BBOQ}\APACrefatitle {An information-theoretic analysis of thompson sampling}
  {An information-theoretic analysis of thompson sampling}.{\BBCQ}
\newblock
\APACjournalVolNumPages{The Journal of Machine Learning
  Research}{17}{1}{2442--2471}.
\PrintBackRefs{\CurrentBib}

\bibitem [\protect \citeauthoryear {%
Sacramento%
, Costa%
, Bengio%
\BCBL {}\ \BBA {} Senn%
}{%
Sacramento%
\ \protect \BOthers {.}}{%
{\protect \APACyear {2018}}%
}]{%
sacramento2018dendritic}
\APACinsertmetastar {%
sacramento2018dendritic}%
\begin{APACrefauthors}%
Sacramento, J.%
, Costa, R\BPBI P.%
, Bengio, Y.%
\BCBL {}\ \BBA {} Senn, W.%
\end{APACrefauthors}%
\unskip\
\newblock
\APACrefYearMonthDay{2018}{}{}.
\newblock
{\BBOQ}\APACrefatitle {Dendritic cortical microcircuits approximate the
  backpropagation algorithm} {Dendritic cortical microcircuits approximate the
  backpropagation algorithm}.{\BBCQ}
\newblock
\BIn{} \APACrefbtitle {Advances in Neural Information Processing Systems}
  {Advances in neural information processing systems}\ (\BPGS\ 8721--8732).
\PrintBackRefs{\CurrentBib}

\bibitem [\protect \citeauthoryear {%
Salimans%
, Ho%
, Chen%
, Sidor%
\BCBL {}\ \BBA {} Sutskever%
}{%
Salimans%
\ \protect \BOthers {.}}{%
{\protect \APACyear {2017}}%
}]{%
salimans2017evolution}
\APACinsertmetastar {%
salimans2017evolution}%
\begin{APACrefauthors}%
Salimans, T.%
, Ho, J.%
, Chen, X.%
, Sidor, S.%
\BCBL {}\ \BBA {} Sutskever, I.%
\end{APACrefauthors}%
\unskip\
\newblock
\APACrefYearMonthDay{2017}{}{}.
\newblock
{\BBOQ}\APACrefatitle {Evolution strategies as a scalable alternative to
  reinforcement learning} {Evolution strategies as a scalable alternative to
  reinforcement learning}.{\BBCQ}
\newblock
\APACjournalVolNumPages{arXiv preprint arXiv:1703.03864}{}{}{}.
\PrintBackRefs{\CurrentBib}

\bibitem [\protect \citeauthoryear {%
Sanborn%
\ \BBA {} Chater%
}{%
Sanborn%
\ \BBA {} Chater%
}{%
{\protect \APACyear {2016}}%
}]{%
sanborn2016Bayesian}
\APACinsertmetastar {%
sanborn2016Bayesian}%
\begin{APACrefauthors}%
Sanborn, A\BPBI N.%
\BCBT {}\ \BBA {} Chater, N.%
\end{APACrefauthors}%
\unskip\
\newblock
\APACrefYearMonthDay{2016}{}{}.
\newblock
{\BBOQ}\APACrefatitle {Bayesian brains without probabilities} {Bayesian brains
  without probabilities}.{\BBCQ}
\newblock
\APACjournalVolNumPages{Trends in Cognitive Sciences}{20}{12}{883--893}.
\PrintBackRefs{\CurrentBib}

\bibitem [\protect \citeauthoryear {%
S{\"a}rkk{\"a}%
}{%
S{\"a}rkk{\"a}%
}{%
{\protect \APACyear {2013}}%
}]{%
sarkka2013Bayesian}
\APACinsertmetastar {%
sarkka2013Bayesian}%
\begin{APACrefauthors}%
S{\"a}rkk{\"a}, S.%
\end{APACrefauthors}%
\unskip\
\newblock
\APACrefYear{2013}.
\newblock
\APACrefbtitle {Bayesian filtering and smoothing} {Bayesian filtering and
  smoothing}\ (\BNUM~3).
\newblock
\APACaddressPublisher{}{Cambridge University Press}.
\PrintBackRefs{\CurrentBib}

\bibitem [\protect \citeauthoryear {%
Scellier%
\ \BBA {} Bengio%
}{%
Scellier%
\ \BBA {} Bengio%
}{%
{\protect \APACyear {2016}}%
}]{%
scellier2016towards}
\APACinsertmetastar {%
scellier2016towards}%
\begin{APACrefauthors}%
Scellier, B.%
\BCBT {}\ \BBA {} Bengio, Y.%
\end{APACrefauthors}%
\unskip\
\newblock
\APACrefYearMonthDay{2016}{}{}.
\newblock
{\BBOQ}\APACrefatitle {Towards a biologically plausible backprop} {Towards a
  biologically plausible backprop}.{\BBCQ}
\newblock
\APACjournalVolNumPages{arXiv preprint arXiv:1602.05179}{914}{}{}.
\PrintBackRefs{\CurrentBib}

\bibitem [\protect \citeauthoryear {%
Scellier%
\ \BBA {} Bengio%
}{%
Scellier%
\ \BBA {} Bengio%
}{%
{\protect \APACyear {2017}}%
}]{%
scellier2017equilibrium}
\APACinsertmetastar {%
scellier2017equilibrium}%
\begin{APACrefauthors}%
Scellier, B.%
\BCBT {}\ \BBA {} Bengio, Y.%
\end{APACrefauthors}%
\unskip\
\newblock
\APACrefYearMonthDay{2017}{}{}.
\newblock
{\BBOQ}\APACrefatitle {Equilibrium propagation: Bridging the gap between
  energy-based models and backpropagation} {Equilibrium propagation: Bridging
  the gap between energy-based models and backpropagation}.{\BBCQ}
\newblock
\APACjournalVolNumPages{Frontiers in computational neuroscience}{11}{}{24}.
\PrintBackRefs{\CurrentBib}

\bibitem [\protect \citeauthoryear {%
Scellier%
, Goyal%
, Binas%
, Mesnard%
\BCBL {}\ \BBA {} Bengio%
}{%
Scellier%
\ \protect \BOthers {.}}{%
{\protect \APACyear {2018}}%
{\protect \APACexlab {{\protect \BCnt {1}}}}}]{%
scellier2018extending}
\APACinsertmetastar {%
scellier2018extending}%
\begin{APACrefauthors}%
Scellier, B.%
, Goyal, A.%
, Binas, J.%
, Mesnard, T.%
\BCBL {}\ \BBA {} Bengio, Y.%
\end{APACrefauthors}%
\unskip\
\newblock
\APACrefYearMonthDay{2018{\protect \BCnt {1}}}{}{}.
\newblock
{\BBOQ}\APACrefatitle {Extending the framework of equilibrium propagation to
  general dynamics} {Extending the framework of equilibrium propagation to
  general dynamics}.{\BBCQ}
\newblock

\PrintBackRefs{\CurrentBib}

\bibitem [\protect \citeauthoryear {%
Scellier%
, Goyal%
, Binas%
, Mesnard%
\BCBL {}\ \BBA {} Bengio%
}{%
Scellier%
\ \protect \BOthers {.}}{%
{\protect \APACyear {2018}}%
{\protect \APACexlab {{\protect \BCnt {2}}}}}]{%
scellier2018generalization}
\APACinsertmetastar {%
scellier2018generalization}%
\begin{APACrefauthors}%
Scellier, B.%
, Goyal, A.%
, Binas, J.%
, Mesnard, T.%
\BCBL {}\ \BBA {} Bengio, Y.%
\end{APACrefauthors}%
\unskip\
\newblock
\APACrefYearMonthDay{2018{\protect \BCnt {2}}}{}{}.
\newblock
{\BBOQ}\APACrefatitle {Generalization of equilibrium propagation to vector
  field dynamics} {Generalization of equilibrium propagation to vector field
  dynamics}.{\BBCQ}
\newblock
\APACjournalVolNumPages{arXiv preprint arXiv:1808.04873}{}{}{}.
\PrintBackRefs{\CurrentBib}

\bibitem [\protect \citeauthoryear {%
Schiess%
, Urbanczik%
\BCBL {}\ \BBA {} Senn%
}{%
Schiess%
\ \protect \BOthers {.}}{%
{\protect \APACyear {2016}}%
}]{%
schiess2016somato}
\APACinsertmetastar {%
schiess2016somato}%
\begin{APACrefauthors}%
Schiess, M.%
, Urbanczik, R.%
\BCBL {}\ \BBA {} Senn, W.%
\end{APACrefauthors}%
\unskip\
\newblock
\APACrefYearMonthDay{2016}{}{}.
\newblock
{\BBOQ}\APACrefatitle {Somato-dendritic synaptic plasticity and
  error-backpropagation in active dendrites} {Somato-dendritic synaptic
  plasticity and error-backpropagation in active dendrites}.{\BBCQ}
\newblock
\APACjournalVolNumPages{PLoS Computational Biology}{12}{2}{}.
\PrintBackRefs{\CurrentBib}

\bibitem [\protect \citeauthoryear {%
Schmidhuber%
}{%
Schmidhuber%
}{%
{\protect \APACyear {1991}}%
}]{%
schmidhuber1991possibility}
\APACinsertmetastar {%
schmidhuber1991possibility}%
\begin{APACrefauthors}%
Schmidhuber, J.%
\end{APACrefauthors}%
\unskip\
\newblock
\APACrefYearMonthDay{1991}{}{}.
\newblock
{\BBOQ}\APACrefatitle {A possibility for implementing curiosity and boredom in
  model-building neural controllers} {A possibility for implementing curiosity
  and boredom in model-building neural controllers}.{\BBCQ}
\newblock
\BIn{} \APACrefbtitle {Proc. of the International Conference on Simulation of
  Adaptive Behavior: From animals to animats} {Proc. of the international
  conference on simulation of adaptive behavior: From animals to animats}\
  (\BPGS\ 222--227).
\PrintBackRefs{\CurrentBib}

\bibitem [\protect \citeauthoryear {%
Schmidhuber%
}{%
Schmidhuber%
}{%
{\protect \APACyear {1999}}%
}]{%
schmidhuber1999artificial}
\APACinsertmetastar {%
schmidhuber1999artificial}%
\begin{APACrefauthors}%
Schmidhuber, J.%
\end{APACrefauthors}%
\unskip\
\newblock
\APACrefYearMonthDay{1999}{}{}.
\newblock
{\BBOQ}\APACrefatitle {Artificial curiosity based on discovering novel
  algorithmic predictability through coevolution} {Artificial curiosity based
  on discovering novel algorithmic predictability through coevolution}.{\BBCQ}
\newblock
\BIn{} \APACrefbtitle {Proceedings of the 1999 Congress on Evolutionary
  Computation-CEC99 (Cat. No. 99TH8406)} {Proceedings of the 1999 congress on
  evolutionary computation-cec99 (cat. no. 99th8406)}\ (\BVOL~3, \BPGS\
  1612--1618).
\PrintBackRefs{\CurrentBib}

\bibitem [\protect \citeauthoryear {%
Schmidhuber%
}{%
Schmidhuber%
}{%
{\protect \APACyear {2007}}%
}]{%
schmidhuber2007simple}
\APACinsertmetastar {%
schmidhuber2007simple}%
\begin{APACrefauthors}%
Schmidhuber, J.%
\end{APACrefauthors}%
\unskip\
\newblock
\APACrefYearMonthDay{2007}{}{}.
\newblock
{\BBOQ}\APACrefatitle {Simple algorithmic principles of discovery, subjective
  beauty, selective attention, curiosity \& creativity} {Simple algorithmic
  principles of discovery, subjective beauty, selective attention, curiosity \&
  creativity}.{\BBCQ}
\newblock
\BIn{} \APACrefbtitle {International Conference on Discovery Science}
  {International conference on discovery science}\ (\BPGS\ 26--38).
\PrintBackRefs{\CurrentBib}

\bibitem [\protect \citeauthoryear {%
Schneider%
}{%
Schneider%
}{%
{\protect \APACyear {1988}}%
}]{%
schneider1988analytical}
\APACinsertmetastar {%
schneider1988analytical}%
\begin{APACrefauthors}%
Schneider, W.%
\end{APACrefauthors}%
\unskip\
\newblock
\APACrefYearMonthDay{1988}{}{}.
\newblock
{\BBOQ}\APACrefatitle {Analytical uses of Kalman filtering in econometrics—A
  survey} {Analytical uses of kalman filtering in econometrics—a
  survey}.{\BBCQ}
\newblock
\APACjournalVolNumPages{Statistical Papers}{29}{1}{3--33}.
\PrintBackRefs{\CurrentBib}

\bibitem [\protect \citeauthoryear {%
Schrittwieser%
\ \protect \BOthers {.}}{%
Schrittwieser%
\ \protect \BOthers {.}}{%
{\protect \APACyear {2019}}%
}]{%
schrittwieser2019mastering}
\APACinsertmetastar {%
schrittwieser2019mastering}%
\begin{APACrefauthors}%
Schrittwieser, J.%
, Antonoglou, I.%
, Hubert, T.%
, Simonyan, K.%
, Sifre, L.%
, Schmitt, S.%
\BDBL {}others%
\end{APACrefauthors}%
\unskip\
\newblock
\APACrefYearMonthDay{2019}{}{}.
\newblock
{\BBOQ}\APACrefatitle {Mastering atari, go, chess and shogi by planning with a
  learned model} {Mastering atari, go, chess and shogi by planning with a
  learned model}.{\BBCQ}
\newblock
\APACjournalVolNumPages{arXiv preprint arXiv:1911.08265}{}{}{}.
\PrintBackRefs{\CurrentBib}

\bibitem [\protect \citeauthoryear {%
Schulman%
, Levine%
, Abbeel%
, Jordan%
\BCBL {}\ \BBA {} Moritz%
}{%
Schulman%
\ \protect \BOthers {.}}{%
{\protect \APACyear {2015}}%
}]{%
schulman2015trust}
\APACinsertmetastar {%
schulman2015trust}%
\begin{APACrefauthors}%
Schulman, J.%
, Levine, S.%
, Abbeel, P.%
, Jordan, M.%
\BCBL {}\ \BBA {} Moritz, P.%
\end{APACrefauthors}%
\unskip\
\newblock
\APACrefYearMonthDay{2015}{}{}.
\newblock
{\BBOQ}\APACrefatitle {Trust region policy optimization} {Trust region policy
  optimization}.{\BBCQ}
\newblock
\BIn{} \APACrefbtitle {International conference on machine learning}
  {International conference on machine learning}\ (\BPGS\ 1889--1897).
\PrintBackRefs{\CurrentBib}

\bibitem [\protect \citeauthoryear {%
Schulman%
, Wolski%
, Dhariwal%
, Radford%
\BCBL {}\ \BBA {} Klimov%
}{%
Schulman%
\ \protect \BOthers {.}}{%
{\protect \APACyear {2017}}%
}]{%
schulman2017proximal}
\APACinsertmetastar {%
schulman2017proximal}%
\begin{APACrefauthors}%
Schulman, J.%
, Wolski, F.%
, Dhariwal, P.%
, Radford, A.%
\BCBL {}\ \BBA {} Klimov, O.%
\end{APACrefauthors}%
\unskip\
\newblock
\APACrefYearMonthDay{2017}{}{}.
\newblock
{\BBOQ}\APACrefatitle {Proximal policy optimization algorithms} {Proximal
  policy optimization algorithms}.{\BBCQ}
\newblock
\APACjournalVolNumPages{arXiv preprint arXiv:1707.06347}{}{}{}.
\PrintBackRefs{\CurrentBib}

\bibitem [\protect \citeauthoryear {%
Schultz%
}{%
Schultz%
}{%
{\protect \APACyear {1998}}%
}]{%
schultz1998predictive}
\APACinsertmetastar {%
schultz1998predictive}%
\begin{APACrefauthors}%
Schultz, W.%
\end{APACrefauthors}%
\unskip\
\newblock
\APACrefYearMonthDay{1998}{}{}.
\newblock
{\BBOQ}\APACrefatitle {Predictive reward signal of dopamine neurons}
  {Predictive reward signal of dopamine neurons}.{\BBCQ}
\newblock
\APACjournalVolNumPages{Journal of neurophysiology}{80}{1}{1--27}.
\PrintBackRefs{\CurrentBib}

\bibitem [\protect \citeauthoryear {%
Schultz%
, Tremblay%
\BCBL {}\ \BBA {} Hollerman%
}{%
Schultz%
\ \protect \BOthers {.}}{%
{\protect \APACyear {1998}}%
}]{%
schultz1998reward}
\APACinsertmetastar {%
schultz1998reward}%
\begin{APACrefauthors}%
Schultz, W.%
, Tremblay, L.%
\BCBL {}\ \BBA {} Hollerman, J\BPBI R.%
\end{APACrefauthors}%
\unskip\
\newblock
\APACrefYearMonthDay{1998}{}{}.
\newblock
{\BBOQ}\APACrefatitle {Reward prediction in primate basal ganglia and frontal
  cortex} {Reward prediction in primate basal ganglia and frontal
  cortex}.{\BBCQ}
\newblock
\APACjournalVolNumPages{Neuropharmacology}{37}{4-5}{421--429}.
\PrintBackRefs{\CurrentBib}

\bibitem [\protect \citeauthoryear {%
Schwartenbeck%
, FitzGerald%
, Dolan%
\BCBL {}\ \BBA {} Friston%
}{%
Schwartenbeck%
\ \protect \BOthers {.}}{%
{\protect \APACyear {2013}}%
}]{%
schwartenbeck2013exploration}
\APACinsertmetastar {%
schwartenbeck2013exploration}%
\begin{APACrefauthors}%
Schwartenbeck, P.%
, FitzGerald, T.%
, Dolan, R.%
\BCBL {}\ \BBA {} Friston, K.%
\end{APACrefauthors}%
\unskip\
\newblock
\APACrefYearMonthDay{2013}{}{}.
\newblock
{\BBOQ}\APACrefatitle {Exploration, novelty, surprise, and free energy
  minimization} {Exploration, novelty, surprise, and free energy
  minimization}.{\BBCQ}
\newblock
\APACjournalVolNumPages{Frontiers in Psychology}{4}{}{710}.
\PrintBackRefs{\CurrentBib}

\bibitem [\protect \citeauthoryear {%
Schwartenbeck%
\ \protect \BOthers {.}}{%
Schwartenbeck%
\ \protect \BOthers {.}}{%
{\protect \APACyear {2015}}%
}]{%
schwartenbeck2015optimal}
\APACinsertmetastar {%
schwartenbeck2015optimal}%
\begin{APACrefauthors}%
Schwartenbeck, P.%
, FitzGerald, T\BPBI H.%
, Mathys, C.%
, Dolan, R.%
, Wurst, F.%
, Kronbichler, M.%
\BCBL {}\ \BBA {} Friston, K.%
\end{APACrefauthors}%
\unskip\
\newblock
\APACrefYearMonthDay{2015}{}{}.
\newblock
{\BBOQ}\APACrefatitle {Optimal inference with suboptimal models: addiction and
  active Bayesian inference} {Optimal inference with suboptimal models:
  addiction and active bayesian inference}.{\BBCQ}
\newblock
\APACjournalVolNumPages{Medical Hypotheses}{84}{2}{109--117}.
\PrintBackRefs{\CurrentBib}

\bibitem [\protect \citeauthoryear {%
Schwartenbeck%
\ \protect \BOthers {.}}{%
Schwartenbeck%
\ \protect \BOthers {.}}{%
{\protect \APACyear {2019}}%
}]{%
schwartenbeck_computational_2019}
\APACinsertmetastar {%
schwartenbeck_computational_2019}%
\begin{APACrefauthors}%
Schwartenbeck, P.%
, Passecker, J.%
, Hauser, T\BPBI U.%
, {FitzGerald}, T\BPBI H.%
, Kronbichler, M.%
\BCBL {}\ \BBA {} Friston, K.%
\end{APACrefauthors}%
\unskip\
\newblock
\APACrefYearMonthDay{2019}{}{}.
\newblock
{\BBOQ}\APACrefatitle {Computational mechanisms of curiosity and goal-directed
  exploration} {Computational mechanisms of curiosity and goal-directed
  exploration}.{\BBCQ}
\newblock
\APACjournalVolNumPages{}{8}{}{e41703}.
\newblock
\begin{APACrefURL} [{2019-11-15}]\url{https://doi.org/10.7554/eLife.41703}
  \end{APACrefURL}
\newblock
\begin{APACrefDOI} \doi{10.7554/eLife.41703} \end{APACrefDOI}
\PrintBackRefs{\CurrentBib}

\bibitem [\protect \citeauthoryear {%
Schw{\"o}bel%
, Kiebel%
\BCBL {}\ \BBA {} Markovi{\'c}%
}{%
Schw{\"o}bel%
\ \protect \BOthers {.}}{%
{\protect \APACyear {2018}}%
}]{%
schwobel2018active}
\APACinsertmetastar {%
schwobel2018active}%
\begin{APACrefauthors}%
Schw{\"o}bel, S.%
, Kiebel, S.%
\BCBL {}\ \BBA {} Markovi{\'c}, D.%
\end{APACrefauthors}%
\unskip\
\newblock
\APACrefYearMonthDay{2018}{}{}.
\newblock
{\BBOQ}\APACrefatitle {Active inference, belief propagation, and the bethe
  approximation} {Active inference, belief propagation, and the bethe
  approximation}.{\BBCQ}
\newblock
\APACjournalVolNumPages{Neural Computation}{30}{9}{2530--2567}.
\PrintBackRefs{\CurrentBib}

\bibitem [\protect \citeauthoryear {%
Seifert%
}{%
Seifert%
}{%
{\protect \APACyear {2008}}%
}]{%
seifert2008stochastic}
\APACinsertmetastar {%
seifert2008stochastic}%
\begin{APACrefauthors}%
Seifert, U.%
\end{APACrefauthors}%
\unskip\
\newblock
\APACrefYearMonthDay{2008}{}{}.
\newblock
{\BBOQ}\APACrefatitle {Stochastic thermodynamics: principles and perspectives}
  {Stochastic thermodynamics: principles and perspectives}.{\BBCQ}
\newblock
\APACjournalVolNumPages{The European Physical Journal B}{64}{3}{423--431}.
\PrintBackRefs{\CurrentBib}

\bibitem [\protect \citeauthoryear {%
Seifert%
}{%
Seifert%
}{%
{\protect \APACyear {2012}}%
}]{%
seifert2012stochastic}
\APACinsertmetastar {%
seifert2012stochastic}%
\begin{APACrefauthors}%
Seifert, U.%
\end{APACrefauthors}%
\unskip\
\newblock
\APACrefYearMonthDay{2012}{}{}.
\newblock
{\BBOQ}\APACrefatitle {Stochastic thermodynamics, fluctuation theorems and
  molecular machines} {Stochastic thermodynamics, fluctuation theorems and
  molecular machines}.{\BBCQ}
\newblock
\APACjournalVolNumPages{Reports on Progress in Physics}{75}{12}{126001}.
\PrintBackRefs{\CurrentBib}

\bibitem [\protect \citeauthoryear {%
Seth%
}{%
Seth%
}{%
{\protect \APACyear {2014}}%
}]{%
seth2014cybernetic}
\APACinsertmetastar {%
seth2014cybernetic}%
\begin{APACrefauthors}%
Seth, A\BPBI K.%
\end{APACrefauthors}%
\unskip\
\newblock
\APACrefYear{2014}.
\newblock
\APACrefbtitle {The cybernetic Bayesian brain} {The cybernetic bayesian brain}.
\newblock
\APACaddressPublisher{}{Open MIND. Frankfurt am Main: MIND Group}.
\newblock
\begin{APACrefURL}
  \url{https://open-mind.net/papers/the-cybernetic-bayesian-brain}
  \end{APACrefURL}
\PrintBackRefs{\CurrentBib}

\bibitem [\protect \citeauthoryear {%
Sethi%
\ \BBA {} Thompson%
}{%
Sethi%
\ \BBA {} Thompson%
}{%
{\protect \APACyear {2000}}%
}]{%
sethi2000optimal}
\APACinsertmetastar {%
sethi2000optimal}%
\begin{APACrefauthors}%
Sethi, S\BPBI P.%
\BCBT {}\ \BBA {} Thompson, G\BPBI L.%
\end{APACrefauthors}%
\unskip\
\newblock
\APACrefYear{2000}.
\newblock
\APACrefbtitle {What is optimal control theory?} {What is optimal control
  theory?}
\newblock
\APACaddressPublisher{}{Springer}.
\PrintBackRefs{\CurrentBib}

\bibitem [\protect \citeauthoryear {%
Seung%
}{%
Seung%
}{%
{\protect \APACyear {2003}}%
}]{%
seung2003learning}
\APACinsertmetastar {%
seung2003learning}%
\begin{APACrefauthors}%
Seung, H\BPBI S.%
\end{APACrefauthors}%
\unskip\
\newblock
\APACrefYearMonthDay{2003}{}{}.
\newblock
{\BBOQ}\APACrefatitle {Learning in spiking neural networks by reinforcement of
  stochastic synaptic transmission} {Learning in spiking neural networks by
  reinforcement of stochastic synaptic transmission}.{\BBCQ}
\newblock
\APACjournalVolNumPages{Neuron}{40}{6}{1063--1073}.
\PrintBackRefs{\CurrentBib}

\bibitem [\protect \citeauthoryear {%
Shanks%
, Tunney%
\BCBL {}\ \BBA {} McCarthy%
}{%
Shanks%
\ \protect \BOthers {.}}{%
{\protect \APACyear {2002}}%
}]{%
shanks2002re}
\APACinsertmetastar {%
shanks2002re}%
\begin{APACrefauthors}%
Shanks, D\BPBI R.%
, Tunney, R\BPBI J.%
\BCBL {}\ \BBA {} McCarthy, J\BPBI D.%
\end{APACrefauthors}%
\unskip\
\newblock
\APACrefYearMonthDay{2002}{}{}.
\newblock
{\BBOQ}\APACrefatitle {A re-examination of probability matching and rational
  choice} {A re-examination of probability matching and rational
  choice}.{\BBCQ}
\newblock
\APACjournalVolNumPages{Journal of Behavioral Decision
  Making}{15}{3}{233--250}.
\PrintBackRefs{\CurrentBib}

\bibitem [\protect \citeauthoryear {%
Shannon%
}{%
Shannon%
}{%
{\protect \APACyear {1948}}%
}]{%
shannon1948mathematical}
\APACinsertmetastar {%
shannon1948mathematical}%
\begin{APACrefauthors}%
Shannon, C\BPBI E.%
\end{APACrefauthors}%
\unskip\
\newblock
\APACrefYearMonthDay{1948}{}{}.
\newblock
{\BBOQ}\APACrefatitle {A mathematical theory of communication} {A mathematical
  theory of communication}.{\BBCQ}
\newblock
\APACjournalVolNumPages{The Bell system technical journal}{27}{3}{379--423}.
\PrintBackRefs{\CurrentBib}

\bibitem [\protect \citeauthoryear {%
Shipp%
}{%
Shipp%
}{%
{\protect \APACyear {2016}}%
}]{%
shipp2016neural}
\APACinsertmetastar {%
shipp2016neural}%
\begin{APACrefauthors}%
Shipp, S.%
\end{APACrefauthors}%
\unskip\
\newblock
\APACrefYearMonthDay{2016}{}{}.
\newblock
{\BBOQ}\APACrefatitle {Neural elements for predictive coding} {Neural elements
  for predictive coding}.{\BBCQ}
\newblock
\APACjournalVolNumPages{Frontiers in Psychology}{7}{}{1792}.
\PrintBackRefs{\CurrentBib}

\bibitem [\protect \citeauthoryear {%
Shipp%
, Adams%
\BCBL {}\ \BBA {} Friston%
}{%
Shipp%
\ \protect \BOthers {.}}{%
{\protect \APACyear {2013}}%
}]{%
shipp2013reflections}
\APACinsertmetastar {%
shipp2013reflections}%
\begin{APACrefauthors}%
Shipp, S.%
, Adams, R\BPBI A.%
\BCBL {}\ \BBA {} Friston, K.%
\end{APACrefauthors}%
\unskip\
\newblock
\APACrefYearMonthDay{2013}{}{}.
\newblock
{\BBOQ}\APACrefatitle {Reflections on agranular architecture: predictive coding
  in the motor cortex} {Reflections on agranular architecture: predictive
  coding in the motor cortex}.{\BBCQ}
\newblock
\APACjournalVolNumPages{Trends in neurosciences}{36}{12}{706--716}.
\PrintBackRefs{\CurrentBib}

\bibitem [\protect \citeauthoryear {%
Shyam%
, Jaśkowski%
\BCBL {}\ \BBA {} Gomez%
}{%
Shyam%
\ \protect \BOthers {.}}{%
{\protect \APACyear {2019}}%
}]{%
shyam_model-based_2019}
\APACinsertmetastar {%
shyam_model-based_2019}%
\begin{APACrefauthors}%
Shyam, P.%
, Jaśkowski, W.%
\BCBL {}\ \BBA {} Gomez, F.%
\end{APACrefauthors}%
\unskip\
\newblock
\APACrefYearMonthDay{2019}{}{}.
\newblock
{\BBOQ}\APACrefatitle {Model-Based Active Exploration} {Model-based active
  exploration}.{\BBCQ}
\newblock
\BIn{} \APACrefbtitle {International Conference on Machine Learning}
  {International conference on machine learning}\ (\BPGS\ 5779--5788).
\newblock
\begin{APACrefURL}
  [{2019-10-11}]\url{http://proceedings.mlr.press/v97/shyam19a.html}
  \end{APACrefURL}
\PrintBackRefs{\CurrentBib}

\bibitem [\protect \citeauthoryear {%
Silver%
\ \protect \BOthers {.}}{%
Silver%
\ \protect \BOthers {.}}{%
{\protect \APACyear {2016}}%
}]{%
silver2016mastering}
\APACinsertmetastar {%
silver2016mastering}%
\begin{APACrefauthors}%
Silver, D.%
, Huang, A.%
, Maddison, C\BPBI J.%
, Guez, A.%
, Sifre, L.%
, Van Den~Driessche, G.%
\BDBL {}others%
\end{APACrefauthors}%
\unskip\
\newblock
\APACrefYearMonthDay{2016}{}{}.
\newblock
{\BBOQ}\APACrefatitle {Mastering the game of Go with deep neural networks and
  tree search} {Mastering the game of go with deep neural networks and tree
  search}.{\BBCQ}
\newblock
\APACjournalVolNumPages{nature}{529}{7587}{484}.
\PrintBackRefs{\CurrentBib}

\bibitem [\protect \citeauthoryear {%
Silver%
\ \protect \BOthers {.}}{%
Silver%
\ \protect \BOthers {.}}{%
{\protect \APACyear {2017}}%
}]{%
silver2017mastering}
\APACinsertmetastar {%
silver2017mastering}%
\begin{APACrefauthors}%
Silver, D.%
, Schrittwieser, J.%
, Simonyan, K.%
, Antonoglou, I.%
, Huang, A.%
, Guez, A.%
\BDBL {}others%
\end{APACrefauthors}%
\unskip\
\newblock
\APACrefYearMonthDay{2017}{}{}.
\newblock
{\BBOQ}\APACrefatitle {Mastering the game of go without human knowledge}
  {Mastering the game of go without human knowledge}.{\BBCQ}
\newblock
\APACjournalVolNumPages{Nature}{550}{7676}{354--359}.
\PrintBackRefs{\CurrentBib}

\bibitem [\protect \citeauthoryear {%
Simoncelli%
}{%
Simoncelli%
}{%
{\protect \APACyear {2009}}%
}]{%
simoncelli2009optimal}
\APACinsertmetastar {%
simoncelli2009optimal}%
\begin{APACrefauthors}%
Simoncelli, E\BPBI P.%
\end{APACrefauthors}%
\unskip\
\newblock
\APACrefYearMonthDay{2009}{}{}.
\newblock
{\BBOQ}\APACrefatitle {Optimal estimation in sensory systems} {Optimal
  estimation in sensory systems}.{\BBCQ}
\newblock
\APACjournalVolNumPages{The Cognitive Neurosciences, IV}{}{}{525--535}.
\PrintBackRefs{\CurrentBib}

\bibitem [\protect \citeauthoryear {%
Singh%
\ \BBA {} Sutton%
}{%
Singh%
\ \BBA {} Sutton%
}{%
{\protect \APACyear {1996}}%
}]{%
singh1996reinforcement}
\APACinsertmetastar {%
singh1996reinforcement}%
\begin{APACrefauthors}%
Singh, S\BPBI P.%
\BCBT {}\ \BBA {} Sutton, R\BPBI S.%
\end{APACrefauthors}%
\unskip\
\newblock
\APACrefYearMonthDay{1996}{}{}.
\newblock
{\BBOQ}\APACrefatitle {Reinforcement learning with replacing eligibility
  traces} {Reinforcement learning with replacing eligibility traces}.{\BBCQ}
\newblock
\APACjournalVolNumPages{Machine Learning}{22}{1}{123--158}.
\PrintBackRefs{\CurrentBib}

\bibitem [\protect \citeauthoryear {%
Spratling%
}{%
Spratling%
}{%
{\protect \APACyear {2008}}%
}]{%
spratling2008reconciling}
\APACinsertmetastar {%
spratling2008reconciling}%
\begin{APACrefauthors}%
Spratling, M\BPBI W.%
\end{APACrefauthors}%
\unskip\
\newblock
\APACrefYearMonthDay{2008}{}{}.
\newblock
{\BBOQ}\APACrefatitle {Reconciling predictive coding and biased competition
  models of cortical function} {Reconciling predictive coding and biased
  competition models of cortical function}.{\BBCQ}
\newblock
\APACjournalVolNumPages{Frontiers in Computational Neuroscience}{2}{}{4}.
\PrintBackRefs{\CurrentBib}

\bibitem [\protect \citeauthoryear {%
Spratling%
}{%
Spratling%
}{%
{\protect \APACyear {2017}}%
}]{%
spratling2017review}
\APACinsertmetastar {%
spratling2017review}%
\begin{APACrefauthors}%
Spratling, M\BPBI W.%
\end{APACrefauthors}%
\unskip\
\newblock
\APACrefYearMonthDay{2017}{}{}.
\newblock
{\BBOQ}\APACrefatitle {A review of predictive coding algorithms} {A review of
  predictive coding algorithms}.{\BBCQ}
\newblock
\APACjournalVolNumPages{Brain and Cognition}{112}{}{92--97}.
\newblock
\begin{APACrefURL}
  \url{https://www.sciencedirect.com/science/article/pii/S027826261530035X?casa_token=zzTchZsrFesAAAAA:5bJNguAnRfn4BOjlCtmGvjiQT0Mkk3CE1By9JsrGrDIT0qY-CUKLUwVROkHB9S_kUx6mtH-nc74}
  \end{APACrefURL}
\PrintBackRefs{\CurrentBib}

\bibitem [\protect \citeauthoryear {%
Steil%
}{%
Steil%
}{%
{\protect \APACyear {2004}}%
}]{%
steil2004backpropagation}
\APACinsertmetastar {%
steil2004backpropagation}%
\begin{APACrefauthors}%
Steil, J\BPBI J.%
\end{APACrefauthors}%
\unskip\
\newblock
\APACrefYearMonthDay{2004}{}{}.
\newblock
{\BBOQ}\APACrefatitle {Backpropagation-decorrelation: online recurrent learning
  with O (N) complexity} {Backpropagation-decorrelation: online recurrent
  learning with o (n) complexity}.{\BBCQ}
\newblock
\BIn{} \APACrefbtitle {2004 IEEE International Joint Conference on Neural
  Networks (IEEE Cat. No. 04CH37541)} {2004 ieee international joint conference
  on neural networks (ieee cat. no. 04ch37541)}\ (\BVOL~2, \BPGS\ 843--848).
\PrintBackRefs{\CurrentBib}

\bibitem [\protect \citeauthoryear {%
Stengel%
}{%
Stengel%
}{%
{\protect \APACyear {1994}}%
}]{%
stengel1994optimal}
\APACinsertmetastar {%
stengel1994optimal}%
\begin{APACrefauthors}%
Stengel, R\BPBI F.%
\end{APACrefauthors}%
\unskip\
\newblock
\APACrefYear{1994}.
\newblock
\APACrefbtitle {Optimal control and estimation} {Optimal control and
  estimation}.
\newblock
\APACaddressPublisher{}{Courier Corporation}.
\PrintBackRefs{\CurrentBib}

\bibitem [\protect \citeauthoryear {%
Still%
\ \BBA {} Precup%
}{%
Still%
\ \BBA {} Precup%
}{%
{\protect \APACyear {2012}}%
}]{%
still2012information}
\APACinsertmetastar {%
still2012information}%
\begin{APACrefauthors}%
Still, S.%
\BCBT {}\ \BBA {} Precup, D.%
\end{APACrefauthors}%
\unskip\
\newblock
\APACrefYearMonthDay{2012}{}{}.
\newblock
{\BBOQ}\APACrefatitle {An information-theoretic approach to curiosity-driven
  reinforcement learning} {An information-theoretic approach to
  curiosity-driven reinforcement learning}.{\BBCQ}
\newblock
\APACjournalVolNumPages{Theory in Biosciences}{131}{3}{139--148}.
\PrintBackRefs{\CurrentBib}

\bibitem [\protect \citeauthoryear {%
Stuart%
, Spruston%
, Sakmann%
\BCBL {}\ \BBA {} H{\"a}usser%
}{%
Stuart%
\ \protect \BOthers {.}}{%
{\protect \APACyear {1997}}%
}]{%
stuart1997action}
\APACinsertmetastar {%
stuart1997action}%
\begin{APACrefauthors}%
Stuart, G.%
, Spruston, N.%
, Sakmann, B.%
\BCBL {}\ \BBA {} H{\"a}usser, M.%
\end{APACrefauthors}%
\unskip\
\newblock
\APACrefYearMonthDay{1997}{}{}.
\newblock
{\BBOQ}\APACrefatitle {Action potential initiation and backpropagation in
  neurons of the mammalian CNS} {Action potential initiation and
  backpropagation in neurons of the mammalian cns}.{\BBCQ}
\newblock
\APACjournalVolNumPages{Trends in Neurosciences}{20}{3}{125--131}.
\PrintBackRefs{\CurrentBib}

\bibitem [\protect \citeauthoryear {%
Such%
\ \protect \BOthers {.}}{%
Such%
\ \protect \BOthers {.}}{%
{\protect \APACyear {2017}}%
}]{%
such2017deep}
\APACinsertmetastar {%
such2017deep}%
\begin{APACrefauthors}%
Such, F\BPBI P.%
, Madhavan, V.%
, Conti, E.%
, Lehman, J.%
, Stanley, K\BPBI O.%
\BCBL {}\ \BBA {} Clune, J.%
\end{APACrefauthors}%
\unskip\
\newblock
\APACrefYearMonthDay{2017}{}{}.
\newblock
{\BBOQ}\APACrefatitle {Deep neuroevolution: Genetic algorithms are a
  competitive alternative for training deep neural networks for reinforcement
  learning} {Deep neuroevolution: Genetic algorithms are a competitive
  alternative for training deep neural networks for reinforcement
  learning}.{\BBCQ}
\newblock
\APACjournalVolNumPages{arXiv preprint arXiv:1712.06567}{}{}{}.
\PrintBackRefs{\CurrentBib}

\bibitem [\protect \citeauthoryear {%
Sun%
, Gomez%
\BCBL {}\ \BBA {} Schmidhuber%
}{%
Sun%
\ \protect \BOthers {.}}{%
{\protect \APACyear {2011}}%
}]{%
sun_planning_2011}
\APACinsertmetastar {%
sun_planning_2011}%
\begin{APACrefauthors}%
Sun, Y.%
, Gomez, F.%
\BCBL {}\ \BBA {} Schmidhuber, J.%
\end{APACrefauthors}%
\unskip\
\newblock
\APACrefYearMonthDay{2011}{}{}.
\newblock
{\BBOQ}\APACrefatitle {Planning to be surprised: Optimal bayesian exploration
  in dynamic environments} {Planning to be surprised: Optimal bayesian
  exploration in dynamic environments}.{\BBCQ}
\newblock
\BIn{} \APACrefbtitle {International Conference on Artificial General
  Intelligence} {International conference on artificial general intelligence}\
  (\BPGS\ 41--51).
\PrintBackRefs{\CurrentBib}

\bibitem [\protect \citeauthoryear {%
Sussman%
\ \BBA {} Wisdom%
}{%
Sussman%
\ \BBA {} Wisdom%
}{%
{\protect \APACyear {2015}}%
}]{%
sussman2015structure}
\APACinsertmetastar {%
sussman2015structure}%
\begin{APACrefauthors}%
Sussman, G\BPBI J.%
\BCBT {}\ \BBA {} Wisdom, J.%
\end{APACrefauthors}%
\unskip\
\newblock
\APACrefYear{2015}.
\newblock
\APACrefbtitle {Structure and interpretation of classical mechanics} {Structure
  and interpretation of classical mechanics}.
\newblock
\APACaddressPublisher{}{The MIT Press}.
\PrintBackRefs{\CurrentBib}

\bibitem [\protect \citeauthoryear {%
Sutton%
}{%
Sutton%
}{%
{\protect \APACyear {1988}}%
}]{%
sutton1988learning}
\APACinsertmetastar {%
sutton1988learning}%
\begin{APACrefauthors}%
Sutton, R\BPBI S.%
\end{APACrefauthors}%
\unskip\
\newblock
\APACrefYearMonthDay{1988}{}{}.
\newblock
{\BBOQ}\APACrefatitle {Learning to predict by the methods of temporal
  differences} {Learning to predict by the methods of temporal
  differences}.{\BBCQ}
\newblock
\APACjournalVolNumPages{Machine Learning}{3}{1}{9--44}.
\PrintBackRefs{\CurrentBib}

\bibitem [\protect \citeauthoryear {%
Sutton%
}{%
Sutton%
}{%
{\protect \APACyear {1990}}%
}]{%
sutton1990integrated}
\APACinsertmetastar {%
sutton1990integrated}%
\begin{APACrefauthors}%
Sutton, R\BPBI S.%
\end{APACrefauthors}%
\unskip\
\newblock
\APACrefYearMonthDay{1990}{}{}.
\newblock
{\BBOQ}\APACrefatitle {Integrated architectures for learning, planning, and
  reacting based on approximating dynamic programming} {Integrated
  architectures for learning, planning, and reacting based on approximating
  dynamic programming}.{\BBCQ}
\newblock
\BIn{} \APACrefbtitle {Machine learning proceedings 1990} {Machine learning
  proceedings 1990}\ (\BPGS\ 216--224).
\newblock
\APACaddressPublisher{}{Elsevier}.
\PrintBackRefs{\CurrentBib}

\bibitem [\protect \citeauthoryear {%
Sutton%
}{%
Sutton%
}{%
{\protect \APACyear {1991}}%
}]{%
sutton1991dyna}
\APACinsertmetastar {%
sutton1991dyna}%
\begin{APACrefauthors}%
Sutton, R\BPBI S.%
\end{APACrefauthors}%
\unskip\
\newblock
\APACrefYearMonthDay{1991}{}{}.
\newblock
{\BBOQ}\APACrefatitle {Dyna, an integrated architecture for learning, planning,
  and reacting} {Dyna, an integrated architecture for learning, planning, and
  reacting}.{\BBCQ}
\newblock
\APACjournalVolNumPages{ACM Sigart Bulletin}{2}{4}{160--163}.
\PrintBackRefs{\CurrentBib}

\bibitem [\protect \citeauthoryear {%
Sutton%
}{%
Sutton%
}{%
{\protect \APACyear {1996}}%
}]{%
sutton1996generalization}
\APACinsertmetastar {%
sutton1996generalization}%
\begin{APACrefauthors}%
Sutton, R\BPBI S.%
\end{APACrefauthors}%
\unskip\
\newblock
\APACrefYearMonthDay{1996}{}{}.
\newblock
{\BBOQ}\APACrefatitle {Generalization in reinforcement learning: Successful
  examples using sparse coarse coding} {Generalization in reinforcement
  learning: Successful examples using sparse coarse coding}.{\BBCQ}
\newblock
\APACjournalVolNumPages{Advances in Neural Information Processing
  Systems}{}{}{1038--1044}.
\PrintBackRefs{\CurrentBib}

\bibitem [\protect \citeauthoryear {%
Sutton%
\ \BBA {} Barto%
}{%
Sutton%
\ \BBA {} Barto%
}{%
{\protect \APACyear {2018}}%
}]{%
sutton2018reinforcement}
\APACinsertmetastar {%
sutton2018reinforcement}%
\begin{APACrefauthors}%
Sutton, R\BPBI S.%
\BCBT {}\ \BBA {} Barto, A\BPBI G.%
\end{APACrefauthors}%
\unskip\
\newblock
\APACrefYear{2018}.
\newblock
\APACrefbtitle {Reinforcement learning: An introduction} {Reinforcement
  learning: An introduction}.
\newblock
\APACaddressPublisher{}{MIT press}.
\PrintBackRefs{\CurrentBib}

\bibitem [\protect \citeauthoryear {%
Sutton%
, Barto%
\BCBL {}\ \protect \BOthers {.}}{%
Sutton%
\ \protect \BOthers {.}}{%
{\protect \APACyear {1998}}%
}]{%
sutton1998introduction}
\APACinsertmetastar {%
sutton1998introduction}%
\begin{APACrefauthors}%
Sutton, R\BPBI S.%
, Barto, A\BPBI G.%
\BCBL {}\ \BOthersPeriod {.}\end{APACrefauthors}%
\unskip\
\newblock
\APACrefYear{1998}.
\newblock
\APACrefbtitle {Introduction to reinforcement learning} {Introduction to
  reinforcement learning}\ (\BVOL~135).
\newblock
\APACaddressPublisher{}{MIT press Cambridge}.
\PrintBackRefs{\CurrentBib}

\bibitem [\protect \citeauthoryear {%
Tallec%
\ \BBA {} Ollivier%
}{%
Tallec%
\ \BBA {} Ollivier%
}{%
{\protect \APACyear {2017}}%
}]{%
tallec2017unbiased}
\APACinsertmetastar {%
tallec2017unbiased}%
\begin{APACrefauthors}%
Tallec, C.%
\BCBT {}\ \BBA {} Ollivier, Y.%
\end{APACrefauthors}%
\unskip\
\newblock
\APACrefYearMonthDay{2017}{}{}.
\newblock
{\BBOQ}\APACrefatitle {Unbiased online recurrent optimization} {Unbiased online
  recurrent optimization}.{\BBCQ}
\newblock
\APACjournalVolNumPages{arXiv preprint arXiv:1702.05043}{}{}{}.
\PrintBackRefs{\CurrentBib}

\bibitem [\protect \citeauthoryear {%
Tesauro%
}{%
Tesauro%
}{%
{\protect \APACyear {1994}}%
}]{%
tesauro1994td}
\APACinsertmetastar {%
tesauro1994td}%
\begin{APACrefauthors}%
Tesauro, G.%
\end{APACrefauthors}%
\unskip\
\newblock
\APACrefYearMonthDay{1994}{}{}.
\newblock
{\BBOQ}\APACrefatitle {TD-Gammon, a self-teaching backgammon program, achieves
  master-level play} {Td-gammon, a self-teaching backgammon program, achieves
  master-level play}.{\BBCQ}
\newblock
\APACjournalVolNumPages{Neural Computation}{6}{2}{215--219}.
\PrintBackRefs{\CurrentBib}

\bibitem [\protect \citeauthoryear {%
E.~Theodorou%
\ \BBA {} Todorov%
}{%
E.~Theodorou%
\ \BBA {} Todorov%
}{%
{\protect \APACyear {2012}}%
}]{%
theodorou2012relative}
\APACinsertmetastar {%
theodorou2012relative}%
\begin{APACrefauthors}%
Theodorou, E.%
\BCBT {}\ \BBA {} Todorov, E.%
\end{APACrefauthors}%
\unskip\
\newblock
\APACrefYearMonthDay{2012}{}{}.
\newblock
{\BBOQ}\APACrefatitle {Relative entropy and free energy dualities: Connections
  to path integral and kl control} {Relative entropy and free energy dualities:
  Connections to path integral and kl control}.{\BBCQ}
\newblock
\BIn{} \APACrefbtitle {2012 IEEE 51st IEEE Conference on Decision and Control
  (CDC)} {2012 ieee 51st ieee conference on decision and control (cdc)}\
  (\BPGS\ 1466--1473).
\PrintBackRefs{\CurrentBib}

\bibitem [\protect \citeauthoryear {%
J.~Theodorou Evangelosnd~Buchli%
\ \BBA {} Schaal%
}{%
J.~Theodorou Evangelosnd~Buchli%
\ \BBA {} Schaal%
}{%
{\protect \APACyear {2010}}%
{\protect \APACexlab {{\protect \BCnt {1}}}}}]{%
theodorou2010generalized}
\APACinsertmetastar {%
theodorou2010generalized}%
\begin{APACrefauthors}%
Theodorou, J., Evangelosnd~Buchli%
\BCBT {}\ \BBA {} Schaal, S.%
\end{APACrefauthors}%
\unskip\
\newblock
\APACrefYearMonthDay{2010{\protect \BCnt {1}}}{}{}.
\newblock
{\BBOQ}\APACrefatitle {A generalized path integral control approach to
  reinforcement learning} {A generalized path integral control approach to
  reinforcement learning}.{\BBCQ}
\newblock
\APACjournalVolNumPages{journal of machine learning
  research}{11}{Nov}{3137--3181}.
\PrintBackRefs{\CurrentBib}

\bibitem [\protect \citeauthoryear {%
J.~Theodorou Evangelosnd~Buchli%
\ \BBA {} Schaal%
}{%
J.~Theodorou Evangelosnd~Buchli%
\ \BBA {} Schaal%
}{%
{\protect \APACyear {2010}}%
{\protect \APACexlab {{\protect \BCnt {2}}}}}]{%
theodorou2010reinforcement}
\APACinsertmetastar {%
theodorou2010reinforcement}%
\begin{APACrefauthors}%
Theodorou, J., Evangelosnd~Buchli%
\BCBT {}\ \BBA {} Schaal, S.%
\end{APACrefauthors}%
\unskip\
\newblock
\APACrefYearMonthDay{2010{\protect \BCnt {2}}}{}{}.
\newblock
{\BBOQ}\APACrefatitle {Reinforcement learning of motor skills in high
  dimensions: A path integral approach} {Reinforcement learning of motor skills
  in high dimensions: A path integral approach}.{\BBCQ}
\newblock
\BIn{} \APACrefbtitle {2010 IEEE International Conference on Robotics and
  Automation} {2010 ieee international conference on robotics and automation}\
  (\BPGS\ 2397--2403).
\PrintBackRefs{\CurrentBib}

\bibitem [\protect \citeauthoryear {%
Tishby%
, Pereira%
\BCBL {}\ \BBA {} Bialek%
}{%
Tishby%
\ \protect \BOthers {.}}{%
{\protect \APACyear {2000}}%
}]{%
tishby2000information}
\APACinsertmetastar {%
tishby2000information}%
\begin{APACrefauthors}%
Tishby, N.%
, Pereira, F\BPBI C.%
\BCBL {}\ \BBA {} Bialek, W.%
\end{APACrefauthors}%
\unskip\
\newblock
\APACrefYearMonthDay{2000}{}{}.
\newblock
{\BBOQ}\APACrefatitle {The information bottleneck method} {The information
  bottleneck method}.{\BBCQ}
\newblock
\APACjournalVolNumPages{arXiv preprint physics/0004057}{}{}{}.
\PrintBackRefs{\CurrentBib}

\bibitem [\protect \citeauthoryear {%
Tishby%
\ \BBA {} Polani%
}{%
Tishby%
\ \BBA {} Polani%
}{%
{\protect \APACyear {2011}}%
}]{%
tishby2011information}
\APACinsertmetastar {%
tishby2011information}%
\begin{APACrefauthors}%
Tishby, N.%
\BCBT {}\ \BBA {} Polani, D.%
\end{APACrefauthors}%
\unskip\
\newblock
\APACrefYearMonthDay{2011}{}{}.
\newblock
{\BBOQ}\APACrefatitle {Information theory of decisions and actions}
  {Information theory of decisions and actions}.{\BBCQ}
\newblock
\BIn{} \APACrefbtitle {Perception-Action Cycle} {Perception-action cycle}\
  (\BPGS\ 601--636).
\newblock
\APACaddressPublisher{}{Springer}.
\PrintBackRefs{\CurrentBib}

\bibitem [\protect \citeauthoryear {%
Todorov%
}{%
Todorov%
}{%
{\protect \APACyear {2004}}%
}]{%
todorov2004optimality}
\APACinsertmetastar {%
todorov2004optimality}%
\begin{APACrefauthors}%
Todorov, E.%
\end{APACrefauthors}%
\unskip\
\newblock
\APACrefYearMonthDay{2004}{}{}.
\newblock
{\BBOQ}\APACrefatitle {Optimality principles in sensorimotor control}
  {Optimality principles in sensorimotor control}.{\BBCQ}
\newblock
\APACjournalVolNumPages{Nature Neuroscience}{7}{9}{907}.
\PrintBackRefs{\CurrentBib}

\bibitem [\protect \citeauthoryear {%
Todorov%
}{%
Todorov%
}{%
{\protect \APACyear {2008}}%
}]{%
todorov2008general}
\APACinsertmetastar {%
todorov2008general}%
\begin{APACrefauthors}%
Todorov, E.%
\end{APACrefauthors}%
\unskip\
\newblock
\APACrefYearMonthDay{2008}{}{}.
\newblock
{\BBOQ}\APACrefatitle {General duality between optimal control and estimation}
  {General duality between optimal control and estimation}.{\BBCQ}
\newblock
\BIn{} \APACrefbtitle {2008 47th IEEE Conference on Decision and Control} {2008
  47th ieee conference on decision and control}\ (\BPGS\ 4286--4292).
\PrintBackRefs{\CurrentBib}

\bibitem [\protect \citeauthoryear {%
Toussaint%
\ \BBA {} Storkey%
}{%
Toussaint%
\ \BBA {} Storkey%
}{%
{\protect \APACyear {2006}}%
}]{%
toussaint2006probabilistic}
\APACinsertmetastar {%
toussaint2006probabilistic}%
\begin{APACrefauthors}%
Toussaint, M.%
\BCBT {}\ \BBA {} Storkey, A.%
\end{APACrefauthors}%
\unskip\
\newblock
\APACrefYearMonthDay{2006}{}{}.
\newblock
{\BBOQ}\APACrefatitle {Probabilistic inference for solving discrete and
  continuous state Markov Decision Processes} {Probabilistic inference for
  solving discrete and continuous state markov decision processes}.{\BBCQ}
\newblock
\BIn{} \APACrefbtitle {Proceedings of the 23rd international conference on
  Machine learning} {Proceedings of the 23rd international conference on
  machine learning}\ (\BPGS\ 945--952).
\PrintBackRefs{\CurrentBib}

\bibitem [\protect \citeauthoryear {%
Tran%
, Dusenberry%
, van~der Wilk%
\BCBL {}\ \BBA {} Hafner%
}{%
Tran%
\ \protect \BOthers {.}}{%
{\protect \APACyear {2018}}%
}]{%
tran2018Bayesian}
\APACinsertmetastar {%
tran2018Bayesian}%
\begin{APACrefauthors}%
Tran, D.%
, Dusenberry, M\BPBI W.%
, van~der Wilk, M.%
\BCBL {}\ \BBA {} Hafner, D.%
\end{APACrefauthors}%
\unskip\
\newblock
\APACrefYearMonthDay{2018}{}{}.
\newblock
{\BBOQ}\APACrefatitle {Bayesian layers: A module for neural network
  uncertainty} {Bayesian layers: A module for neural network
  uncertainty}.{\BBCQ}
\newblock
\APACjournalVolNumPages{arXiv preprint arXiv:1812.03973}{}{}{}.
\PrintBackRefs{\CurrentBib}

\bibitem [\protect \citeauthoryear {%
Tschantz%
, Baltieri%
, Seth%
\BCBL {}\ \BBA {} Buckley%
}{%
Tschantz%
, Baltieri%
\BCBL {}\ \protect \BOthers {.}}{%
{\protect \APACyear {2020}}%
}]{%
tschantz_scaling_2019}
\APACinsertmetastar {%
tschantz_scaling_2019}%
\begin{APACrefauthors}%
Tschantz, A.%
, Baltieri, M.%
, Seth, A\BPBI K.%
\BCBL {}\ \BBA {} Buckley, C\BPBI L.%
\end{APACrefauthors}%
\unskip\
\newblock
\APACrefYearMonthDay{2020}{}{}.
\newblock
{\BBOQ}\APACrefatitle {Scaling active inference} {Scaling active
  inference}.{\BBCQ}
\newblock
\BIn{} \APACrefbtitle {2020 International Joint Conference on Neural Networks
  (IJCNN)} {2020 international joint conference on neural networks (ijcnn)}\
  (\BPGS\ 1--8).
\PrintBackRefs{\CurrentBib}

\bibitem [\protect \citeauthoryear {%
Tschantz%
, Millidge%
, Seth%
\BCBL {}\ \BBA {} Buckley%
}{%
Tschantz%
, Millidge%
\BCBL {}\ \protect \BOthers {.}}{%
{\protect \APACyear {2020}}%
{\protect \APACexlab {{\protect \BCnt {1}}}}}]{%
tschantz2020control}
\APACinsertmetastar {%
tschantz2020control}%
\begin{APACrefauthors}%
Tschantz, A.%
, Millidge, B.%
, Seth, A\BPBI K.%
\BCBL {}\ \BBA {} Buckley, C\BPBI L.%
\end{APACrefauthors}%
\unskip\
\newblock
\APACrefYearMonthDay{2020{\protect \BCnt {1}}}{}{}.
\newblock
{\BBOQ}\APACrefatitle {Control as hybrid inference} {Control as hybrid
  inference}.{\BBCQ}
\newblock
\APACjournalVolNumPages{arXiv preprint arXiv:2007.05838}{}{}{}.
\PrintBackRefs{\CurrentBib}

\bibitem [\protect \citeauthoryear {%
Tschantz%
, Millidge%
, Seth%
\BCBL {}\ \BBA {} Buckley%
}{%
Tschantz%
, Millidge%
\BCBL {}\ \protect \BOthers {.}}{%
{\protect \APACyear {2020}}%
{\protect \APACexlab {{\protect \BCnt {2}}}}}]{%
tschantz2020reinforcement}
\APACinsertmetastar {%
tschantz2020reinforcement}%
\begin{APACrefauthors}%
Tschantz, A.%
, Millidge, B.%
, Seth, A\BPBI K.%
\BCBL {}\ \BBA {} Buckley, C\BPBI L.%
\end{APACrefauthors}%
\unskip\
\newblock
\APACrefYearMonthDay{2020{\protect \BCnt {2}}}{}{}.
\newblock
{\BBOQ}\APACrefatitle {Reinforcement Learning through Active Inference}
  {Reinforcement learning through active inference}.{\BBCQ}
\newblock
\APACjournalVolNumPages{arXiv preprint arXiv:2002.12636}{}{}{}.
\PrintBackRefs{\CurrentBib}

\bibitem [\protect \citeauthoryear {%
Tversky%
\ \BBA {} Kahneman%
}{%
Tversky%
\ \BBA {} Kahneman%
}{%
{\protect \APACyear {1974}}%
}]{%
tversky1974judgment}
\APACinsertmetastar {%
tversky1974judgment}%
\begin{APACrefauthors}%
Tversky, A.%
\BCBT {}\ \BBA {} Kahneman, D.%
\end{APACrefauthors}%
\unskip\
\newblock
\APACrefYearMonthDay{1974}{}{}.
\newblock
{\BBOQ}\APACrefatitle {Judgment under uncertainty: Heuristics and biases}
  {Judgment under uncertainty: Heuristics and biases}.{\BBCQ}
\newblock
\APACjournalVolNumPages{science}{185}{4157}{1124--1131}.
\PrintBackRefs{\CurrentBib}

\bibitem [\protect \citeauthoryear {%
Ueltzhöffer%
}{%
Ueltzhöffer%
}{%
{\protect \APACyear {2018}}%
}]{%
ueltzhoffer_deep_2018}
\APACinsertmetastar {%
ueltzhoffer_deep_2018}%
\begin{APACrefauthors}%
Ueltzhöffer, K.%
\end{APACrefauthors}%
\unskip\
\newblock
\APACrefYearMonthDay{2018}{}{}.
\newblock
{\BBOQ}\APACrefatitle {Deep Active Inference} {Deep active inference}.{\BBCQ}
\newblock
\APACjournalVolNumPages{}{112}{6}{547--573}.
\newblock
\begin{APACrefURL} [{2019-10-19}]\url{http://arxiv.org/abs/1709.02341}
  \end{APACrefURL}
\newblock
\begin{APACrefDOI} \doi{10.1007/s00422-018-0785-7} \end{APACrefDOI}
\PrintBackRefs{\CurrentBib}

\bibitem [\protect \citeauthoryear {%
Van~Merri{\"e}nboer%
, Breuleux%
, Bergeron%
\BCBL {}\ \BBA {} Lamblin%
}{%
Van~Merri{\"e}nboer%
\ \protect \BOthers {.}}{%
{\protect \APACyear {2018}}%
}]{%
van2018automatic}
\APACinsertmetastar {%
van2018automatic}%
\begin{APACrefauthors}%
Van~Merri{\"e}nboer, B.%
, Breuleux, O.%
, Bergeron, A.%
\BCBL {}\ \BBA {} Lamblin, P.%
\end{APACrefauthors}%
\unskip\
\newblock
\APACrefYearMonthDay{2018}{}{}.
\newblock
{\BBOQ}\APACrefatitle {Automatic differentiation in ML: Where we are and where
  we should be going} {Automatic differentiation in ml: Where we are and where
  we should be going}.{\BBCQ}
\newblock
\BIn{} \APACrefbtitle {Advances in neural Information Processing Systems}
  {Advances in neural information processing systems}\ (\BPGS\ 8757--8767).
\PrintBackRefs{\CurrentBib}

\bibitem [\protect \citeauthoryear {%
Vaswani%
\ \protect \BOthers {.}}{%
Vaswani%
\ \protect \BOthers {.}}{%
{\protect \APACyear {2017}}%
}]{%
vaswani2017attention}
\APACinsertmetastar {%
vaswani2017attention}%
\begin{APACrefauthors}%
Vaswani, A.%
, Shazeer, N.%
, Parmar, N.%
, Uszkoreit, J.%
, Jones, L.%
, Gomez, A\BPBI N.%
\BDBL {}Polosukhin, I.%
\end{APACrefauthors}%
\unskip\
\newblock
\APACrefYearMonthDay{2017}{}{}.
\newblock
{\BBOQ}\APACrefatitle {Attention is all you need} {Attention is all you
  need}.{\BBCQ}
\newblock
\BIn{} \APACrefbtitle {Advances in neural Information Processing Systems}
  {Advances in neural information processing systems}\ (\BPGS\ 5998--6008).
\PrintBackRefs{\CurrentBib}

\bibitem [\protect \citeauthoryear {%
Vinyals%
\ \protect \BOthers {.}}{%
Vinyals%
\ \protect \BOthers {.}}{%
{\protect \APACyear {2019}}%
}]{%
vinyals2019grandmaster}
\APACinsertmetastar {%
vinyals2019grandmaster}%
\begin{APACrefauthors}%
Vinyals, O.%
, Babuschkin, I.%
, Czarnecki, W\BPBI M.%
, Mathieu, M.%
, Dudzik, A.%
, Chung, J.%
\BDBL {}others%
\end{APACrefauthors}%
\unskip\
\newblock
\APACrefYearMonthDay{2019}{}{}.
\newblock
{\BBOQ}\APACrefatitle {Grandmaster level in StarCraft II using multi-agent
  reinforcement learning} {Grandmaster level in starcraft ii using multi-agent
  reinforcement learning}.{\BBCQ}
\newblock
\APACjournalVolNumPages{Nature}{575}{7782}{350--354}.
\PrintBackRefs{\CurrentBib}

\bibitem [\protect \citeauthoryear {%
Vulkan%
}{%
Vulkan%
}{%
{\protect \APACyear {2000}}%
}]{%
vulkan2000economist}
\APACinsertmetastar {%
vulkan2000economist}%
\begin{APACrefauthors}%
Vulkan, N.%
\end{APACrefauthors}%
\unskip\
\newblock
\APACrefYearMonthDay{2000}{}{}.
\newblock
{\BBOQ}\APACrefatitle {An economist’s perspective on probability matching}
  {An economist’s perspective on probability matching}.{\BBCQ}
\newblock
\APACjournalVolNumPages{Journal of economic surveys}{14}{1}{101--118}.
\PrintBackRefs{\CurrentBib}

\bibitem [\protect \citeauthoryear {%
Wainwright%
\ \BBA {} Jordan%
}{%
Wainwright%
\ \BBA {} Jordan%
}{%
{\protect \APACyear {2008}}%
}]{%
wainwright2008graphical}
\APACinsertmetastar {%
wainwright2008graphical}%
\begin{APACrefauthors}%
Wainwright, M\BPBI J.%
\BCBT {}\ \BBA {} Jordan, M\BPBI I.%
\end{APACrefauthors}%
\unskip\
\newblock
\APACrefYear{2008}.
\newblock
\APACrefbtitle {Graphical models, exponential families, and variational
  inference} {Graphical models, exponential families, and variational
  inference}.
\newblock
\APACaddressPublisher{}{Now Publishers Inc}.
\PrintBackRefs{\CurrentBib}

\bibitem [\protect \citeauthoryear {%
Walsh%
, McGovern%
, Clark%
\BCBL {}\ \BBA {} O'Connell%
}{%
Walsh%
\ \protect \BOthers {.}}{%
{\protect \APACyear {2020}}%
}]{%
walsh2020evaluating}
\APACinsertmetastar {%
walsh2020evaluating}%
\begin{APACrefauthors}%
Walsh, K\BPBI S.%
, McGovern, D\BPBI P.%
, Clark, A.%
\BCBL {}\ \BBA {} O'Connell, R\BPBI G.%
\end{APACrefauthors}%
\unskip\
\newblock
\APACrefYearMonthDay{2020}{}{}.
\newblock
{\BBOQ}\APACrefatitle {Evaluating the neurophysiological evidence for
  predictive processing as a model of perception} {Evaluating the
  neurophysiological evidence for predictive processing as a model of
  perception}.{\BBCQ}
\newblock
\APACjournalVolNumPages{Annals of the New York Academy of
  Sciences}{1464}{1}{242}.
\PrintBackRefs{\CurrentBib}

\bibitem [\protect \citeauthoryear {%
Wan%
\ \BBA {} Van Der~Merwe%
}{%
Wan%
\ \BBA {} Van Der~Merwe%
}{%
{\protect \APACyear {2000}}%
}]{%
wan2000unscented}
\APACinsertmetastar {%
wan2000unscented}%
\begin{APACrefauthors}%
Wan, E\BPBI A.%
\BCBT {}\ \BBA {} Van Der~Merwe, R.%
\end{APACrefauthors}%
\unskip\
\newblock
\APACrefYearMonthDay{2000}{}{}.
\newblock
{\BBOQ}\APACrefatitle {The unscented Kalman filter for nonlinear estimation}
  {The unscented kalman filter for nonlinear estimation}.{\BBCQ}
\newblock
\BIn{} \APACrefbtitle {Proceedings of the IEEE 2000 Adaptive Systems for Signal
  Processing, Communications, and Control Symposium (Cat. No. 00EX373)}
  {Proceedings of the ieee 2000 adaptive systems for signal processing,
  communications, and control symposium (cat. no. 00ex373)}\ (\BPGS\ 153--158).
\PrintBackRefs{\CurrentBib}

\bibitem [\protect \citeauthoryear {%
Wang%
\ \protect \BOthers {.}}{%
Wang%
\ \protect \BOthers {.}}{%
{\protect \APACyear {2016}}%
}]{%
wang2016dueling}
\APACinsertmetastar {%
wang2016dueling}%
\begin{APACrefauthors}%
Wang, Z.%
, Schaul, T.%
, Hessel, M.%
, Hasselt, H.%
, Lanctot, M.%
\BCBL {}\ \BBA {} Freitas, N.%
\end{APACrefauthors}%
\unskip\
\newblock
\APACrefYearMonthDay{2016}{}{}.
\newblock
{\BBOQ}\APACrefatitle {Dueling network architectures for deep reinforcement
  learning} {Dueling network architectures for deep reinforcement
  learning}.{\BBCQ}
\newblock
\BIn{} \APACrefbtitle {International Conference on machine learning}
  {International conference on machine learning}\ (\BPGS\ 1995--2003).
\PrintBackRefs{\CurrentBib}

\bibitem [\protect \citeauthoryear {%
Watanabe%
, Kitaoka%
, Sakamoto%
, Yasugi%
\BCBL {}\ \BBA {} Tanaka%
}{%
Watanabe%
\ \protect \BOthers {.}}{%
{\protect \APACyear {2018}}%
}]{%
watanabe2018illusory}
\APACinsertmetastar {%
watanabe2018illusory}%
\begin{APACrefauthors}%
Watanabe, E.%
, Kitaoka, A.%
, Sakamoto, K.%
, Yasugi, M.%
\BCBL {}\ \BBA {} Tanaka, K.%
\end{APACrefauthors}%
\unskip\
\newblock
\APACrefYearMonthDay{2018}{}{}.
\newblock
{\BBOQ}\APACrefatitle {Illusory motion reproduced by deep neural networks
  trained for prediction} {Illusory motion reproduced by deep neural networks
  trained for prediction}.{\BBCQ}
\newblock
\APACjournalVolNumPages{Frontiers in Psychology}{9}{}{345}.
\PrintBackRefs{\CurrentBib}

\bibitem [\protect \citeauthoryear {%
Watkins%
\ \BBA {} Dayan%
}{%
Watkins%
\ \BBA {} Dayan%
}{%
{\protect \APACyear {1992}}%
}]{%
watkins1992q}
\APACinsertmetastar {%
watkins1992q}%
\begin{APACrefauthors}%
Watkins, C\BPBI J.%
\BCBT {}\ \BBA {} Dayan, P.%
\end{APACrefauthors}%
\unskip\
\newblock
\APACrefYearMonthDay{1992}{}{}.
\newblock
{\BBOQ}\APACrefatitle {Q-learning} {Q-learning}.{\BBCQ}
\newblock
\APACjournalVolNumPages{Machine learning}{8}{3-4}{279--292}.
\PrintBackRefs{\CurrentBib}

\bibitem [\protect \citeauthoryear {%
Watter%
, Springenberg%
, Boedecker%
\BCBL {}\ \BBA {} Riedmiller%
}{%
Watter%
\ \protect \BOthers {.}}{%
{\protect \APACyear {2015}}%
}]{%
nagabandi_neural_2017}
\APACinsertmetastar {%
nagabandi_neural_2017}%
\begin{APACrefauthors}%
Watter, M.%
, Springenberg, J\BPBI T.%
, Boedecker, J.%
\BCBL {}\ \BBA {} Riedmiller, M.%
\end{APACrefauthors}%
\unskip\
\newblock
\APACrefYearMonthDay{2015}{}{}.
\newblock
{\BBOQ}\APACrefatitle {Embed to control: A locally linear latent dynamics model
  for control from raw images} {Embed to control: A locally linear latent
  dynamics model for control from raw images}.{\BBCQ}
\newblock
\APACjournalVolNumPages{arXiv preprint arXiv:1506.07365}{}{}{}.
\PrintBackRefs{\CurrentBib}

\bibitem [\protect \citeauthoryear {%
Weilnhammer%
, Stuke%
, Hesselmann%
, Sterzer%
\BCBL {}\ \BBA {} Schmack%
}{%
Weilnhammer%
\ \protect \BOthers {.}}{%
{\protect \APACyear {2017}}%
}]{%
weilnhammer2017predictive}
\APACinsertmetastar {%
weilnhammer2017predictive}%
\begin{APACrefauthors}%
Weilnhammer, V.%
, Stuke, H.%
, Hesselmann, G.%
, Sterzer, P.%
\BCBL {}\ \BBA {} Schmack, K.%
\end{APACrefauthors}%
\unskip\
\newblock
\APACrefYearMonthDay{2017}{}{}.
\newblock
{\BBOQ}\APACrefatitle {A predictive coding account of bistable perception-a
  model-based fMRI study} {A predictive coding account of bistable perception-a
  model-based fmri study}.{\BBCQ}
\newblock
\APACjournalVolNumPages{PLoS Computational Biology}{13}{5}{e1005536}.
\PrintBackRefs{\CurrentBib}

\bibitem [\protect \citeauthoryear {%
Welling%
\ \BBA {} Teh%
}{%
Welling%
\ \BBA {} Teh%
}{%
{\protect \APACyear {2011}}%
}]{%
welling2011Bayesian}
\APACinsertmetastar {%
welling2011Bayesian}%
\begin{APACrefauthors}%
Welling, M.%
\BCBT {}\ \BBA {} Teh, Y\BPBI W.%
\end{APACrefauthors}%
\unskip\
\newblock
\APACrefYearMonthDay{2011}{}{}.
\newblock
{\BBOQ}\APACrefatitle {Bayesian learning via stochastic gradient Langevin
  dynamics} {Bayesian learning via stochastic gradient langevin
  dynamics}.{\BBCQ}
\newblock
\BIn{} \APACrefbtitle {Proceedings of the 28th International Conference on
  Machine Learning (ICML-11)} {Proceedings of the 28th international conference
  on machine learning (icml-11)}\ (\BPGS\ 681--688).
\PrintBackRefs{\CurrentBib}

\bibitem [\protect \citeauthoryear {%
West%
\ \BBA {} Stanovich%
}{%
West%
\ \BBA {} Stanovich%
}{%
{\protect \APACyear {2003}}%
}]{%
west2003probability}
\APACinsertmetastar {%
west2003probability}%
\begin{APACrefauthors}%
West, R\BPBI F.%
\BCBT {}\ \BBA {} Stanovich, K\BPBI E.%
\end{APACrefauthors}%
\unskip\
\newblock
\APACrefYearMonthDay{2003}{}{}.
\newblock
{\BBOQ}\APACrefatitle {Is probability matching smart? Associations between
  probabilistic choices and cognitive ability} {Is probability matching smart?
  associations between probabilistic choices and cognitive ability}.{\BBCQ}
\newblock
\APACjournalVolNumPages{Memory \& Cognition}{31}{2}{243--251}.
\PrintBackRefs{\CurrentBib}

\bibitem [\protect \citeauthoryear {%
Whittington%
\ \BBA {} Bogacz%
}{%
Whittington%
\ \BBA {} Bogacz%
}{%
{\protect \APACyear {2017}}%
}]{%
whittington2017approximation}
\APACinsertmetastar {%
whittington2017approximation}%
\begin{APACrefauthors}%
Whittington, J\BPBI C.%
\BCBT {}\ \BBA {} Bogacz, R.%
\end{APACrefauthors}%
\unskip\
\newblock
\APACrefYearMonthDay{2017}{}{}.
\newblock
{\BBOQ}\APACrefatitle {An approximation of the error backpropagation algorithm
  in a predictive coding network with local hebbian synaptic plasticity} {An
  approximation of the error backpropagation algorithm in a predictive coding
  network with local hebbian synaptic plasticity}.{\BBCQ}
\newblock
\APACjournalVolNumPages{Neural Computation}{29}{5}{1229--1262}.
\PrintBackRefs{\CurrentBib}

\bibitem [\protect \citeauthoryear {%
Wiener%
}{%
Wiener%
}{%
{\protect \APACyear {2019}}%
}]{%
wiener2019cybernetics}
\APACinsertmetastar {%
wiener2019cybernetics}%
\begin{APACrefauthors}%
Wiener, N.%
\end{APACrefauthors}%
\unskip\
\newblock
\APACrefYear{2019}.
\newblock
\APACrefbtitle {Cybernetics or Control and Communication in the Animal and the
  Machine} {Cybernetics or control and communication in the animal and the
  machine}.
\newblock
\APACaddressPublisher{}{MIT press}.
\PrintBackRefs{\CurrentBib}

\bibitem [\protect \citeauthoryear {%
D.~Williams%
}{%
D.~Williams%
}{%
{\protect \APACyear {2018}}%
}]{%
williams2018predictive}
\APACinsertmetastar {%
williams2018predictive}%
\begin{APACrefauthors}%
Williams, D.%
\end{APACrefauthors}%
\unskip\
\newblock
\APACrefYearMonthDay{2018}{}{}.
\newblock
{\BBOQ}\APACrefatitle {Predictive processing and the representation wars}
  {Predictive processing and the representation wars}.{\BBCQ}
\newblock
\APACjournalVolNumPages{Minds and Machines}{28}{1}{141--172}.
\newblock
\begin{APACrefURL}
  \url{https://link.springer.com/article/10.1007/s11023-017-9441-6}
  \end{APACrefURL}
\PrintBackRefs{\CurrentBib}

\bibitem [\protect \citeauthoryear {%
D.~Williams%
}{%
D.~Williams%
}{%
{\protect \APACyear {2020}}%
}]{%
williams2020brain}
\APACinsertmetastar {%
williams2020brain}%
\begin{APACrefauthors}%
Williams, D.%
\end{APACrefauthors}%
\unskip\
\newblock
\APACrefYearMonthDay{2020}{}{}.
\newblock
{\BBOQ}\APACrefatitle {Is the Brain an Organ for Prediction Error
  Minimization?} {Is the brain an organ for prediction error
  minimization?}{\BBCQ}
\newblock

\PrintBackRefs{\CurrentBib}

\bibitem [\protect \citeauthoryear {%
G.~Williams%
, Aldrich%
\BCBL {}\ \BBA {} Theodorou%
}{%
G.~Williams%
, Aldrich%
\BCBL {}\ \BBA {} Theodorou%
}{%
{\protect \APACyear {2017}}%
}]{%
williams2017model}
\APACinsertmetastar {%
williams2017model}%
\begin{APACrefauthors}%
Williams, G.%
, Aldrich, A.%
\BCBL {}\ \BBA {} Theodorou, E.%
\end{APACrefauthors}%
\unskip\
\newblock
\APACrefYearMonthDay{2017}{}{}.
\newblock
{\BBOQ}\APACrefatitle {Model predictive path integral control: From theory to
  parallel computation} {Model predictive path integral control: From theory to
  parallel computation}.{\BBCQ}
\newblock
\APACjournalVolNumPages{Journal of Guidance, Control, and
  Dynamics}{40}{2}{344--357}.
\PrintBackRefs{\CurrentBib}

\bibitem [\protect \citeauthoryear {%
G.~Williams%
, Drews%
, Goldfain%
, Rehg%
\BCBL {}\ \BBA {} Theodorou%
}{%
G.~Williams%
\ \protect \BOthers {.}}{%
{\protect \APACyear {2016}}%
}]{%
williams2016aggressive}
\APACinsertmetastar {%
williams2016aggressive}%
\begin{APACrefauthors}%
Williams, G.%
, Drews, P.%
, Goldfain, B.%
, Rehg, J\BPBI M.%
\BCBL {}\ \BBA {} Theodorou, E.%
\end{APACrefauthors}%
\unskip\
\newblock
\APACrefYearMonthDay{2016}{}{}.
\newblock
{\BBOQ}\APACrefatitle {Aggressive driving with model predictive path integral
  control} {Aggressive driving with model predictive path integral
  control}.{\BBCQ}
\newblock
\BIn{} \APACrefbtitle {2016 IEEE International Conference on Robotics and
  Automation (ICRA)} {2016 ieee international conference on robotics and
  automation (icra)}\ (\BPGS\ 1433--1440).
\PrintBackRefs{\CurrentBib}

\bibitem [\protect \citeauthoryear {%
G.~Williams%
, Wagener%
\BCBL {}\ \protect \BOthers {.}}{%
G.~Williams%
, Wagener%
\BCBL {}\ \protect \BOthers {.}}{%
{\protect \APACyear {2017}}%
}]{%
williams2017information}
\APACinsertmetastar {%
williams2017information}%
\begin{APACrefauthors}%
Williams, G.%
, Wagener, N.%
, Goldfain, B.%
, Drews, P.%
, Rehg, J\BPBI M.%
, Boots, B.%
\BCBL {}\ \BBA {} Theodorou, E.%
\end{APACrefauthors}%
\unskip\
\newblock
\APACrefYearMonthDay{2017}{}{}.
\newblock
{\BBOQ}\APACrefatitle {Information theoretic MPC for model-based reinforcement
  learning} {Information theoretic mpc for model-based reinforcement
  learning}.{\BBCQ}
\newblock
\BIn{} \APACrefbtitle {2017 IEEE International Conference on Robotics and
  Automation (ICRA)} {2017 ieee international conference on robotics and
  automation (icra)}\ (\BPGS\ 1714--1721).
\PrintBackRefs{\CurrentBib}

\bibitem [\protect \citeauthoryear {%
R\BPBI J.~Williams%
\ \BBA {} Zipser%
}{%
R\BPBI J.~Williams%
\ \BBA {} Zipser%
}{%
{\protect \APACyear {1989}}%
{\protect \APACexlab {{\protect \BCnt {1}}}}}]{%
williams1989experimental}
\APACinsertmetastar {%
williams1989experimental}%
\begin{APACrefauthors}%
Williams, R\BPBI J.%
\BCBT {}\ \BBA {} Zipser, D.%
\end{APACrefauthors}%
\unskip\
\newblock
\APACrefYearMonthDay{1989{\protect \BCnt {1}}}{}{}.
\newblock
{\BBOQ}\APACrefatitle {Experimental analysis of the real-time recurrent
  learning algorithm} {Experimental analysis of the real-time recurrent
  learning algorithm}.{\BBCQ}
\newblock
\APACjournalVolNumPages{Connection Science}{1}{1}{87--111}.
\PrintBackRefs{\CurrentBib}

\bibitem [\protect \citeauthoryear {%
R\BPBI J.~Williams%
\ \BBA {} Zipser%
}{%
R\BPBI J.~Williams%
\ \BBA {} Zipser%
}{%
{\protect \APACyear {1989}}%
{\protect \APACexlab {{\protect \BCnt {2}}}}}]{%
williams1989learning}
\APACinsertmetastar {%
williams1989learning}%
\begin{APACrefauthors}%
Williams, R\BPBI J.%
\BCBT {}\ \BBA {} Zipser, D.%
\end{APACrefauthors}%
\unskip\
\newblock
\APACrefYearMonthDay{1989{\protect \BCnt {2}}}{}{}.
\newblock
{\BBOQ}\APACrefatitle {A learning algorithm for continually running fully
  recurrent neural networks} {A learning algorithm for continually running
  fully recurrent neural networks}.{\BBCQ}
\newblock
\APACjournalVolNumPages{Neural Computation}{1}{2}{270--280}.
\PrintBackRefs{\CurrentBib}

\bibitem [\protect \citeauthoryear {%
Wolpert%
}{%
Wolpert%
}{%
{\protect \APACyear {1997}}%
}]{%
wolpert1997computational}
\APACinsertmetastar {%
wolpert1997computational}%
\begin{APACrefauthors}%
Wolpert, D\BPBI M.%
\end{APACrefauthors}%
\unskip\
\newblock
\APACrefYearMonthDay{1997}{}{}.
\newblock
{\BBOQ}\APACrefatitle {Computational approaches to motor control}
  {Computational approaches to motor control}.{\BBCQ}
\newblock
\APACjournalVolNumPages{Trends in Cognitive Sciences}{1}{6}{209--216}.
\PrintBackRefs{\CurrentBib}

\bibitem [\protect \citeauthoryear {%
Xiao%
, Rasul%
\BCBL {}\ \BBA {} Vollgraf%
}{%
Xiao%
\ \protect \BOthers {.}}{%
{\protect \APACyear {2017}}%
{\protect \APACexlab {{\protect \BCnt {1}}}}}]{%
xiao2017fashion}
\APACinsertmetastar {%
xiao2017fashion}%
\begin{APACrefauthors}%
Xiao, H.%
, Rasul, K.%
\BCBL {}\ \BBA {} Vollgraf, R.%
\end{APACrefauthors}%
\unskip\
\newblock
\APACrefYearMonthDay{2017{\protect \BCnt {1}}}{}{}.
\newblock
{\BBOQ}\APACrefatitle {Fashion-mnist: a novel image dataset for benchmarking
  machine learning algorithms} {Fashion-mnist: a novel image dataset for
  benchmarking machine learning algorithms}.{\BBCQ}
\newblock
\APACjournalVolNumPages{arXiv preprint arXiv:1708.07747}{}{}{}.
\PrintBackRefs{\CurrentBib}

\bibitem [\protect \citeauthoryear {%
Xiao%
, Rasul%
\BCBL {}\ \BBA {} Vollgraf%
}{%
Xiao%
\ \protect \BOthers {.}}{%
{\protect \APACyear {2017}}%
{\protect \APACexlab {{\protect \BCnt {2}}}}}]{%
xiao2017online}
\APACinsertmetastar {%
xiao2017online}%
\begin{APACrefauthors}%
Xiao, H.%
, Rasul, K.%
\BCBL {}\ \BBA {} Vollgraf, R.%
\end{APACrefauthors}%
\unskip\
\newblock
\APACrefYearMonthDay{2017{\protect \BCnt {2}}}{}{}.
\newblock
{\BBOQ}\APACrefatitle {Fashion-MNIST: a Novel Image Dataset for Benchmarking
  Machine Learning Algorithms} {Fashion-mnist: a novel image dataset for
  benchmarking machine learning algorithms}.{\BBCQ}
\newblock

\PrintBackRefs{\CurrentBib}

\bibitem [\protect \citeauthoryear {%
Yedidia%
}{%
Yedidia%
}{%
{\protect \APACyear {2011}}%
}]{%
yedidia2011message}
\APACinsertmetastar {%
yedidia2011message}%
\begin{APACrefauthors}%
Yedidia, J\BPBI S.%
\end{APACrefauthors}%
\unskip\
\newblock
\APACrefYearMonthDay{2011}{}{}.
\newblock
{\BBOQ}\APACrefatitle {Message-passing algorithms for inference and
  optimization} {Message-passing algorithms for inference and
  optimization}.{\BBCQ}
\newblock
\APACjournalVolNumPages{Journal of Statistical Physics}{145}{4}{860--890}.
\PrintBackRefs{\CurrentBib}

\bibitem [\protect \citeauthoryear {%
Yedidia%
, Freeman%
\BCBL {}\ \BBA {} Weiss%
}{%
Yedidia%
\ \protect \BOthers {.}}{%
{\protect \APACyear {2001}}%
}]{%
yedidia2001generalized}
\APACinsertmetastar {%
yedidia2001generalized}%
\begin{APACrefauthors}%
Yedidia, J\BPBI S.%
, Freeman, W\BPBI T.%
\BCBL {}\ \BBA {} Weiss, Y.%
\end{APACrefauthors}%
\unskip\
\newblock
\APACrefYearMonthDay{2001}{}{}.
\newblock
{\BBOQ}\APACrefatitle {Generalized belief propagation} {Generalized belief
  propagation}.{\BBCQ}
\newblock
\BIn{} \APACrefbtitle {Advances in neural Information Processing Systems}
  {Advances in neural information processing systems}\ (\BPGS\ 689--695).
\PrintBackRefs{\CurrentBib}

\bibitem [\protect \citeauthoryear {%
Yedidia%
, Freeman%
\BCBL {}\ \BBA {} Weiss%
}{%
Yedidia%
\ \protect \BOthers {.}}{%
{\protect \APACyear {2005}}%
}]{%
yedidia2005constructing}
\APACinsertmetastar {%
yedidia2005constructing}%
\begin{APACrefauthors}%
Yedidia, J\BPBI S.%
, Freeman, W\BPBI T.%
\BCBL {}\ \BBA {} Weiss, Y.%
\end{APACrefauthors}%
\unskip\
\newblock
\APACrefYearMonthDay{2005}{}{}.
\newblock
{\BBOQ}\APACrefatitle {Constructing free-energy approximations and generalized
  belief propagation algorithms} {Constructing free-energy approximations and
  generalized belief propagation algorithms}.{\BBCQ}
\newblock
\APACjournalVolNumPages{IEEE Transactions on information
  theory}{51}{7}{2282--2312}.
\PrintBackRefs{\CurrentBib}

\bibitem [\protect \citeauthoryear {%
Yuan%
\ \BBA {} Ao%
}{%
Yuan%
\ \BBA {} Ao%
}{%
{\protect \APACyear {2012}}%
}]{%
yuan2012beyond}
\APACinsertmetastar {%
yuan2012beyond}%
\begin{APACrefauthors}%
Yuan, R.%
\BCBT {}\ \BBA {} Ao, P.%
\end{APACrefauthors}%
\unskip\
\newblock
\APACrefYearMonthDay{2012}{}{}.
\newblock
{\BBOQ}\APACrefatitle {Beyond it{\^o} versus stratonovich} {Beyond it{\^o}
  versus stratonovich}.{\BBCQ}
\newblock
\APACjournalVolNumPages{Journal of Statistical Mechanics: Theory and
  Experiment}{2012}{07}{P07010}.
\PrintBackRefs{\CurrentBib}

\bibitem [\protect \citeauthoryear {%
Yuan%
, Ma%
, Yuan%
\BCBL {}\ \BBA {} Ao%
}{%
Yuan%
\ \protect \BOthers {.}}{%
{\protect \APACyear {2010}}%
}]{%
yuan2010constructive}
\APACinsertmetastar {%
yuan2010constructive}%
\begin{APACrefauthors}%
Yuan, R.%
, Ma, Y.%
, Yuan, B.%
\BCBL {}\ \BBA {} Ao, P.%
\end{APACrefauthors}%
\unskip\
\newblock
\APACrefYearMonthDay{2010}{}{}.
\newblock
{\BBOQ}\APACrefatitle {Constructive proof of global lyapunov function as
  potential function} {Constructive proof of global lyapunov function as
  potential function}.{\BBCQ}
\newblock
\APACjournalVolNumPages{arXiv preprint arXiv:1012.2721}{}{}{}.
\PrintBackRefs{\CurrentBib}

\bibitem [\protect \citeauthoryear {%
Yuan%
, Ma%
, Yuan%
\BCBL {}\ \BBA {} Ao%
}{%
Yuan%
\ \protect \BOthers {.}}{%
{\protect \APACyear {2011}}%
}]{%
yuan2011potential}
\APACinsertmetastar {%
yuan2011potential}%
\begin{APACrefauthors}%
Yuan, R.%
, Ma, Y.%
, Yuan, B.%
\BCBL {}\ \BBA {} Ao, P.%
\end{APACrefauthors}%
\unskip\
\newblock
\APACrefYearMonthDay{2011}{}{}.
\newblock
{\BBOQ}\APACrefatitle {Potential function in dynamical systems and the relation
  with Lyapunov function} {Potential function in dynamical systems and the
  relation with lyapunov function}.{\BBCQ}
\newblock
\BIn{} \APACrefbtitle {Proceedings of the 30th Chinese Control Conference}
  {Proceedings of the 30th chinese control conference}\ (\BPGS\ 6573--6580).
\PrintBackRefs{\CurrentBib}

\bibitem [\protect \citeauthoryear {%
Yuan%
, Tang%
\BCBL {}\ \BBA {} Ao%
}{%
Yuan%
\ \protect \BOthers {.}}{%
{\protect \APACyear {2017}}%
}]{%
yuan2017sde}
\APACinsertmetastar {%
yuan2017sde}%
\begin{APACrefauthors}%
Yuan, R.%
, Tang, Y.%
\BCBL {}\ \BBA {} Ao, P.%
\end{APACrefauthors}%
\unskip\
\newblock
\APACrefYearMonthDay{2017}{}{}.
\newblock
{\BBOQ}\APACrefatitle {SDE decomposition and A-type stochastic interpretation
  in nonequilibrium processes} {Sde decomposition and a-type stochastic
  interpretation in nonequilibrium processes}.{\BBCQ}
\newblock
\APACjournalVolNumPages{Frontiers of Physics}{12}{6}{1--9}.
\PrintBackRefs{\CurrentBib}

\bibitem [\protect \citeauthoryear {%
Zago%
, McIntyre%
, Senot%
\BCBL {}\ \BBA {} Lacquaniti%
}{%
Zago%
\ \protect \BOthers {.}}{%
{\protect \APACyear {2008}}%
}]{%
zago2008internal}
\APACinsertmetastar {%
zago2008internal}%
\begin{APACrefauthors}%
Zago, M.%
, McIntyre, J.%
, Senot, P.%
\BCBL {}\ \BBA {} Lacquaniti, F.%
\end{APACrefauthors}%
\unskip\
\newblock
\APACrefYearMonthDay{2008}{}{}.
\newblock
{\BBOQ}\APACrefatitle {Internal models and prediction of visual gravitational
  motion} {Internal models and prediction of visual gravitational
  motion}.{\BBCQ}
\newblock
\APACjournalVolNumPages{Vision Research}{48}{14}{1532--1538}.
\PrintBackRefs{\CurrentBib}

\bibitem [\protect \citeauthoryear {%
Zenke%
\ \BBA {} Ganguli%
}{%
Zenke%
\ \BBA {} Ganguli%
}{%
{\protect \APACyear {2018}}%
}]{%
zenke2018superspike}
\APACinsertmetastar {%
zenke2018superspike}%
\begin{APACrefauthors}%
Zenke, F.%
\BCBT {}\ \BBA {} Ganguli, S.%
\end{APACrefauthors}%
\unskip\
\newblock
\APACrefYearMonthDay{2018}{}{}.
\newblock
{\BBOQ}\APACrefatitle {Superspike: Supervised learning in multilayer spiking
  neural networks} {Superspike: Supervised learning in multilayer spiking
  neural networks}.{\BBCQ}
\newblock
\APACjournalVolNumPages{Neural Computation}{30}{6}{1514--1541}.
\PrintBackRefs{\CurrentBib}

\bibitem [\protect \citeauthoryear {%
Zwanzig%
}{%
Zwanzig%
}{%
{\protect \APACyear {2001}}%
}]{%
zwanzig2001nonequilibrium}
\APACinsertmetastar {%
zwanzig2001nonequilibrium}%
\begin{APACrefauthors}%
Zwanzig, R.%
\end{APACrefauthors}%
\unskip\
\newblock
\APACrefYear{2001}.
\newblock
\APACrefbtitle {Nonequilibrium statistical mechanics} {Nonequilibrium
  statistical mechanics}.
\newblock
\APACaddressPublisher{}{Oxford University Press}.
\PrintBackRefs{\CurrentBib}

\end{thebibliography}
